\documentclass[10pt,twocolumn,letterpaper]{article}

\newif\ifshowtodos        % toggle switch
\showtodostrue            % set to \showtodosfalse to hide all TODOs

\usepackage[pagenumbers]{cvpr} % To force page numbers, e.g. for an arXiv version

\usepackage[accsupp]{axessibility}  % Improves PDF readability for those with disabilities.

\usepackage{graphicx}
\usepackage{svg}
\usepackage{comment}
\usepackage{import}
\usepackage{float}
\usepackage{array}
\usepackage{adjustbox}
\usepackage{capt-of}   % for \captionof

\definecolor{cvprblue}{rgb}{0.21,0.49,0.74}
\usepackage[pagebackref,breaklinks,colorlinks,allcolors=cvprblue]{hyperref}

\def\paperID{} % *** Enter the Paper ID here
\def\confName{CVPR}
\def\confYear{2026}

\title{Inferring Compositional 4D Scenes without Ever Seeing One}

\author{
Ahmet Berke Gökmen \quad
Ajad Chhatkuli \quad
Luc Van Gool \quad
Danda Pani Paudel \\
INSAIT, Sofia University ``St. Kliment Ohridski'', Bulgaria \\
{\tt\small \{berke.gokmen, ajad.chhatkuli, luc.vangool, danda.paudel\}@insait.ai}
}

\begin{document}

\twocolumn[{%
\renewcommand\twocolumn[1][]{#1}%
\maketitle
\vspace{-1cm}
\begin{center}
    \centering
    \captionsetup{type=figure}
    \includegraphics[width=0.99\textwidth]{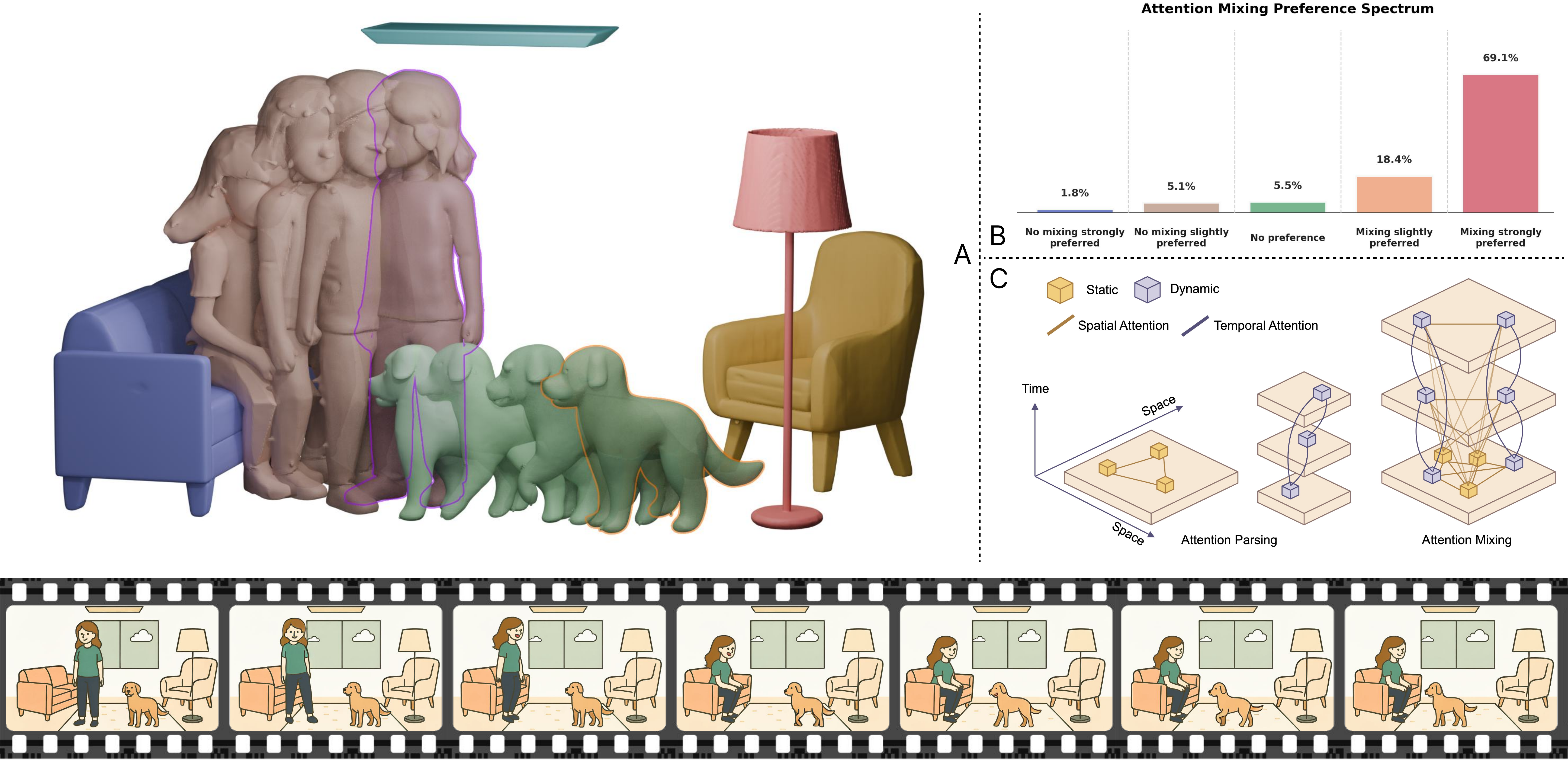}
    \captionof{figure}{ Given a single video (bottom), our method reconstructs the entire 3D scene along with the individual dynamic objects (A), while maintaining spatial and temporal consistency through spatio-temporal attention mixing (C). The silhouettes (\textcolor{magenta}{purple} for human and \textcolor{orange}{orange} for dog) correspond to the beginning of dynamic sequences. Our user study (for spatial correctness and temporal coherence) shows that the reconstructions obtained using the proposed attention mixing mechanism are clearly preferred over the baseline without mixing (B). 
    %Our approach accurately reconstructs object geometry, spatial placement, and temporal evolution from a single video input.
}
    \label{fig:teaser}
\end{center}%
}]

\begin{abstract}
Scenes in the real world are often composed of several static and dynamic objects. 
Capturing their 4-dimensional structures, composition and spatio-temporal configuration in-the-wild, though extremely interesting, is equally hard.
Therefore, existing works often focus on one object at a time, while relying on some category-specific parametric shape model for dynamic objects. This can lead to inconsistent scene configurations, in addition to being limited to the modeled object categories. We propose COM4D (Compositional 4D), a method that consistently and jointly predicts the structure and spatio-temporal configuration of 4D/3D objects using only static multi-object or dynamic single object supervision. We achieve this by a carefully designed training of spatial and temporal attentions on 2D video input. The training is disentangled into learning from object compositions on the one hand, and single object dynamics throughout the video on the other, thus completely avoiding reliance on 4D compositional training data.
At inference time, our proposed attention mixing mechanism combines these independently learned attentions, without requiring any 4D composition examples.
By alternating between spatial and temporal reasoning, COM4D reconstructs complete and persistent 4D scenes with multiple interacting objects directly from monocular videos.
%This unified formulation not only breaks the reliance on 4D supervision and category priors but also establishes a new direction for learning holistic scene dynamics purely from compositional reasoning. 
Furthermore,
COM4D provides state-of-the-art results in existing separate problems of 4D object and composed 3D reconstruction despite being purely data-driven. Code is available at \href{https://github.com/insait-institute/COM4D}{https://github.com/insait-institute/COM4D}.

%The diffusion-forcing based approach used in such training then allows spatio-temporal attention mixing at the inference time. Thus, despite not seeing any 4D composed scenes at training, the method can reconstruct accurately the 4D of the observed video for any static-dynamic object composition. 
%Single object Reconstructing a scene as decomposed dynamic objects and their static objects, in a single framework is a grand challenge in AR/VR and digital twin applications. Despite the recent improvements in single object 3D inference, disentangled multi-object reconstruction remains highly challenging. A recent work approaches multi-object reconstruction using multi token diffusion via self and cross-token attention, each token representing an object shape. However, this approach fails when there is a dynamic object changing shape, for both the static objects as well as the new dynamic object. In this work we use the diffusion forcing approach proposed for generating better video frames and adopt it for the 4D shape generation task.
\end{abstract} 
\section{Introduction}

% \begin{figure}[t]
%   \centering
%   \includegraphics[width=\columnwidth]{figures/training/general.pdf}
%   \caption{}
%   \label{fig:general}
% \end{figure}

\label{sec:intro}
Real scenes consist of multiple static and dynamic objects whose structures, compositional relationships, and spatio-temporal configurations evolve continuously over time. Capturing these factors jointly, without restrictive assumptions, requires solving reconstruction, decomposition, and temporal reasoning simultaneously. Existing approaches often sidestep this by focusing on a single object at a time or relying on category-specific parametric models for dynamic entities, which may result in inconsistent scene geometry, limited generalization, and strong inefficiencies whenever real-world objects or motions deviate from modeled priors.
Consequently, despite its importance, object-decomposed 4D scene modeling in-the-wild remains highly challenging.

Single-object 3D from an image~\cite{liu2023zero,li2025triposghighfidelity3dshape,zhao2025hunyuan3d,zhang2024gaussiancube,zhang2024clay} has advanced rapidly, thanks to the large scale object \cite{deitke2022objaverseuniverseannotated3d}, and powerful generative pipelines~\cite{ho2020denoising,dhariwal2021diffusion,lipman2023flow,liu2022flow} combining self-supervised visual priors \cite{oquab2024dinov,wang2025vggt,el2024probing}, Variational Autoencoder (VAE) shape spaces \cite{zhang20233dshape2vecset,weng2025scaling,li2025triposghighfidelity3dshape,xiang2025structured}, and diffusion transformers \cite{peebles2023scalable}. Object-agnostic point-based 3D/4D scene reconstruction has progressed through self-supervised image priors, predicting the joint structure~\cite{dust3r_cvpr24,mast3r_eccv24,keetha2025mapanything,NEURIPS2024_l4gm,cut3r} or structure with parametric 4D fitting~\cite{chen2025human3r}. These successes hinge on simplifying assumptions: objects are static, or of known parametric class, data scale is massive, and structure is globally consistent.

Similarly restrictive, most approaches to dynamic 4D reconstruction are limited to a single deforming object~\cite{chen2025v2m44dmeshanimation,zhang2025gaussianvariationfielddiffusion,ren2024l4gmlarge4dgaussian,peng2024animatable,cao2024motion2vecsets}, multiple objects~\cite{jiang2020mpshape,chen2025human3r} defined by category-specific parametric models~\cite{loper2015smpl,smplx2019}, or are object-unaware ~\cite{cut3r,Wu2024Deblur4DGS4G,Wang_2025_C4D_ICCV,zhang2024monst3r,Chen_2025_ICCV}. Typically, motion in training data is captured in controlled environments using active sensors, dedicated motion-tracking rigs, or via physical deformation models or carefully designed synthetic assets. Yet, real-world scenes violate all of these conveniences: multiple static and dynamic objects coexist, interact, occlude one another, and exhibit heterogeneous geometric and motion patterns that defy any unified prior. Consequently, as soon as objects move behind others, exhibit complex interactions, or undergo significant viewpoint changes, 4D structure can no longer be captured in many approaches, resulting in fragile representations. In other words, most 4D scene recovery often struggle with both consistency and persistence of objects. For the same reasons, large-scale, in-the-wild 4D multi-object data~\cite{panoptic} is extremely scarce, making learning severely under-constrained. As a result, progress on multi-object 4D scene reconstruction has lagged far behind that of simpler settings. To the best of our knowledge, no existing model can infer a complete and persistent 4D representation of multiple static and dynamic objects in-the-wild using only monocular videos, without test-time optimization.

% todo: What are the flaws of the current generative reconstruction? 1. Fixed length frame input. 2. ...

%Existing 4D reconstruction approaches either focus on a single dynamic object, leveraging tracking or strong category-specific priors, or they handle multiple objects but only reconstruct the geometry that remains continuously visible \cite{chen2025v2m44dmeshanimation,zhang2025gaussianvariationfielddiffusion,ren2024l4gmlarge4dgaussian}. As soon as objects move behind others, exhibit complex interactions, or undergo significant viewpoint changes, their 4D structure is no longer maintained — resulting in partial and temporally fragile representations. These methods further rely on clean surfaces, stable correspondences, or controlled capture conditions, making them poorly suited for cluttered real scenes. To the best of our knowledge, no existing method can reconstruct a complete and persistent 4D representation of multiple static and dynamic objects in-the-wild using only monocular videos.

To break through this barrier, we pursue a fundamentally different approach. We show that the required spatio-temporal reasoning can be learned in the form of attentions. We do so separately from two sources that are easy to obtain: static multi-object observations for spatial structure, and single-object animations for temporal dynamics. Thus, we introduce COM4D, a compositional 4D reconstruction method that unifies these independently learned attentions at inference time, guided by a simple but powerful physical assumption: \emph{at every time instant, all scene elements are momentarily static, and their dynamics unfold by propagating object states forward in time}. By iteratively alternating spatial and temporal reasoning — a process we call attention mixing (as illustrated in Figure~\ref{fig:teaser}.C), COM4D implicitly recovers the complete and persistent 4D structure of multiple interacting objects without ever seeing a single example of such data during training. Realizing this intuition required a series of careful design choices in representation, architecture, supervision, and inference, which collectively allow us to address a problem long considered exceptionally hard. We summarize our main contributions as follows:
\begin{itemize}
\item We introduce attention parsing, a simple yet effective strategy that disentangles the learning of spatial and temporal reasoning from separate, complementary data sources, without compromising the quality of either.
\item We propose attention mixing, which unifies these independently learned attentions at inference time on an input of a video of any length, to achieve compositional 4D scene reconstruction. Thus, our model recovers multi-object, spatio-temporal structure despite never being explicitly trained on such data.
\item We show that this unified attention framework generalizes across diverse in-the-wild scenes, achieving coherent and persistent 4D reconstructions of multiple interacting objects, significantly outperforming specialized dynamic single-object or static-scene baselines.
\end{itemize}

\section{Related Works}
%We briefly describe previous works on the task of 4D object/scene and generative reconstructions.

\paragraph{4D Object Reconstruction.}
In earlier works, termed as non-rigid structure-from-motion, 4D reconstruction approaches primarily relied on low-rank assumptions~\cite{Bregler2000RecoveringN3, NIPS2003_8db92642, paladini2009factorization, Dai2012, novotny2019c3dpo} and physics-based priors~\cite{salzmann2009, Parashar_2016_CVPR, AgudoACM14}. Due to more practical results, subsequent works favored category-specific approaches~\cite{zuffi20173d,hassan2019resolving,goel2023humans,lei2024gart,hampali2021monte,paudel2024ihuman} using parametric human shape models~\cite{loper2015smpl,smplx2019}. In order to handle any shape, methods~\cite{jiang2024consistentd,Bahmani_2024_CVPR} later opted the score-distillation~\cite{poole2023dreamfusion,wang2023prolificdreamer,huang2024_icm}. Considering both speed and generalizability to any shape, generative or diffusion-based approaches~\cite{liang2024diffusion4d,chen2025v2m44dmeshanimation, zhang2025gaussianvariationfielddiffusion, ren2024l4gmlarge4dgaussian} have become popular, as they have the potential to capture any shape persistently (including the occluded regions), while avoiding expensive test time adaptations. Noteably, \cite{ren2024l4gmlarge4dgaussian} proposed to learn temporal self-attention in a local-global manner similar to \cite{rw2019timm} instead of borrowing the pretrained priors from video diffusion models~\cite{Blattmann_2023_CVPR,emuvideo2024}.
%todo explanations of temporal attention.

\paragraph{Composed 4D Scene Reconstruction.}
The steady advancement of 3D scene reconstruction~\cite{triggs1999bundle, nister2004efficient, snavely2008bundler, schonberger2016structure, mast3r_eccv24, dust3r_cvpr24, wang2025vggt} have enabled learning-based approaches to unify static-dynamic scene 4D reconstruction~\cite{zhao2024pgdvs, HuangGS_2024_CVPR, zhang2024monst3r,cut3r,chen2025human3r, st4rtrack2025}. This has been largely possible through the novel unified appearance-3D representations~\cite{mildenhall2021nerf,kerbl20233d}. Still, most previous works target human motion~\cite{hassan2019resolving,whamcvpr2024,TRACE2023,liu2025joint,chen2025human3r,shen2024gvhmr} or do not tackle the persistent object decomposed reconstruction~\cite{som2024,Lei_2025_CVPR,cut3r,Chen_2025_ICCV,Wang_2025_C4D_ICCV,SelfYuan_2025_ICCV,stream3r2025}. Alternately, many current approaches~\cite{Li_2025_CVPR,li2025wonderplay,ZhuomanLiu_2025_CVPR,Lei_2025_CVPR,som2024} for 4D scenes use an online optimization paradigm limiting their efficiency. Specifically \cite{dreamscene4d} reconstructs static-dynamic objects in their respective coordinates and recomposes them together with their poses and depth, through test-time optimization. Test-time optimization is also favored by other 4D complete scene reconstruction approaches~\cite{som2024,Lei_2025_CVPR,Chen_2025_ICCV}.

\paragraph{Generative Scene Reconstruction.}   Despite novel representations~\cite{mildenhall2021nerf,kerbl20233d} unlocking new capabilities~\cite{VRNeRF2023,huang_telepresence2025}, many real-world applications still favor persistent one-mesh-per-object explicit geometry~\cite{zhao2025hunyuan3d,zhang2024clay}. Thanks to large scale data~\cite{deitke2022objaverseuniverseannotated3d,wu2023omniobject3d}, incredible advancements in fast single object reconstruction~\cite{liu2023zero, zhang20233dshape2vecset, li2025triposghighfidelity3dshape,zhang2024gaussiancube, xiang2025structured, zhang2024clay, zhao2025hunyuan3d} have been made. However, surprisingly few works have tried integrating generative approaches for multi-object scene reconstruction, despite their potential in persistent, complete and object-aware results. Among the few, MIDI~\cite{huang2025midimultiinstancediffusionsingle} proposes multi-instance attention in order to learn relative placement of objects from the object-wise masked conditional image embeddings. A similar approach with multi-instance attention and object-wise mask inputs is followed by \cite{meng2025scenegen}. Closest to our approach, the recent PartCrafter~\cite{lin2025partcrafterstructured3dmesh} circumvents object-wise masks, and instead denoises multiple latents by alternating in-part and inter-part attention with the part embeddings for object localization. Nevertheless, to the best of our knowledge, previous test-time optimization-free approaches (diffusion or single-step) do not solve object decomposed 4D reconstruction of scenes from video.
% can be taken to the method part.
\begin{comment}
    
\paragraph{Spatio-temporal Attention.}
Spatial and temporal attention mechanisms are an integral aspect of 4D scene reconstruction. Here, the interesting aspect is neither the of the two types of attentions but how they operate together in a single transformer. MIDI~\cite{midi} proposes multi-instance attention, which is simply the self-attention between the object latents. PartCrafter uses a more sophisticated attention mechanism that first uses local self-attention within the components of a part, followed by the global self-attention between parts. The local attention is designed to capture each object's shape while the global self-attention is designed to learn the relative positioning of objects. Temporal attention in 4D object reconstruction is commonly used to enforce temporal smoothness or learn smooth deformation. However, previous approaches do not use both of them together for the lack of training data to train both spatial and temporal attention at the same time. On the other hand, naively training them with different data does not work as we show in this work. 

\end{comment}
\section{Method}
%In the following, we describe in detail our approach to compositional scene 4D reconstruction.
\paragraph{Problem.}
Consider a 4D scene composed of $N$ static (indexed by $i$) and $M$ dynamic (indexed by $j$) objects, recorded in a fixed camera monocular video over $F$ frames (indexed by $f$). We represent the images by their DINOv2~\cite{oquab2024dinov} embeddings $\{ \smash{{}^{f}}\!\mathbf{y}^j\}$, with the static objects separated in the embedding $\mathbf{y}$. The goal of compositional 4D is to reconstruct the static object geometry latents $\mathcal{S} = \{\mathbf{z}^i\}$ and the dynamic ones $\mathcal{D}= \{\!{}^{f}\!\mathbf{z}^j\}$ conditioned on the image embeddings. Note that, we use the VAE latent quantity $z$ to also refer to an object geometry, for convenience.
\paragraph{Overview.}
In this work, we aim to learn a compositional generative model for $\{\mathcal{S}, \mathcal{D}\}$ without ever seeing such compositions.
Consider a single Diffusion Transformer (DiT), $\mathbf{v}_\theta$ with parameters $\theta$, trained on the target distributions $p(\{\mathbf{z}^i\}|\mathbf{y})$: the static geometry composition distribution conditioned on the image embedding, and $p(\{{}^{f}\!\mathbf{z}|{}^{f}\!\mathbf{y}\})$: the dynamic single object shape distribution over each video frame, conditioned on each frame embedding. Given the trainings, we want $\mathbf{v}_\theta$ to also naturally model the target joint static-dynamic distribution $p(\mathcal{S}, \mathcal{D} | \mathbf{y},\{{}^{f}\!\mathbf{y}^j\} )$. Here, $\mathcal{D}$ may consist of multiple dynamic objects, unseen during the training of $\mathbf{v}_\theta$. In the following, we describe how we tackle the problem by designing $\theta$ and its corresponding learning.

\begin{figure*}[ht]
  \centering
  \includegraphics[width=\textwidth,
                   trim=6mm 4mm 6mm 3mm, clip]{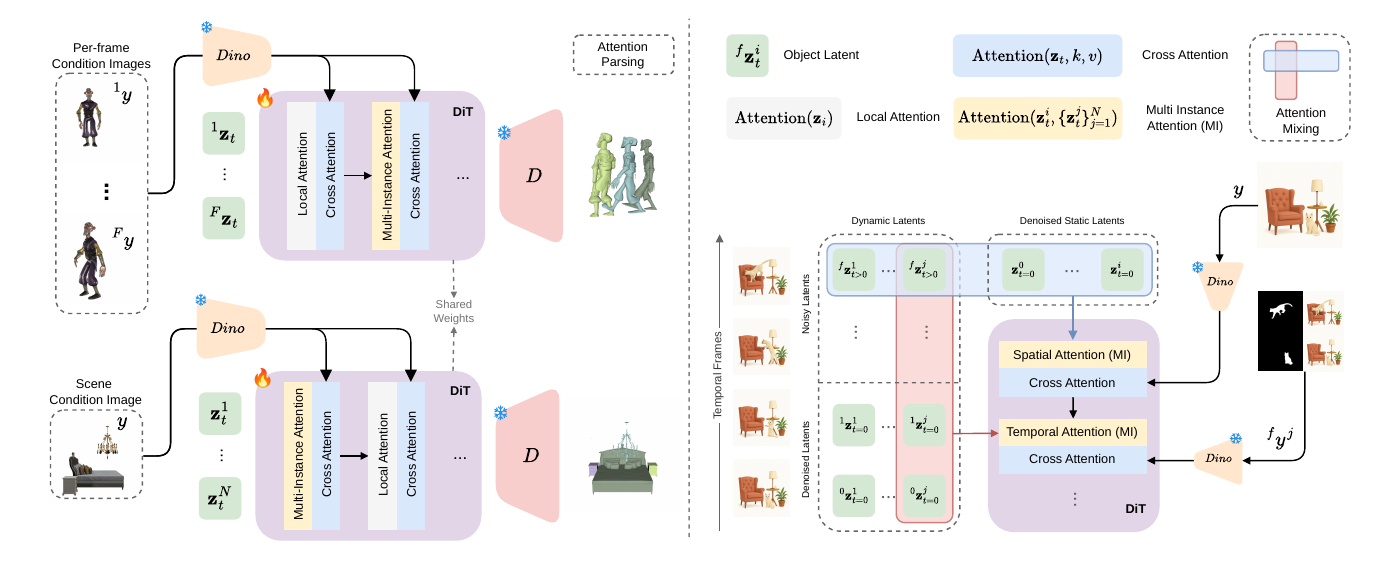}
  \caption{Our attention parsing and mixing strategy. A single DiT model with shared weights is trained jointly on two datasets. (Top) During training with samples from DeformingThings~\cite{li20214dcomplete}, odd-indexed blocks perform multi-frame attention to capture temporal dynamics. (Bottom) When training with samples from 3D-FRONT~\cite{fu20213dfront3dfurnishedrooms}, even-indexed blocks perform multi-instance attention to model spatial part decomposition. At inference, the same model applies an attention mixing mechanism. In each layer, spatial blocks (even-indexed) aggregate all latents from a single frame and process them jointly, conditioned on the full-scene image $y$ at that timestep. Temporal blocks (odd-indexed) then operate over all frames of each dynamic object separately, conditioned on their corresponding masked images. Masks are extracted from the video for each dynamic object using SAM~\cite{sam}, enabling temporally consistent object-specific reasoning.}
  \label{fig:diagram_method}
\end{figure*}

The core of our method is a single DiT architecture that first learns to perform two distinct but complementary tasks: object aware reconstruction of 3D scenes and modeling the temporal dynamics of a deformable object. We achieve this through a dual-objective training strategy called \emph{Attention Parsing}. At inference, a novel \emph{Attention Mixing} mechanism allows us to combine these learned capabilities to conditionally generate complex 4D scenes with both static and dynamic components, a task for which the model was never explicitly trained. Finally, we enhance the temporal coherence of our generations by fine-tuning with \emph{Diffusion Forcing}, cf. next section. \cref{fig:diagram_method} illustrates how \emph{Attention Parsing} and \emph{Attention Mixing} works.

\subsection{Preliminaries}

\noindent\textbf{Diffusion Transformer for 3D.} Our work builds upon TripoSG~\cite{li2025triposghighfidelity3dshape}, a state-of-the-art image-to-mesh generative model. It consists of a shape Variational Autoencoder (VAE). The encoder of the VAE converts an object mesh into a latent feature which can then be decoded into the Signed Distance Field (SDF) through its decoder similar to \cite{zhang20233dshape2vecset}. A DiT~\cite{peebles2023scalable,li2025triposghighfidelity3dshape} conditioned on the DinoV2~\cite{oquab2024dinov} image embedding is then trained to denoise a noisy latent to the object shape's VAE latent on Objaverse~\cite{deitke2022objaverseuniverseannotated3d} and ShapeNet~\cite{chang2015shapenet}.

\noindent\textbf{Diffusion Forcing.}
Different from the standard diffusion-based learning~\cite{lipman2023flow,ho2020denoising}, Diffusion Forcing~\cite{chen2024diffusionforcingnexttokenprediction} applies independent noise to the different latent vectors in the same data and denoises them together. The ability to process such mixed-noise inputs, where some latents are clean and others are noisy, is what makes this training scheme critical for our method's requirements. First, it directly addresses the need for \emph{stable static guidance} by training the model to handle conditioning on fully denoised static objects ($t=0$) while generating the remaining noisy dynamic latents. Second, it is essential for \emph{history-guided generation}, as it enables a previously denoised frame $f-1$ to serve as a clean ($t=0$) context for generating the subsequent frame $f$. This ensures strong temporal consistency and coherent evolution of the dynamic objects.

\subsection{Attention Parsing: Dual-Objective Training}
\label{sec:attention_parsing}
Our key insight is to train a single DiT model such that it understands both spatial composition and dynamics by alternately training the model on two distinct datasets, with the transformer blocks performing different roles in each.

\noindent\textbf{Dual-Dataset Strategy.} At each training step, we sample with equal probability from either the 3D-FRONT~\cite{fu20213dfront3dfurnishedrooms} dataset, which provides static scenes with object-level decompositions, or the DeformingThings~\cite{li20214dcomplete} dataset, which contains dynamic sequences of a single deforming object.

\noindent\textbf{Alternating Block Roles.} The DiT backbone consists of 21 transformer blocks. We assign complementary roles to these blocks depending on the data source. When training on a 3D-FRONT~\cite{fu20213dfront3dfurnishedrooms} sample, the even-indexed blocks are configured to perform multi-instance attention, enabling them to reason about the spatial relationships between different object parts in a scene. Conversely, when training on a DeformingThings~\cite{li20214dcomplete} sample, the odd-indexed blocks are tasked with multi-frame attention, allowing them to capture the temporal dependencies across different frames of a sequence. The blocks not assigned to multi-instance attention (even blocks in DeformingThings~\cite{li20214dcomplete}) or multi-frame attention (odd blocks in 3D-Front~\cite{fu20213dfront3dfurnishedrooms}) default to local self-attention, see \cref{fig:diagram_method}.A.

\noindent\textbf{Compositional Latent Space.} Following our problem formulation, a scene is represented as a collection of $N+M$ latents (objects and/or frames). Each latent/token is a tensor $\mathbf{z} \in \mathbb{R}^{K \times C}$. To distinguish between the different objects/frames, we add a unique, learnable embedding in each token: an object embedding $e^i$ for 3D-FRONT samples and a frame embedding ${{}^f}\!e$ for DeformingThings samples. Similarly, a single frame embedding is added to the 3D-FRONT object latents and a single object embedding is added to the DeformingThings sample. %For simplicity, we denote the latents $\mathbf{z}$ as inclusive of the embeddings $e$. %The global tokens for the scene are formed by concatenating the tokens of all components, $\mathbf{Z} = \{\mathbf{z}^1, \dots, \mathbf{z}^N\} \in \mathbb{R}^{NK \times C}$.

\noindent\textbf{Diffusion Training Objective.} Our architecture's ability to reason about multiple components is enabled by the dual attention strategy. Local self-attention is applied independently to each latent's tokens $\mathbf{z}$, while global reasoning is handled by transforming specific self-attention layers into multi-instance attention layers. The multi-instance attention allows the latent of each object $\mathbf{z}^i$ to attend to the rest $\{\mathbf{z}^l\}_{l=1}^N$ while updating itself. Similarly, the multi-frame attention each latent ${{}^f}\!\mathbf{z}$ using the rest ${{}^F_{l=1}}\!\{{{}^l}\!\mathbf{z}\}$. We write the multi-instance attention for multi-objects as below.
\begin{equation}
\mathbf{z}^{i_\text{out}} = \text{Attention}(\mathbf{z}^i, \{\mathbf{z}^l\}_{l=1}^N).
\end{equation}
The multi-frame attention follows similarly.

To train this architecture, we adapt the rectified flow objective for our dual-task. For the sake of brevity and correctness, we describe how the process works for static multiple objects in a scene. The rectified flow process for the multi-frame object latents follows similarly. 

Crucially, to enable Diffusion Forcing~\cite{chen2024diffusionforcingnexttokenprediction}, we sample an \emph{independent} time step $t_i\! \in\! [0, 1]$ for each latent among $N$. Each clean latent $\mathbf{z}_0^i$ is then perturbed along its own linear trajectory by $\boldsymbol{\epsilon}^i \sim \mathcal{N}(0, \mathbf{I})$, a random Gaussian noise tensor:
\begin{equation}
\mathbf{z}_{t_i}^i = t_i \mathbf{z}_0^i + (1 - t_i) \boldsymbol{\epsilon}^i,
\end{equation}
The flow network $\mathbf{v}_\theta$ is trained to predict the velocity vector $\boldsymbol{\epsilon}^i - \mathbf{z}_0^i$ for each component based on its unique noisy state. The overall loss is the sum over all $N$ latents:
\begin{equation}
\label{eq:static_loss}
\mathcal{L}_S = \mathbb{E} \left[ \sum_{i=1}^{N} \left\| (\boldsymbol{\epsilon}^i - \mathbf{z}_0^i) - \mathbf{v}_\theta(\mathbf{z}_{t_i}^i, t_i, \mathbf{y}) \right\|^2 \right].
\end{equation}

The conditioning tensor $\mathbf{y}$ is the background subtracted static object image embedding for eq.~\eqref{eq:static_loss}, where the samples are selected from the 3D-Front dataset~\cite{fu20213dfront3dfurnishedrooms}. During temporal training (on DeformingThings, see the equation in the supplementary), the conditioning tensor is ${{}^f}\!\mathbf{y}$, the unique image embedding corresponding to the $f$-th frame. The independent noising used in training the spatial and temporal denoising is vital, as it forces the model to handle latents at different stages of the denoising process, directly preparing it for history-guided generation at inference. The exact loss used for training is $\mathcal{L}_{S/T/R}$: static, temporal or the standard loss from TripoSG~\cite{li2025triposghighfidelity3dshape} for regularization.

\subsection{Attention Mixing: Compositional 4D Denoising}
As a consequence of Attention Parsing, our model can separate spatial and temporal reasoning. At the inference time, on a given compositional video input, the \emph{Attention Mixing} strategy enables the joint denoising of complex 4D scenes of any combination of static and dynamic entities by modulating the information flow through the transformer blocks.

% \begin{figure}[h]
%   \centering
%   \includegraphics[width=1.0\columnwidth]{figures/training/mixing.pdf}
%   \caption{The \textbf{Attention Mixing} mechanism for inference. At each layer, \textbf{spatial blocks} (even-indexed) process all latents (static and dynamic) together, using scene-level features for cross-attention to enforce global structure. \textbf{Temporal blocks} (odd-indexed) then process each dynamic object's latent sequence independently, using frame-level features for cross-attention to model its unique temporal evolution.}
%   \label{fig:attention-mixing}
% \end{figure}

During a single denoising step at inference, the DiT blocks alternate their function to cohesively generate a 4D scene. First, the even-indexed spatial blocks reason about the global scene layout. To do this, they receive the concatenated latents of all static objects along with the latent representing the \emph{current frame} of each dynamic object. This combined set forms a complete snapshot of the scene at a single moment. The block performs multi-instance attention across this entire set, using cross-attention keys and values derived from the single, global scene image to ensure all elements are placed correctly relative to one another. Subsequently, the odd-indexed temporal blocks model motion and deformation. These blocks process the latent sequence for each dynamic object separately, performing multi-frame attention over its history. The cross-attention keys and values for this operation are derived from the corresponding per-frame conditioning embeddings for that specific object, allowing the model to capture its unique temporal dynamics. Note that the strategy can handle any number of frames by sequentially propagating attention through a sliding temporal window (red block in \cref{fig:diagram_method} right) over all the frames, thus maintaining temporal consistency. Static object latents pass through these temporal blocks without being processed for motion. This flexible information routing allows the model to satisfy both the spatial constraints learned from 3D-FRONT~\cite{fu20213dfront3dfurnishedrooms} and the temporal dynamics learned from DeformingThings~\cite{li20214dcomplete} within a single denoising pass. Algo.~1 describes Attention Mixing in a simple code.

\begin{figure}[h]
\label{algo:attention_mixing}
\hrule
\vspace{1mm}
\textbf{Algorithm 1} Single Denoising Pass with Attention Mixing 
\vspace{1mm}
\hrule
\definecolor{commentcolor}{rgb}{0.1, 0.5, 0.3}
\definecolor{keywordcolor}{rgb}{0.8, 0.1, 0.5}
\begin{flushleft} 
\ttfamily
\vspace{-1mm}
\scriptsize
\textcolor{commentcolor}{\# (N,M): Number of (Static, Dynamic) Objects}\\
\textcolor{commentcolor}{\# y: Full scene condition; Z: Latent matrix ((N+M)xF)}\\
\textcolor{commentcolor}{\# Y: Masked frame-wise conditions (MxF)}\\
\vspace{1mm}
\textcolor{keywordcolor}{for} block \textcolor{keywordcolor}{in} transformer.blocks: \\
    \hspace{4mm} \textcolor{keywordcolor}{if} is\_spatial(block): \\
        \hspace{8mm} \textcolor{keywordcolor}{for} f \textcolor{keywordcolor}{in} 1..F: \\
            \hspace{12mm} Z'[:,f] = block(Z[:,f],y,num\_instances=N+M) \\
            
    \hspace{4mm} \textcolor{keywordcolor}{else}: \\
        \hspace{8mm} \textcolor{keywordcolor}{for} i \textcolor{keywordcolor}{in} 1..N: \\
            \hspace{12mm} Z'[i,:] = block(Z[i,:],y,num\_instances=1) \\
        \vspace{2mm}
        \hspace{8mm} \textcolor{keywordcolor}{for} j \textcolor{keywordcolor}{in} N..M: \\
            \hspace{12mm} Z'[j,:]=block(Z[j,:],Y[j-N,:],num\_instances=F) \\
\vspace{2mm}
    \hspace{4mm} Z = Z' \\
\vspace{2mm}
\textcolor{keywordcolor}{return} Z \\
\vspace{2mm}
\hrule
\end{flushleft}
%\hypertarget{algo:attention_mixing}{}
%\caption{}
\vspace{-8mm}
\end{figure}

%\vspace{-8mm}

\begin{figure*}[ht]
  \centering
  
  \setlength{\tabcolsep}{2pt} % Adjust horizontal gap between frames
  \renewcommand{\arraystretch}{1} % Reset default spacing

  % ==================== MASTER TABLE START ====================
  % Use m{1.5em} for the first column to vertically center the labels
  \begin{tabular}{@{}m{1em}@{\hspace{2pt}}c@{\hspace{15pt}}c@{}}

    % ==================== FIRST EXAMPLE ROW (Input) ====================
    \rotatebox{90}{Input} &
    % Left Group (DATE - INPUT)
    \begin{tabular}{@{}cccc@{}}
      \includegraphics[width=0.22\columnwidth]{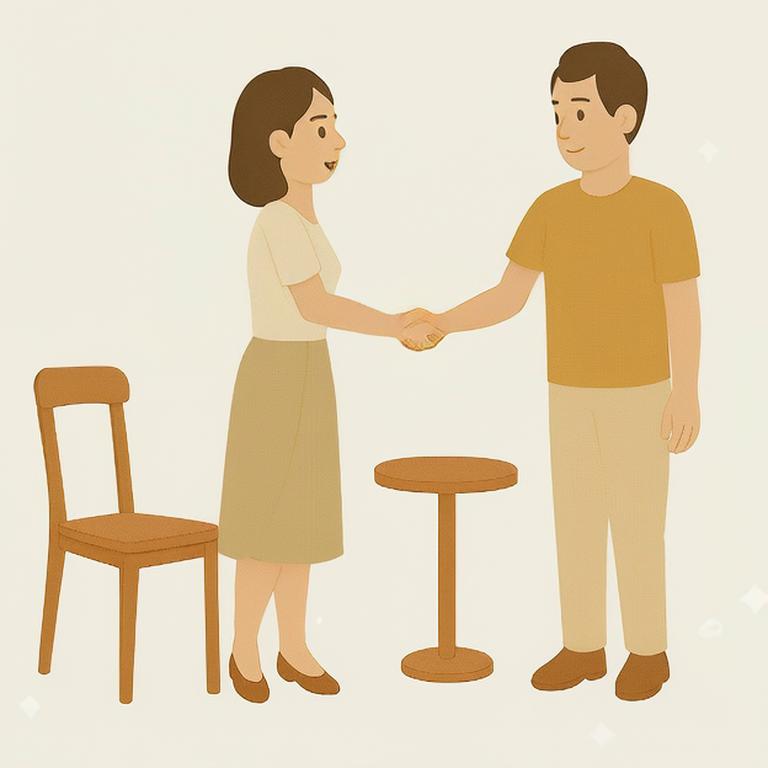} &
      \includegraphics[width=0.22\columnwidth]{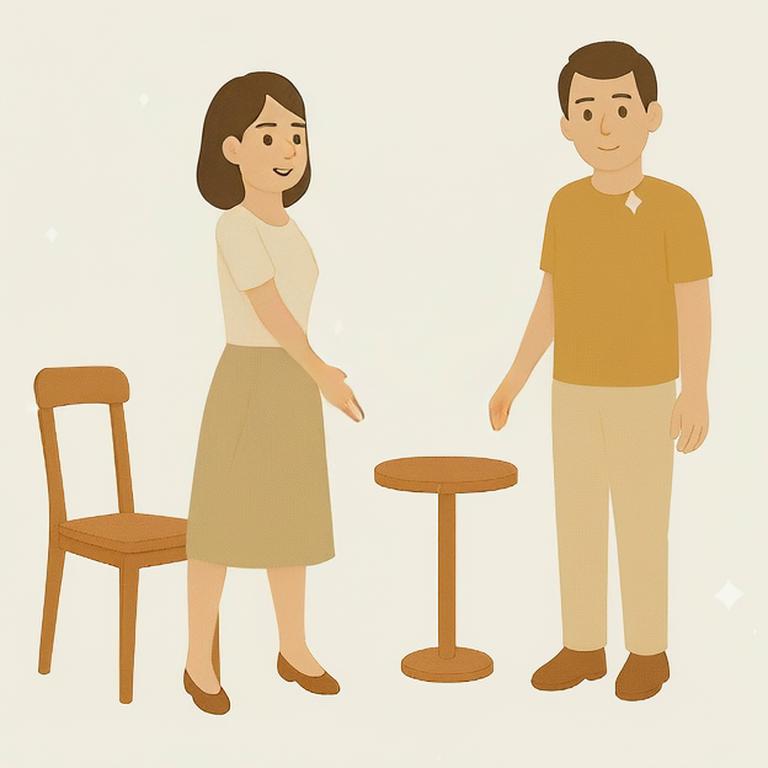} &
      \includegraphics[width=0.22\columnwidth]{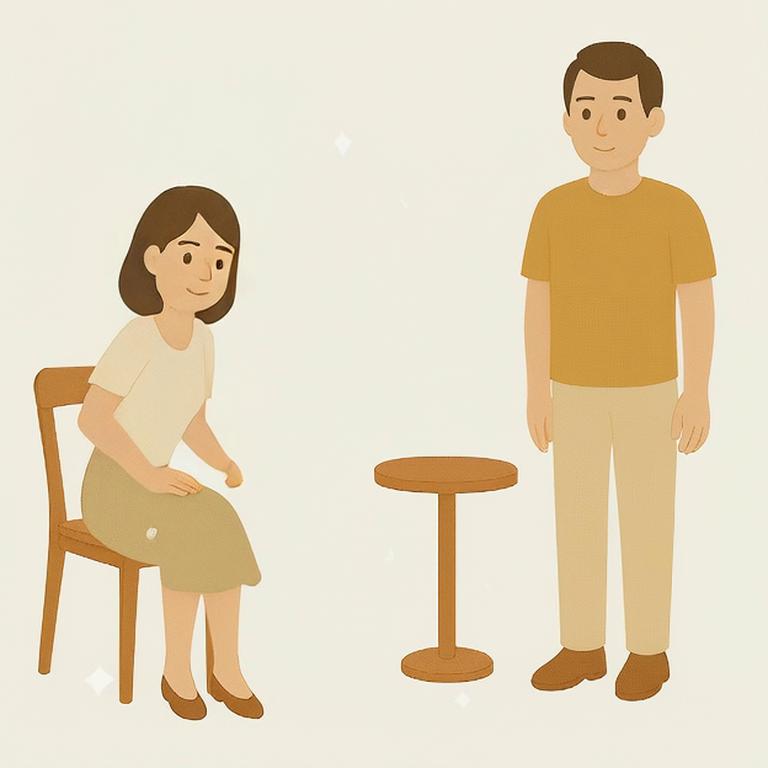} &
      \includegraphics[width=0.22\columnwidth]{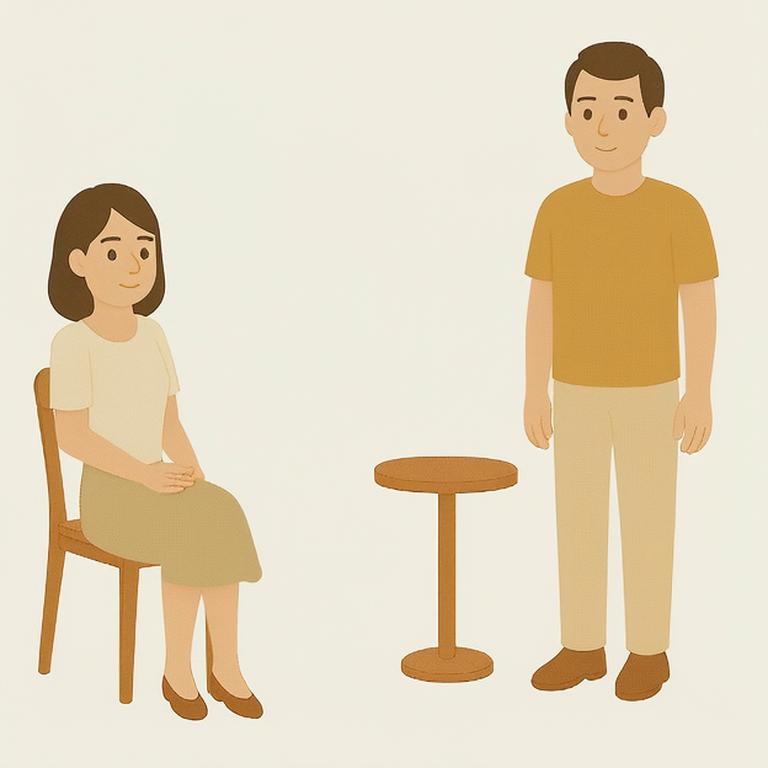}
    \end{tabular} &
    % Right Group (LIVING_ROOM - INPUT)
    \begin{tabular}{@{}cccc@{}}
      \includegraphics[width=0.22\columnwidth]{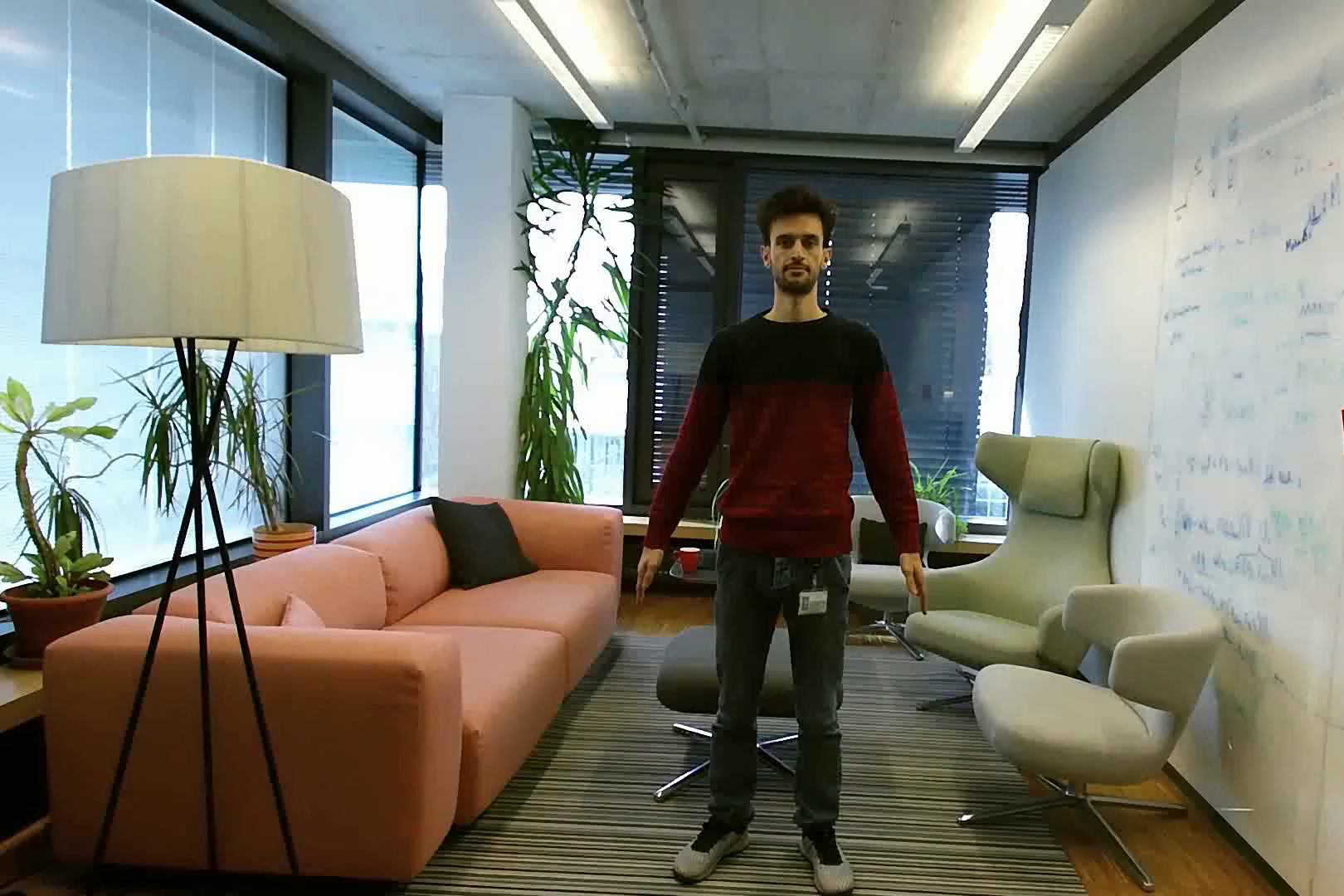} &
      \includegraphics[width=0.22\columnwidth]{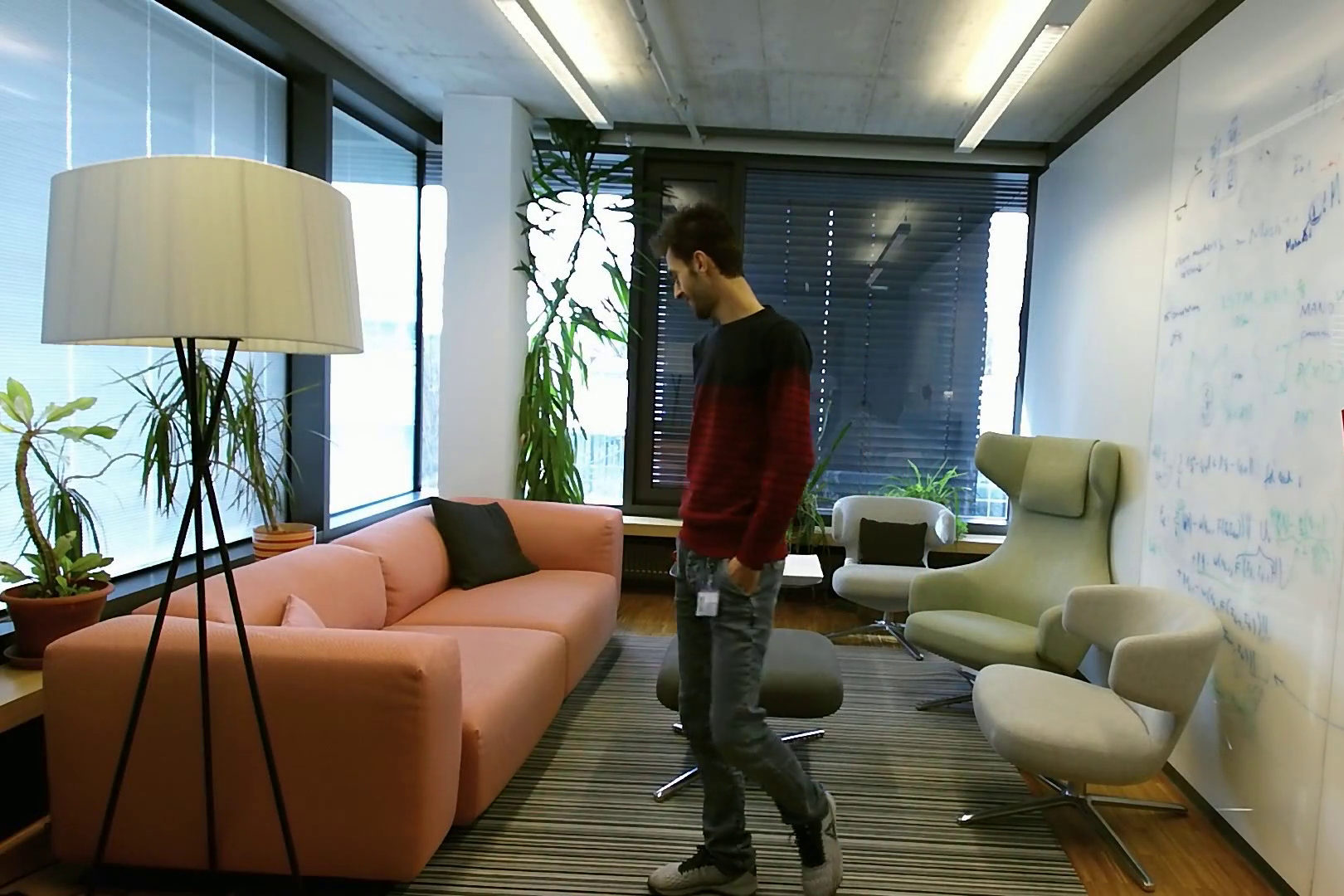} &
      \includegraphics[width=0.22\columnwidth]{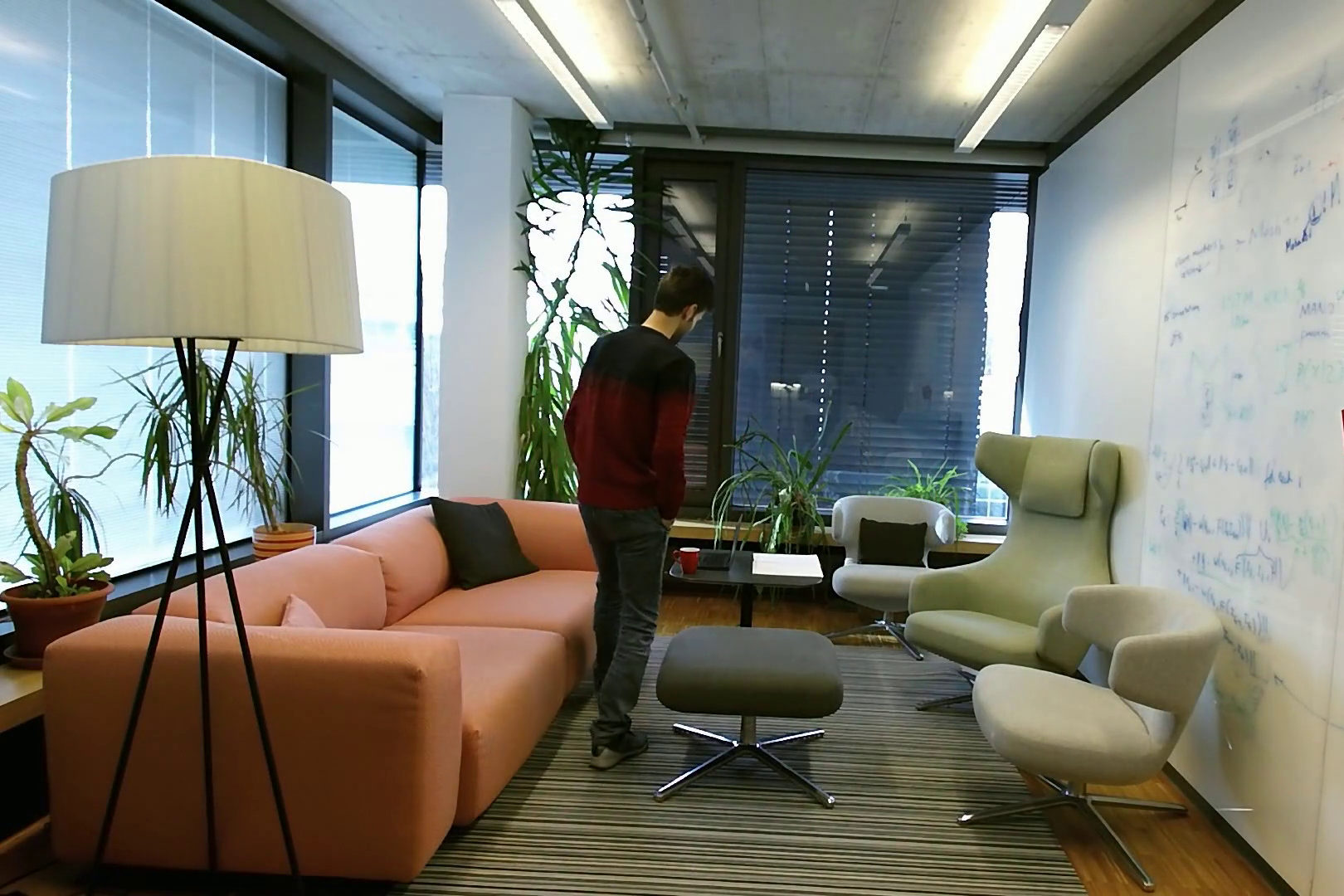} &
      \includegraphics[width=0.22\columnwidth]{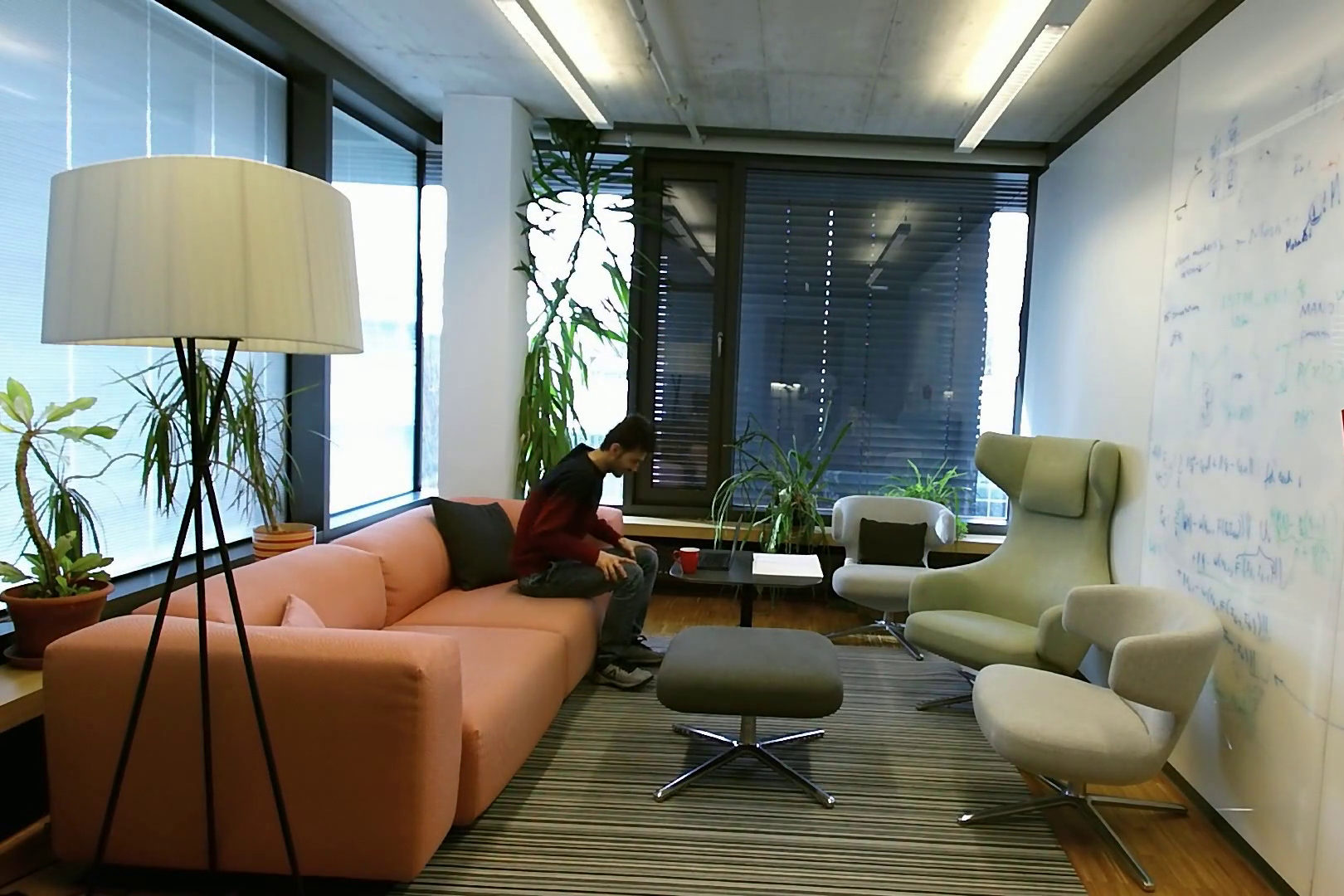}
    \end{tabular} \\
    
    % ==================== FIRST EXAMPLE ROW (Ours) ====================
    \rotatebox{90}{\textbf{Ours}} &
    % Left Group (DATE - OURS)
    \begin{tabular}{@{}cccc@{}}
      \begin{tabular}{@{}c@{}} \includegraphics[width=0.22\columnwidth]{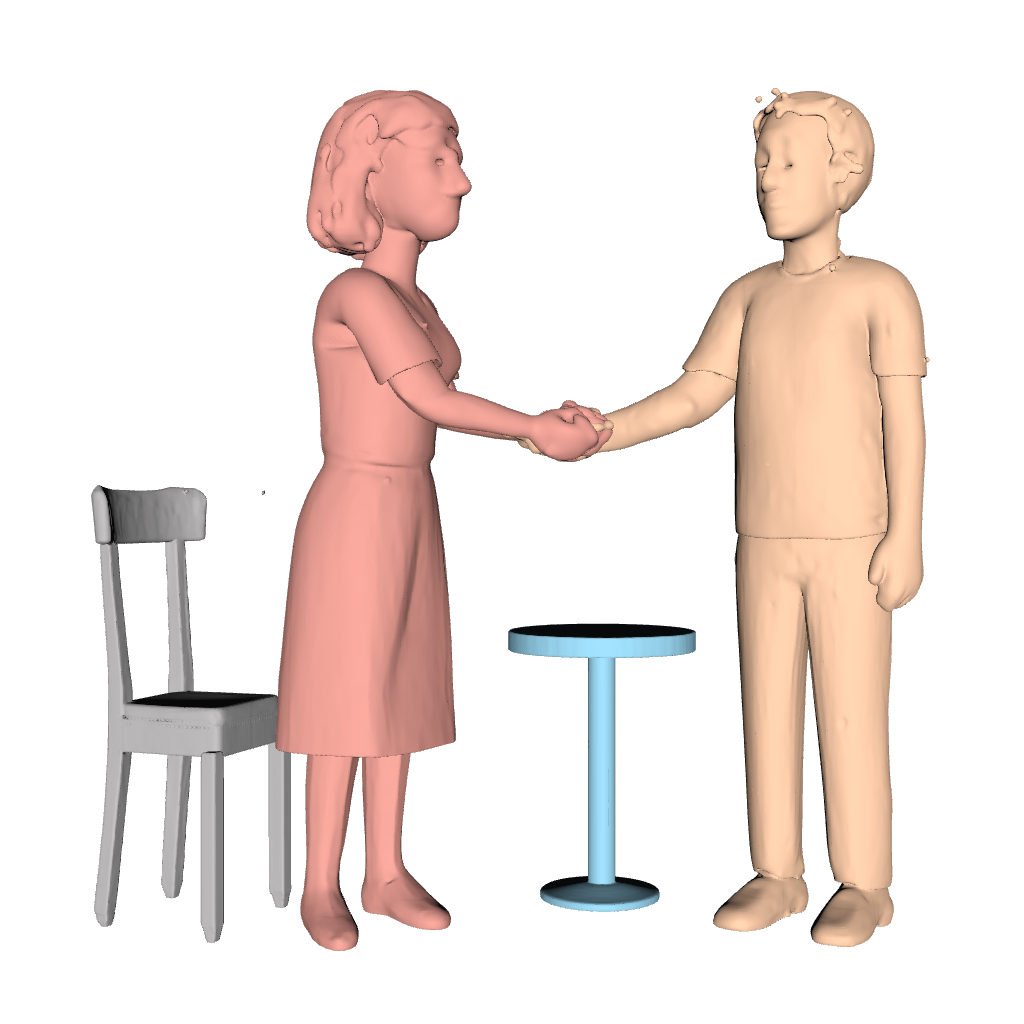} \\[-1.5pt] \includegraphics[width=0.22\columnwidth]{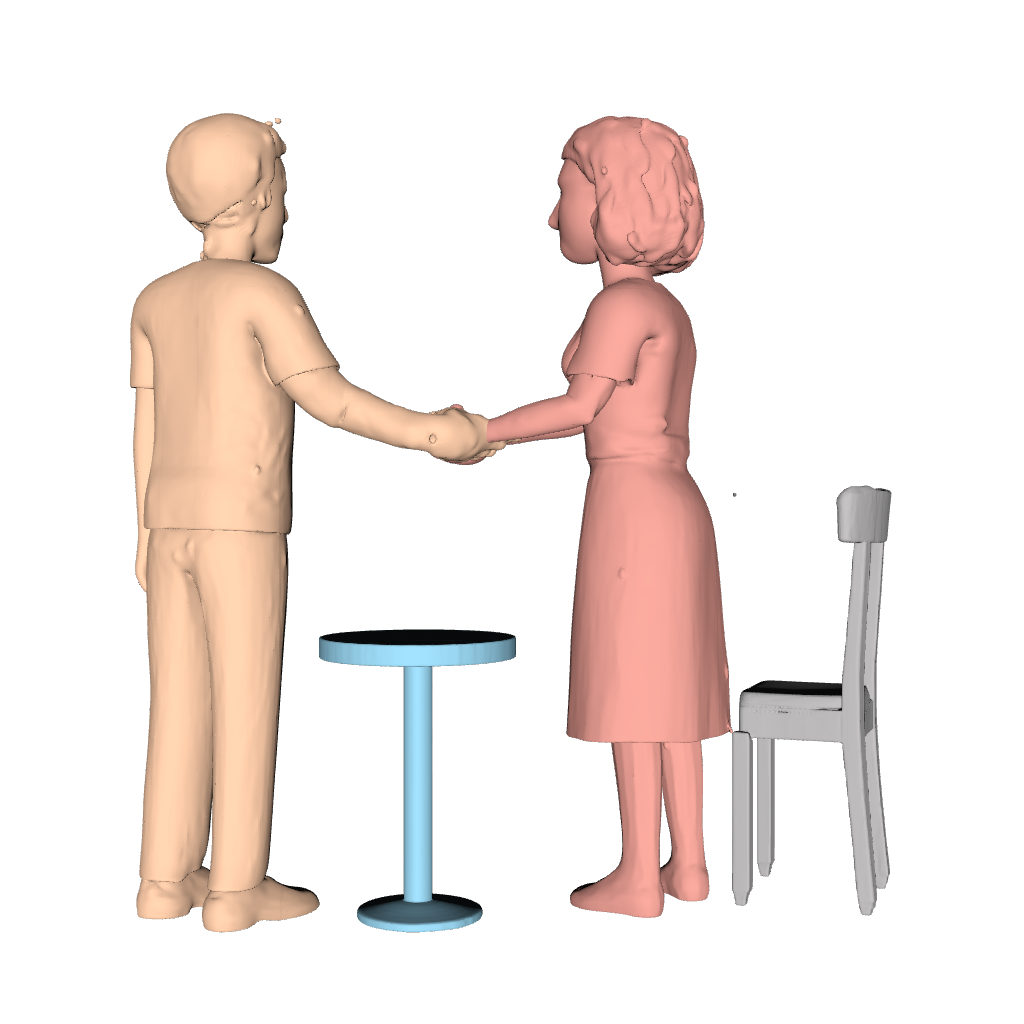} \end{tabular} &
      \begin{tabular}{@{}c@{}} \includegraphics[width=0.22\columnwidth]{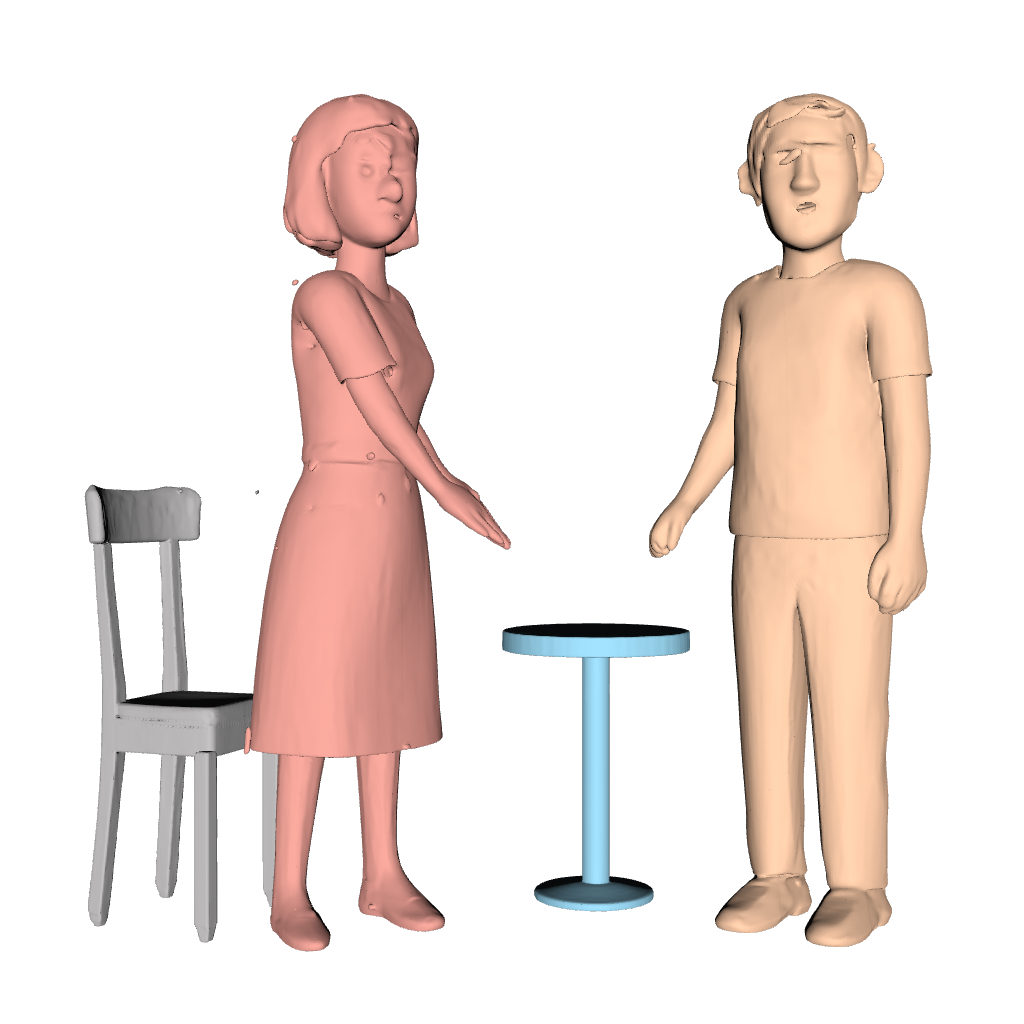} \\[-1.5pt] \includegraphics[width=0.22\columnwidth]{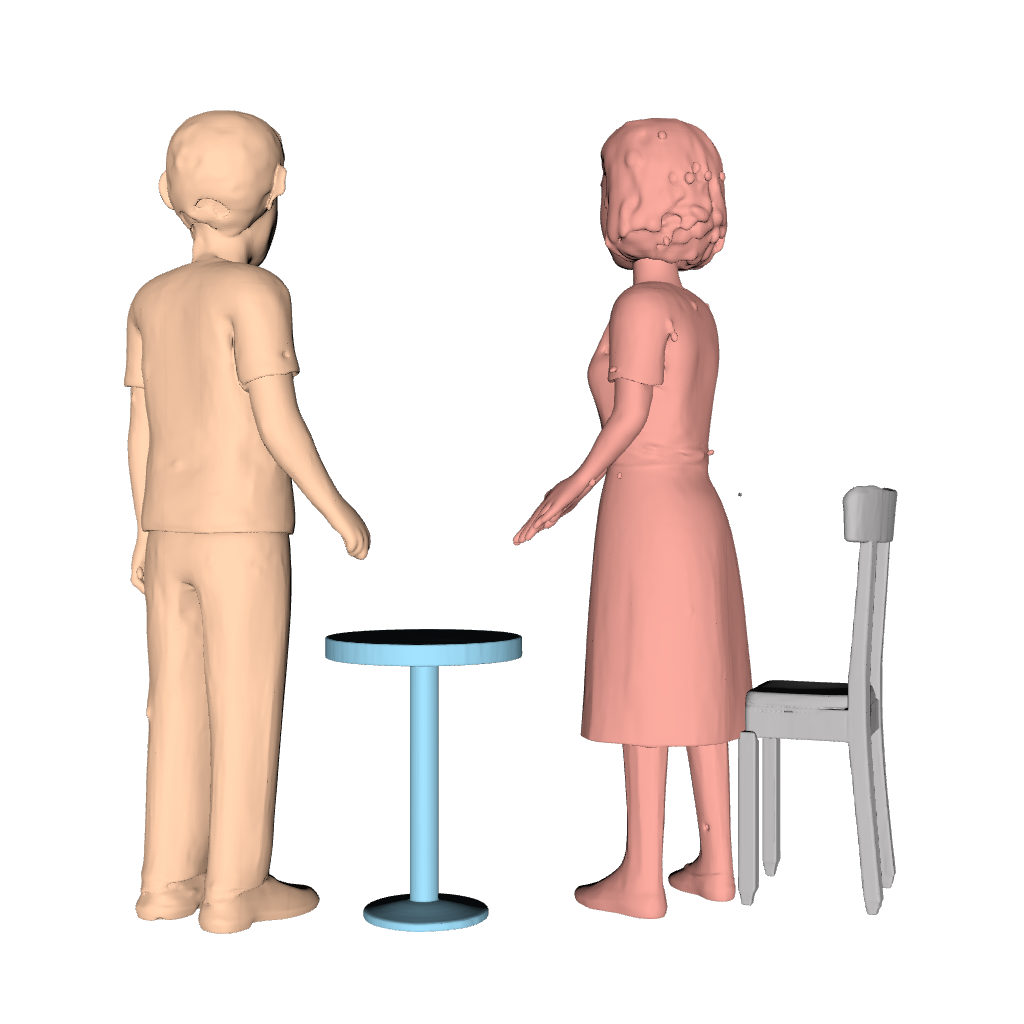} \end{tabular} &
      \begin{tabular}{@{}c@{}} \includegraphics[width=0.22\columnwidth]{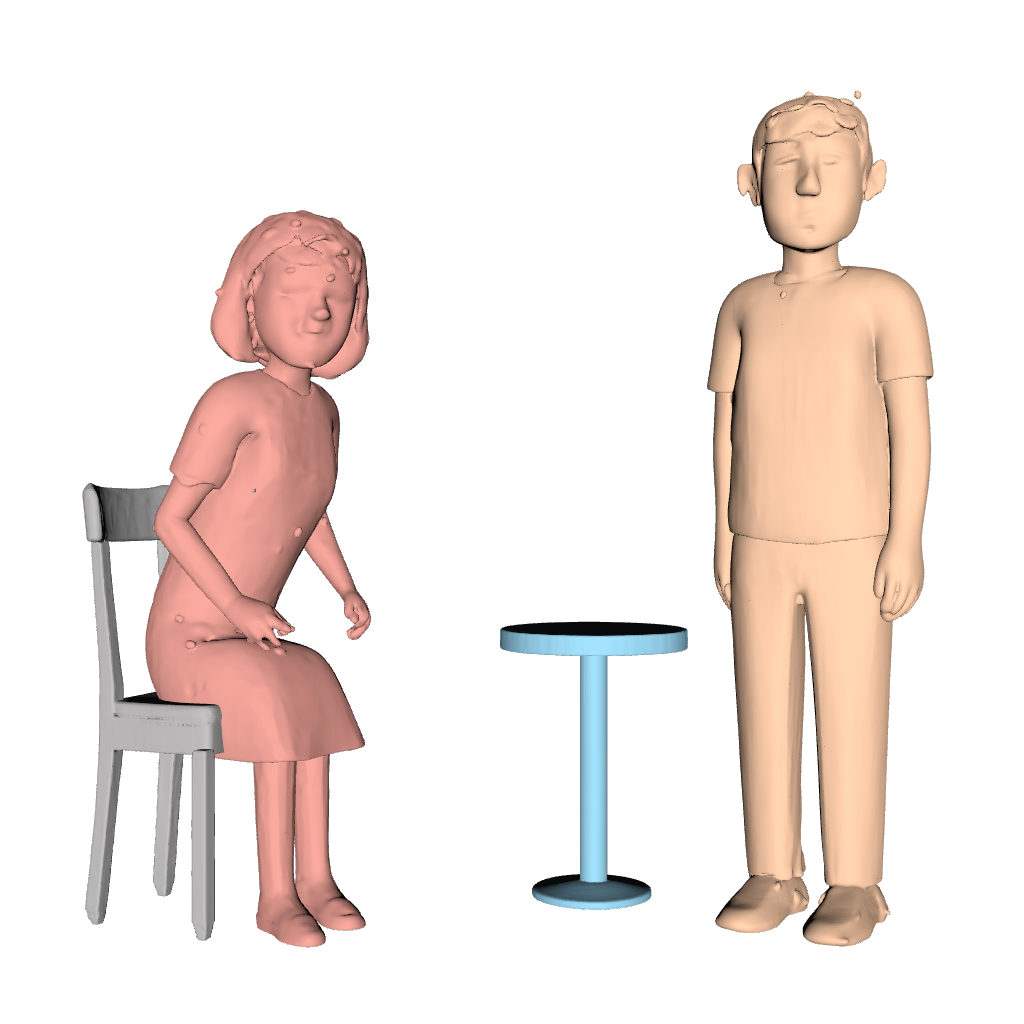} \\[-1.5pt] \includegraphics[width=0.22\columnwidth]{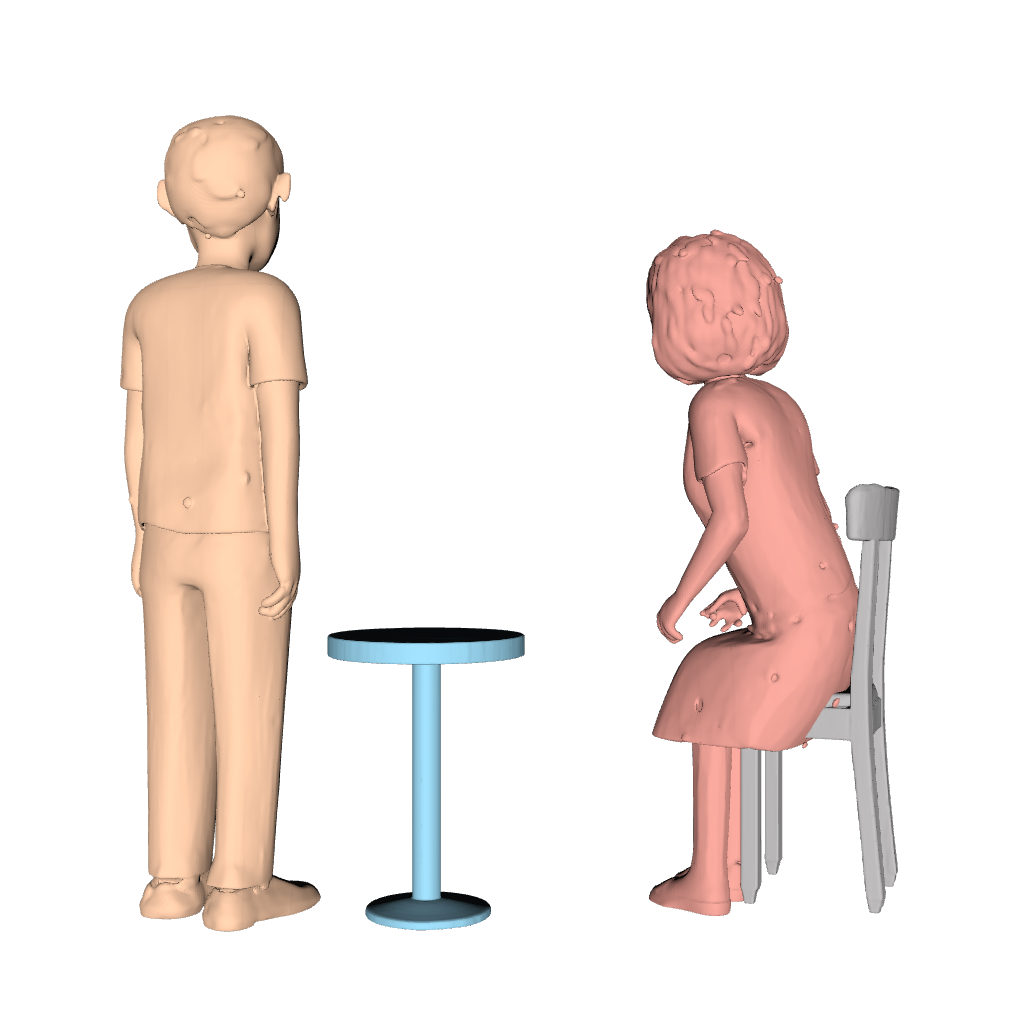} \end{tabular} &
      \begin{tabular}{@{}c@{}} \includegraphics[width=0.22\columnwidth]{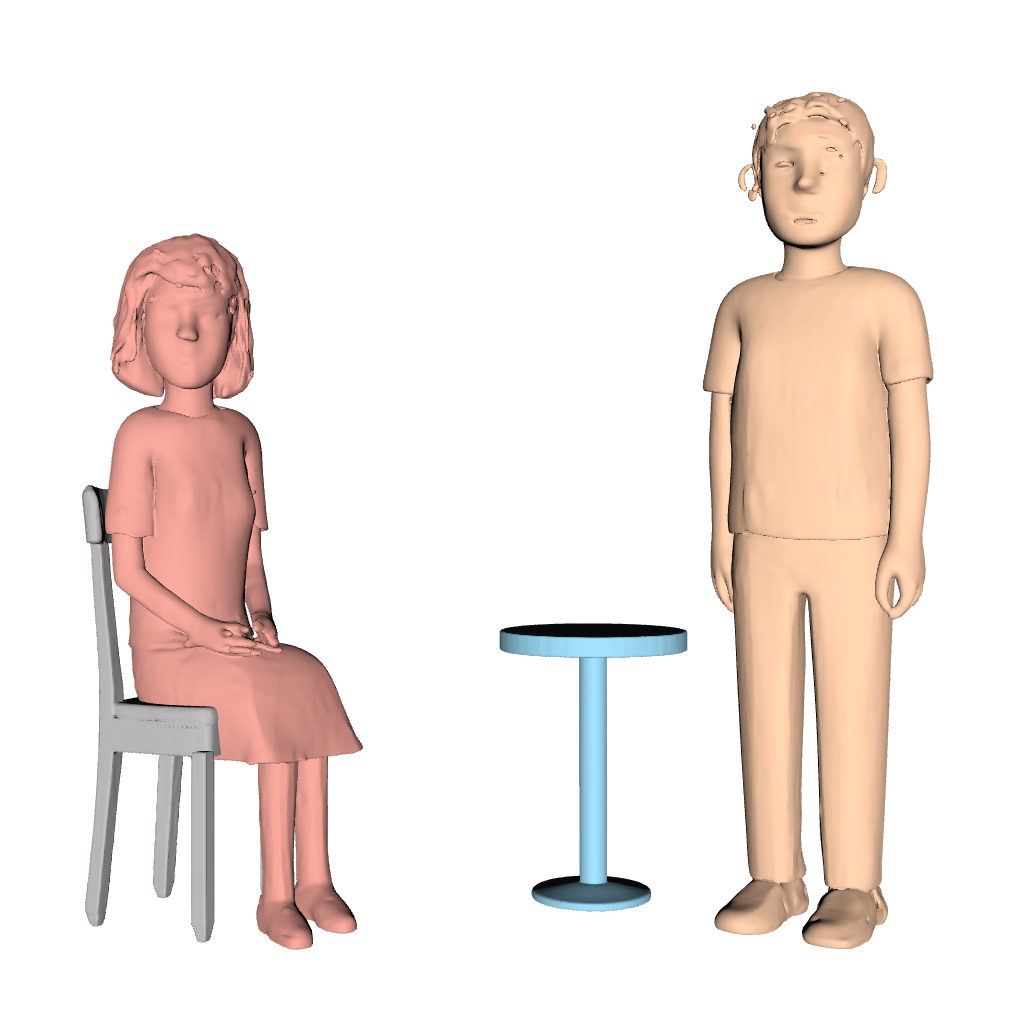} \\[-1.5pt] \includegraphics[width=0.22\columnwidth]{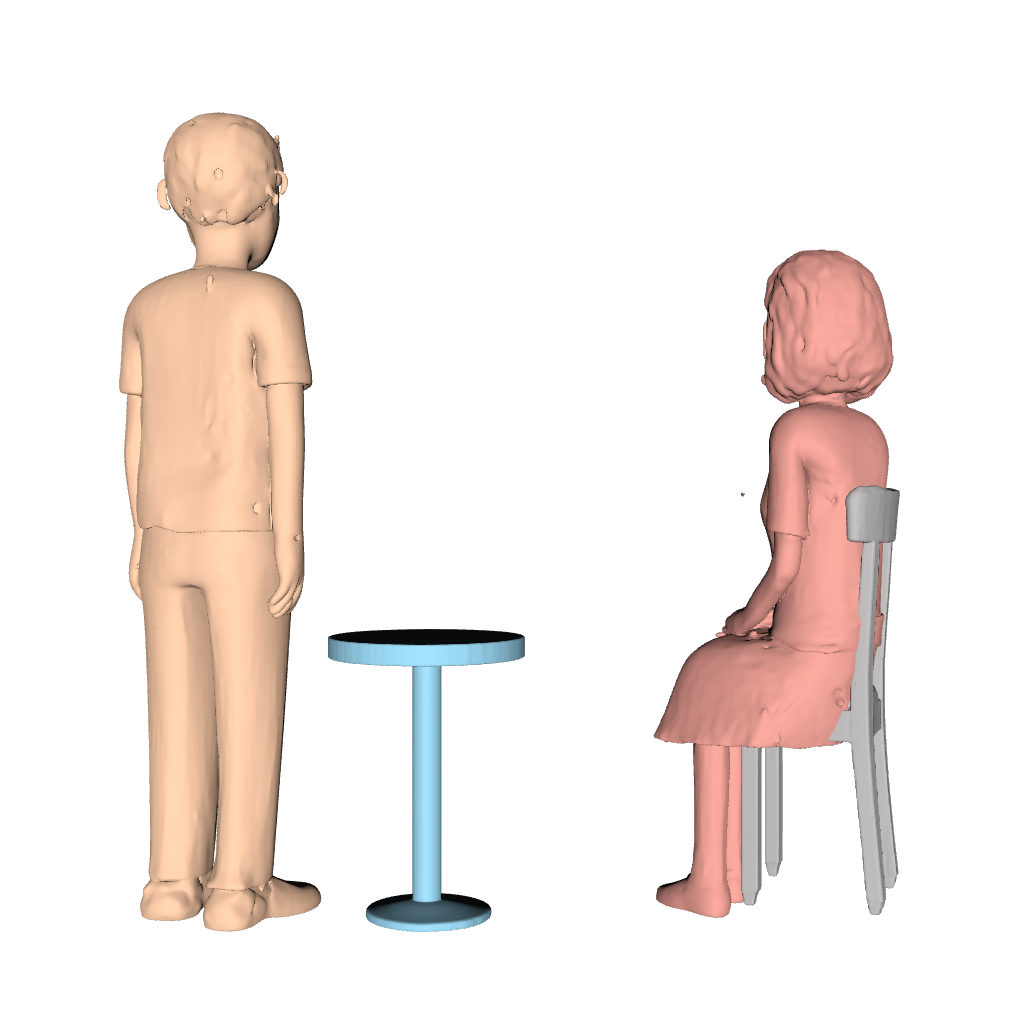} \end{tabular}
    \end{tabular} &
    % Right Group (LIVING_ROOM - OURS)
    \begin{tabular}{@{}cccc@{}}
      \begin{tabular}{@{}c@{}} \includegraphics[width=0.22\columnwidth]{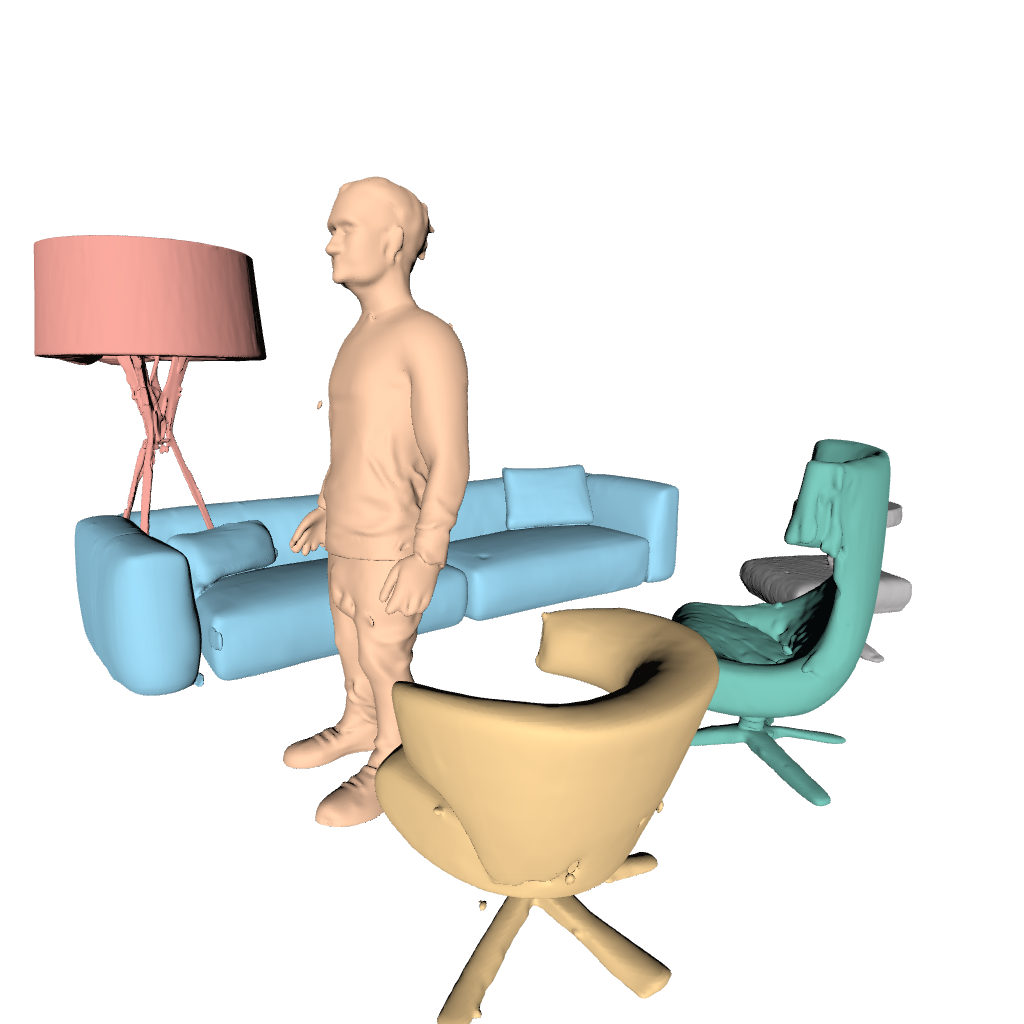} \\[-1.5pt] \includegraphics[width=0.22\columnwidth]{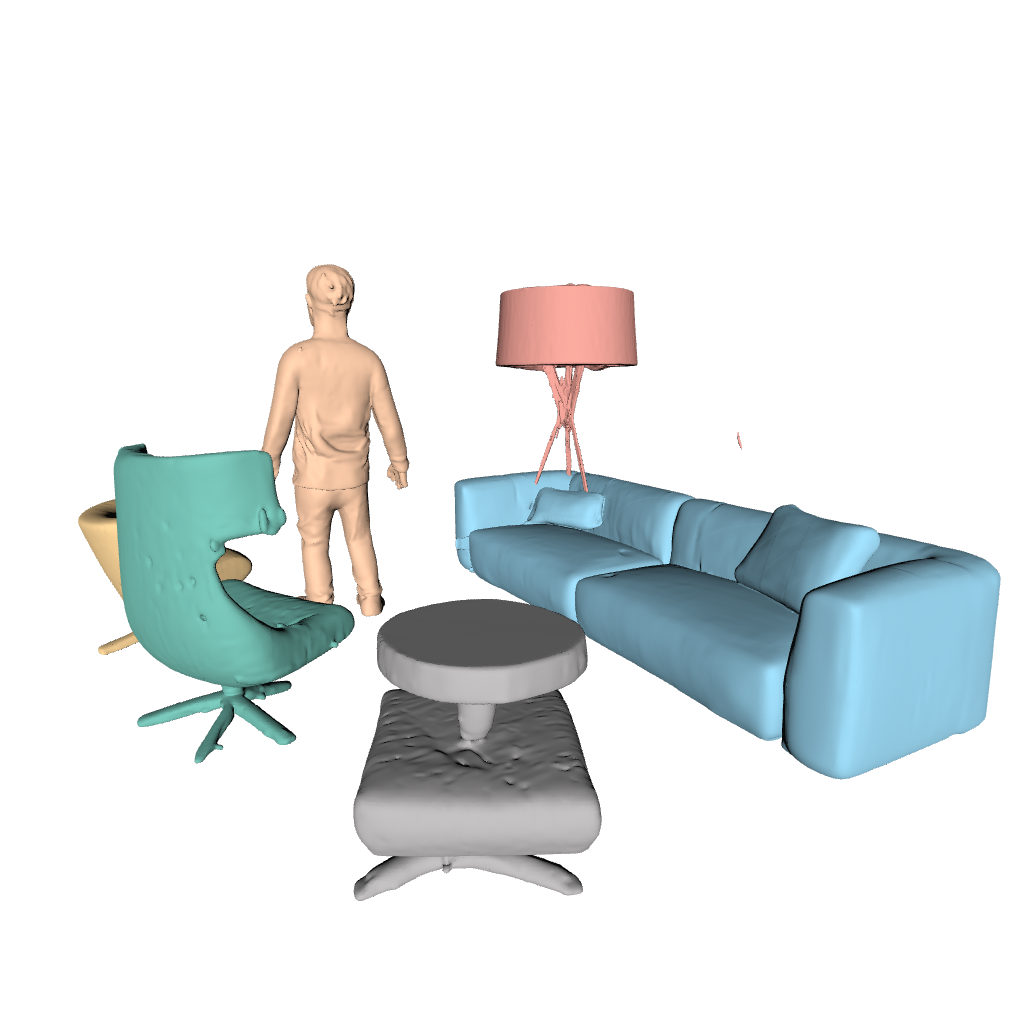} \end{tabular} &
      \begin{tabular}{@{}c@{}} \includegraphics[width=0.22\columnwidth]{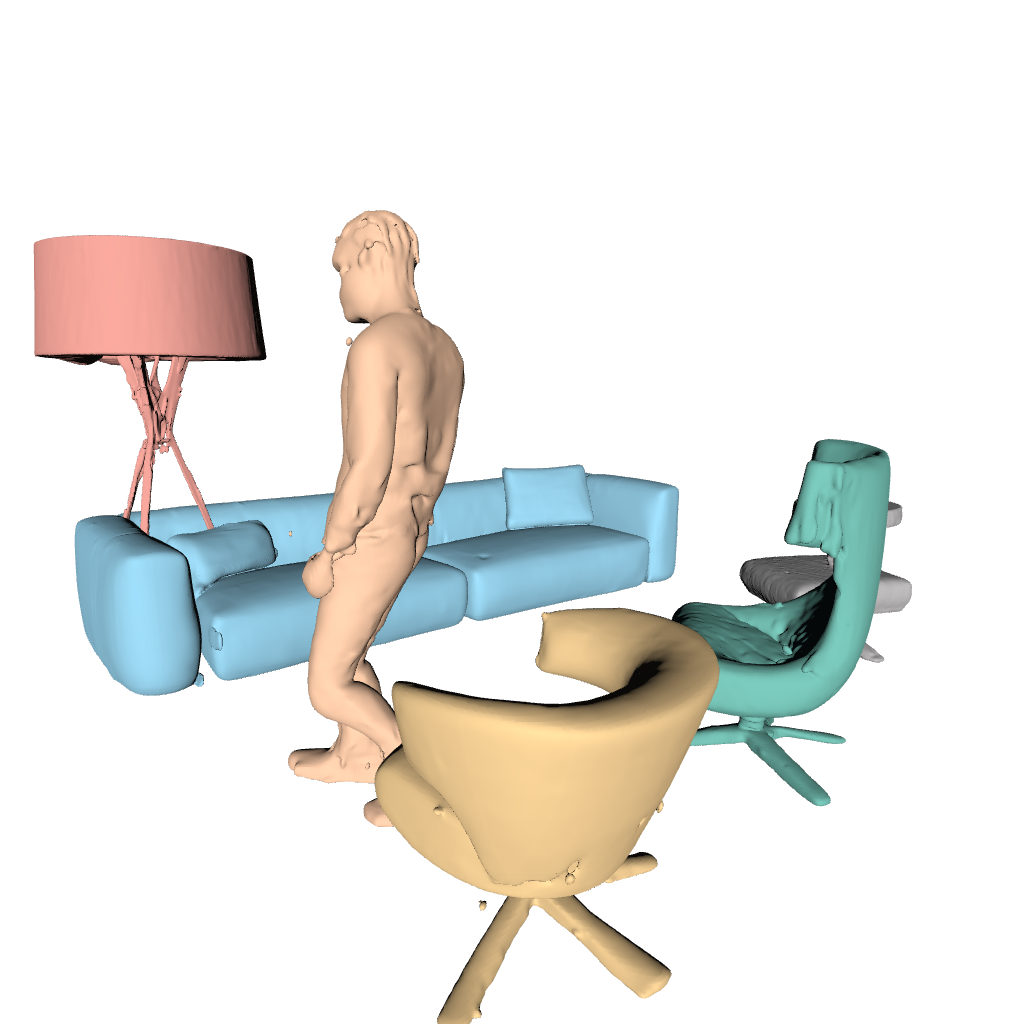} \\[-1.5pt] \includegraphics[width=0.22\columnwidth]{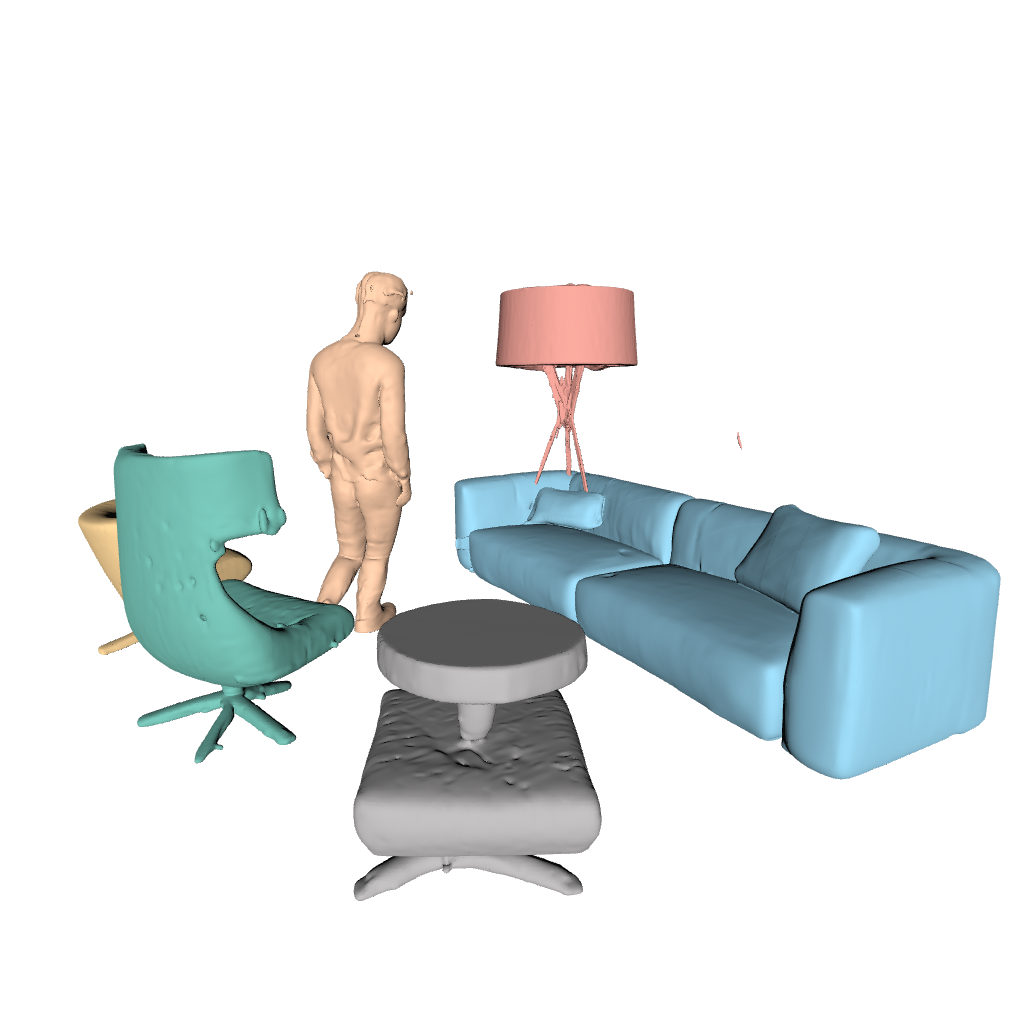} \end{tabular} &
      \begin{tabular}{@{}c@{}} \includegraphics[width=0.22\columnwidth]{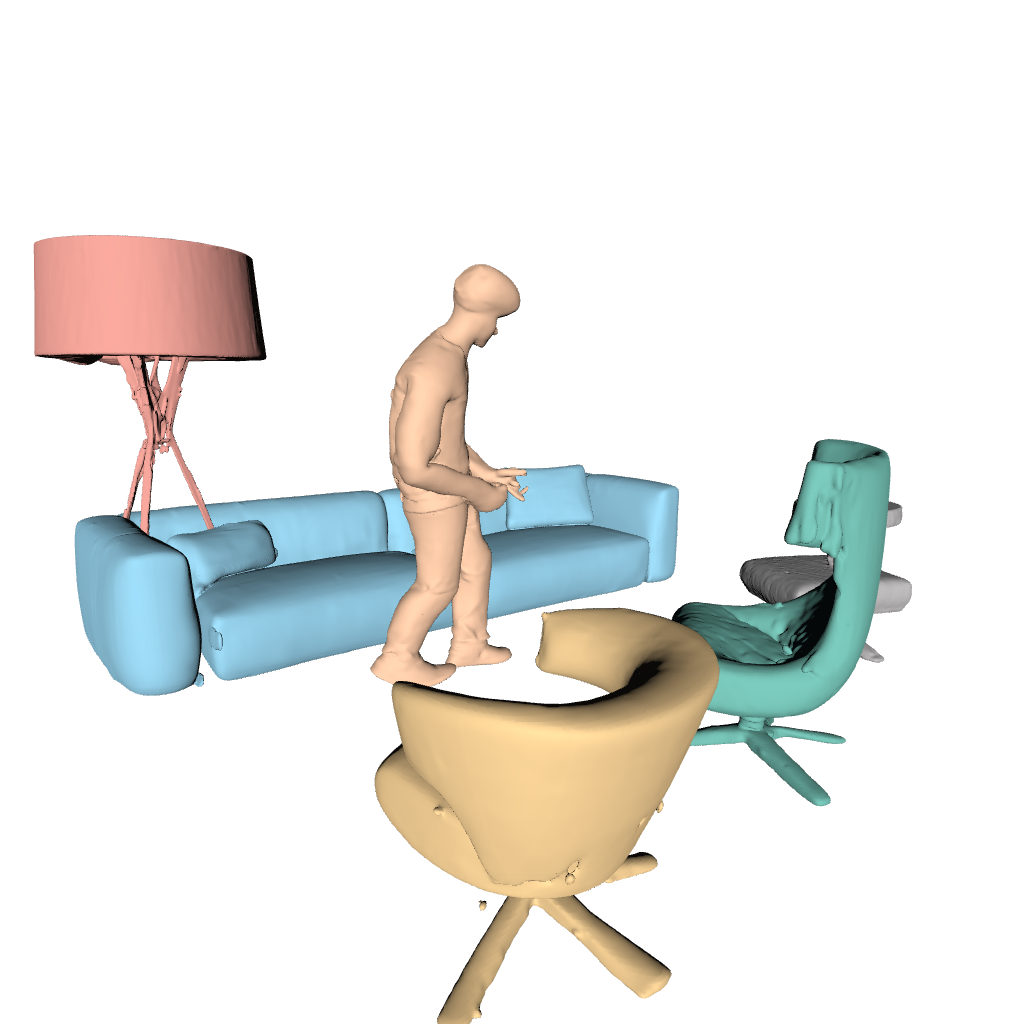} \\[-1.5pt] \includegraphics[width=0.22\columnwidth]{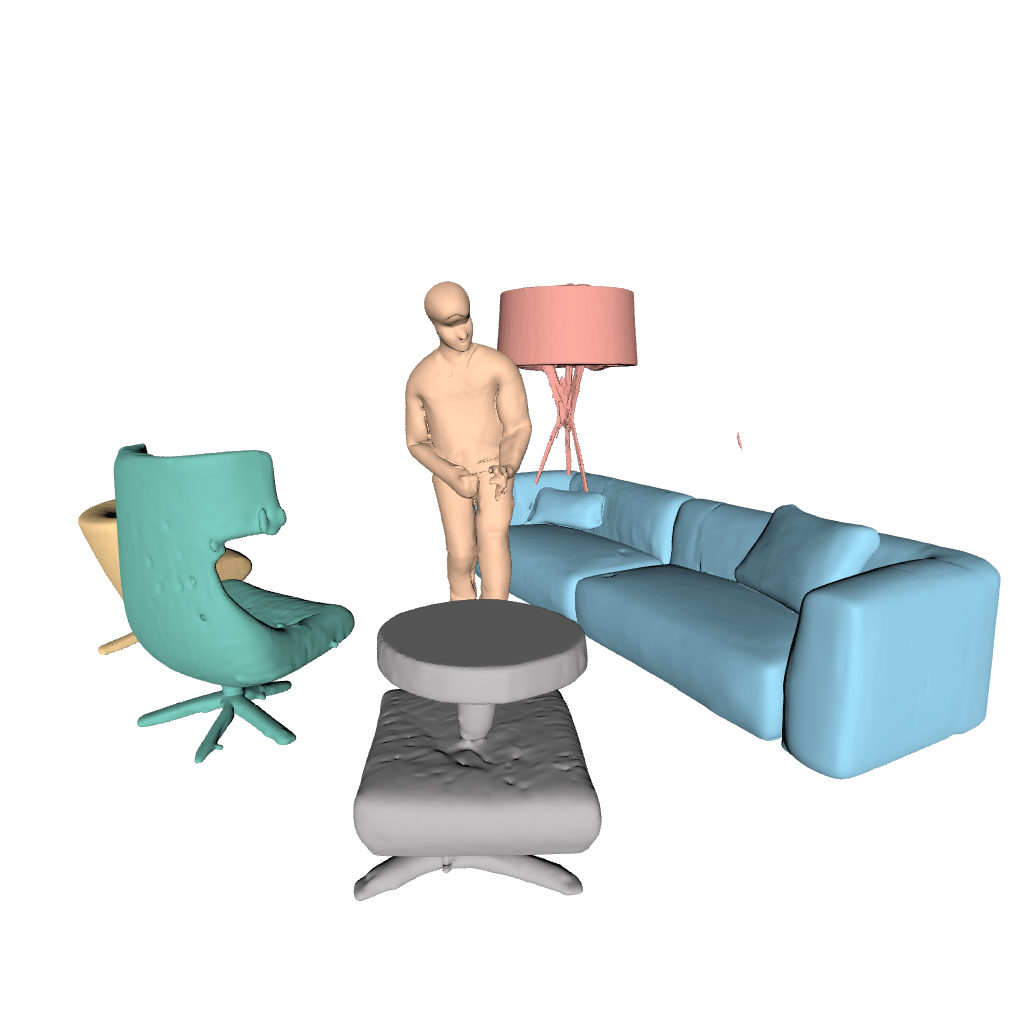} \end{tabular} &
      \begin{tabular}{@{}c@{}} \includegraphics[width=0.22\columnwidth]{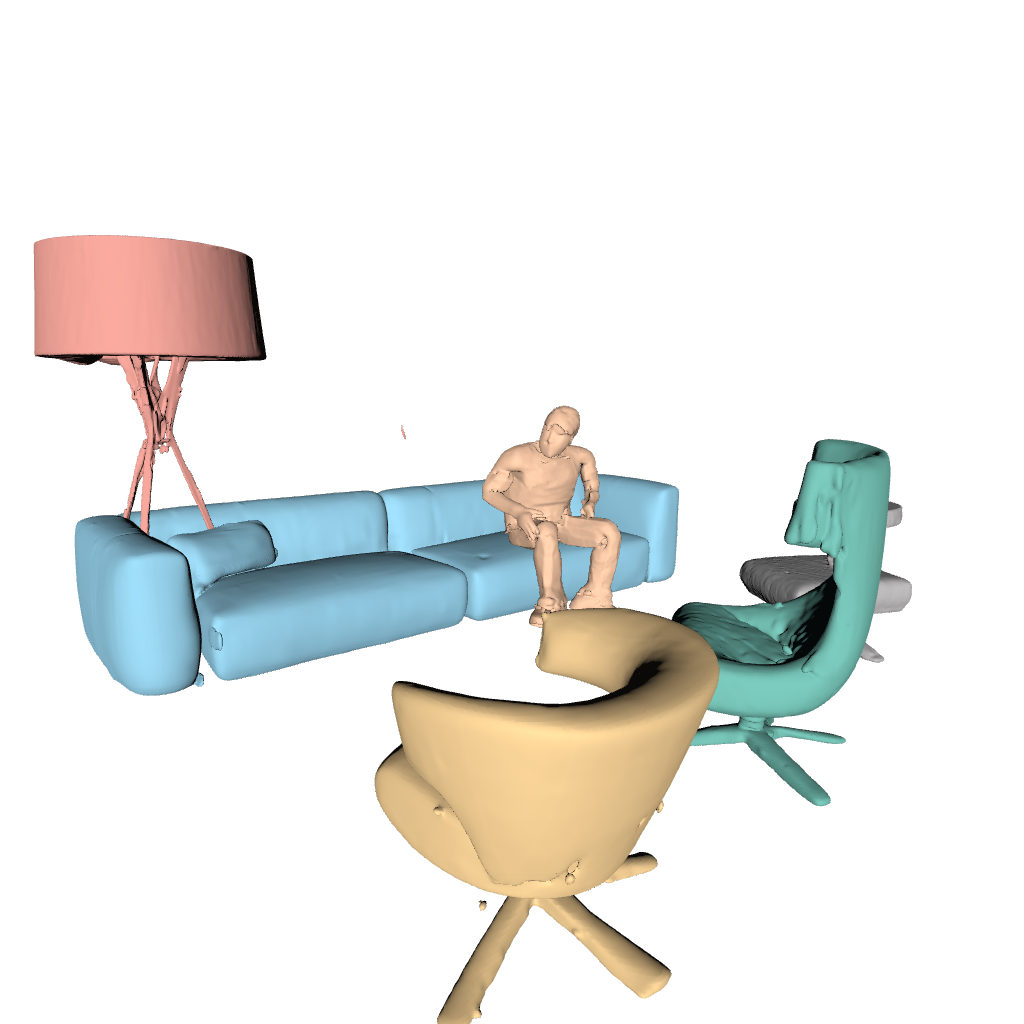} \\[-1.5pt] \includegraphics[width=0.22\columnwidth]{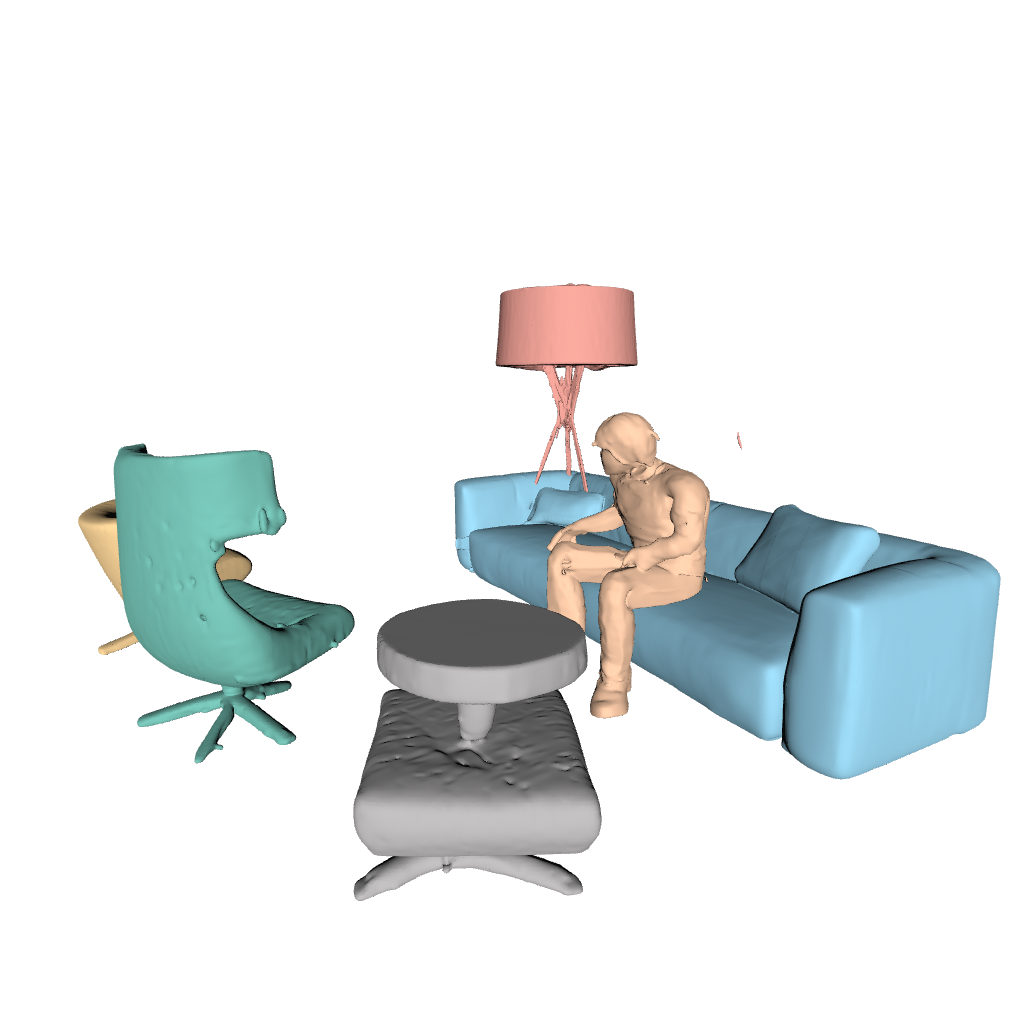} \end{tabular}
    \end{tabular} \\

    % ==================== SECOND EXAMPLE ROW (Input) ====================
    \rotatebox{90}{Input} &
    % Left Group (IAN - INPUT)
    \begin{tabular}{@{}cccc@{}}
      \includegraphics[width=0.22\columnwidth]{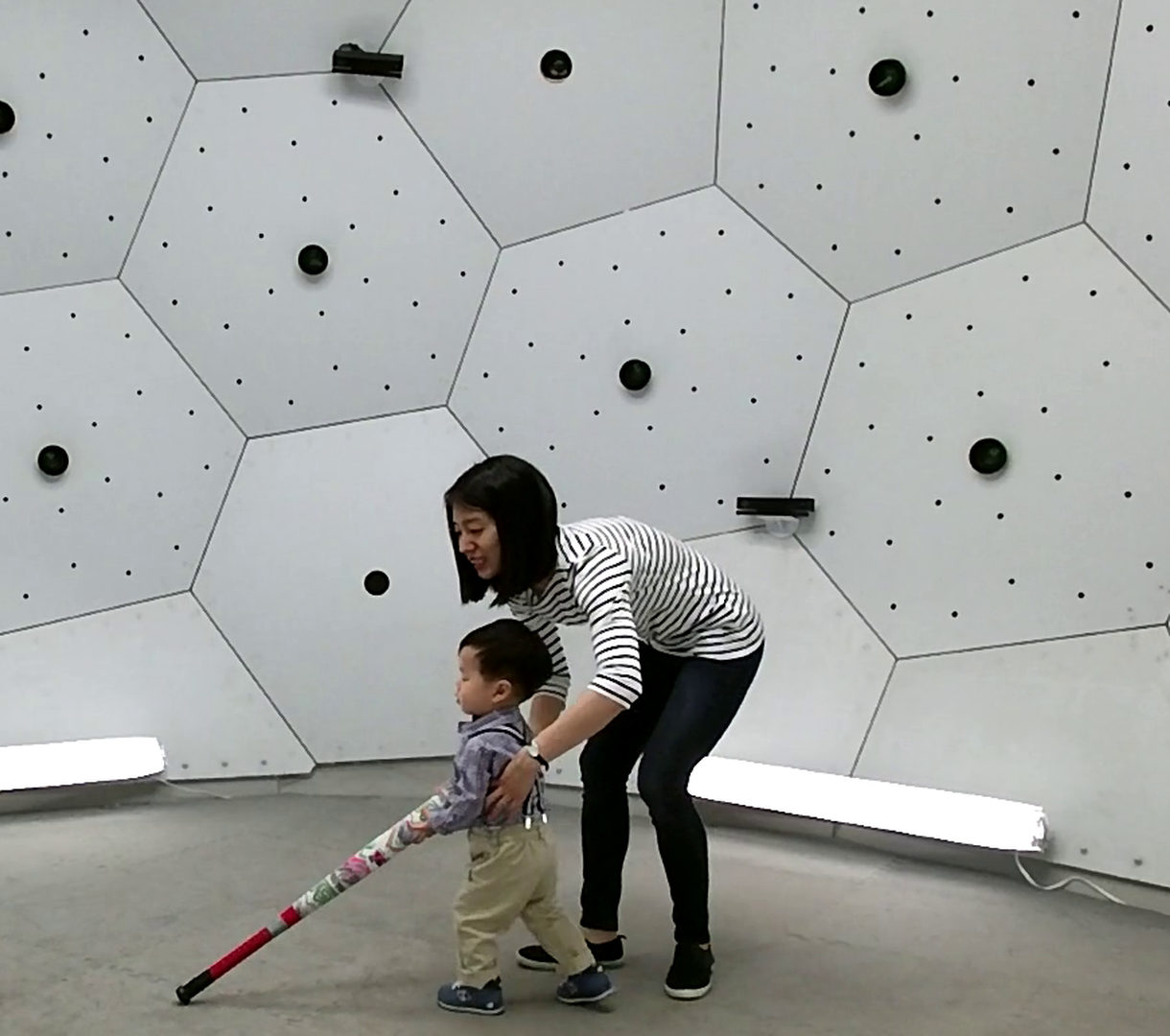} &
      \includegraphics[width=0.22\columnwidth]{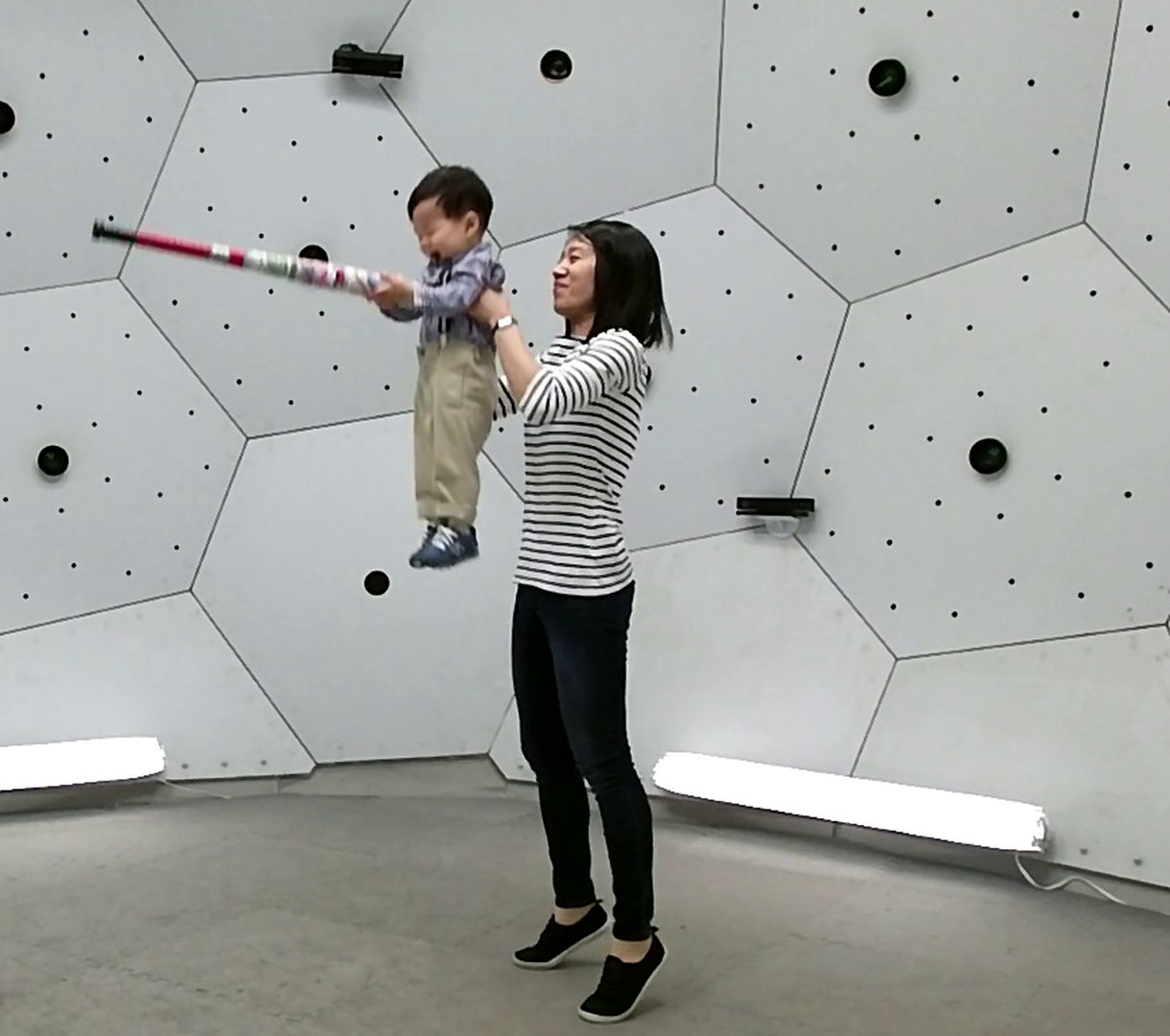} &
      \includegraphics[width=0.22\columnwidth]{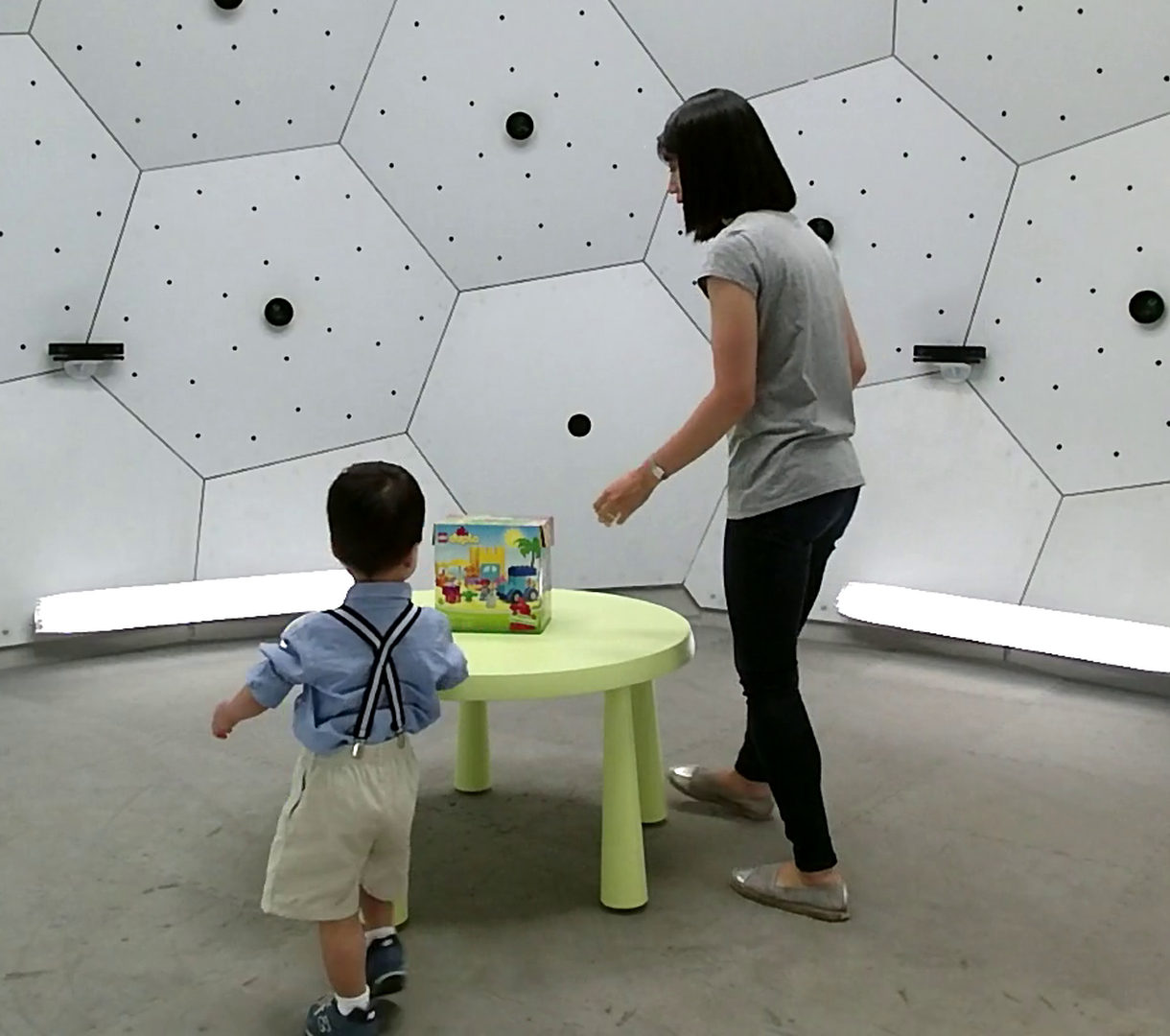} &
      \includegraphics[width=0.22\columnwidth]{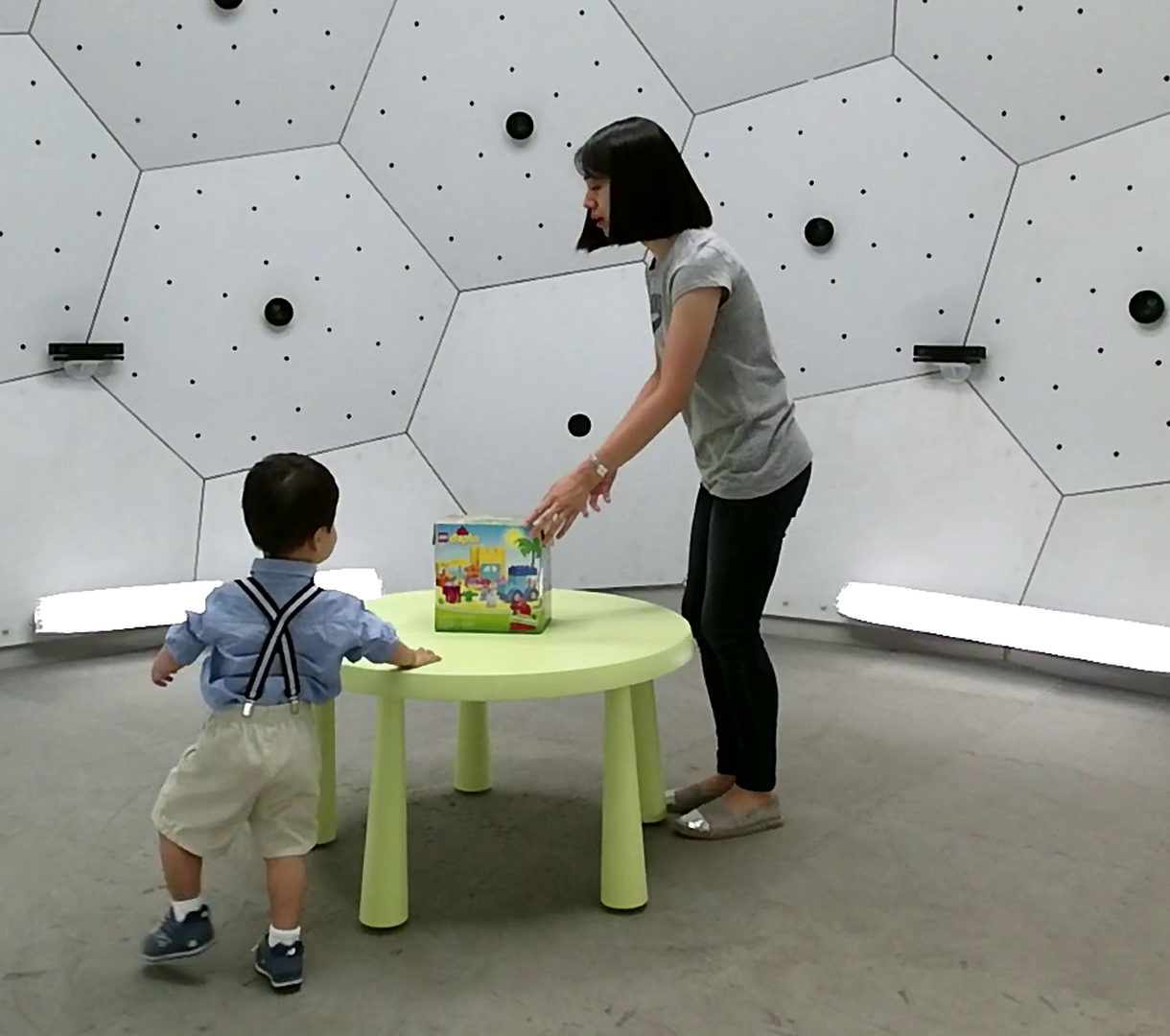}
    \end{tabular} &
    % Right Group (CAT - INPUT)
    \begin{tabular}{@{}cccc@{}}
      \includegraphics[width=0.22\columnwidth]{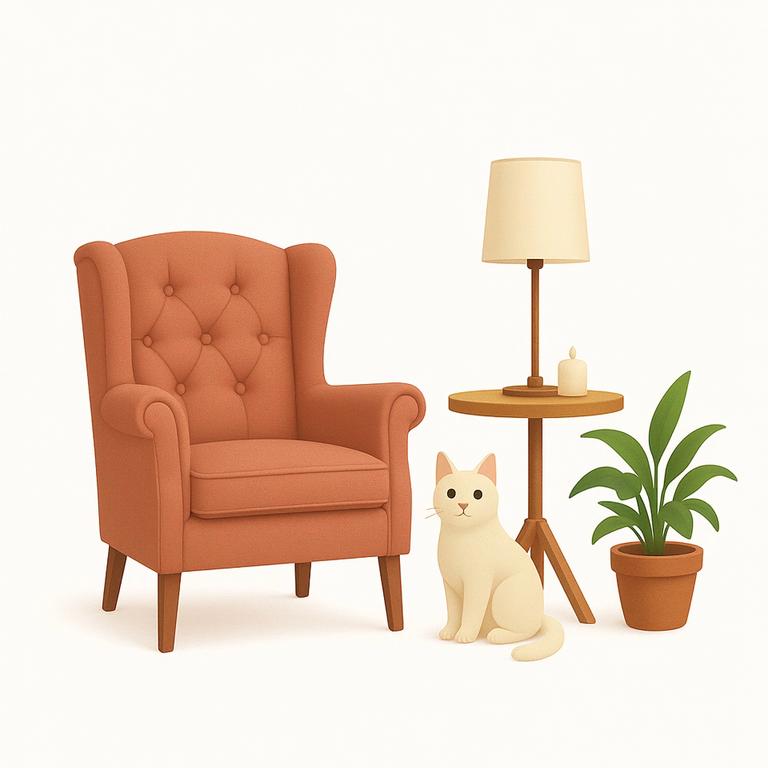} &
      \includegraphics[width=0.22\columnwidth]{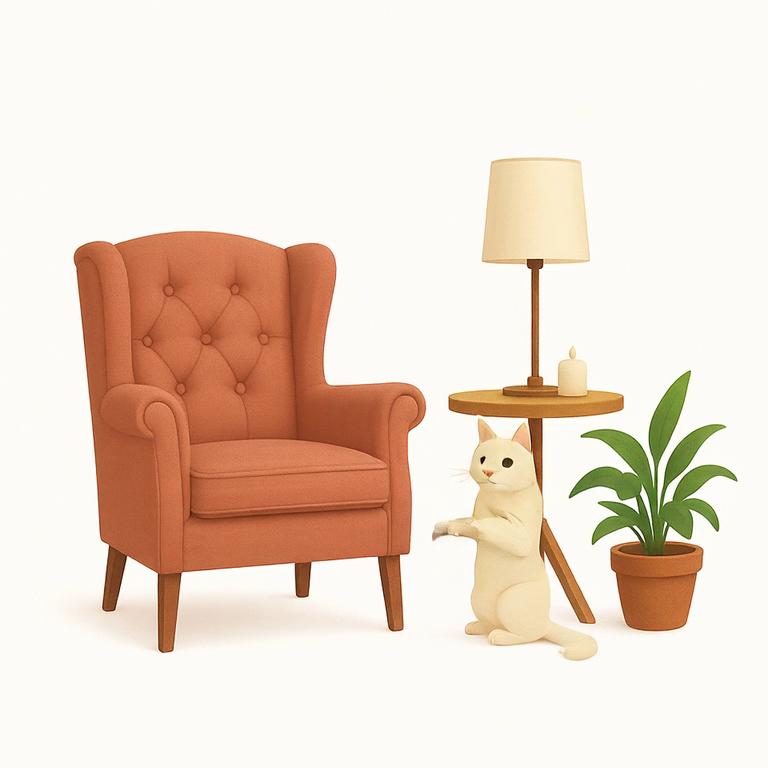} &
      \includegraphics[width=0.22\columnwidth]{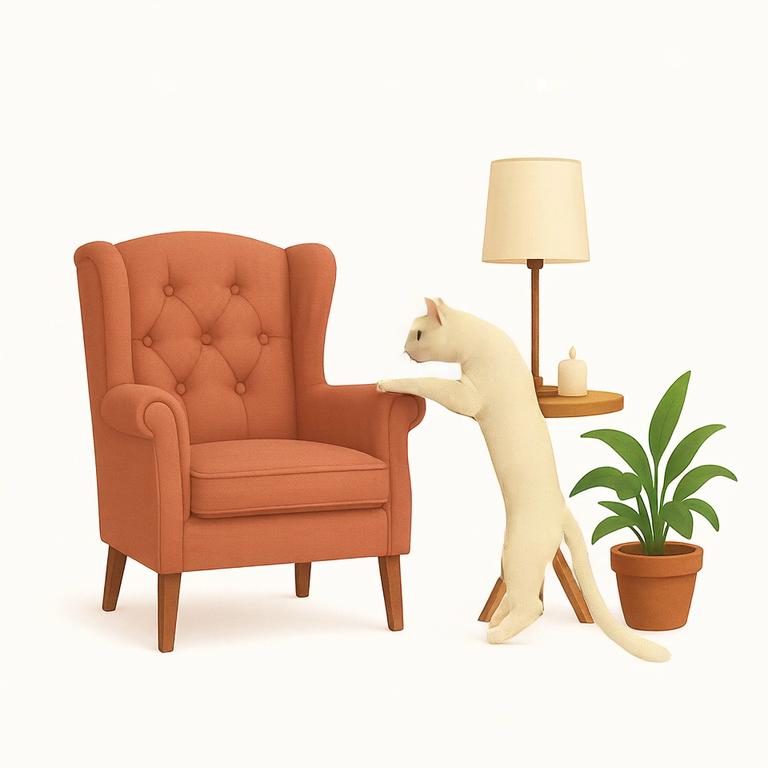} &
      \includegraphics[width=0.22\columnwidth]{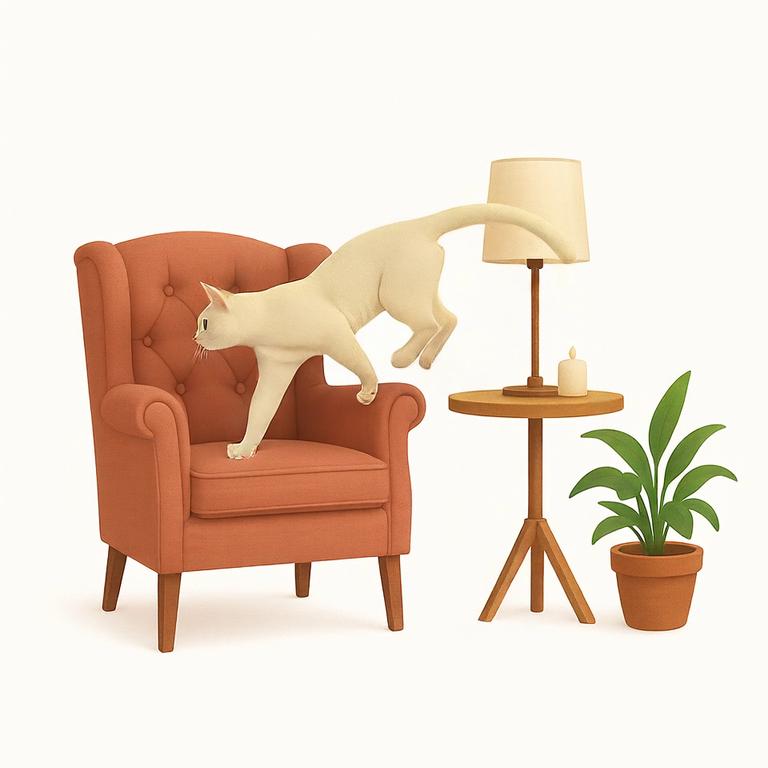}
    \end{tabular} \\

    % ==================== SECOND EXAMPLE ROW (Ours) ====================
    \rotatebox{90}{\textbf{Ours}} &
    % Left Group (IAN - OURS)
    \begin{tabular}{@{}cccc@{}}
      \begin{tabular}{@{}c@{}} \includegraphics[width=0.22\columnwidth]{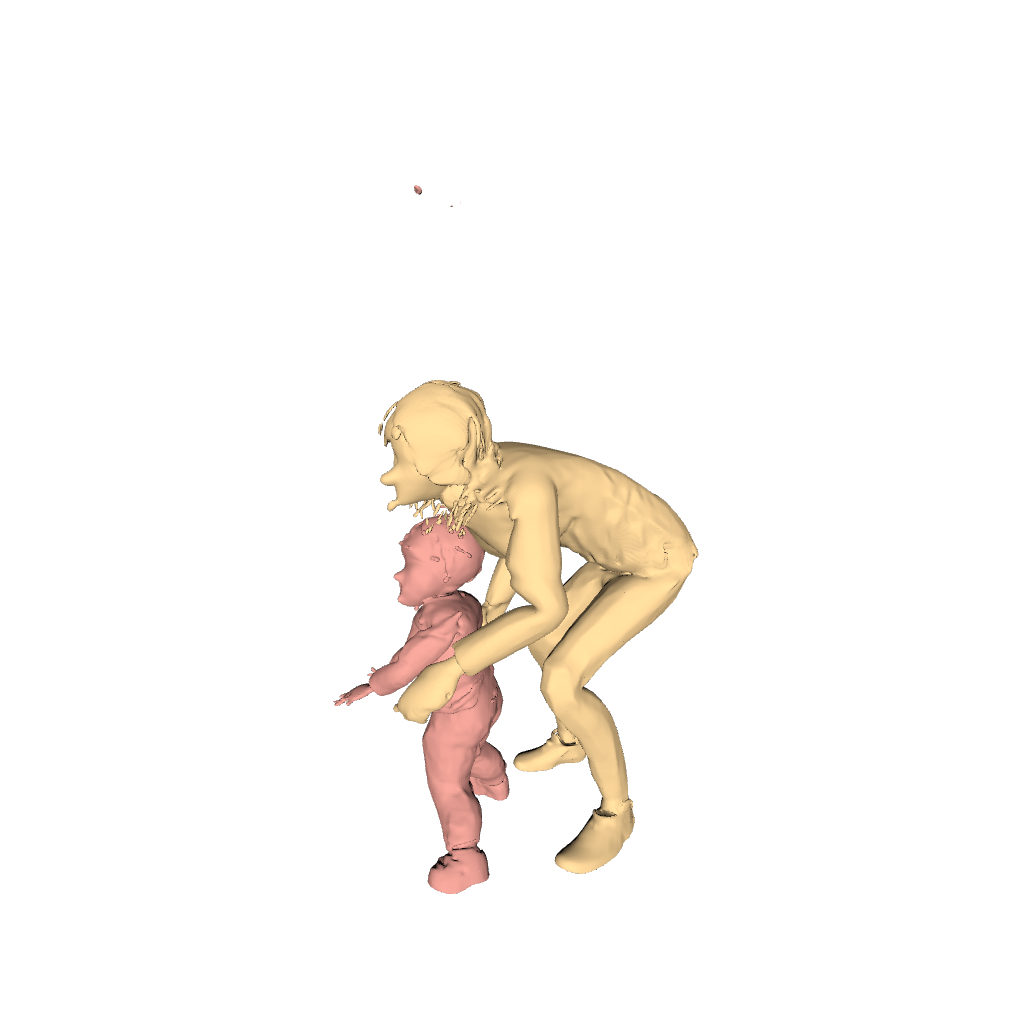} \\[-1.5pt] \includegraphics[width=0.22\columnwidth]{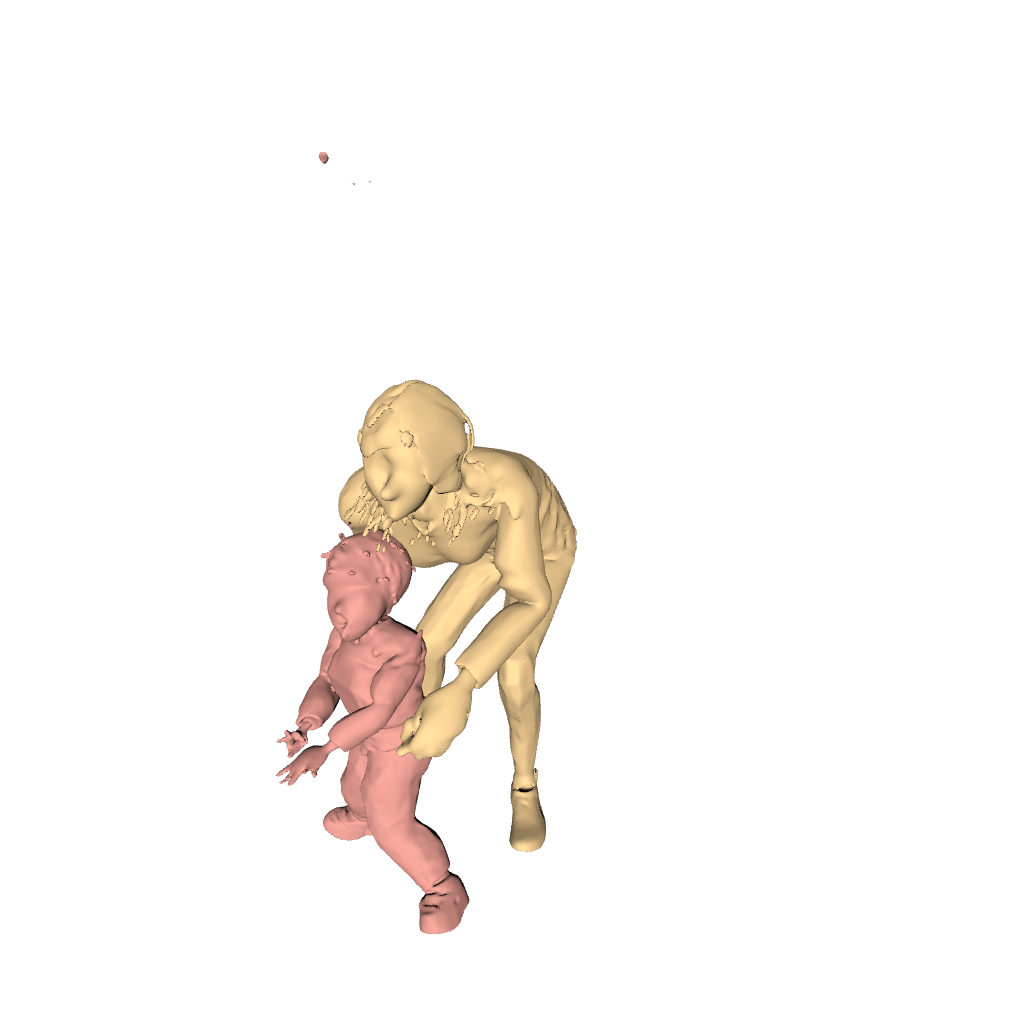} \end{tabular} &
      \begin{tabular}{@{}c@{}} \includegraphics[width=0.22\columnwidth]{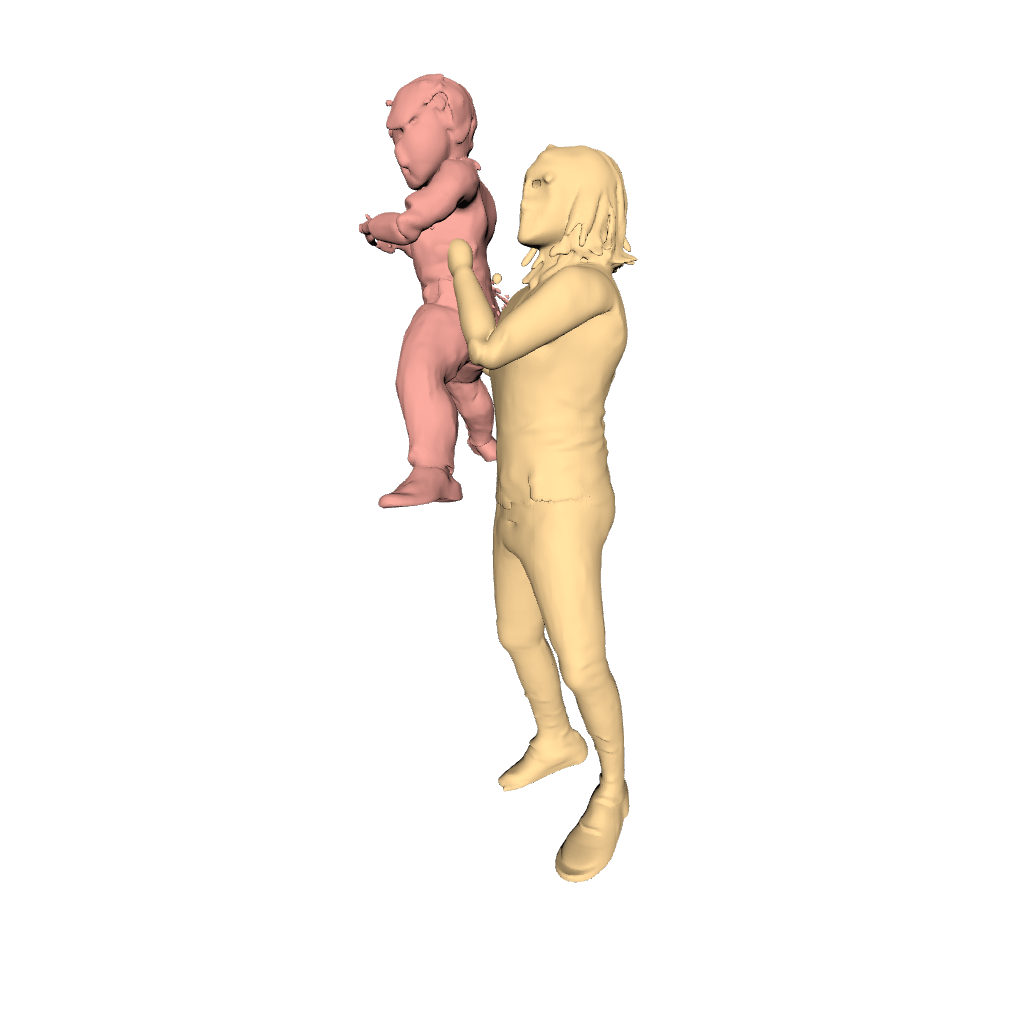} \\[-1.5pt] \includegraphics[width=0.22\columnwidth]{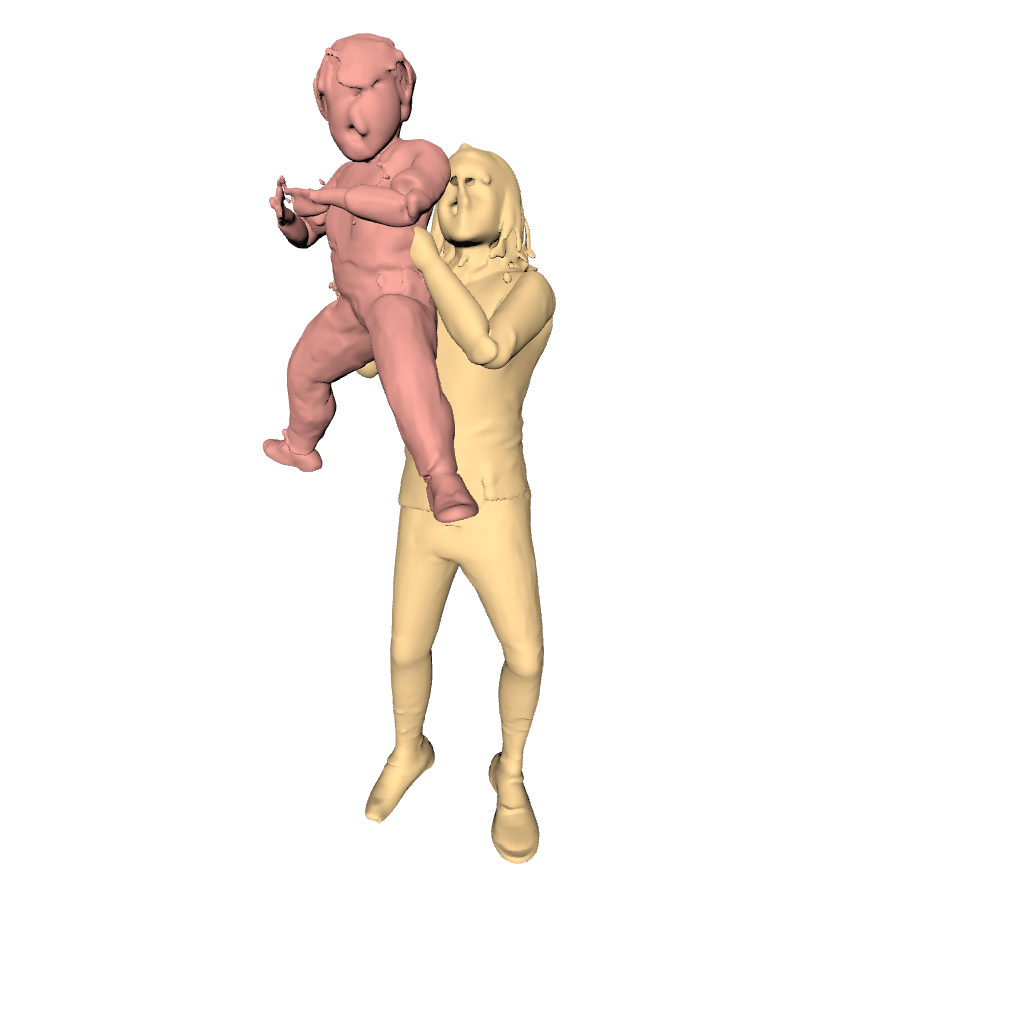} \end{tabular} &
      \begin{tabular}{@{}c@{}} \includegraphics[width=0.22\columnwidth]{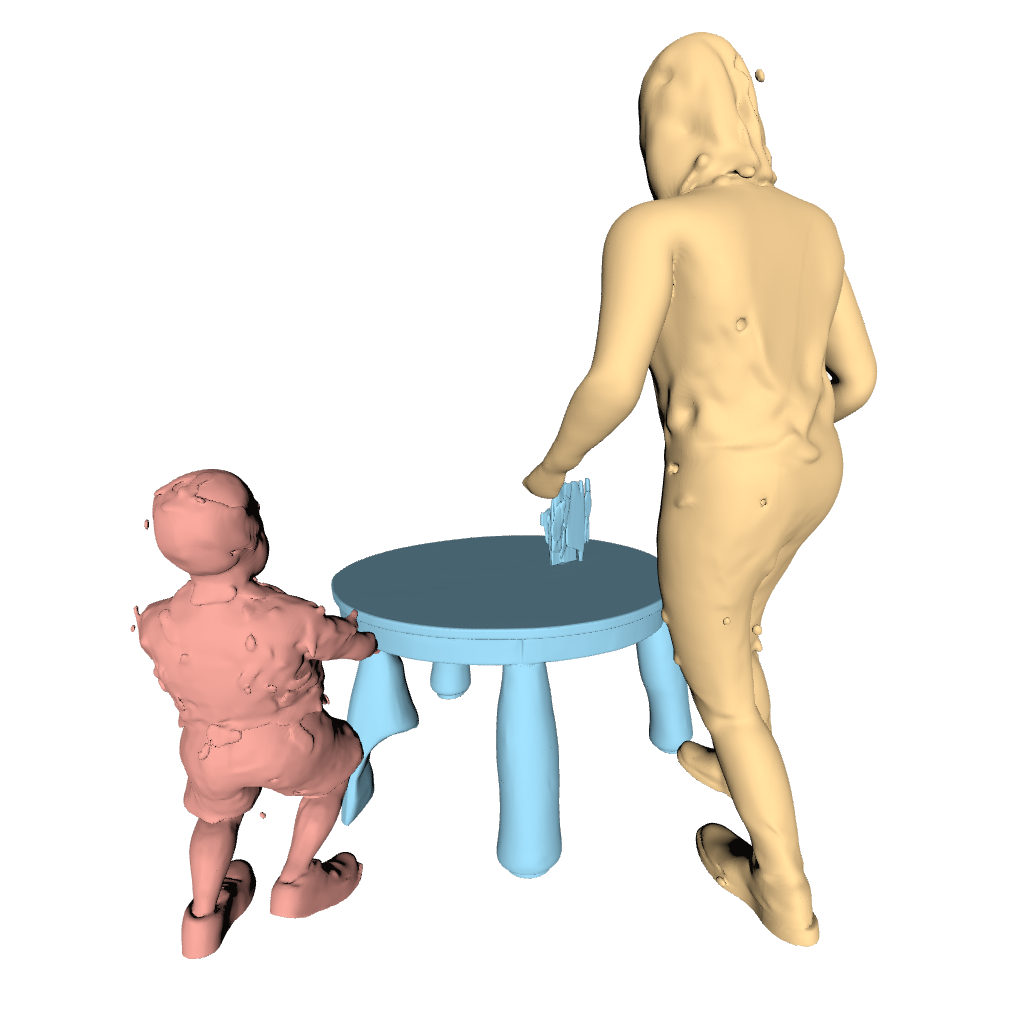} \\[-1.5pt] \includegraphics[width=0.22\columnwidth]{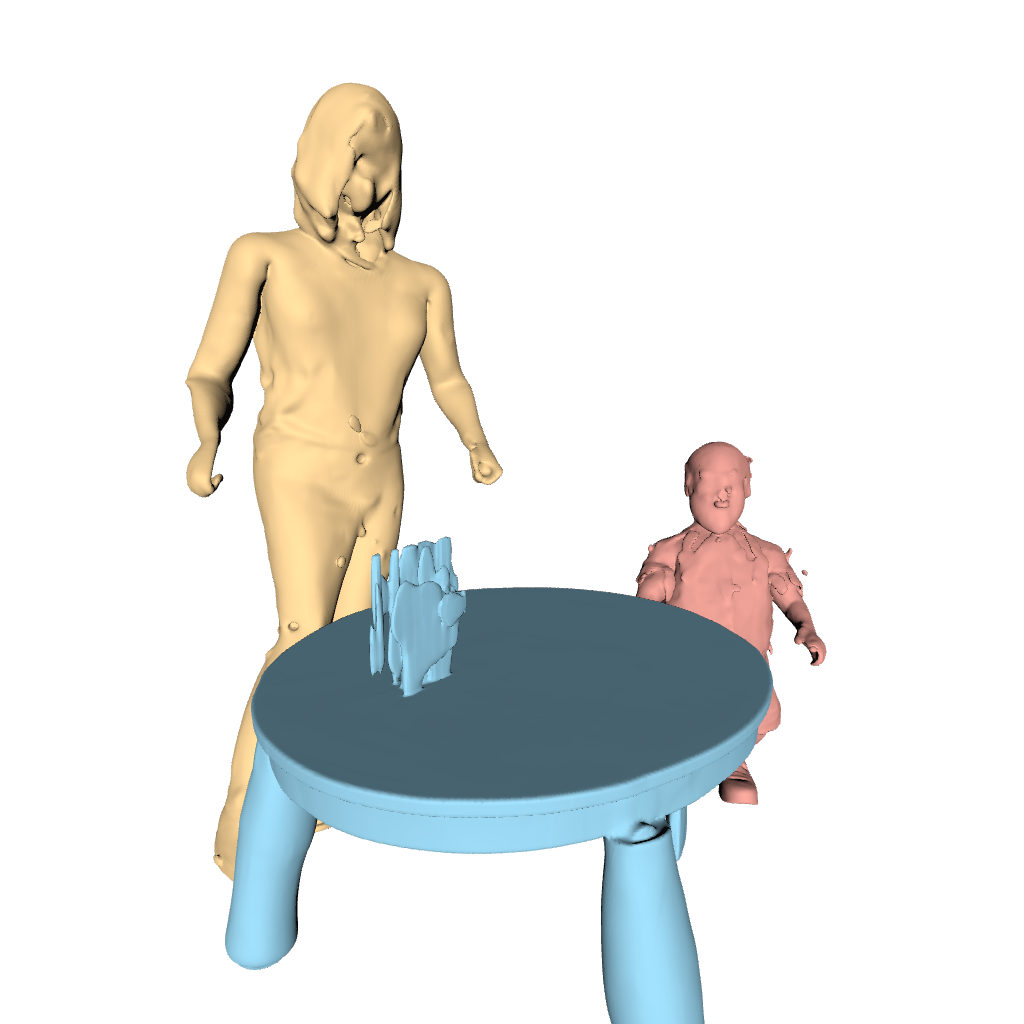} \end{tabular} &
      \begin{tabular}{@{}c@{}} \includegraphics[width=0.22\columnwidth]{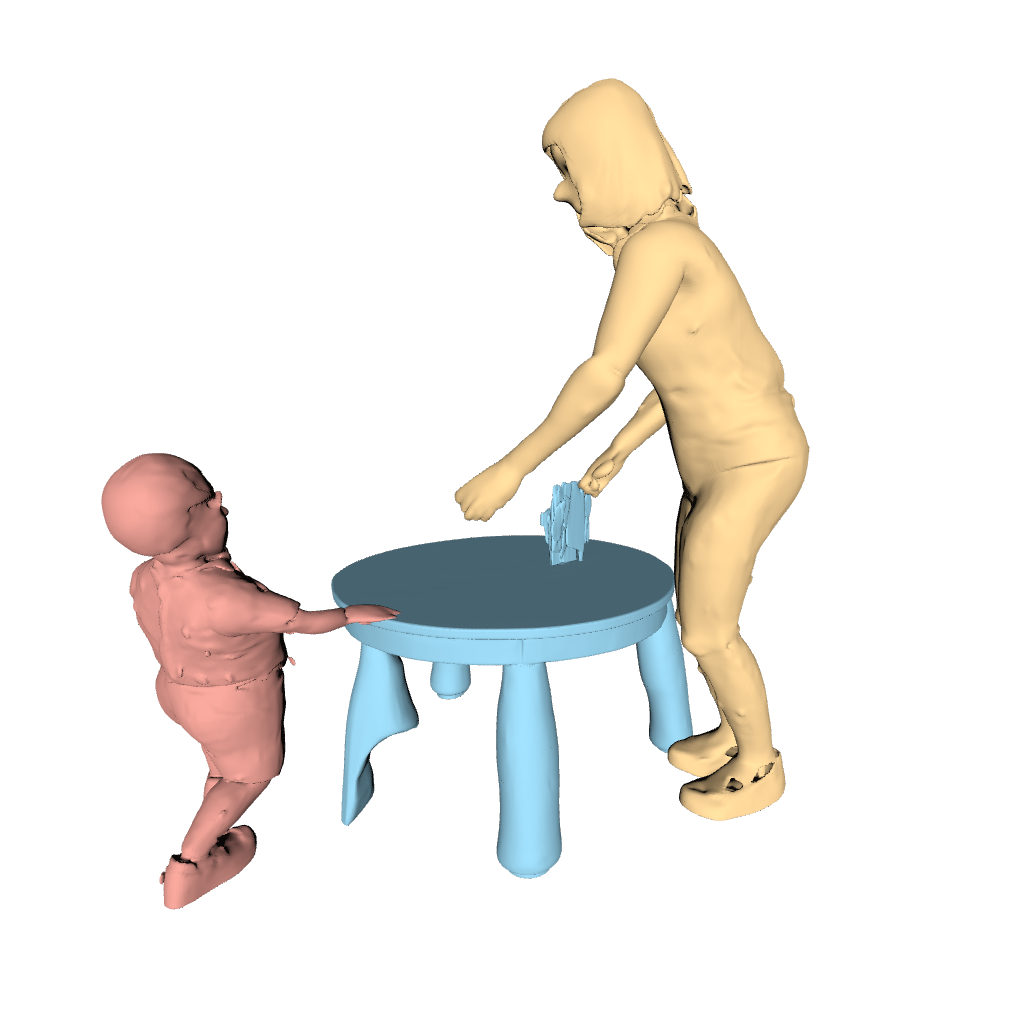} \\[-1.5pt] \includegraphics[width=0.22\columnwidth]{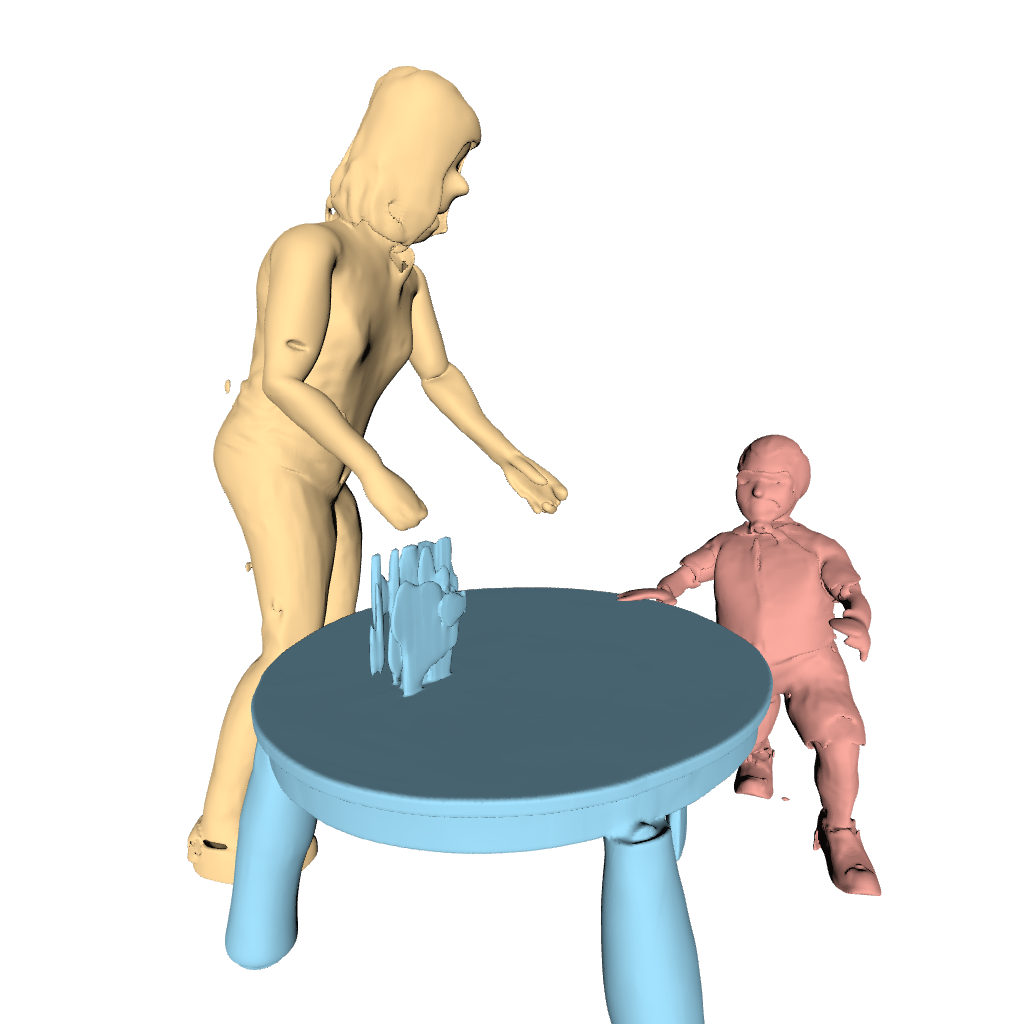} \end{tabular}
    \end{tabular} &
    % Right Group (CAT - OURS)
    \begin{tabular}{@{}cccc@{}}
      \begin{tabular}{@{}c@{}} \adjincludegraphics[width=0.22\columnwidth, trim={{0.15\width} {0.15\height} {0.15\width} {0.15\height}}, clip]{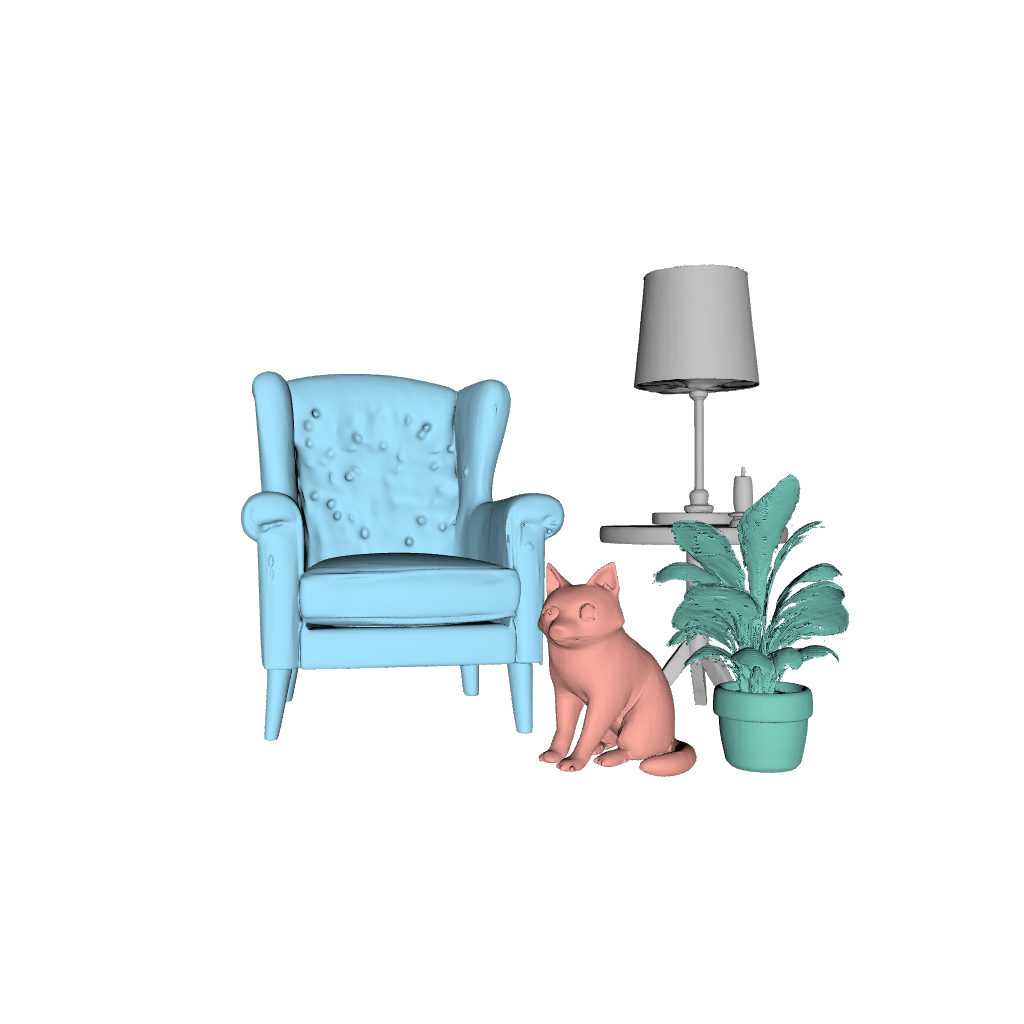} \\[-1.5pt] \adjincludegraphics[width=0.22\columnwidth, trim={{0.15\width} {0.15\height} {0.15\width} {0.15\height}}, clip]{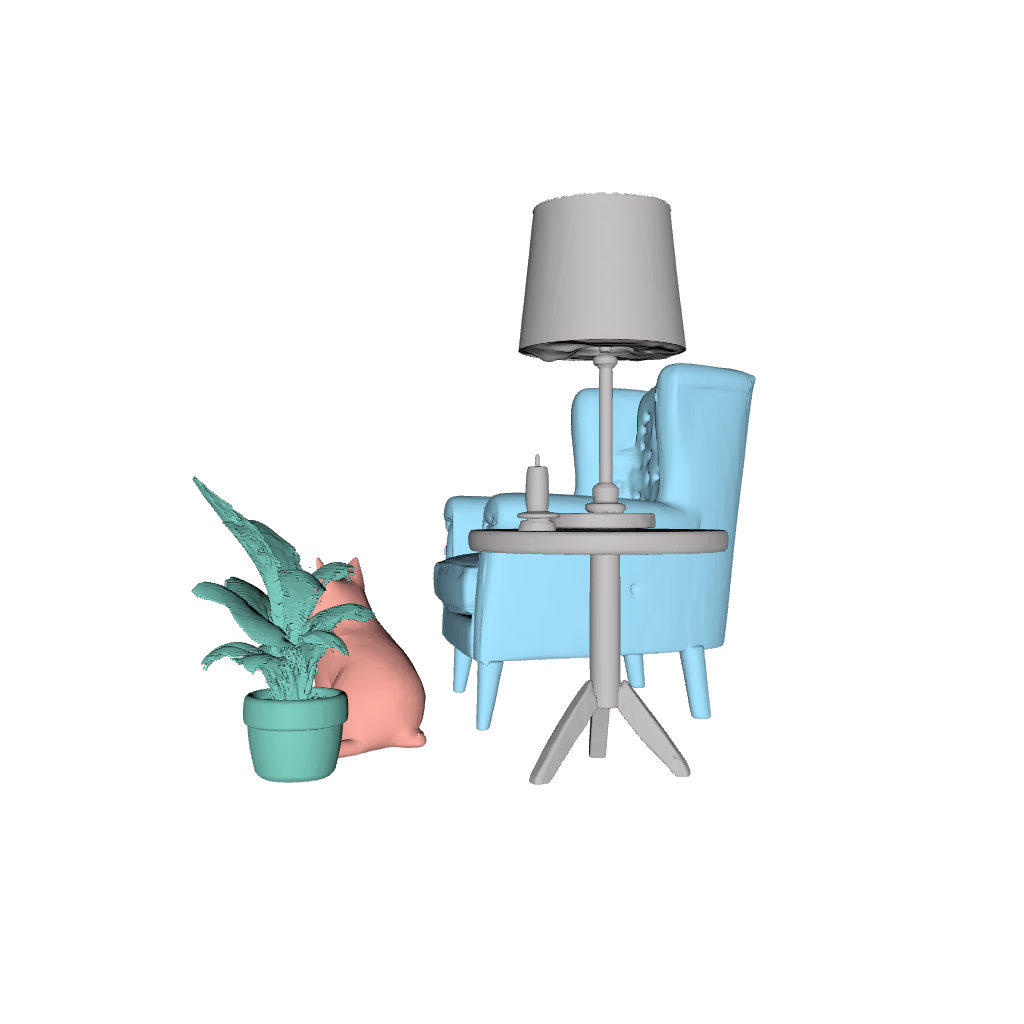} \end{tabular} &
      \begin{tabular}{@{}c@{}} \adjincludegraphics[width=0.22\columnwidth, trim={{0.15\width} {0.15\height} {0.15\width} {0.15\height}}, clip]{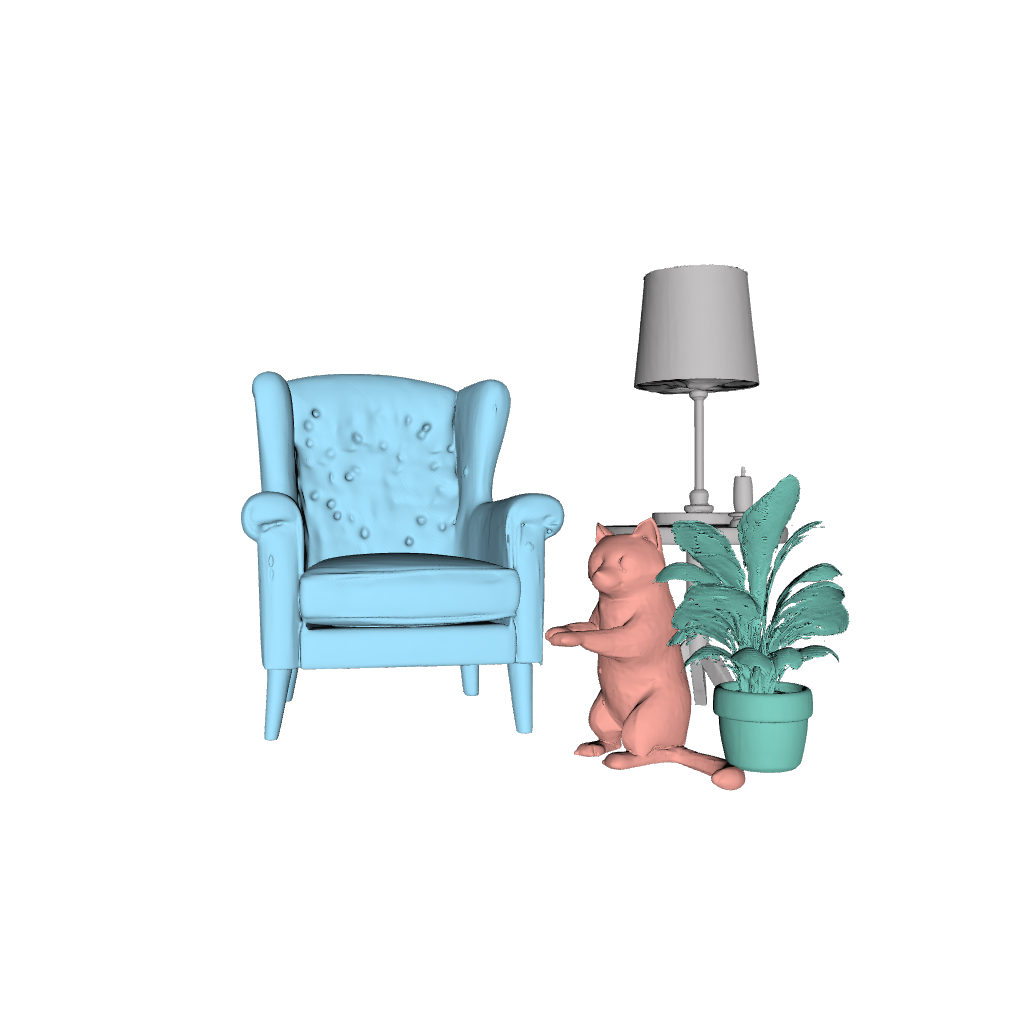} \\[-1.5pt] \adjincludegraphics[width=0.22\columnwidth, trim={{0.15\width} {0.15\height} {0.15\width} {0.15\height}}, clip]{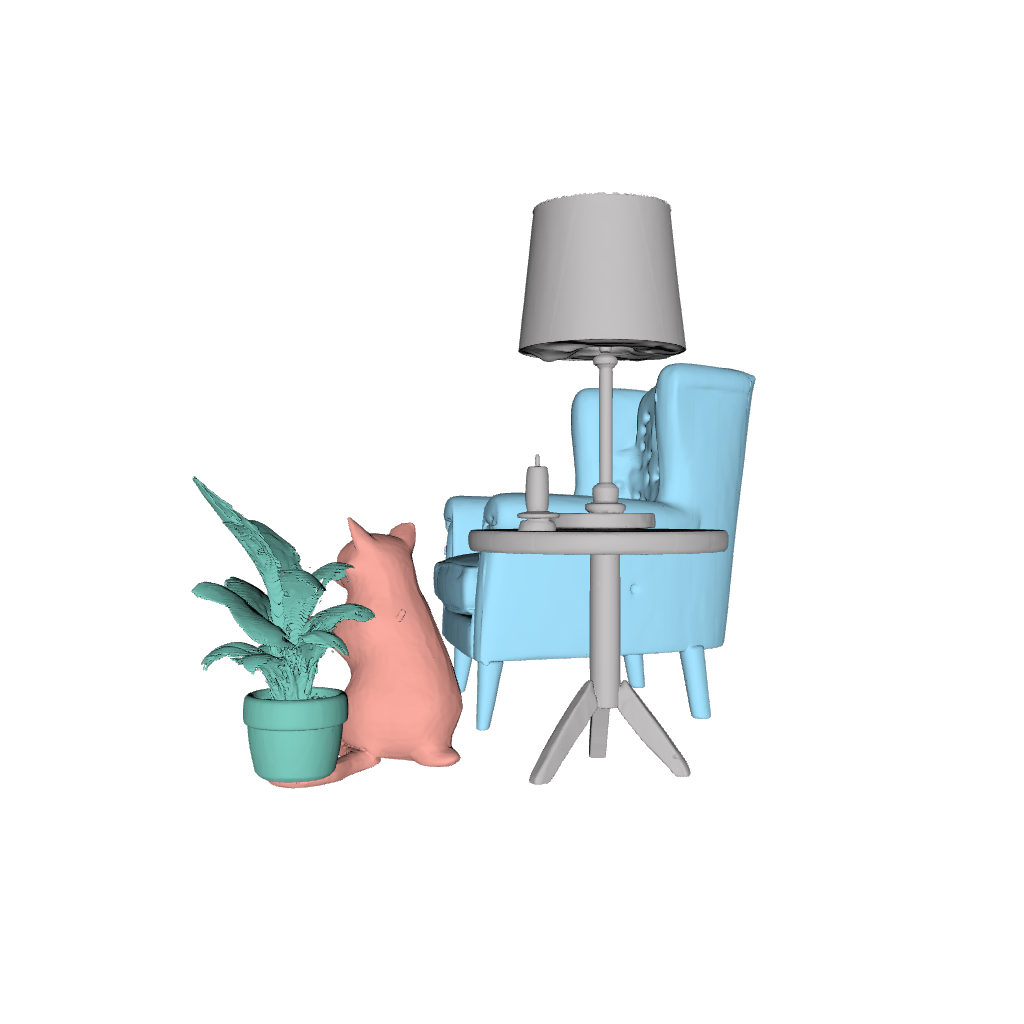} \end{tabular} &
      \begin{tabular}{@{}c@{}} \adjincludegraphics[width=0.22\columnwidth, trim={{0.15\width} {0.15\height} {0.15\width} {0.15\height}}, clip]{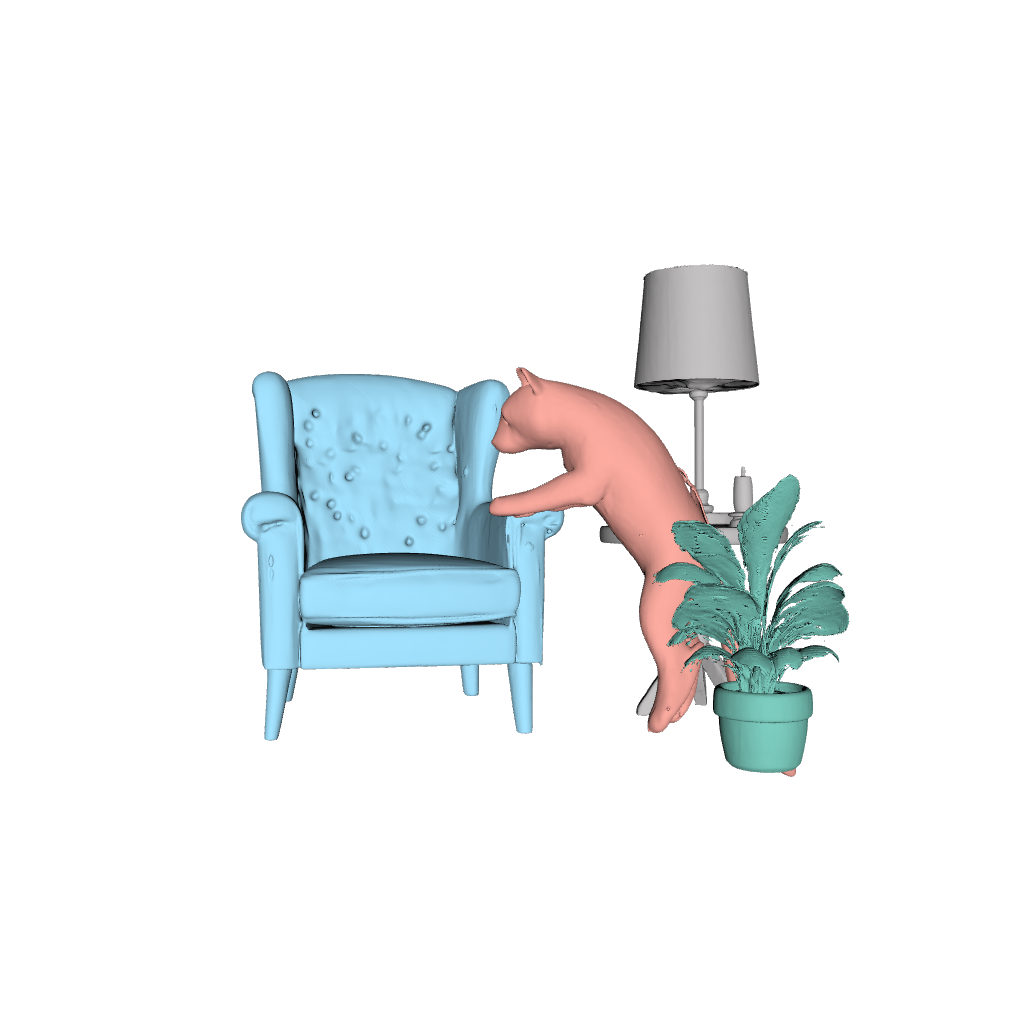} \\[-1.5pt] \adjincludegraphics[width=0.22\columnwidth, trim={{0.15\width} {0.15\height} {0.15\width} {0.15\height}}, clip]{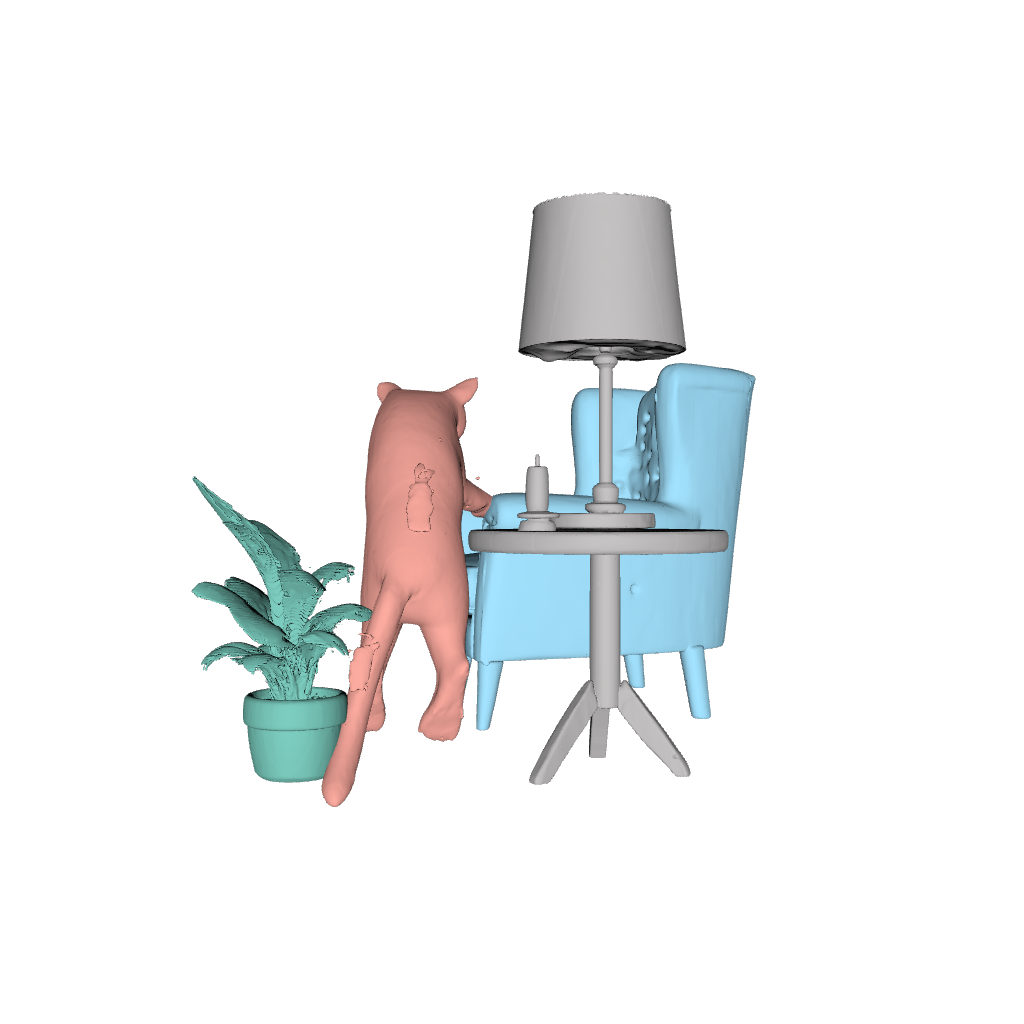} \end{tabular} &
      \begin{tabular}{@{}c@{}} \adjincludegraphics[width=0.22\columnwidth, trim={{0.15\width} {0.15\height} {0.15\width} {0.15\height}}, clip]{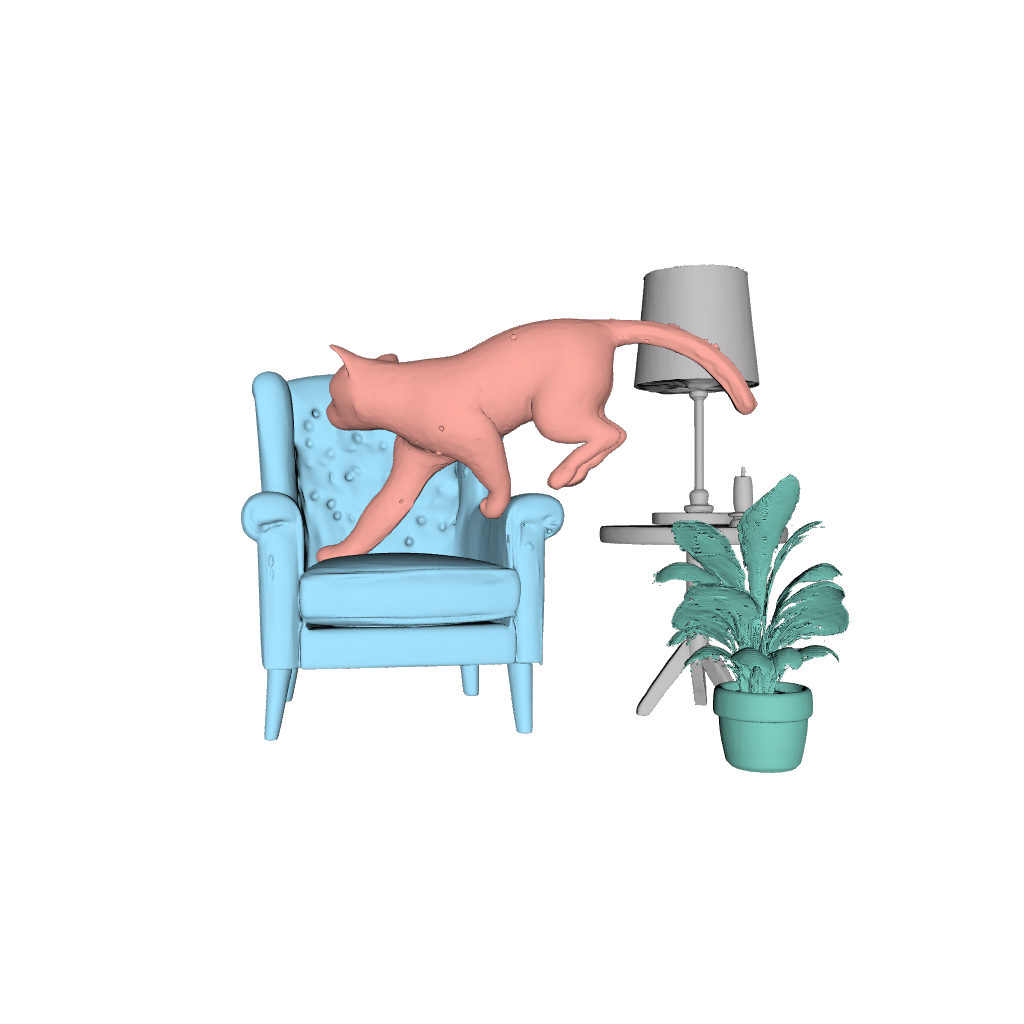} \\[-1.5pt] \adjincludegraphics[width=0.22\columnwidth, trim={{0.15\width} {0.15\height} {0.15\width} {0.15\height}}, clip]{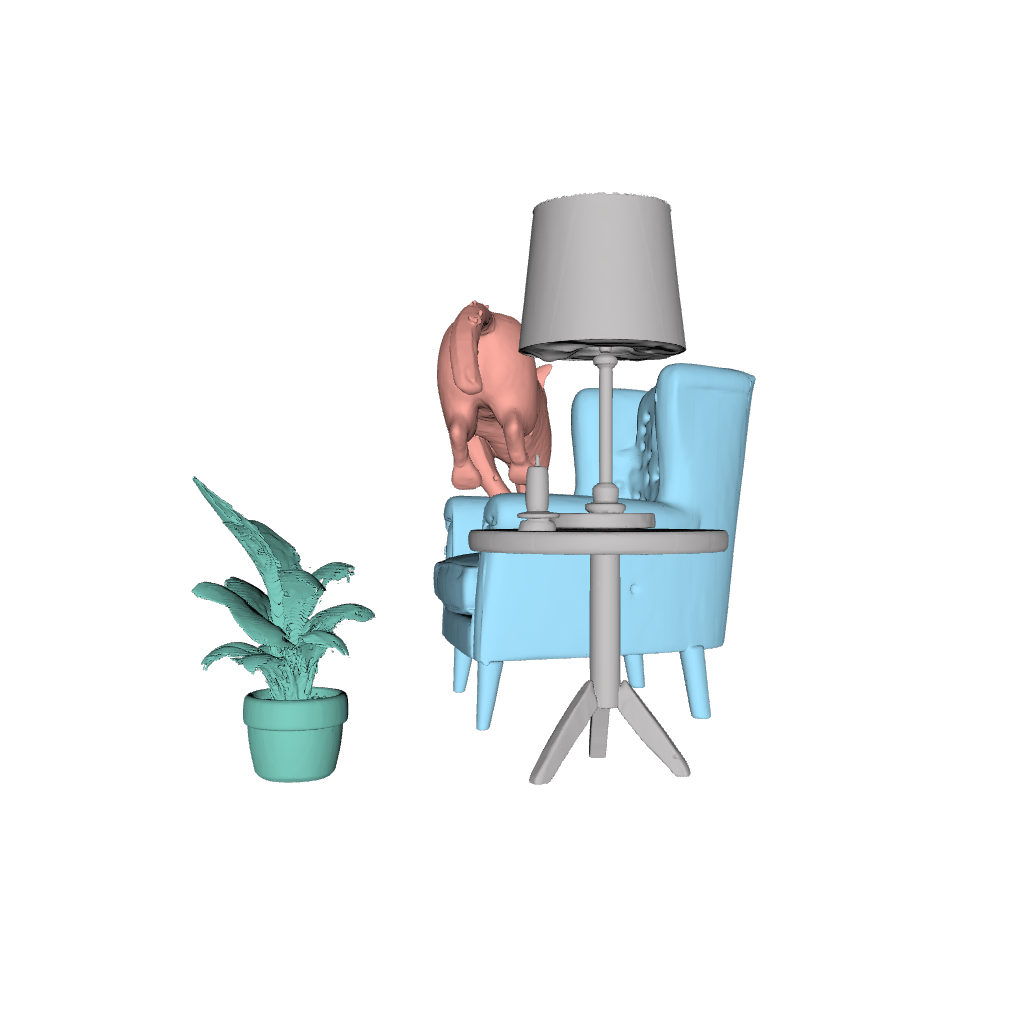} \end{tabular}
    \end{tabular} \\
  \end{tabular}
  \caption{Qualitative results on temporal sequences. Top rows show input frames; bottom rows show our generated reconstructions from two vertically stacked camera views. The examples shown are from content produced by ChatGPT~\cite{chatgpt} and animated with Wan~\cite{wan}, the CMU Panoptic dataset~\cite{panoptic} (sequences \texttt{160401\_ian3} and \texttt{160906\_ian2}), and the PROX dataset~\cite{prox} (\texttt{N3OpenArea\_00158\_02}). Our method maintains temporal consistency and spatial realism across both real and synthetic sources.}
  \label{fig:temporal_combined}
  % ==================== MASTER TABLE END ====================
\end{figure*}

\subsection{Implementation Details}
We build on the 21-block DiT backbone of TripoSG~\cite{li2025triposghighfidelity3dshape}. Even-indexed blocks operate in the \emph{spatial} mode, while odd-indexed blocks operate in the \emph{temporal} mode. This depth-wise alternation follows the multi-instance reasoning intuition~\cite{lin2025partcrafterstructured3dmesh} while keeping the original DiT~\cite{li2025triposghighfidelity3dshape} depth and width unchanged. We fully fine-tune all weights of the pretrained TripoSG DiT. We cap the supervision targets at 8 parts for the spatial path and 8 frames for the temporal. At each iteration, with probability $0.3$ we train on a \emph{monolithic} sample (single part, single frame) to regularize the model and preserve its learned object prior ($\mathcal{L}_R$).

Training uses two stages on a single NVIDIA H200 GPU. In stage 1 (8k steps), we fine-tune the model with a batch size of $50$ and a learning rate of $1\times 10^{-4}$. In stage 2 (12k steps), we keep the batch size at $50$, enable Diffusion Forcing~\cite{chen2024diffusionforcingnexttokenprediction}, and lower the learning rate to $1\times 10^{-5}$. The full training time is about 2 days.
\section{Experiments}
%We first describe the datasets and the evaluation protocol used in the experiments, followed by the results in each task.

\paragraph{Test Datasets.}
For 3D evaluation, we use the 3D-FRONT~\cite{fu20213dfront3dfurnishedrooms} test set following MIDI’s protocol~\cite{huang2025midimultiinstancediffusionsingle}. For 4D evaluation, we construct two complementary sets: (i) an Objaverse-based subset starting from the animated object list released with \textsc{Puppet-Master}~\cite{li2025puppetmasterscalinginteractivevideo}, from which we select 40 high-quality assets (clean geometry and faithful textures) from Objaverse~\cite{deitke2022objaverseuniverseannotated3d}; and (ii) a DeformingThings4D~\cite{li20214dcomplete} subset comprising 30 human and 30 animal animations. For every object or sequence, we render image sequences from a single fixed, calibrated camera and use the resulting frames as inputs for evaluation.
\vspace{-2mm}
\paragraph{Evaluation Protocol.}
We perform evaluation on three cases: Compositional 4D, 4D object and 3D scenes. For all methods we report runtime, fidelity, and task-specific structural quality. Runtimes are computed from the average inference time in NVIDIA H200.

For the static 3D scenes, we compare against \textsc{MIDI-3D}~\cite{huang2025midimultiinstancediffusionsingle} and \textsc{PartCrafter}~\cite{lin2025partcrafterstructured3dmesh}. For a fair comparison, we provide \textsc{MIDI-3D} with ground-truth instance masks at inference, a step not required by \textsc{PartCrafter} or our method. For the 4D animation task, we include state-of-the-art generative approaches \textsc{L4GM}~\cite{ren2024l4gmlarge4dgaussian} and \textsc{GVFD}~\cite{zhang2025gaussianvariationfielddiffusion}, the mesh-based \textsc{V2M4}~\cite{chen2025v2m44dmeshanimation}, and a frame-wise \textsc{TripoSG}~\cite{li2025triposghighfidelity3dshape} baseline to ablate the effects of temporal modeling. The token budget is fixed at 512 per part for all 3D methods and 1024 per frame for all mesh-based 4D methods.

Geometric fidelity is measured using the Chamfer Distance (CD)~\cite{barrow77,Chamfer88} and F-Score~\cite{vanrijsbergen1979information} (at a 0.1 threshold), where lower CD and higher F-Score are better. For 3D scenes, we concatenate all parts into a single mesh before evaluation, whereas for 4D sequences, we compute the metrics per-frame and then average. To assess structural quality, we use two Intersection-over-Union (IoU) metrics based on a consistent $64^3$ voxelization. In 3D, we measure part independence via a pairwise IoU, where lower scores indicate better separation between parts. In 4D, we evaluate accuracy with a per-frame IoU against the ground truth, where higher scores are better. Note that for Gaussian-based baselines (\textsc{L4GM}, \textsc{GVFD}), we convert their outputs to point clouds for CD and F-Score evaluation, but omit the IoU metric for lack of reliable watertight meshes. All reported scores are the mean values over the test set.

% Put the table here and the large figure in the next page.

\subsection{Compositional 4D Reconstruction}
Compositional 4D forms our primary experimental setup and is therefore our first task. We provide experimental results using various real and synthetic sequences. Fig.~\ref{fig:temporal_combined} provides the qualitative results on the synthetic, \ie, generated video on the top row and on the CMU Panoptic sequences~\cite{panoptic} in the bottom row. Fig.~\ref{fig:temporal_combined} shows that our method completes the reconstruction capturing the inter-object interactions with surprising accuracy, thanks to the attention mixing strategy where dynamic objects can attend to its history as well as the static objects in the scene. We provide more qualitative/quantitative results for our method, with the impact of attention mixing in \S\ref{sec:ablation_study}. %We provide additional synthetic data experiment in the supplementary Sec.~I. %We construct a synthetic compositional 4D dataset and provide quantitative results on it in the Appendix.

\begin{figure}[h]
  \centering
  \setlength{\tabcolsep}{2pt} % Adjust horizontal "small gap" between frames
  
  % ==================== TABLE START = "==================="
  \begin{tabular}{@{}cccc@{}}
    % -------- Headers --------
    \scriptsize{w/o Mixing} & \scriptsize{w/ Mixing} & \scriptsize{w/o Mixing} & \scriptsize{w/ Mixing} \\
    
    % -------- First Pair of Images (ian sequence) --------
    \raisebox{-.5\height}{\includegraphics[width=0.24\columnwidth]{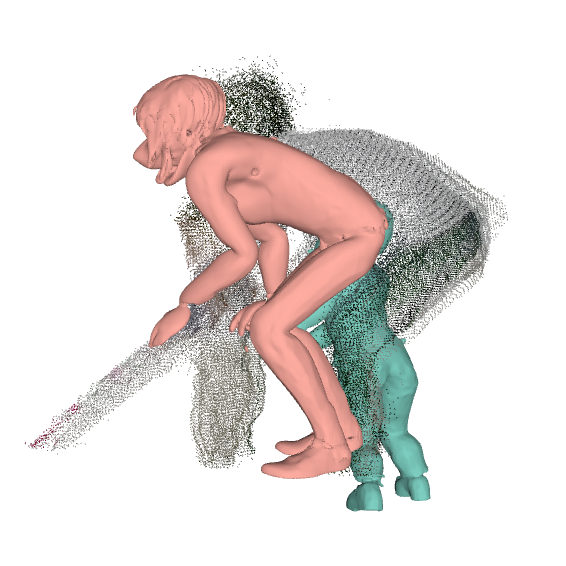}} &
    \raisebox{-.5\height}{\includegraphics[width=0.24\columnwidth]{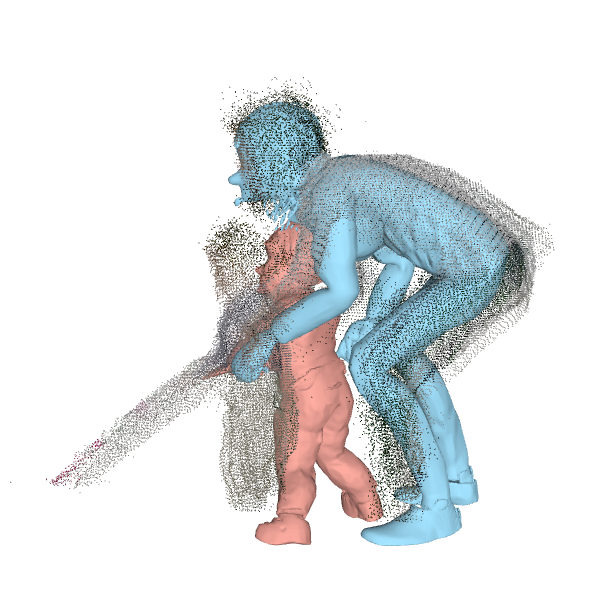}} &
    
    % -------- Second Pair of Images (office sequence) --------
    \raisebox{-.5\height}{\includegraphics[width=0.24\columnwidth]{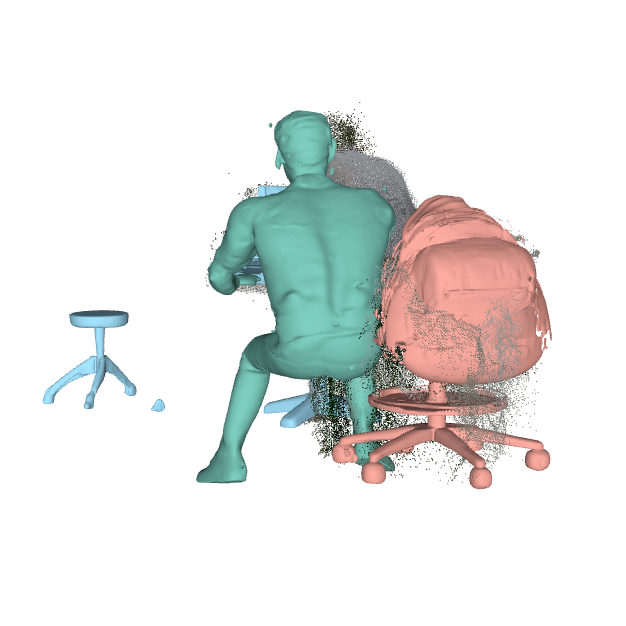}} &
    \raisebox{-.5\height}{\includegraphics[width=0.24\columnwidth]{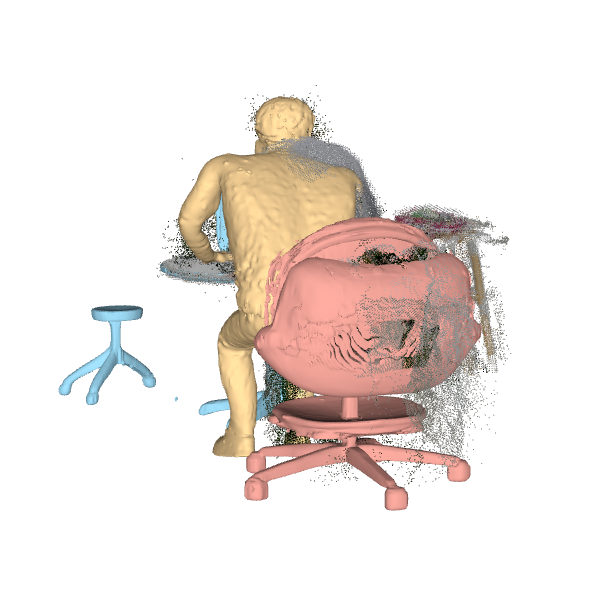}} \\
    
  \end{tabular}
  \caption{Visualizations with and without our Attention Mixing strategy. Results are for \texttt{160401\_ian3} at frame 1180 (starting frame: 1100) and \texttt{170915\_office1} at frame 670 (starting frame: 590). Gray points denote ground truth.}
  % ==================== TABLE END ====================
  \label{fig:panoptic_quant}
\end{figure}

\vspace{-1mm}

\subsection{Single Object 4D}
\begin{figure}[t]
  \centering
  % --- MODIFICATIONS ---
  % Zero horizontal spacing between images
  \setlength{\tabcolsep}{0pt} 
  % Remove default extra padding in rows
  \renewcommand{\arraystretch}{0} 

  % ==================== TABLE START ====================
  \begin{tabular}{@{}cccccc@{}}
    \small{Input} &
    \textbf{\small{Ours}} &
    \small{TripoSG} &
    \small{V2M4} &
    \small{GVFD} &
    \small{L4GM} \\[3pt] % Keep a small gap between titles and images

% -------- Ninja, Time 1 --------
\includegraphics[width=0.16\columnwidth, trim=0 0 0 0, clip]{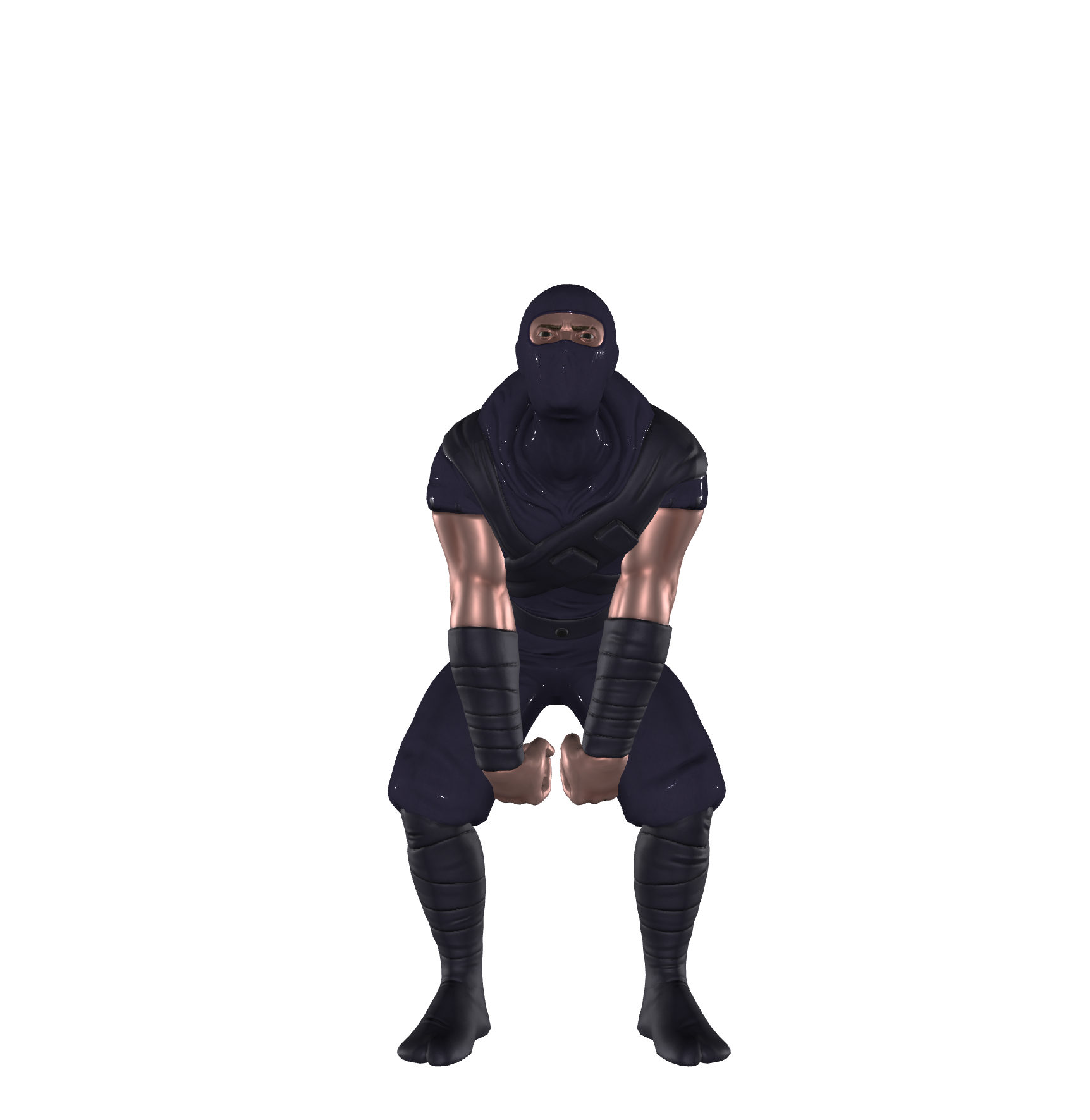} &
\includegraphics[width=0.16\columnwidth, trim=0 0 0 0, clip]{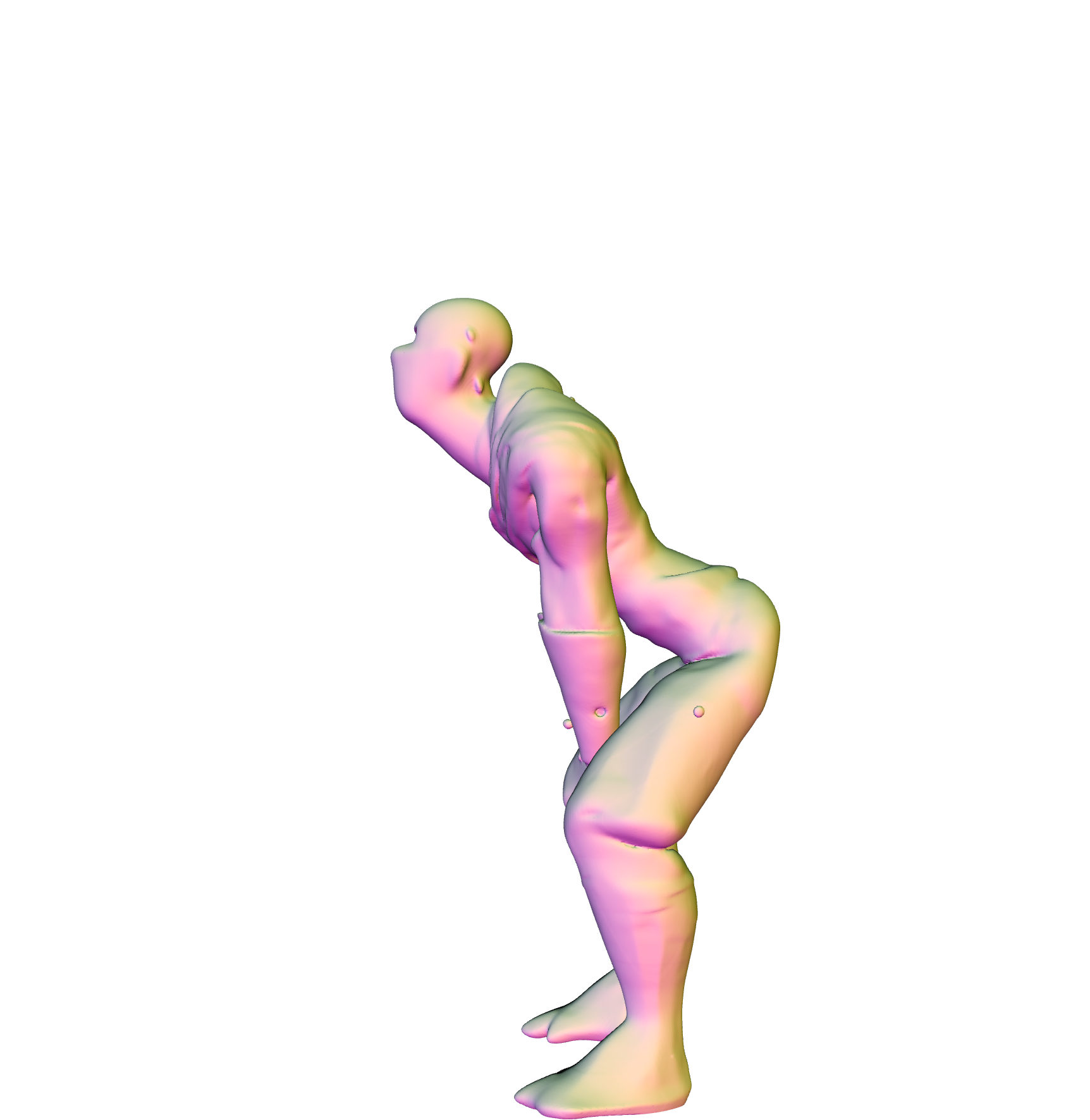} &
\includegraphics[width=0.16\columnwidth, trim=0 0 0 0, clip]{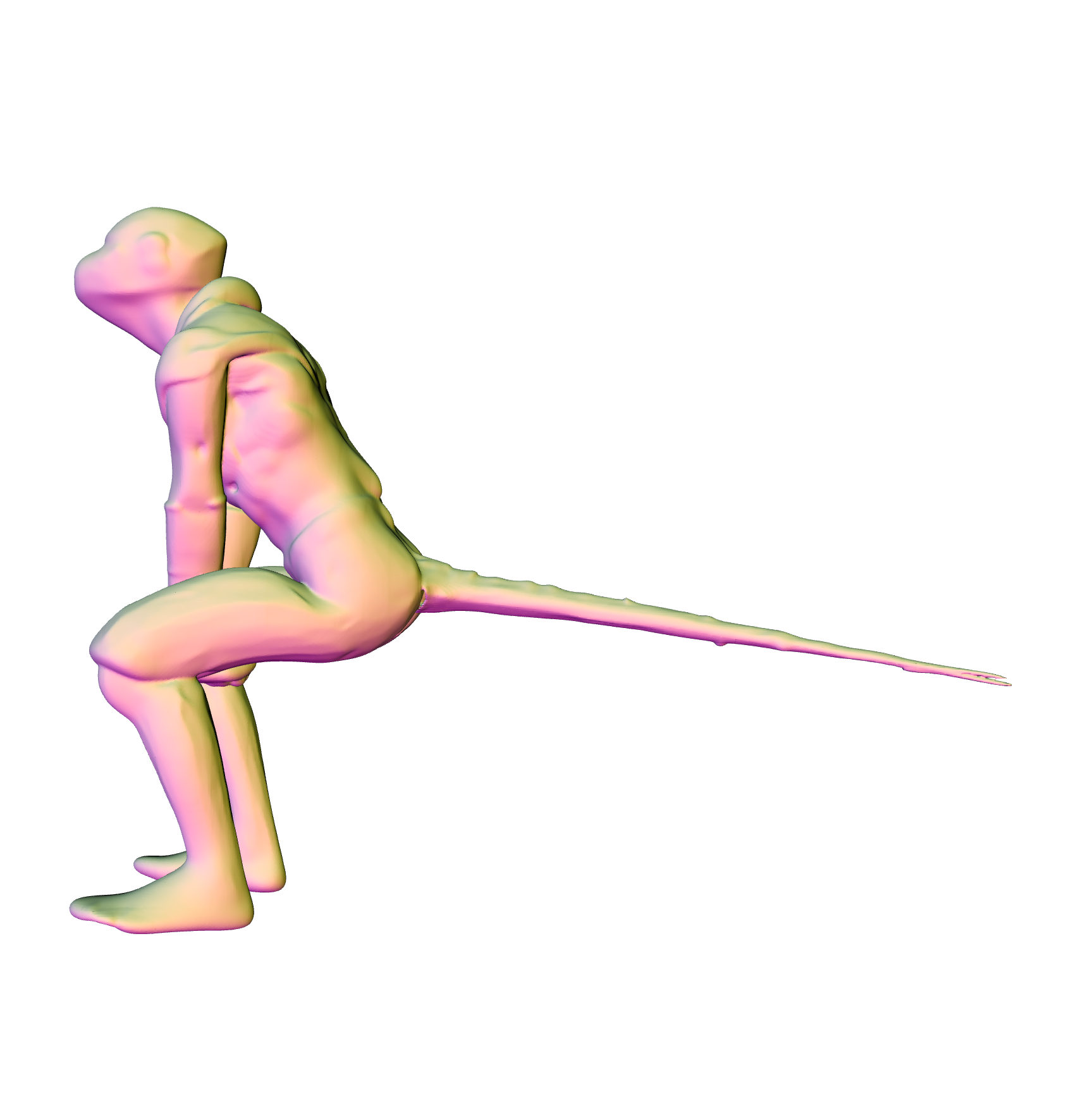} &
\includegraphics[width=0.16\columnwidth, trim=0 0 0 0, clip]{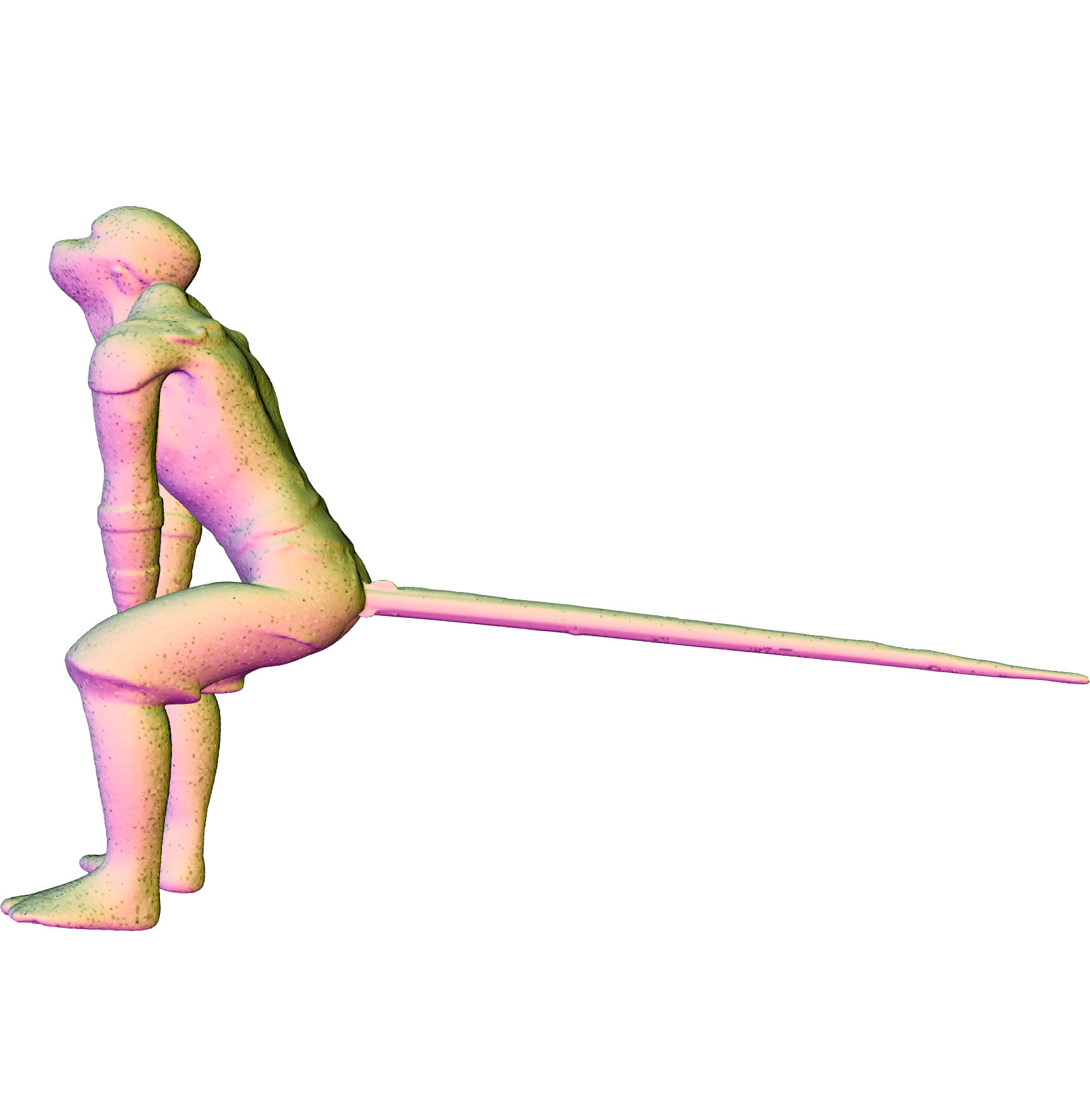} &
\includegraphics[width=0.16\columnwidth, trim=0 0 0 0, clip]{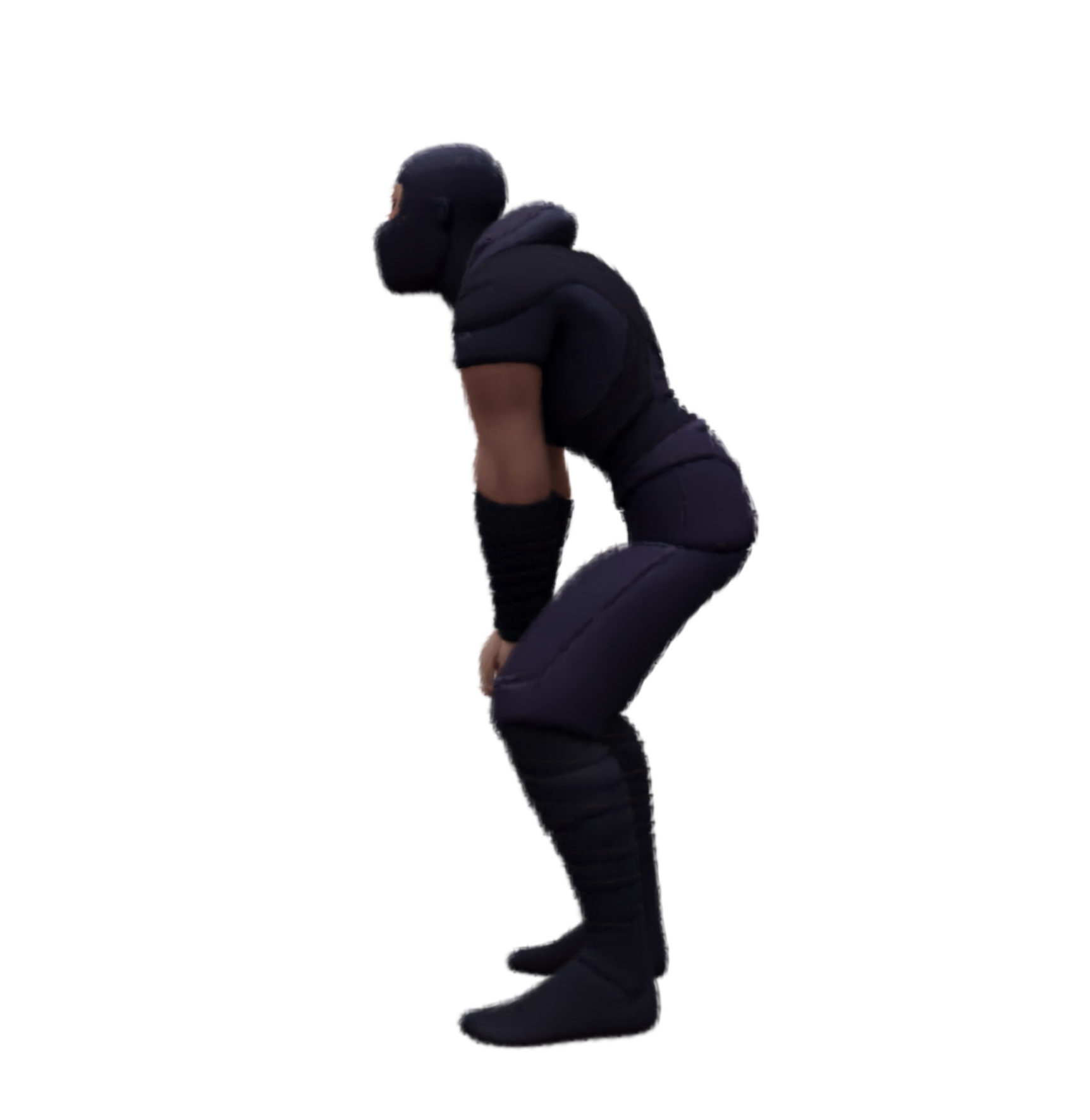} &
\includegraphics[width=0.16\columnwidth, trim=0 0 0 0, clip]{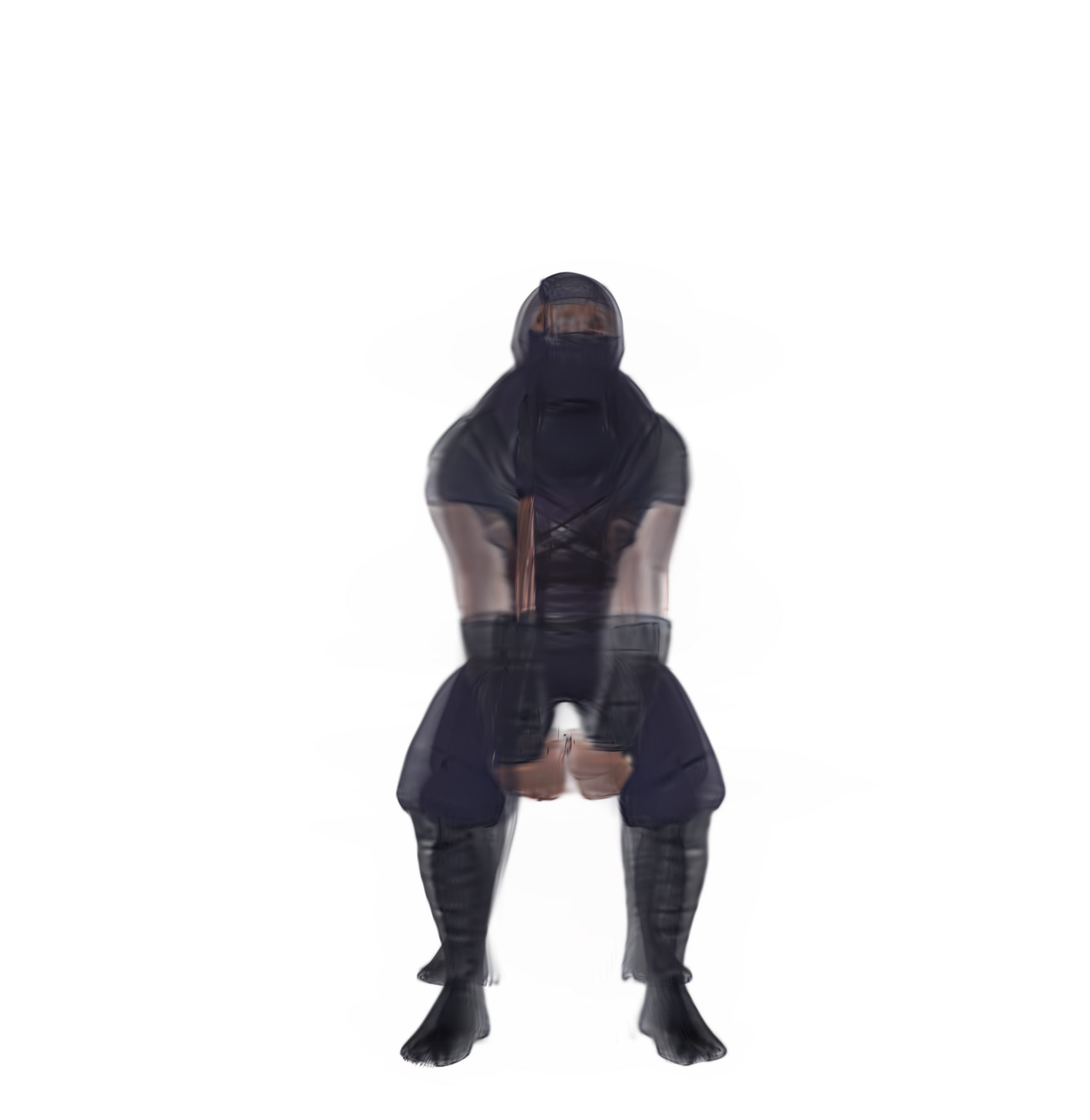} \\[-1.5pt]

% -------- Ninja, Time 2 --------
\includegraphics[width=0.16\columnwidth, trim=0 0 0 0, clip]{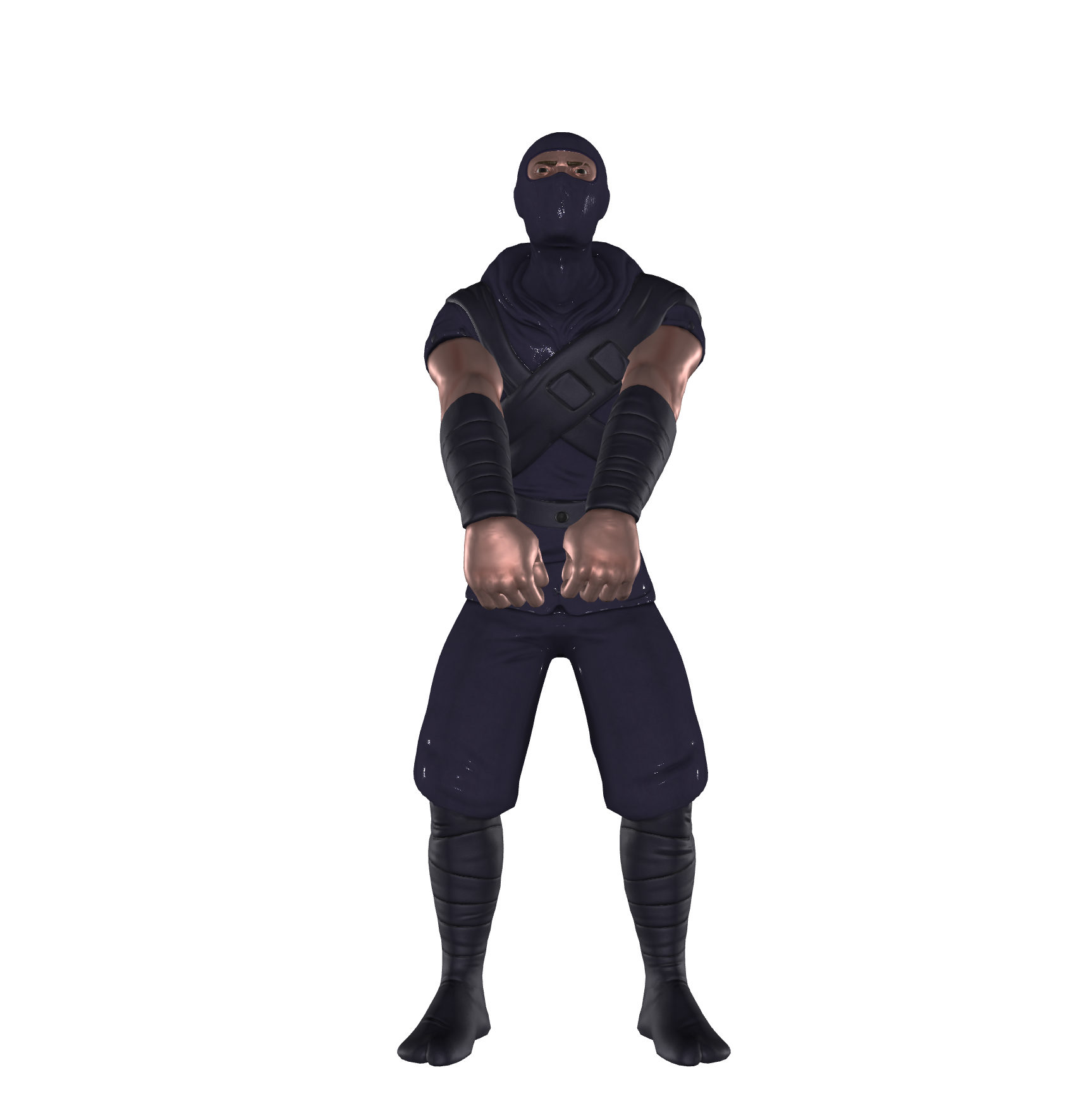} &
\includegraphics[width=0.16\columnwidth, trim=0 0 0 0, clip]{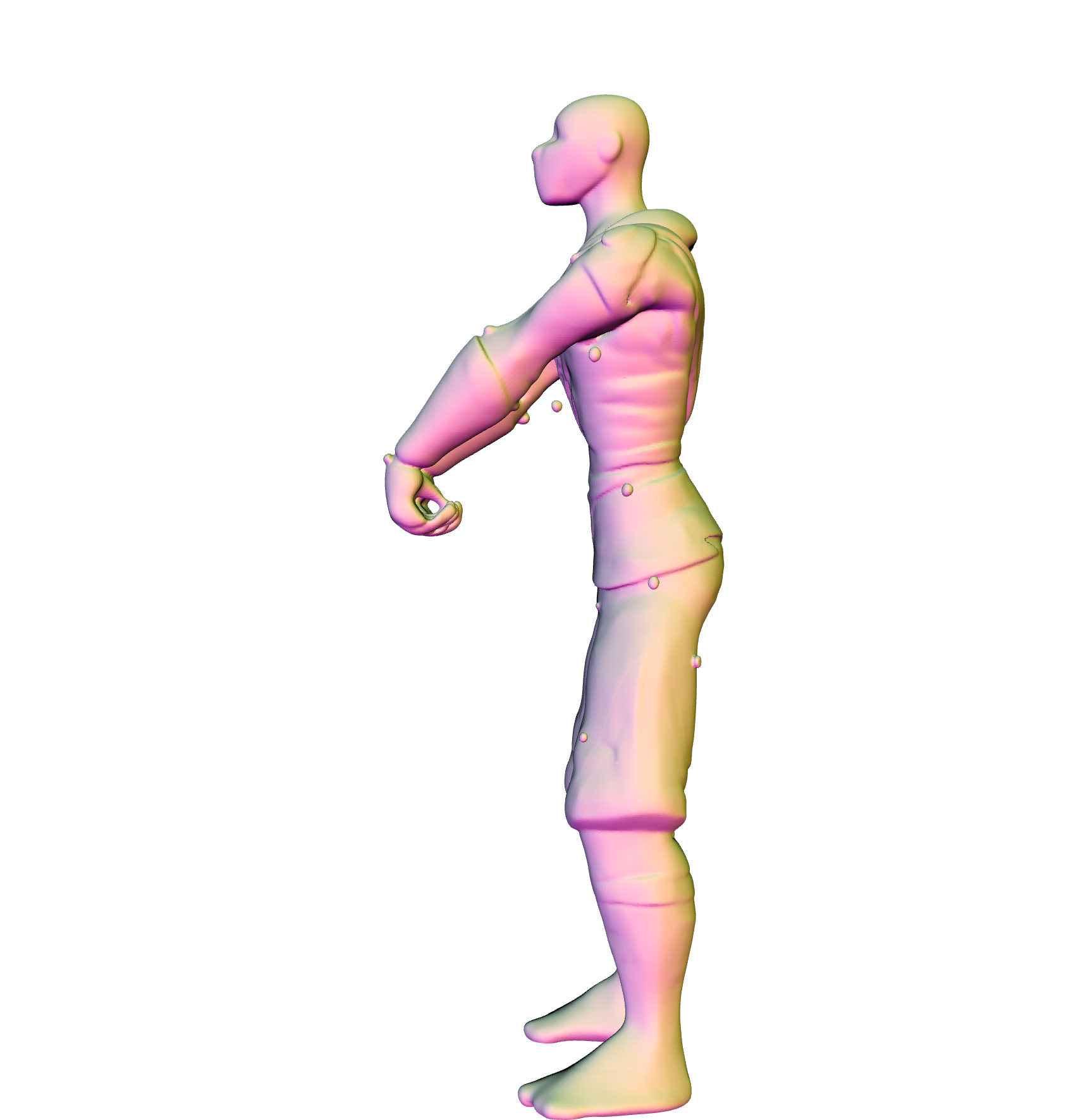} &
\includegraphics[width=0.16\columnwidth, trim=0 0 0 0, clip]{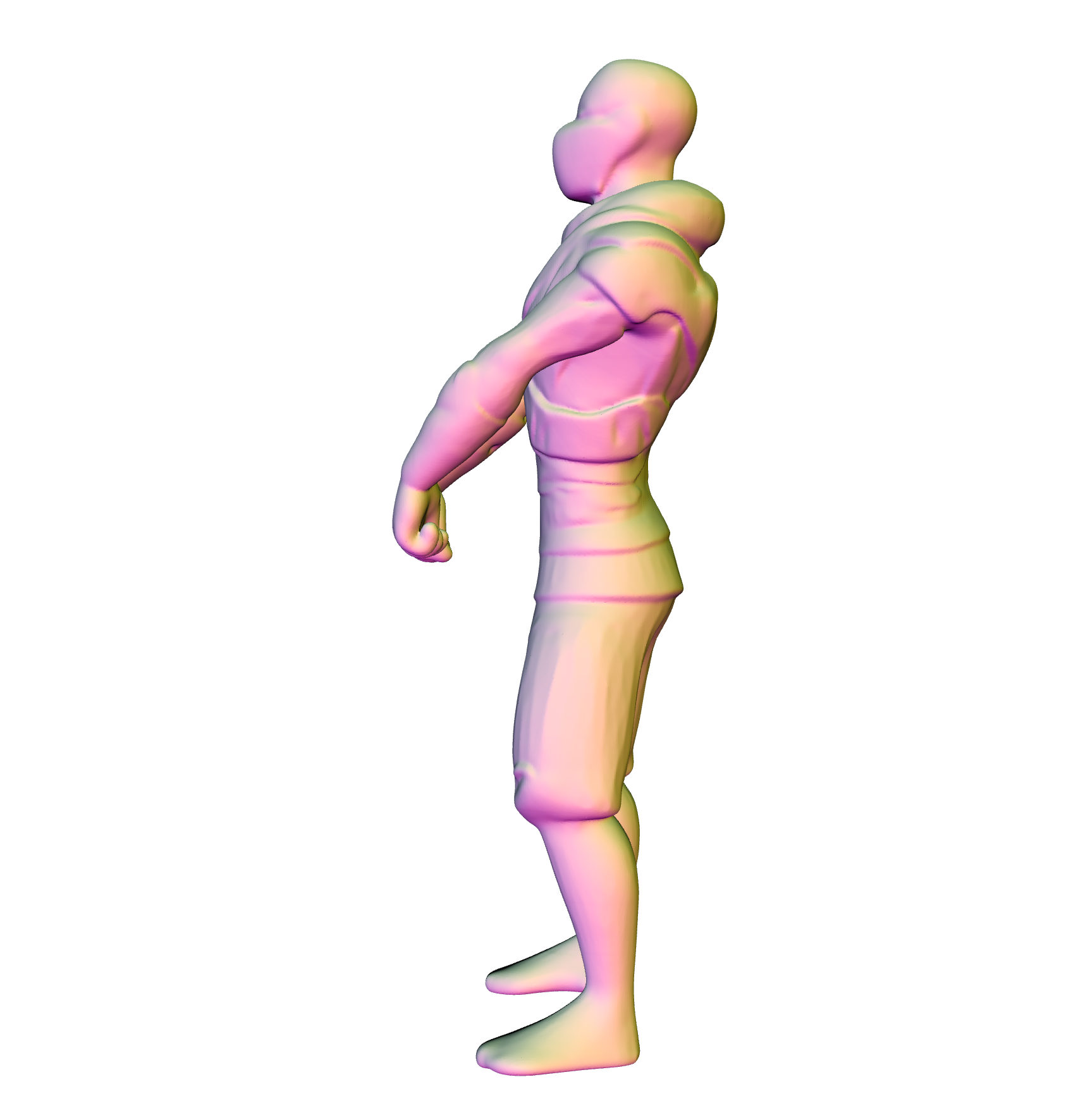} &
\includegraphics[width=0.16\columnwidth, trim=0 0 0 0, clip]{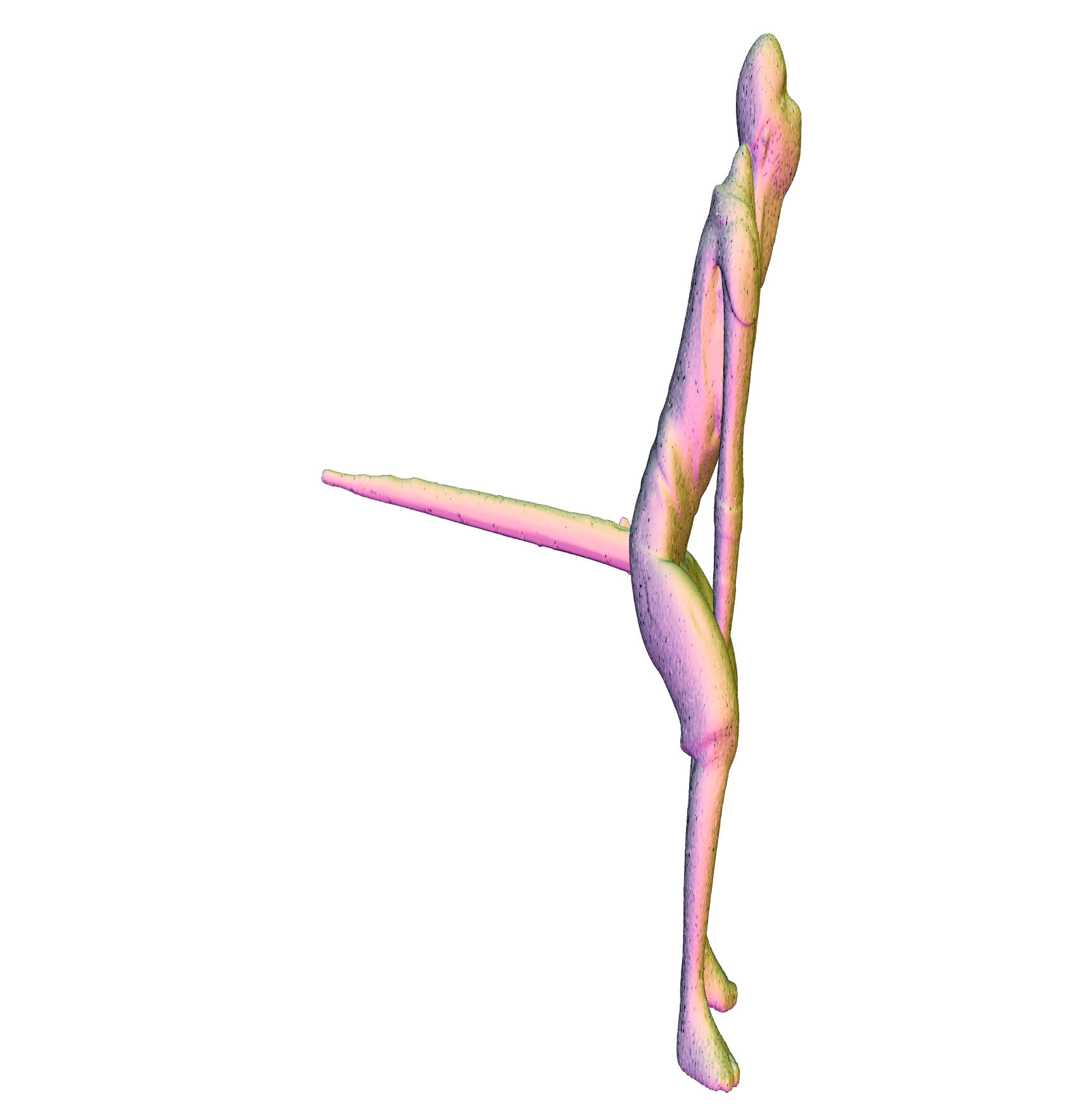} &
\includegraphics[width=0.16\columnwidth, trim=0 0 0 0, clip]{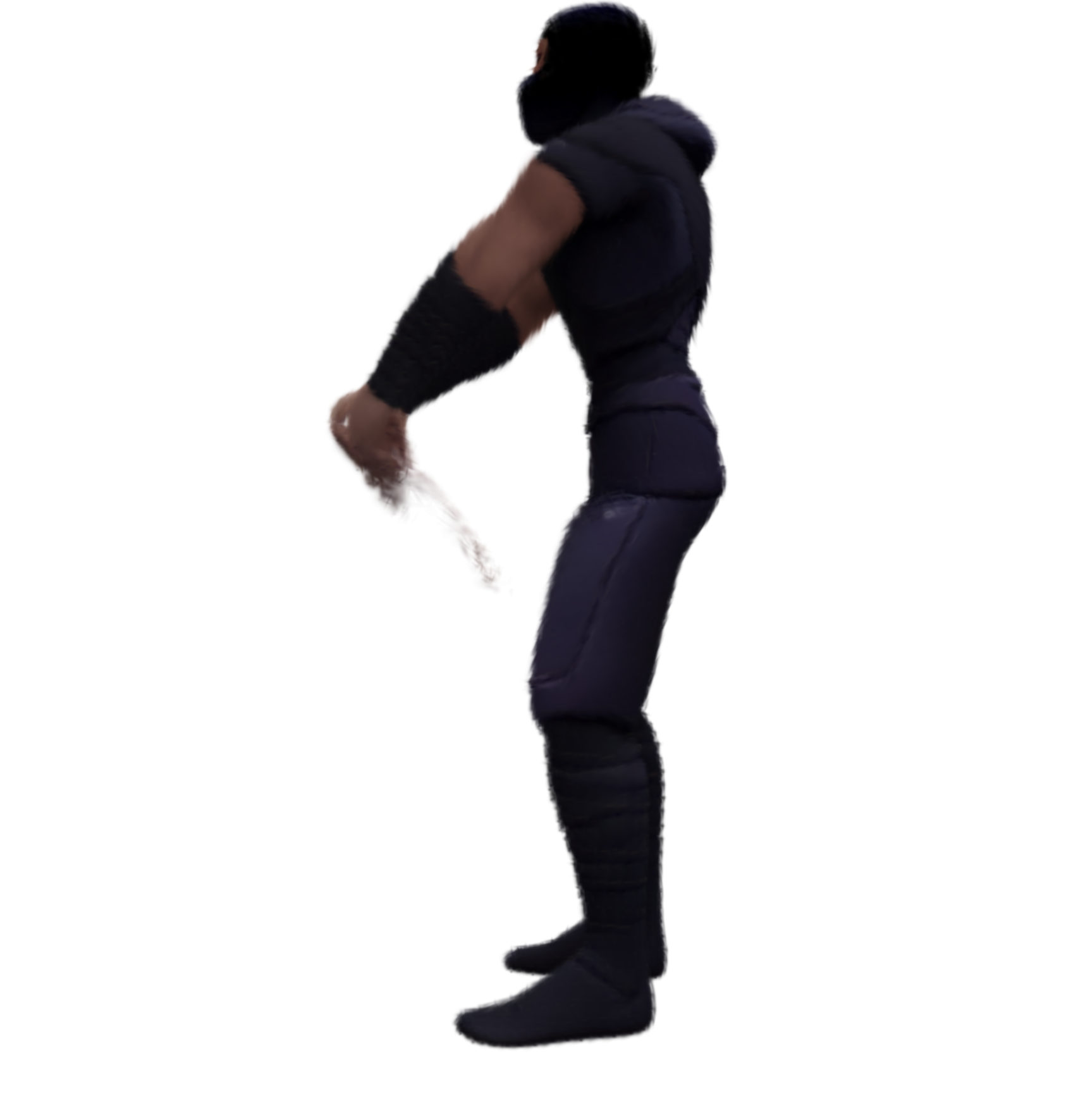} &
\includegraphics[width=0.16\columnwidth, trim=0 0 0 0, clip]{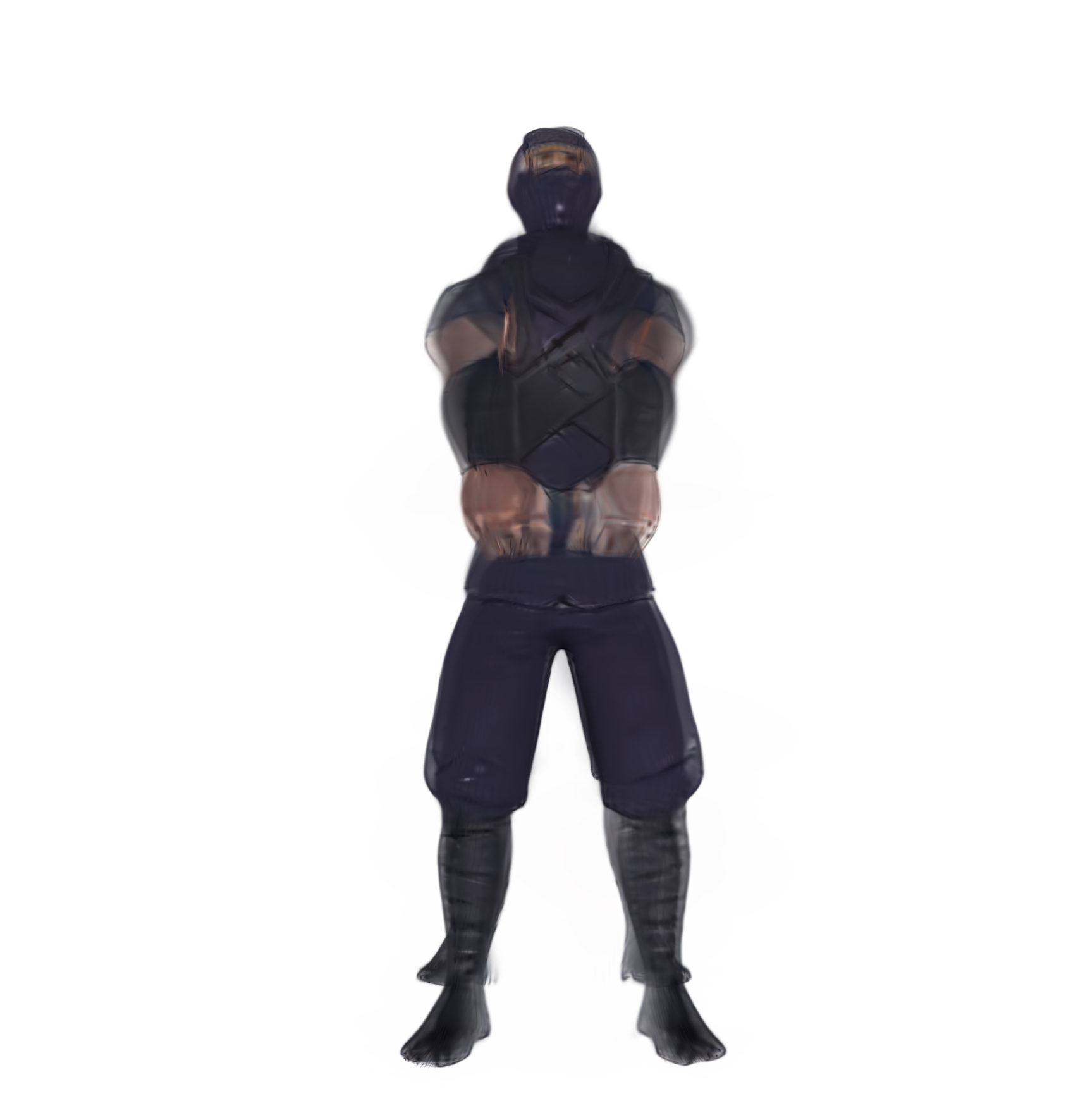} \\[-1.5pt]

% -------- Amy, Time 1 --------
\includegraphics[width=0.16\columnwidth, trim=0 0 0 0, clip]{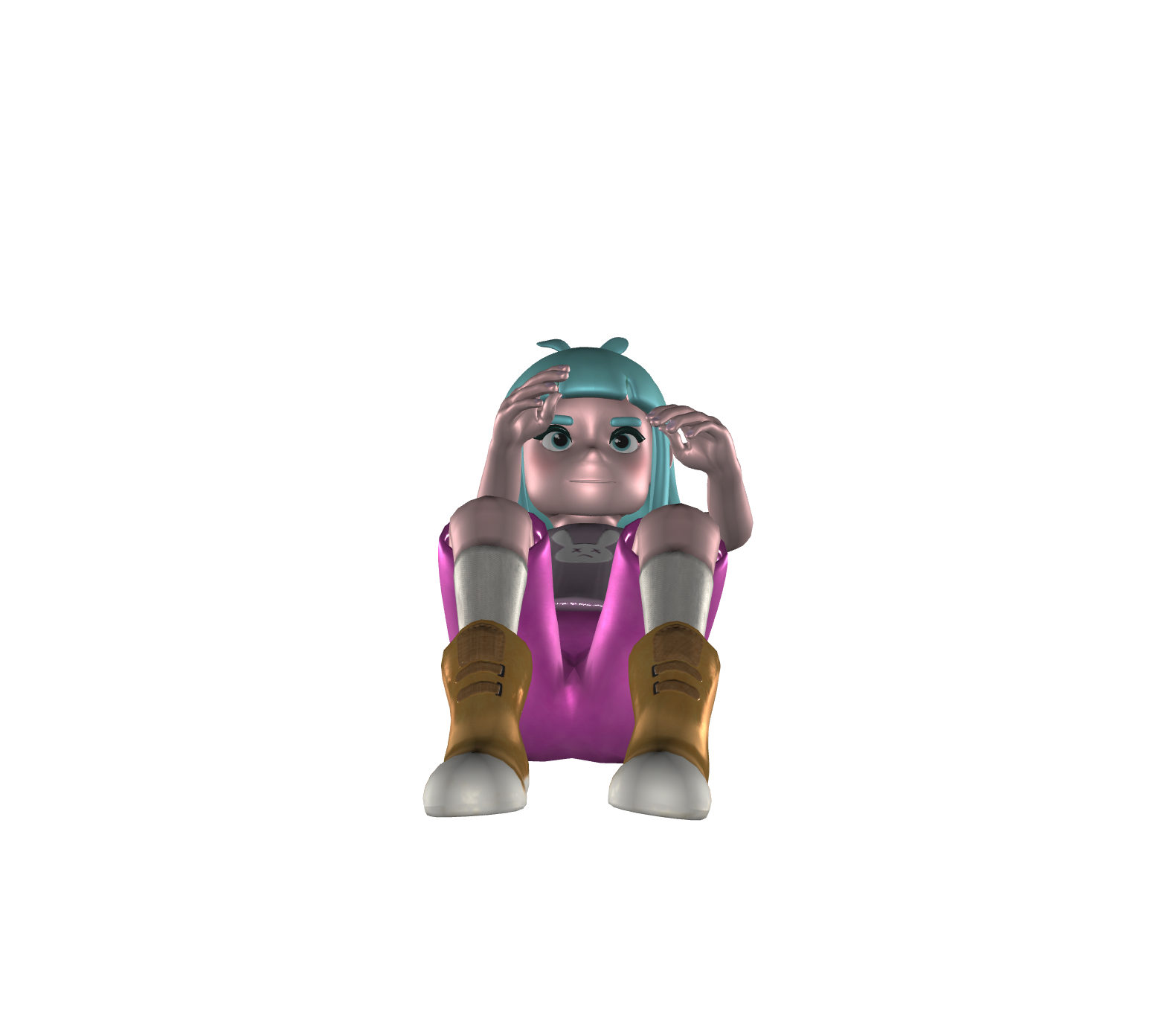} &
\includegraphics[width=0.16\columnwidth, trim=0 0 0 0, clip]{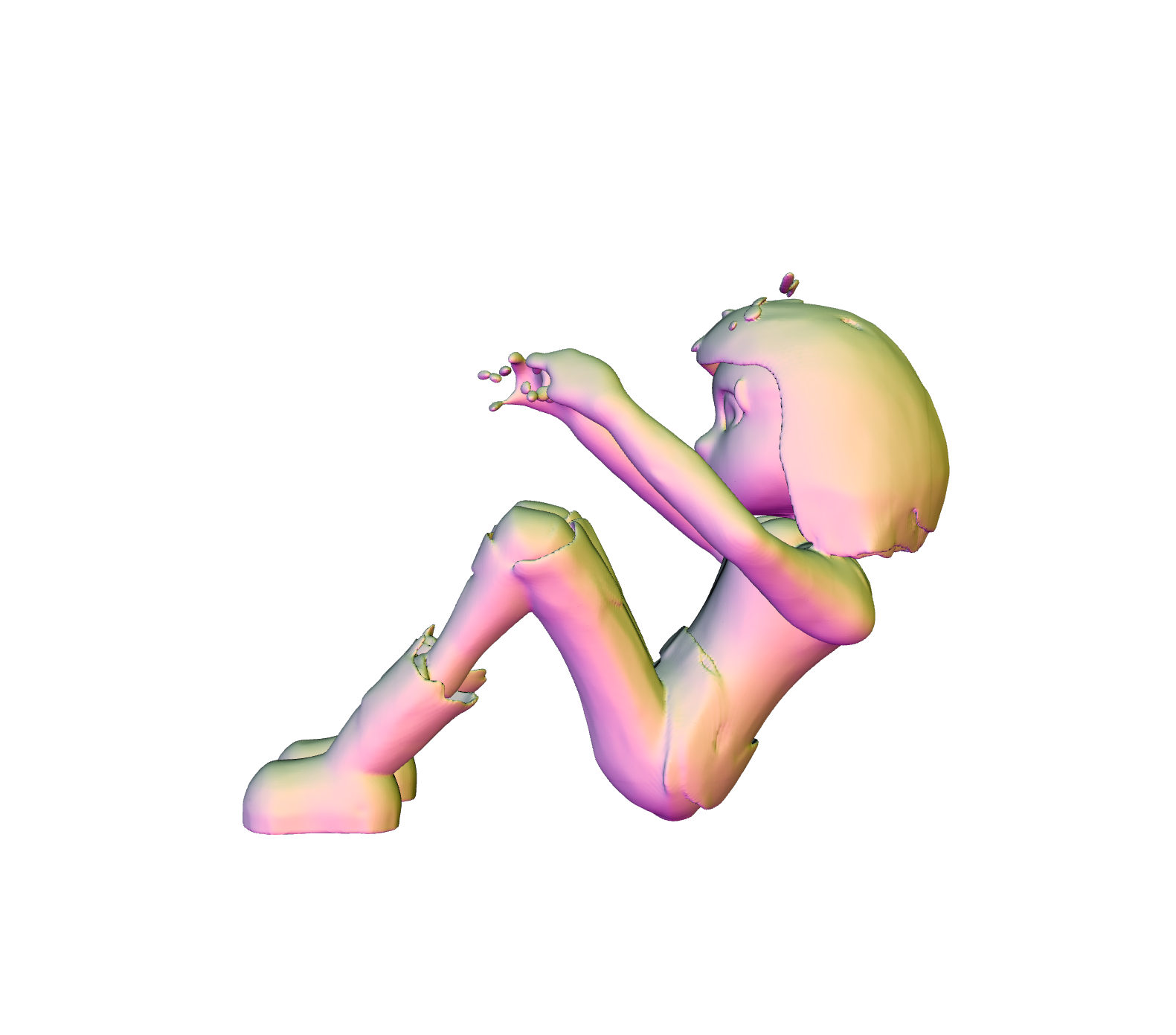} &
\includegraphics[width=0.16\columnwidth, trim=0 0 0 0, clip]{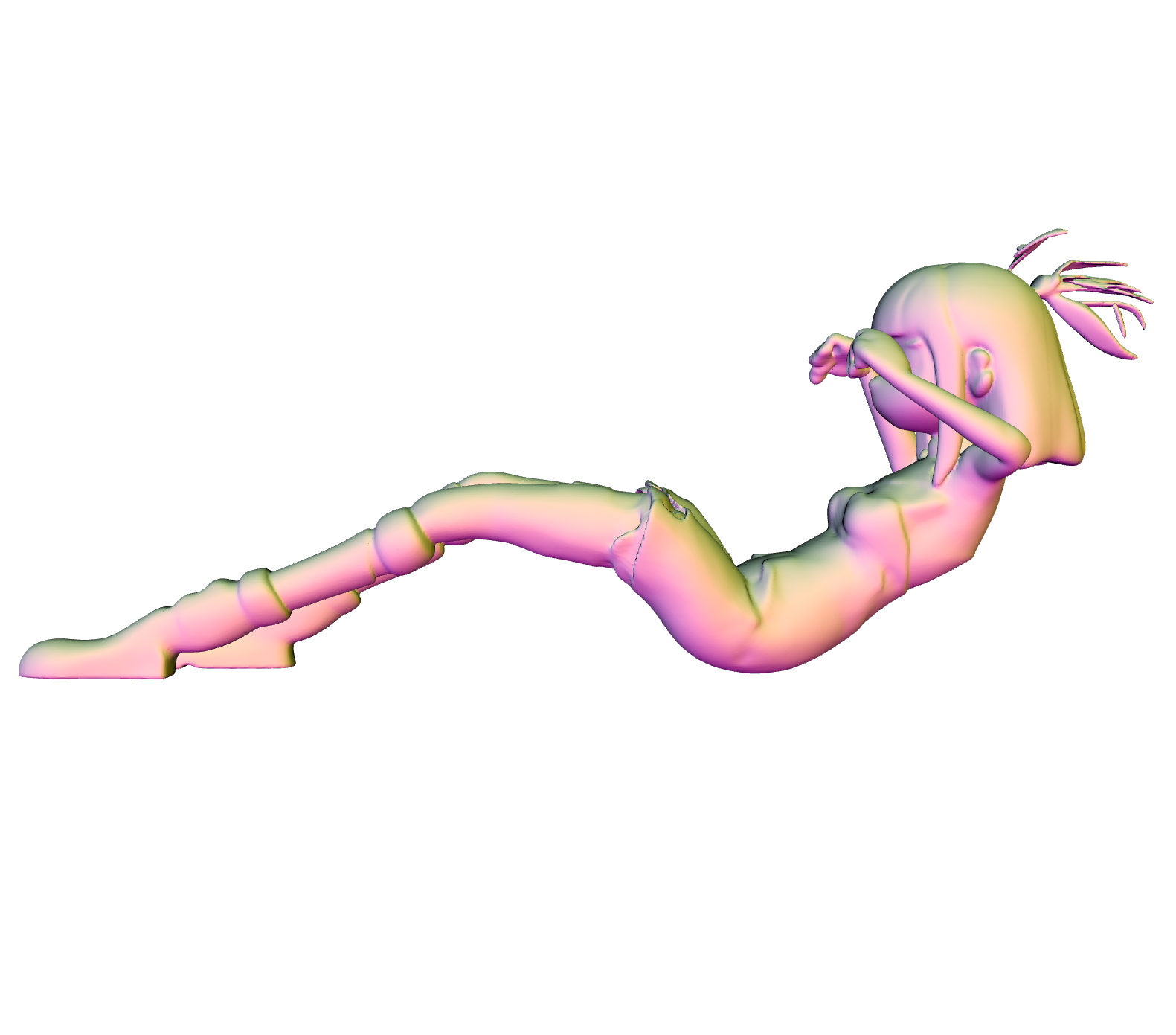} &
\includegraphics[width=0.16\columnwidth, trim=0 0 0 0, clip]{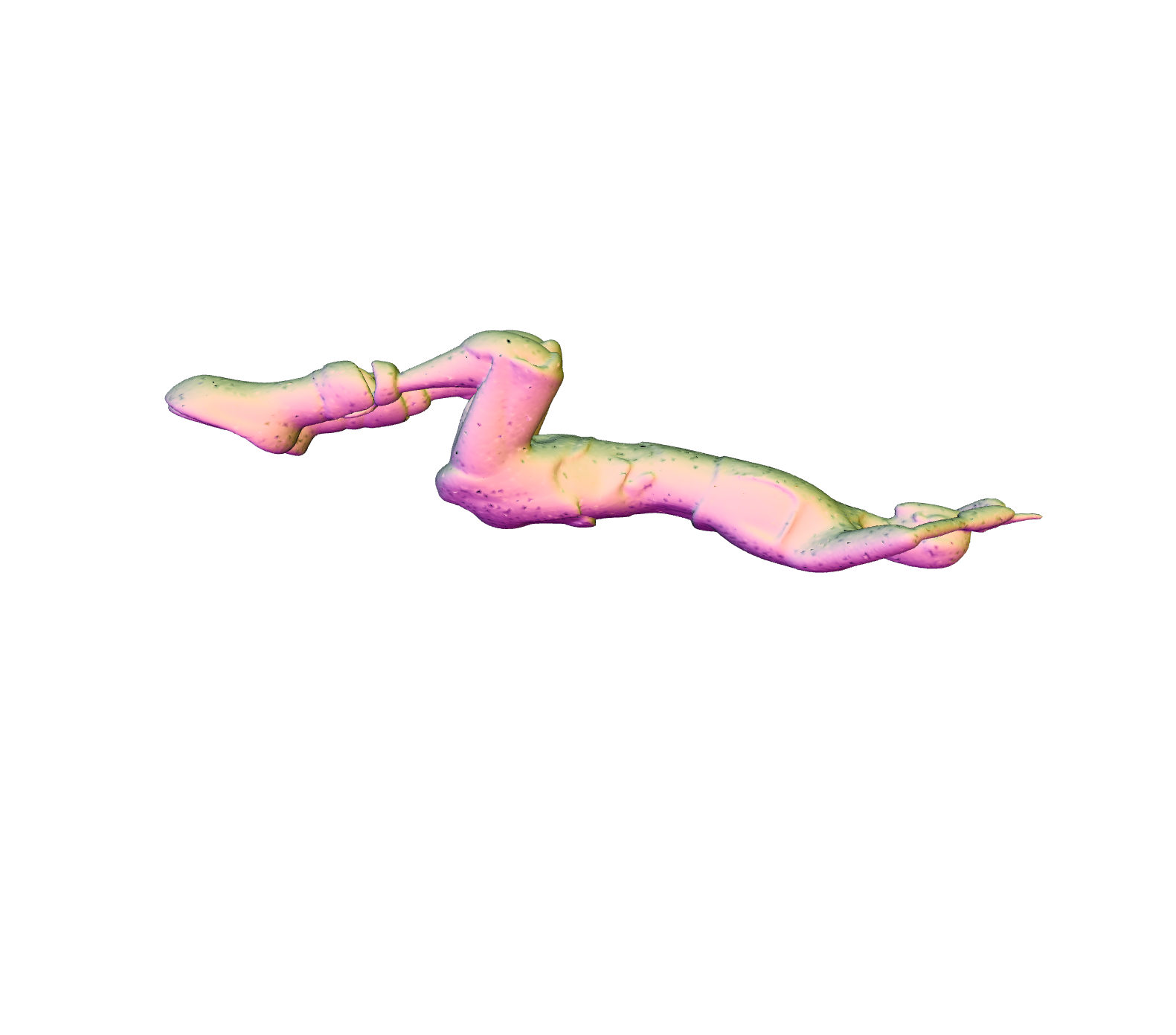} &
\includegraphics[width=0.16\columnwidth, trim=0 0 0 0, clip]{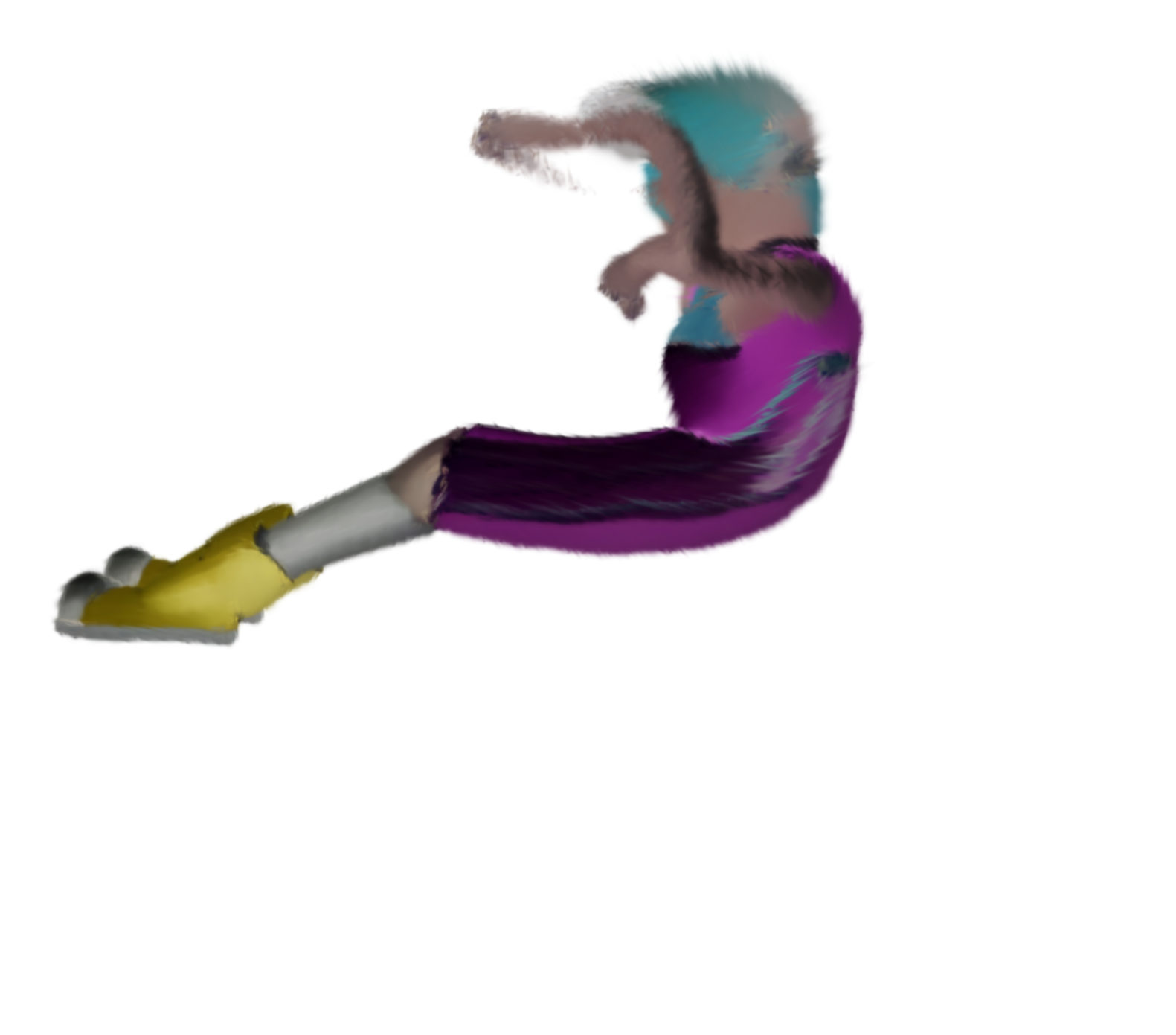} &
\includegraphics[width=0.16\columnwidth, trim=0 0 0 0, clip]{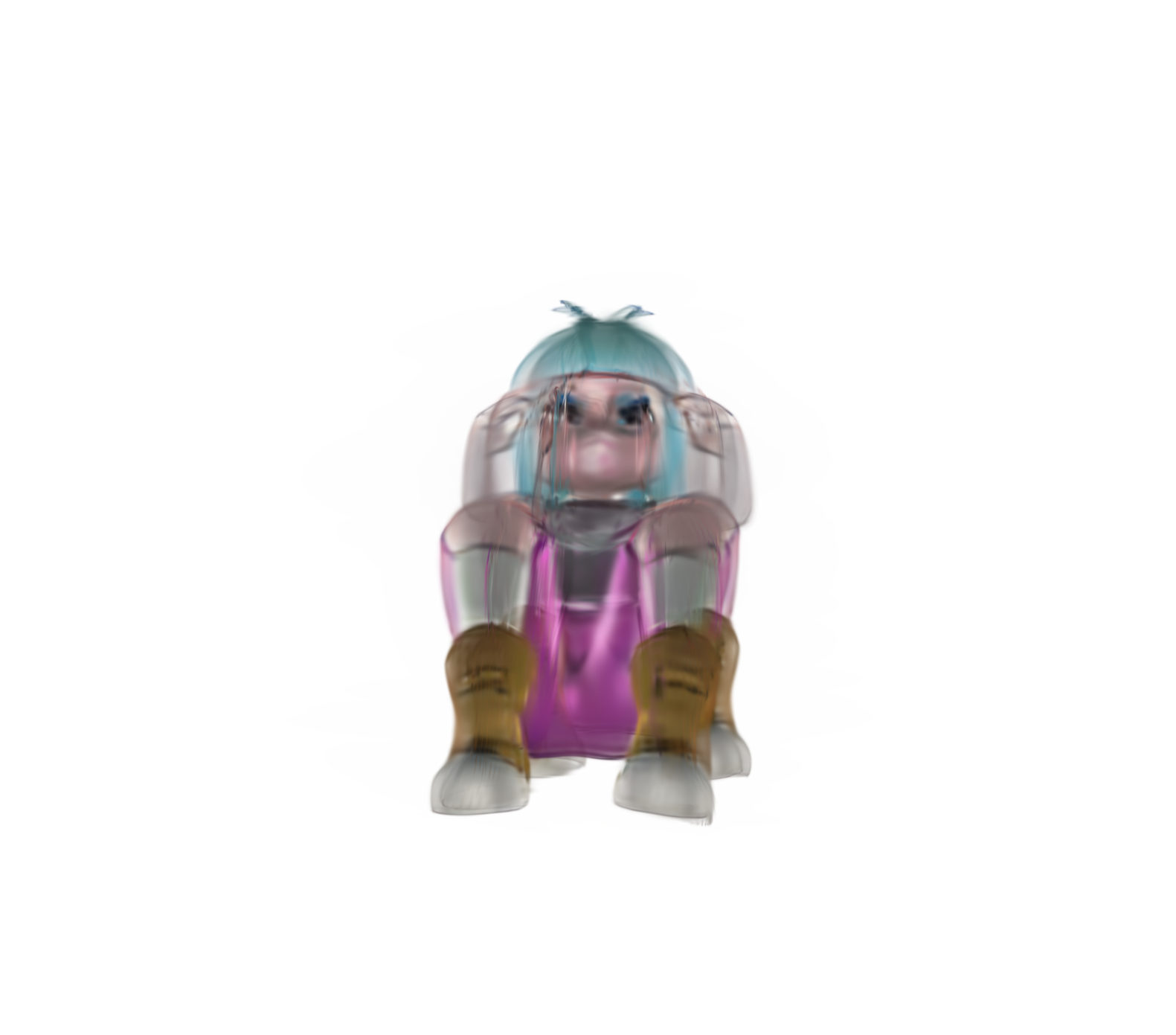} \\[-1.5pt]

% -------- Amy, Time 2 --------
\includegraphics[width=0.16\columnwidth, trim=0 0 0 0, clip]{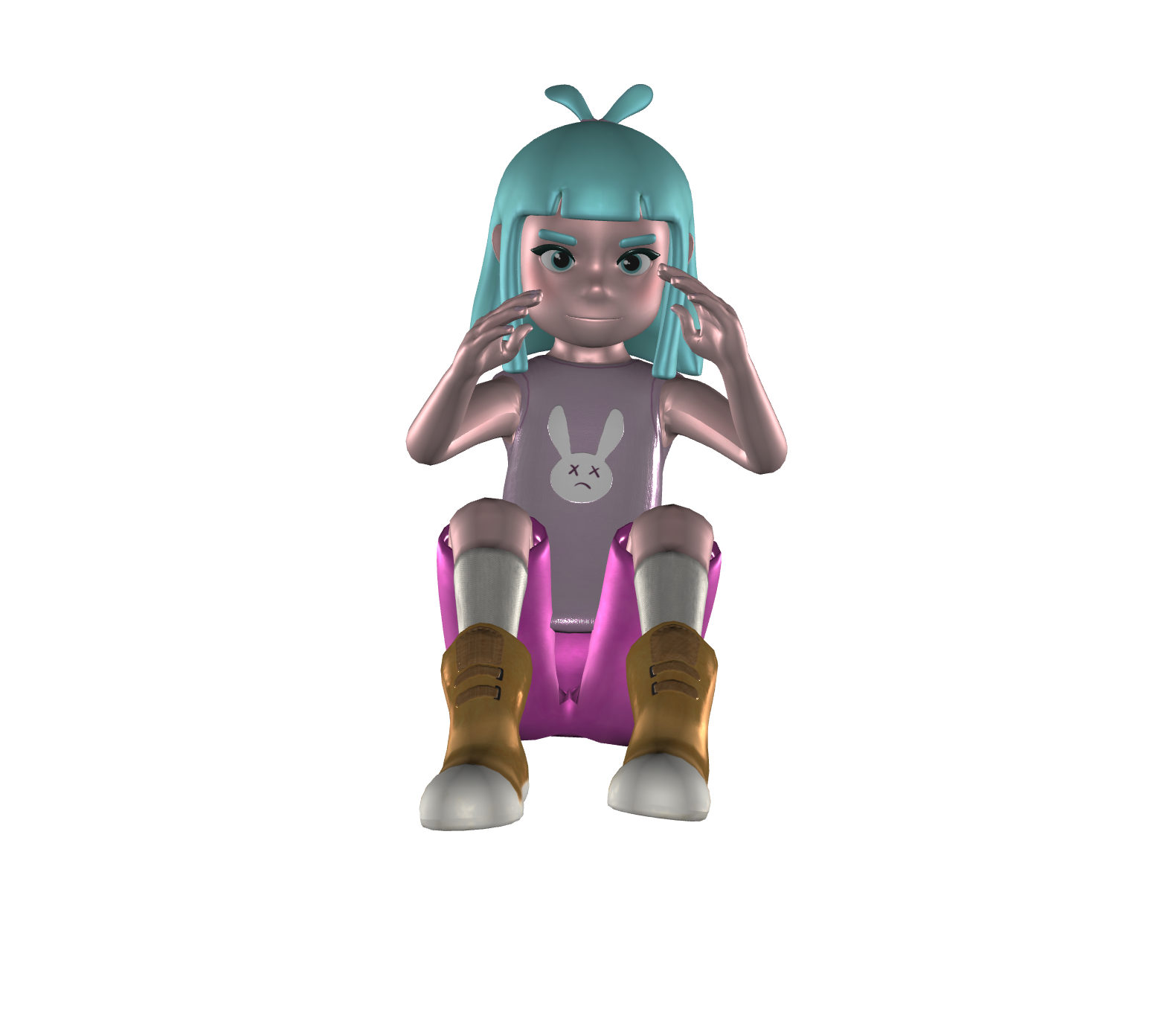} &
\includegraphics[width=0.16\columnwidth, trim=0 0 0 0, clip]{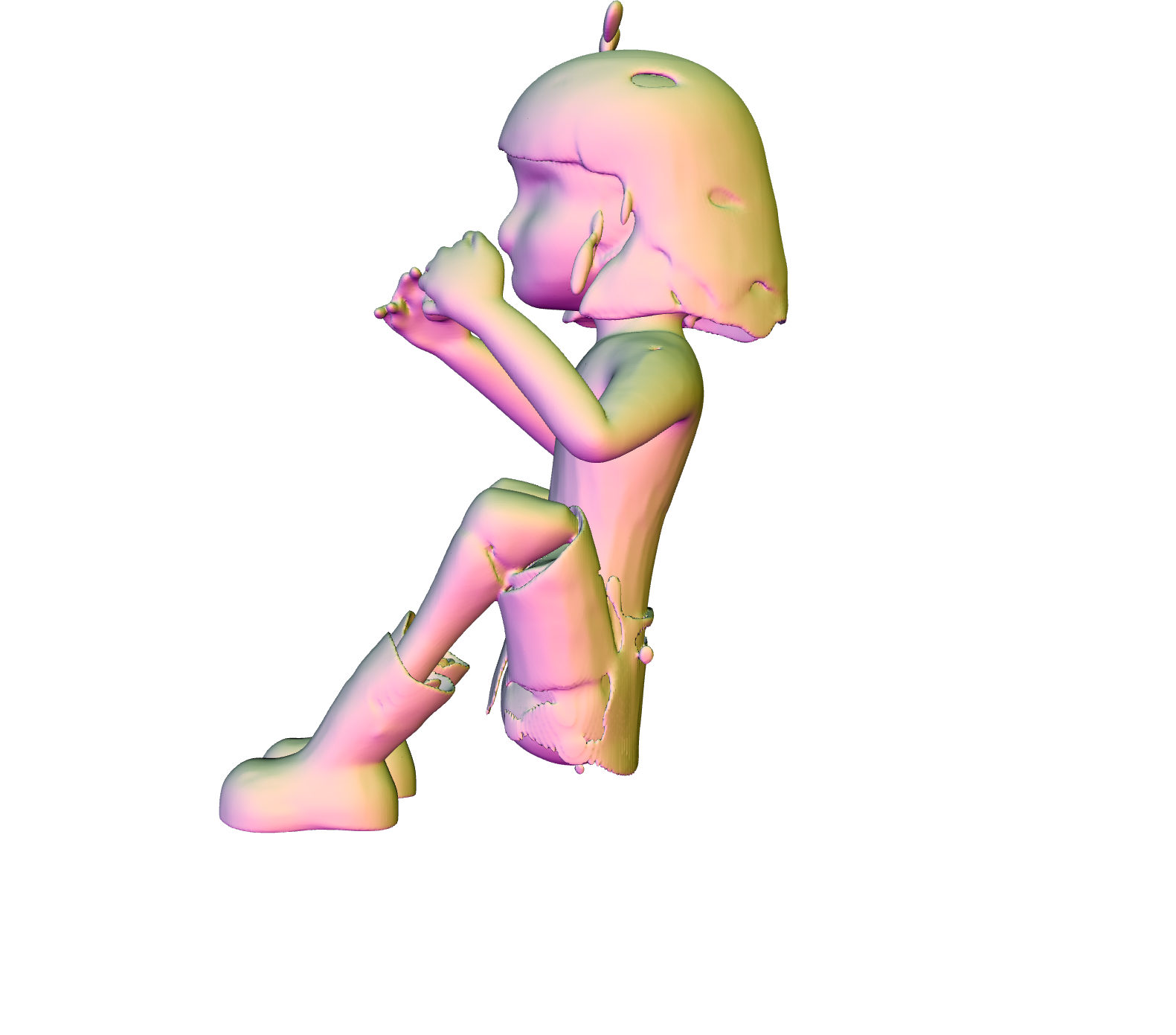} &
\includegraphics[width=0.16\columnwidth, trim=0 0 0 0, clip]{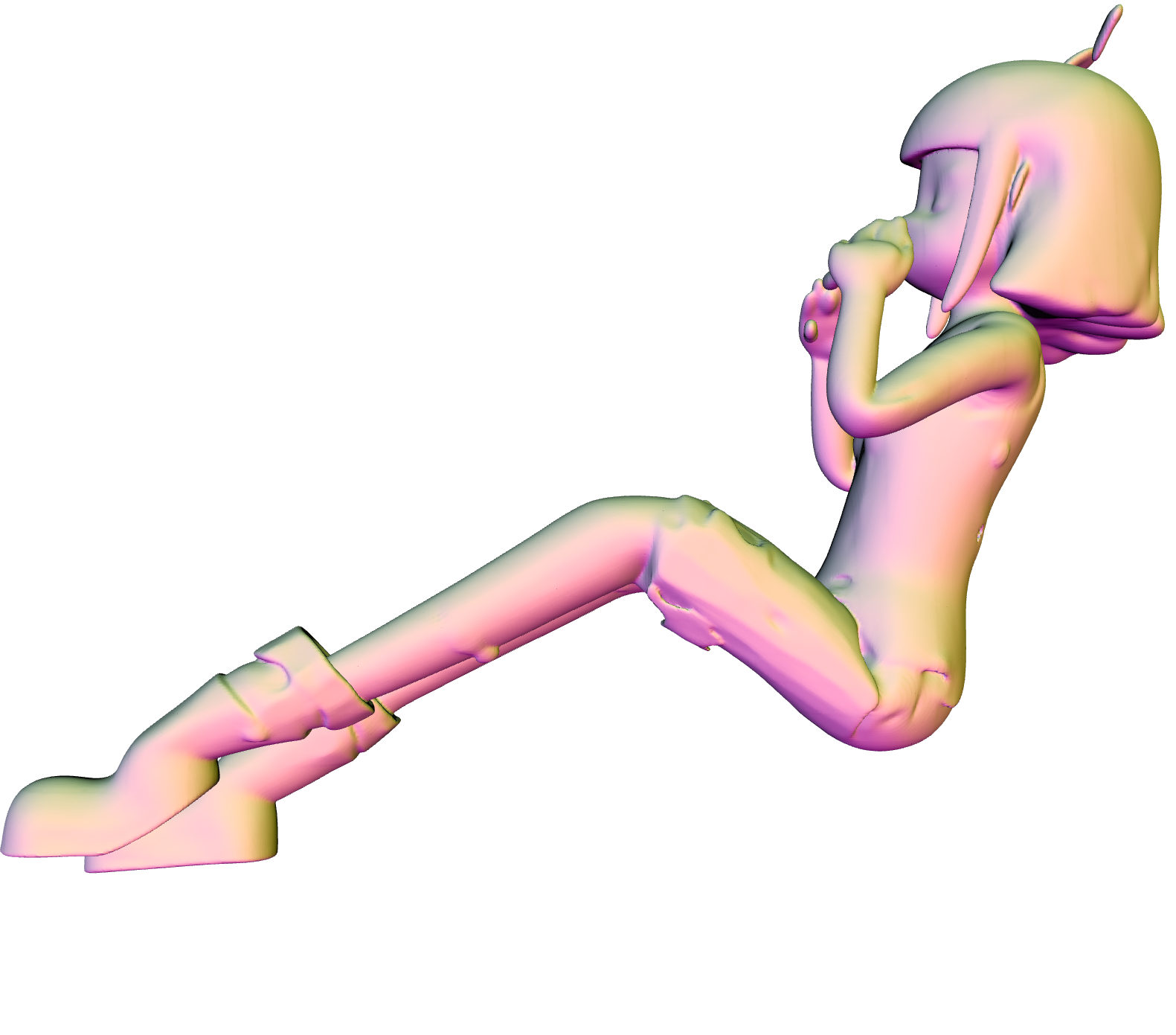} &
\includegraphics[width=0.16\columnwidth, trim=0 0 0 0, clip]{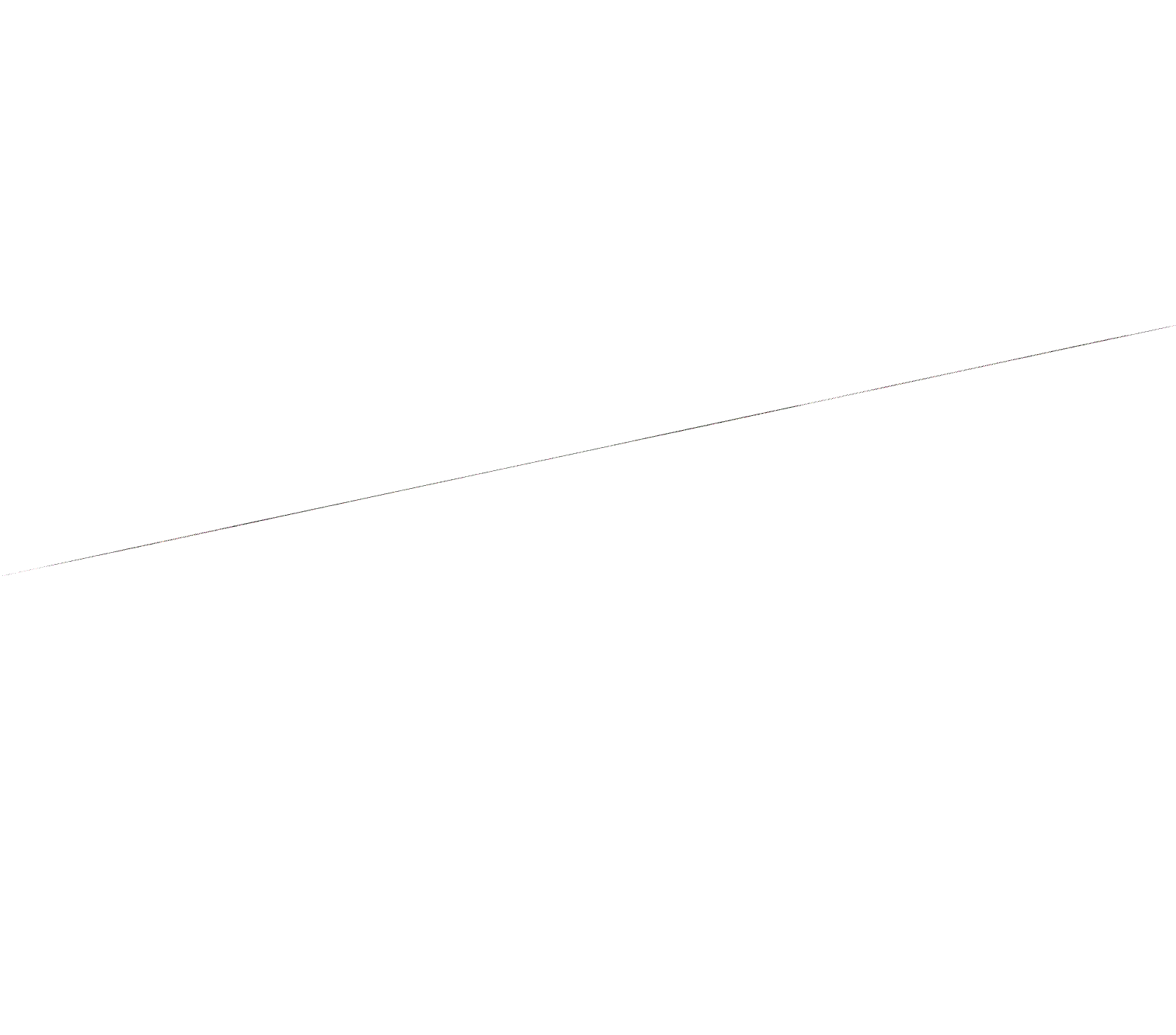} &
\includegraphics[width=0.16\columnwidth, trim=0 0 0 0, clip]{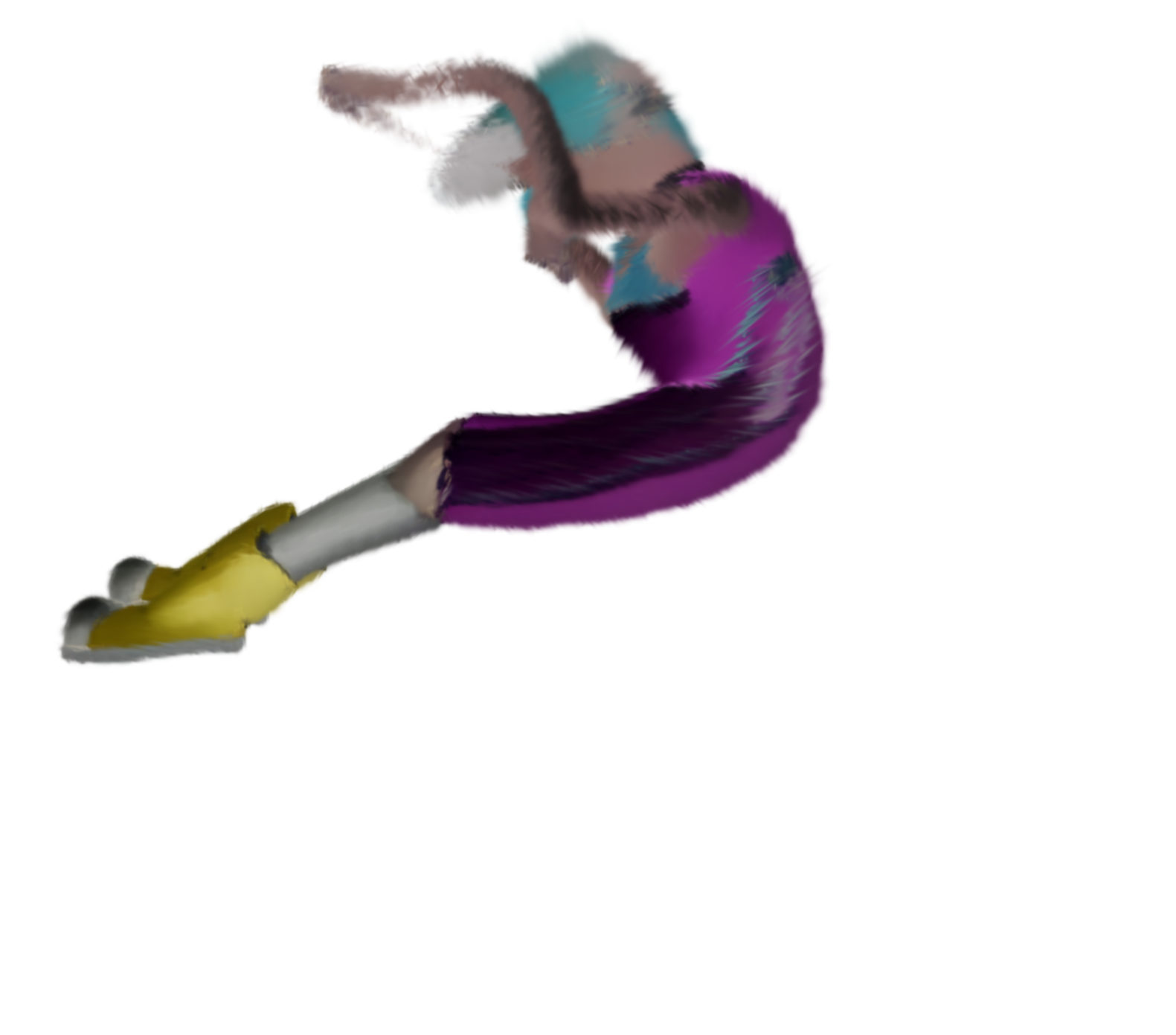} &
\includegraphics[width=0.16\columnwidth, trim=0 0 0 0, clip]{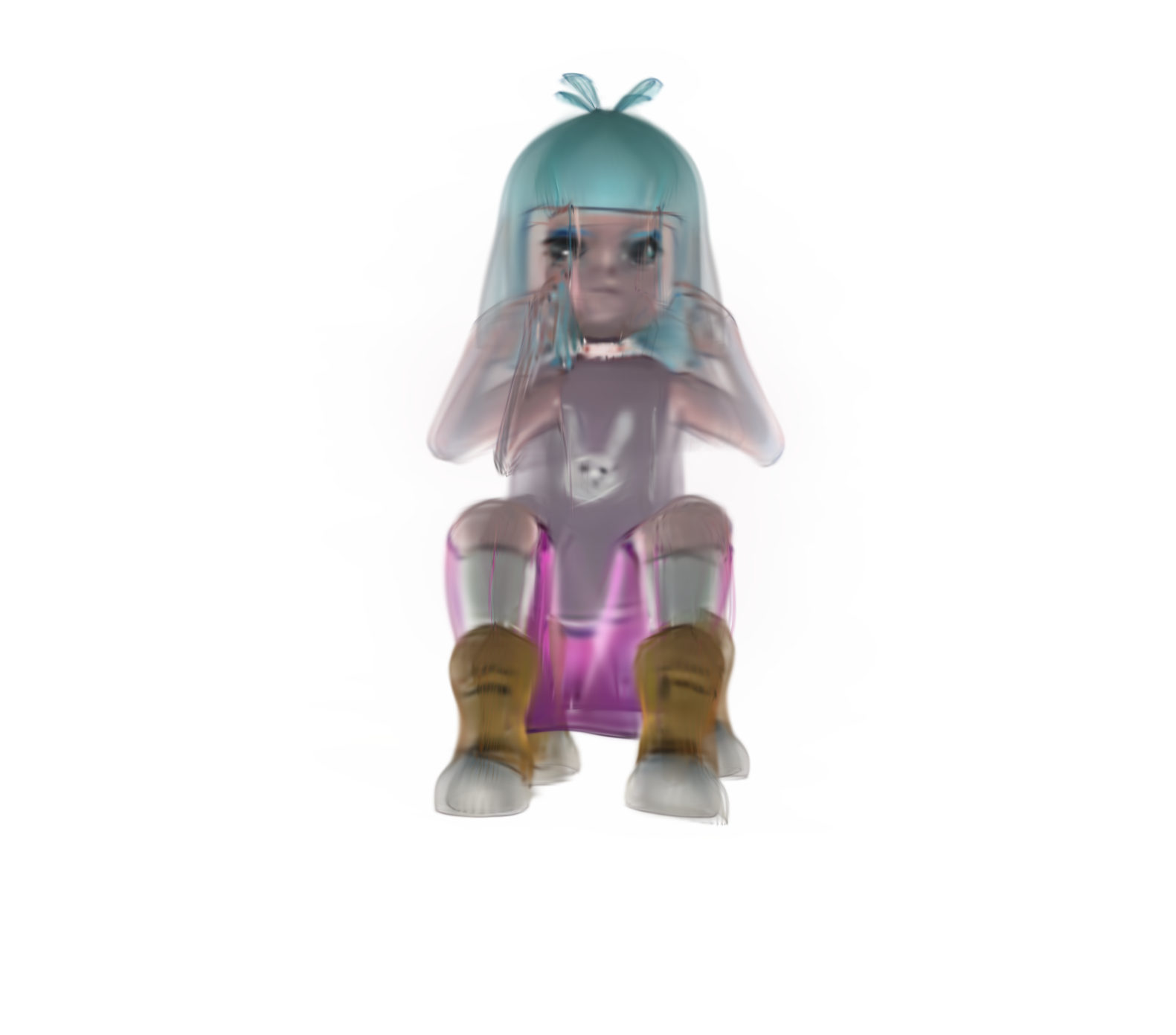} \\

  \end{tabular}
  % ==================== TABLE END ====================

  \vspace{-3pt}
  \caption{
  Qualitative 4D generation comparisons for two subjects (top two rows: Ninja, bottom two rows: Amy) at two time steps. The first column shows the input frames, and subsequent columns show a fixed pose rendered view from each method. V2M4 fails in a few samples, \eg, the last row input.}
  
  \label{fig:qual_comp_4d}
\end{figure}

Unlike the compositional case, several approaches exist for 4D object-aware generative reconstruction. We show qualitative comparisons in \cref{fig:qual_comp_4d} in a novel view. We observe that our method captures the shape details accurately. TripoSG~\cite{li2025triposghighfidelity3dshape} provides strong static reconstruction, while it along with V2M4~\cite{chen2025v2m44dmeshanimation}, struggle on temporal consistency for certain frames. Similarly, GVFD~\cite{zhang2025gaussianvariationfielddiffusion} and L4GM~\cite{ren2024l4gmlarge4dgaussian} produce textured Gaussians that look plausible from the input view, but their underlying 3D geometry shows inconsistency when rendered from novel viewpoints.

The quantitative evaluations in \cref{tab:quant_comp_4d} reflect the same observations. On DeformingThings~\cite{li20214dcomplete}, our approach significantly outperforms all baselines across every metric; for instance, our IoU of 0.4191 shows a substantial gain over the next best method. On Objaverse~\cite{deitke2022objaverseuniverseannotated3d}, our approach again achieves the best F-Score and IoU, indicating better shape completeness and overlap, while remaining highly competitive in Chamfer Distance against TripoSG~\cite{li2025triposghighfidelity3dshape}. Overall, these results validate the geometric accuracy and completeness of our method's reconstructions on both moderate and large dynamics.

\begin{table}[h]
  \centering
  \caption{Method comparison across datasets. Lower is better for CD; higher is better for F-Score and IoU. The best score is shown bold while the second best is shown italicized. Runtimes were averaged to reflect the reconstruction of a single frame.}
  \label{tab:methods-datasets}
  \setlength{\tabcolsep}{6pt}
  \resizebox{\columnwidth}{!}{%
  \begin{tabular}{l ccc ccc c}
    \toprule
    & \multicolumn{3}{c}{DeformingThings~\cite{li20214dcomplete}} & \multicolumn{3}{c}{Objaverse~\cite{deitke2022objaverseuniverseannotated3d}} & \multicolumn{1}{c}{Runtime} \\
    \cmidrule(lr){2-4} \cmidrule(lr){5-7} \cmidrule(lr){8-8}
    Method & CD $\downarrow$ & F-Score $\uparrow$ & IoU $\uparrow$ & CD $\downarrow$ & F-Score $\uparrow$ & IoU $\uparrow$ & Sec. $\downarrow$ \\
    \midrule
    V2M4~\cite{chen2025v2m44dmeshanimation}     & 0.1678 & 0.4759 & 0.1533 & 0.1595 & 0.4903 & 0.2135 & 61.98 \\
    L4GM~\cite{ren2024l4gmlarge4dgaussian}      & 0.2633 & 0.3167 & ---    & 0.2308 & 0.3236 & ---    & \textbf{1.22} \\
    GVFD~\cite{zhang2025gaussianvariationfielddiffusion}     & 0.2806 & 0.2706 & ---    & 0.2728 & 0.2856 & ---    & 2.87 \\
    TripoSG~\cite{li2025triposghighfidelity3dshape} & \textit{0.1558} & \textit{0.5179} & \textit{0.1784} & \textbf{0.1107} & \textit{0.6585} & \textit{0.2874} & \textit{2.59} \\
    Ours                                         & \textbf{0.1144} & \textbf{0.8388} & \textbf{0.4191} & \textit{0.1205} & \textbf{0.7349} & \textbf{0.3413} & 4.48 \\
    \bottomrule
  \end{tabular}%
  }
  \label{tab:quant_comp_4d}
\end{table}

\subsection{3D Scene Reconstruction}
\begin{figure}[h]
  \centering
  \setlength{\tabcolsep}{0.2pt} % horizontal space
  \renewcommand{\arraystretch}{0} % vertical compression
  \begin{tabular}{@{}cccc@{}}
    Input &
    \textbf{Ours} &
    PartCrafter~\cite{lin2025partcrafterstructured3dmesh} &
    MIDI~\cite{huang2025midimultiinstancediffusionsingle} \\
    % Row 1
    \raisebox{-0.5\height}
    {\includegraphics[width=0.23\linewidth]{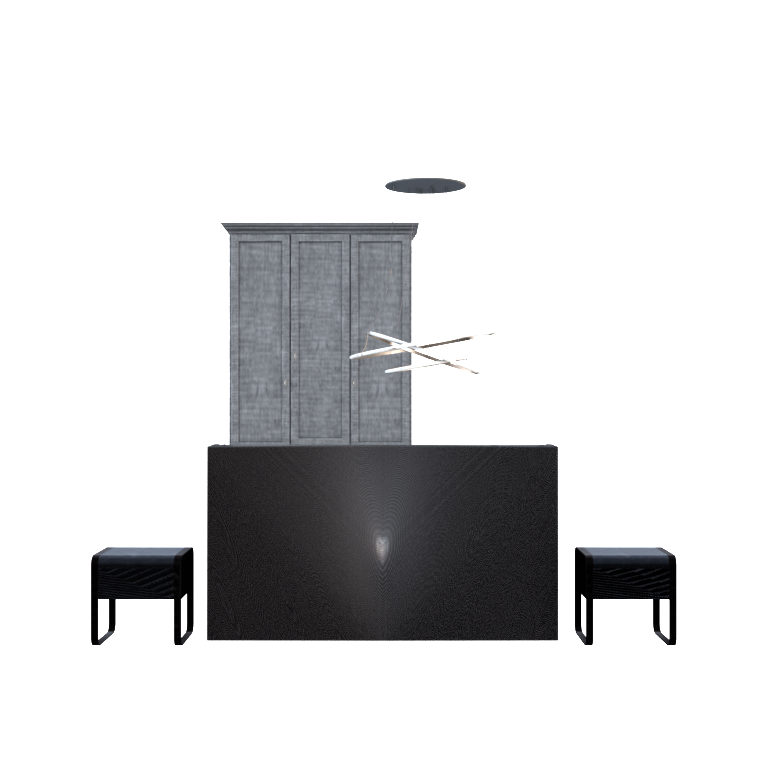}} &
    \raisebox{-0.5\height}{\includegraphics[width=0.23\linewidth]{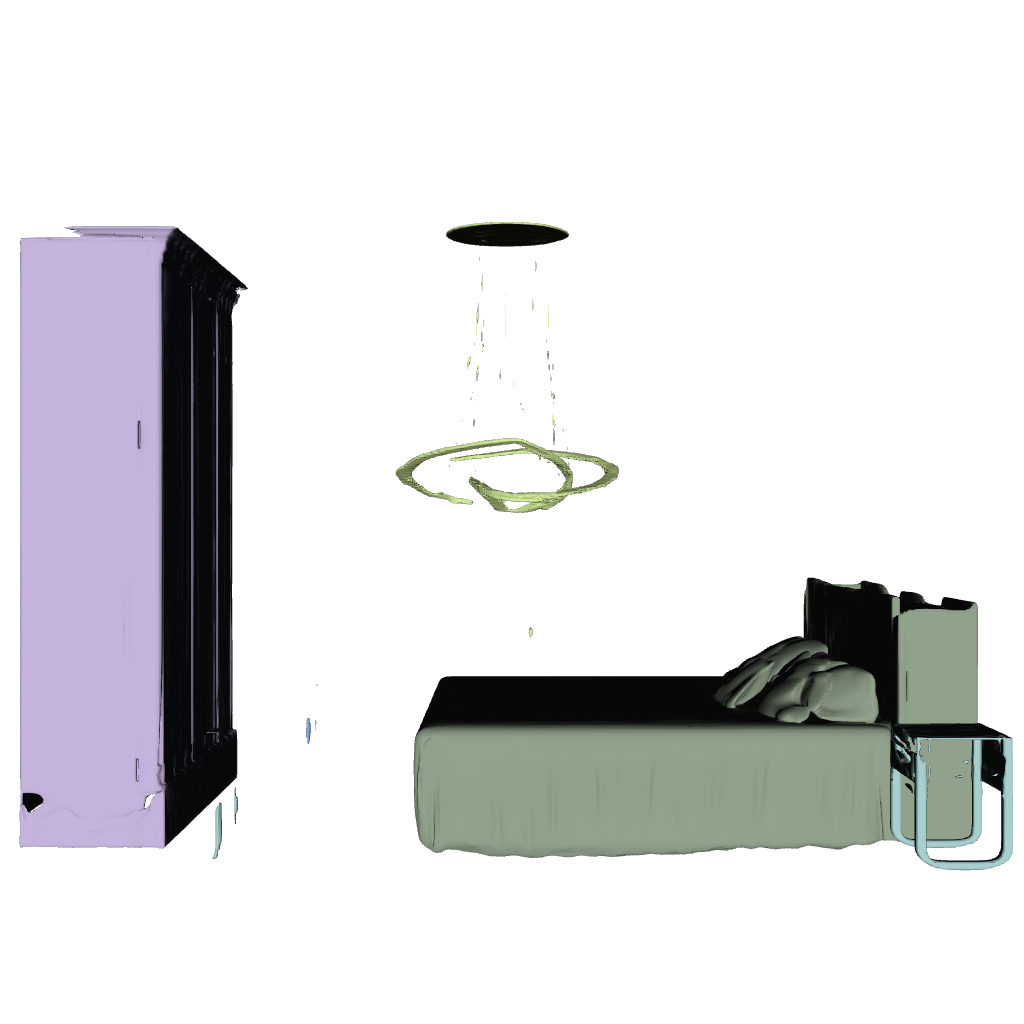}} &
    \raisebox{-0.5\height}{\includegraphics[width=0.23\linewidth]{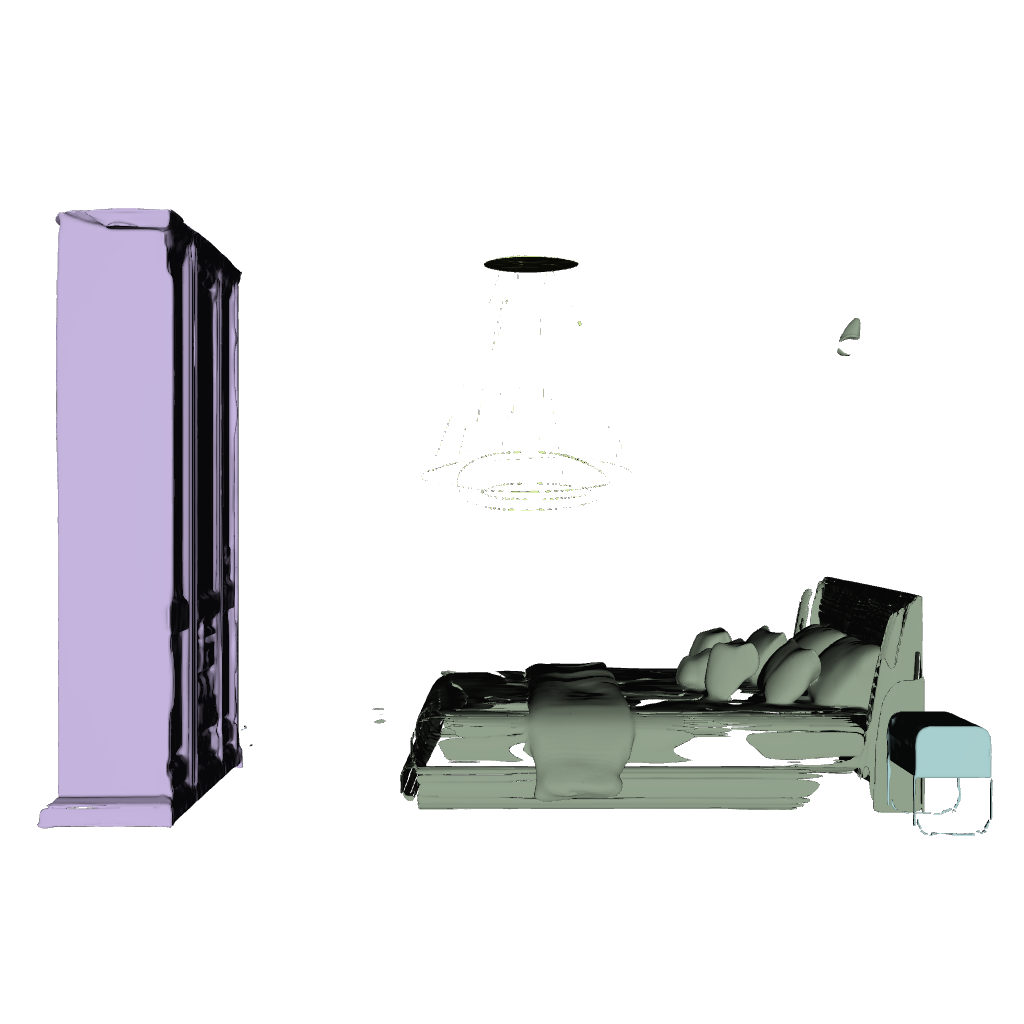}} &
    \raisebox{-0.5\height}{\includegraphics[width=0.23\linewidth]{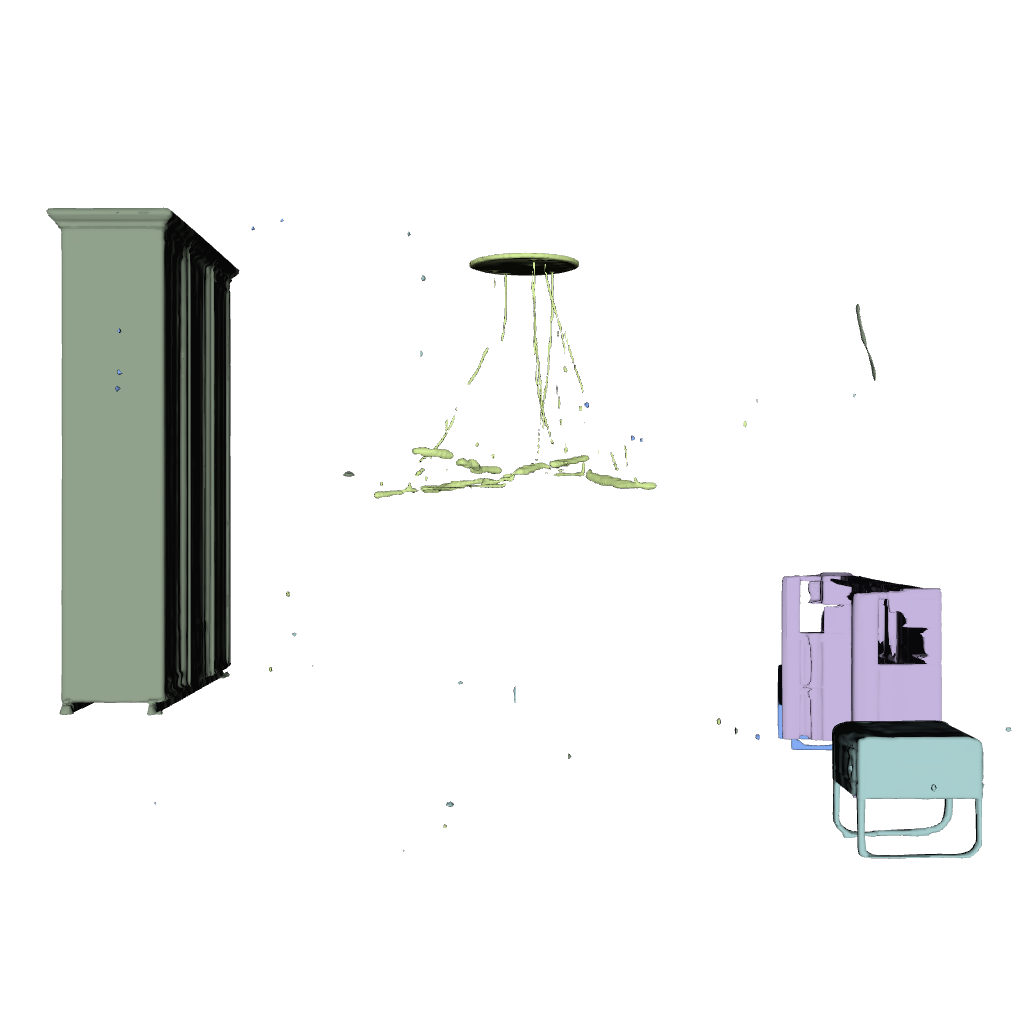}} \\%
    
    % \raisebox{-0.5\height}{\adjincludegraphics[width=0.23\linewidth, trim={{0.15\width} {0.15\height} {0.15\width} {0.15\height}}, clip]{figures/comparisons/3d/00212__e901e594-bf48-4f10-b837-7575f8a22df3__MasterBedroom-15662/input.jpg}} &
    % \raisebox{-0.5\height}{\adjincludegraphics[width=0.23\linewidth, trim={{0.15\width} {0.15\height} {0.15\width} {0.15\height}}, clip]{figures/comparisons/3d/00212__e901e594-bf48-4f10-b837-7575f8a22df3__MasterBedroom-15662/ours/file_00.jpg}} &
    % \raisebox{-0.5\height}{\adjincludegraphics[width=0.23\linewidth, trim={{0.15\width} {0.15\height} {0.15\width} {0.15\height}}, clip]{figures/comparisons/3d/00212__e901e594-bf48-4f10-b837-7575f8a22df3__MasterBedroom-15662/partcrafter/file_00.jpg}} &
    % \raisebox{-0.5\height}{\adjincludegraphics[width=0.23\linewidth, trim={{0.15\width} {0.15\height} {0.15\width} {0.15\height}}, clip]{figures/comparisons/3d/00212__e901e594-bf48-4f10-b837-7575f8a22df3__MasterBedroom-15662/midi/file_00.jpg}} \\
    
    \raisebox{-0.5\height}{\includegraphics[width=0.23\linewidth]{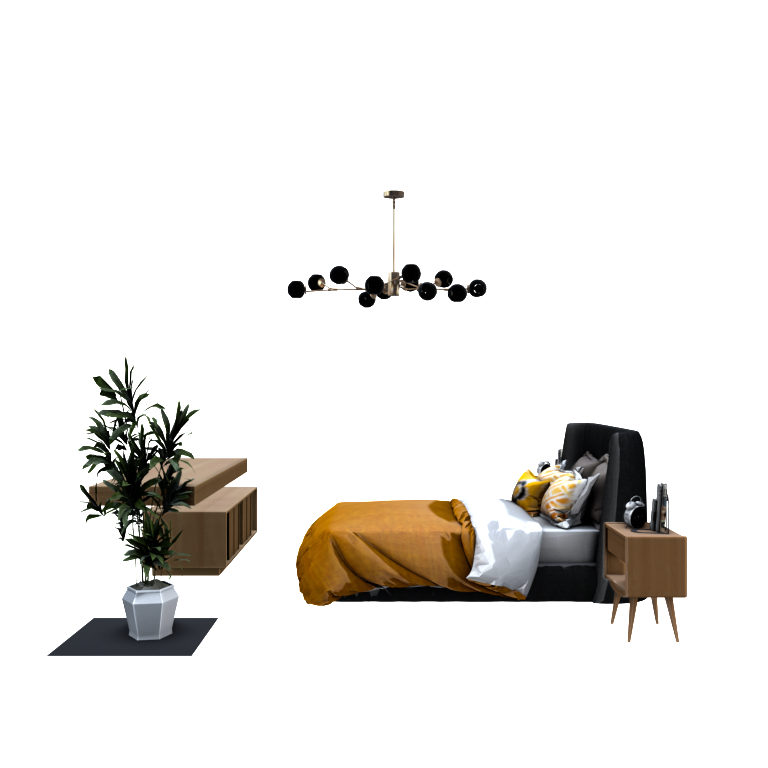}} &
    \raisebox{-0.5\height}{\includegraphics[width=0.23\linewidth]{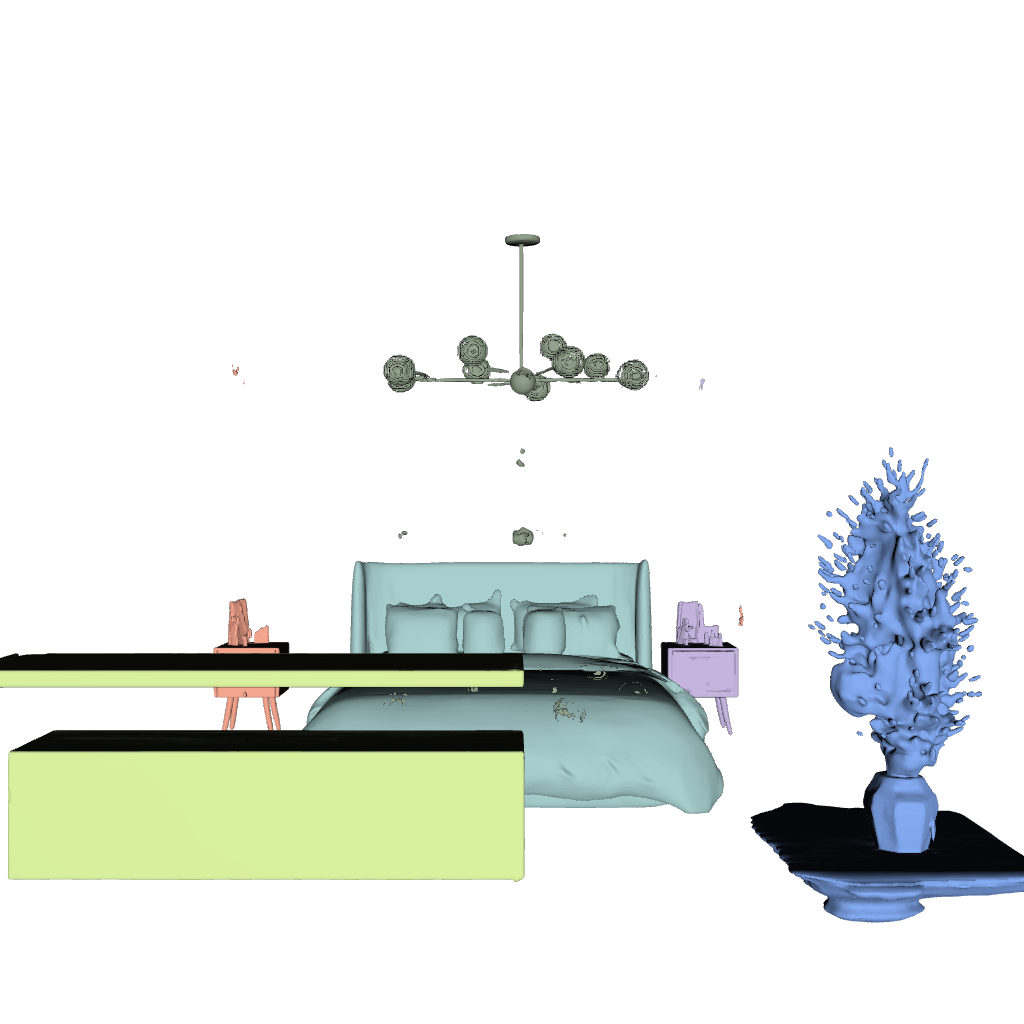}} &
    \raisebox{-0.5\height}{\includegraphics[width=0.23\linewidth]{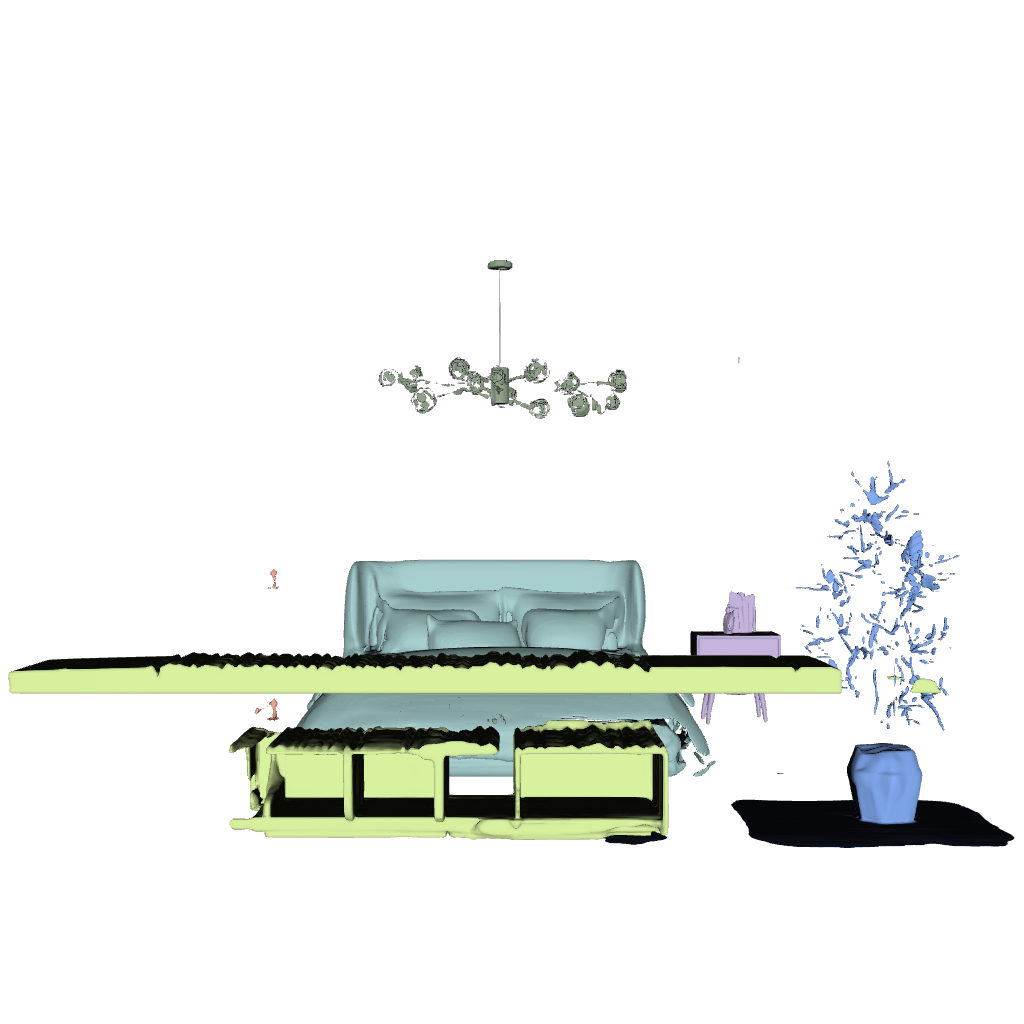}} &
    \raisebox{-0.5\height}{\includegraphics[width=0.23\linewidth]{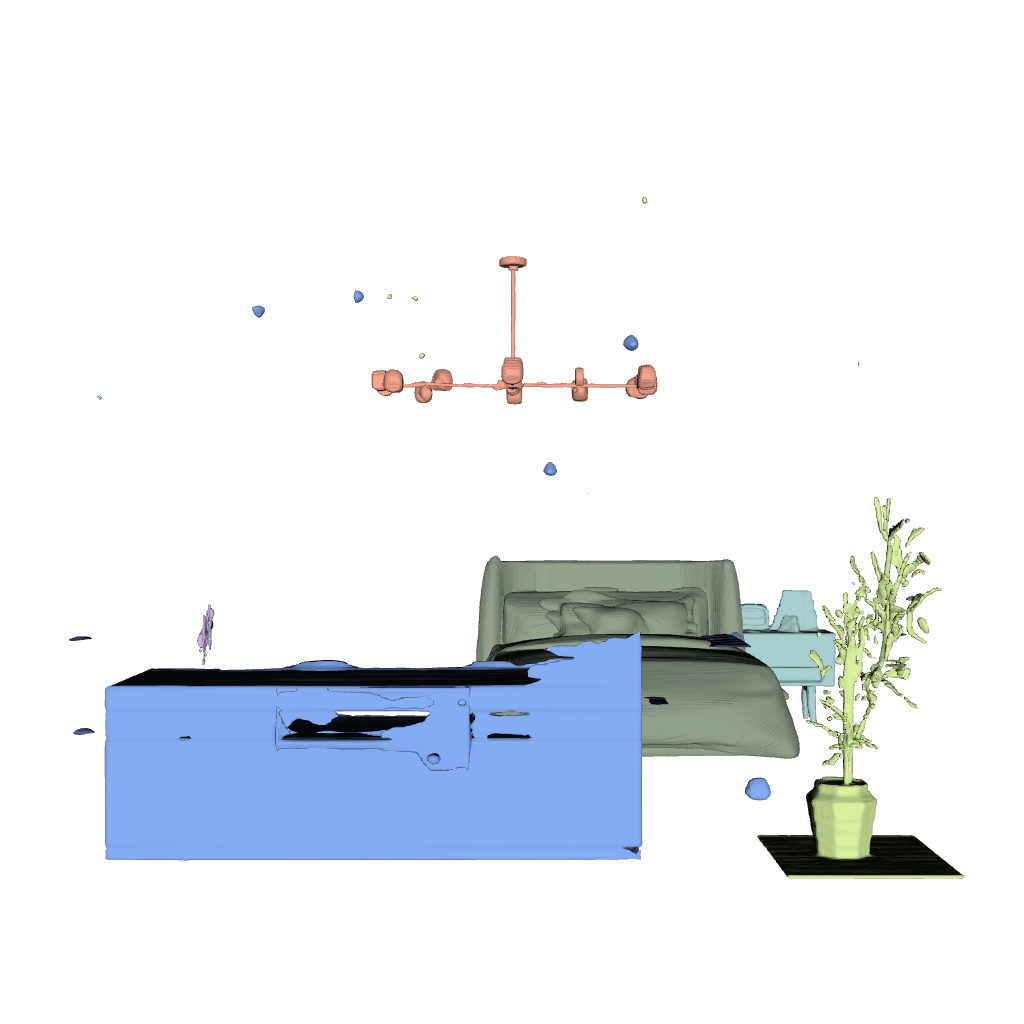}} \\

    % \raisebox{-0.5\height}{\includegraphics[width=0.23\linewidth]{figures/comparisons/3d/00337__f108224b-4b3f-4919-9ca1-7b4306523f17__MasterBedroom-13063/input.jpg}} &
    % \raisebox{-0.5\height}{\includegraphics[width=0.23\linewidth]{figures/comparisons/3d/00337__f108224b-4b3f-4919-9ca1-7b4306523f17__MasterBedroom-13063/ours/file_00.jpg}} &
    % \raisebox{-0.5\height}{\includegraphics[width=0.23\linewidth]{figures/comparisons/3d/00337__f108224b-4b3f-4919-9ca1-7b4306523f17__MasterBedroom-13063/partcrafter/file_00.jpg}} &
    % \raisebox{-0.5\height}{\includegraphics[width=0.23\linewidth]{figures/comparisons/3d/00337__f108224b-4b3f-4919-9ca1-7b4306523f17__MasterBedroom-13063/midi/file_00.jpg}} \\
    
  \end{tabular}

  \caption{
  Qualitative comparison across methods.
  Our approach generates more consistent and detailed structures compared to PartCrafter~\cite{lin2025partcrafterstructured3dmesh} and MIDI~\cite{huang2025midimultiinstancediffusionsingle}.
  }
  \label{fig:qual_comp_3d}
\end{figure}

\cref{fig:qual_comp_3d} provides a visual comparison of our method against previous generative approaches: PartCrafter and MIDI, on the 3D-Front~\cite{fu20213dfront3dfurnishedrooms} examples. Our method reconstructs complete and detailed object layouts, being more faithful to the input. In contrast, PartCrafter~\cite{lin2025partcrafterstructured3dmesh} occasionally produces partial or low-quality meshes, while MIDI~\cite{huang2025midimultiinstancediffusionsingle} struggles with occlusions.

\begin{table}[h]
  \centering
  \caption{3D Scene Generation on 3D-FRONT and 3D-FRONT-Occluded. Lower is better for CD and IoU; higher is better for F-Score. The best score is shown bold while the second best is shown italicized.}
  \label{tab:front-all-one}
  \resizebox{\linewidth}{!}{%
    \begin{tabular}{lccc cc c}
      \toprule
      & \multicolumn{3}{c}{3D-FRONT~\cite{fu20213dfront3dfurnishedrooms}} & \multicolumn{2}{c}{3D-FRONT-Occluded~\cite{fu20213dfront3dfurnishedrooms}} & \multicolumn{1}{c}{Runtime} \\
      \cmidrule(lr){2-4} \cmidrule(lr){5-6} \cmidrule(lr){7-7}
      Method & CD $\downarrow$ & F-Score $\uparrow$ & IoU $\downarrow$ & CD $\downarrow$ & F-Score $\uparrow$ & Seconds $\downarrow$ \\
      \midrule
      MIDI~\cite{huang2025midimultiinstancediffusionsingle}        & \textit{0.1445} & \textit{0.7829} & 0.0027 & \textit{0.1781} & 0.7387 & 3.51 \\
      PartCrafter~\cite{lin2025partcrafterstructured3dmesh}         & 0.1751 & 0.7569 & \textbf{0.0017} & 0.1951 & \textit{0.7461} & \textbf{2.60} \\
      Ours                                                         & \textbf{0.0909} & \textbf{0.8069} & \textit{0.0018} & \textbf{0.1256} & \textbf{0.7521} & \textit{2.63} \\
      \bottomrule
    \end{tabular}%
  }
  \label{tab:quant_comp_3d}
\end{table}

The quantitative evaluation makes these even clearer. \cref{tab:quant_comp_3d} compares our method with recent baselines on the 3D-FRONT dataset~\cite{fu20213dfront3dfurnishedrooms}. Our approach demonstrates state-of-the-art performance, achieving the best Chamfer Distance (0.0909) and F-Score (0.8069). This represents a significant improvement in geometric accuracy, surpassing the second-best method, MIDI~\cite{huang2025midimultiinstancediffusionsingle}. This advantage holds even on the challenging 3D-FRONT-Occluded split.

\begin{comment}

\subsubsection{Quantitive Results}
\cref{tab:quant_comp_3d} compares our method with recent baselines on the 3D-FRONT dataset~\cite{fu20213dfront3dfurnishedrooms}. Our approach demonstrates state-of-the-art performance, achieving the best Chamfer Distance (0.0909) and F-Score (0.8069). This represents a significant improvement in geometric accuracy, surpassing the second-best method, MIDI~\cite{huang2025midimultiinstancediffusionsingle}. This advantage holds even on the challenging 3D-FRONT-Occluded split.

Similarly, for 4D generation in \cref{tab:quant_comp_4d}, our method shows strong performance. On the DeformingThings dataset~\cite{li20214dcomplete}, our approach significantly outperforms all baselines across every metric; for instance, our IoU of 0.4191 shows a substantial gain over the next best method. On the Objaverse dataset~\cite{deitke2022objaverseuniverseannotated3d}, our approach again achieves the best F-Score and IoU, indicating superior shape completeness and overlap, while remaining highly competitive in Chamfer Distance against TripoSG~\cite{li2025triposghighfidelity3dshape}. Overall, these results validate the effectiveness of our approach in producing geometrically accurate and complete shapes for both static and dynamic scenes.

\end{comment}

\subsection{Ablation Study}
\label{sec:ablation_study}
We conduct a series of ablation studies to validate the effectiveness of our key architectural components: the use of distinct static/dynamic embeddings, the Diffusion Forcing training scheme, and our attention mixing strategy.

\noindent\textbf{Quantitative Analysis.}
\noindent The quantitative results in ~\cref{tab:ablation-ours} validate our design choices against a baseline model. Adding the Static/Dynamic Embeddings yields the most significant performance gain; on DeformingThings, this drops the Chamfer Distance (CD) from 0.1525 to 0.1284 and nearly doubles IoU from 0.2018 to 0.4034, highlighting the importance of disentangling static and dynamic representations. Diffusion Forcing also provides a notable improvement, particularly for IoU, which enhances temporal consistency. Our full model COM4D, combining both components, achieves the best overall performance.

\begin{figure}[h]
  \centering
  \setlength{\tabcolsep}{2pt} % Adjust horizontal "small gap" between frames
  
  % ==================== TABLE START ====================
  \begin{tabular}{@{}ccc@{}}
    % -------- Headers --------
    \textbf{Input} & \textbf{w/o Mixing} & \textbf{w/ Mixing} \\
    
    \raisebox{-.5\height}{\includegraphics[width=0.3\columnwidth]{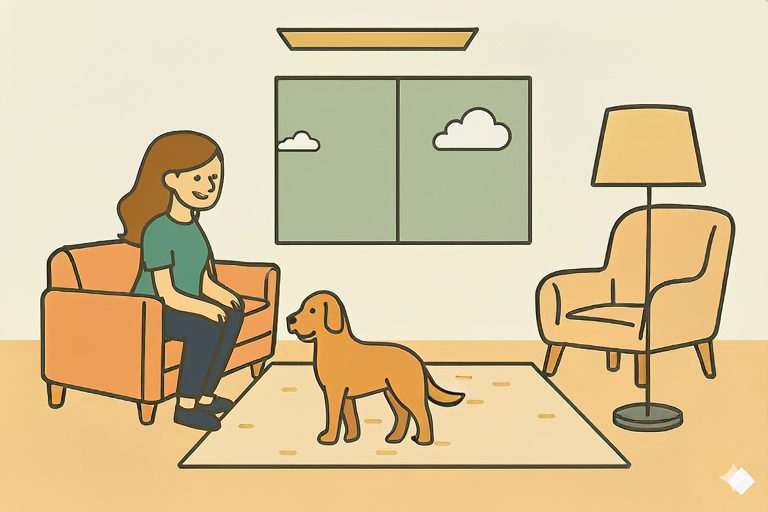}} &
    % w/o Mixing Frame 4 (single bottom image & zoomed)
    \raisebox{-.5\height}{\includegraphics[width=0.3\columnwidth, trim=50 50 50 50, clip]{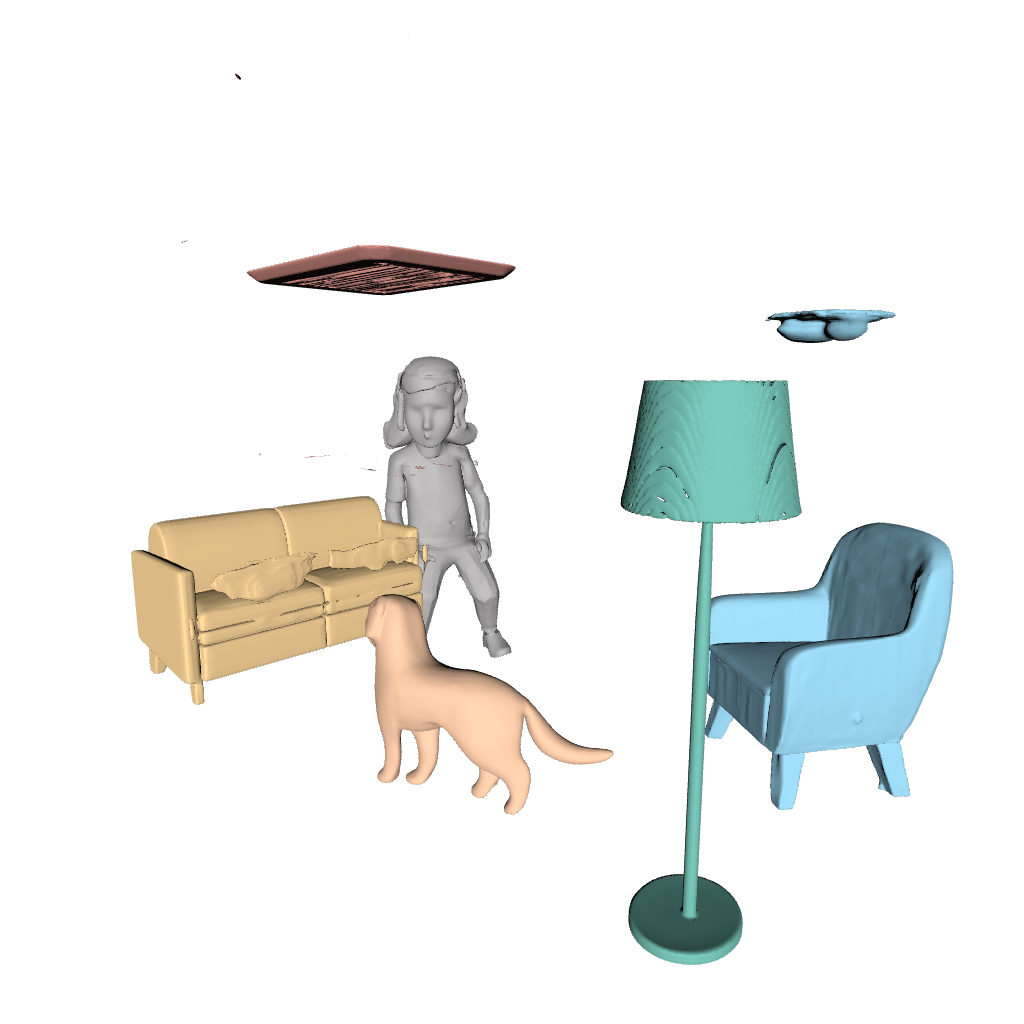}} &
    % Ours Frame 4 (single bottom image & zoomed)
    \raisebox{-.5\height}{\includegraphics[width=0.3\columnwidth, trim=50 50 50 50, clip]{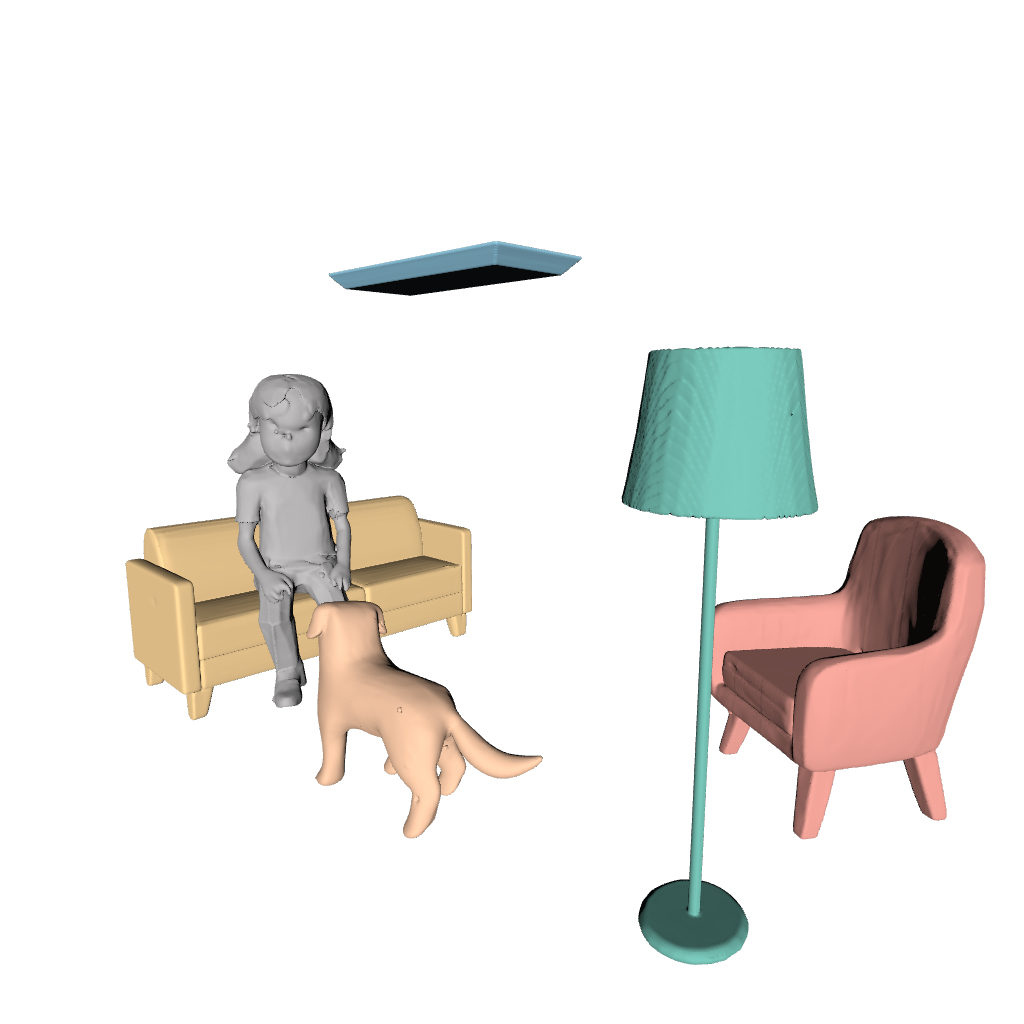}} \\

    % ==================== CAT EXAMPLE - FRAME 3 ====================
    % Input Frame 3
    \raisebox{-.5\height}{\includegraphics[width=0.3\columnwidth]{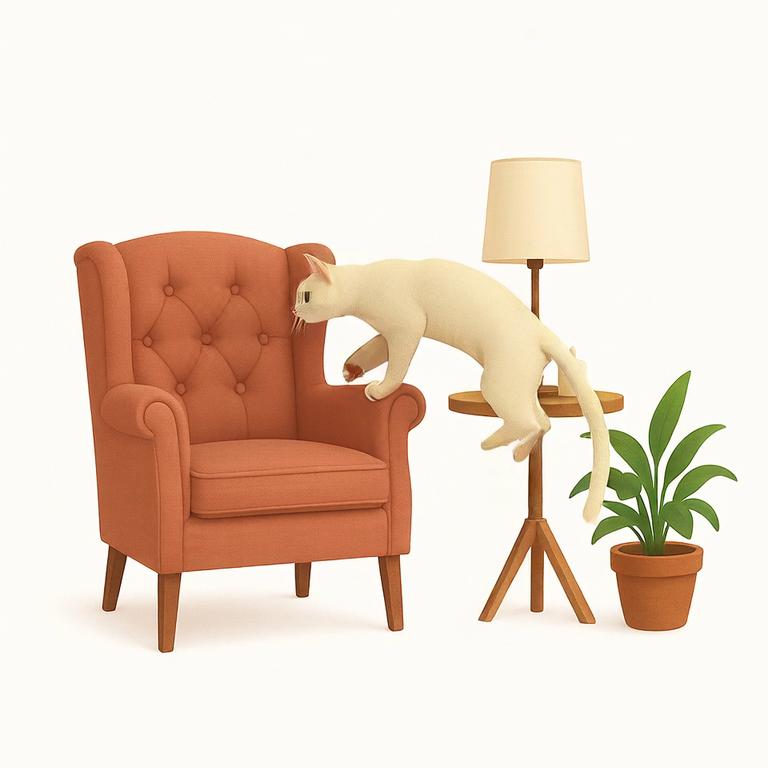}} &
    % w/o Mixing Frame 3 (single bottom image & zoomed)
    \raisebox{-.5\height}{\includegraphics[width=0.3\columnwidth, trim=50 50 50 50, clip]{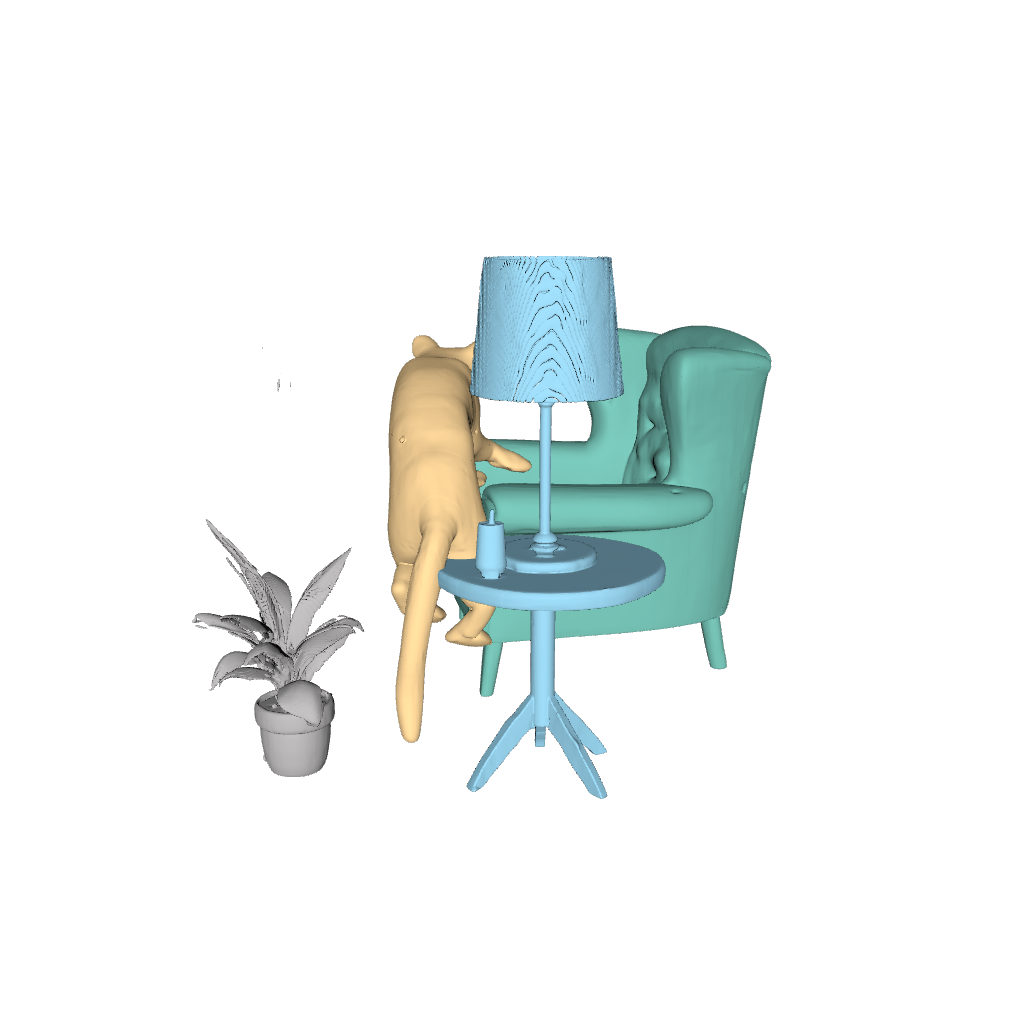}} &
    % Ours Frame 3 (single bottom image & zoomed)
    \raisebox{-.5\height}{\includegraphics[width=0.3\columnwidth, trim=50 50 50 50, clip]{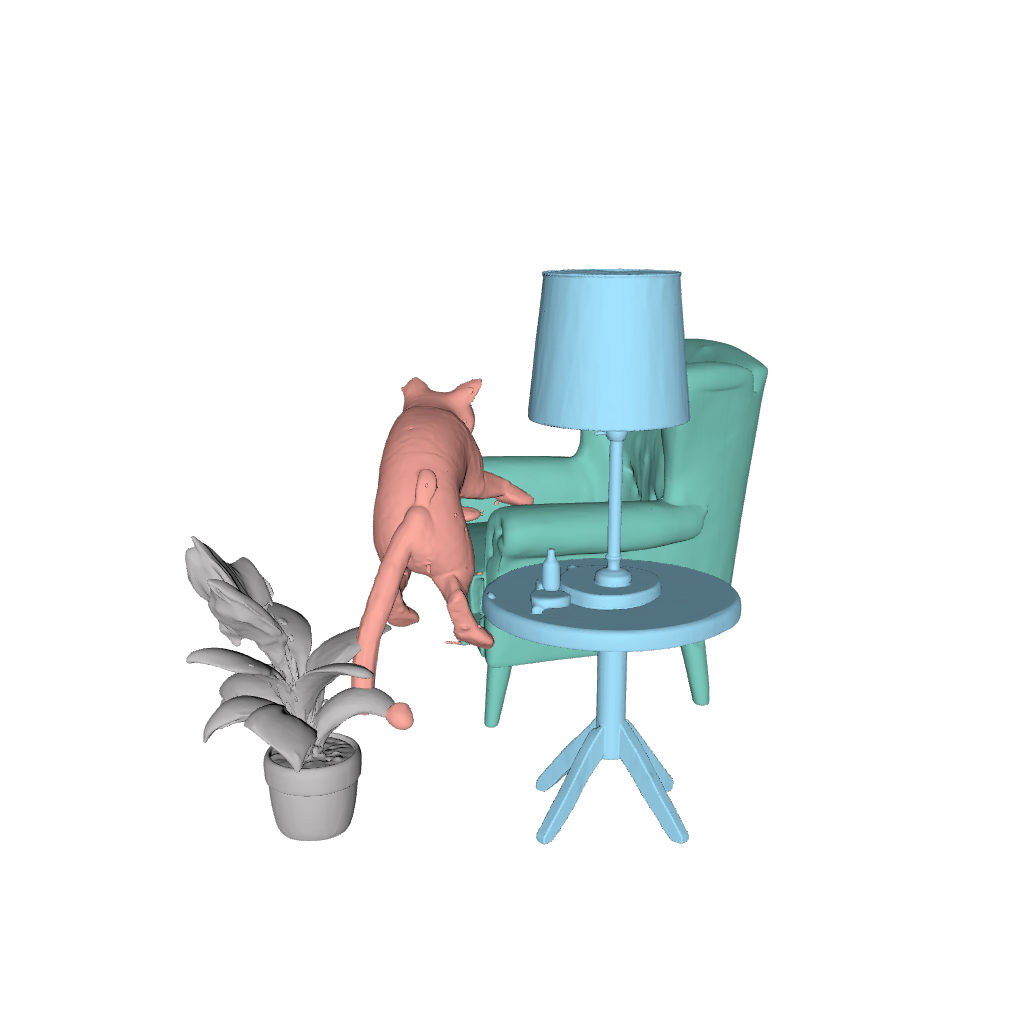}} \\

    % ==================== LIVING ROOM EXAMPLE - FRAME 4 ====================
    % Input Frame 4
    \raisebox{-.5\height}{\includegraphics[width=0.3\columnwidth]{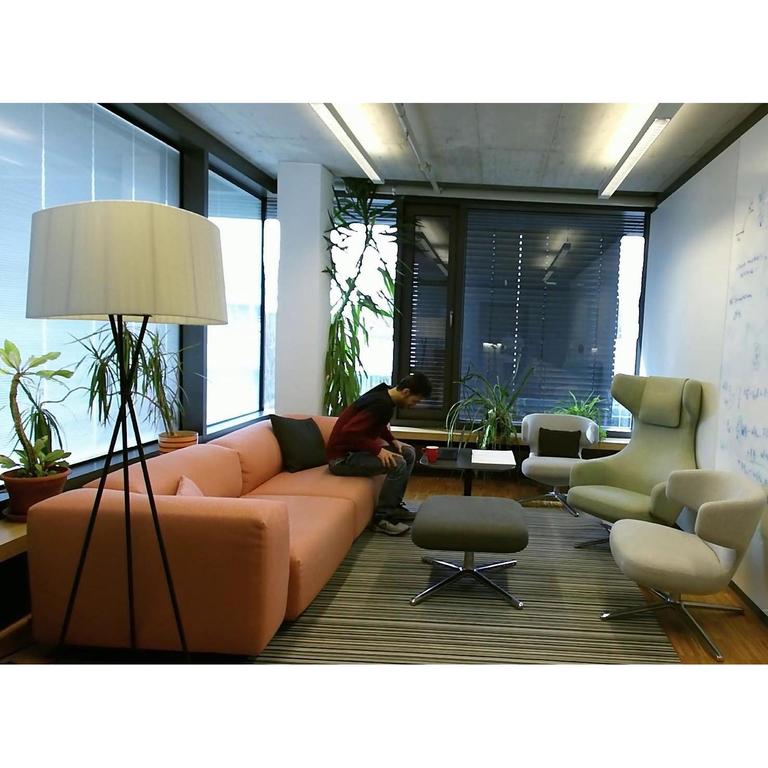}} &
    % w/o Mixing Frame 4 (single bottom image & zoomed)
    \raisebox{-.5\height}{\includegraphics[width=0.3\columnwidth, trim=50 50 50 50, clip]{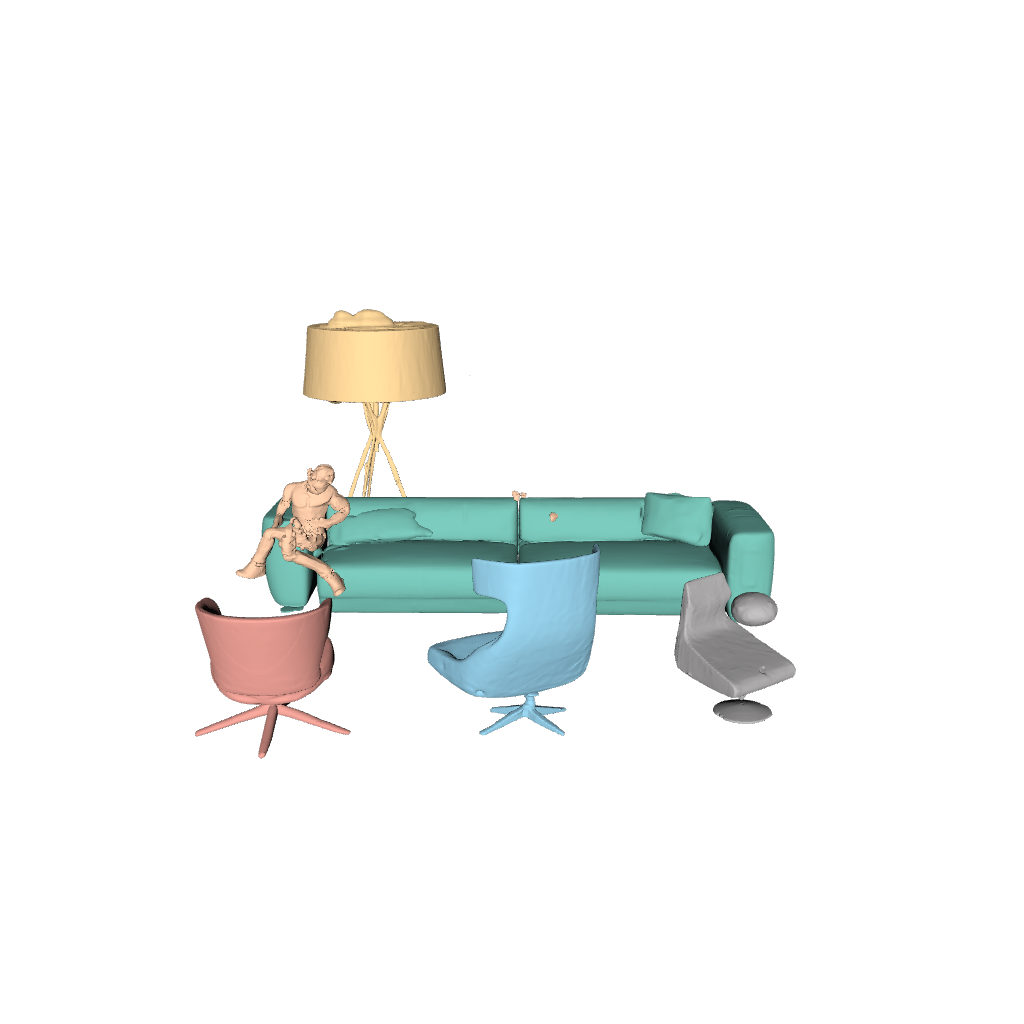}} &
    % Ours Frame 4 (single bottom image & zoomed)
    \raisebox{-.5\height}{\includegraphics[width=0.3\columnwidth, trim=50 50 50 50, clip]{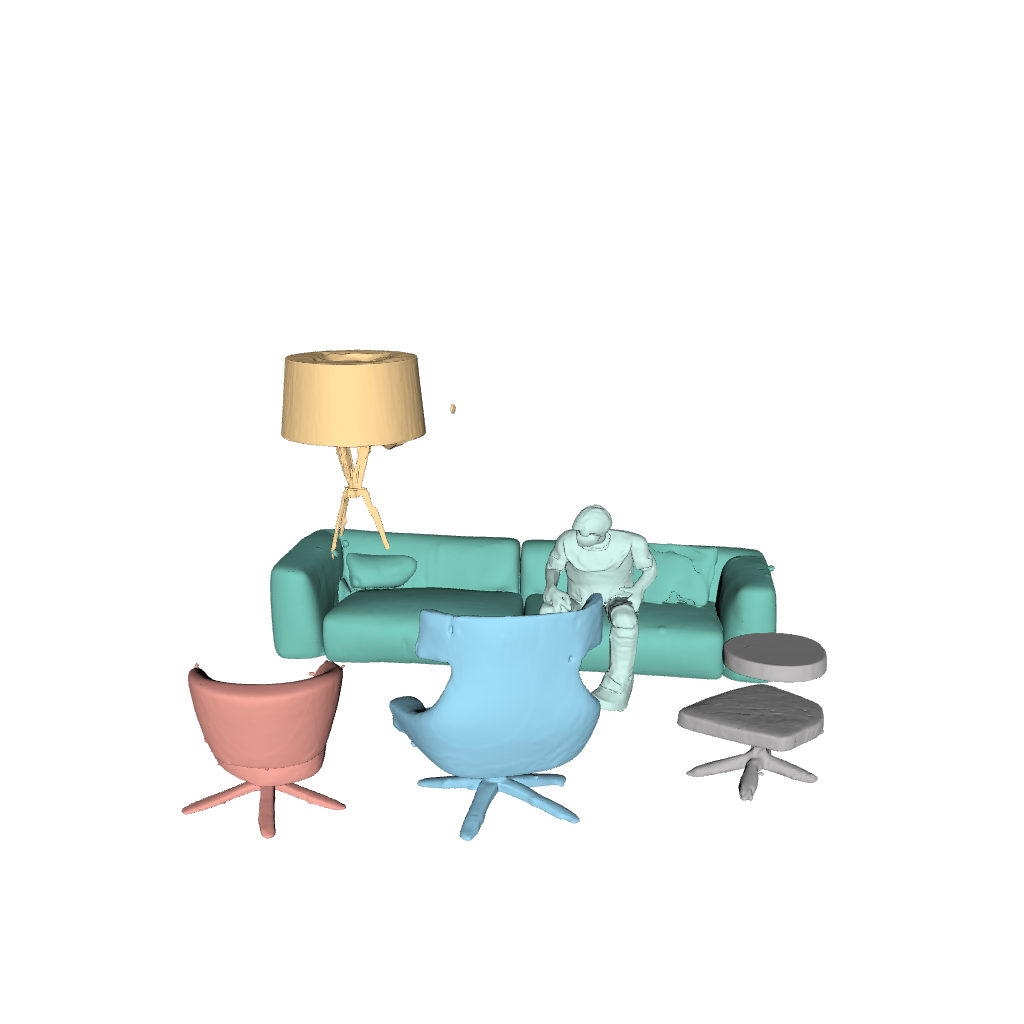}} \\
   
  \end{tabular}
  \caption{Qualitative comparisons on Attention Mixing. Mixing (right) provides far better static-dynamic compositions compared to without Mixing (middle).}
  % ==================== TABLE END ====================
  \label{fig:ablation_final}
\vspace{-5mm}

\end{figure}

\noindent\textbf{Analysis of Attention Mixing.}
Fig.~\ref{fig:ablation_final} qualitatively validates our \emph{Attention Mixing} strategy. Without it, the model fails to ground dynamic objects in their static context, resulting in artifacts where the cat overlaps with the lamp stand and humans are not sitting where they are supposed to. In contrast, by alternating between spatial and temporal attention in the Attention Mixing strategy, our model makes dynamic objects aware of static placements. This leads to reconstructions that are significantly more spatially accurate, validating our strategy for modeling plausible temporal evolutions in multi-object scenes.

We report quantitative results in two forms. Because no compositional 4D dataset with object-level decomposition exists, our main evaluation relies on user studies in \cref{fig:teaser}.B, where our method is preferred over a standard generative baseline (without Attention Mixing) by a factor of 12 (87\% vs. 6.9\%). We also evaluate on CMU Panoptic~\cite{panoptic,Joo_2018_CVPR} point clouds \emph{by registering only the first frame reconstruction}, where Attention Mixing reduces the average CD from 35.91 cm to 7.42 cm. This demonstrates that COM4D can accurately capture multiple dynamic shape evolutions in long sequences $>90$ frames, with surprisingly small drifts. \cref{fig:panoptic_quant} visualizes the reconstructions of ours and the baseline. Further details and additional quantitive analysis on synthetic compositional data appear in the supplementary material.

\begin{table}[h]
  \centering
  \caption{
  Ablation study on 3D-FRONT and DeformingThings datasets.
  ``Baseline'' refers to the model without static/dynamic embeddings or diffusion forcing, while ``Ours'' uses both.
  Lower is better for CD and IoU in 3D-FRONT; higher is better for F-Score and IoU in DeformingThings.
  }
  \vspace{-2mm}
  \label{tab:ablation-ours}
  \resizebox{\linewidth}{!}{%
    \begin{tabular}{l ccc ccc}
      \toprule
      & \multicolumn{3}{c}{3D-FRONT~\cite{fu20213dfront3dfurnishedrooms}} 
      & \multicolumn{3}{c}{DeformingThings~\cite{li20214dcomplete}} \\
      \cmidrule(lr){2-4} \cmidrule(lr){5-7}
      Method & CD $\downarrow$ & F $\uparrow$ & IoU $\downarrow$
             & CD $\downarrow$ & F $\uparrow$ & IoU $\uparrow$ \\
      \midrule
      Baseline & 0.1668 & 0.6347 & 0.0025 & 0.1525 & 0.7854 & 0.2018 \\
      + Static/Dynamic Emb. & 0.1044 & 0.7888 & 0.0017 & 0.1284 & \textbf{0.9350} & 0.4034 \\
      + Diffusion Forcing & 0.1247 & 0.7152 & 0.0021 & 0.1488 & 0.8189 & \textbf{0.4271} \\
      \textbf{Ours} & \textbf{0.0909} & \textbf{0.8069} & \textbf{0.0018} & \textbf{0.1144} & 0.8388 & 0.4191 \\
      \bottomrule
    \end{tabular}%
  }
\end{table}

\section{Conclusion}
In this work, we introduced COM4D, a novel approach for reconstructing compositional 4D scenes from monocular videos, without requiring direct supervision on such scarce data. Our work is motivated by the inherent difficulty of acquiring real world compositional 4D for training, which in fact, may not be resolved in the near future. We instead, address the problem by proposing a unique training strategy called \textit{Attention Parsing}, which factorizes the learning of spatial and temporal priors into disentangled attention pathways. At inference, our \textit{Attention Mixing} mechanism effectively fuses these learned attentions to produce coherent, persistent, and compositionally consistent 4D reconstructions, even in scenes containing multiple interacting static and dynamic objects. COM4D achieves state-of-the-art performance on both 3D compositional scene reconstruction and 4D dynamic object reconstruction, underscoring the versatility and effectiveness of the approach. Despite these promising results, certain limitations remain. The model's understanding of motion is learned from data and does not incorporate an explicit physical causality. Consequently, when objects become occluded, it can struggle to reason about their continued trajectory and interaction in a physically plausible manner. Furthermore, COM4D is designed to operate on videos captured from a fixed camera perspective and does not currently support scenarios with camera motion. Future work could explore integrating physical causality to improve reasoning under occlusions and extending it to dynamic camera inputs.

\paragraph{Acknowledgements.}
This research was partially funded by the Ministry of Education and Science of Bulgaria (support for INSAIT, part of
the Bulgarian National Roadmap for Research Infrastructure). The research was also partly funded by INSAIT-VIVO project on 3D scene understanding and editing.
{
    \small
    \bibliographystyle{ieeenat_fullname}
    \bibliography{main}

\begin{thebibliography}{99}
\providecommand{\natexlab}[1]{#1}
\providecommand{\url}[1]{\texttt{#1}}
\expandafter\ifx\csname urlstyle\endcsname\relax
  \providecommand{\doi}[1]{doi: #1}\else
  \providecommand{\doi}{doi: \begingroup \urlstyle{rm}\Url}\fi

\bibitem[Agudo et~al.(2014)Agudo, Agapito, Calvo, and Montiel]{AgudoACM14}
Antonio Agudo, Lourdes Agapito, Bego{\~{n}}a Calvo, and J.~M.~M. Montiel.
\newblock Good vibrations: {A} modal analysis approach for sequential non-rigid structure from motion.
\newblock In \emph{2014 {IEEE} Conference on Computer Vision and Pattern Recognition, {CVPR} 2014, Columbus, OH, USA, June 23-28, 2014}, pages 1558--1565. {IEEE} Computer Society, 2014.

\bibitem[Bahmani et~al.(2024)Bahmani, Skorokhodov, Rong, Wetzstein, Guibas, Wonka, Tulyakov, Park, Tagliasacchi, and Lindell]{Bahmani_2024_CVPR}
Sherwin Bahmani, Ivan Skorokhodov, Victor Rong, Gordon Wetzstein, Leonidas Guibas, Peter Wonka, Sergey Tulyakov, Jeong~Joon Park, Andrea Tagliasacchi, and David~B. Lindell.
\newblock 4d-fy: Text-to-4d generation using hybrid score distillation sampling.
\newblock In \emph{Proceedings of the IEEE/CVF Conference on Computer Vision and Pattern Recognition (CVPR)}, pages 7996--8006, 2024.

\bibitem[Barrow et~al.(1977)Barrow, Tenenbaum, Bolles, and Wolf]{barrow77}
H.~G. Barrow, J.~M. Tenenbaum, R.~C. Bolles, and H.~C. Wolf.
\newblock Parametric correspondence and chamfer matching: two new techniques for image matching.
\newblock In \emph{Proceedings of the 5th International Joint Conference on Artificial Intelligence - Volume 2}, page 659–663, San Francisco, CA, USA, 1977. Morgan Kaufmann Publishers Inc.

\bibitem[Blattmann et~al.(2023)Blattmann, Rombach, Ling, Dockhorn, Kim, Fidler, and Kreis]{Blattmann_2023_CVPR}
Andreas Blattmann, Robin Rombach, Huan Ling, Tim Dockhorn, Seung~Wook Kim, Sanja Fidler, and Karsten Kreis.
\newblock Align your latents: High-resolution video synthesis with latent diffusion models.
\newblock In \emph{Proceedings of the IEEE/CVF Conference on Computer Vision and Pattern Recognition (CVPR)}, pages 22563--22575, 2023.

\bibitem[Borgefors(1988)]{Chamfer88}
G. Borgefors.
\newblock Hierarchical chamfer matching: a parametric edge matching algorithm.
\newblock \emph{IEEE Transactions on Pattern Analysis and Machine Intelligence}, 10\penalty0 (6):\penalty0 849--865, 1988.

\bibitem[Bregler et~al.(2000)Bregler, Hertzmann, and Biermann]{Bregler2000RecoveringN3}
Christoph Bregler, Aaron Hertzmann, and Henning Biermann.
\newblock Recovering non-rigid 3d shape from image streams.
\newblock \emph{Proceedings IEEE Conference on Computer Vision and Pattern Recognition. CVPR 2000 (Cat. No.PR00662)}, 2:\penalty0 690--696 vol.2, 2000.

\bibitem[Cao et~al.(2024)Cao, Luo, Zhang, Nie{\ss}ner, and Tang]{cao2024motion2vecsets}
Wei Cao, Chang Luo, Biao Zhang, Matthias Nie{\ss}ner, and Jiapeng Tang.
\newblock Motion2vecsets: 4d latent vector set diffusion for non-rigid shape reconstruction and tracking.
\newblock In \emph{Proceedings of the IEEE/CVF conference on computer vision and pattern recognition}, pages 20496--20506, 2024.

\bibitem[Chang et~al.(2015)Chang, Funkhouser, Guibas, Hanrahan, Huang, Li, Savarese, Savva, Song, Su, et~al.]{chang2015shapenet}
Angel~X Chang, Thomas Funkhouser, Leonidas Guibas, Pat Hanrahan, Qixing Huang, Zimo Li, Silvio Savarese, Manolis Savva, Shuran Song, Hao Su, et~al.
\newblock Shapenet: An information-rich 3d model repository.
\newblock \emph{arXiv preprint arXiv:1512.03012}, 2015.

\bibitem[Chen et~al.(2024)Chen, Monso, Du, Simchowitz, Tedrake, and Sitzmann]{chen2024diffusionforcingnexttokenprediction}
Boyuan Chen, Diego~Marti Monso, Yilun Du, Max Simchowitz, Russ Tedrake, and Vincent Sitzmann.
\newblock Diffusion forcing: Next-token prediction meets full-sequence diffusion, 2024.

\bibitem[Chen et~al.(2025{\natexlab{a}})Chen, Zhang, Tang, and Wonka]{chen2025v2m44dmeshanimation}
Jianqi Chen, Biao Zhang, Xiangjun Tang, and Peter Wonka.
\newblock V2m4: 4d mesh animation reconstruction from a single monocular video, 2025{\natexlab{a}}.

\bibitem[Chen et~al.(2025{\natexlab{b}})Chen, Zhang, Wimbauer, Wang, Araslanov, Vedaldi, and Cremers]{Chen_2025_ICCV}
Weirong Chen, Ganlin Zhang, Felix Wimbauer, Rui Wang, Nikita Araslanov, Andrea Vedaldi, and Daniel Cremers.
\newblock Back on track: Bundle adjustment for dynamic scene reconstruction.
\newblock In \emph{Proceedings of the IEEE/CVF International Conference on Computer Vision (ICCV)}, pages 4951--4960, 2025{\natexlab{b}}.

\bibitem[Chen et~al.(2025{\natexlab{c}})Chen, Chen, Xue, Chen, Xiu, and Pons-Moll]{chen2025human3r}
Yue Chen, Xingyu Chen, Yuxuan Xue, Anpei Chen, Yuliang Xiu, and Gerard Pons-Moll.
\newblock Human3r: Everyone everywhere all at once.
\newblock \emph{arXiv preprint arXiv:2510.06219}, 2025{\natexlab{c}}.

\bibitem[Chu et~al.(2024)Chu, Ke, and Fragkiadaki]{dreamscene4d}
Wen-Hsuan Chu, Lei Ke, and Katerina Fragkiadaki.
\newblock Dreamscene4d: Dynamic multi-object scene generation from monocular videos.
\newblock \emph{NeurIPS}, 2024.

\bibitem[Dai et~al.(2012)Dai, Li, and He]{Dai2012}
Yuchao Dai, Hongdong Li, and Mingyi He.
\newblock A simple prior-free method for non-rigid structure-from-motion factorization.
\newblock In \emph{2012 IEEE Conference on Computer Vision and Pattern Recognition}, pages 2018--2025, 2012.

\bibitem[Deitke et~al.(2022)Deitke, Schwenk, Salvador, Weihs, Michel, VanderBilt, Schmidt, Ehsani, Kembhavi, and Farhadi]{deitke2022objaverseuniverseannotated3d}
Matt Deitke, Dustin Schwenk, Jordi Salvador, Luca Weihs, Oscar Michel, Eli VanderBilt, Ludwig Schmidt, Kiana Ehsani, Aniruddha Kembhavi, and Ali Farhadi.
\newblock Objaverse: A universe of annotated 3d objects, 2022.

\bibitem[Dhariwal and Nichol(2021)]{dhariwal2021diffusion}
Prafulla Dhariwal and Alexander Nichol.
\newblock Diffusion models beat gans on image synthesis.
\newblock \emph{Advances in neural information processing systems}, 34:\penalty0 8780--8794, 2021.

\bibitem[El~Banani et~al.(2024)El~Banani, Raj, Maninis, Kar, Li, Rubinstein, Sun, Guibas, Johnson, and Jampani]{el2024probing}
Mohamed El~Banani, Amit Raj, Kevis-Kokitsi Maninis, Abhishek Kar, Yuanzhen Li, Michael Rubinstein, Deqing Sun, Leonidas Guibas, Justin Johnson, and Varun Jampani.
\newblock Probing the 3d awareness of visual foundation models.
\newblock In \emph{Proceedings of the IEEE/CVF Conference on Computer Vision and Pattern Recognition}, pages 21795--21806, 2024.

\bibitem[Feng* et~al.(2025)Feng*, Zhang*, Wang, Ye, Yu, Black, Darrell, and Kanazawa]{st4rtrack2025}
Haiwen Feng*, Junyi Zhang*, Qianqian Wang, Yufei Ye, Pengcheng Yu, Michael~J. Black, Trevor Darrell, and Angjoo Kanazawa.
\newblock St4rtrack: Simultaneous 4d reconstruction and tracking in the world.
\newblock In \emph{Proceedings of the IEEE/CVF International Conference on Computer Vision}, 2025.

\bibitem[Fu et~al.(2021)Fu, Cai, Gao, Zhang, Li, Xun, Sun, Jia, Zhao, and Zhang]{fu20213dfront3dfurnishedrooms}
Huan Fu, Bowen Cai, Lin Gao, Lingxiao Zhang, Jiaming Wang~Cao Li, Zengqi Xun, Chengyue Sun, Rongfei Jia, Binqiang Zhao, and Hao Zhang.
\newblock 3d-front: 3d furnished rooms with layouts and semantics, 2021.

\bibitem[Girdhar et~al.(2024)Girdhar, Singh, Brown, Duval, Azadi, Rambhatla, Shah, Yin, Parikh, and Misra]{emuvideo2024}
Rohit Girdhar, Mannat Singh, Andrew Brown, Quentin Duval, Samaneh Azadi, Sai~Saketh Rambhatla, Akbar Shah, Xi Yin, Devi Parikh, and Ishan Misra.
\newblock Factorizing text-to-video generation by explicit image conditioning.
\newblock In \emph{Computer Vision – ECCV 2024: 18th European Conference, Milan, Italy, September 29–October 4, 2024, Proceedings, Part LXII}, page 205–224, 2024.

\bibitem[Goel et~al.(2023)Goel, Pavlakos, Rajasegaran, Kanazawa, and Malik]{goel2023humans}
Shubham Goel, Georgios Pavlakos, Jathushan Rajasegaran, Angjoo Kanazawa, and Jitendra Malik.
\newblock Humans in 4d: Reconstructing and tracking humans with transformers.
\newblock In \emph{Proceedings of the IEEE/CVF International Conference on Computer Vision}, pages 14783--14794, 2023.

\bibitem[Hampali et~al.(2021)Hampali, Stekovic, Sarkar, Kumar, Fraundorfer, and Lepetit]{hampali2021monte}
Shreyas Hampali, Sinisa Stekovic, Sayan~Deb Sarkar, Chetan~S Kumar, Friedrich Fraundorfer, and Vincent Lepetit.
\newblock Monte carlo scene search for 3d scene understanding.
\newblock In \emph{Proceedings of the IEEE/CVF Conference on Computer Vision and Pattern Recognition}, pages 13804--13813, 2021.

\bibitem[Hassan et~al.(2019{\natexlab{a}})Hassan, Choutas, Tzionas, and Black]{hassan2019resolving}
Mohamed Hassan, Vasileios Choutas, Dimitrios Tzionas, and Michael~J Black.
\newblock Resolving 3d human pose ambiguities with 3d scene constraints.
\newblock In \emph{ICCV}, pages 2282--2292, 2019{\natexlab{a}}.

\bibitem[Hassan et~al.(2019{\natexlab{b}})Hassan, Choutas, Tzionas, and Black]{prox}
Mohamed Hassan, Vasileios Choutas, Dimitrios Tzionas, and Michael~J. Black.
\newblock Resolving {3D} human pose ambiguities with {3D} scene constraints.
\newblock In \emph{International Conference on Computer Vision}, pages 2282--2292, 2019{\natexlab{b}}.

\bibitem[Ho et~al.(2020)Ho, Jain, and Abbeel]{ho2020denoising}
Jonathan Ho, Ajay Jain, and Pieter Abbeel.
\newblock Denoising diffusion probabilistic models.
\newblock \emph{Advances in neural information processing systems}, 33:\penalty0 6840--6851, 2020.

\bibitem[Huang et~al.(2024{\natexlab{a}})Huang, Sun, Wang, Qin, Xiong, Zhang, Wan, Zhang, and Jia]{huang2024_icm}
Shuo Huang, Shikun Sun, Zixuan Wang, Xiaoyu Qin, Yanmin Xiong, Yuan Zhang, Pengfei Wan, Di Zhang, and Jia Jia.
\newblock Placiddreamer: Advancing harmony in text-to-3d generation.
\newblock In \emph{Proceedings of the 32nd ACM International Conference on Multimedia}, page 6880–6889, 2024{\natexlab{a}}.

\bibitem[Huang et~al.(2025{\natexlab{a}})Huang, Frehlich, Xia, Gholami, and Xiao]{huang_telepresence2025}
Xincheng Huang, Dieter Frehlich, Ziyi Xia, Peyman Gholami, and Robert Xiao.
\newblock Gaussiannexus: Room-scale real-time ar/vr telepresence with gaussian splatting.
\newblock In \emph{Proceedings of the 38th Annual ACM Symposium on User Interface Software and Technology}, New York, NY, USA, 2025{\natexlab{a}}. Association for Computing Machinery.

\bibitem[Huang et~al.(2024{\natexlab{b}})Huang, Sun, Yang, Lyu, Cao, and Qi]{HuangGS_2024_CVPR}
Yi-Hua Huang, Yang-Tian Sun, Ziyi Yang, Xiaoyang Lyu, Yan-Pei Cao, and Xiaojuan Qi.
\newblock Sc-gs: Sparse-controlled gaussian splatting for editable dynamic scenes.
\newblock In \emph{Proceedings of the IEEE/CVF Conference on Computer Vision and Pattern Recognition (CVPR)}, pages 4220--4230, 2024{\natexlab{b}}.

\bibitem[Huang et~al.(2025{\natexlab{b}})Huang, Guo, An, Yang, Li, Zou, Liang, Liu, Cao, and Sheng]{huang2025midimultiinstancediffusionsingle}
Zehuan Huang, Yuan-Chen Guo, Xingqiao An, Yunhan Yang, Yangguang Li, Zi-Xin Zou, Ding Liang, Xihui Liu, Yan-Pei Cao, and Lu Sheng.
\newblock Midi: Multi-instance diffusion for single image to 3d scene generation, 2025{\natexlab{b}}.

\bibitem[Jiang et~al.(2020)Jiang, Kolotouros, Pavlakos, Zhou, and Daniilidis]{jiang2020mpshape}
Wen Jiang, Nikos Kolotouros, Georgios Pavlakos, Xiaowei Zhou, and Kostas Daniilidis.
\newblock Coherent reconstruction of multiple humans from a single image.
\newblock In \emph{CVPR}, 2020.

\bibitem[Jiang et~al.(2024)Jiang, Zhang, Gao, Hu, and Yao]{jiang2024consistentd}
Yanqin Jiang, Li Zhang, Jin Gao, Weiming Hu, and Yao Yao.
\newblock Consistent4d: Consistent 360{\textdegree} dynamic object generation from monocular video.
\newblock In \emph{The Twelfth International Conference on Learning Representations}, 2024.

\bibitem[Joo et~al.(2017)Joo, Simon, Li, Liu, Tan, Gui, Banerjee, Godisart, Nabbe, Matthews, Kanade, Nobuhara, and Sheikh]{panoptic}
Hanbyul Joo, Tomas Simon, Xulong Li, Hao Liu, Lei Tan, Lin Gui, Sean Banerjee, Timothy~Scott Godisart, Bart Nabbe, Iain Matthews, Takeo Kanade, Shohei Nobuhara, and Yaser Sheikh.
\newblock Panoptic studio: A massively multiview system for social interaction capture.
\newblock \emph{IEEE Transactions on Pattern Analysis and Machine Intelligence}, 2017.

\bibitem[Joo et~al.(2018)Joo, Simon, and Sheikh]{Joo_2018_CVPR}
Hanbyul Joo, Tomas Simon, and Yaser Sheikh.
\newblock Total capture: A 3d deformation model for tracking faces, hands, and bodies.
\newblock In \emph{Proceedings of the IEEE Conference on Computer Vision and Pattern Recognition (CVPR)}, 2018.

\bibitem[Keetha et~al.(2025)Keetha, M{\"u}ller, Sch{\"o}nberger, Porzi, Zhang, Fischer, Knapitsch, Zauss, Weber, Antunes, et~al.]{keetha2025mapanything}
Nikhil Keetha, Norman M{\"u}ller, Johannes Sch{\"o}nberger, Lorenzo Porzi, Yuchen Zhang, Tobias Fischer, Arno Knapitsch, Duncan Zauss, Ethan Weber, Nelson Antunes, et~al.
\newblock Mapanything: Universal feed-forward metric 3d reconstruction.
\newblock \emph{arXiv preprint arXiv:2509.13414}, 2025.

\bibitem[Kerbl et~al.(2023)Kerbl, Kopanas, Leimk{\"u}hler, and Drettakis]{kerbl20233d}
Bernhard Kerbl, Georgios Kopanas, Thomas Leimk{\"u}hler, and George Drettakis.
\newblock 3d gaussian splatting for real-time radiance field rendering.
\newblock \emph{ACM Trans. Graph.}, 42\penalty0 (4):\penalty0 139--1, 2023.

\bibitem[Kirillov et~al.(2023)Kirillov, Mintun, Ravi, Mao, Rolland, Gustafson, Xiao, Whitehead, Berg, Lo, Doll{\'a}r, and Girshick]{sam}
Alexander Kirillov, Eric Mintun, Nikhila Ravi, Hanzi Mao, Chloe Rolland, Laura Gustafson, Tete Xiao, Spencer Whitehead, Alexander~C. Berg, Wan-Yen Lo, Piotr Doll{\'a}r, and Ross Girshick.
\newblock Segment anything.
\newblock \emph{arXiv:2304.02643}, 2023.

\bibitem[Lan et~al.(2025)Lan, Luo, Hong, Zhou, Chen, Lyu, Yang, Dai, Loy, and Pan]{stream3r2025}
Yushi Lan, Yihang Luo, Fangzhou Hong, Shangchen Zhou, Honghua Chen, Zhaoyang Lyu, Shuai Yang, Bo Dai, Chen~Change Loy, and Xingang Pan.
\newblock Stream3r: Scalable sequential 3d reconstruction with causal transformer.
\newblock 2025.

\bibitem[Lei et~al.(2024)Lei, Wang, Pavlakos, Liu, and Daniilidis]{lei2024gart}
Jiahui Lei, Yufu Wang, Georgios Pavlakos, Lingjie Liu, and Kostas Daniilidis.
\newblock Gart: Gaussian articulated template models.
\newblock In \emph{Proceedings of the IEEE/CVF conference on computer vision and pattern recognition}, pages 19876--19887, 2024.

\bibitem[Lei et~al.(2025)Lei, Weng, Harley, Guibas, and Daniilidis]{Lei_2025_CVPR}
Jiahui Lei, Yijia Weng, Adam~W. Harley, Leonidas Guibas, and Kostas Daniilidis.
\newblock Mosca: Dynamic gaussian fusion from casual videos via 4d motion scaffolds.
\newblock In \emph{Proceedings of the IEEE/CVF Conference on Computer Vision and Pattern Recognition (CVPR)}, pages 6165--6177, 2025.

\bibitem[Leroy et~al.(2024)Leroy, Cabon, and Revaud]{mast3r_eccv24}
Vincent Leroy, Yohann Cabon, and Jerome Revaud.
\newblock Grounding image matching in 3d with mast3r, 2024.

\bibitem[Li et~al.(2025{\natexlab{a}})Li, Zheng, Rupprecht, and Vedaldi]{li2025puppetmasterscalinginteractivevideo}
Ruining Li, Chuanxia Zheng, Christian Rupprecht, and Andrea Vedaldi.
\newblock Puppet-master: Scaling interactive video generation as a motion prior for part-level dynamics, 2025{\natexlab{a}}.

\bibitem[Li et~al.(2021)Li, Takehara, Taketomi, Zheng, and Nießner]{li20214dcomplete}
Yang Li, Hikari Takehara, Takafumi Taketomi, Bo Zheng, and Matthias Nießner.
\newblock 4dcomplete: Non-rigid motion estimation beyond the observable surface.
\newblock \emph{IEEE International Conference on Computer Vision (ICCV)}, 2021.

\bibitem[Li et~al.(2025{\natexlab{b}})Li, Zou, Liu, Wang, Liang, Yu, Liu, Guo, Liang, Ouyang, and Cao]{li2025triposghighfidelity3dshape}
Yangguang Li, Zi-Xin Zou, Zexiang Liu, Dehu Wang, Yuan Liang, Zhipeng Yu, Xingchao Liu, Yuan-Chen Guo, Ding Liang, Wanli Ouyang, and Yan-Pei Cao.
\newblock Triposg: High-fidelity 3d shape synthesis using large-scale rectified flow models, 2025{\natexlab{b}}.

\bibitem[Li et~al.(2025{\natexlab{c}})Li, Tucker, Cole, Wang, Jin, Ye, Kanazawa, Holynski, and Snavely]{Li_2025_CVPR}
Zhengqi Li, Richard Tucker, Forrester Cole, Qianqian Wang, Linyi Jin, Vickie Ye, Angjoo Kanazawa, Aleksander Holynski, and Noah Snavely.
\newblock Megasam: Accurate, fast and robust structure and motion from casual dynamic videos.
\newblock In \emph{Proceedings of the Computer Vision and Pattern Recognition Conference (CVPR)}, pages 10486--10496, 2025{\natexlab{c}}.

\bibitem[Li et~al.(2025{\natexlab{d}})Li, Yu, Liu, Yang, Herrmann, Wetzstein, and Wu]{li2025wonderplay}
Zizhang Li, Hong-Xing Yu, Wei Liu, Yin Yang, Charles Herrmann, Gordon Wetzstein, and Jiajun Wu.
\newblock Wonderplay: Dynamic 3d scene generation from a single image and actions.
\newblock 2025{\natexlab{d}}.

\bibitem[Liang et~al.(2024)Liang, Yin, Xu, Liang, Wang, Plataniotis, Zhao, and Wei]{liang2024diffusion4d}
Hanwen Liang, Yuyang Yin, Dejia Xu, Hanxue Liang, Zhangyang Wang, Konstantinos~N Plataniotis, Yao Zhao, and Yunchao Wei.
\newblock Diffusion4d: Fast spatial-temporal consistent 4d generation via video diffusion models.
\newblock \emph{arXiv preprint arXiv:2405.16645}, 2024.

\bibitem[Lin et~al.(2025)Lin, Lin, Pan, Yan, Feng, Mu, and Fragkiadaki]{lin2025partcrafterstructured3dmesh}
Yuchen Lin, Chenguo Lin, Panwang Pan, Honglei Yan, Yiqiang Feng, Yadong Mu, and Katerina Fragkiadaki.
\newblock Partcrafter: Structured 3d mesh generation via compositional latent diffusion transformers, 2025.

\bibitem[Lipman et~al.(2023)Lipman, Chen, Ben-Hamu, Nickel, and Le]{lipman2023flow}
Yaron Lipman, Ricky T.~Q. Chen, Heli Ben-Hamu, Maximilian Nickel, and Matthew Le.
\newblock Flow matching for generative modeling.
\newblock In \emph{The Eleventh International Conference on Learning Representations}, 2023.

\bibitem[Liu et~al.(2023)Liu, Wu, Van~Hoorick, Tokmakov, Zakharov, and Vondrick]{liu2023zero}
Ruoshi Liu, Rundi Wu, Basile Van~Hoorick, Pavel Tokmakov, Sergey Zakharov, and Carl Vondrick.
\newblock Zero-1-to-3: Zero-shot one image to 3d object.
\newblock In \emph{Proceedings of the IEEE/CVF international conference on computer vision}, pages 9298--9309, 2023.

\bibitem[Liu et~al.(2022)Liu, Gong, and Liu]{liu2022flow}
Xingchao Liu, Chengyue Gong, and Qiang Liu.
\newblock Flow straight and fast: Learning to generate and transfer data with rectified flow.
\newblock \emph{arXiv preprint arXiv:2209.03003}, 2022.

\bibitem[Liu et~al.(2025{\natexlab{a}})Liu, Lin, Wu, and Zhou]{liu2025joint}
Zhizheng Liu, Joe Lin, Wayne Wu, and Bolei Zhou.
\newblock Joint optimization for 4d human-scene reconstruction in the wild.
\newblock \emph{arXiv preprint arXiv:2501.02158}, 2025{\natexlab{a}}.

\bibitem[Liu et~al.(2025{\natexlab{b}})Liu, Ye, Luximon, Wan, and Zhang]{ZhuomanLiu_2025_CVPR}
Zhuoman Liu, Weicai Ye, Yan Luximon, Pengfei Wan, and Di Zhang.
\newblock Unleashing the potential of multi-modal foundation models and video diffusion for 4d dynamic physical scene simulation.
\newblock In \emph{Proceedings of the IEEE/CVF Conference on Computer Vision and Pattern Recognition (CVPR)}, pages 11016--11025, 2025{\natexlab{b}}.

\bibitem[Loper et~al.(2015)Loper, Mahmood, Romero, Pons-Moll, and Black]{loper2015smpl}
Matthew Loper, Naureen Mahmood, Javier Romero, Gerard Pons-Moll, and Michael~J. Black.
\newblock {SMPL}: A skinned multi-person linear model.
\newblock \emph{ACM Trans. Graphics (Proc. SIGGRAPH Asia)}, 34\penalty0 (6):\penalty0 248:1--248:16, 2015.

\bibitem[Meng et~al.(2025)Meng, Wu, Zhang, and Xie]{meng2025scenegen}
Yanxu Meng, Haoning Wu, Ya Zhang, and Weidi Xie.
\newblock Scenegen: Single-image 3d scene generation in one feedforward pass.
\newblock 2025.

\bibitem[Mildenhall et~al.(2021)Mildenhall, Srinivasan, Tancik, Barron, Ramamoorthi, and Ng]{mildenhall2021nerf}
Ben Mildenhall, Pratul~P Srinivasan, Matthew Tancik, Jonathan~T Barron, Ravi Ramamoorthi, and Ren Ng.
\newblock Nerf: Representing scenes as neural radiance fields for view synthesis.
\newblock \emph{Communications of the ACM}, 65\penalty0 (1):\penalty0 99--106, 2021.

\bibitem[Nist{\'e}r(2004)]{nister2004efficient}
David Nist{\'e}r.
\newblock An efficient solution to the five-point relative pose problem.
\newblock \emph{IEEE transactions on pattern analysis and machine intelligence}, 26\penalty0 (6):\penalty0 756--770, 2004.

\bibitem[Novotny et~al.(2019)Novotny, Ravi, Graham, Neverova, and Vedaldi]{novotny2019c3dpo}
David Novotny, Nikhila Ravi, Benjamin Graham, Natalia Neverova, and Andrea Vedaldi.
\newblock C3dpo: Canonical 3d pose networks for non-rigid structure from motion.
\newblock In \emph{Proceedings of the IEEE International Conference on Computer Vision}, 2019.

\bibitem[OpenAI(2025)]{chatgpt}
OpenAI.
\newblock Chatgpt (gpt-5) conversation with the author.
\newblock \url{https://chat.openai.com}, 2025.
\newblock Accessed November 2025.

\bibitem[Oquab et~al.(2024)Oquab, Darcet, Moutakanni, Vo, Szafraniec, Khalidov, Fernandez, HAZIZA, Massa, El-Nouby, Assran, Ballas, Galuba, Howes, Huang, Li, Misra, Rabbat, Sharma, Synnaeve, Xu, Jegou, Mairal, Labatut, Joulin, and Bojanowski]{oquab2024dinov}
Maxime Oquab, Timoth{\'e}e Darcet, Th{\'e}o Moutakanni, Huy~V. Vo, Marc Szafraniec, Vasil Khalidov, Pierre Fernandez, Daniel HAZIZA, Francisco Massa, Alaaeldin El-Nouby, Mido Assran, Nicolas Ballas, Wojciech Galuba, Russell Howes, Po-Yao Huang, Shang-Wen Li, Ishan Misra, Michael Rabbat, Vasu Sharma, Gabriel Synnaeve, Hu Xu, Herve Jegou, Julien Mairal, Patrick Labatut, Armand Joulin, and Piotr Bojanowski.
\newblock {DINO}v2: Learning robust visual features without supervision.
\newblock \emph{Transactions on Machine Learning Research}, 2024.
\newblock Featured Certification.

\bibitem[Paladini et~al.(2009)Paladini, Del~Bue, Stosic, Dodig, Xavier, and Agapito]{paladini2009factorization}
Marco Paladini, Alessio Del~Bue, Marko Stosic, Marija Dodig, Joao Xavier, and Lourdes Agapito.
\newblock Factorization for non-rigid and articulated structure using metric projections.
\newblock In \emph{2009 IEEE Conference on Computer Vision and Pattern Recognition}, pages 2898--2905. IEEE, 2009.

\bibitem[Parashar et~al.(2016)Parashar, Pizarro, and Bartoli]{Parashar_2016_CVPR}
Shaifali Parashar, Daniel Pizarro, and Adrien Bartoli.
\newblock Isometric non-rigid shape-from-motion in linear time.
\newblock In \emph{Proceedings of the IEEE Conference on Computer Vision and Pattern Recognition (CVPR)}, 2016.

\bibitem[Paudel et~al.(2024)Paudel, Khanal, Paudel, Tandukar, and Chhatkuli]{paudel2024ihuman}
Pramish Paudel, Anubhav Khanal, Danda~Pani Paudel, Jyoti Tandukar, and Ajad Chhatkuli.
\newblock ihuman: Instant animatable digital humans from monocular videos.
\newblock In \emph{European Conference on Computer Vision}, pages 304--323. Springer, 2024.

\bibitem[Pavlakos et~al.(2019)Pavlakos, Choutas, Ghorbani, Bolkart, Osman, Tzionas, and Black]{smplx2019}
Georgios Pavlakos, Vasileios Choutas, Nima Ghorbani, Timo Bolkart, Ahmed A.~A. Osman, Dimitrios Tzionas, and Michael~J. Black.
\newblock Expressive body capture: {3D} hands, face, and body from a single image.
\newblock In \emph{Proceedings IEEE Conf. on Computer Vision and Pattern Recognition (CVPR)}, pages 10975--10985, 2019.

\bibitem[Peebles and Xie(2023)]{peebles2023scalable}
William Peebles and Saining Xie.
\newblock Scalable diffusion models with transformers.
\newblock In \emph{Proceedings of the IEEE/CVF international conference on computer vision}, pages 4195--4205, 2023.

\bibitem[Peng et~al.(2024)Peng, Xu, Dong, Wang, Zhang, Shuai, Bao, and Zhou]{peng2024animatable}
Sida Peng, Zhen Xu, Junting Dong, Qianqian Wang, Shangzhan Zhang, Qing Shuai, Hujun Bao, and Xiaowei Zhou.
\newblock Animatable implicit neural representations for creating realistic avatars from videos.
\newblock \emph{TPAMI}, 2024.

\bibitem[Poole et~al.(2023)Poole, Jain, Barron, and Mildenhall]{poole2023dreamfusion}
Ben Poole, Ajay Jain, Jonathan~T. Barron, and Ben Mildenhall.
\newblock Dreamfusion: Text-to-3d using 2d diffusion.
\newblock In \emph{The Eleventh International Conference on Learning Representations}, 2023.

\bibitem[Ren et~al.(2024{\natexlab{a}})Ren, Xie, Mirzaei, Liang, Zeng, Kreis, Liu, Torralba, Fidler, Kim, and Ling]{NEURIPS2024_l4gm}
Jiawei Ren, Kevin Xie, Ashkan Mirzaei, Hanxue Liang, Xiaohui Zeng, Karsten Kreis, Ziwei Liu, Antonio Torralba, Sanja Fidler, Seung~Wook Kim, and Huan Ling.
\newblock L4gm: Large 4d gaussian reconstruction model.
\newblock In \emph{Advances in Neural Information Processing Systems}, pages 56828--56858. Curran Associates, Inc., 2024{\natexlab{a}}.

\bibitem[Ren et~al.(2024{\natexlab{b}})Ren, Xie, Mirzaei, Liang, Zeng, Kreis, Liu, Torralba, Fidler, Kim, and Ling]{ren2024l4gmlarge4dgaussian}
Jiawei Ren, Kevin Xie, Ashkan Mirzaei, Hanxue Liang, Xiaohui Zeng, Karsten Kreis, Ziwei Liu, Antonio Torralba, Sanja Fidler, Seung~Wook Kim, and Huan Ling.
\newblock L4gm: Large 4d gaussian reconstruction model, 2024{\natexlab{b}}.

\bibitem[Salzmann and Fua(2009)]{salzmann2009}
Mathieu Salzmann and Pascal Fua.
\newblock Reconstructing sharply folding surfaces: A convex formulation.
\newblock In \emph{2009 IEEE Conference on Computer Vision and Pattern Recognition}, pages 1054--1061, 2009.

\bibitem[Schonberger and Frahm(2016)]{schonberger2016structure}
Johannes~L Schonberger and Jan-Michael Frahm.
\newblock Structure-from-motion revisited.
\newblock In \emph{Proceedings of the IEEE conference on computer vision and pattern recognition}, pages 4104--4113, 2016.

\bibitem[Shen et~al.(2024)Shen, Pi, Xia, Cen, Peng, Hu, Bao, Hu, and Zhou]{shen2024gvhmr}
Zehong Shen, Huaijin Pi, Yan Xia, Zhi Cen, Sida Peng, Zechen Hu, Hujun Bao, Ruizhen Hu, and Xiaowei Zhou.
\newblock World-grounded human motion recovery via gravity-view coordinates.
\newblock In \emph{SIGGRAPH Asia Conference Proceedings}, 2024.

\bibitem[Shin et~al.(2024)Shin, Kim, Halilaj, and Black]{whamcvpr2024}
Soyong Shin, Juyong Kim, Eni Halilaj, and Michael~J. Black.
\newblock {WHAM}: Reconstructing world-grounded humans with accurate {3D} motion.
\newblock In \emph{IEEE/CVF Conf.~on Computer Vision and Pattern Recognition (CVPR)}, 2024.

\bibitem[Snavely et~al.(2008)Snavely, Seitz, and Szeliski]{snavely2008bundler}
N. Snavely, S.~M. Seitz, and R. Szeliski.
\newblock Modeling the world from {Internet} photo collections.
\newblock \emph{International Journal of Computer Vision}, 80\penalty0 (2):\penalty0 189--210, 2008.

\bibitem[Sun et~al.(2023)Sun, Bao, Liu, Mei, and Black]{TRACE2023}
Yu Sun, Qian Bao, Wu Liu, Tao Mei, and Michael~J. Black.
\newblock {TRACE: 5D Temporal Regression of Avatars with Dynamic Cameras in 3D Environments}.
\newblock In \emph{IEEE/CVF Conf.~on Computer Vision and Pattern Recognition (CVPR)}, 2023.

\bibitem[Torresani et~al.(2003)Torresani, Hertzmann, and Bregler]{NIPS2003_8db92642}
Lorenzo Torresani, Aaron Hertzmann, and Christoph Bregler.
\newblock Learning non-rigid 3d shape from 2d motion.
\newblock In \emph{Advances in Neural Information Processing Systems}. MIT Press, 2003.

\bibitem[Triggs et~al.(1999)Triggs, McLauchlan, Hartley, and Fitzgibbon]{triggs1999bundle}
Bill Triggs, Philip~F McLauchlan, Richard~I Hartley, and Andrew~W Fitzgibbon.
\newblock Bundle adjustment—a modern synthesis.
\newblock In \emph{International workshop on vision algorithms}, pages 298--372. Springer, 1999.

\bibitem[van Rijsbergen(1979)]{vanrijsbergen1979information}
C.~J. van Rijsbergen.
\newblock \emph{Information Retrieval}.
\newblock Butterworth-Heinemann, London, 2 edition, 1979.

\bibitem[Wan et~al.(2025)Wan, Wang, Ai, Wen, Mao, Xie, Chen, Yu, Zhao, Yang, Zeng, Wang, Zhang, Zhou, Wang, Chen, Zhu, Zhao, Yan, Huang, Feng, Zhang, Li, Wu, Chu, Feng, Zhang, Sun, Fang, Wang, Gui, Weng, Shen, Lin, Wang, Wang, Zhou, Wang, Shen, Yu, Shi, Huang, Xu, Kou, Lv, Li, Liu, Wang, Zhang, Huang, Li, Wu, Liu, Pan, Zheng, Hong, Shi, Feng, Jiang, Han, Wu, and Liu]{wan}
Team Wan, Ang Wang, Baole Ai, Bin Wen, Chaojie Mao, Chen-Wei Xie, Di Chen, Feiwu Yu, Haiming Zhao, Jianxiao Yang, Jianyuan Zeng, Jiayu Wang, Jingfeng Zhang, Jingren Zhou, Jinkai Wang, Jixuan Chen, Kai Zhu, Kang Zhao, Keyu Yan, Lianghua Huang, Mengyang Feng, Ningyi Zhang, Pandeng Li, Pingyu Wu, Ruihang Chu, Ruili Feng, Shiwei Zhang, Siyang Sun, Tao Fang, Tianxing Wang, Tianyi Gui, Tingyu Weng, Tong Shen, Wei Lin, Wei Wang, Wei Wang, Wenmeng Zhou, Wente Wang, Wenting Shen, Wenyuan Yu, Xianzhong Shi, Xiaoming Huang, Xin Xu, Yan Kou, Yangyu Lv, Yifei Li, Yijing Liu, Yiming Wang, Yingya Zhang, Yitong Huang, Yong Li, You Wu, Yu Liu, Yulin Pan, Yun Zheng, Yuntao Hong, Yupeng Shi, Yutong Feng, Zeyinzi Jiang, Zhen Han, Zhi-Fan Wu, and Ziyu Liu.
\newblock Wan: Open and advanced large-scale video generative models.
\newblock \emph{arXiv preprint arXiv:2503.20314}, 2025.

\bibitem[Wang et~al.(2025{\natexlab{a}})Wang, Chen, Karaev, Vedaldi, Rupprecht, and Novotny]{wang2025vggt}
Jianyuan Wang, Minghao Chen, Nikita Karaev, Andrea Vedaldi, Christian Rupprecht, and David Novotny.
\newblock Vggt: Visual geometry grounded transformer.
\newblock In \emph{Proceedings of the Computer Vision and Pattern Recognition Conference}, pages 5294--5306, 2025{\natexlab{a}}.

\bibitem[Wang et~al.(2025{\natexlab{b}})Wang, Ye, Gao, Zeng, Austin, Li, and Kanazawa]{som2024}
Qianqian Wang, Vickie Ye, Hang Gao, Weijia Zeng, Jake Austin, Zhengqi Li, and Angjoo Kanazawa.
\newblock Shape of motion: 4d reconstruction from a single video.
\newblock In \emph{International Conference on Computer Vision (ICCV)}, 2025{\natexlab{b}}.

\bibitem[Wang* et~al.(2025)Wang*, Zhang*, Holynski, Efros, and Kanazawa]{cut3r}
Qianqian Wang*, Yifei Zhang*, Aleksander Holynski, Alexei~A. Efros, and Angjoo Kanazawa.
\newblock Continuous 3d perception model with persistent state.
\newblock In \emph{CVPR}, 2025.

\bibitem[Wang et~al.(2024)Wang, Leroy, Cabon, Chidlovskii, and Revaud]{dust3r_cvpr24}
Shuzhe Wang, Vincent Leroy, Yohann Cabon, Boris Chidlovskii, and Jerome Revaud.
\newblock Dust3r: Geometric 3d vision made easy.
\newblock In \emph{CVPR}, 2024.

\bibitem[Wang et~al.(2025)Wang, Jiang, Yang, and Wang]{Wang_2025_C4D_ICCV}
Shizun Wang, Zhenxiang Jiang, Xingyi Yang, and Xinchao Wang.
\newblock C4d: 4d made from 3d through dual correspondences.
\newblock In \emph{Proceedings of the IEEE/CVF International Conference on Computer Vision (ICCV)}, pages 7570--7580, 2025.

\bibitem[Wang et~al.(2023)Wang, Lu, Wang, Bao, Li, Su, and Zhu]{wang2023prolificdreamer}
Zhengyi Wang, Cheng Lu, Yikai Wang, Fan Bao, Chongxuan Li, Hang Su, and Jun Zhu.
\newblock Prolificdreamer: High-fidelity and diverse text-to-3d generation with variational score distillation.
\newblock In \emph{Advances in Neural Information Processing Systems (NeurIPS)}, 2023.

\bibitem[Weng et~al.(2025)Weng, Zhao, Lei, Yang, Liu, Lai, Chen, Liu, Jiang, Guo, et~al.]{weng2025scaling}
Haohan Weng, Zibo Zhao, Biwen Lei, Xianghui Yang, Jian Liu, Zeqiang Lai, Zhuo Chen, Yuhong Liu, Jie Jiang, Chunchao Guo, et~al.
\newblock Scaling mesh generation via compressive tokenization.
\newblock In \emph{CVPR}, pages 11093--11103, 2025.

\bibitem[Wightman(2019)]{rw2019timm}
Ross Wightman.
\newblock Pytorch image models.
\newblock \url{https://github.com/rwightman/pytorch-image-models}, 2019.

\bibitem[Wu et~al.(2024)Wu, Zhang, Chen, Fan, Yan, and Zuo]{Wu2024Deblur4DGS4G}
Ren-Rong Wu, Zhilu Zhang, Mingyang Chen, Xiaopeng Fan, Zifei Yan, and Wangmeng Zuo.
\newblock Deblur4dgs: 4d gaussian splatting from blurry monocular video.
\newblock \emph{ArXiv}, abs/2412.06424, 2024.

\bibitem[Wu et~al.(2023)Wu, Zhang, Fu, Wang, Jiawei~Ren, Wu, Yang, Wang, Qian, Lin, and Liu]{wu2023omniobject3d}
Tong Wu, Jiarui Zhang, Xiao Fu, Yuxin Wang, Liang~Pan Jiawei~Ren, Wayne Wu, Lei Yang, Jiaqi Wang, Chen Qian, Dahua Lin, and Ziwei Liu.
\newblock Omniobject3d: Large-vocabulary 3d object dataset for realistic perception, reconstruction and generation.
\newblock In \emph{IEEE/CVF Conference on Computer Vision and Pattern Recognition (CVPR)}, 2023.

\bibitem[Xiang et~al.(2025)Xiang, Lv, Xu, Deng, Wang, Zhang, Chen, Tong, and Yang]{xiang2025structured}
Jianfeng Xiang, Zelong Lv, Sicheng Xu, Yu Deng, Ruicheng Wang, Bowen Zhang, Dong Chen, Xin Tong, and Jiaolong Yang.
\newblock Structured 3d latents for scalable and versatile 3d generation.
\newblock In \emph{Proceedings of the Computer Vision and Pattern Recognition Conference}, pages 21469--21480, 2025.

\bibitem[Xu et~al.(2023)Xu, Agrawal, Laney, Garcia, Bansal, Kim, Rota~Bulò, Porzi, Kontschieder, Božič, Lin, Zollhöfer, and Richardt]{VRNeRF2023}
Linning Xu, Vasu Agrawal, William Laney, Tony Garcia, Aayush Bansal, Changil Kim, Samuel Rota~Bulò, Lorenzo Porzi, Peter Kontschieder, Aljaž Božič, Dahua Lin, Michael Zollhöfer, and Christian Richardt.
\newblock {VR-NeRF}: High-fidelity virtualized walkable spaces.
\newblock In \emph{SIGGRAPH Asia Conference Proceedings}, 2023.

\bibitem[Yuan et~al.(2025)Yuan, Chen, Yi, and Gao]{SelfYuan_2025_ICCV}
Chengbo Yuan, Geng Chen, Li Yi, and Yang Gao.
\newblock Self-supervised monocular 4d scene reconstruction for egocentric videos.
\newblock In \emph{Proceedings of the IEEE/CVF International Conference on Computer Vision (ICCV)}, pages 8863--8874, 2025.

\bibitem[Zhang et~al.(2023)Zhang, Tang, Niessner, and Wonka]{zhang20233dshape2vecset}
Biao Zhang, Jiapeng Tang, Matthias Niessner, and Peter Wonka.
\newblock 3dshape2vecset: A 3d shape representation for neural fields and generative diffusion models.
\newblock \emph{ACM Transactions On Graphics (TOG)}, 42\penalty0 (4):\penalty0 1--16, 2023.

\bibitem[Zhang et~al.(2024{\natexlab{a}})Zhang, Cheng, Yang, Wang, Zhao, Tang, Chen, and Guo]{zhang2024gaussiancube}
Bowen Zhang, Yiji Cheng, Jiaolong Yang, Chunyu Wang, Feng Zhao, Yansong Tang, Dong Chen, and Baining Guo.
\newblock Gaussiancube: A structured and explicit radiance representation for 3d generative modeling.
\newblock \emph{arXiv preprint arXiv:2403.19655}, 2024{\natexlab{a}}.

\bibitem[Zhang et~al.(2025)Zhang, Xu, Wang, Yang, Zhao, Chen, and Guo]{zhang2025gaussianvariationfielddiffusion}
Bowen Zhang, Sicheng Xu, Chuxin Wang, Jiaolong Yang, Feng Zhao, Dong Chen, and Baining Guo.
\newblock Gaussian variation field diffusion for high-fidelity video-to-4d synthesis, 2025.

\bibitem[Zhang et~al.(2024{\natexlab{b}})Zhang, Herrmann, Hur, Jampani, Darrell, Cole, Sun, and Yang]{zhang2024monst3r}
Junyi Zhang, Charles Herrmann, Junhwa Hur, Varun Jampani, Trevor Darrell, Forrester Cole, Deqing Sun, and Ming-Hsuan Yang.
\newblock Monst3r: A simple approach for estimating geometry in the presence of motion.
\newblock \emph{arXiv preprint arxiv:2410.03825}, 2024{\natexlab{b}}.

\bibitem[Zhang et~al.(2024{\natexlab{c}})Zhang, Wang, Zhang, Qiu, Pang, Jiang, Yang, Xu, and Yu]{zhang2024clay}
Longwen Zhang, Ziyu Wang, Qixuan Zhang, Qiwei Qiu, Anqi Pang, Haoran Jiang, Wei Yang, Lan Xu, and Jingyi Yu.
\newblock Clay: A controllable large-scale generative model for creating high-quality 3d assets.
\newblock \emph{ACM Transactions on Graphics (TOG)}, 43\penalty0 (4):\penalty0 1--20, 2024{\natexlab{c}}.

\bibitem[Zhao et~al.(2024)Zhao, Colburn, Ma, Ángel Bautista, Susskind, and Schwing]{zhao2024pgdvs}
Xiaoming Zhao, Alex Colburn, Fangchang Ma, Miguel Ángel Bautista, Joshua~M. Susskind, and Alexander~G. Schwing.
\newblock {Pseudo-Generalized Dynamic View Synthesis from a Video}.
\newblock In \emph{ICLR}, 2024.

\bibitem[Zhao et~al.(2025)Zhao, Lai, Lin, Zhao, Liu, Yang, Feng, Yang, Zhang, Yang, et~al.]{zhao2025hunyuan3d}
Zibo Zhao, Zeqiang Lai, Qingxiang Lin, Yunfei Zhao, Haolin Liu, Shuhui Yang, Yifei Feng, Mingxin Yang, Sheng Zhang, Xianghui Yang, et~al.
\newblock Hunyuan3d 2.0: Scaling diffusion models for high resolution textured 3d assets generation.
\newblock \emph{arXiv preprint arXiv:2501.12202}, 2025.

\bibitem[Zuffi et~al.(2017)Zuffi, Kanazawa, Jacobs, and Black]{zuffi20173d}
Silvia Zuffi, Angjoo Kanazawa, David~W Jacobs, and Michael~J Black.
\newblock 3d menagerie: Modeling the 3d shape and pose of animals.
\newblock In \emph{Proceedings of the IEEE conference on computer vision and pattern recognition}, pages 6365--6373, 2017.

\end{thebibliography}
}
\clearpage
\setcounter{page}{1}
\maketitlesupplementary
\setcounter{section}{0}
\renewcommand{\thesection}{\Alph{section}}
\renewcommand{\thesubsection}{\thesection.\arabic{subsection}}
\renewcommand{\thesubsubsection}{\thesubsection.\arabic{subsubsection}}

\section{Training Details}
\label{sec:training_details}
We provide details of the full training losses described in \cref{sec:attention_parsing}. Specifically, eq.~\eqref{eq:static_loss} provides the equation for $\mathcal{L}_S$ and thereafter describes $\mathcal{L}_T$ in text. Here, we write $\mathcal{L}_T$ formally:
\begin{equation}
\label{eq:dynamic_loss}
\mathcal{L}_T = \mathbb{E} \left[ \sum_{f=1}^{F} \left\| ({{}^f}\!\boldsymbol{\epsilon} - {{}^f}\!\mathbf{z}_0) - \mathbf{v}_\theta({{}^f}\!\mathbf{z}_{t_f}, t_f, {{}^f}\!\mathbf{y}) \right\|^2 \right].
\end{equation}

The above expression in eq.~\eqref{eq:dynamic_loss} for the temporal loss is almost identical to the spatial expression in eq.~\eqref{eq:static_loss}. The obvious change is the the frame index $f$ replacing the object index $i$, as each data: $({{}^f}\!\mathbf{z}_0, {{}^f}\!\mathbf{y})$ is sampled from DeformingThings~\cite{li20214dcomplete}. Additionally, the conditional image embeddings ${{}^f}\!\mathbf{y}$ are separate for each frame among $F$, unlike in the spatial loss expression, where all $N$ objects use the same image embedding $\mathbf{y}$.

Finally for completeness, we formally write the regularization loss, \ie, the TripoSG~\cite{li2025triposghighfidelity3dshape} loss\footnote{The training loss for DiT is not mentioned explicitly in the reference. Please refer to Algorithm 1 in \cite{liu2022flow} for the rectified flow loss.} as follows:
\begin{equation}
\label{eq:regularization}
\mathcal{L}_R = \mathbb{E} \left[ \left\| (\boldsymbol{\epsilon} - \mathbf{z}_0) - \mathbf{v}_\theta(\mathbf{z}_{t}, t, \mathbf{y}) \right\|^2 \right].
\end{equation}
Note that, each sample in eq.~\eqref{eq:regularization} only consists of one object with no temporal evolution. Samples $(\mathbf{z}_0, \mathbf{y})$ are obtained from the Objaverse training set~\cite{deitke2022objaverseuniverseannotated3d}.
Finally, the overall loss is $\mathcal{L}_{S/T/R}$: $\mathcal{L}_S, \mathcal{L}_T, \mathcal{L}_R$ are sampled with a ratio of $0.35: 0.35: 0.3$ respectively. The regularization and its sampling ratio of 0.3 is also used by PartCrafter~\cite{lin2025partcrafterstructured3dmesh} and MIDI~\cite{huang2025midimultiinstancediffusionsingle}.

\section{User Study Details}
\label{sec:supp_user_study}

To quantitatively evaluate our method's perceptual quality, we conducted a user preference study. The study was administered via Google Forms and compared our full model (with Attention Mixing) against an ablation baseline (without Attention Mixing).

\begin{figure}[h]
  \centering
  \includegraphics[width=0.9\linewidth]{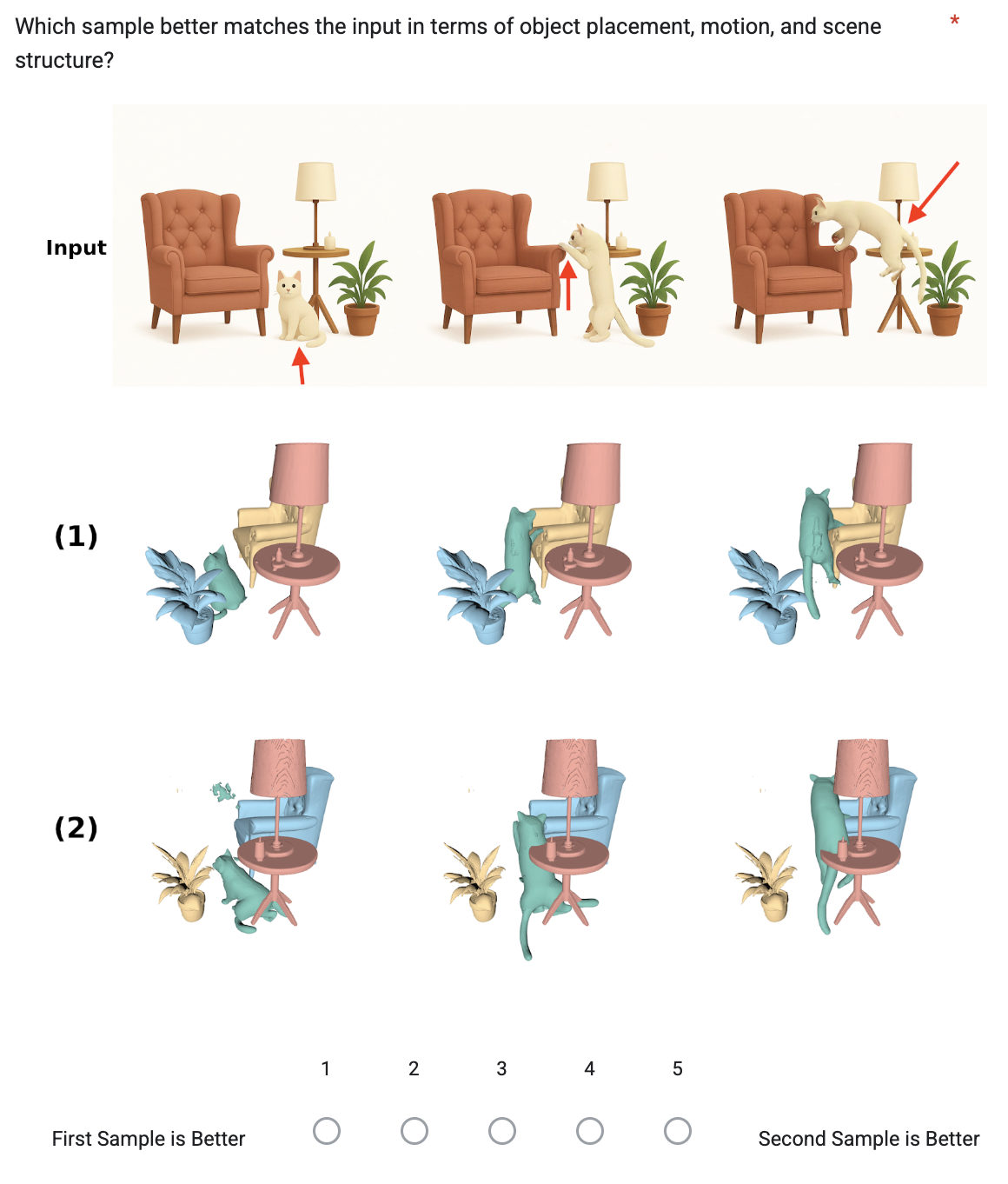} 
  \caption{\textbf{Our user study interface, run on Google Forms.} Participants viewed a 2D input sequence (top) and two 3D results (middle, bottom), then rated their preference on a 5-point scale.}
  \label{fig:supp_ui}
\end{figure}

\paragraph{Procedure.}
As shown in \cref{fig:supp_ui}, participants were presented with a 2D input sequence indicating intended object motion and two corresponding 3D animated samples (labeled (1) and (2)). For each comparison, they answered the question: \textit{"Which sample better matches the input in terms of object placement, motion, and scene structure?"} by rating their preference on a 5-point Likert scale (1: Sample 1 is better, 3: Both are about the same, 5: Sample 2 is better). To prevent bias, the assignment of our method to Sample (1) or (2) and the order of the scenes were randomized for each participant.

\paragraph{Data Analysis.}
All ratings were included in our final analysis. The preference scores reported in the main paper show the complete distribution of judgments across the 5-point Likert scale.

\section{Evaluation on CMU Panoptic Dataset}
\label{sec:supp_cmu_eval}

To quantitatively assess the temporal consistency and motion accuracy of our model, we performed an evaluation on the CMU Panoptic dataset~\cite{panoptic}. This section details our protocol for preparing the data and computing the metrics.

\paragraph{Ground Truth Point Cloud Generation.}
We first generated ground truth (GT) point clouds from the raw RGB-D Kinect data provided in the dataset. To ensure a clean and fair comparison, we pre-processed these GT clouds in two steps:
\begin{enumerate}
    \item \textbf{Ground Removal:} The ground plane was removed using a simple height threshold.
    \item \textbf{Denoising:} We applied a statistical outlier removal filter to eliminate stray, floating points in the cloud.
\end{enumerate}

\paragraph{Alignment and Metric Computation.}
A key challenge in evaluating generative models is that their outputs are not inherently aligned with the GT coordinate system; they may have an arbitrary scale, rotation, and translation. To address this, we adopted a first-frame alignment protocol.

For each sequence, we independently registered the initial generated mesh (frame 1) from both our full model and the baseline (without Attention Mixing) to the corresponding ground truth point cloud. This one-time alignment transformation (capturing scale, rotation, and translation) was then applied uniformly to all subsequent frames generated by that method for the entire sequence.

\begin{figure}[h!]
  \centering
  % --- MODIFICATIONS ---
  \setlength{\tabcolsep}{0pt}      % NO horizontal space between columns
  \renewcommand{\arraystretch}{0} % NO vertical stretching of rows

  % ==================== TABLE START ====================
  \begin{tabular}{@{}c@{}ccccc@{}} % Use @{} to remove all column padding

    % --- Row 1: GT Frames ---
    \rotatebox{90}{\small ~~~Input} & % Rotated label with manual spacing
    \includegraphics[width=0.18\linewidth]{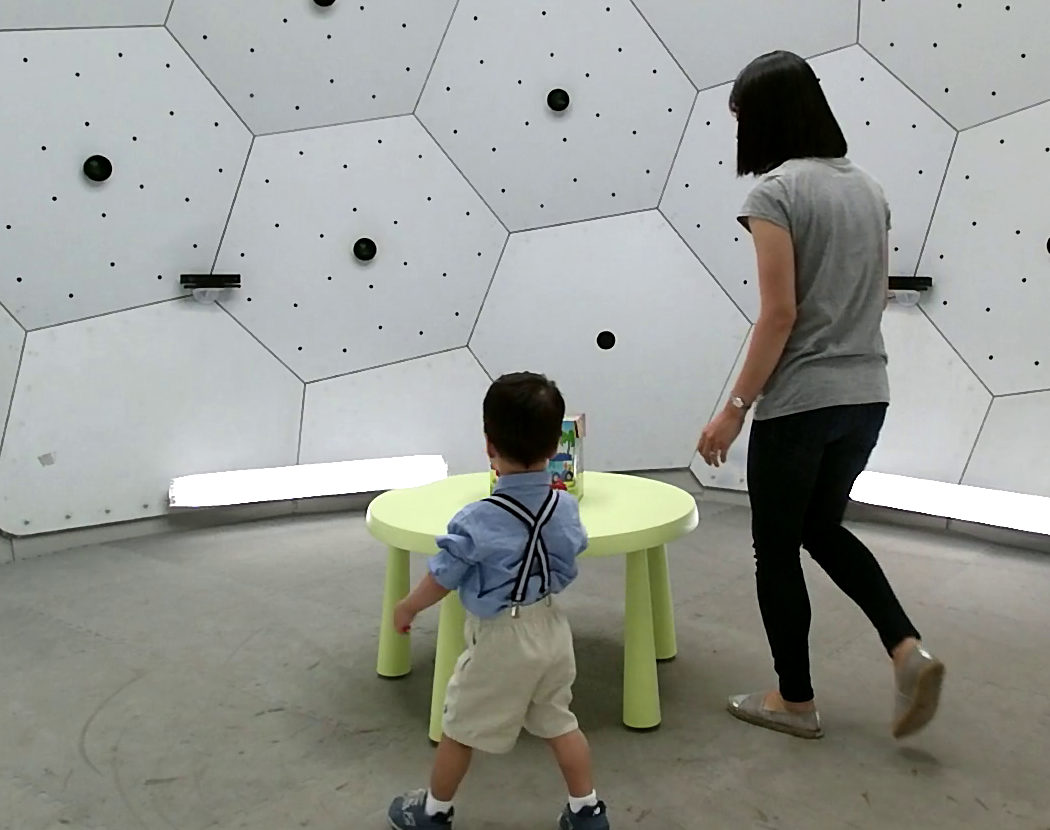} &
    \includegraphics[width=0.18\linewidth]{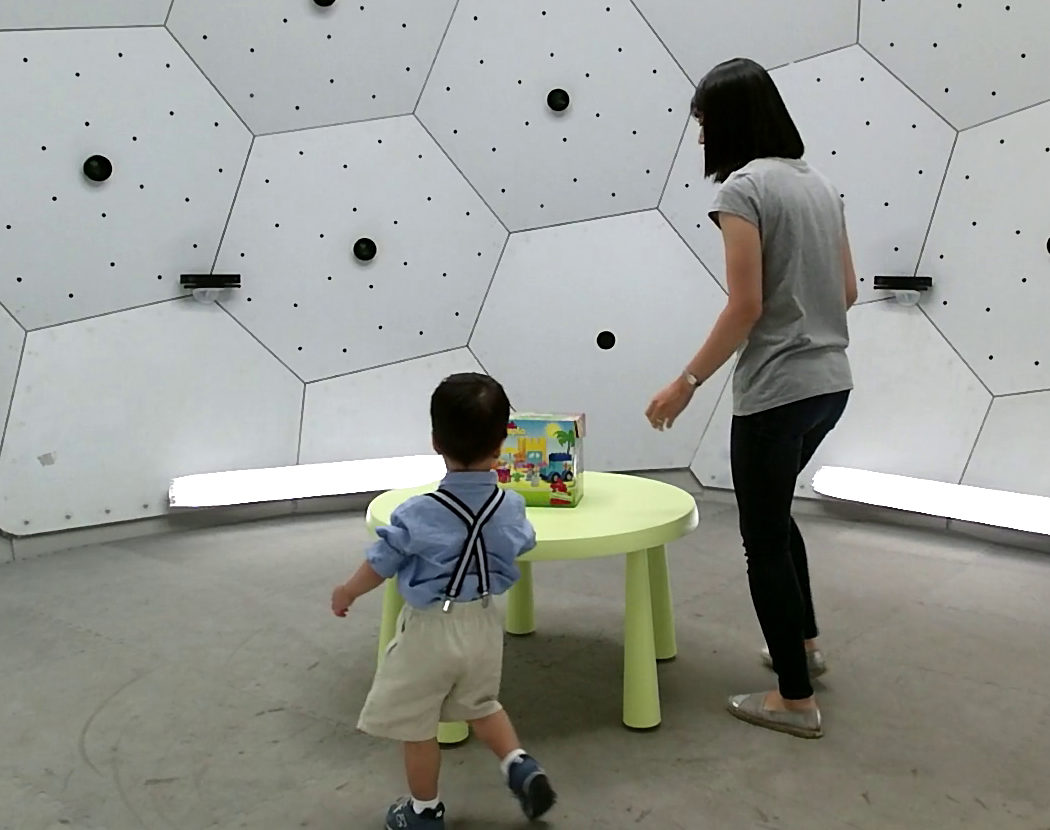} &
    \includegraphics[width=0.18\linewidth]{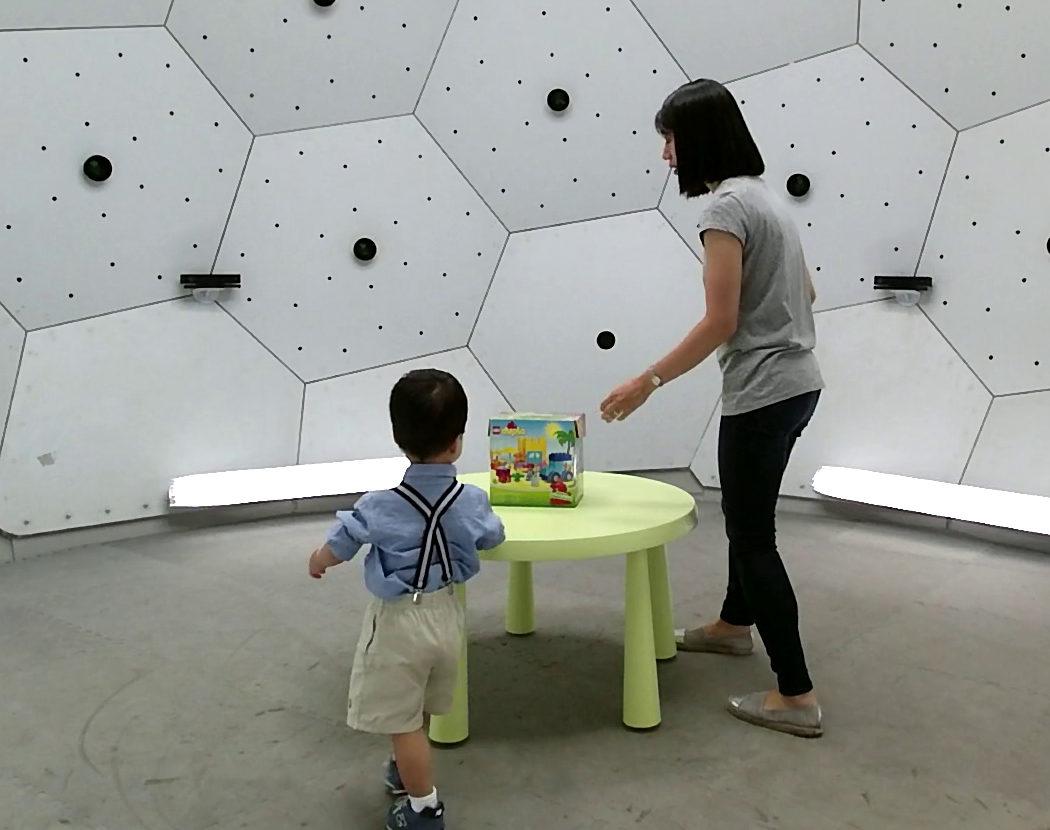} &
    \includegraphics[width=0.18\linewidth]{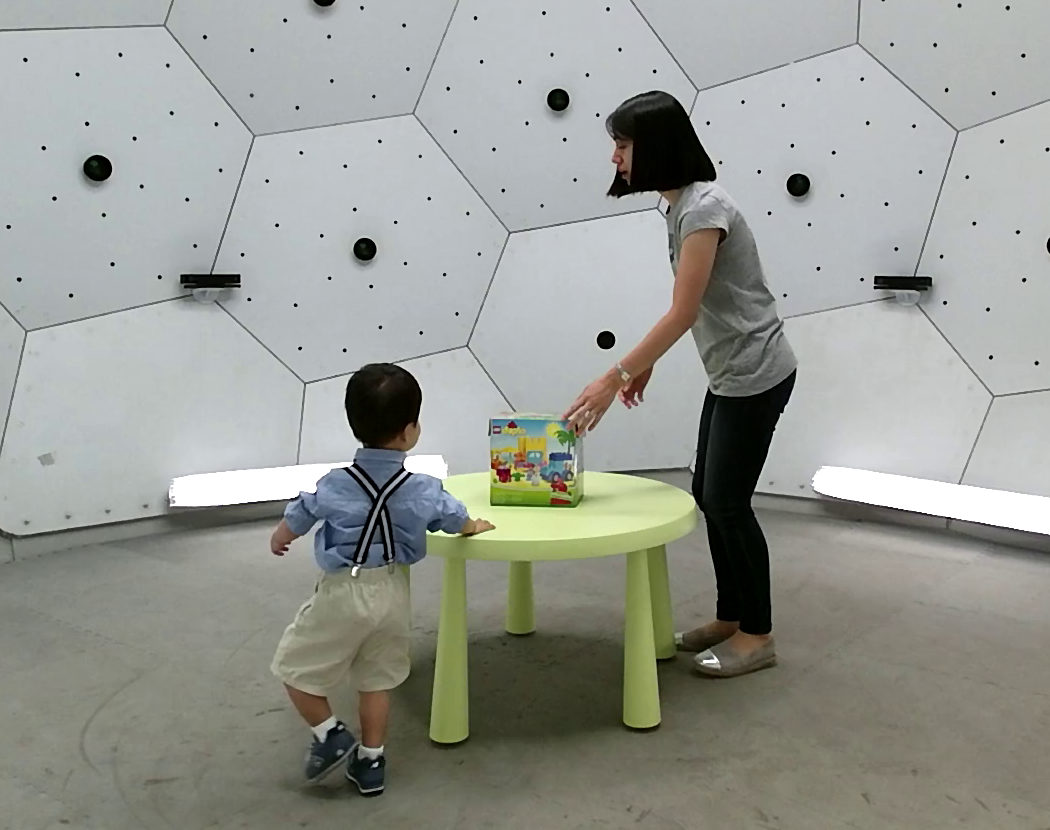} &
    \includegraphics[width=0.18\linewidth]{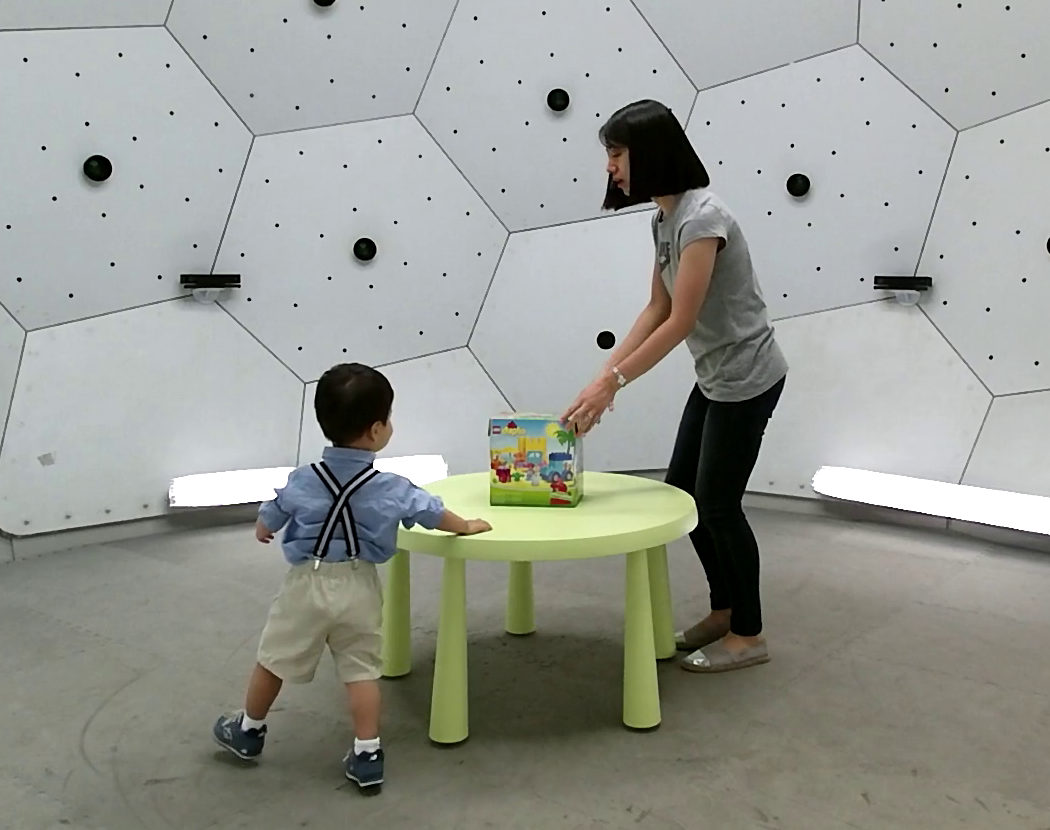} \\

    % --- Row 2: w/ Mixing View 1 ---
    \rotatebox{90}{\small w/ Mixing} &
    \includegraphics[width=0.18\linewidth]{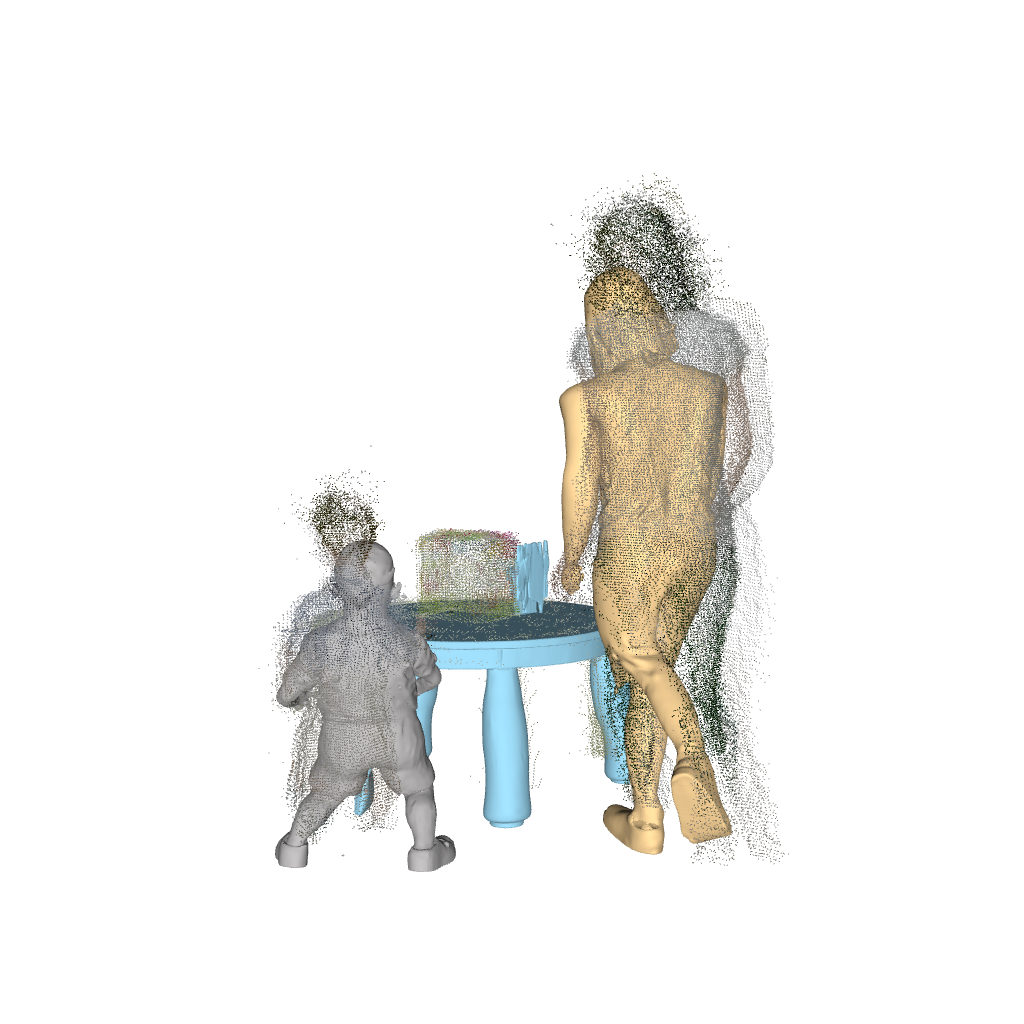} &
    \includegraphics[width=0.18\linewidth]{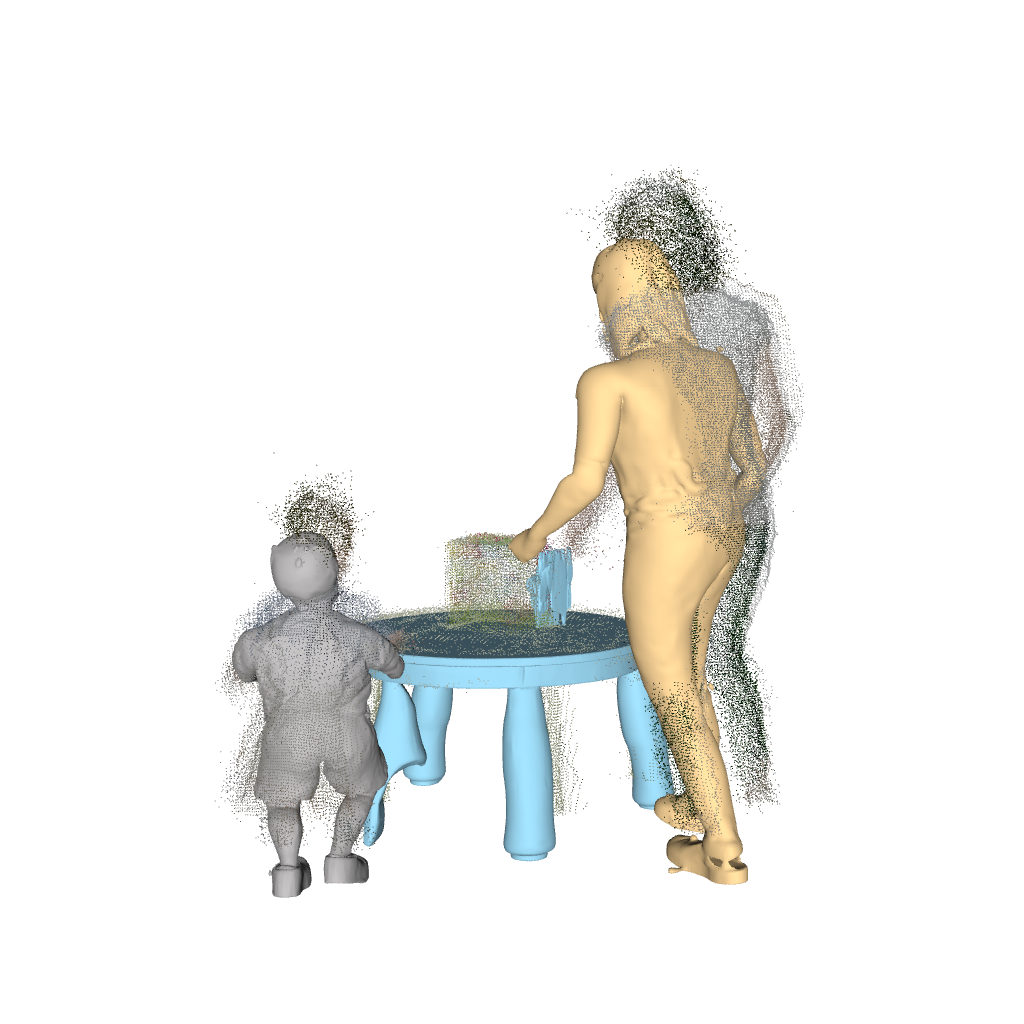} &
    \includegraphics[width=0.18\linewidth]{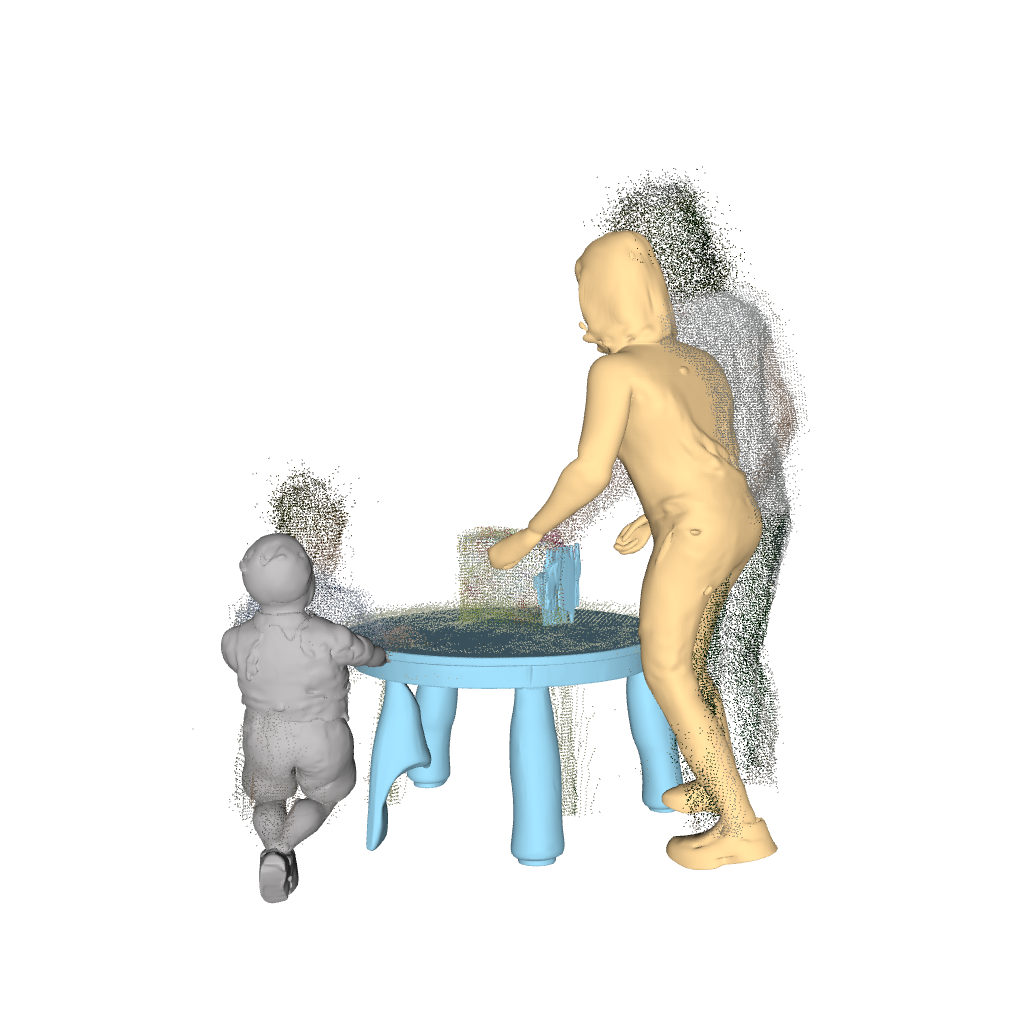} &
    \includegraphics[width=0.18\linewidth]{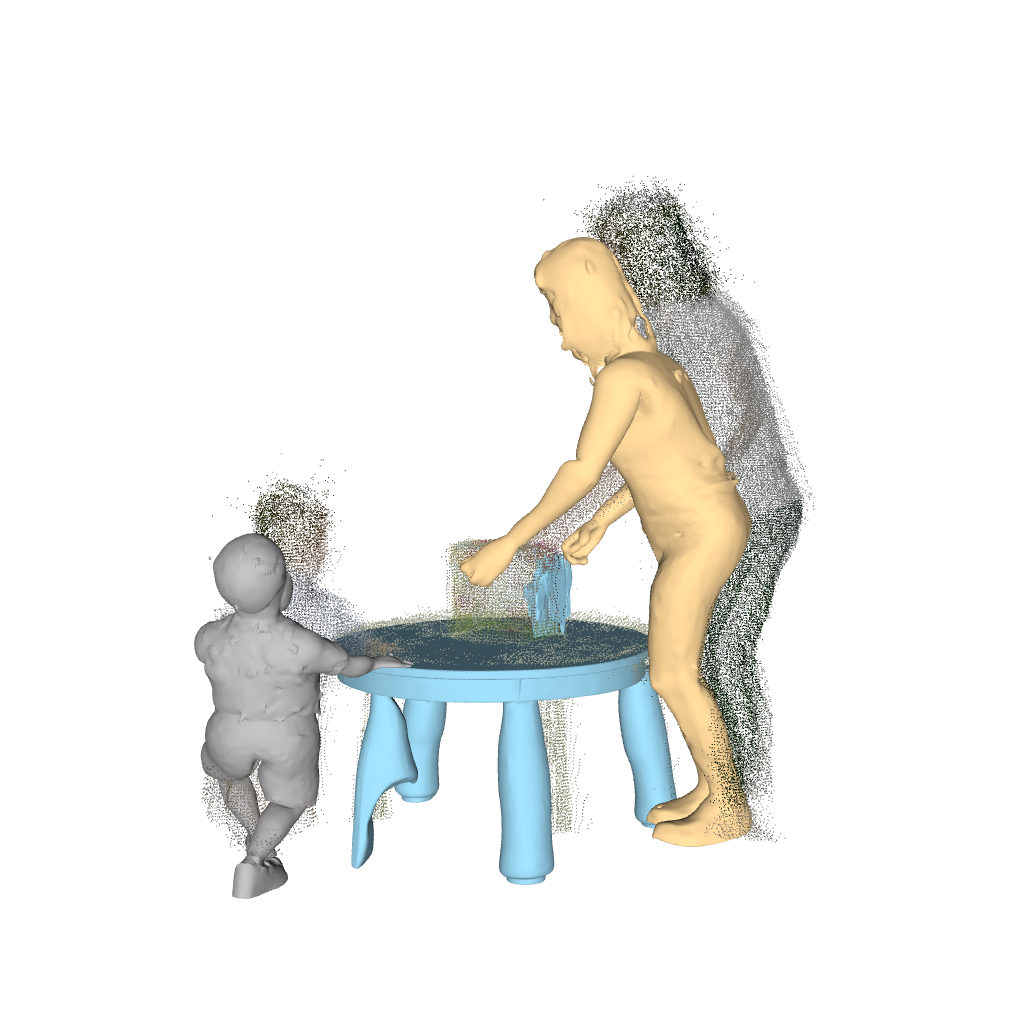} &
    \includegraphics[width=0.18\linewidth]{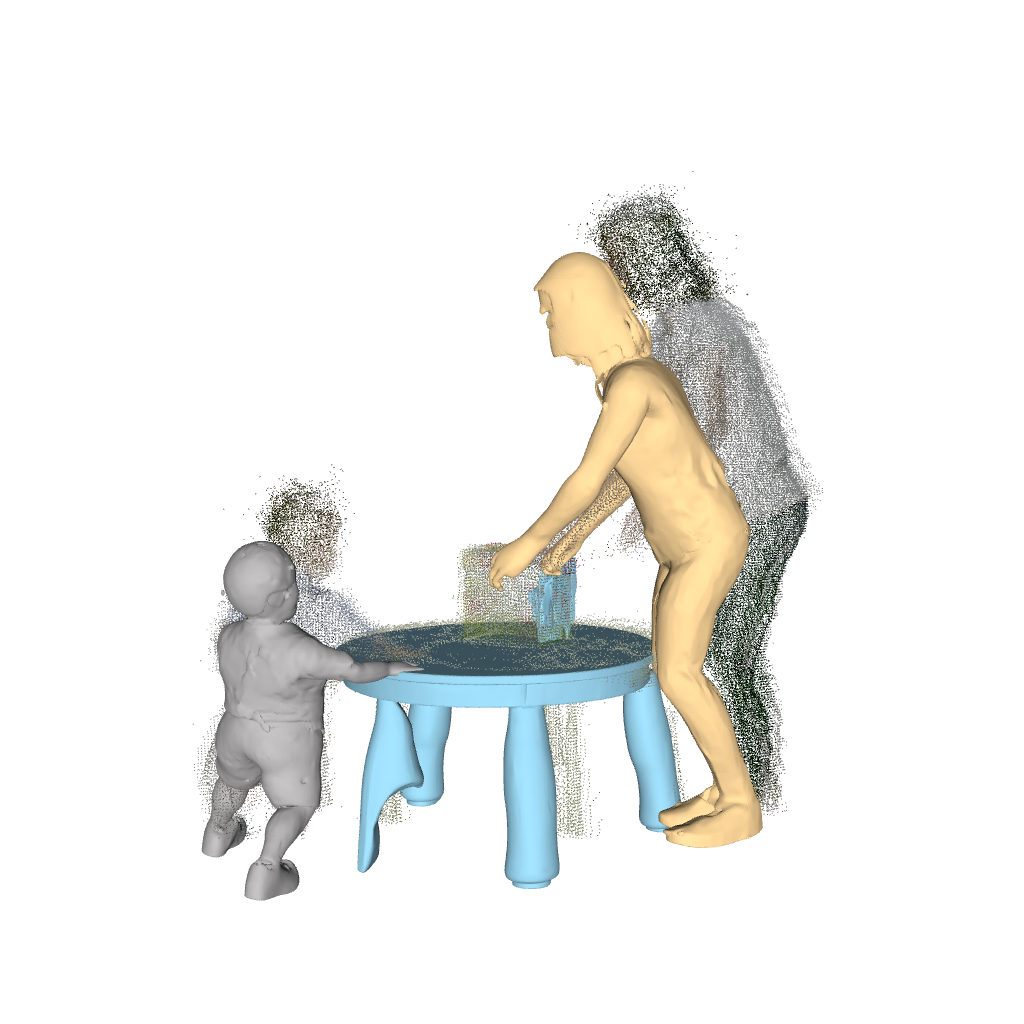} \\

    % --- Row 3: w/ Mixing View 2 ---
    \rotatebox{90}{\small} &
    \includegraphics[width=0.18\linewidth]{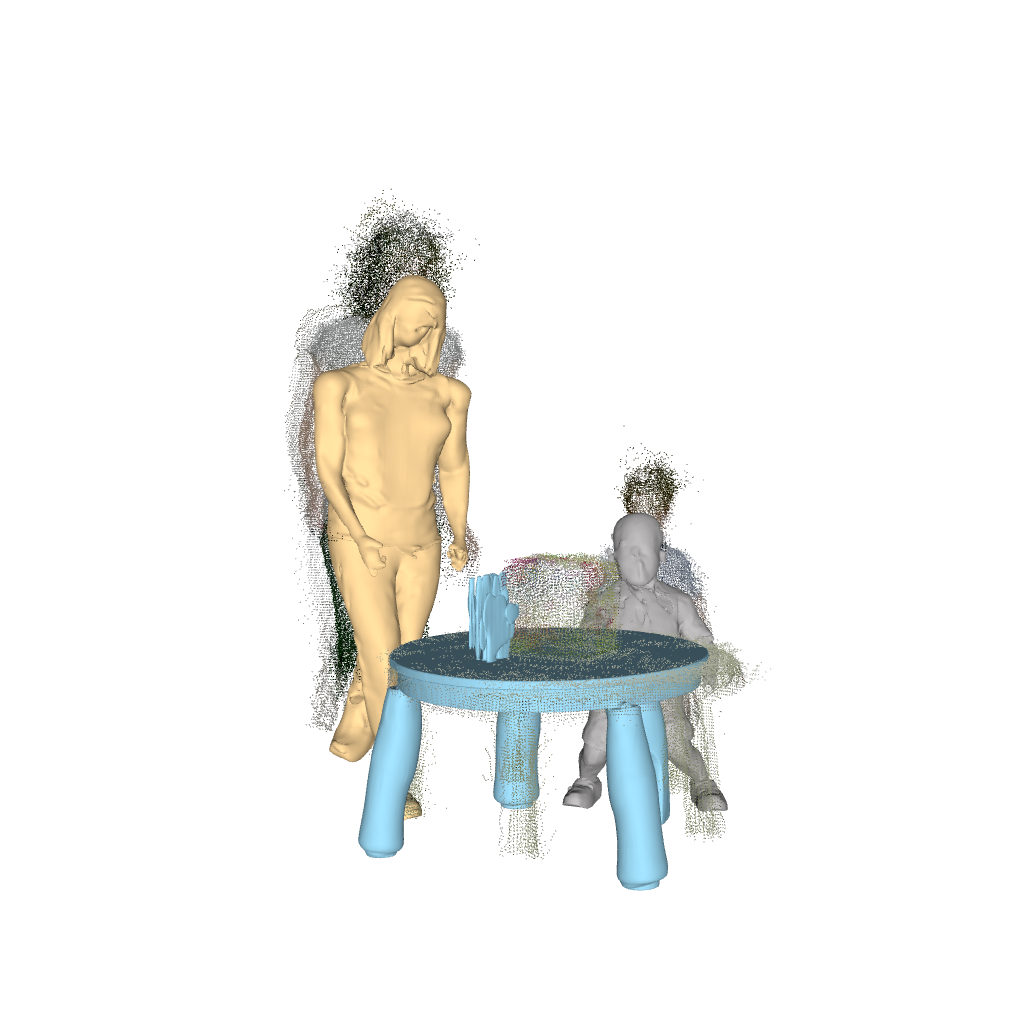} &
    \includegraphics[width=0.18\linewidth]{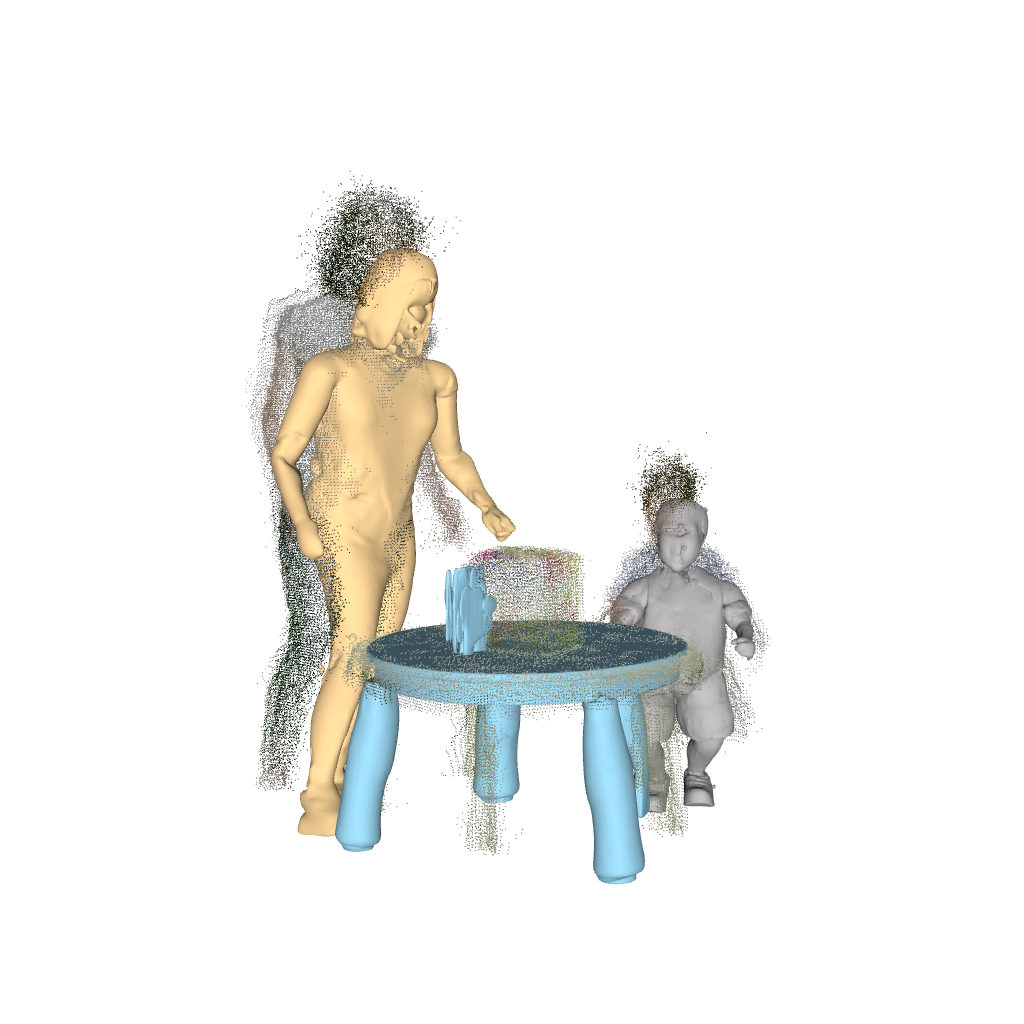} &
    \includegraphics[width=0.18\linewidth]{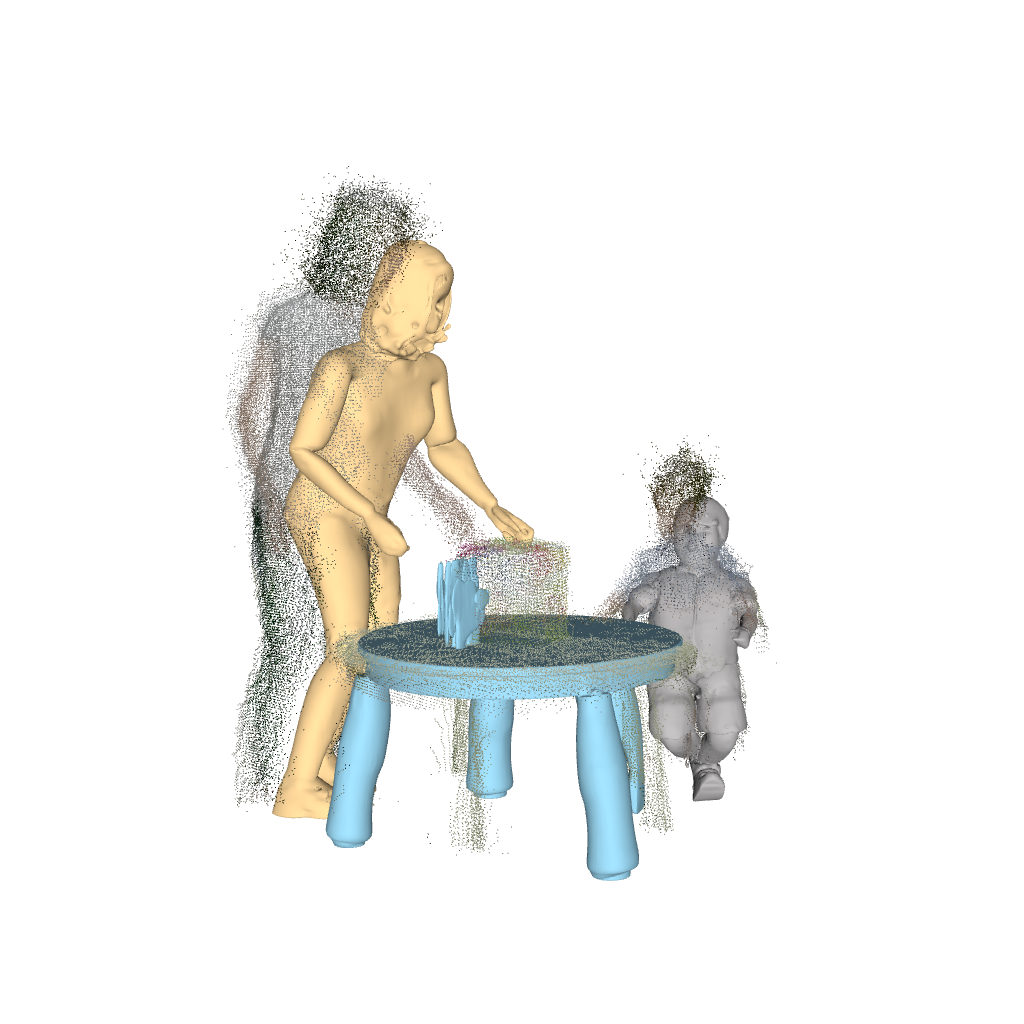} &
    \includegraphics[width=0.18\linewidth]{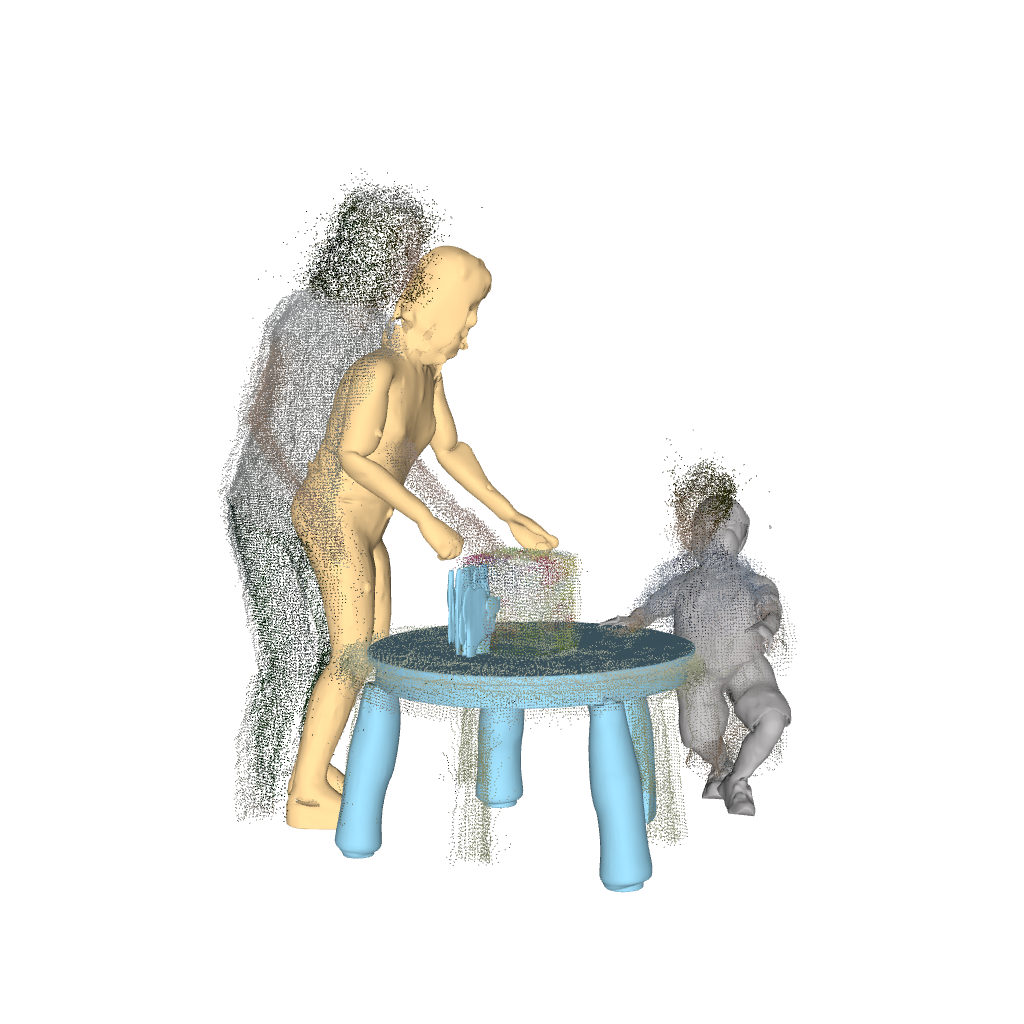} &
    \includegraphics[width=0.18\linewidth]{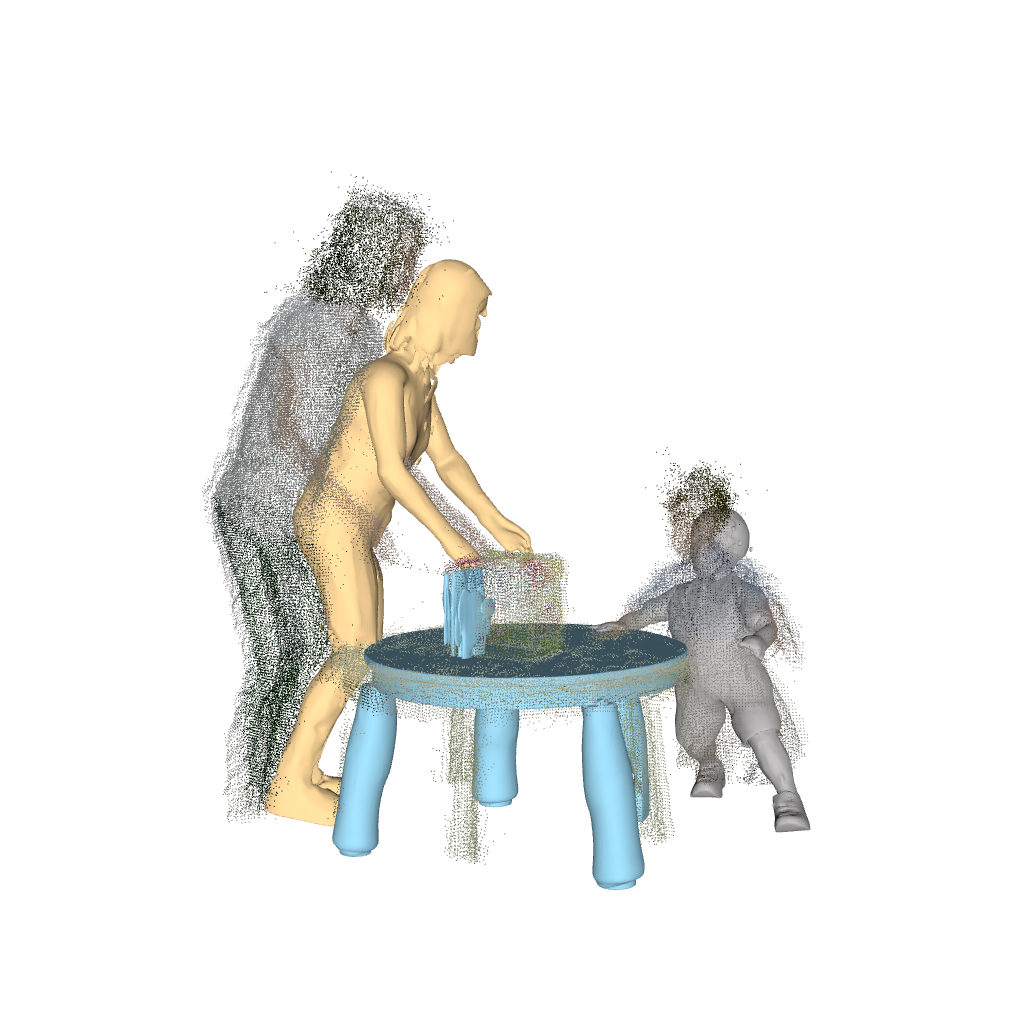} \\

    % --- Row 4: w/o Mixing View 1 ---
    \rotatebox{90}{\small w/o Mixing} &
    \includegraphics[width=0.18\linewidth]{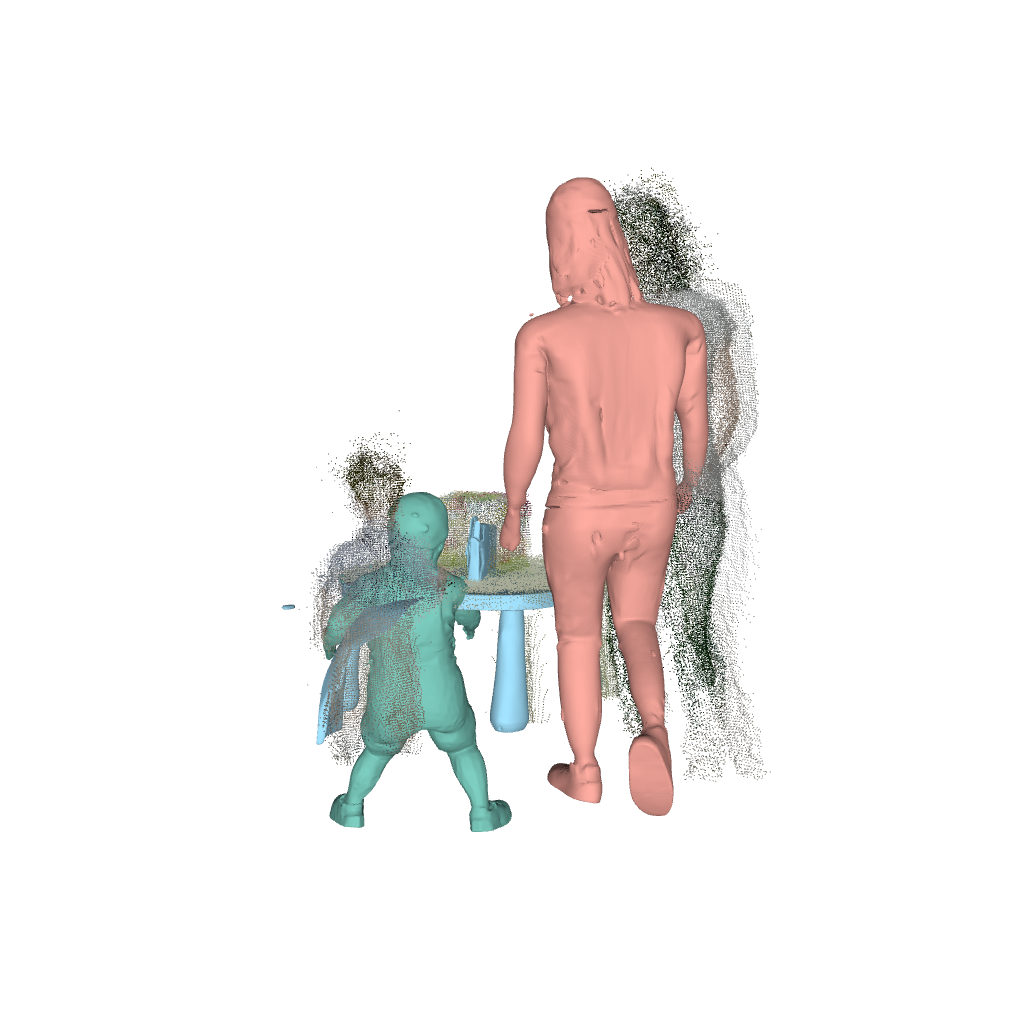} &
    \includegraphics[width=0.18\linewidth]{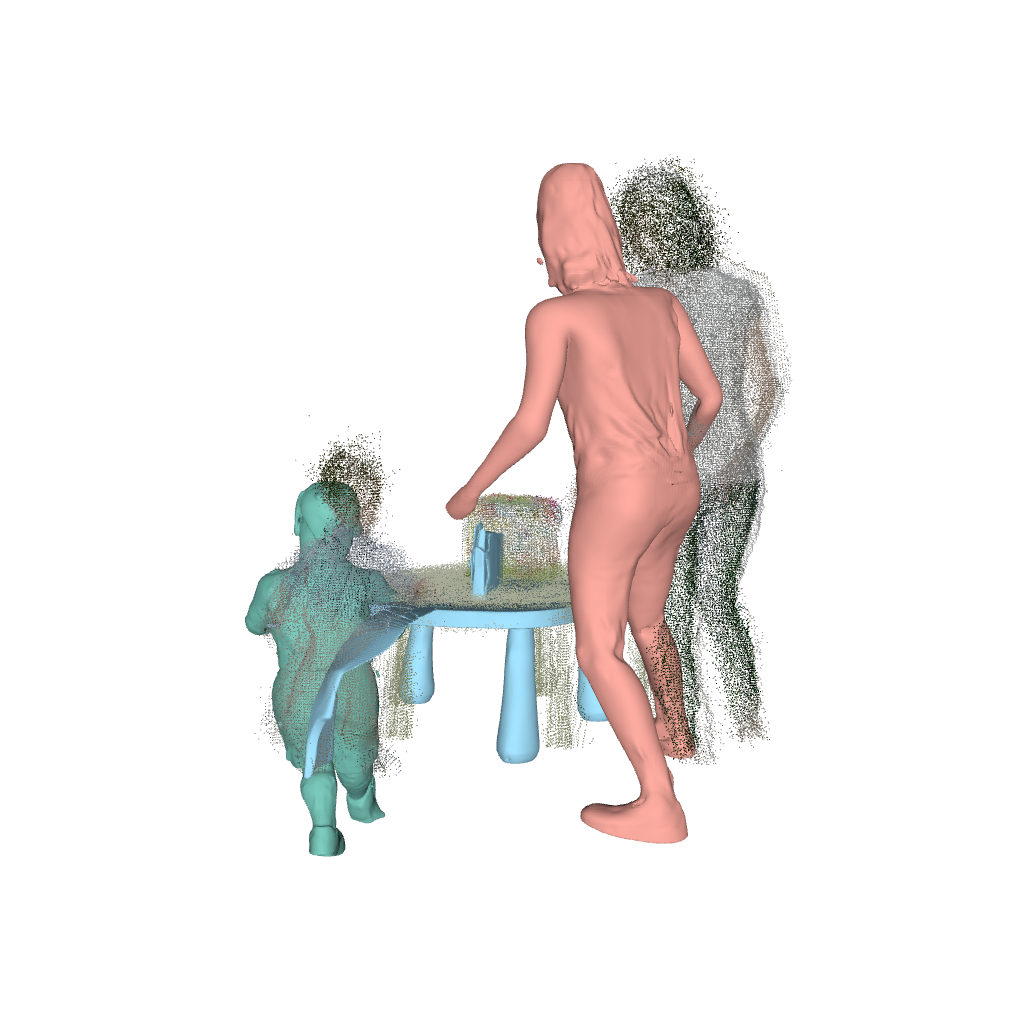} &
    \includegraphics[width=0.18\linewidth]{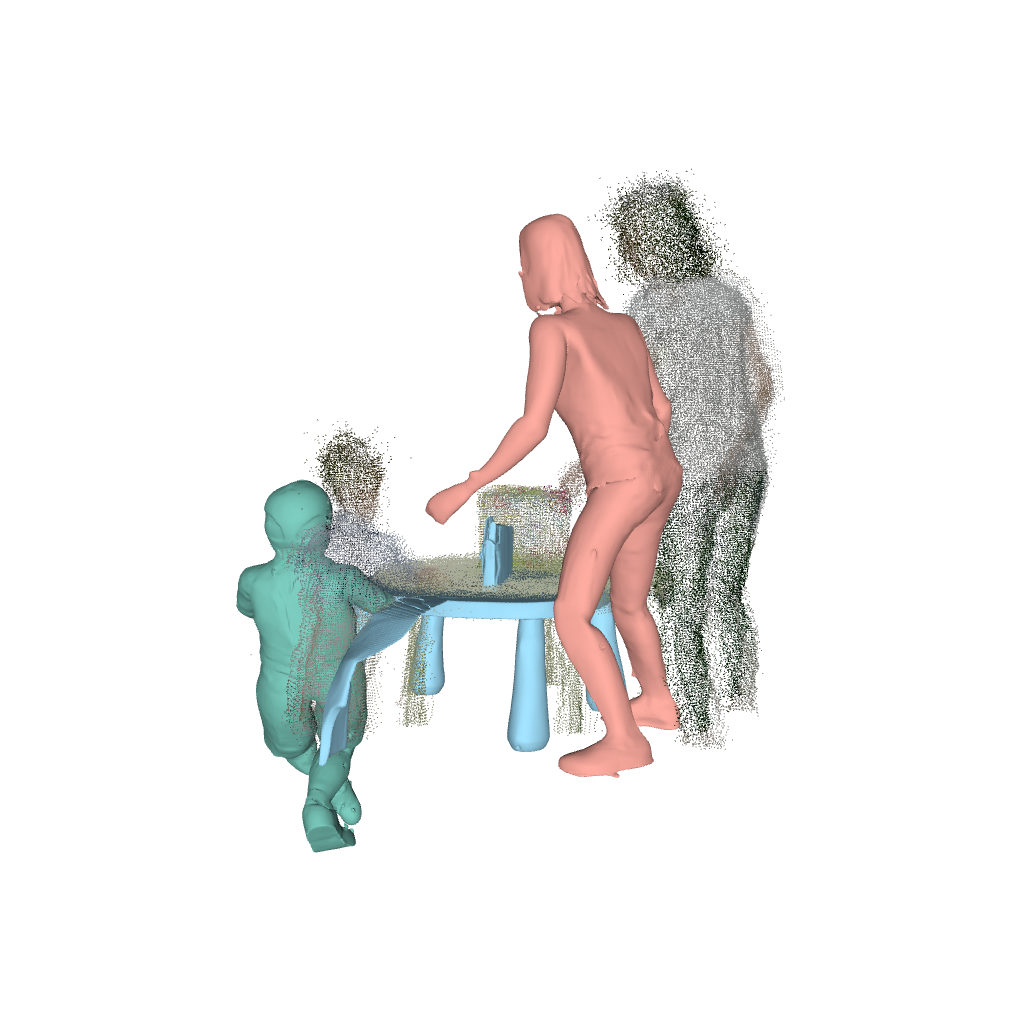} &
    \includegraphics[width=0.18\linewidth]{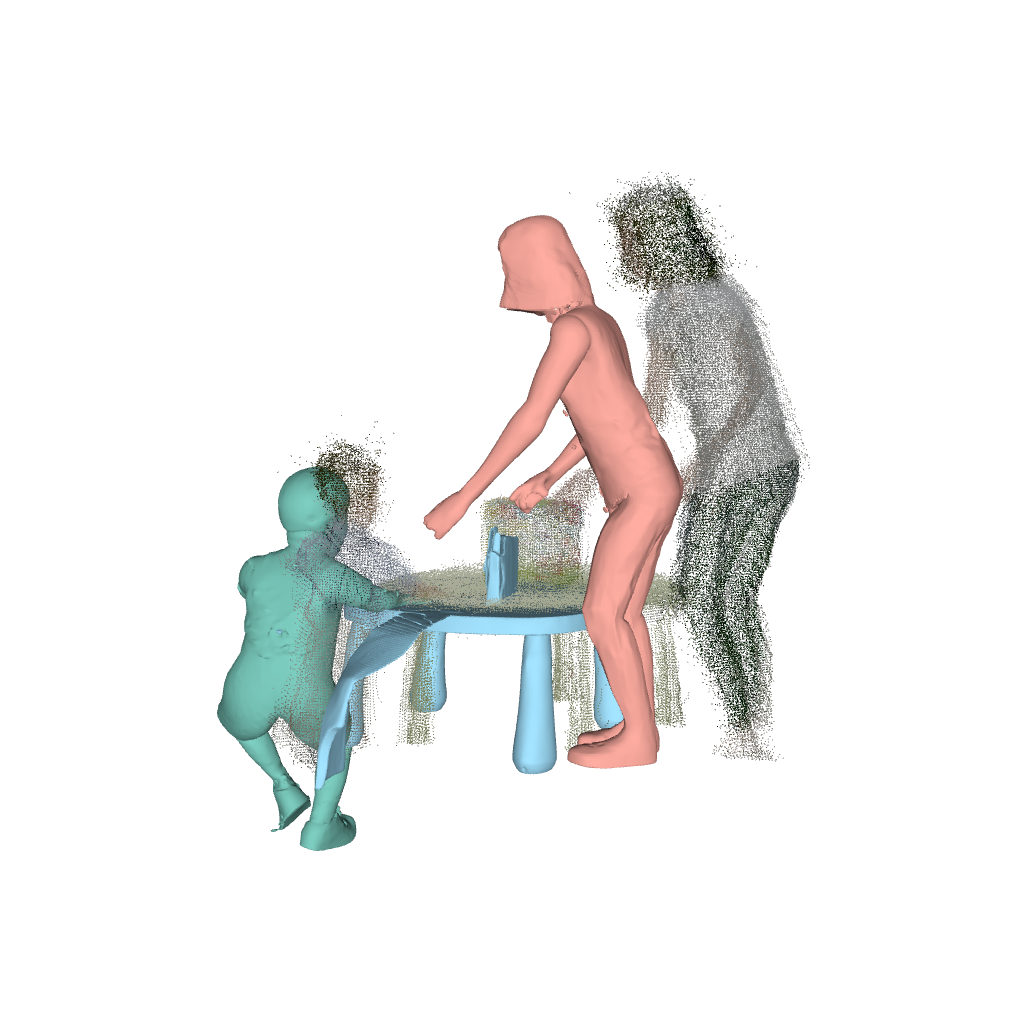} &
    \includegraphics[width=0.18\linewidth]{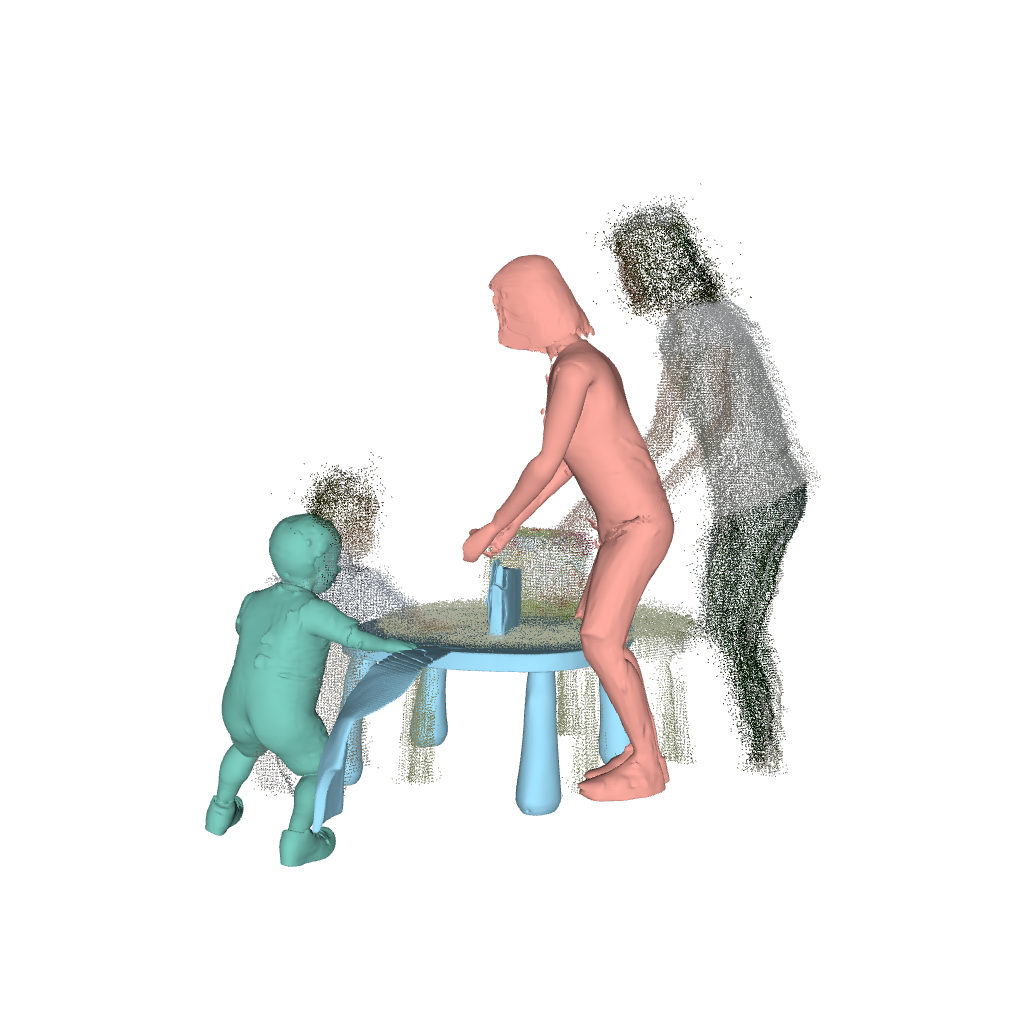} \\

    % --- Row 5: w/o Mixing View 2 ---
    \rotatebox{90}{\small} &
    \includegraphics[width=0.18\linewidth]{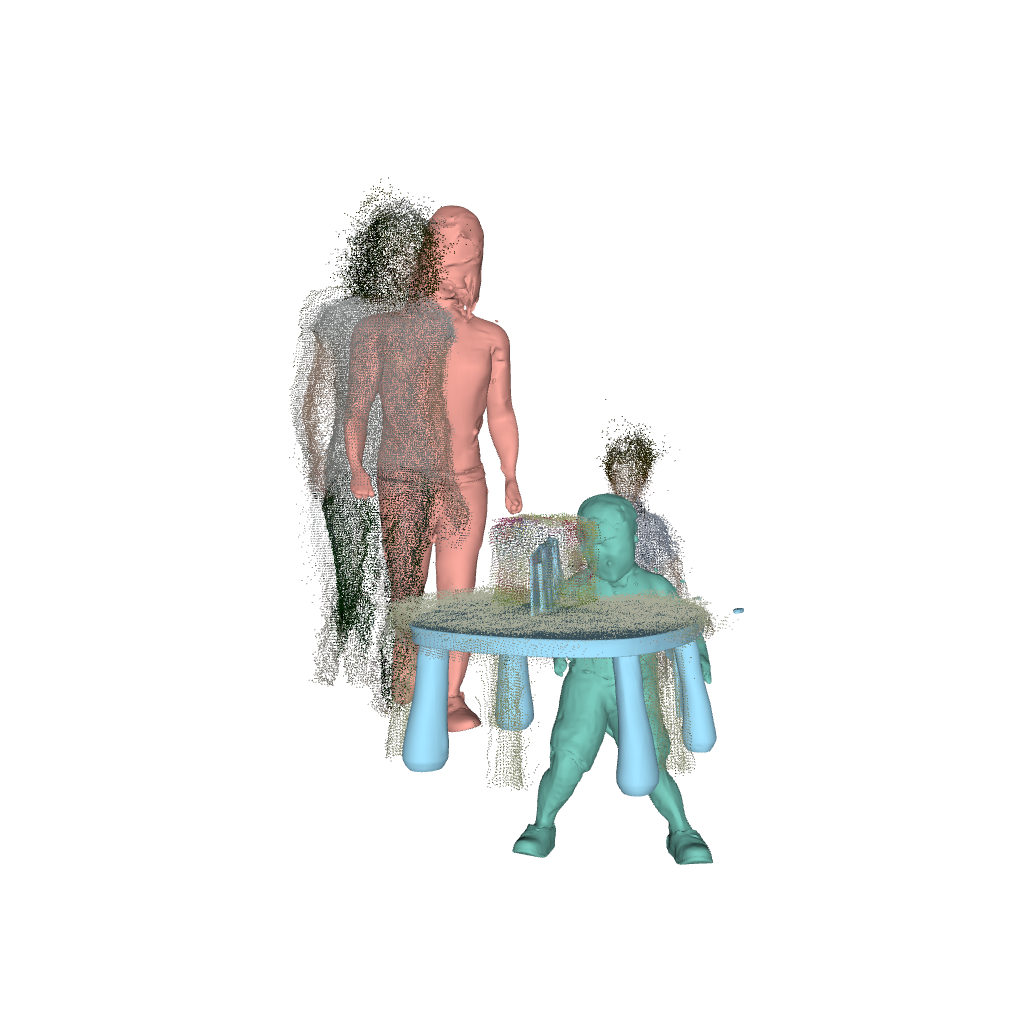} &
    \includegraphics[width=0.18\linewidth]{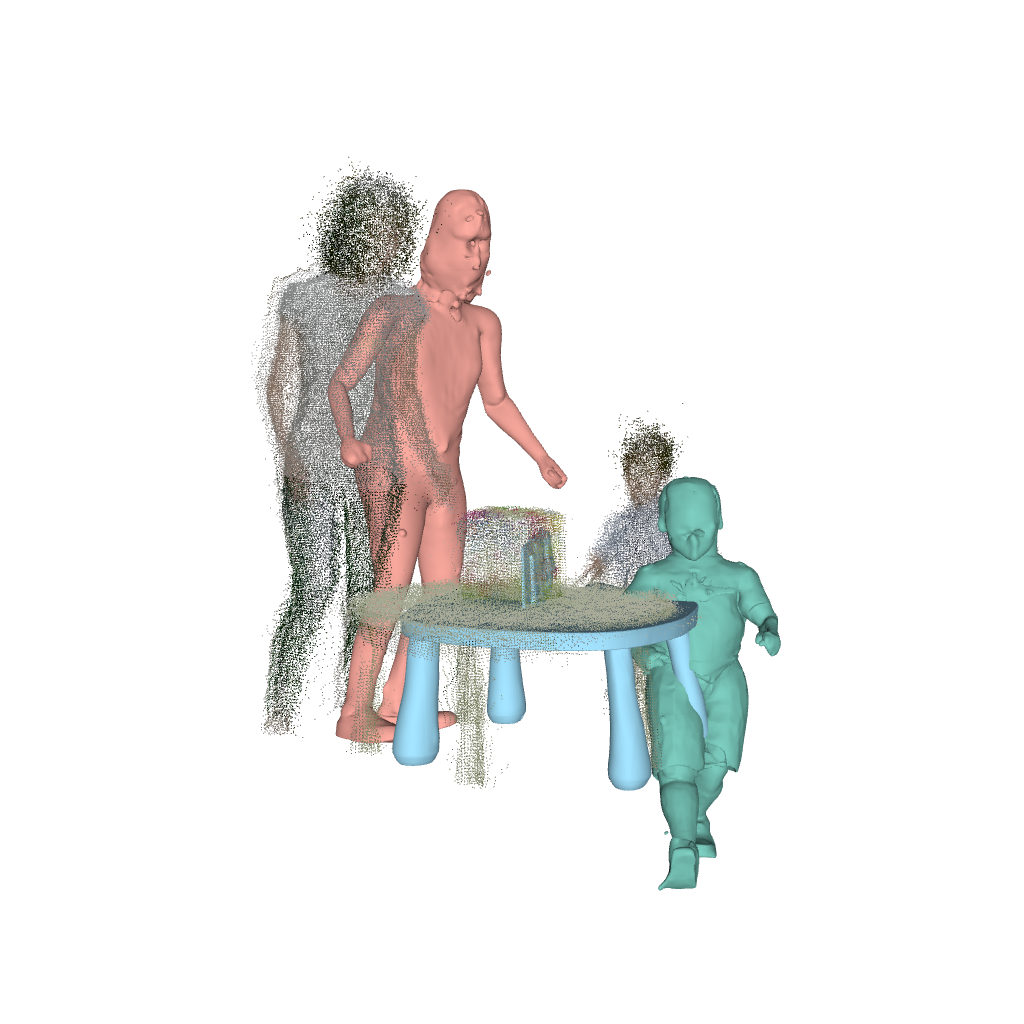} &
    \includegraphics[width=0.18\linewidth]{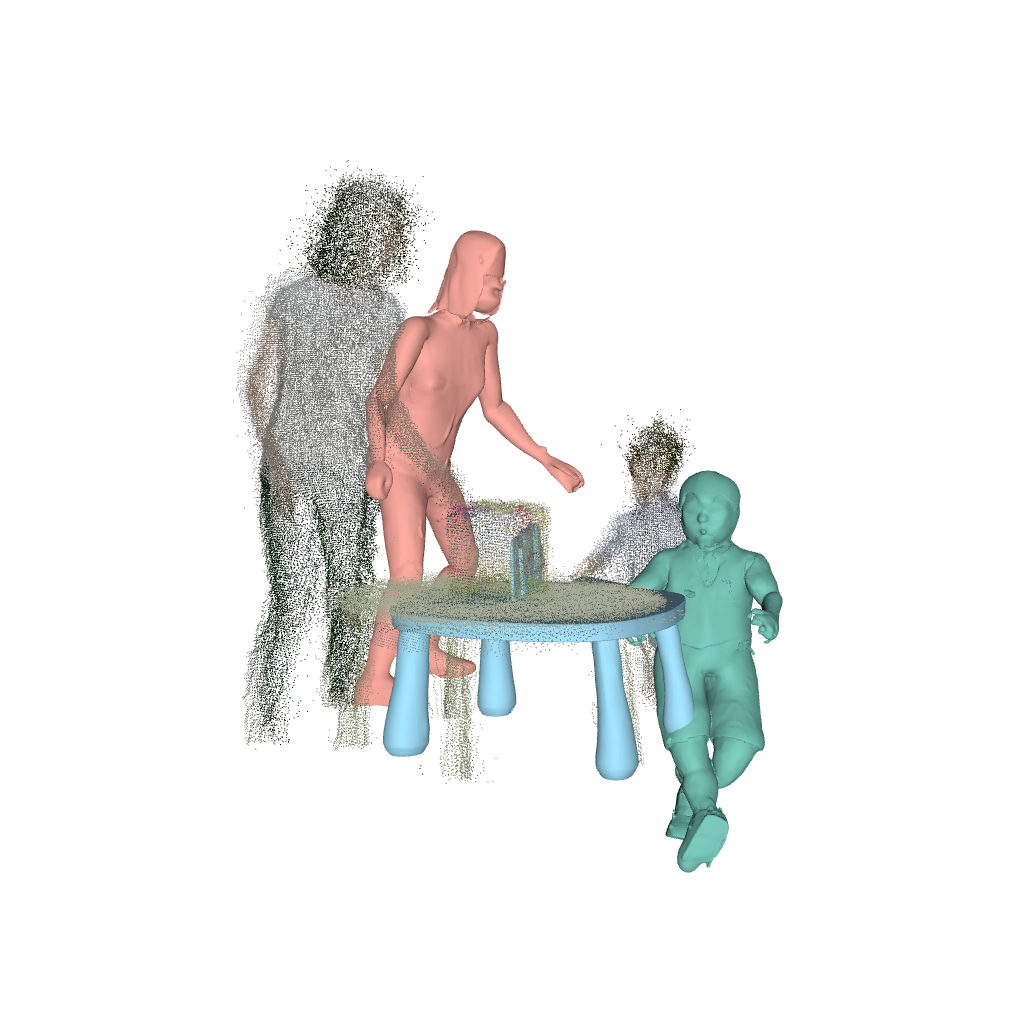} &
    \includegraphics[width=0.18\linewidth]{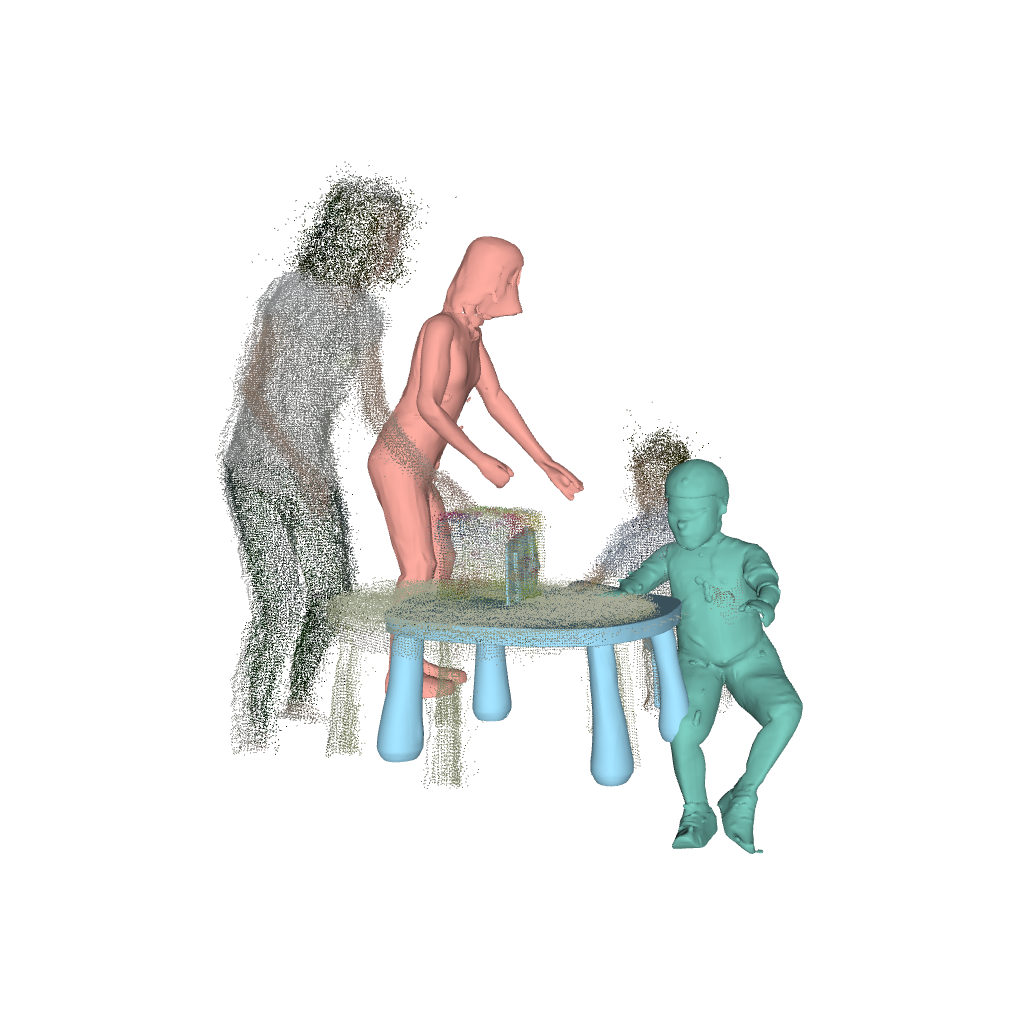} &
    \includegraphics[width=0.18\linewidth]{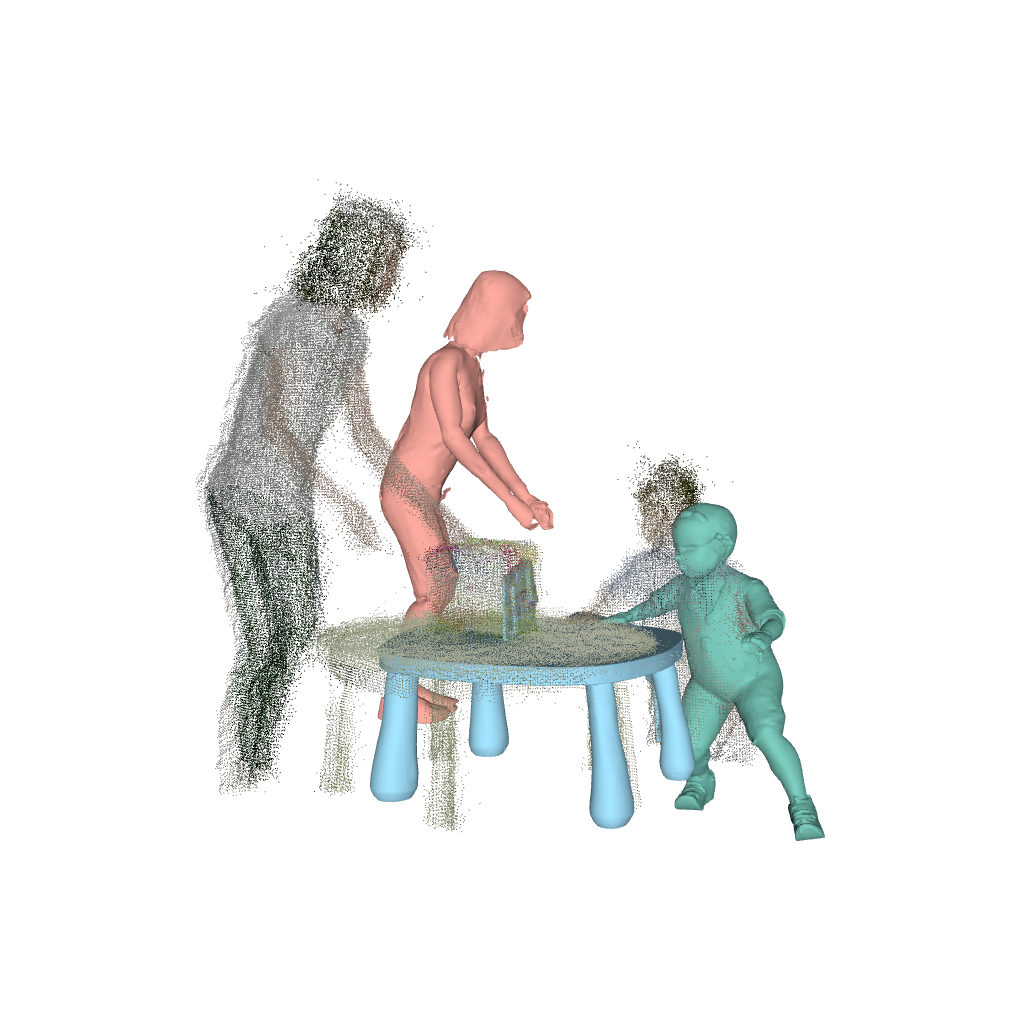} \\

  \end{tabular}
  % ==================== TABLE END ====================
  
  \vspace{-3pt}
  \caption{
    Ablation study on mixing components across five time steps in the CMU Panoptic~\cite{panoptic} sample. The top row shows the ground truth frames, followed by two views with our mixing strategy and two views without. Gray points denote the ground truth point cloud.
  }
  \label{fig:supp_toddler_ian_matching}
\end{figure}

\begin{figure}[h!]
  \centering
  % --- MODIFICATIONS ---
  \setlength{\tabcolsep}{0pt}      % NO horizontal space between columns
  \renewcommand{\arraystretch}{0} % NO vertical stretching of rows

  % ==================== TABLE START ====================
  \begin{tabular}{@{}c@{}ccccc@{}} % Use @{} to remove all column padding

    % --- Row 1: GT Frames ---
    \rotatebox{90}{\small ~~~~~Input} & % Rotated label with manual spacing
    \includegraphics[width=0.18\linewidth]{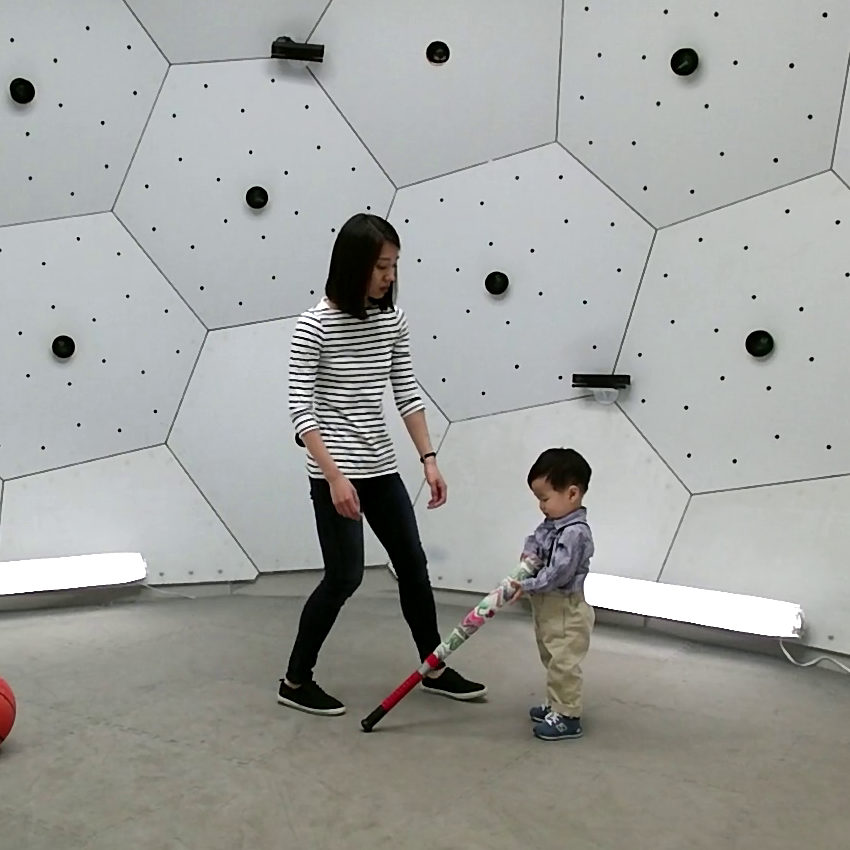} &
    \includegraphics[width=0.18\linewidth]{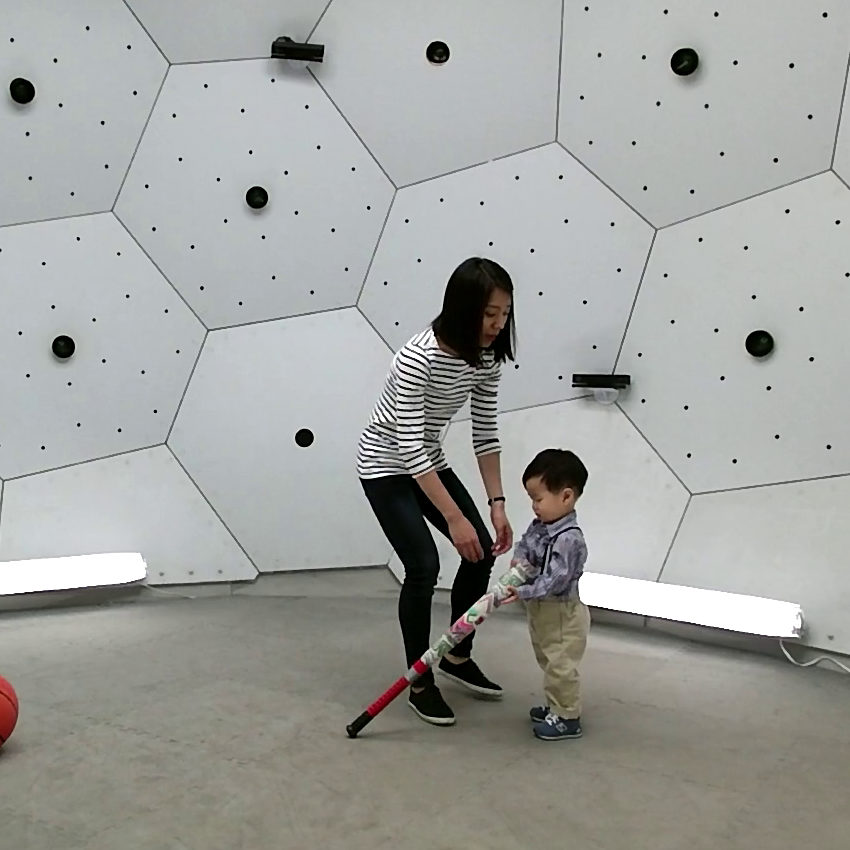} &
    \includegraphics[width=0.18\linewidth]{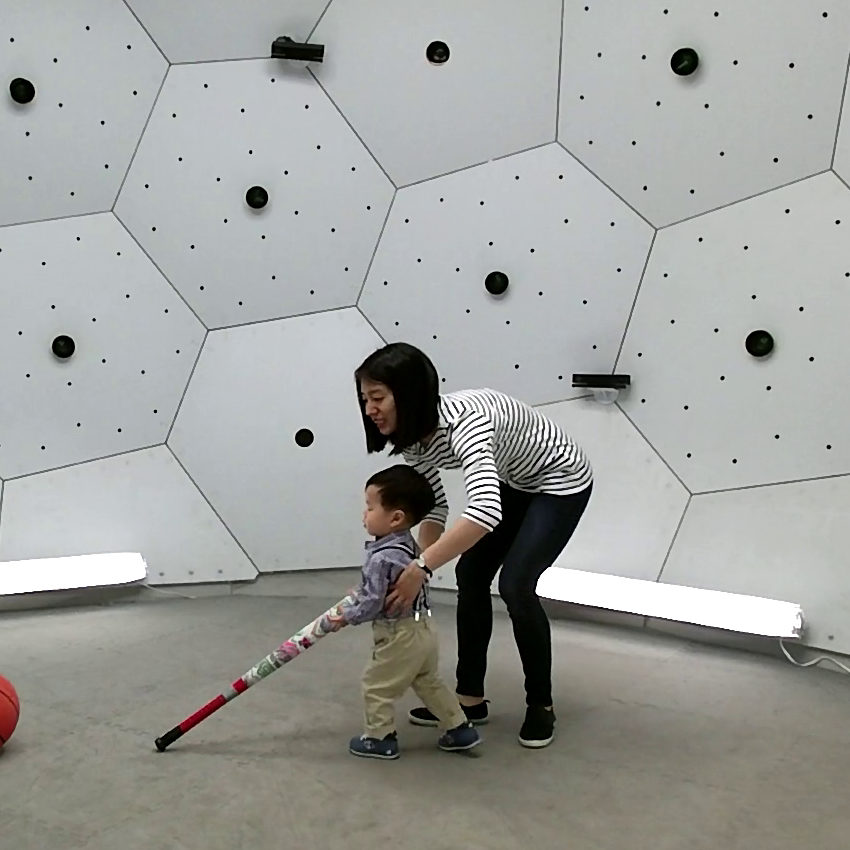} &
    \includegraphics[width=0.18\linewidth]{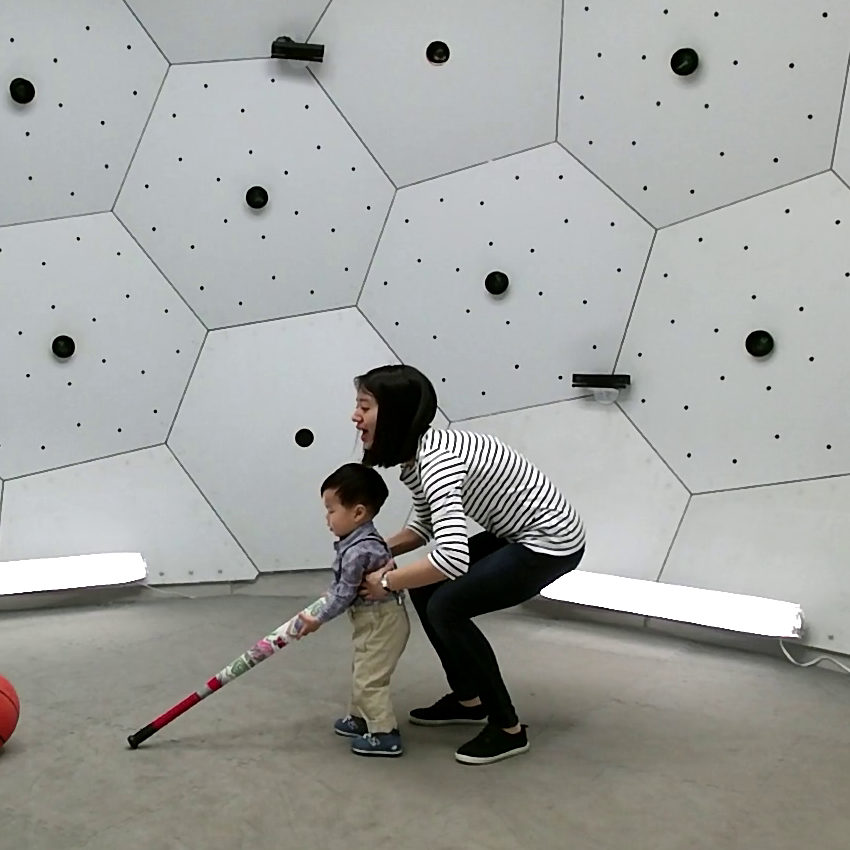} &
    \includegraphics[width=0.18\linewidth]{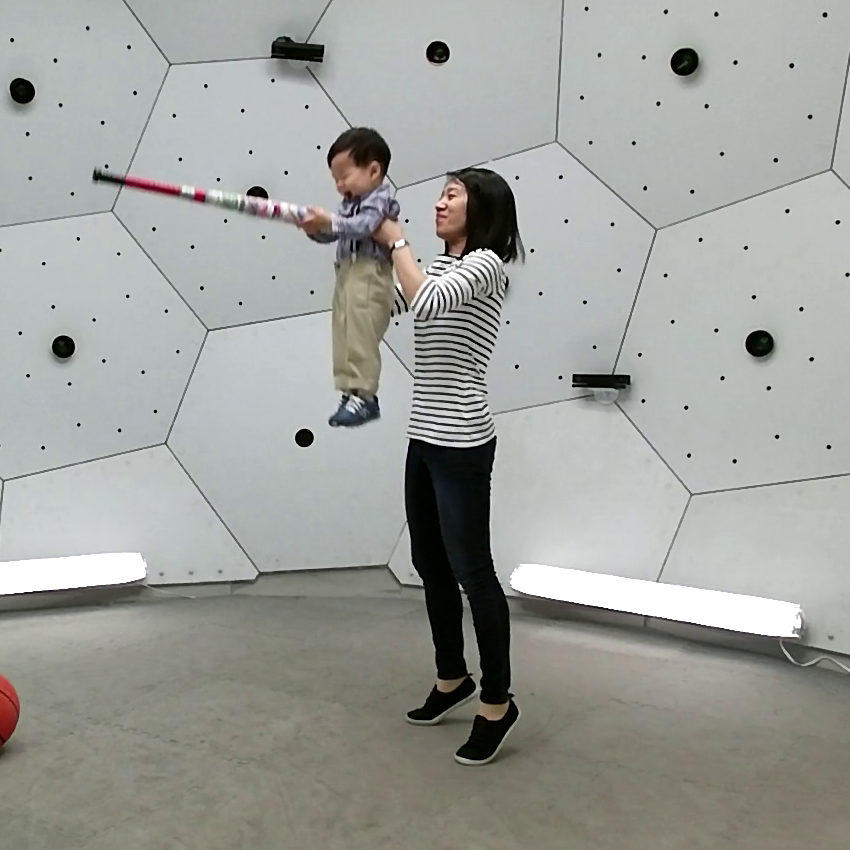} \\

    % --- Row 2: w/ Mixing View 1 ---
    \rotatebox{90}{\small w/ Mixing} &
    \includegraphics[width=0.18\linewidth]{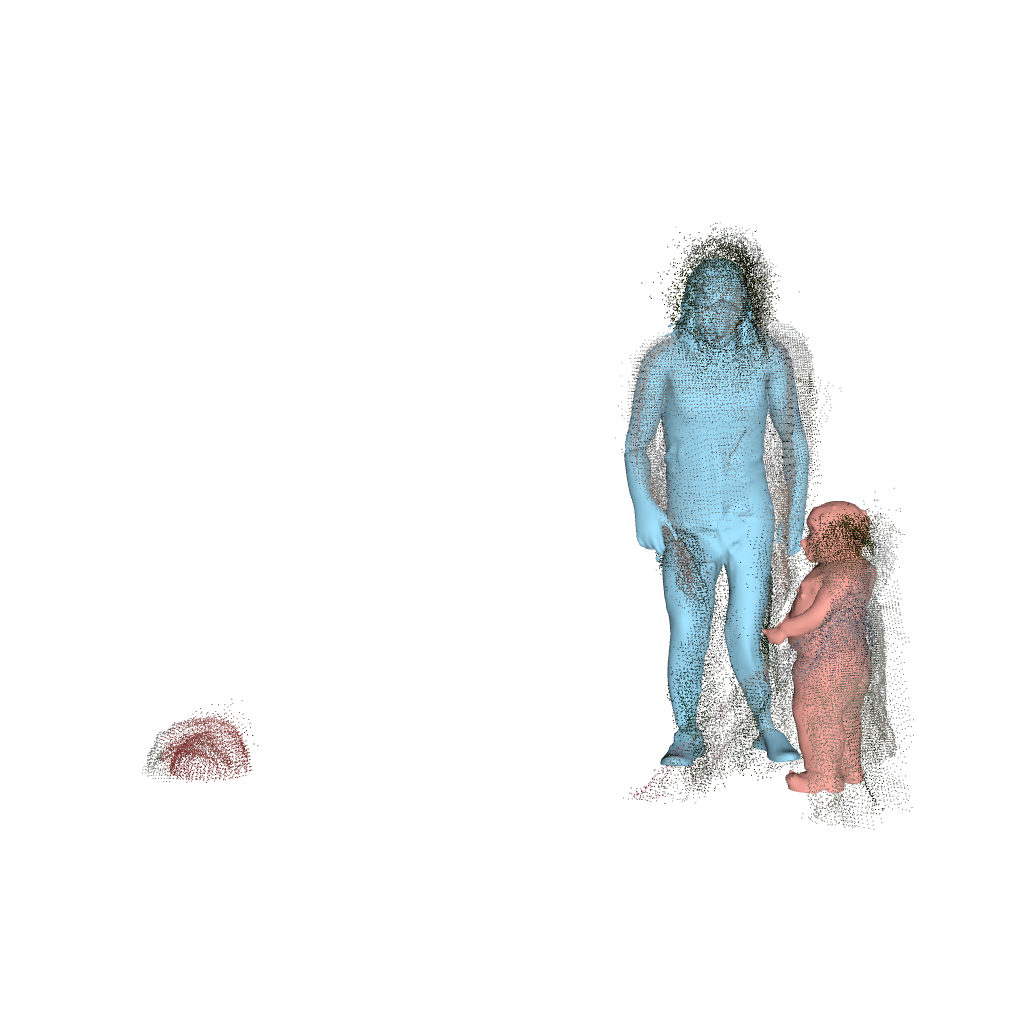} &
    \includegraphics[width=0.18\linewidth]{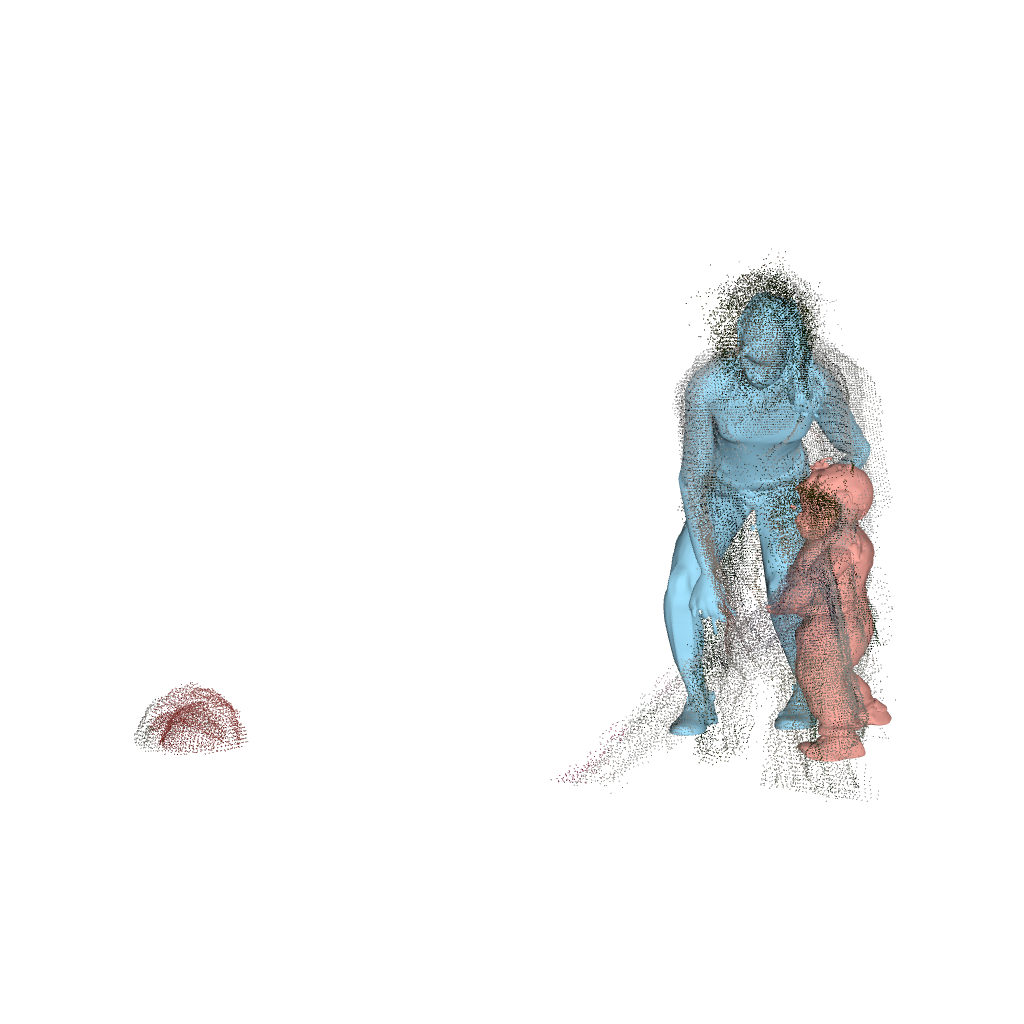} &
    \includegraphics[width=0.18\linewidth]{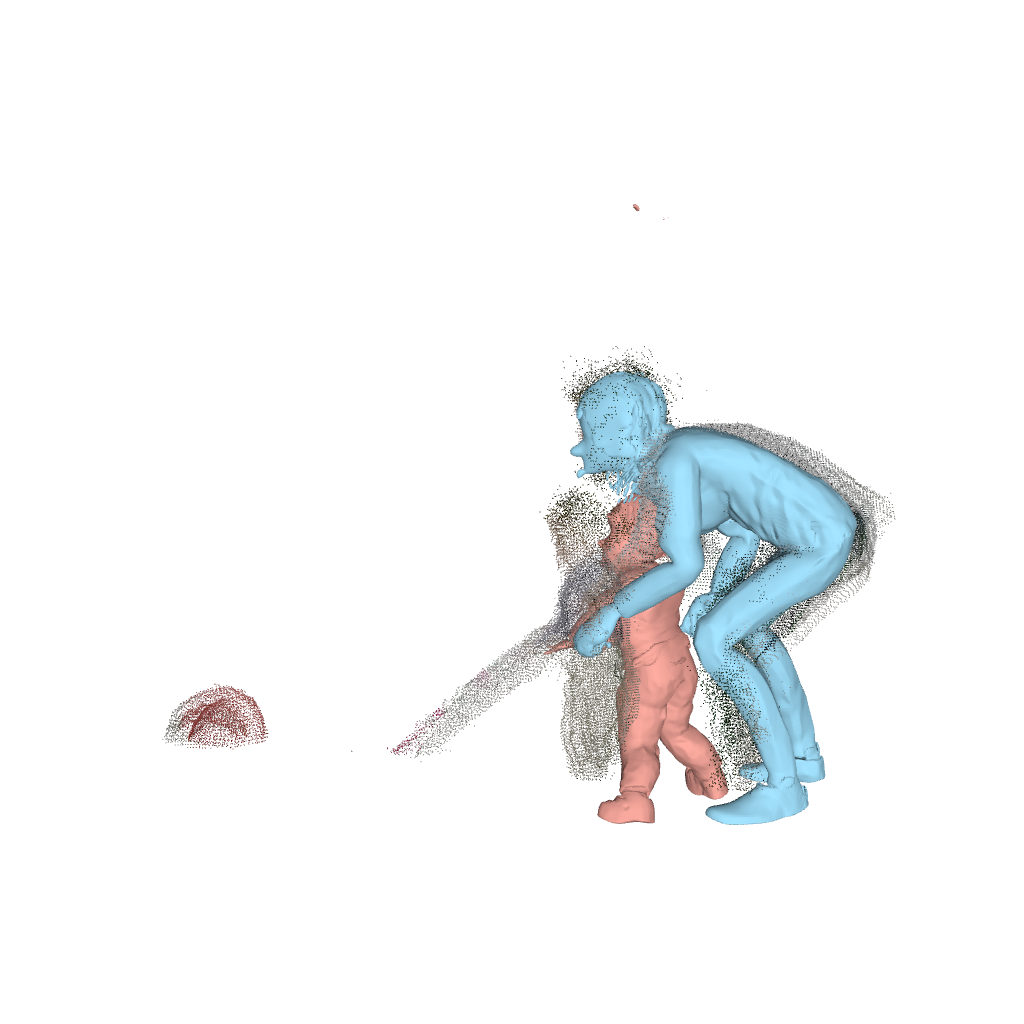} &
    \includegraphics[width=0.18\linewidth]{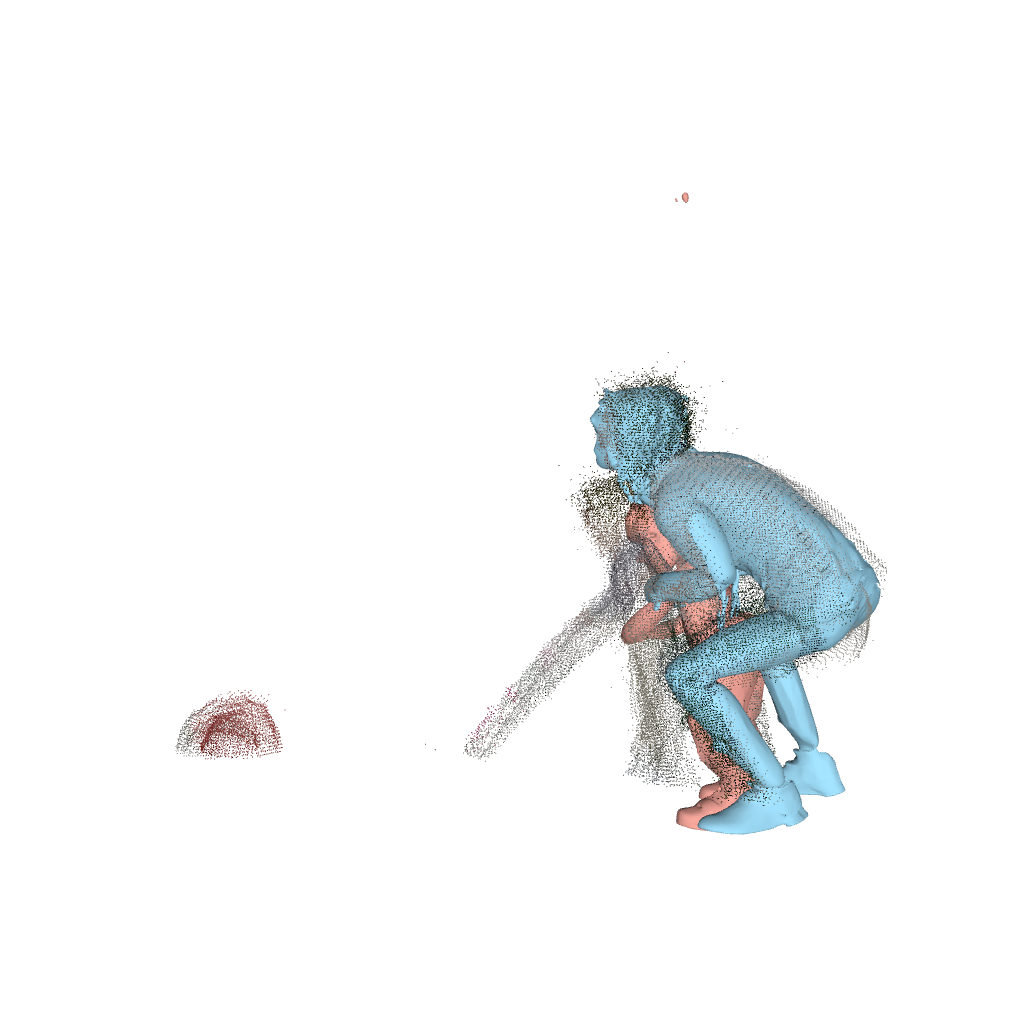} &
    \includegraphics[width=0.18\linewidth]{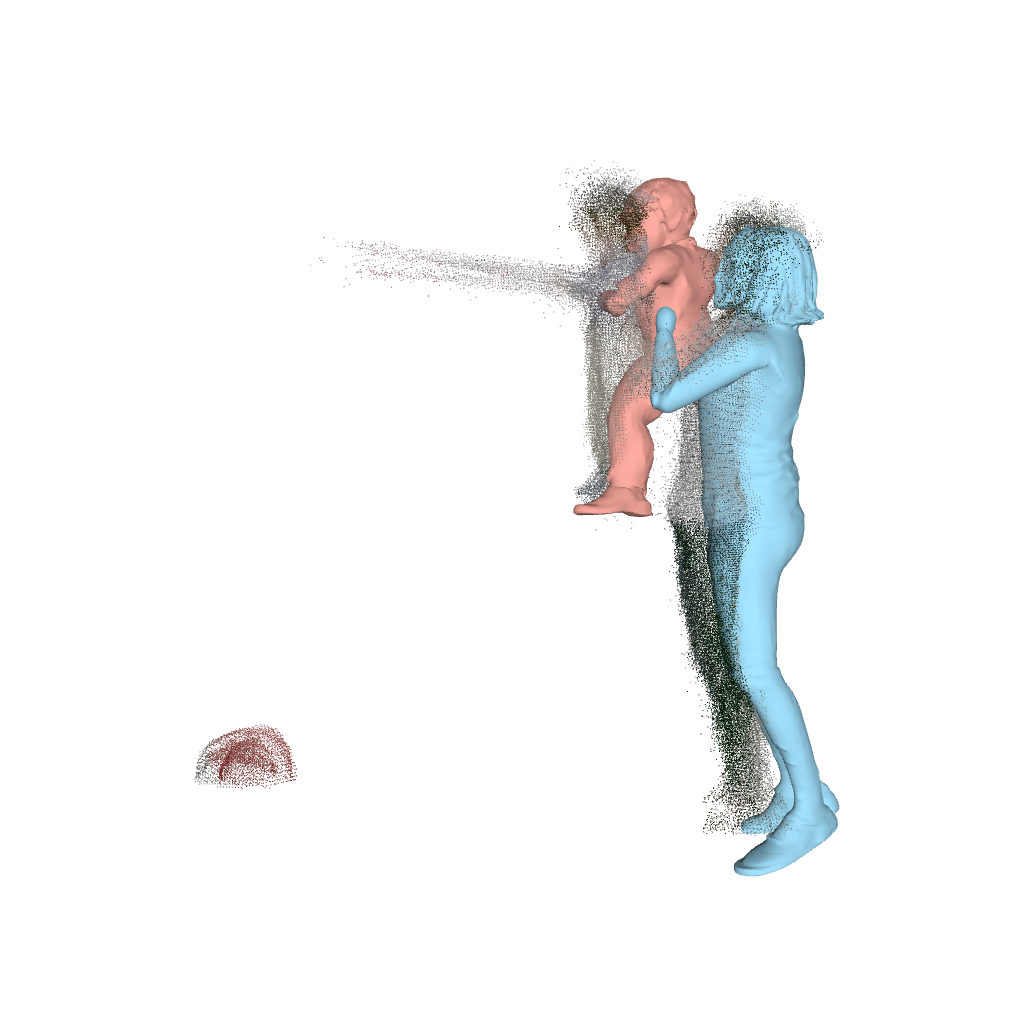} \\

    % --- Row 3: w/ Mixing View 2 ---
    \rotatebox{90}{\small} &
    \includegraphics[width=0.18\linewidth]{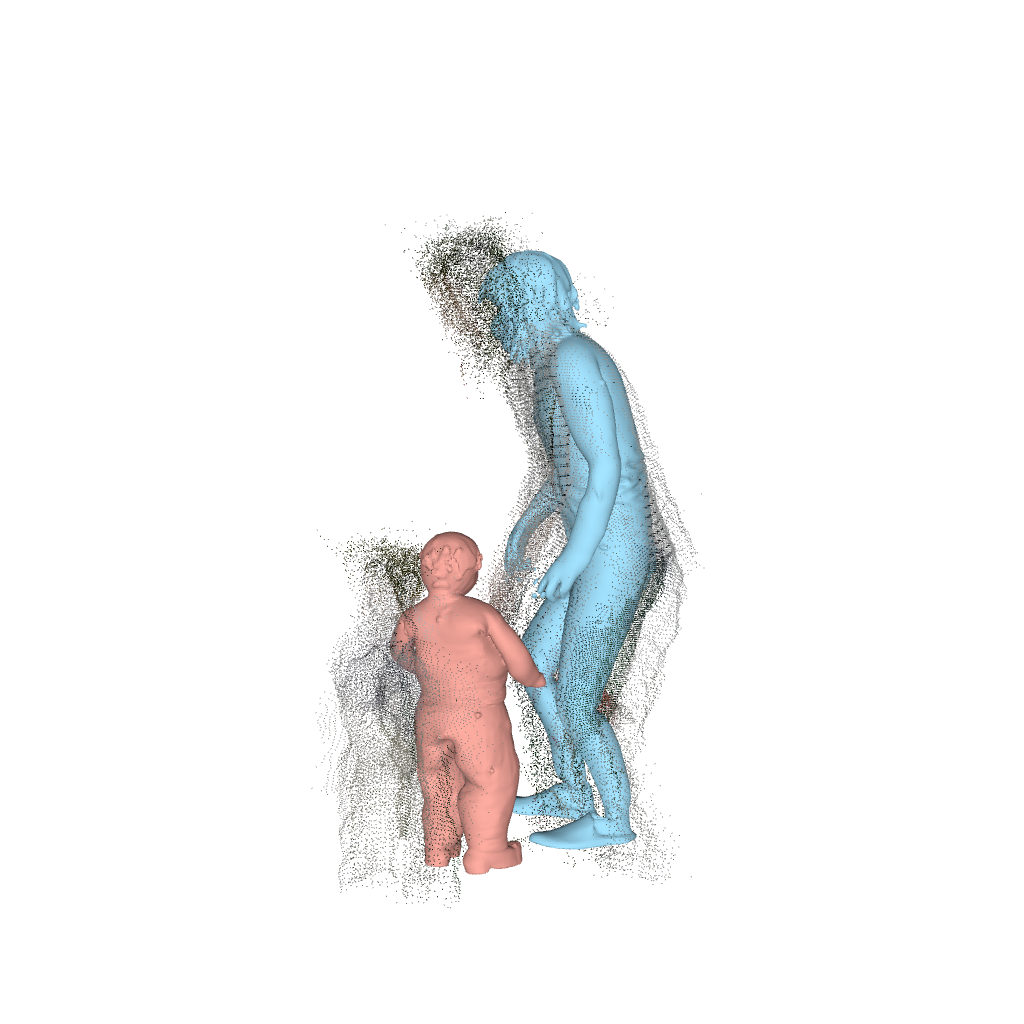} &
    \includegraphics[width=0.18\linewidth]{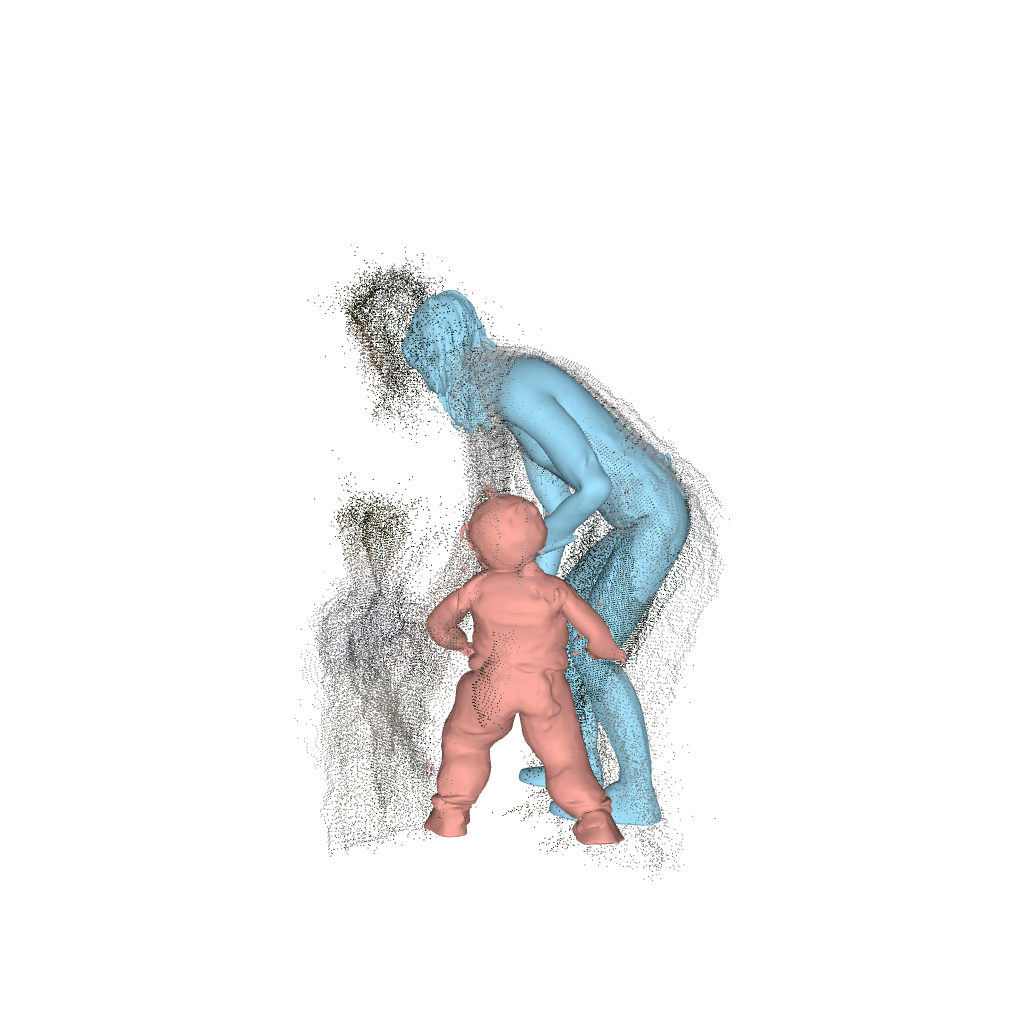} &
    \includegraphics[width=0.18\linewidth]{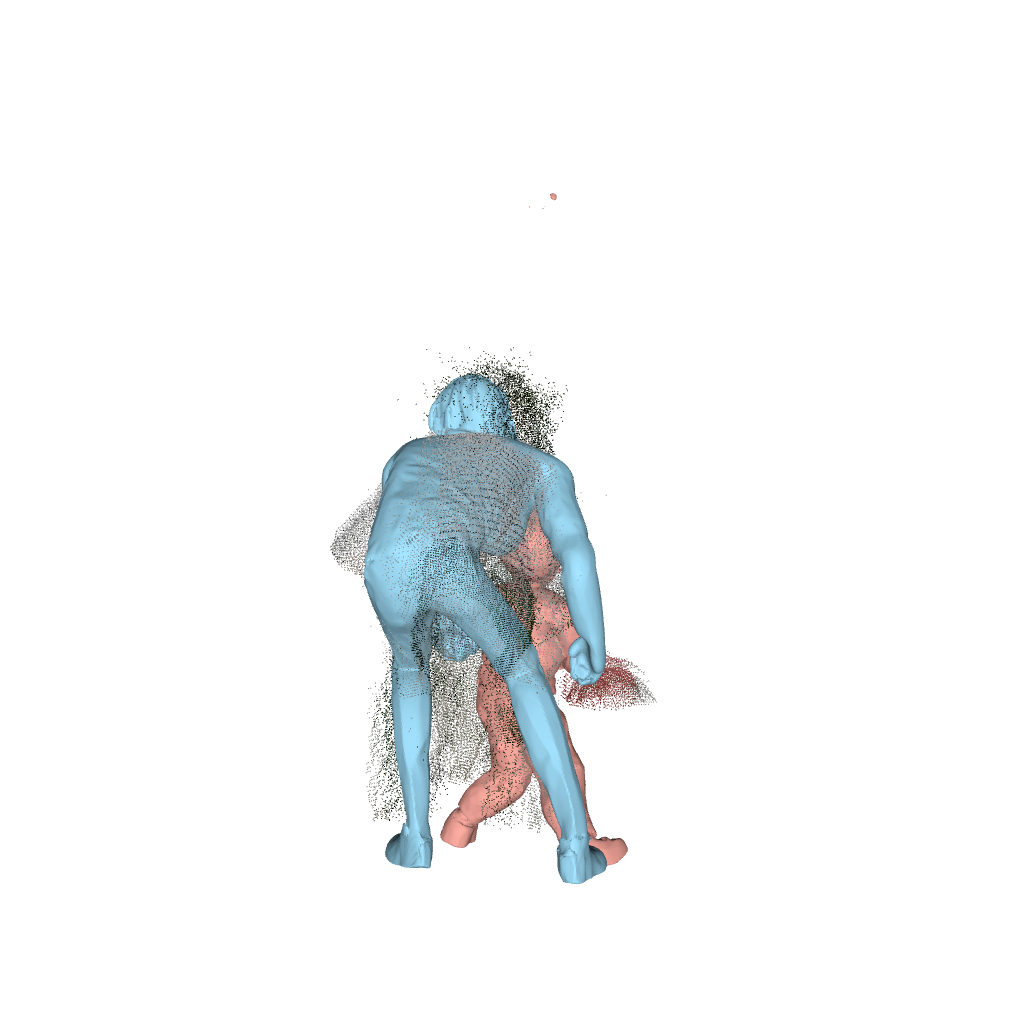} &
    \includegraphics[width=0.18\linewidth]{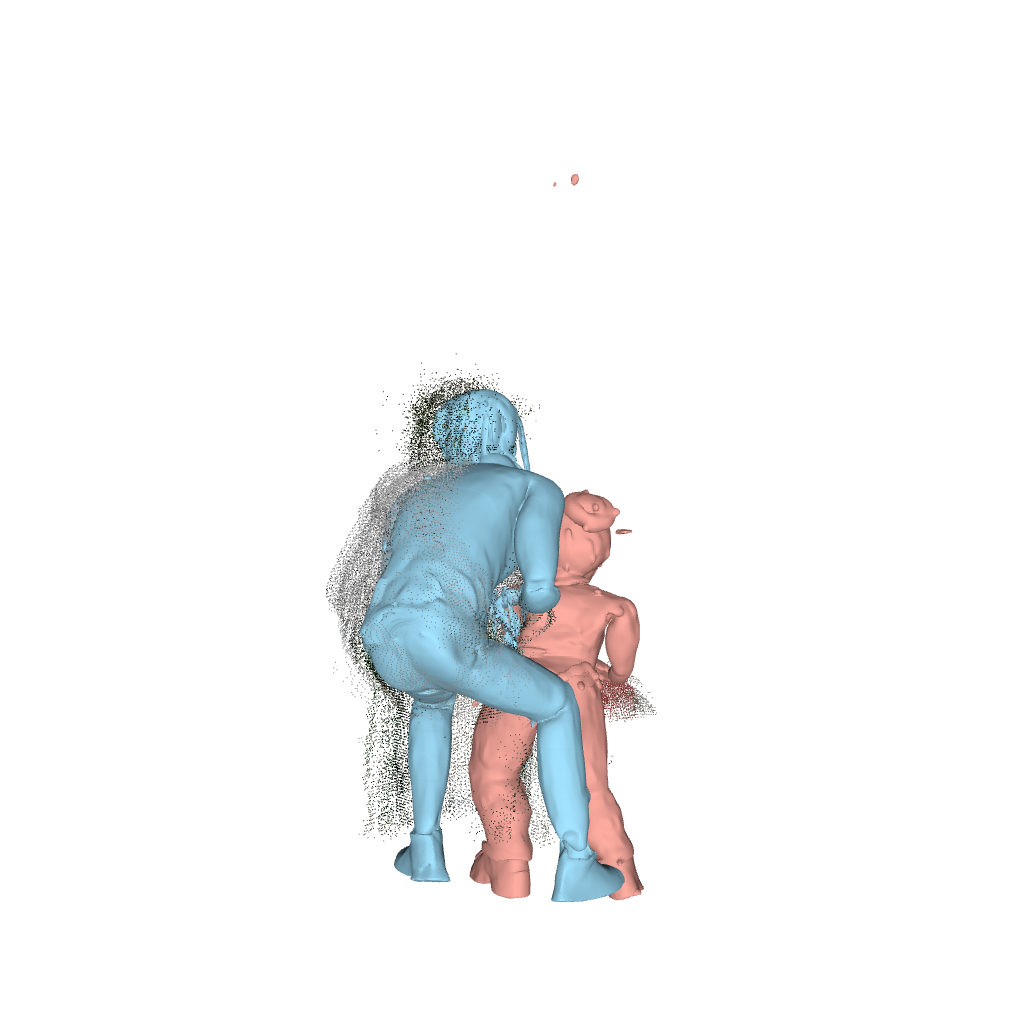} &
    \includegraphics[width=0.18\linewidth]{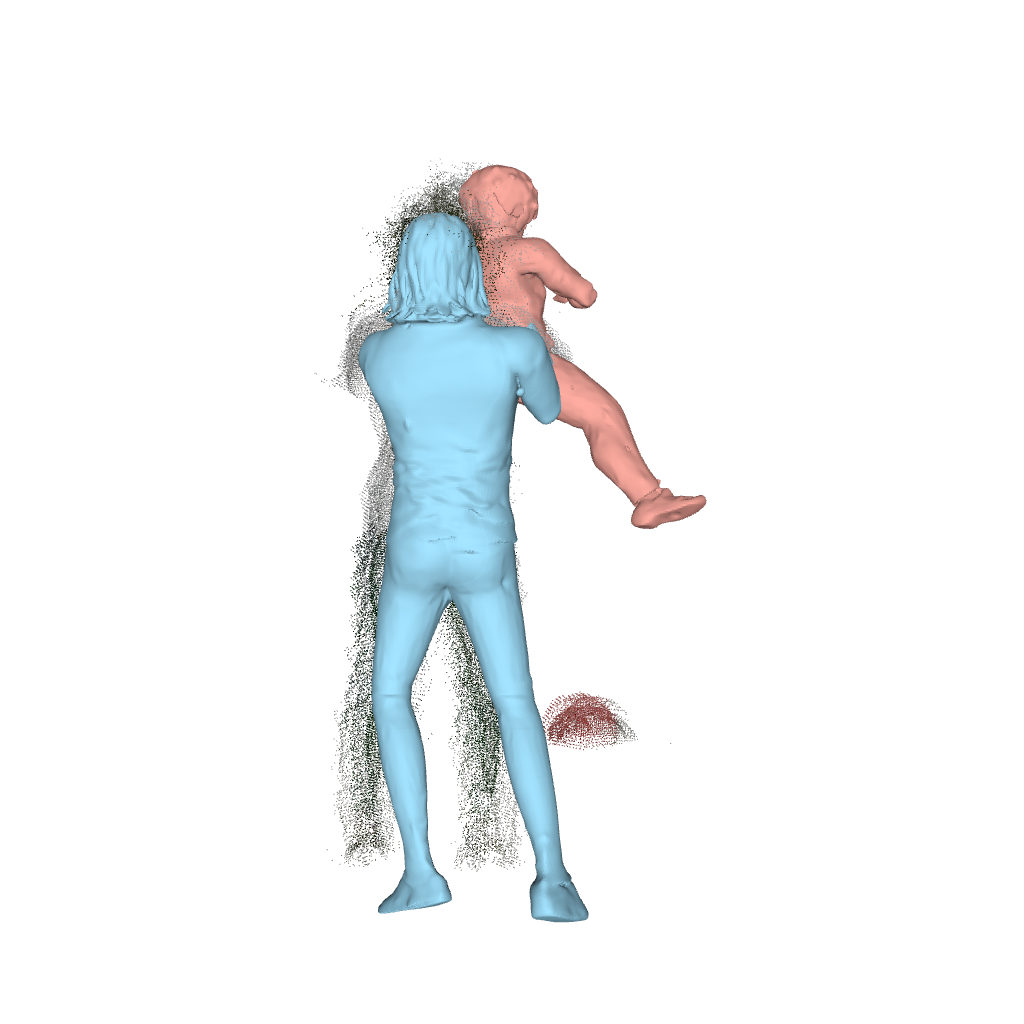} \\

    % --- Row 4: w/o Mixing View 1 ---
    \rotatebox{90}{\small w/o Mixing} &
    \includegraphics[width=0.18\linewidth]{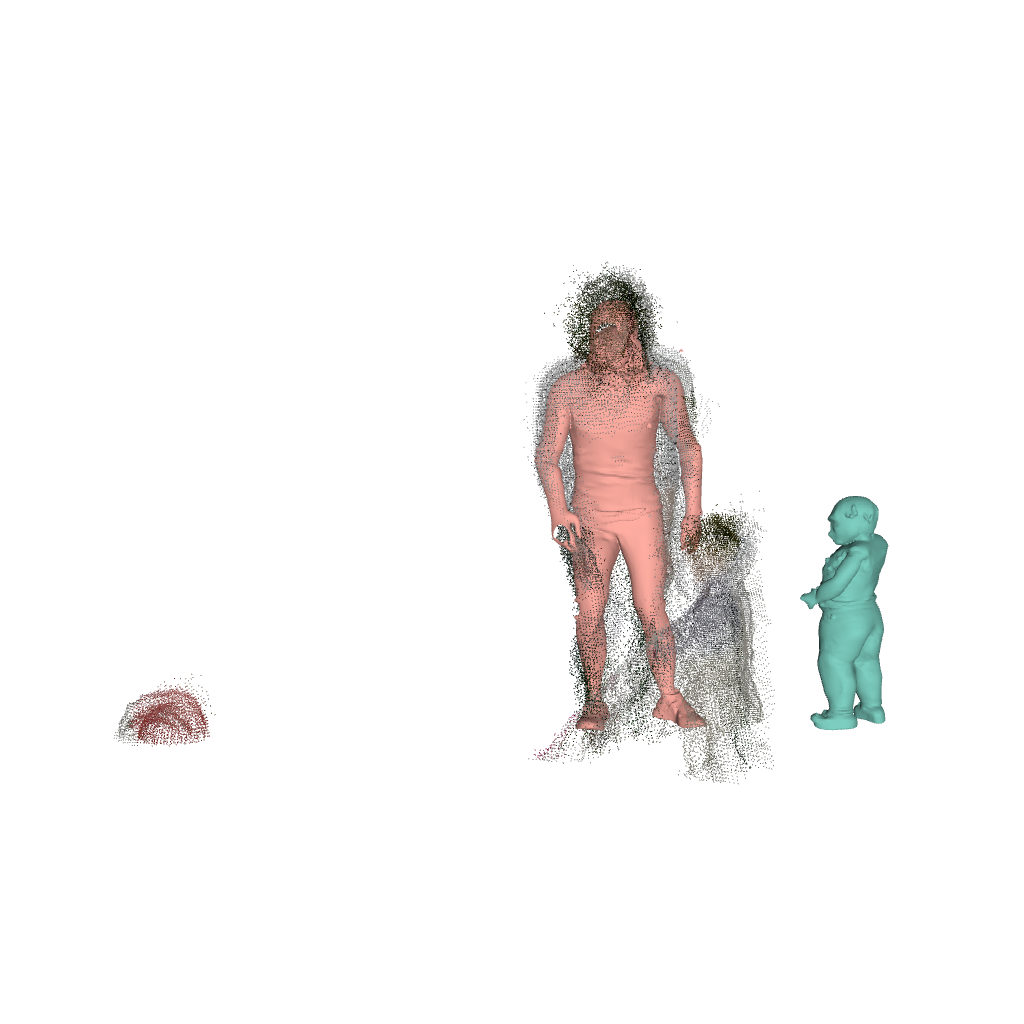} &
    \includegraphics[width=0.18\linewidth]{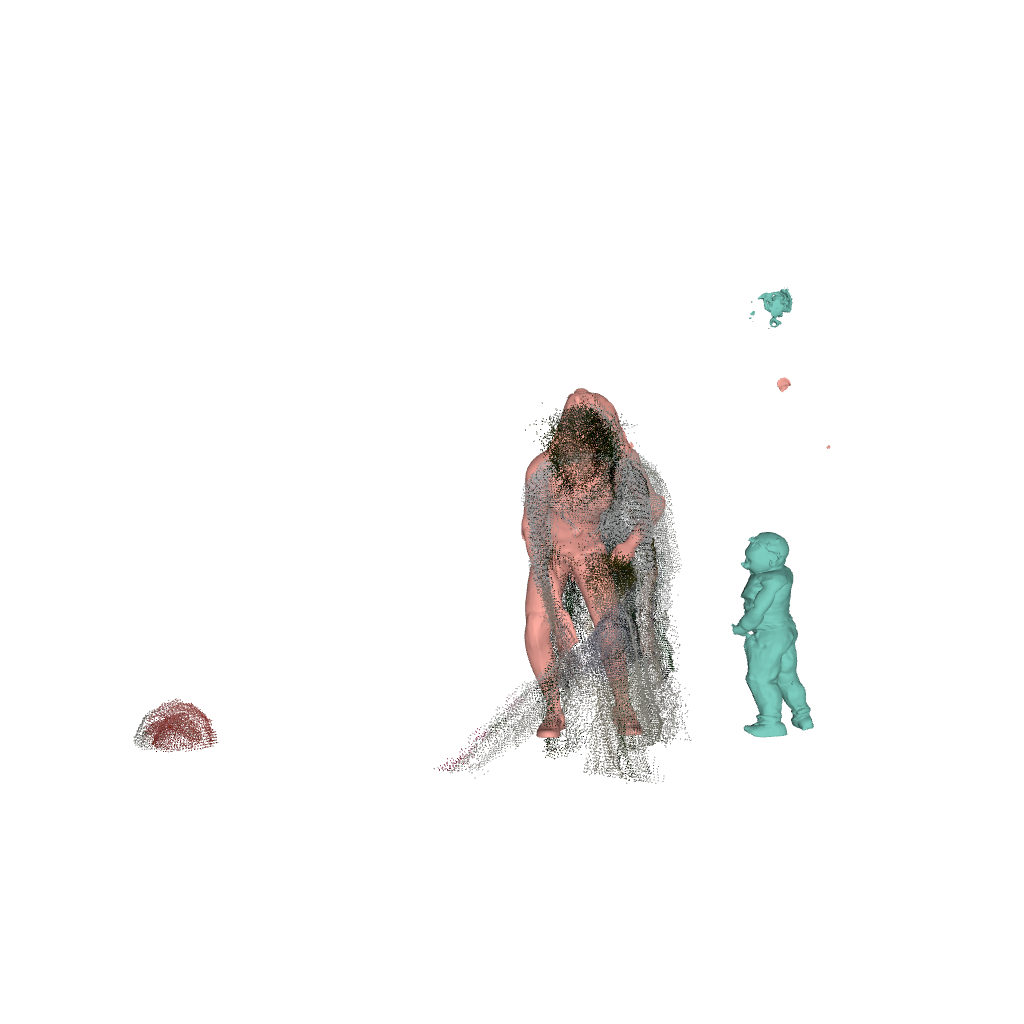} &
    \includegraphics[width=0.18\linewidth]{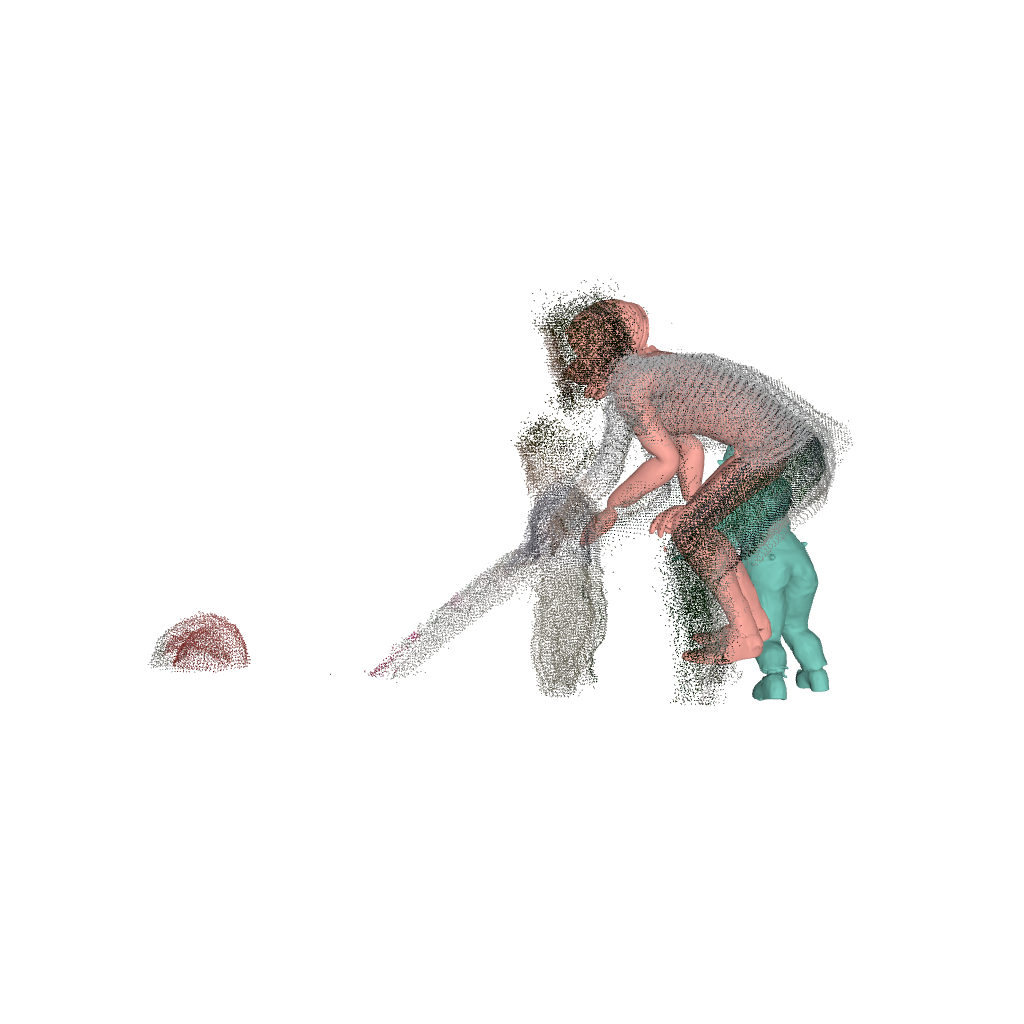} &
    \includegraphics[width=0.18\linewidth]{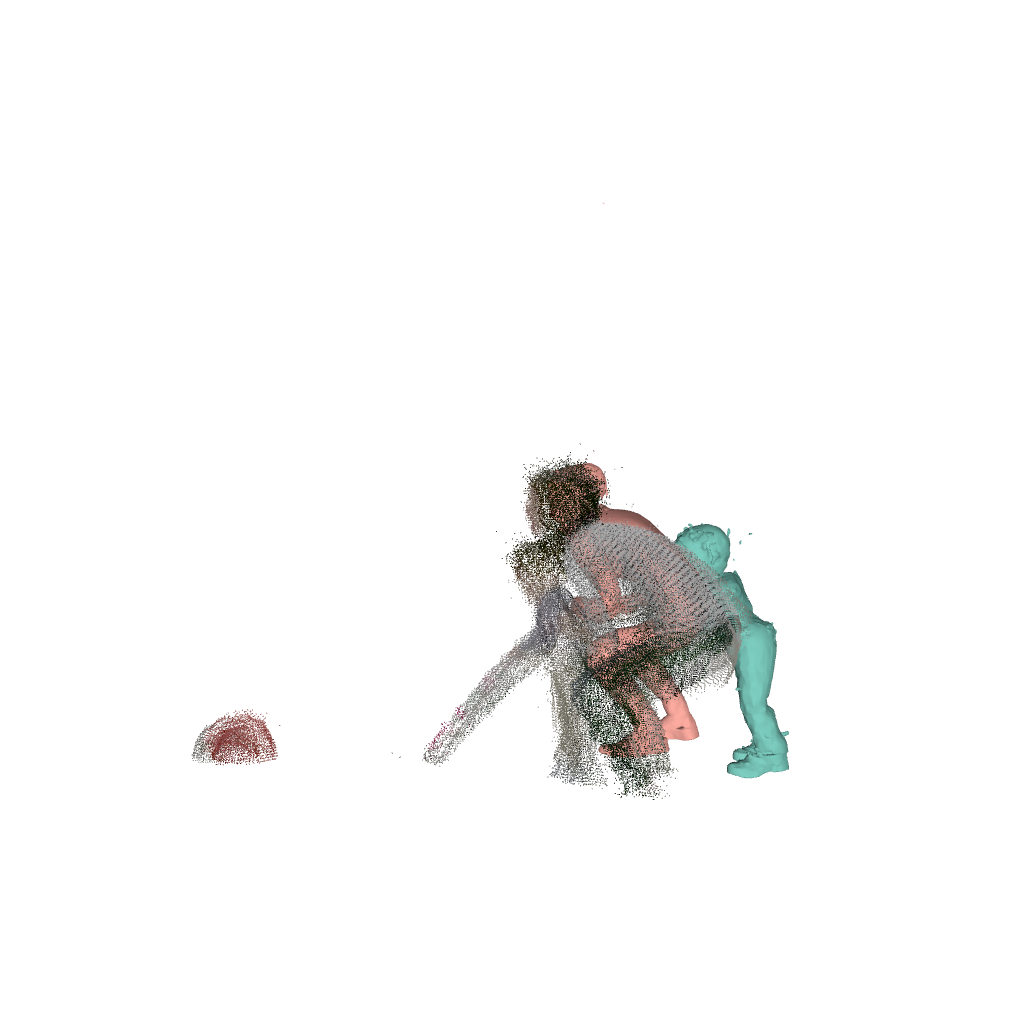} &
    \includegraphics[width=0.18\linewidth]{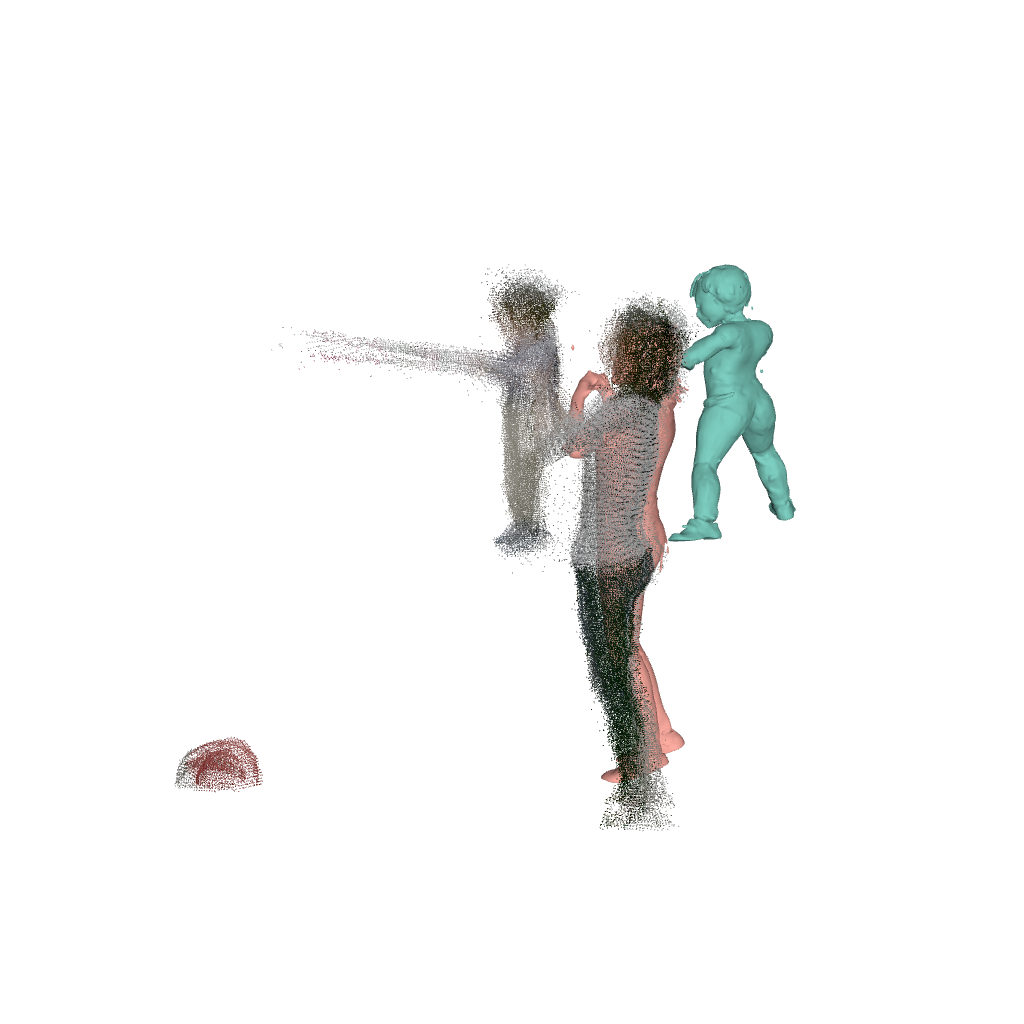} \\

    % --- Row 5: w/o Mixing View 2 ---
    \rotatebox{90}{\small} &
    \includegraphics[width=0.18\linewidth]{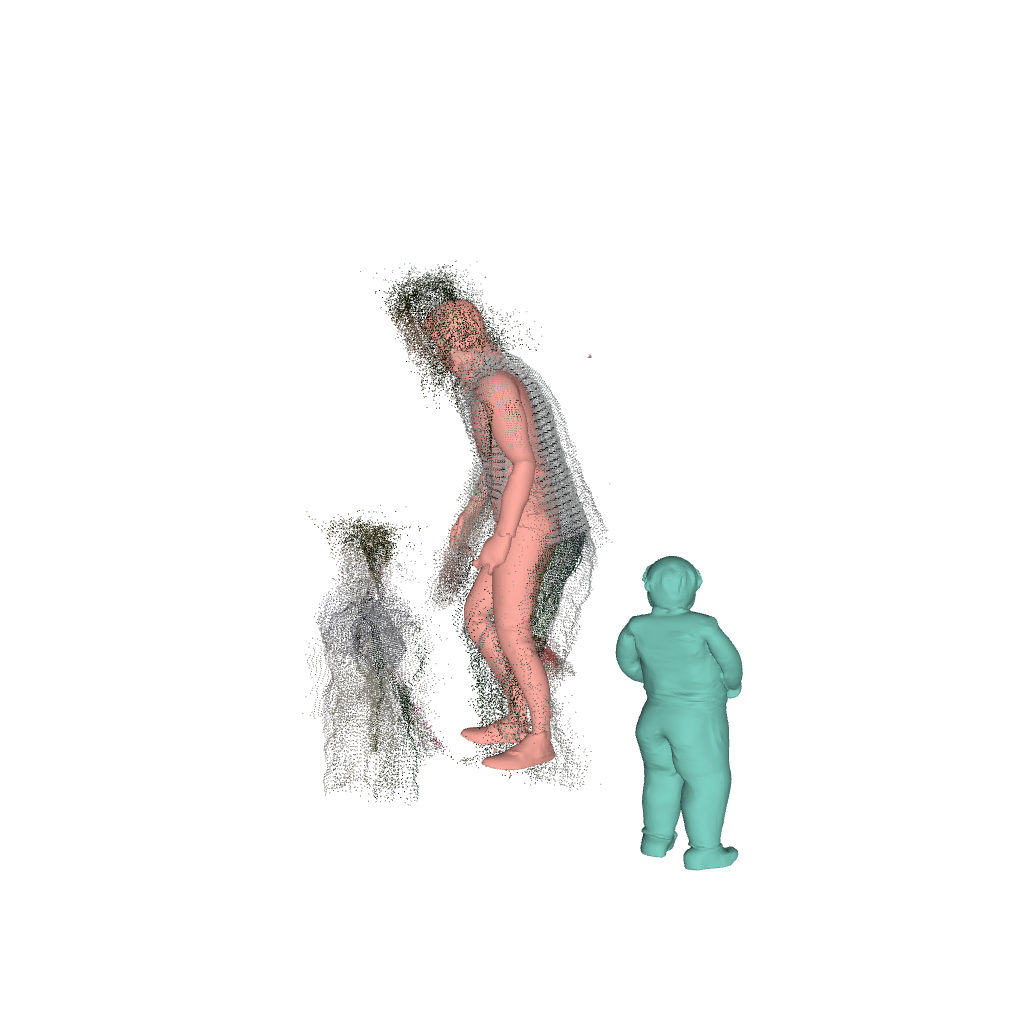} &
    \includegraphics[width=0.18\linewidth]{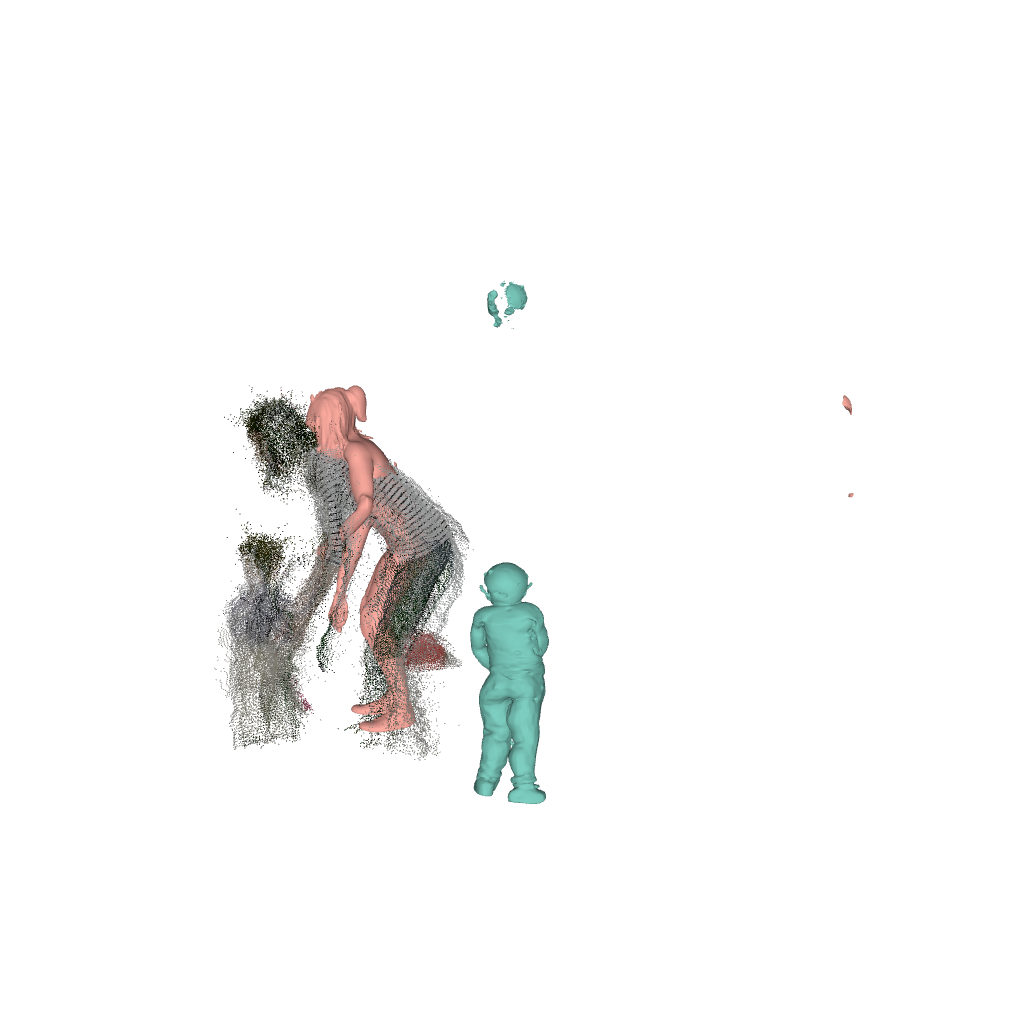} &
    \includegraphics[width=0.18\linewidth]{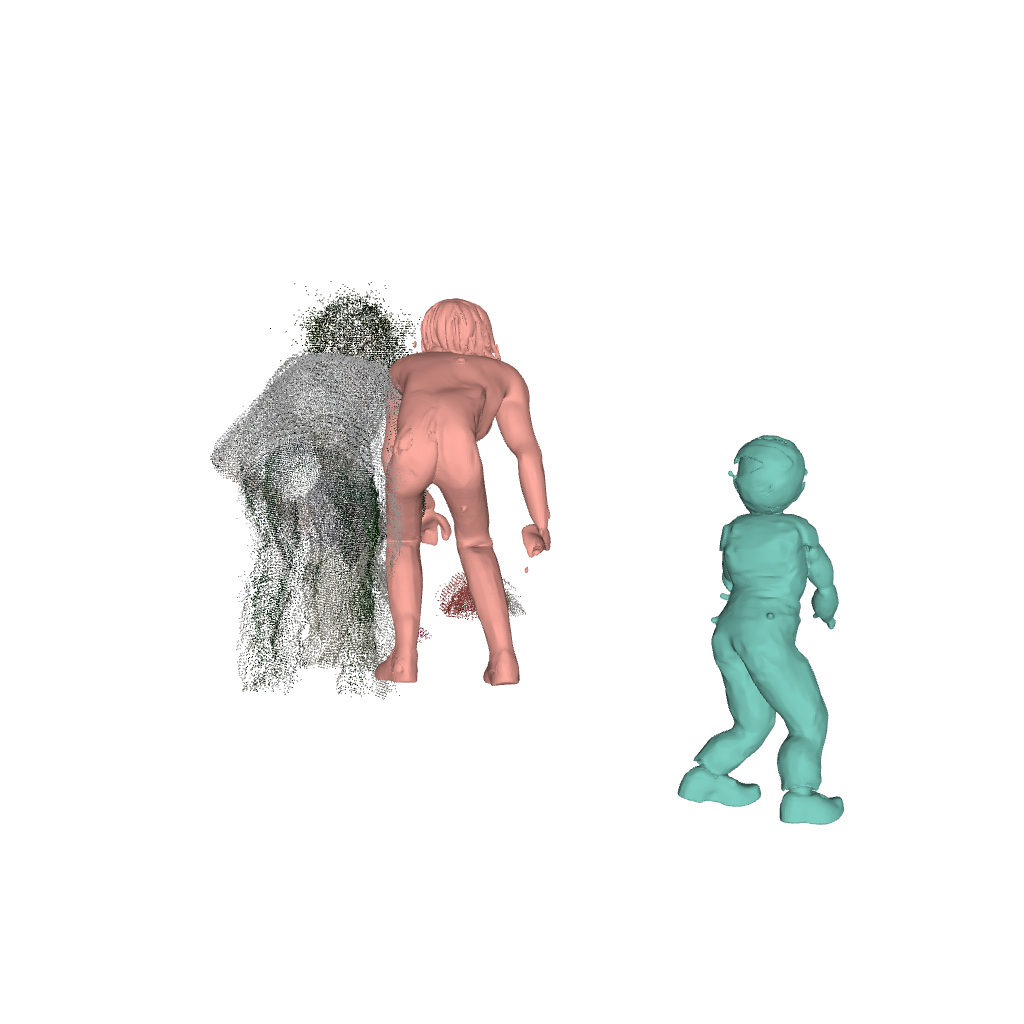} &
    \includegraphics[width=0.18\linewidth]{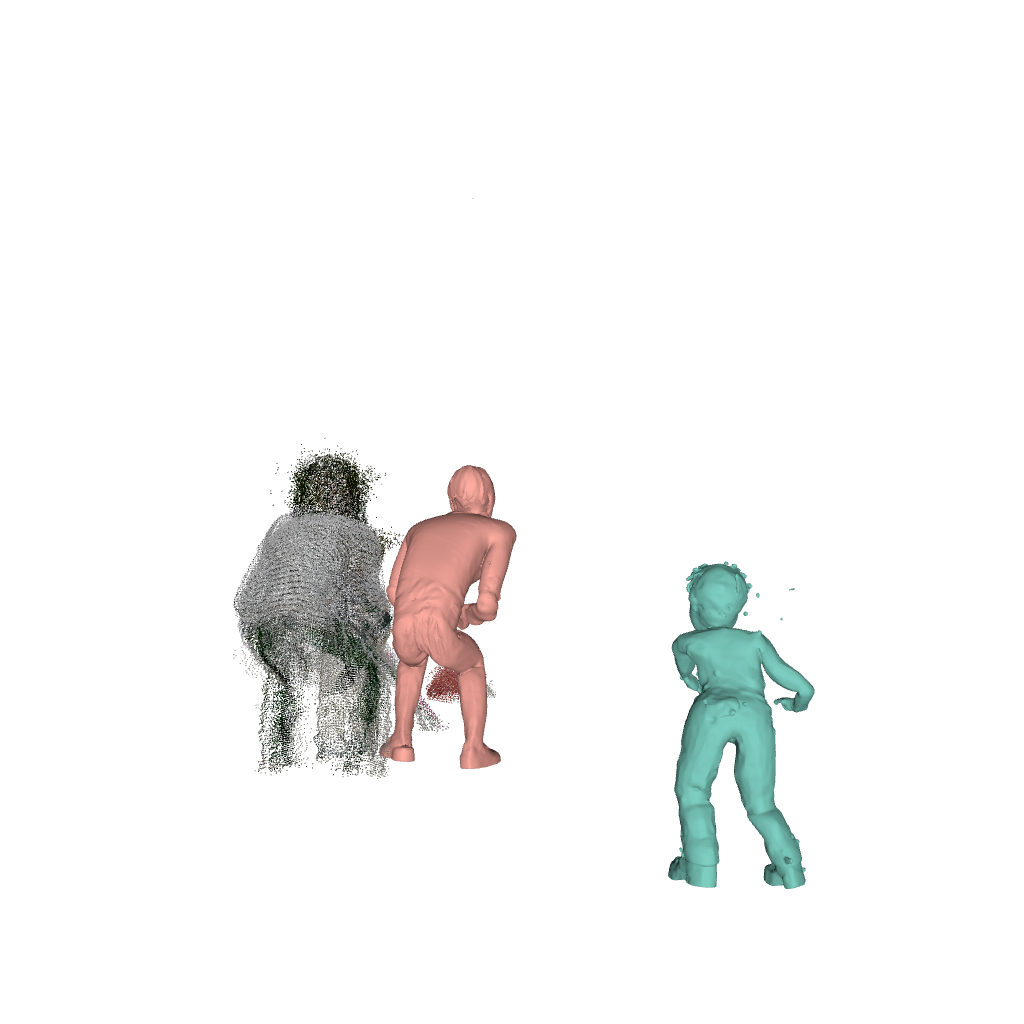} &
    \includegraphics[width=0.18\linewidth]{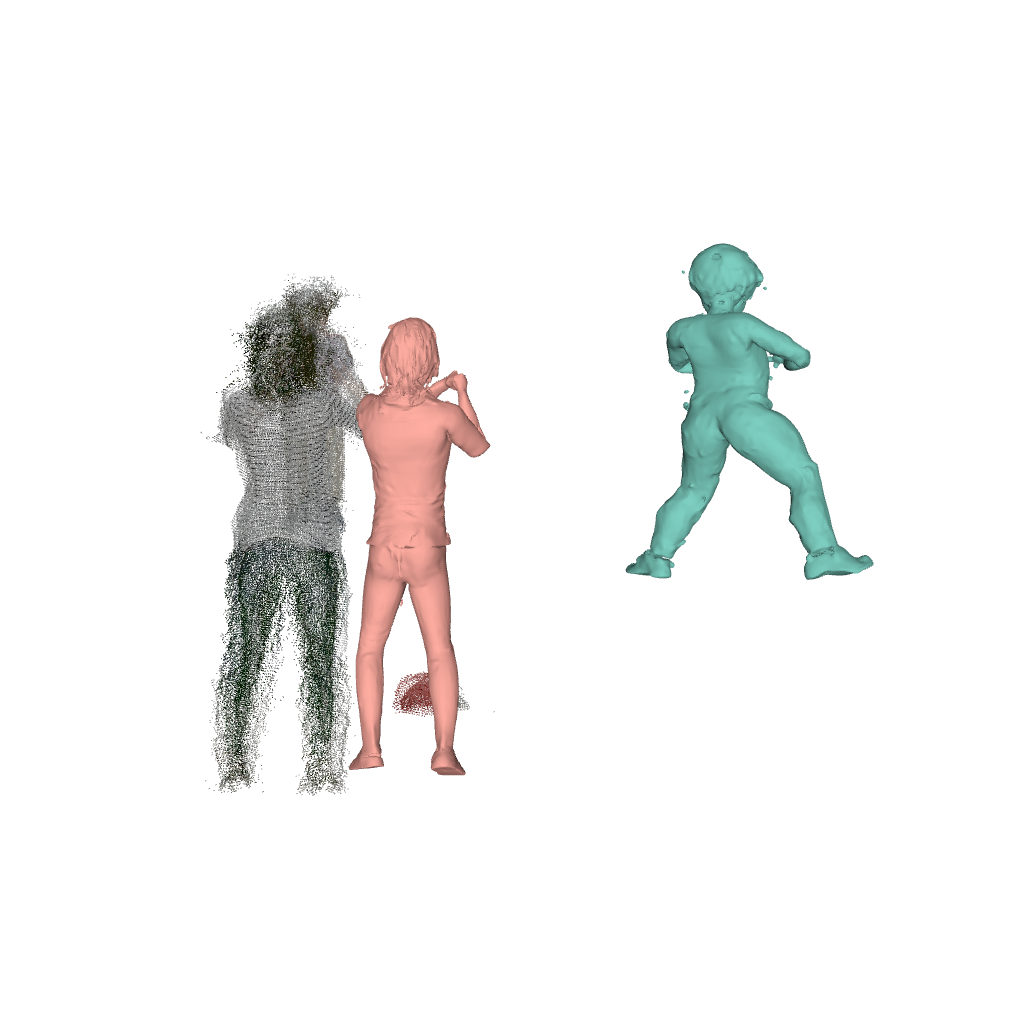} \\

  \end{tabular}
  % ==================== TABLE END ====================
  
  \vspace{-3pt}
  \caption{
    Ablation study on mixing components across five time steps. The top row shows the ground truth frames, followed by two views with our mixing strategy and two views without. Gray points denote the ground truth point cloud.
  }
  \label{fig:supp_ian_look_3_matching}
\end{figure}

\begin{figure}[h!]
  \centering
  % --- MODIFICATIONS ---
  \setlength{\tabcolsep}{0pt}      % NO horizontal space between columns
  \renewcommand{\arraystretch}{0} % NO vertical stretching of rows

  % ==================== TABLE START ====================
  \begin{tabular}{@{}c@{}ccccc@{}} % Use @{} to remove all column padding

    % --- Row 1: GT Frames ---
    \rotatebox{90}{\small ~~~Input} & % Rotated label with manual spacing
    \includegraphics[width=0.18\linewidth]{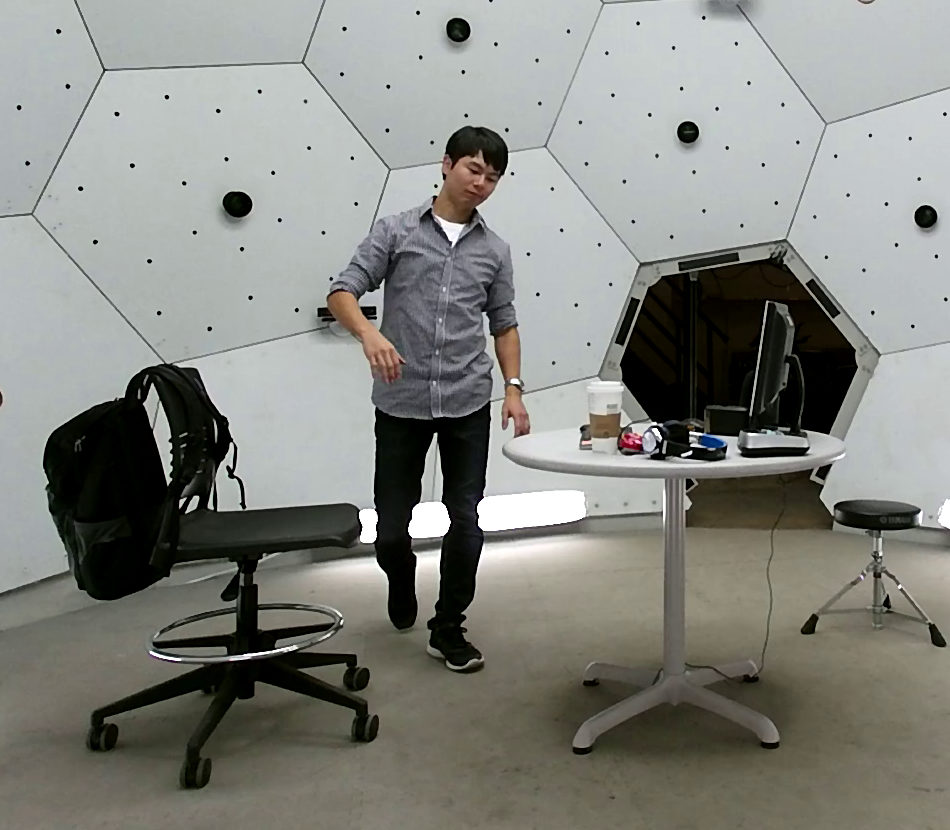} &
    \includegraphics[width=0.18\linewidth]{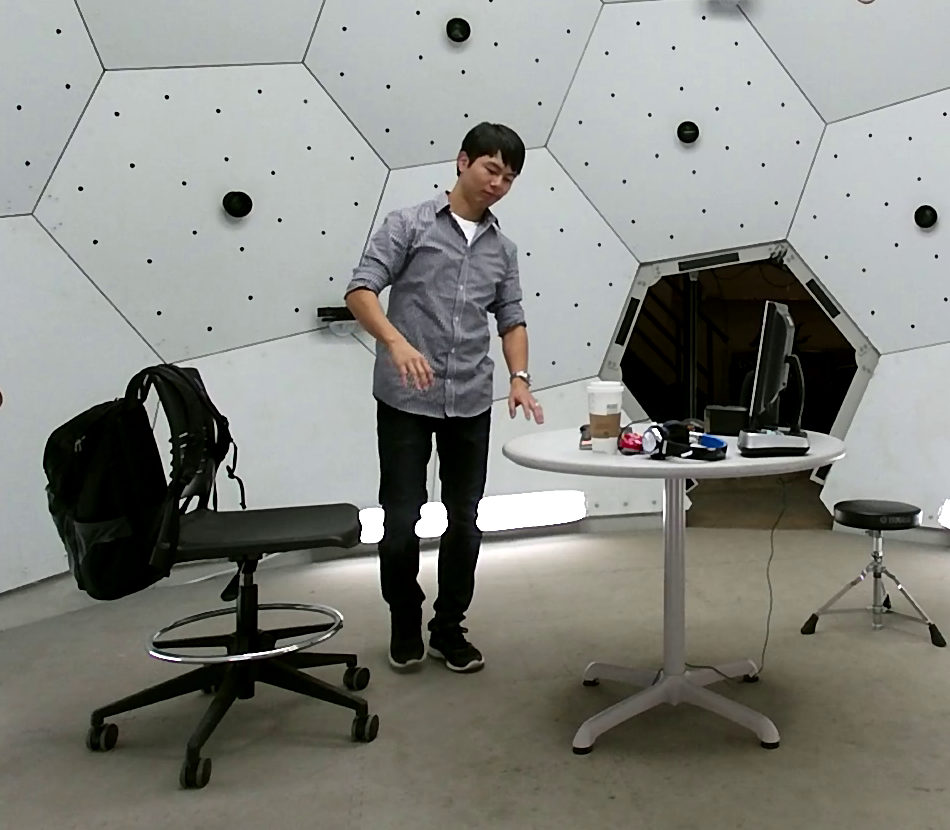} &
    \includegraphics[width=0.18\linewidth]{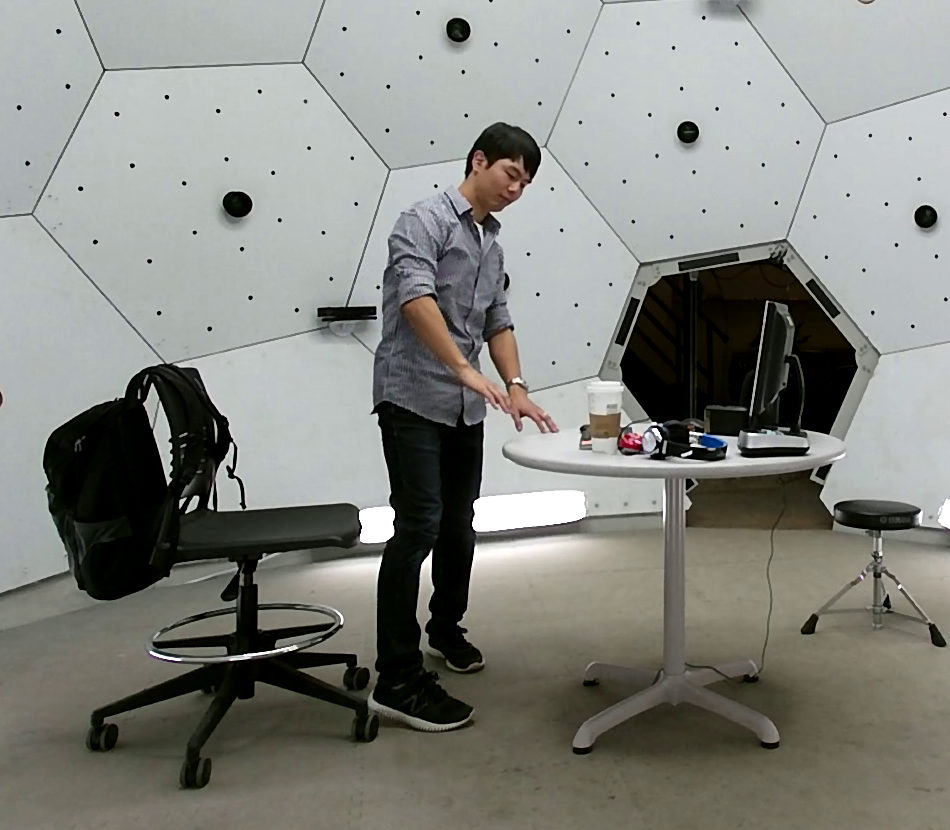} &
    \includegraphics[width=0.18\linewidth]{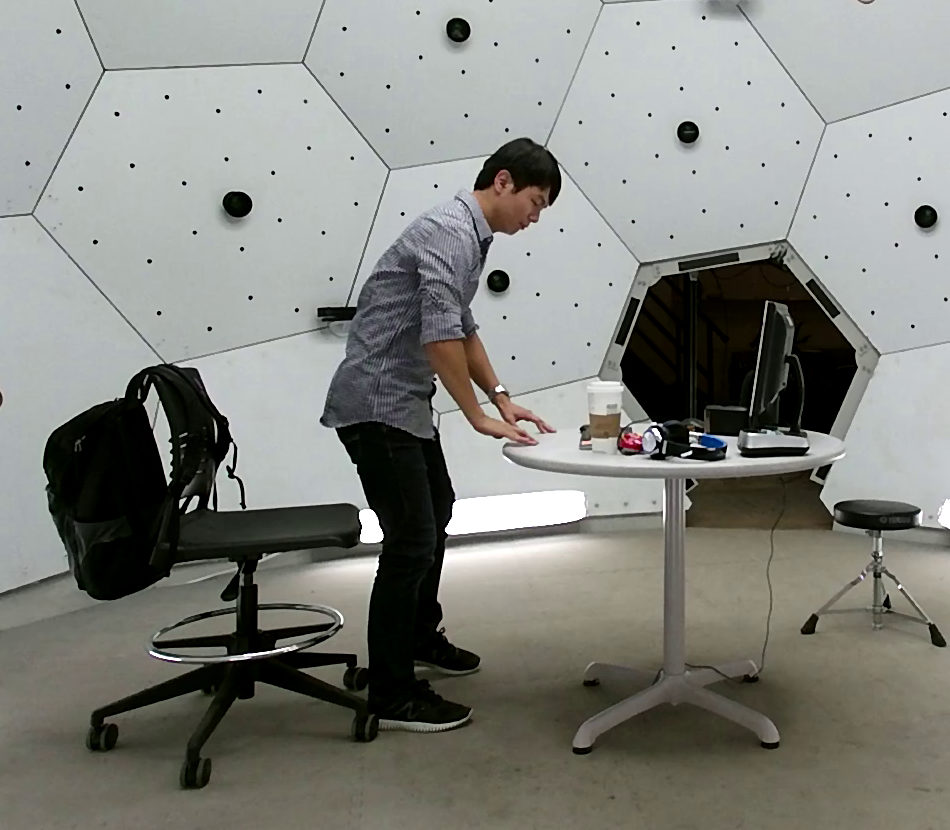} &
    \includegraphics[width=0.18\linewidth]{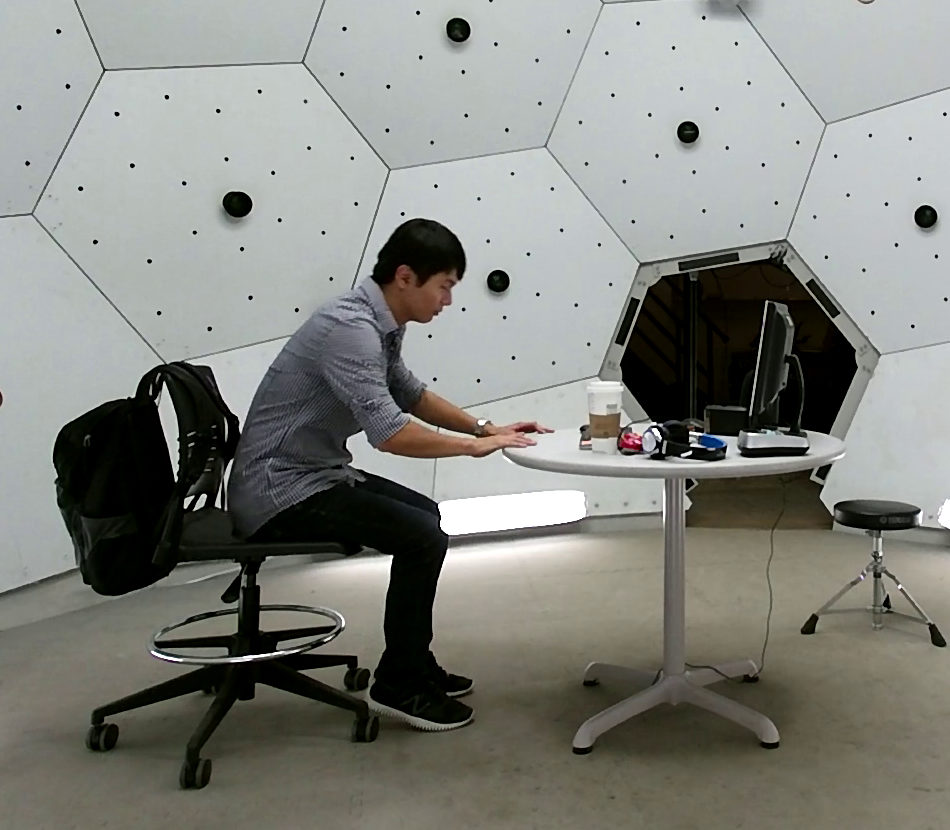} \\

    % --- Row 2: w/ Mixing View 1 ---
    \rotatebox{90}{\small w/ Mixing} &
    \includegraphics[width=0.18\linewidth]{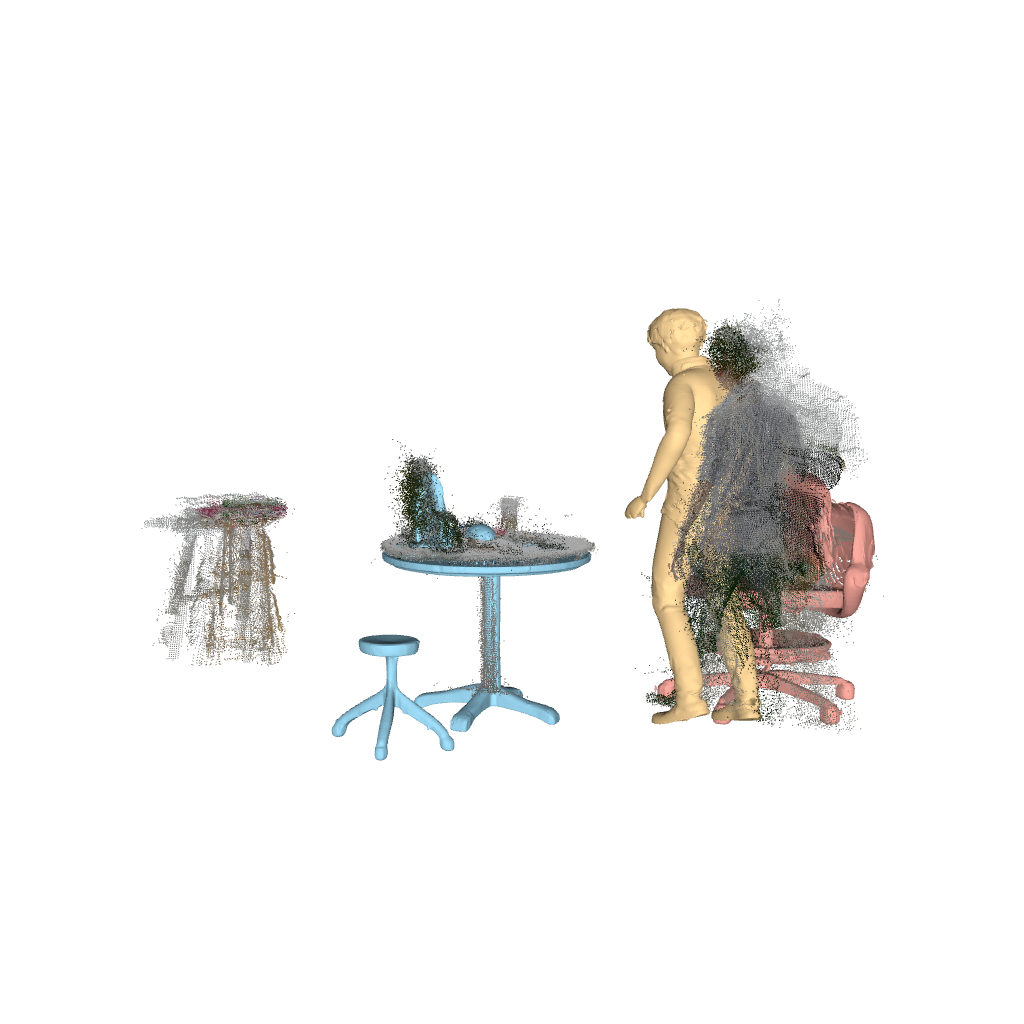} &
    \includegraphics[width=0.18\linewidth]{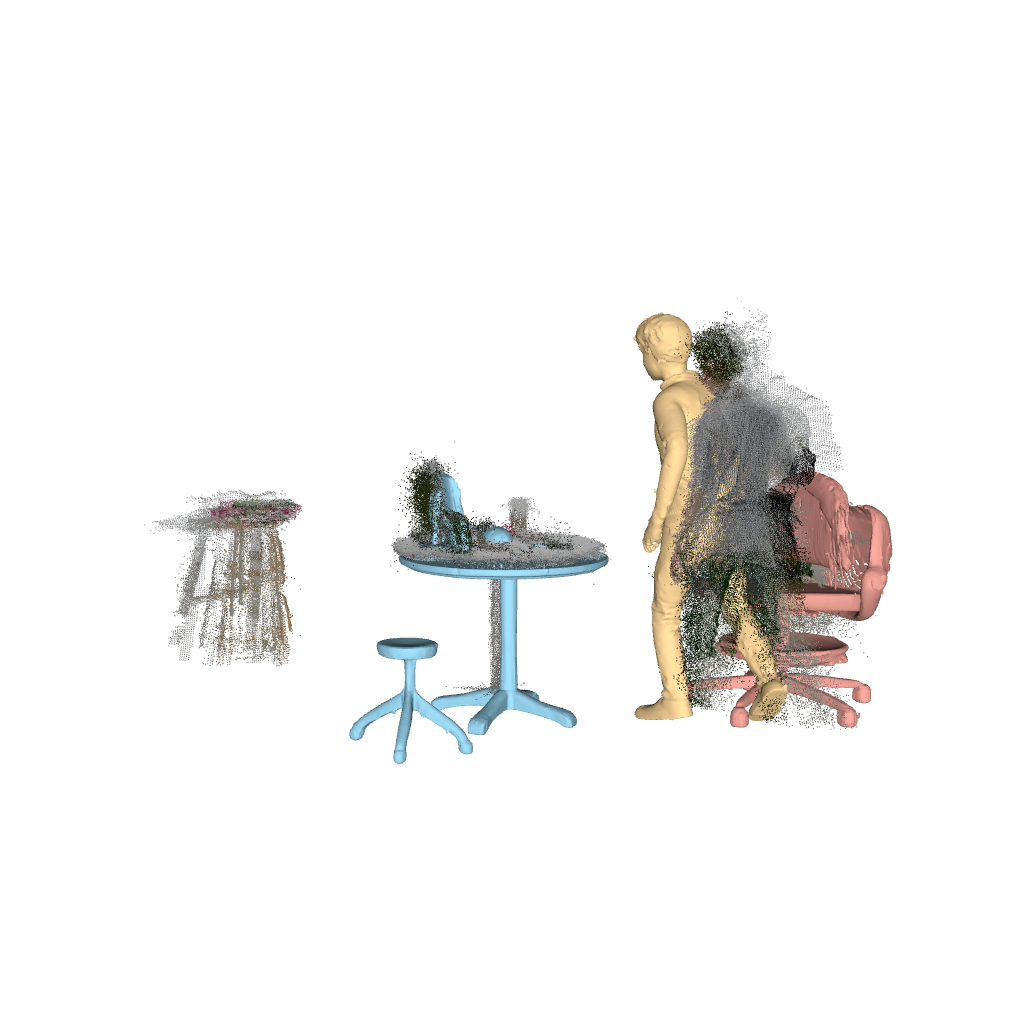} &
    \includegraphics[width=0.18\linewidth]{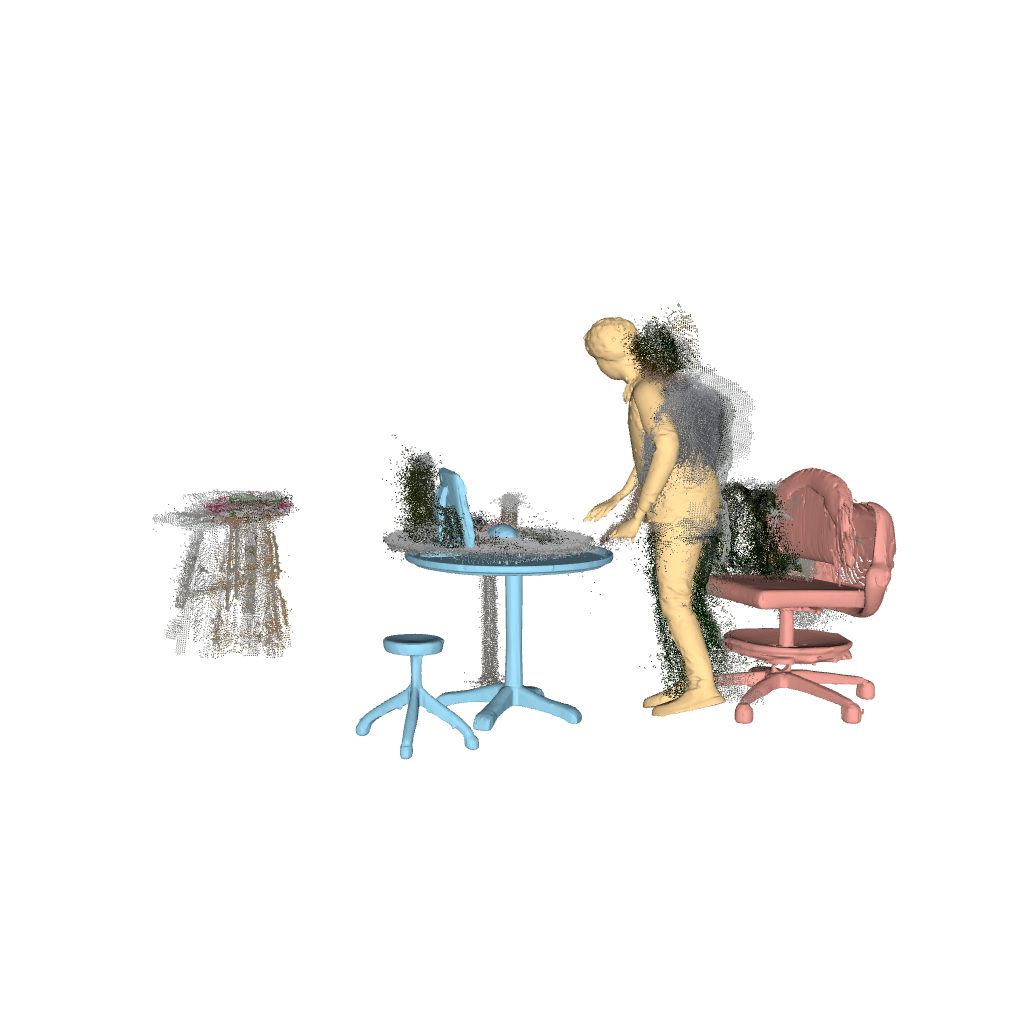} &
    \includegraphics[width=0.18\linewidth]{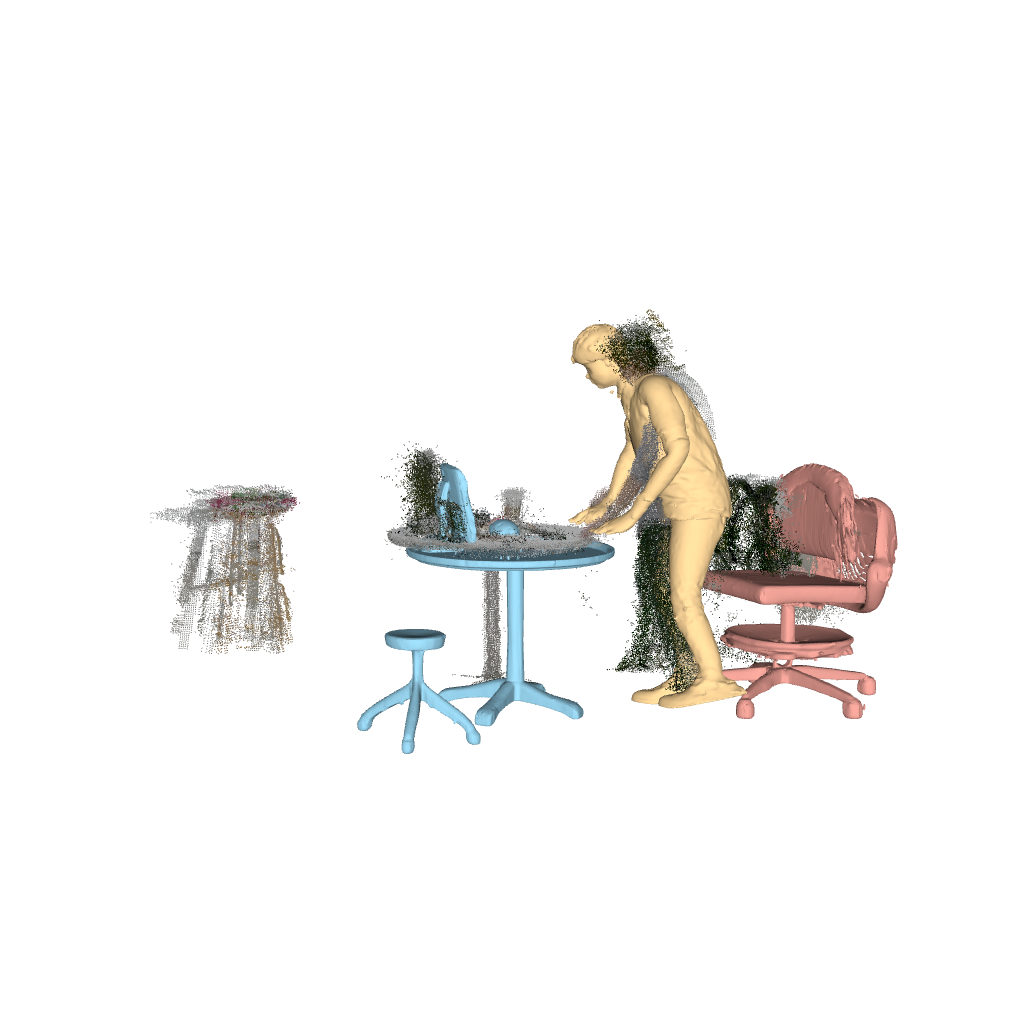} &
    \includegraphics[width=0.18\linewidth]{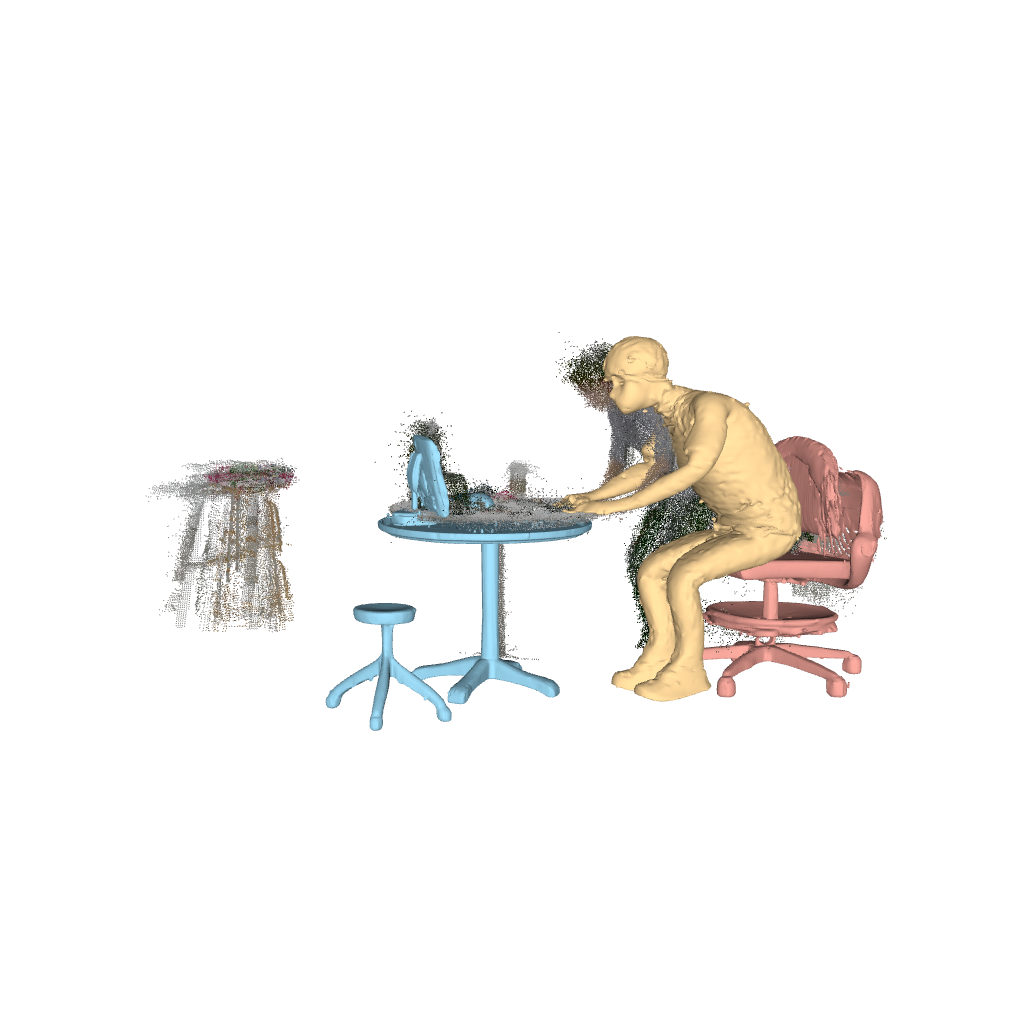} \\

    % --- Row 3: w/ Mixing View 2 ---
    \rotatebox{90}{\small} &
    \includegraphics[width=0.18\linewidth]{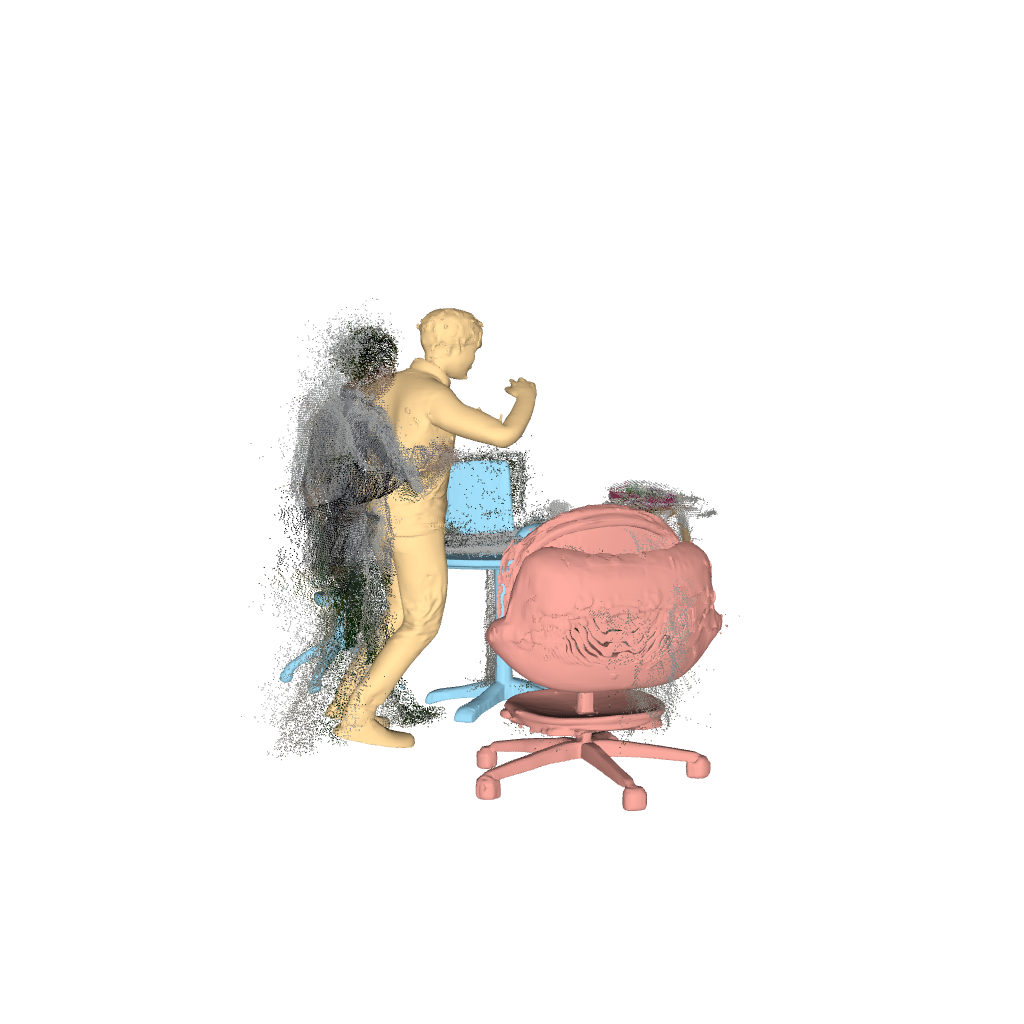} &
    \includegraphics[width=0.18\linewidth]{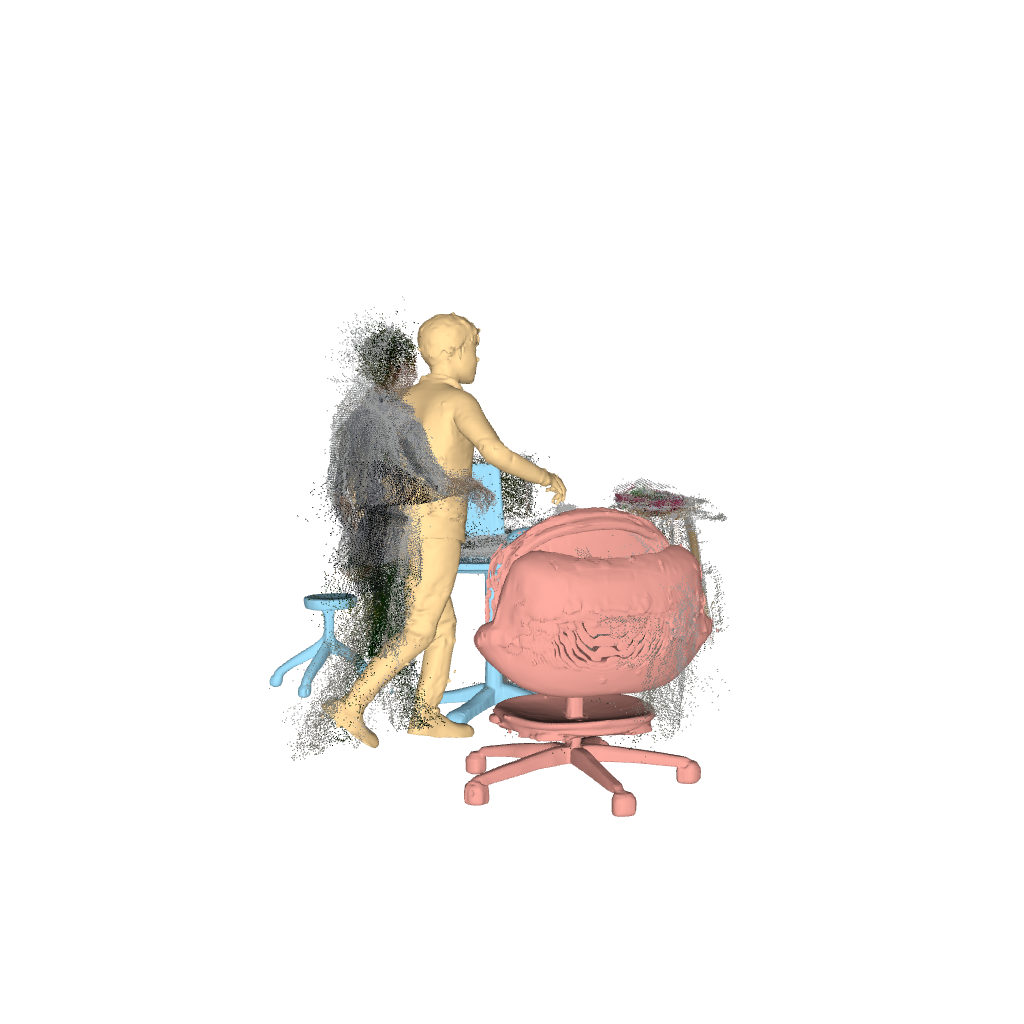} &
    \includegraphics[width=0.18\linewidth]{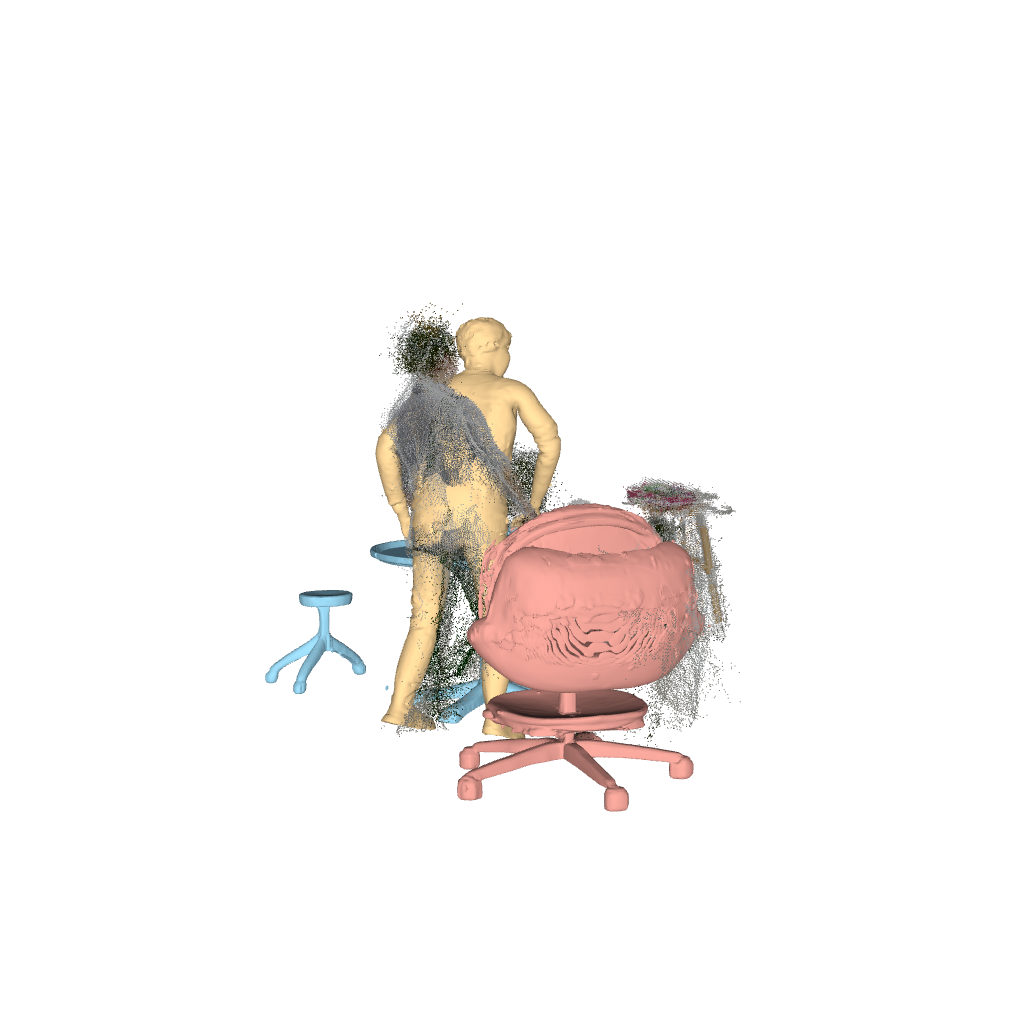} &
    \includegraphics[width=0.18\linewidth]{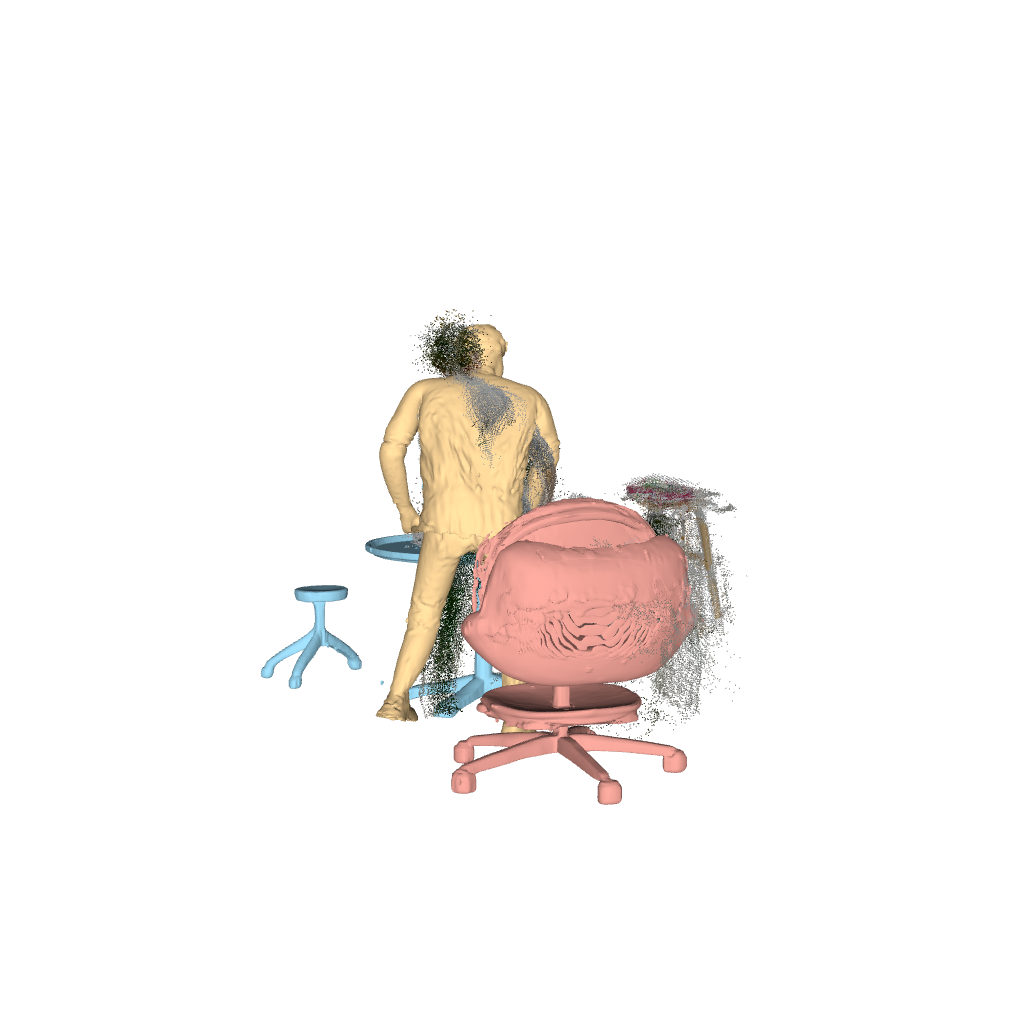} &
    \includegraphics[width=0.18\linewidth]{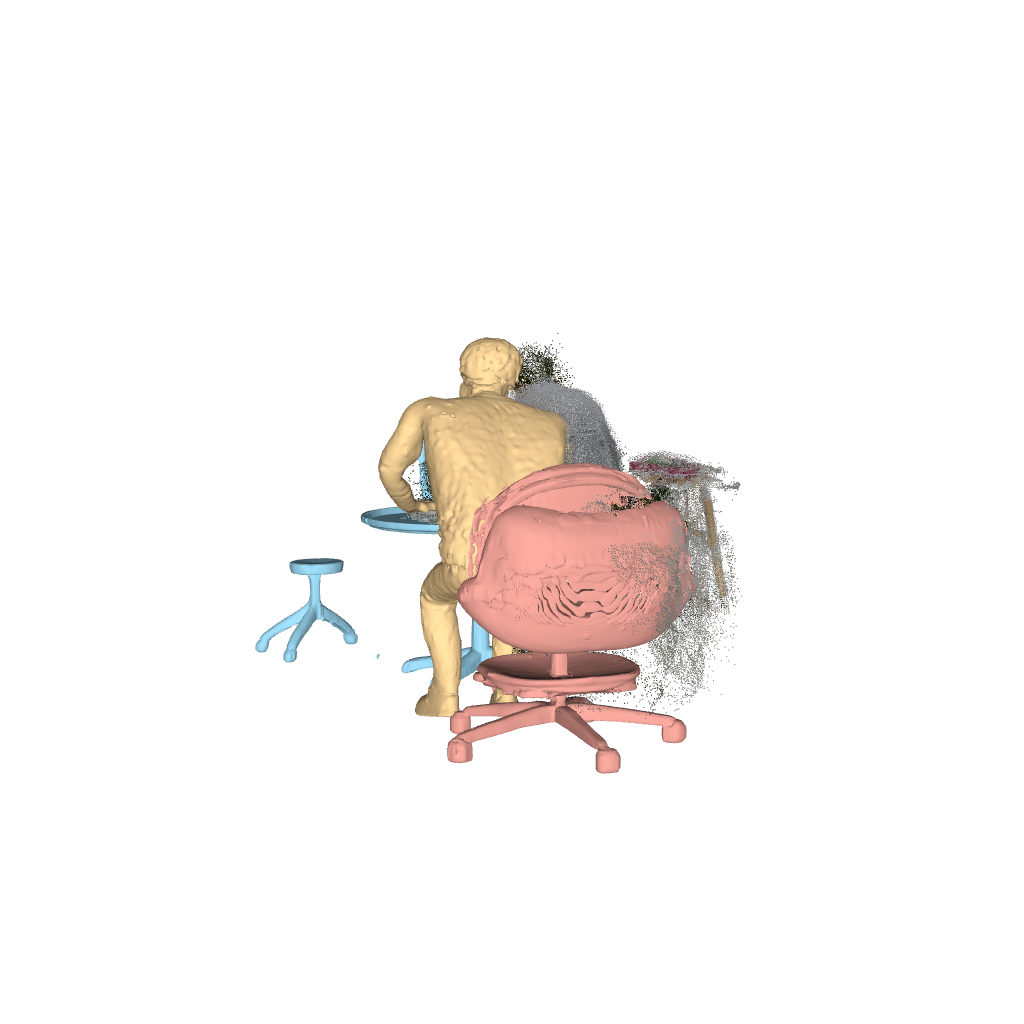} \\

    % --- Row 4: w/o Mixing View 1 ---
    \rotatebox{90}{\small w/o Mixing} &
    \includegraphics[width=0.18\linewidth]{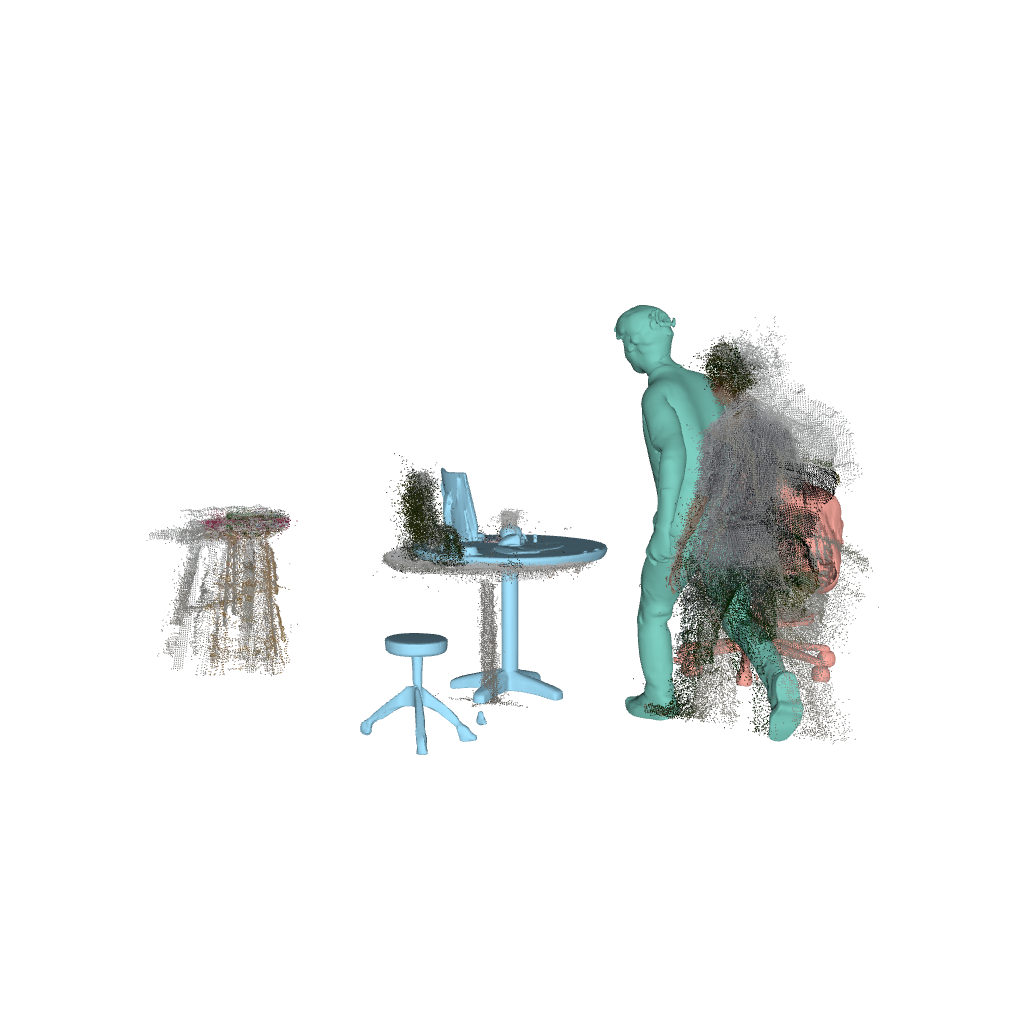} &
    \includegraphics[width=0.18\linewidth]{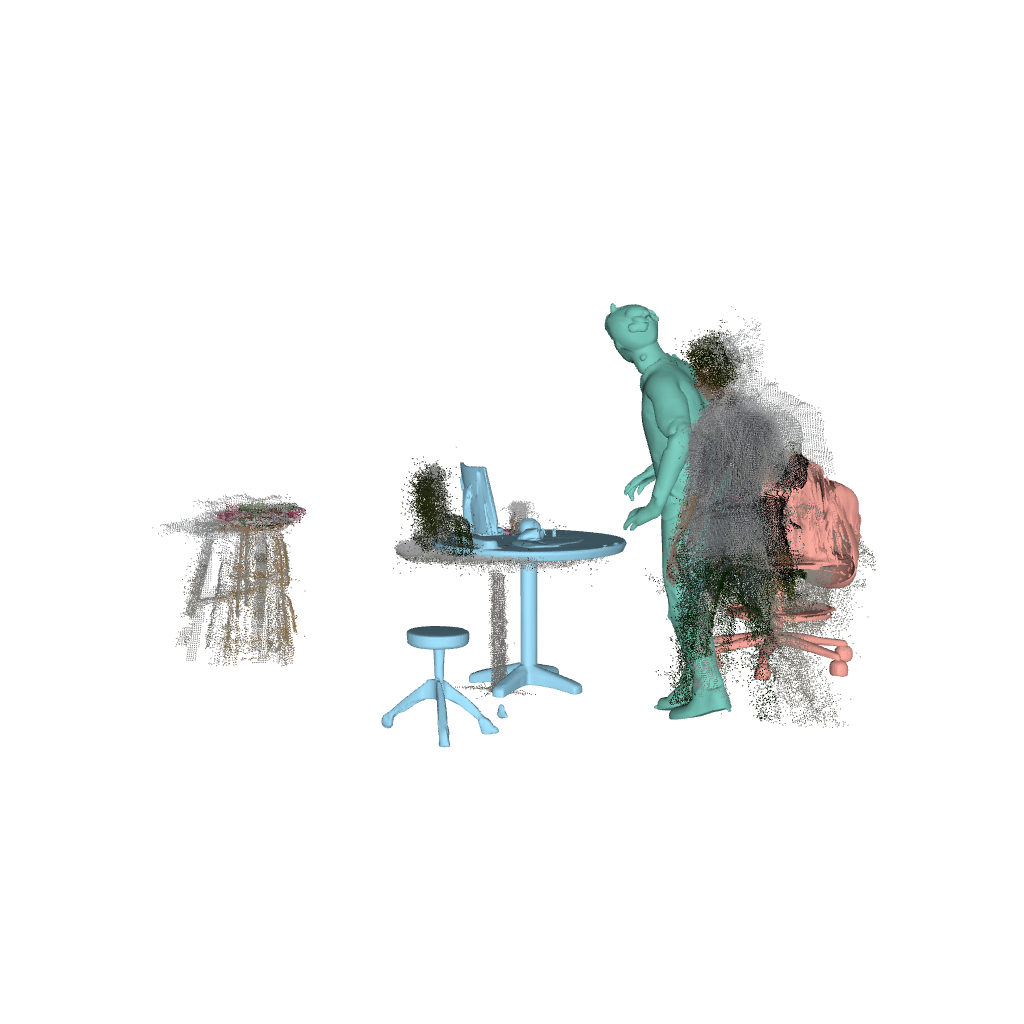} &
    \includegraphics[width=0.18\linewidth]{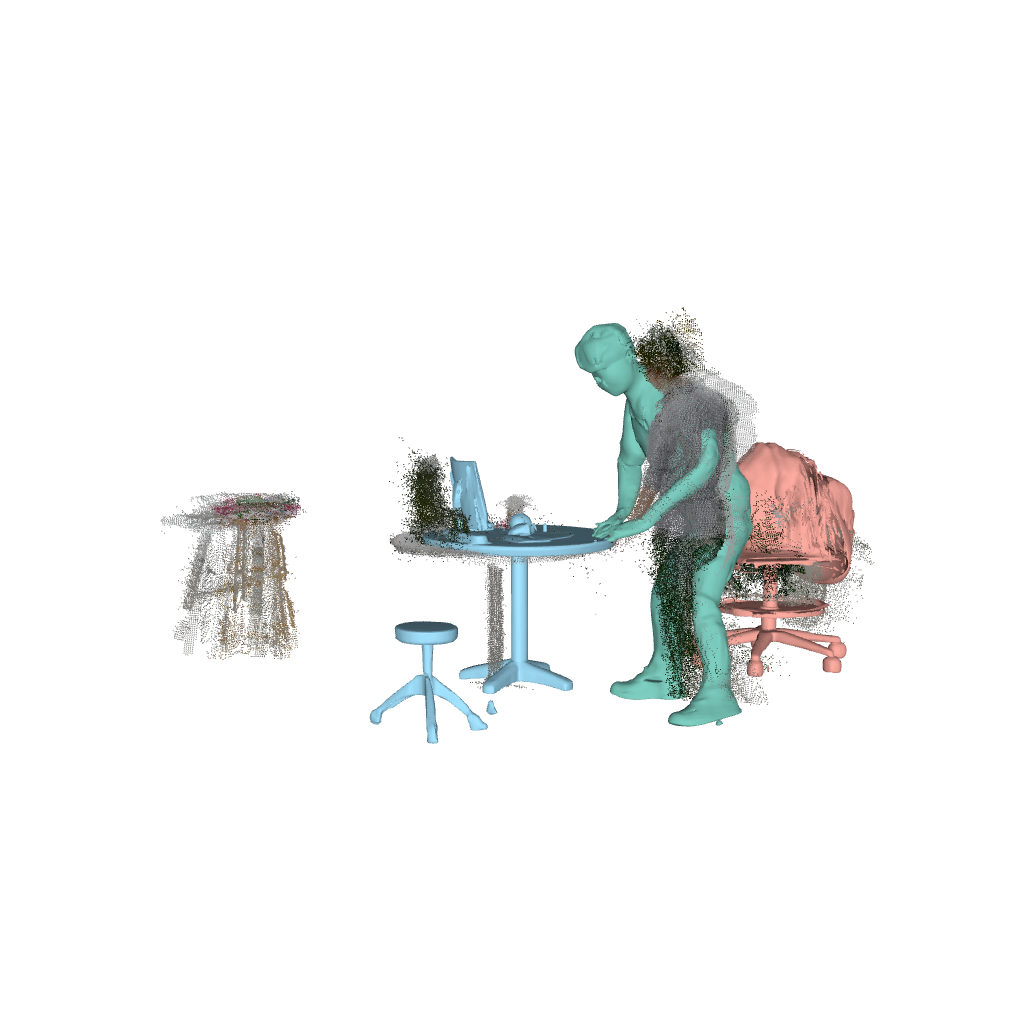} &
    \includegraphics[width=0.18\linewidth]{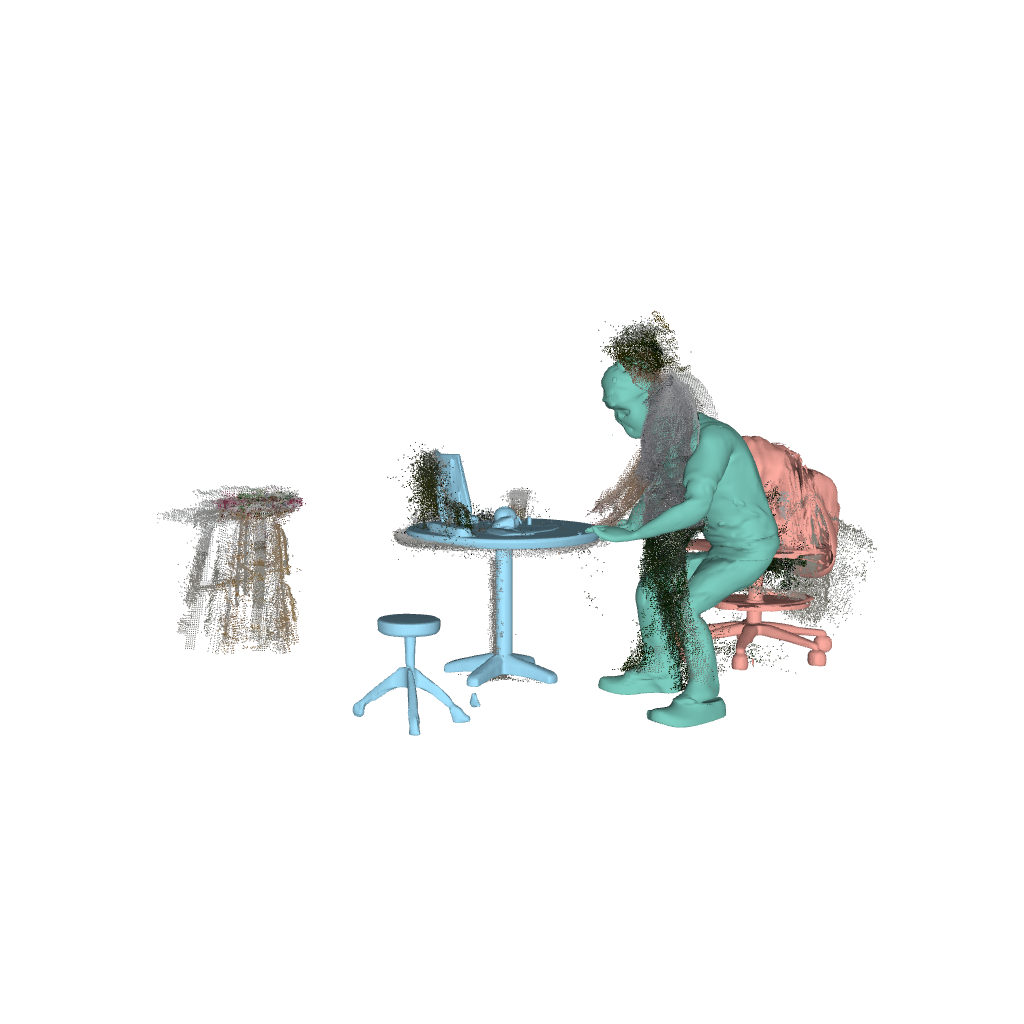} &
    \includegraphics[width=0.18\linewidth]{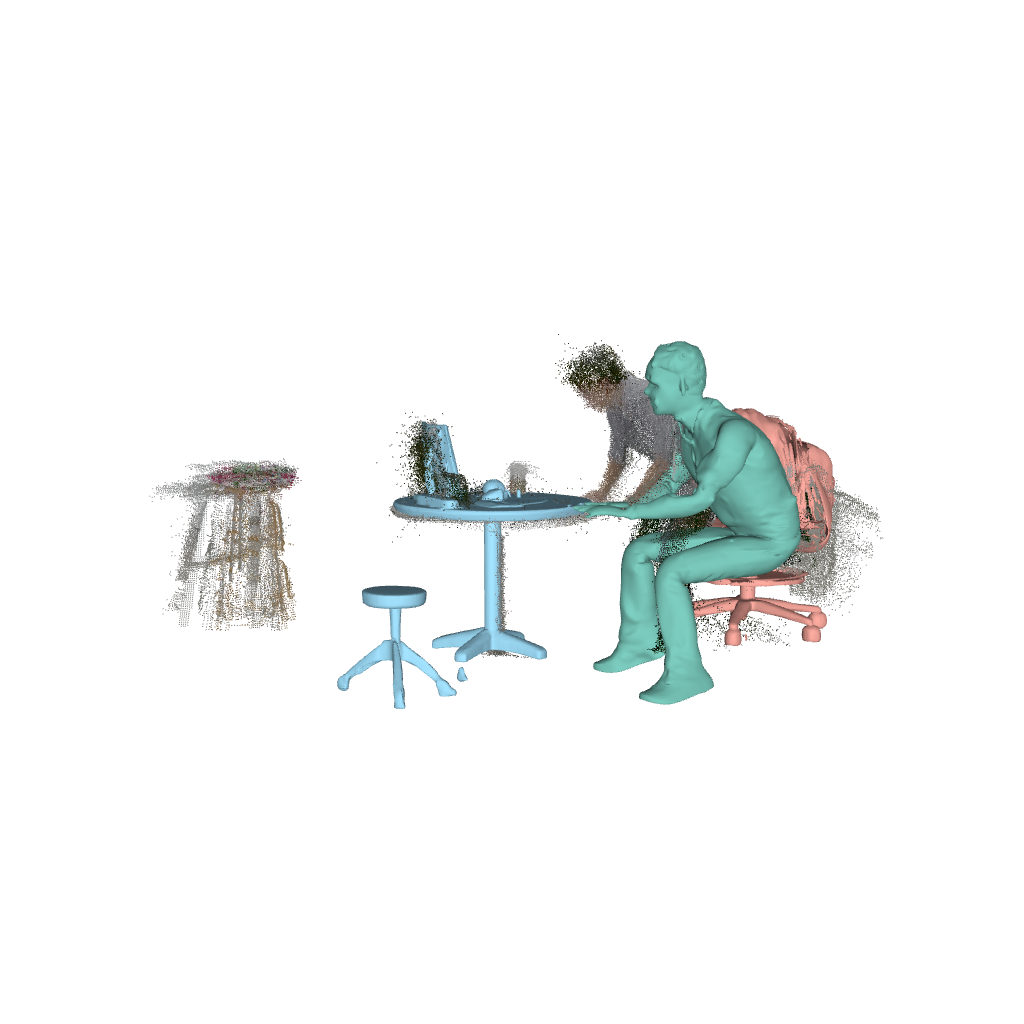} \\

    % --- Row 5: w/o Mixing View 2 ---
    \rotatebox{90}{\small} &
    \includegraphics[width=0.18\linewidth]{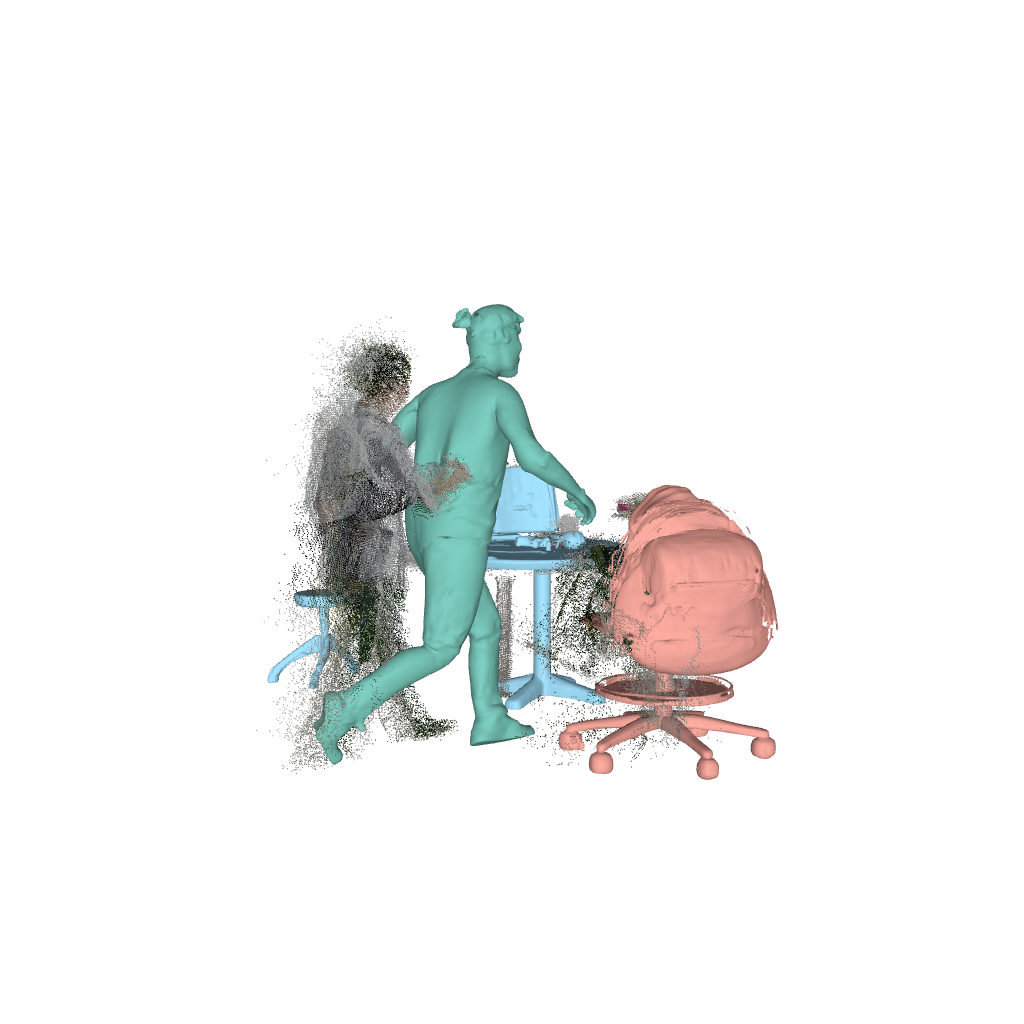} &
    \includegraphics[width=0.18\linewidth]{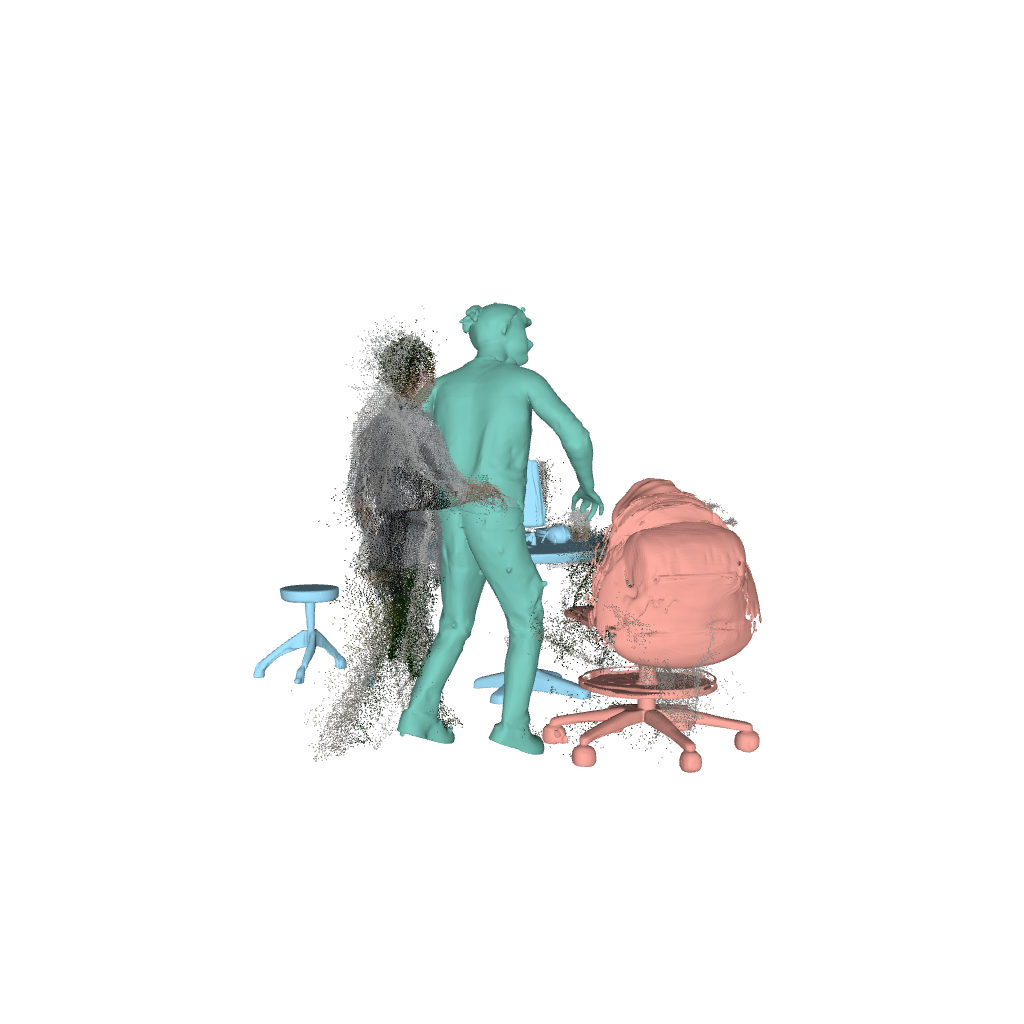} &
    \includegraphics[width=0.18\linewidth]{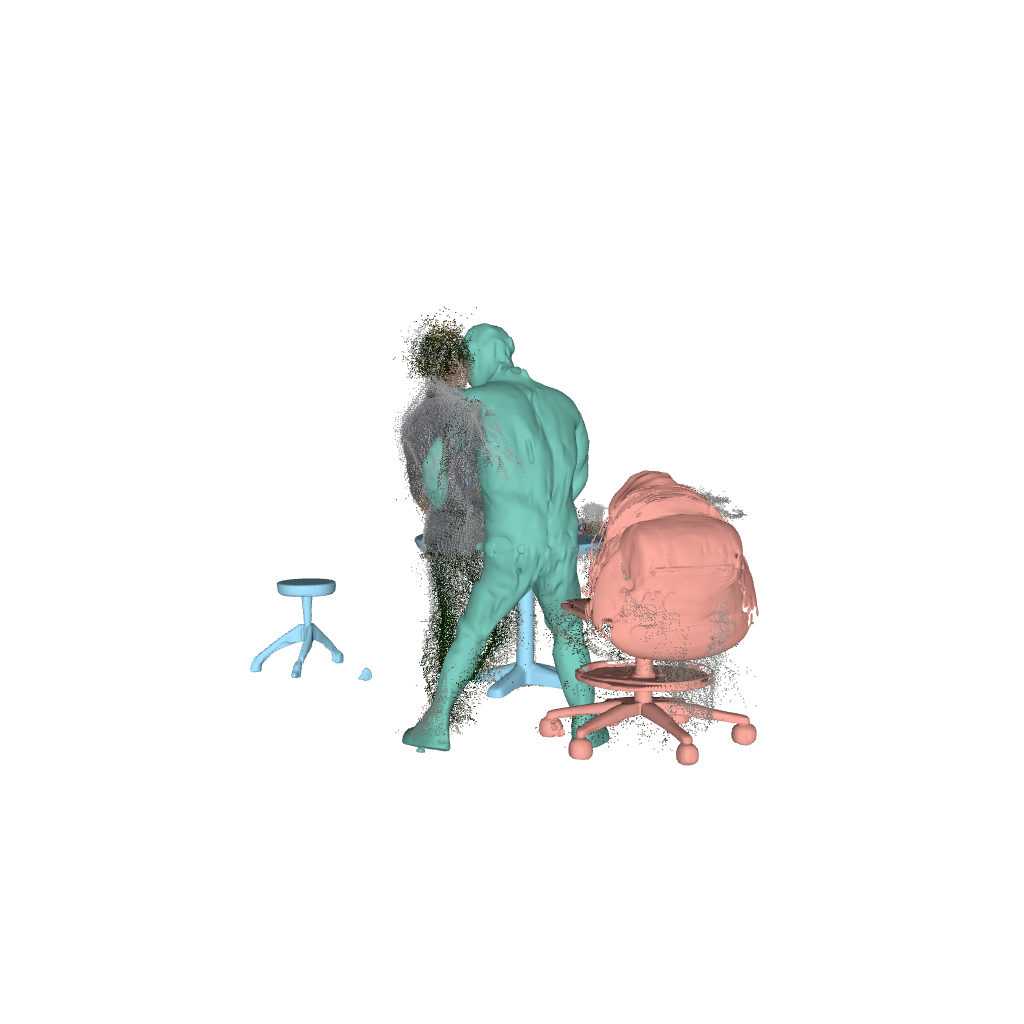} &
    \includegraphics[width=0.18\linewidth]{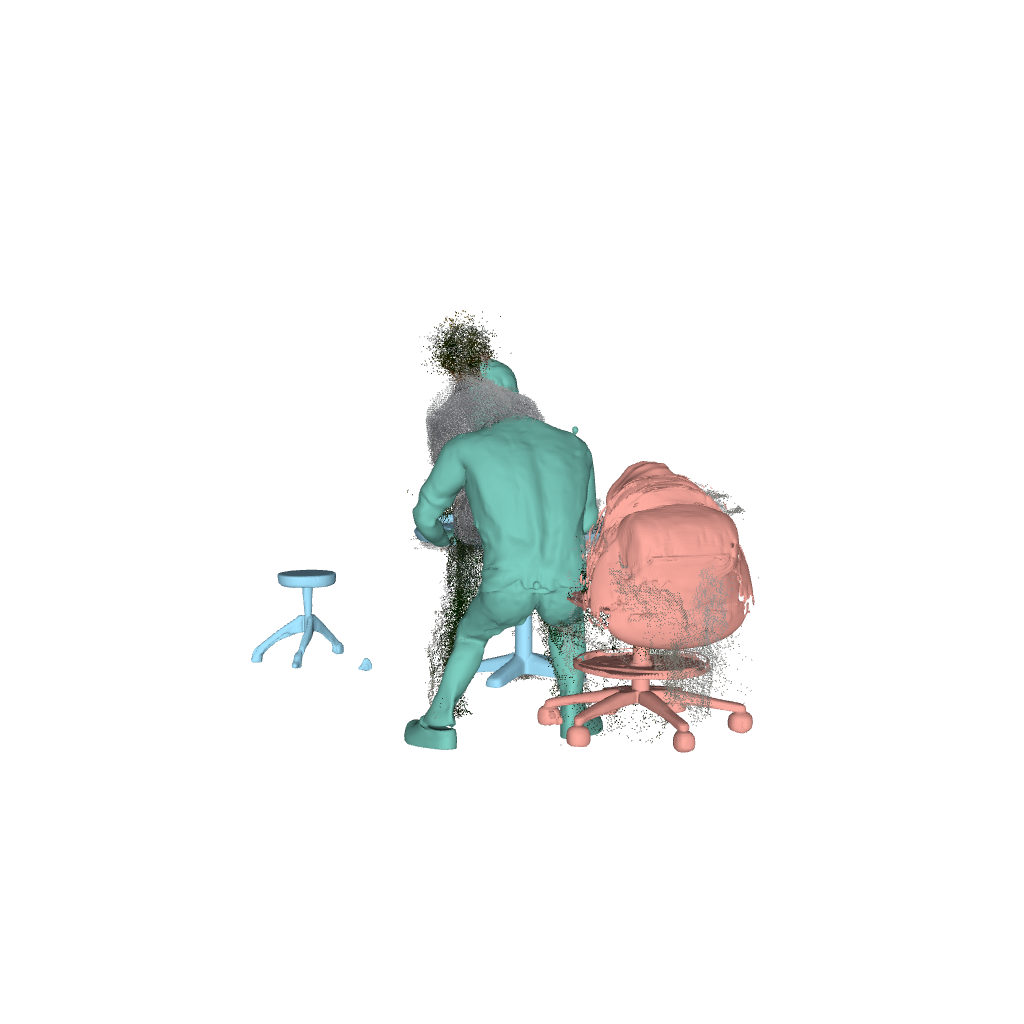} &
    \includegraphics[width=0.18\linewidth]{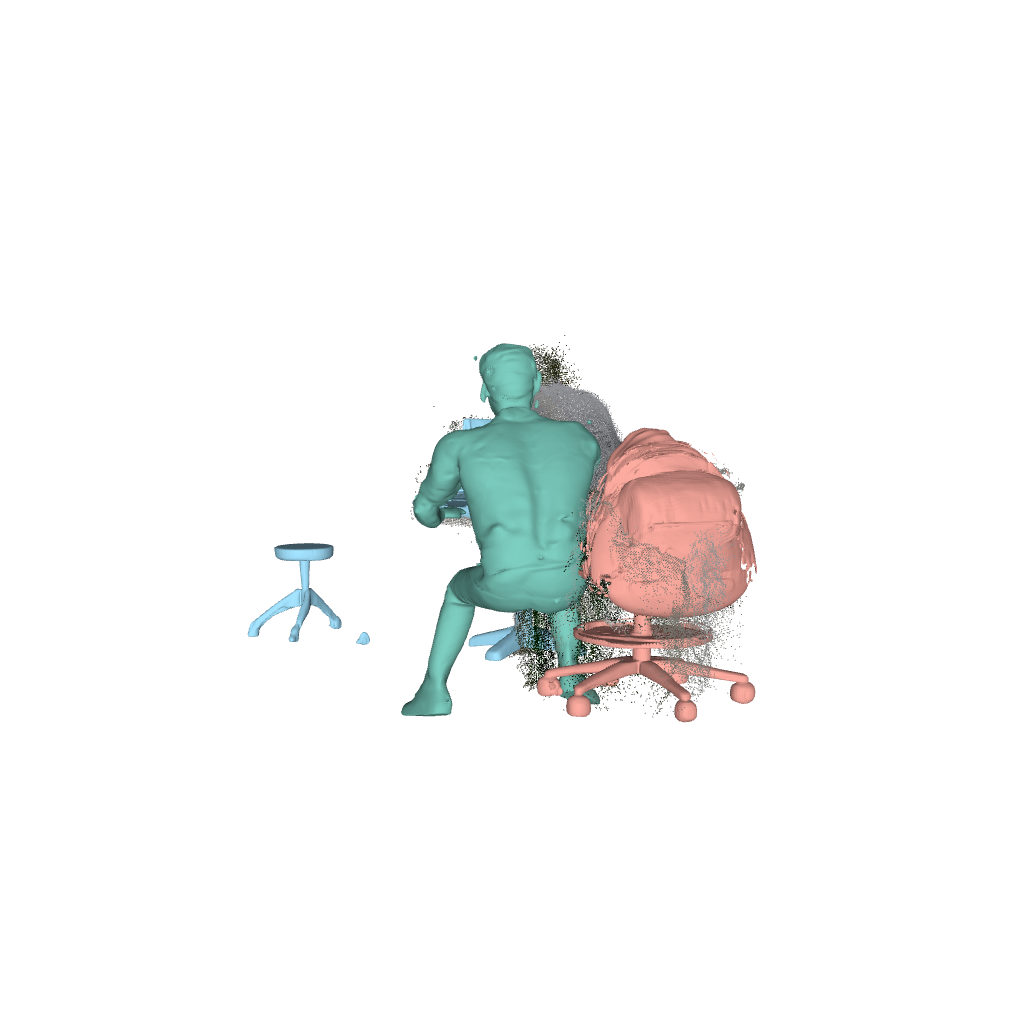} \\

  \end{tabular}
  % ==================== TABLE END ====================
  
  \vspace{-3pt}
  \caption{
    Ablation study on mixing components across five time steps. The top row shows the ground truth frames, followed by two views with our mixing strategy and two views without. Gray points denote the ground truth point cloud.
  }
  \label{fig:supp_office_sample_matching}
\end{figure}

Finally, we computed the Chamfer Distance (CD) between our transformed per-frame reconstructions and the GT point clouds. This metric effectively measures how much the predicted motion deviates from the ground truth over time, given an initial registration. A lower accumulated CD indicates a more accurate and temporally consistent motion prediction. Visual examples of these per-frame alignments are shown in \cref{fig:supp_toddler_ian_matching}, \cref{fig:supp_ian_look_3_matching} and \cref{fig:supp_office_sample_matching}. We can observe that not only registered $1^{st}$ frame, but also the remaining frames align well with the GT pointcloud despite the motion, with surprisingly small drift.

\newpage
\newpage

\section{Scalability with Respect to Object Count}
COM4D supports denoising with up to 8 parts, and up to 16 parts with additional finetuning. In the static dataset, scenes contain on average 6.19 objects, with the most common configuration being 5 objects (16.67\%). 

We observe that performance is moderately sensitive to both the number of dynamic objects and the complexity of the static scene. As the number of interacting objects increases, the compositional reasoning task becomes more challenging, leading to gradual degradation in reconstruction quality.

\begin{figure}[h!]
    \section{Additional Qualitative Results on Compositional 4D}
    
  \centering
  % --- MODIFICATIONS ---
  \setlength{\tabcolsep}{0pt}      % NO horizontal space between columns
  \renewcommand{\arraystretch}{0} % NO vertical stretching of rows

  % ==================== TABLE START ====================
  \begin{tabular}{@{}c@{\hspace{2pt}}ccc@{}}

    % ==================== FIRST SAMPLE ====================
    % --- Row 1: Input ---
    \rotatebox{90}{\small ~~~~~Input} &
    \includegraphics[width=0.3\linewidth]{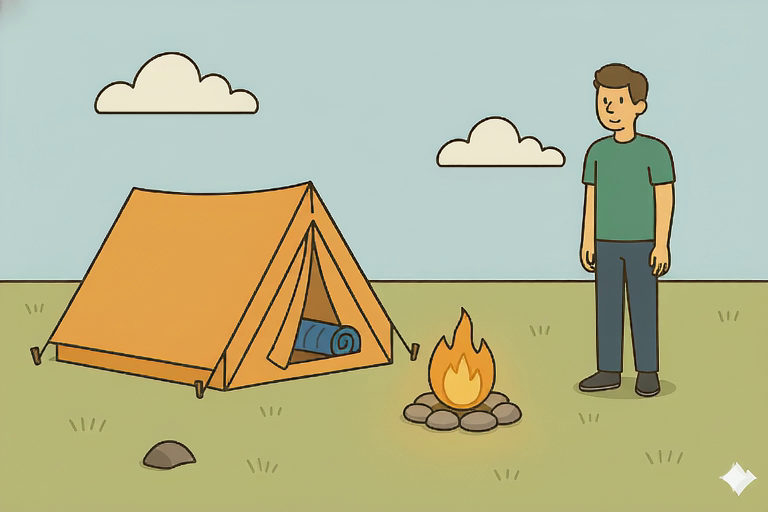} &
    \includegraphics[width=0.3\linewidth]{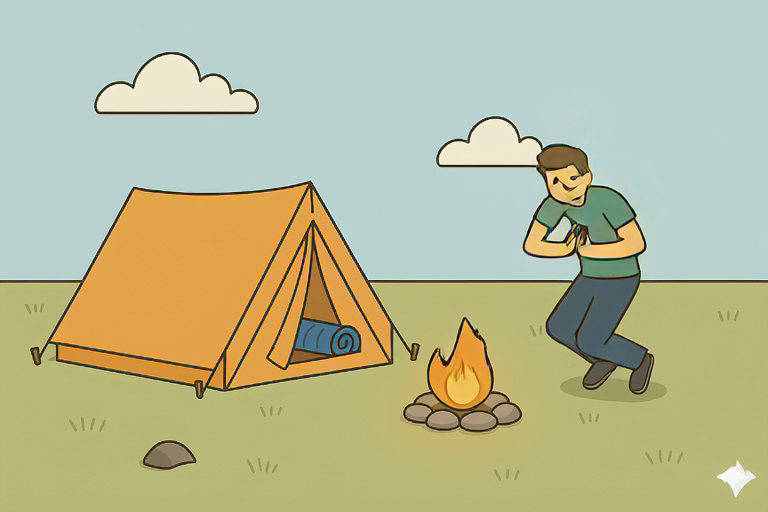} &
    \includegraphics[width=0.3\linewidth]{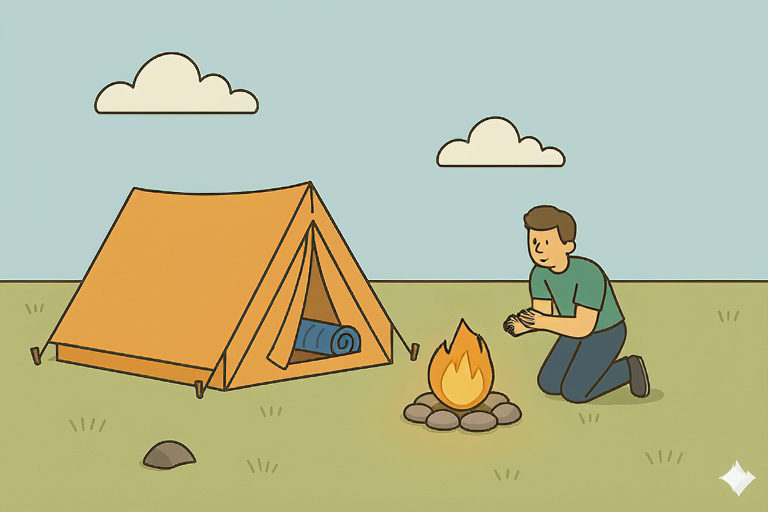} \\[-0.5pt]

    % --- Row 2: w/ Mixing ---
    \rotatebox{90}{\small ~~w/ Mixing} &
    \includegraphics[width=0.3\linewidth]{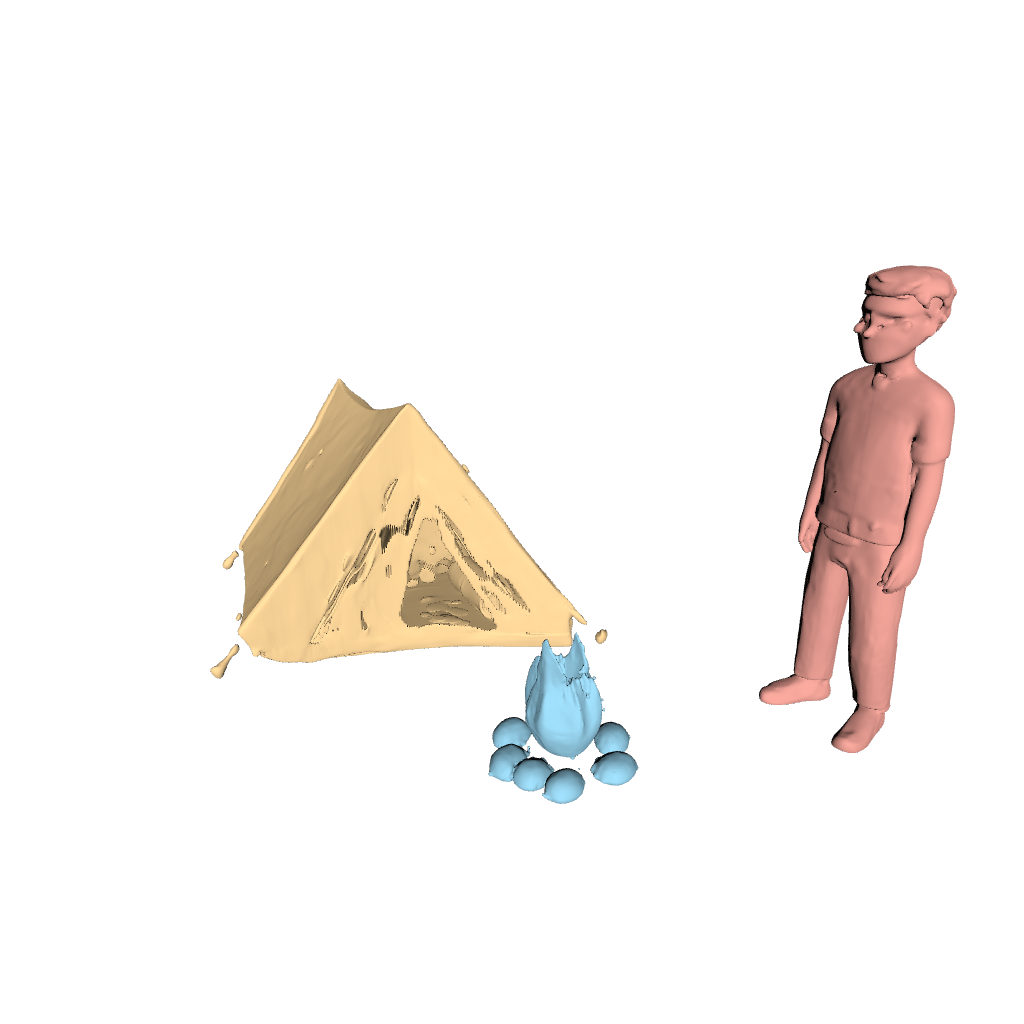} &
    \includegraphics[width=0.3\linewidth]{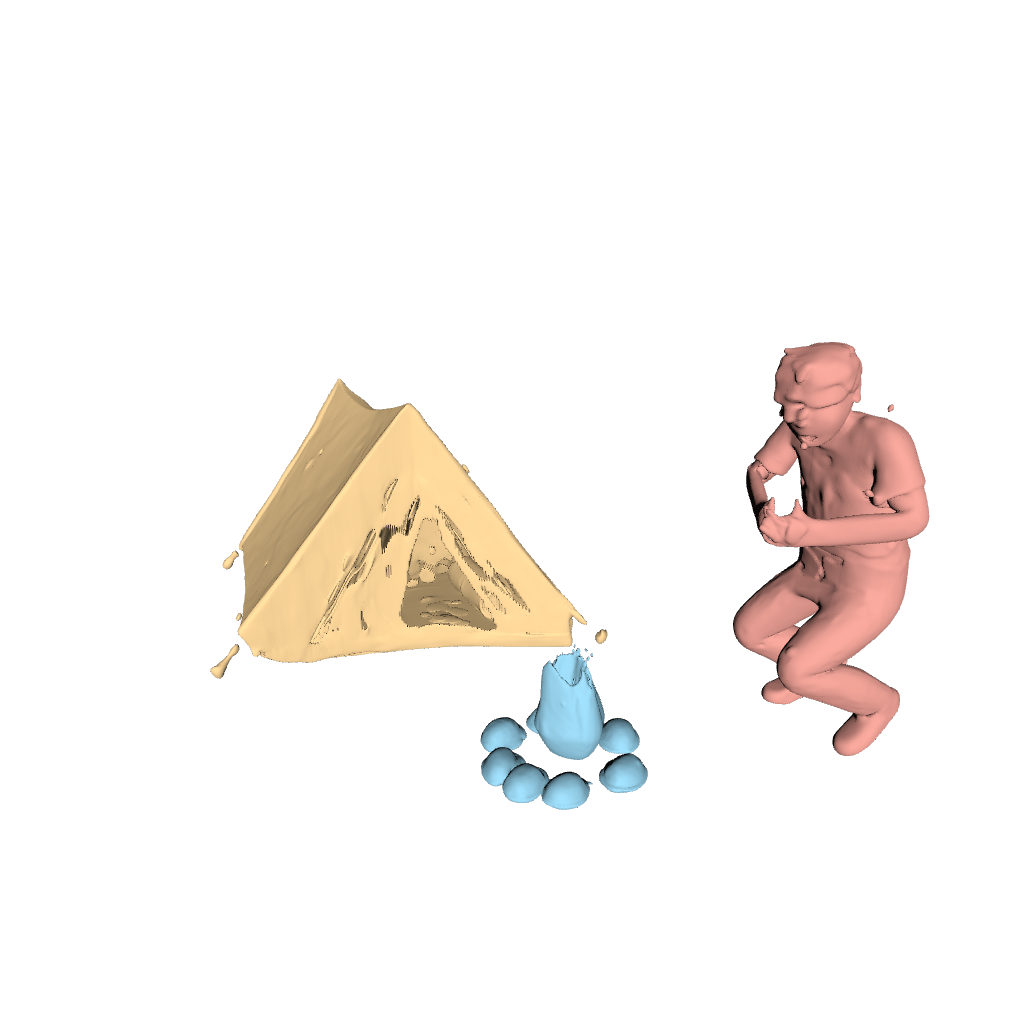} &
    \includegraphics[width=0.3\linewidth]{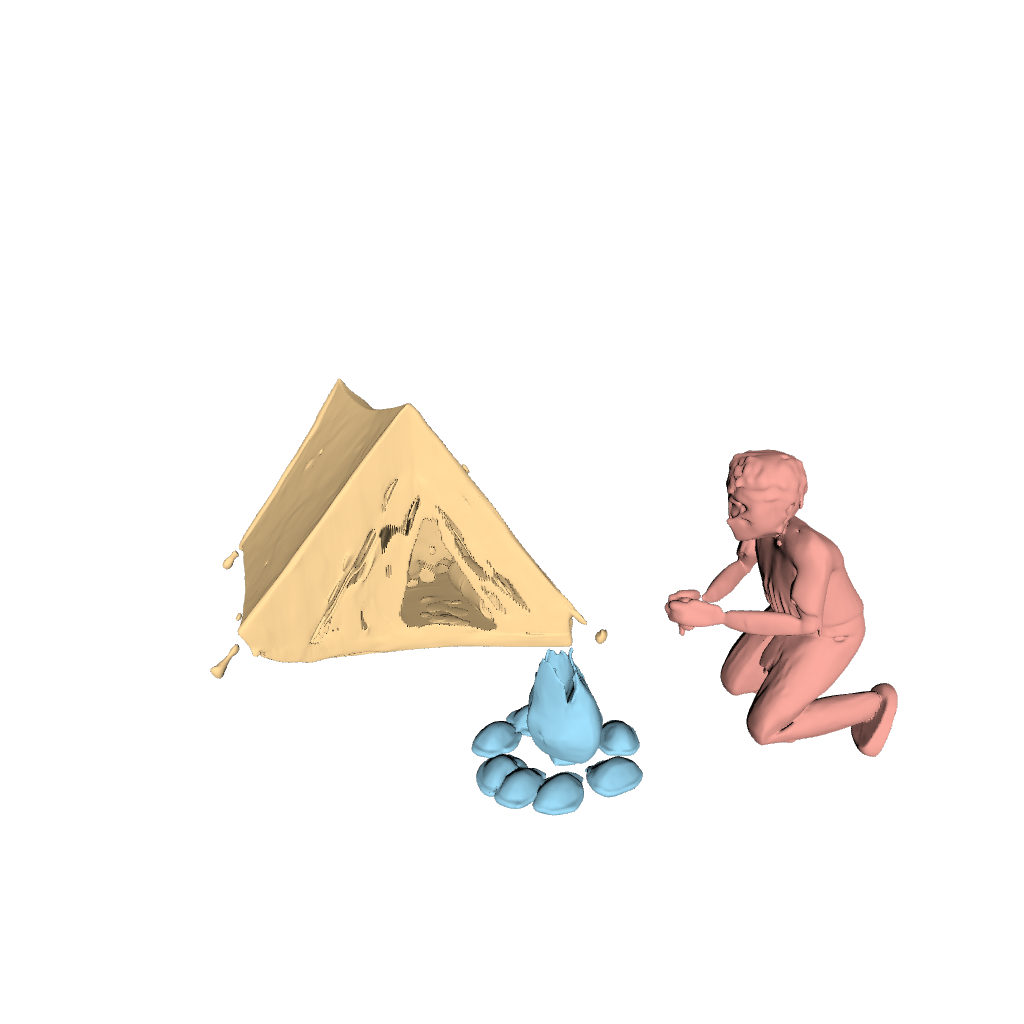} \\[-0.5pt]

    % --- Row 3: w/o Mixing ---
    \rotatebox{90}{\small ~~w/o Mixing} &
    \includegraphics[width=0.3\linewidth]{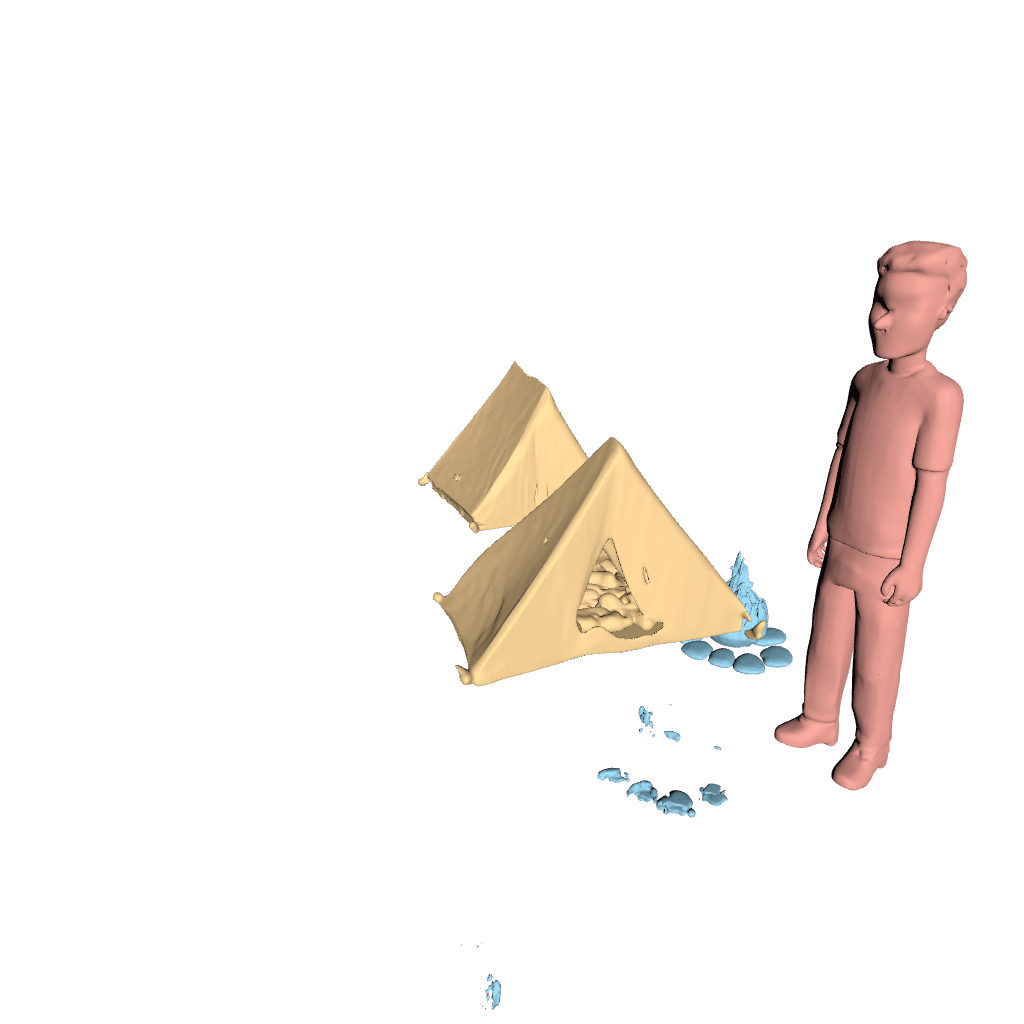} &
    \includegraphics[width=0.3\linewidth]{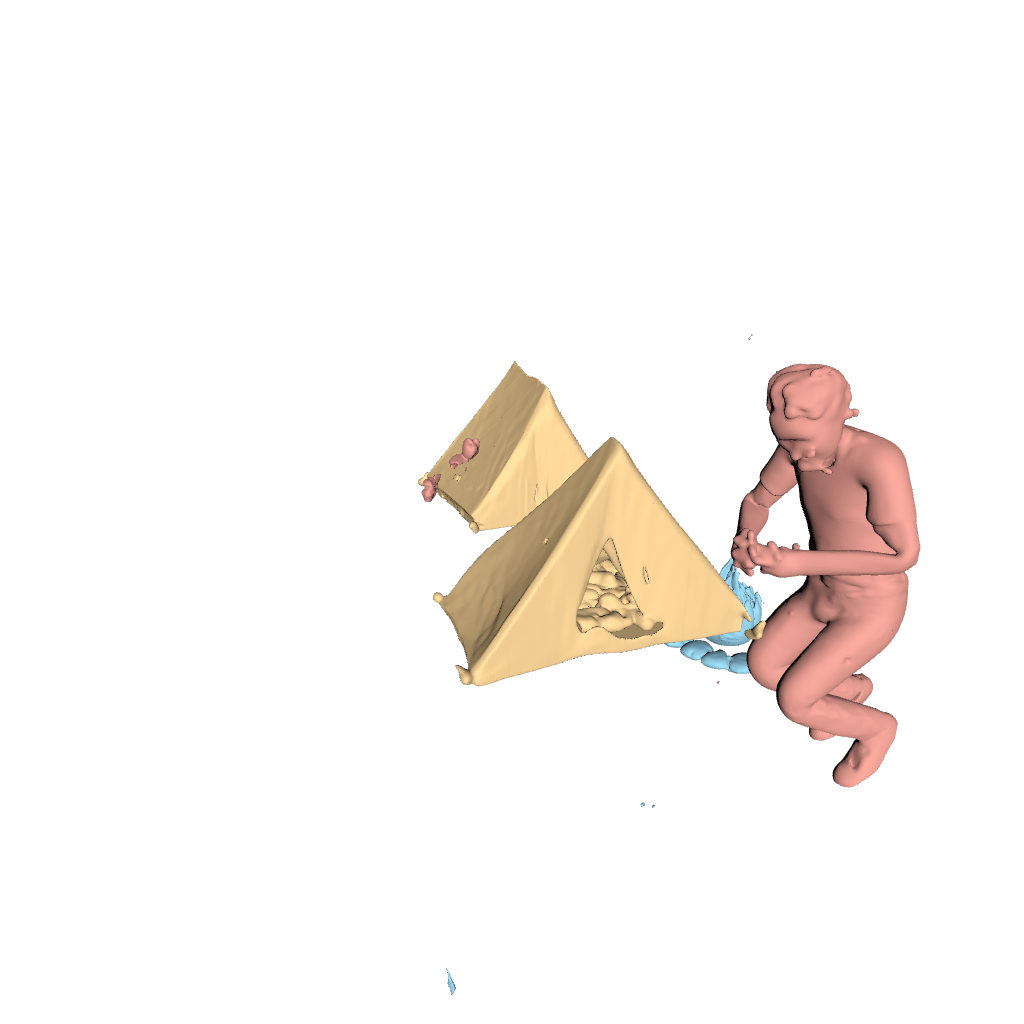} &
    \includegraphics[width=0.3\linewidth]{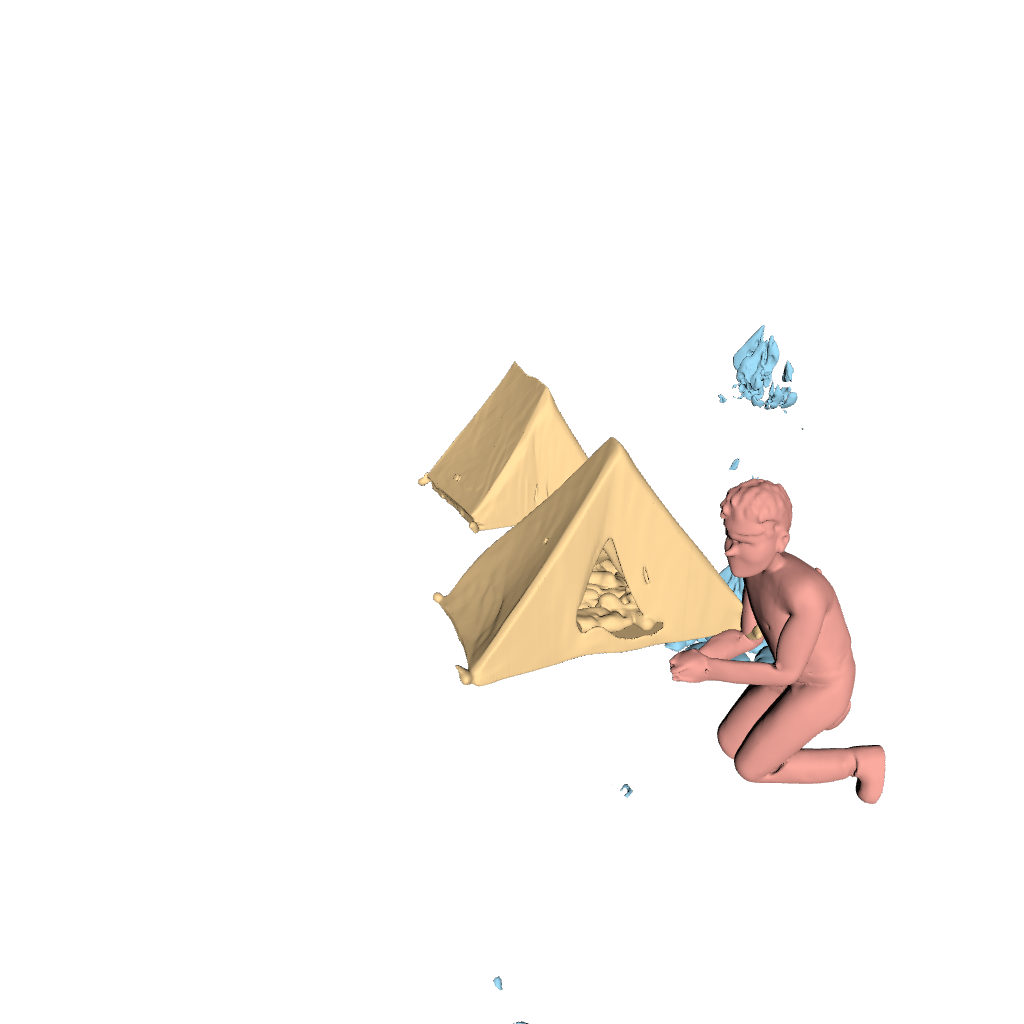} \\[8pt] % Space between samples

    % % ==================== SECOND SAMPLE ====================
    % % --- Row 1: Input ---
    % \rotatebox{90}{\small ~~~~~~Input} &
    % \includegraphics[width=0.3\linewidth]{figures/appendix/ablation/other/cat/input/0.jpg} &
    % \includegraphics[width=0.3\linewidth]{figures/appendix/ablation/other/cat/input/1.jpg} &
    % \includegraphics[width=0.3\linewidth]{figures/appendix/ablation/other/cat/input/2.jpg} \\[-0.5pt]

    % % --- Row 2: w/ Mixing ---
    % \rotatebox{90}{\small ~~w/ Mixing} &
    % \includegraphics[width=0.3\linewidth]{figures/appendix/ablation/other/cat/mixing/0.jpg} &
    % \includegraphics[width=0.3\linewidth]{figures/appendix/ablation/other/cat/mixing/1.jpg} &
    % \includegraphics[width=0.3\linewidth]{figures/appendix/ablation/other/cat/mixing/2.jpg} \\[-0.5pt]

    % % --- Row 3: w/o Mixing ---
    % \rotatebox{90}{\small ~~w/o Mixing} &
    % \includegraphics[width=0.3\linewidth]{figures/appendix/ablation/other/cat/no_mixing/0.jpg} &
    % \includegraphics[width=0.3\linewidth]{figures/appendix/ablation/other/cat/no_mixing/1.jpg} &
    % \includegraphics[width=0.3\linewidth]{figures/appendix/ablation/other/cat/no_mixing/2.jpg} \\[8pt] % Space between samples

    % ==================== THIRD SAMPLE ====================
    % --- Row 1: Input ---
    \rotatebox{90}{\small ~~~~~~~~~Input} &
    \includegraphics[width=0.3\linewidth]{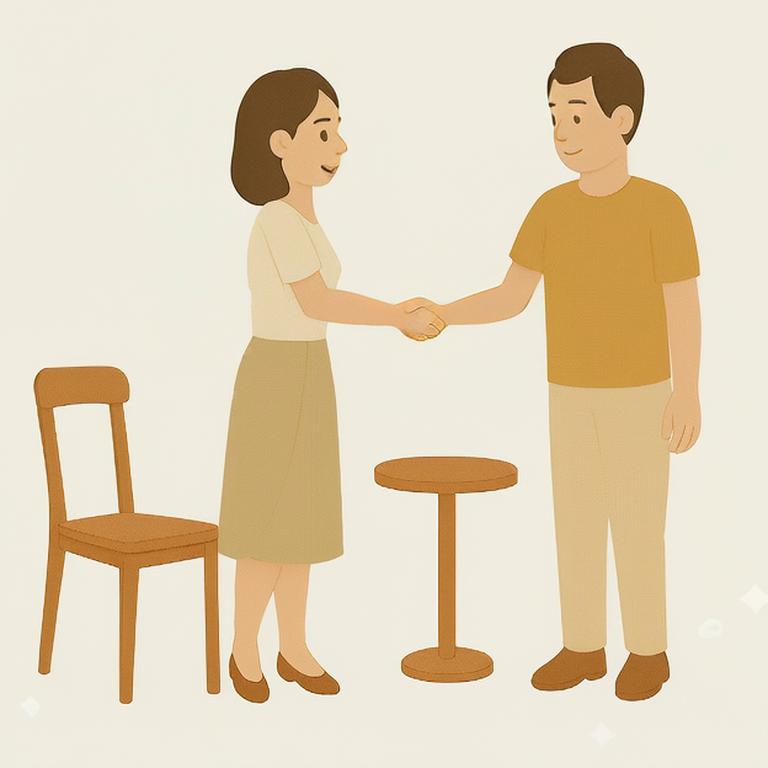} &
    \includegraphics[width=0.3\linewidth]{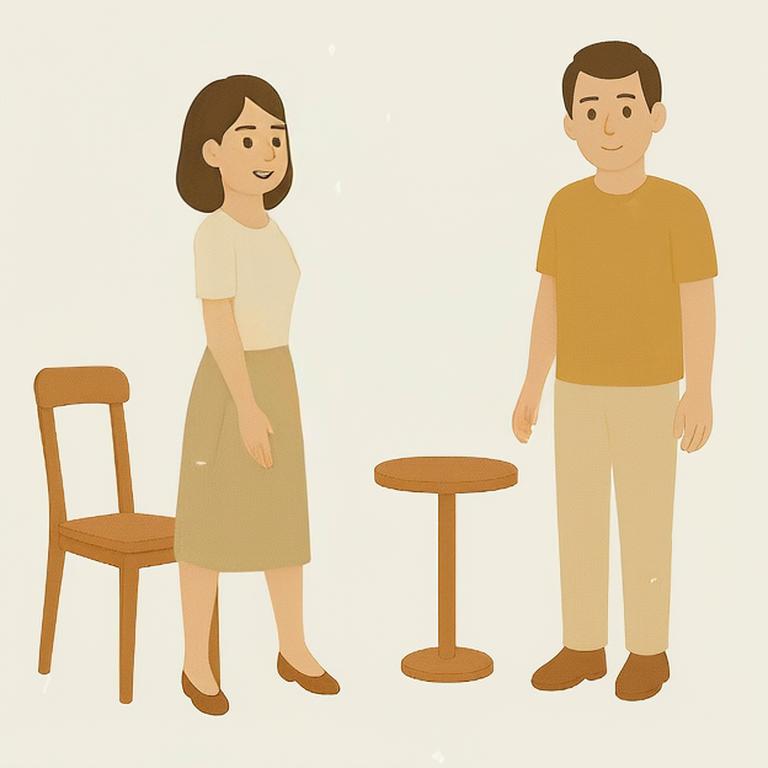} &
    \includegraphics[width=0.3\linewidth]{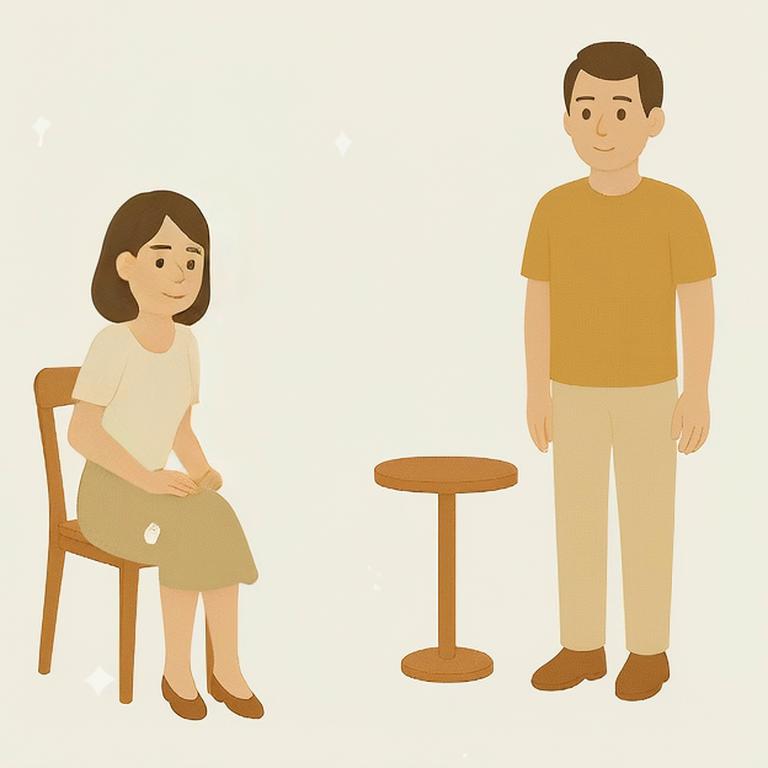} \\[-0.5pt]

    % --- Row 2: w/ Mixing ---
    \rotatebox{90}{\small ~~w/ Mixing} &
    \includegraphics[width=0.3\linewidth]{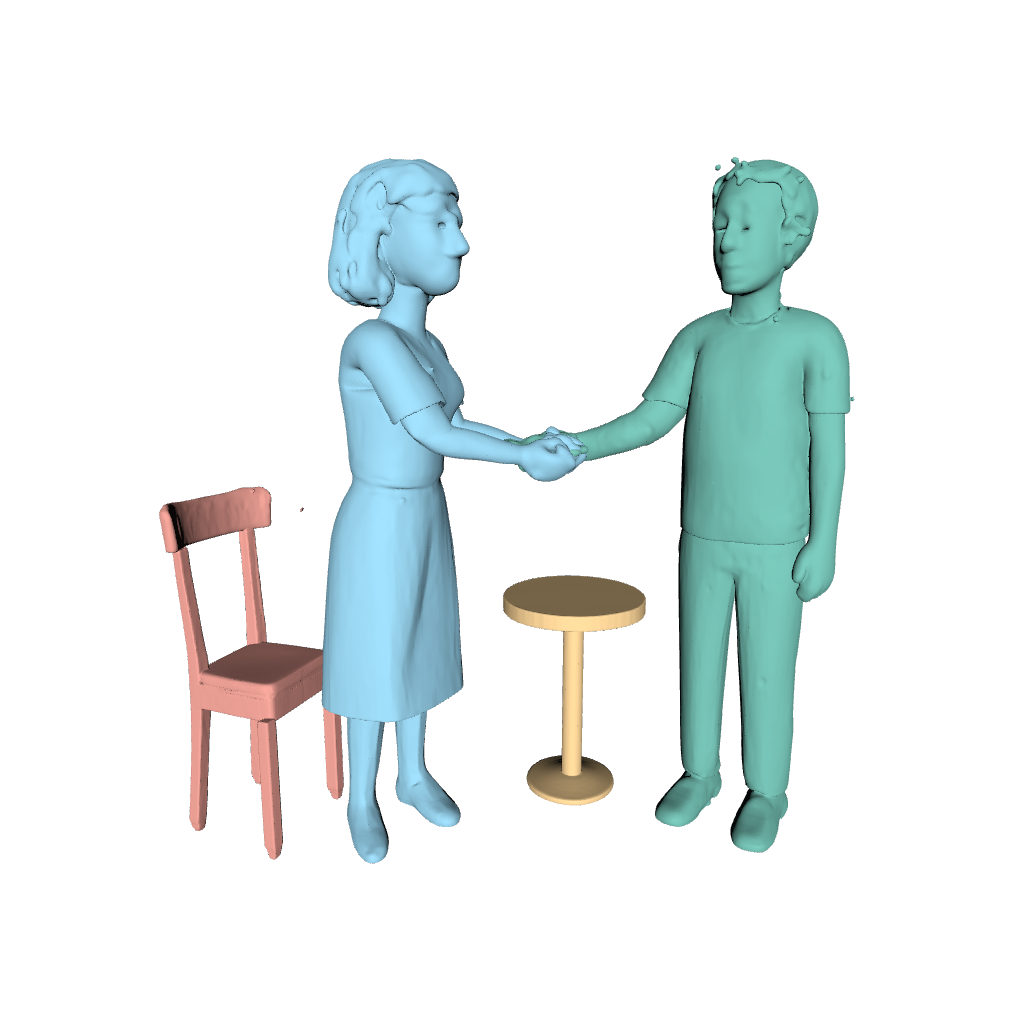} &
    \includegraphics[width=0.3\linewidth]{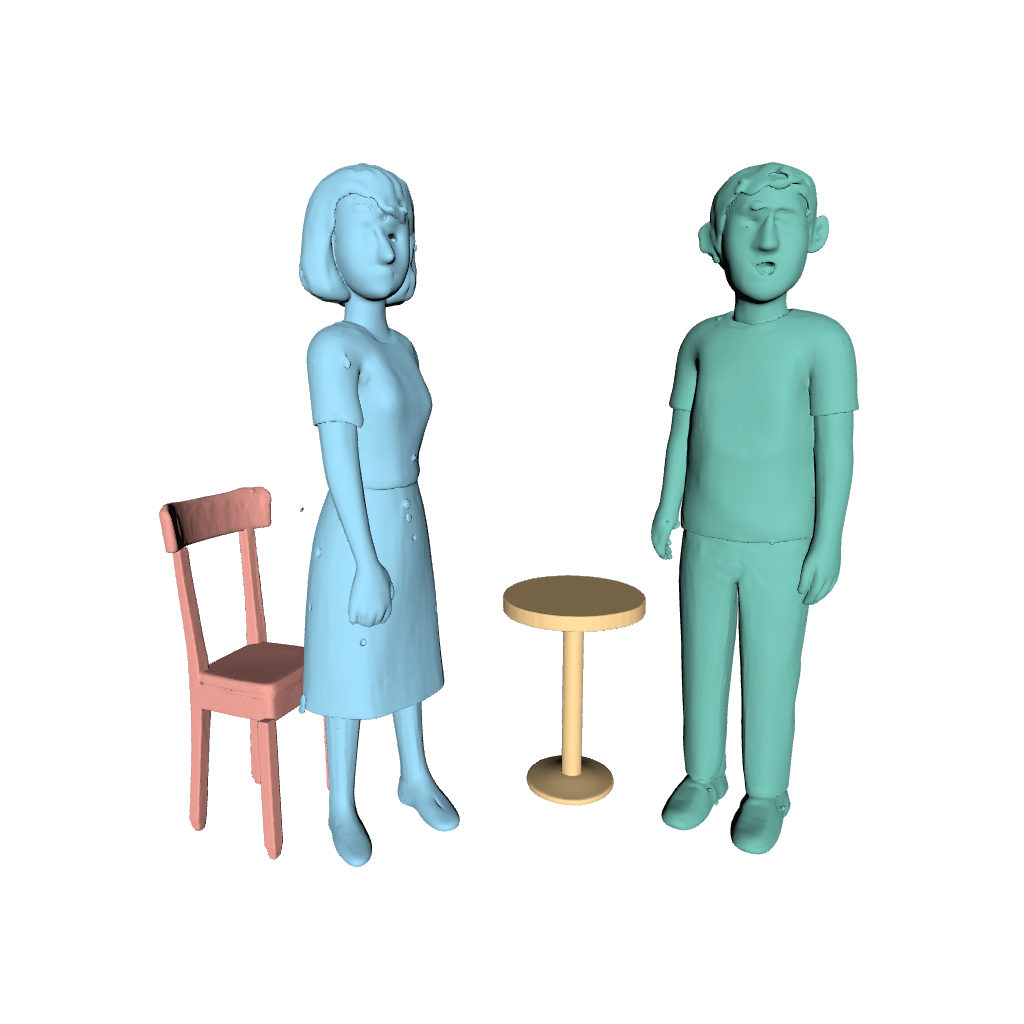} &
    \includegraphics[width=0.3\linewidth]{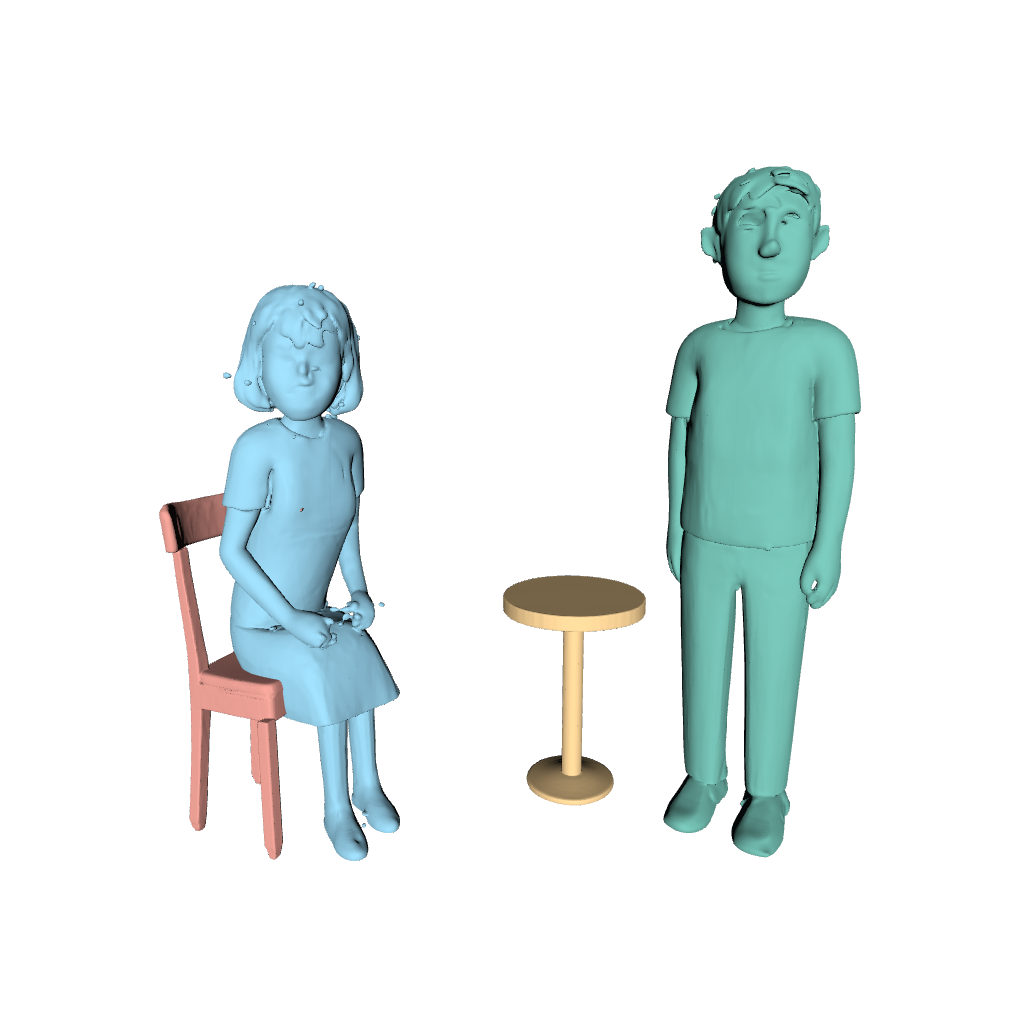} \\[-0.5pt]

    % --- Row 3: w/o Mixing ---
    \rotatebox{90}{\small ~~w/o Mixing} &
    \includegraphics[width=0.3\linewidth]{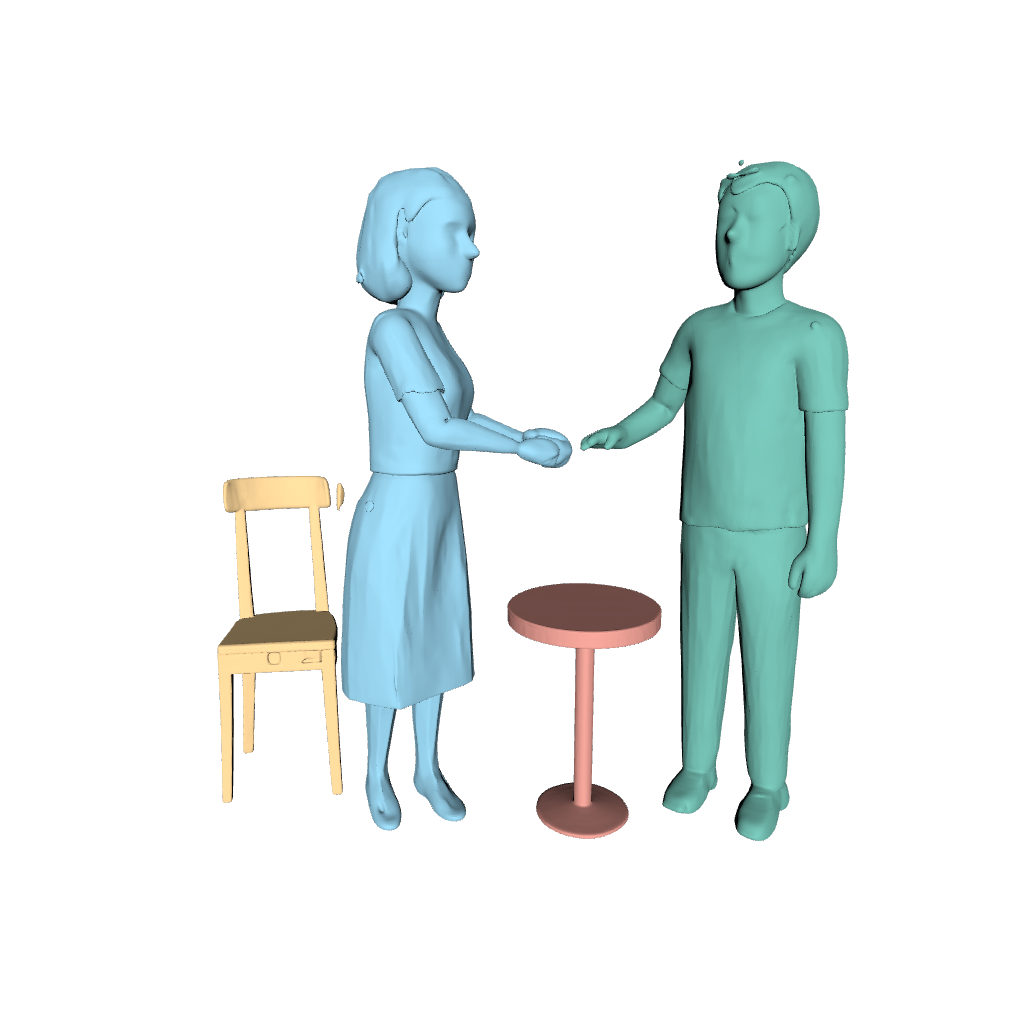} &
    \includegraphics[width=0.3\linewidth]{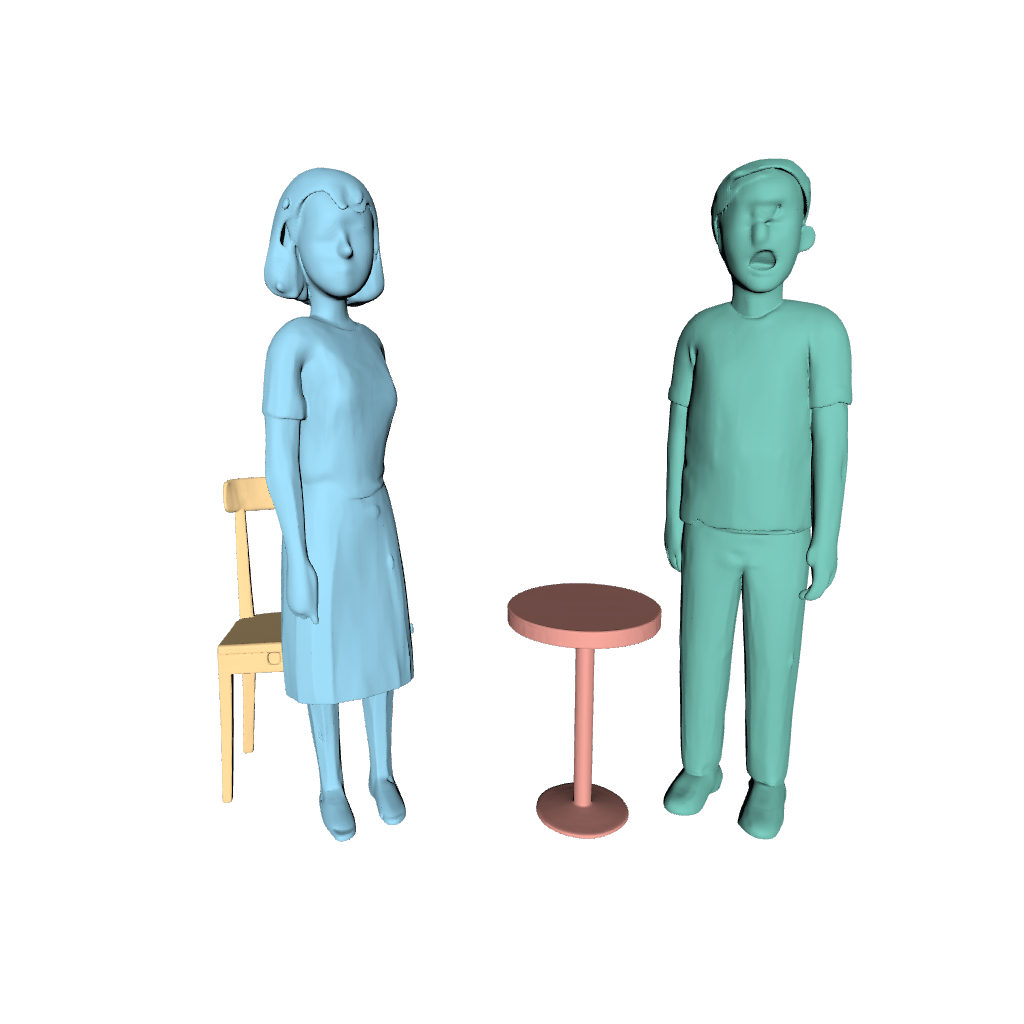} &
    \includegraphics[width=0.3\linewidth]{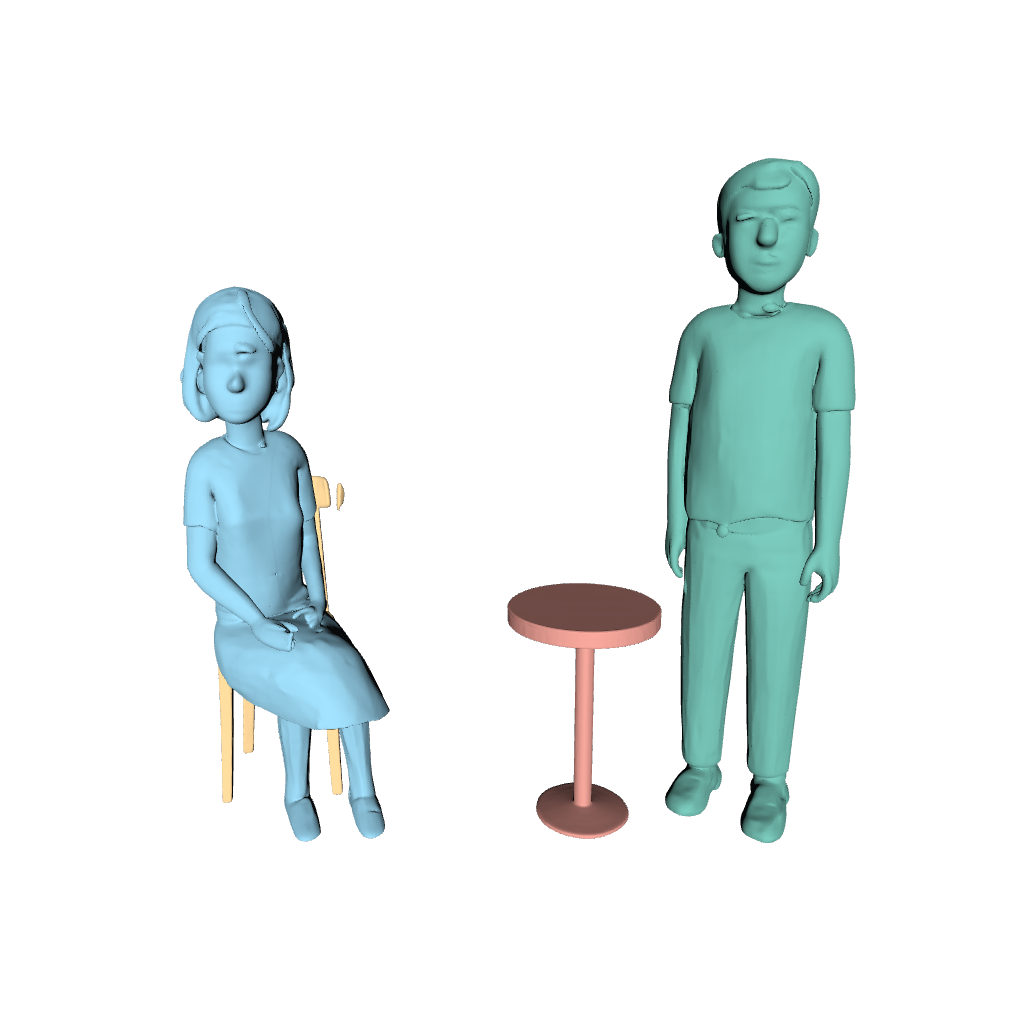} \\

  \end{tabular}
  % ==================== TABLE END ====================
  
  \vspace{-3pt}
  \caption{
    Ablation study on mixing components for various sequences. For each sample, we show input frames, results with our mixing strategy, and results without. In particular, the chair pose and the interaction between the dynamics (the lady shaking hands with the man) and the dynamic and static (lady and the chair) are captured incorrectly without mixing.
  }
  \label{fig:com4d_appendix_1}
\end{figure}

\begin{figure}[h]
    \section{Additional Qualitative Results with Moving Camera}
    COM4D assumes a fixed camera when reconstructing compositional scenes containing both static and dynamic objects, as camera motion is inherently entangled with static scene geometry. In such settings, disentangling camera motion from scene structure is fundamentally ambiguous without additional constraints or supervision.
    
    That said, COM4D is not inherently restricted to static-camera scenarios. In cases where the scene consists only of dynamic objects, the model can still recover consistent relative 4D structure and motion over time. This is because COM4D relies on learning relative spatial relationships and temporal coherence, rather than absolute camera-referenced geometry. 

    \vspace{0.5cm}
    
    \centering
    \setlength{\tabcolsep}{0pt}
    \renewcommand{\arraystretch}{0.0}
    \begin{tabular}{c c c c c}
        % Row 1: Inputs
        \rotatebox{90}{\footnotesize ~~~~~~~Input} &
        \includegraphics[width=0.22\columnwidth]
        {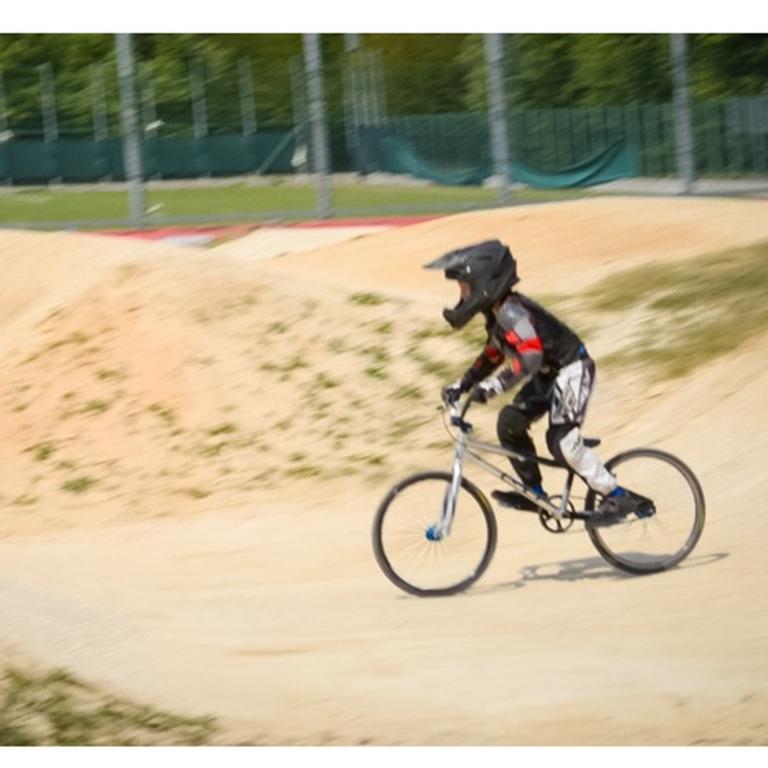} &
        \includegraphics[width=0.22\columnwidth]
        {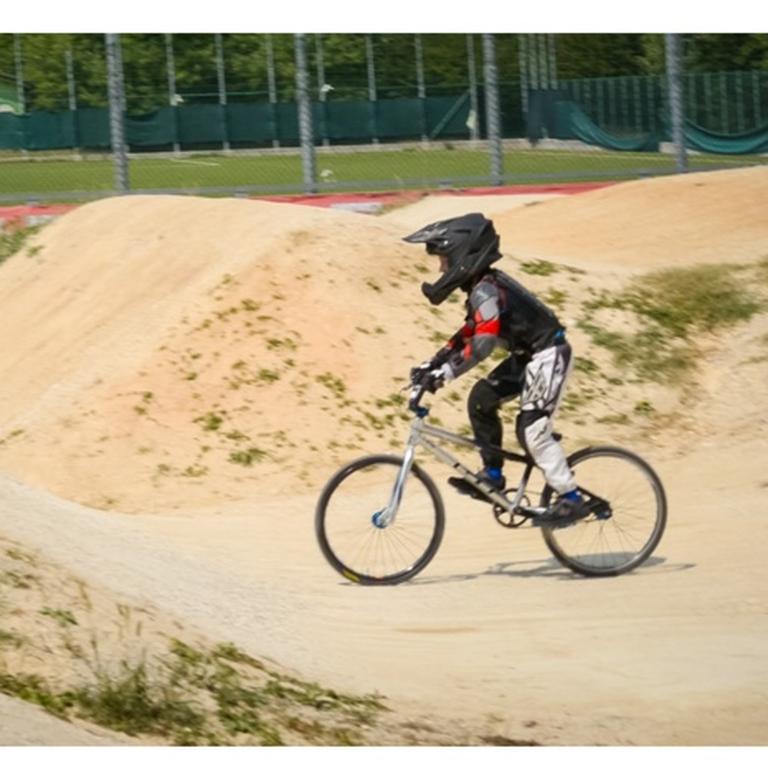} &
        \includegraphics[width=0.22\columnwidth]
        {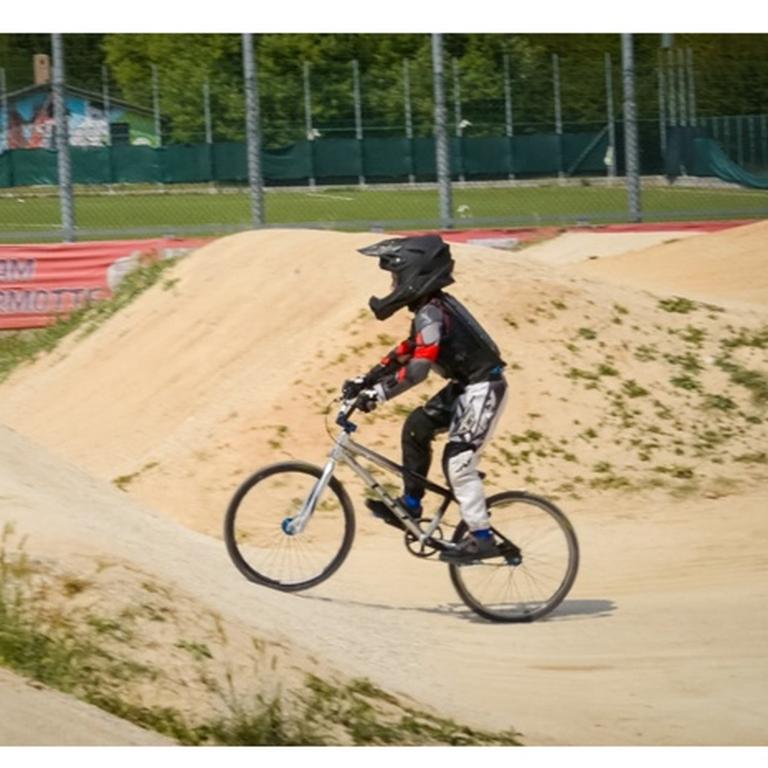} &
        \includegraphics[width=0.22\columnwidth]
        {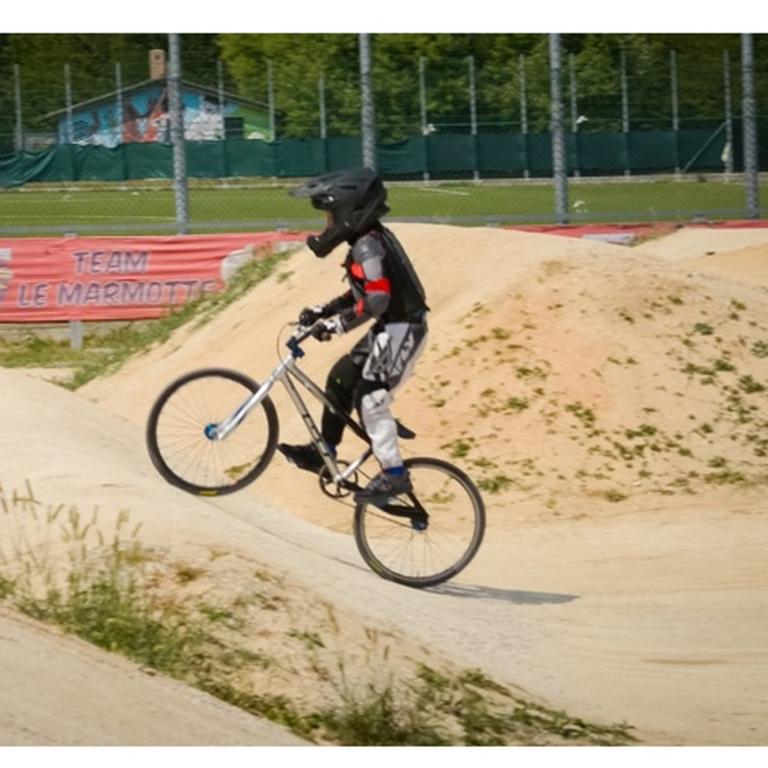} \\
    
        % Row 2: Reconstructions
        \rotatebox{90}{\footnotesize ~~~~~Reconstr.} &
        \includegraphics[width=0.22\columnwidth, trim=75 75 75 75, clip]
        {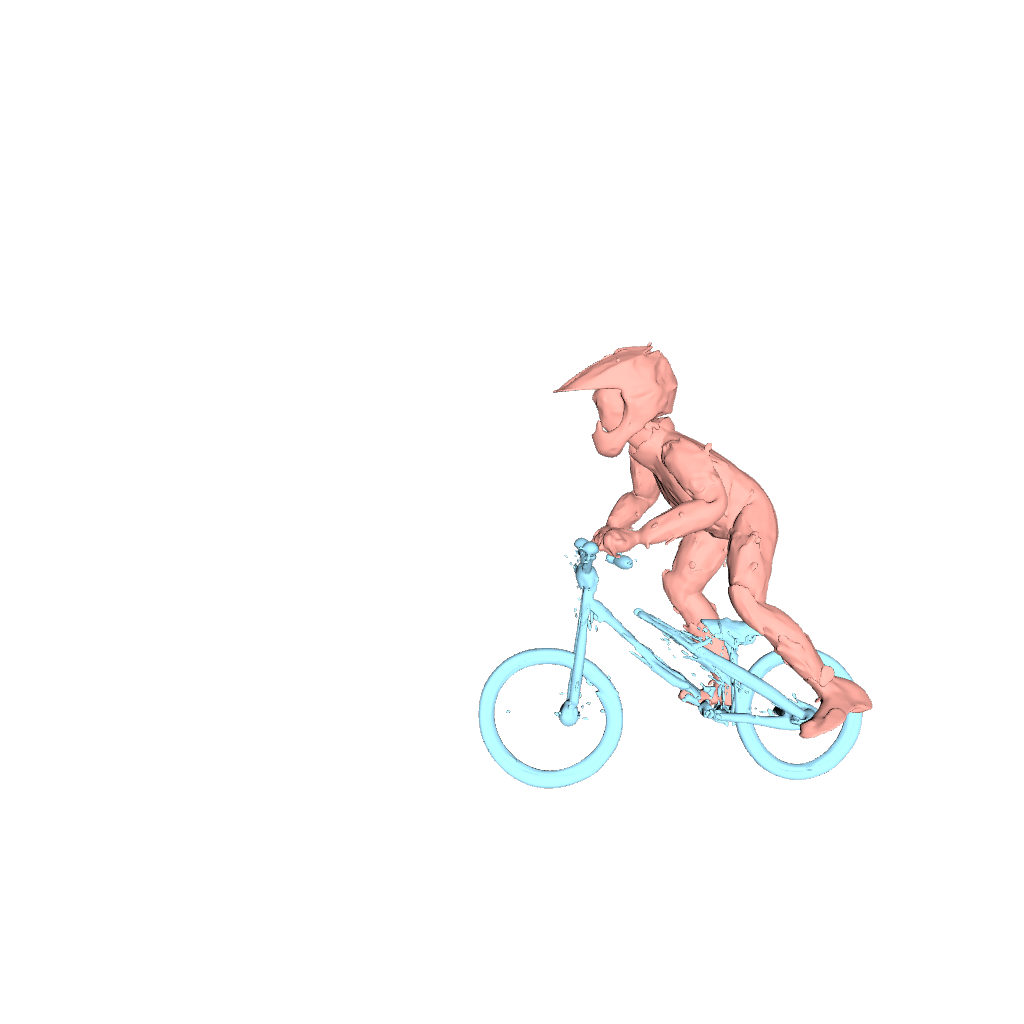} &
        \includegraphics[width=0.22\columnwidth, trim=75 75 75 75, clip]
        {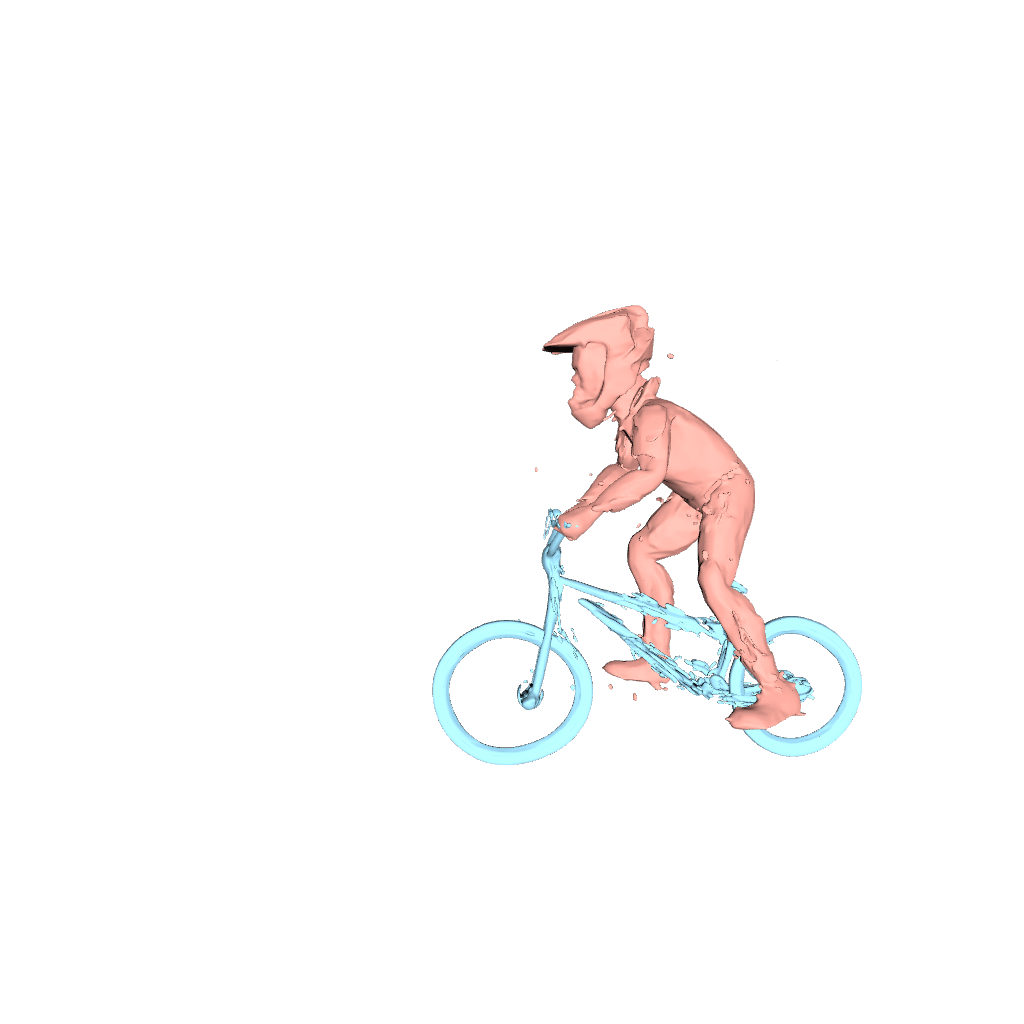} &
        \includegraphics[width=0.22\columnwidth, trim=75 75 75 75, clip]
        {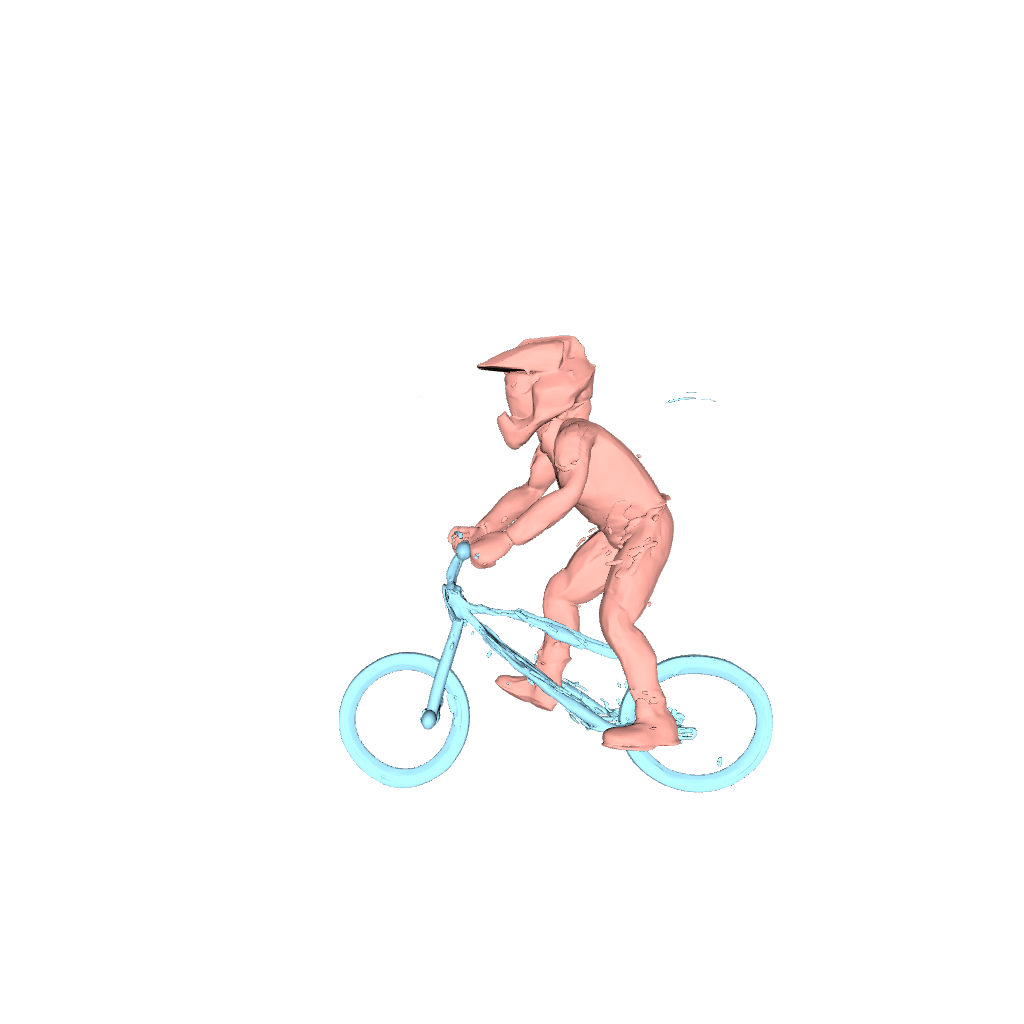} &
        \includegraphics[width=0.22\columnwidth, trim=75 75 75 75, clip]
        {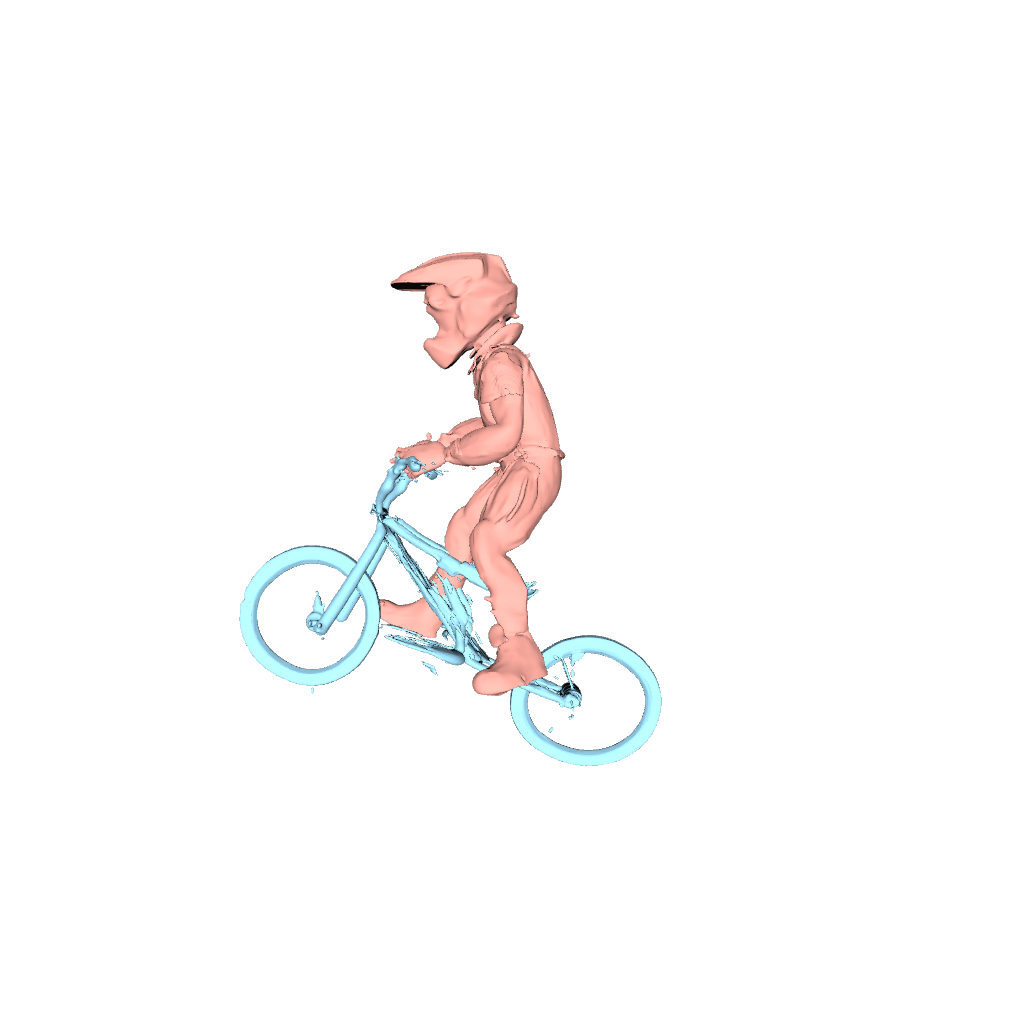} \\
    
        % Row 3: Novel Views
        \rotatebox{90}{\footnotesize ~~~Novel View} &
        \includegraphics[width=0.22\columnwidth, trim=125 125 125 125, clip]
        {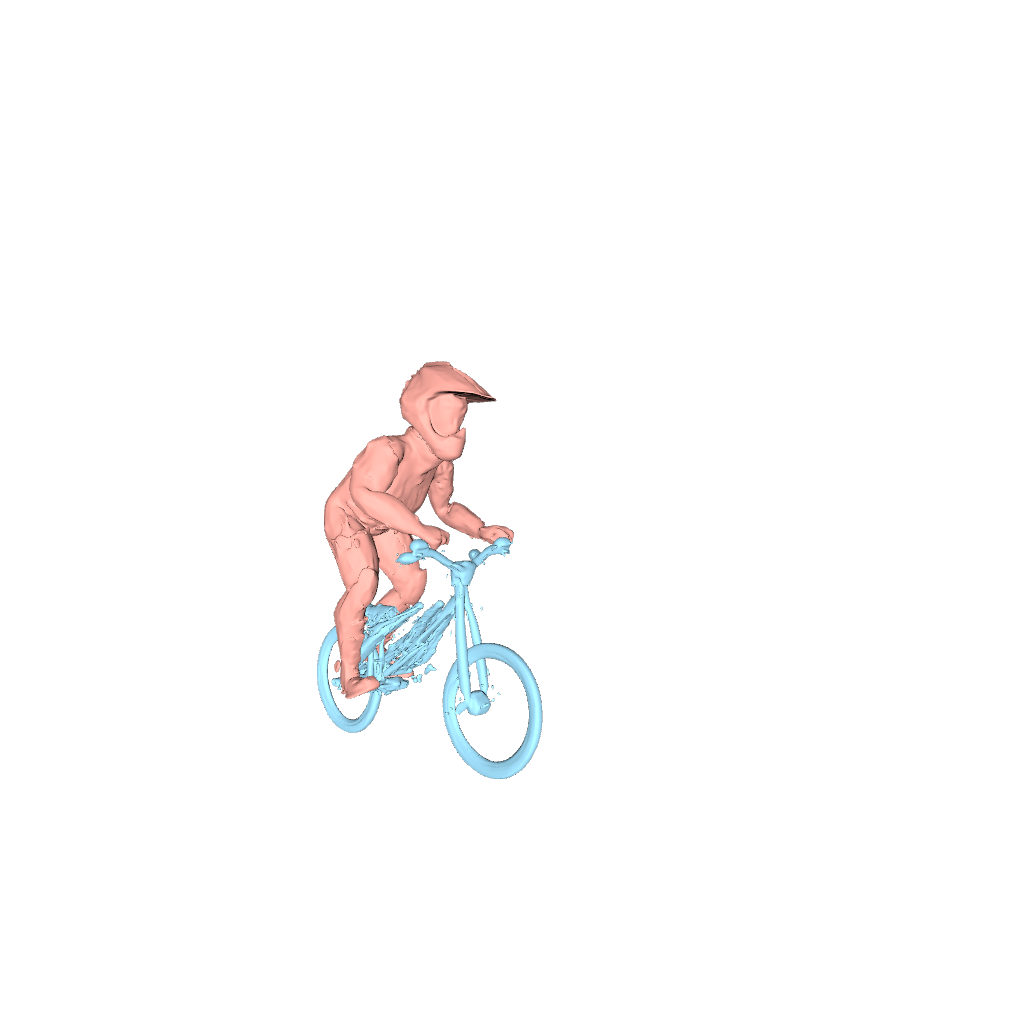} &
        \includegraphics[width=0.22\columnwidth, trim=125 125 125 125, clip]
        {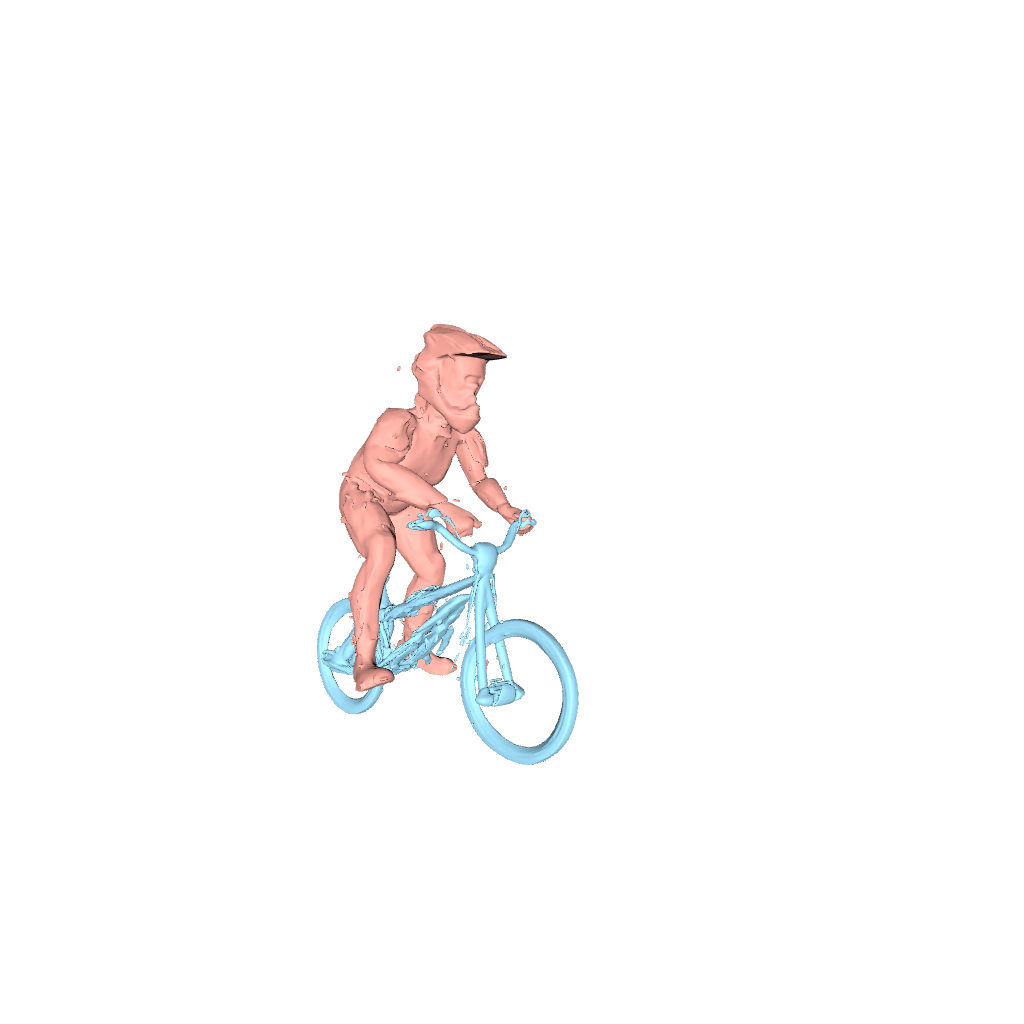} &
        \includegraphics[width=0.22\columnwidth, trim=125 125 125 125, clip]
        {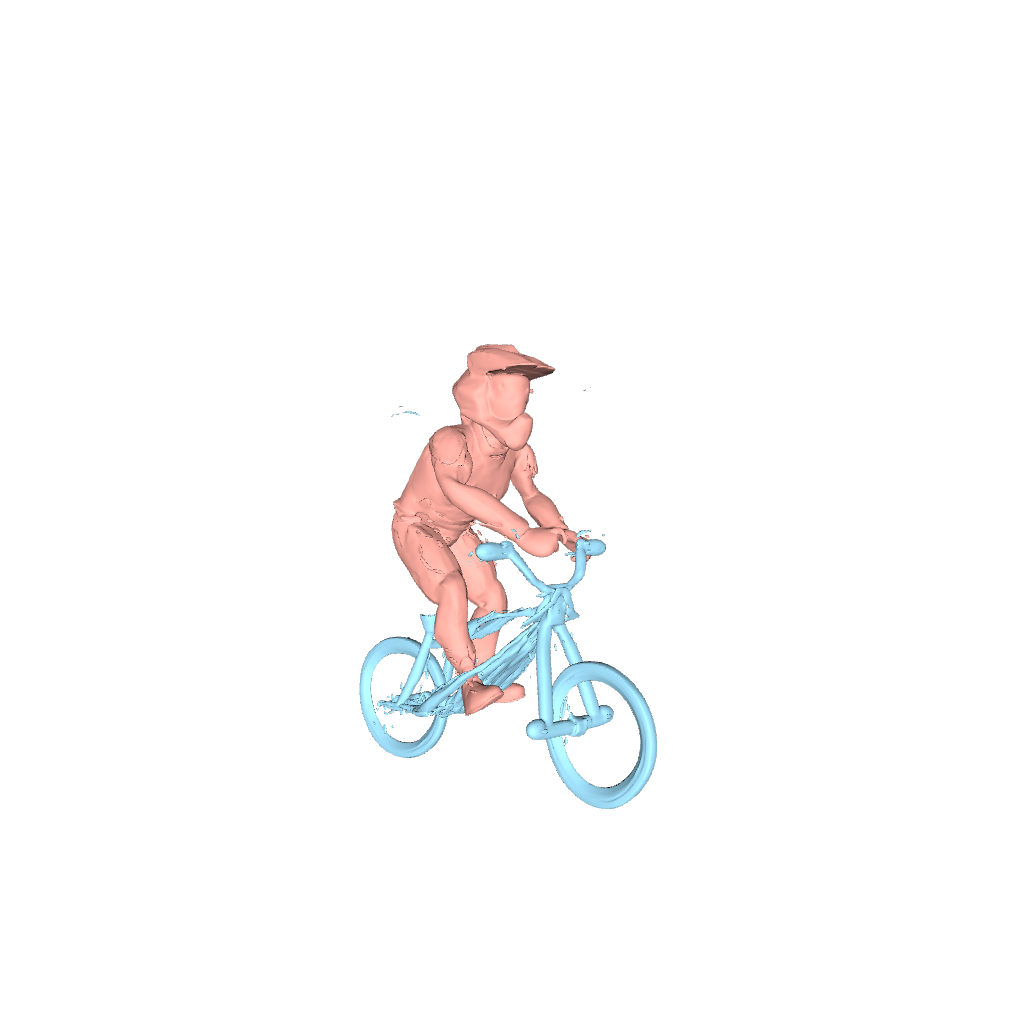} &
        \includegraphics[width=0.22\columnwidth, trim=125 125 125 125, clip]
        {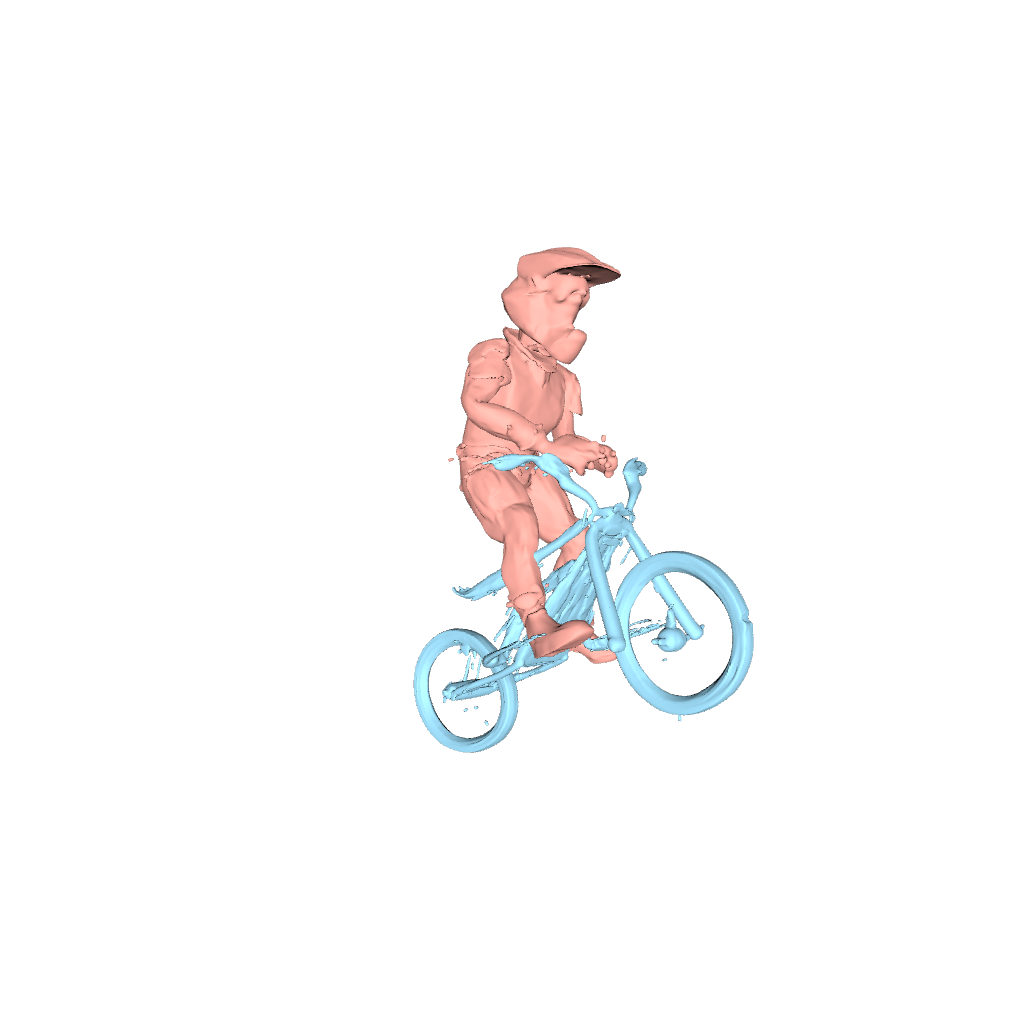} \\
    \end{tabular}
    \caption{Qualitative results for a sample with dynamic camera.}
    \label{fig:dynamic_camera_appendix}
\end{figure}

\begin{figure}[h]
    \section{Physical Plausibility and Object Interactions}
    COM4D is a fully data-driven framework and does not explicitly enforce physical constraints such as contact or collision avoidance. Instead, interactions between multiple objects are modeled implicitly through attention mechanisms in the latent space.

    To assess whether this implicit modeling yields physically plausible reconstructions, we evaluate mesh interpenetration across 8 video sequences by measuring the intersection-over-union (IoU) between reconstructed object meshes. We observe a low average overlap of $\text{IoU} = 0.0096$, indicating minimal interpenetration between interacting objects.
    
    For reference, we report an IoU of $0.0018$ on 3D-FRONT, where ground-truth scenes exhibit little to no physical contact between objects (see \cref{tab:quant_comp_3d}). The small gap between these values suggests that COM4D maintains a comparable level of physical plausibility despite not explicitly modeling physical constraints.

    \centering
    \setlength{\tabcolsep}{0pt}   % no horizontal padding
    \renewcommand{\arraystretch}{0}
    \begin{tabular}{c c c c c}
        % Row 1: input frames
        \includegraphics[width=0.20\columnwidth, trim=200 40 200 0, clip]{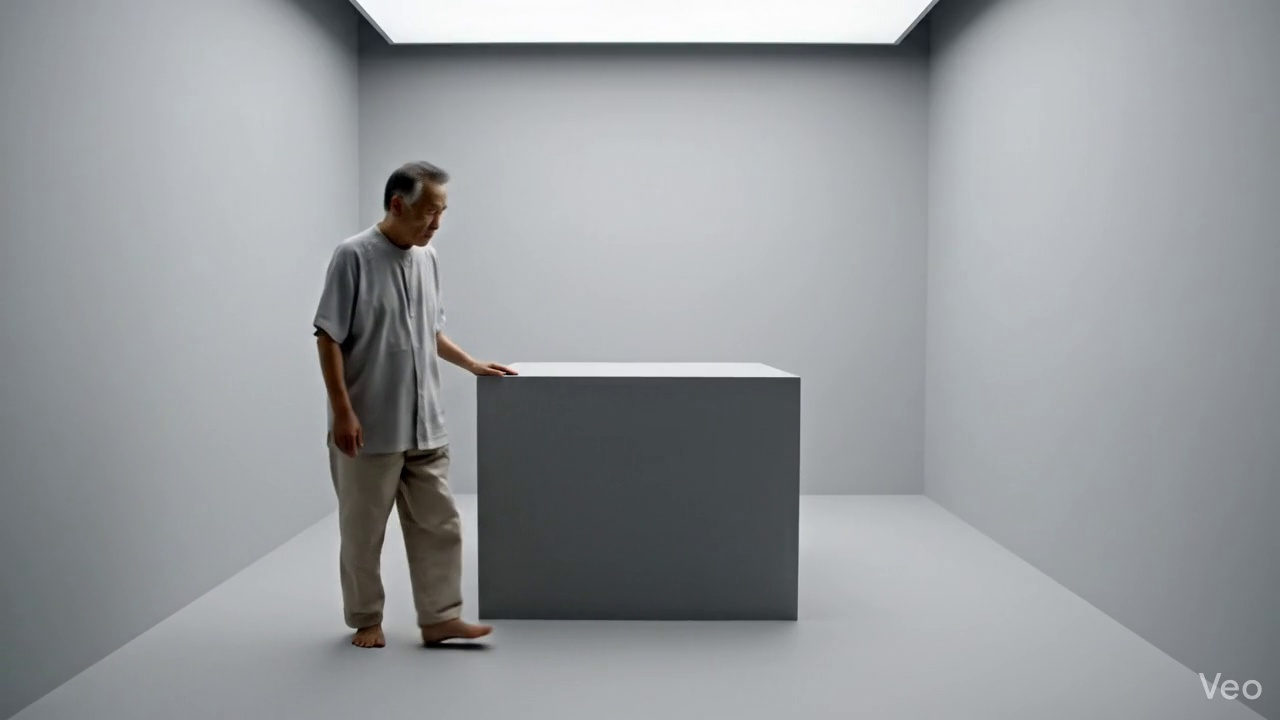} &
        \includegraphics[width=0.20\columnwidth, trim=50 100 50 0, clip]{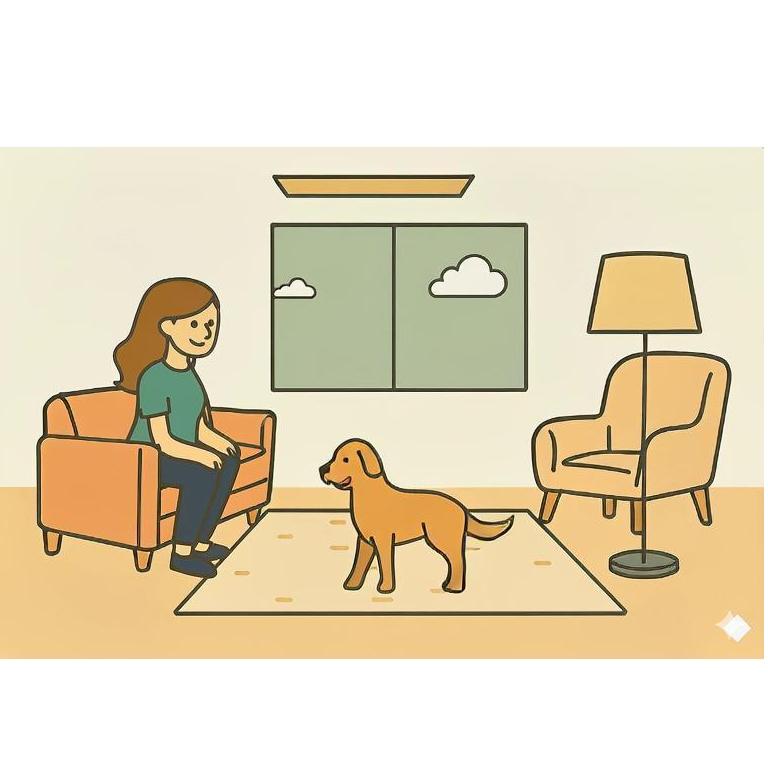} &
        \includegraphics[width=0.20\columnwidth, trim=500 200 300 0, clip]{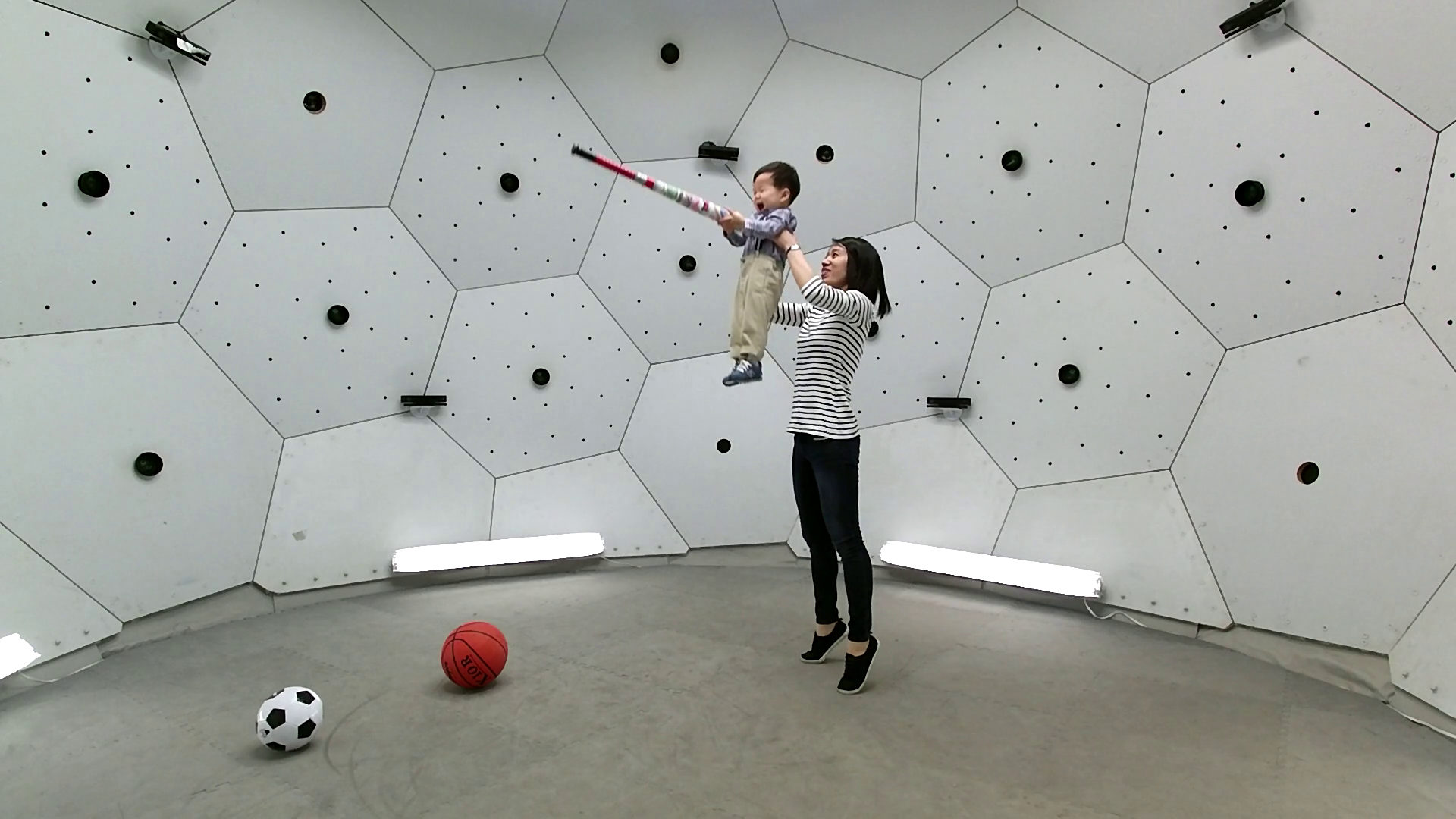} &
        \includegraphics[width=0.20\columnwidth, trim=0 100 500 0, clip]{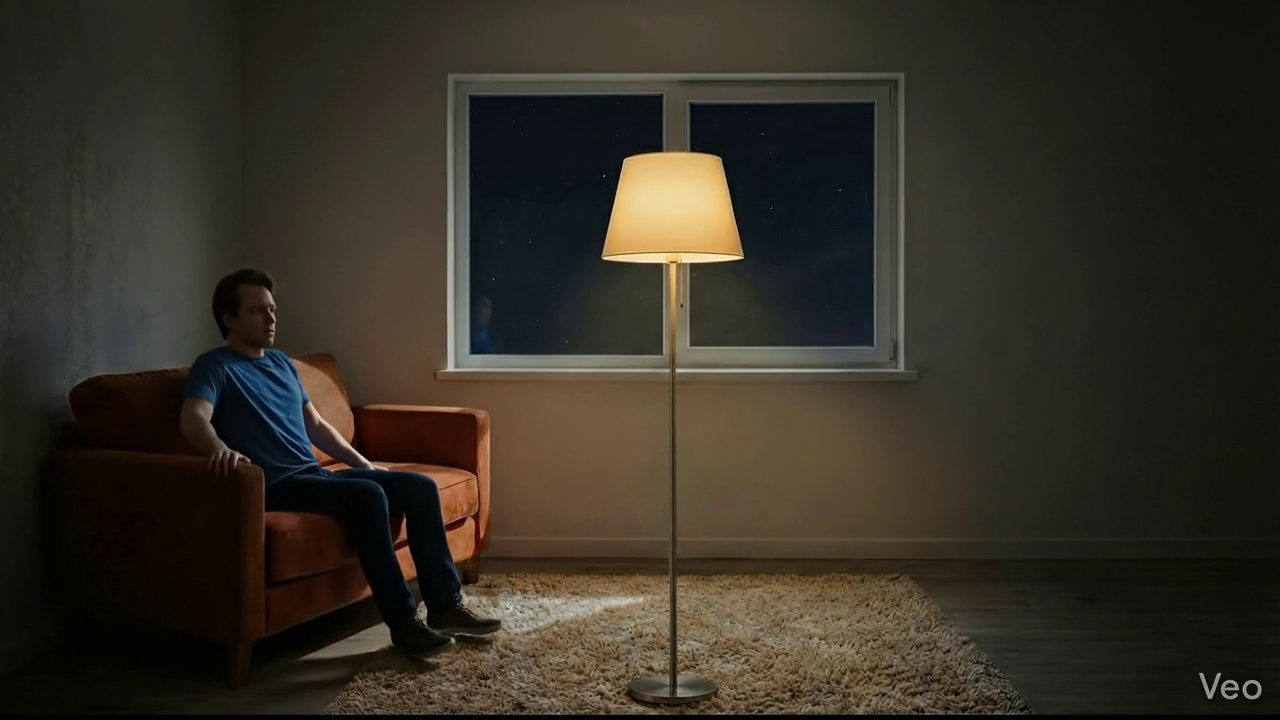} &
        \includegraphics[width=0.20\columnwidth, trim=100 0 450 0, clip]{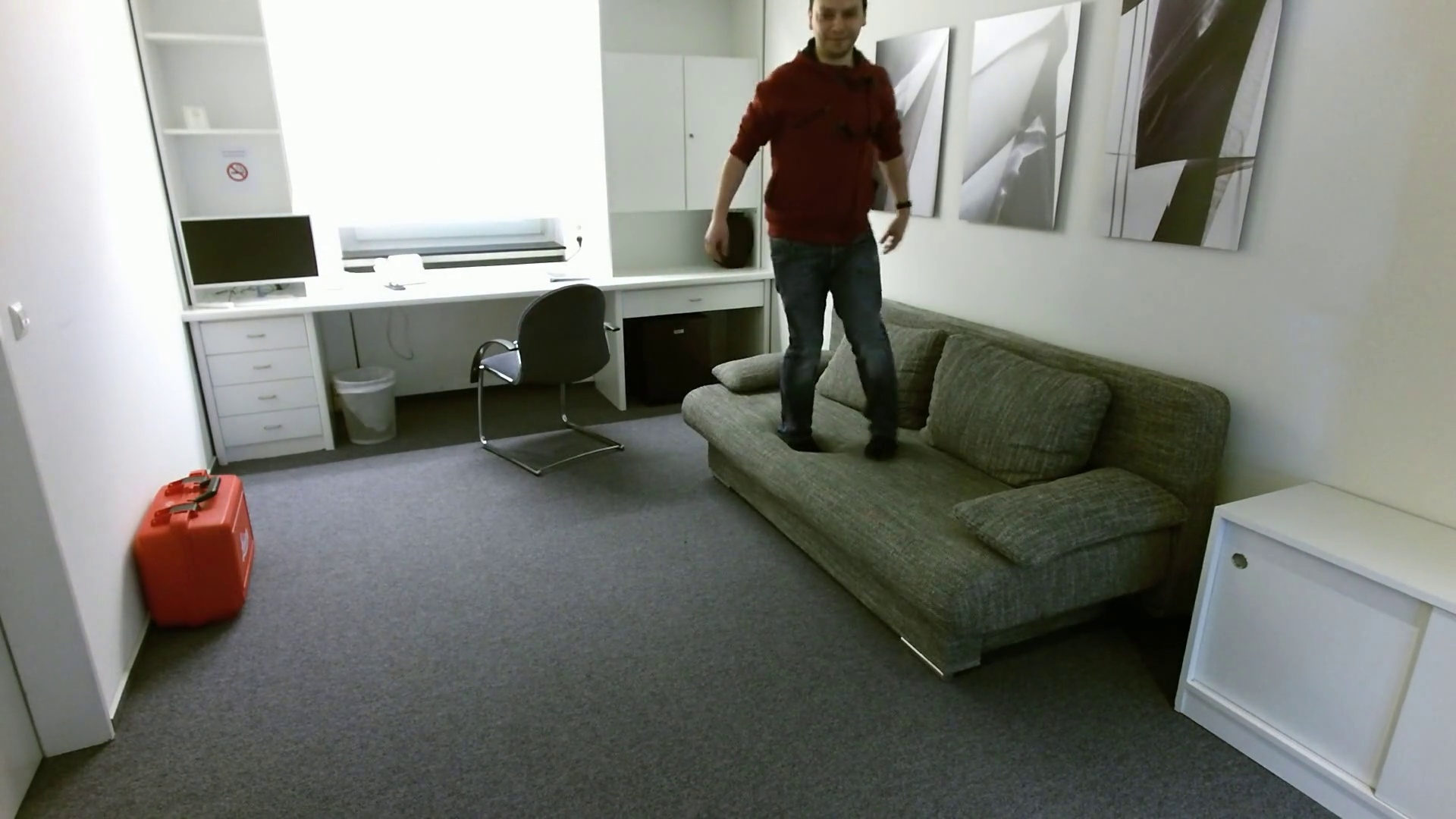} \\

        % Row 2: renderings
        \includegraphics[width=0.20\columnwidth, trim=100 0 100 0, clip]{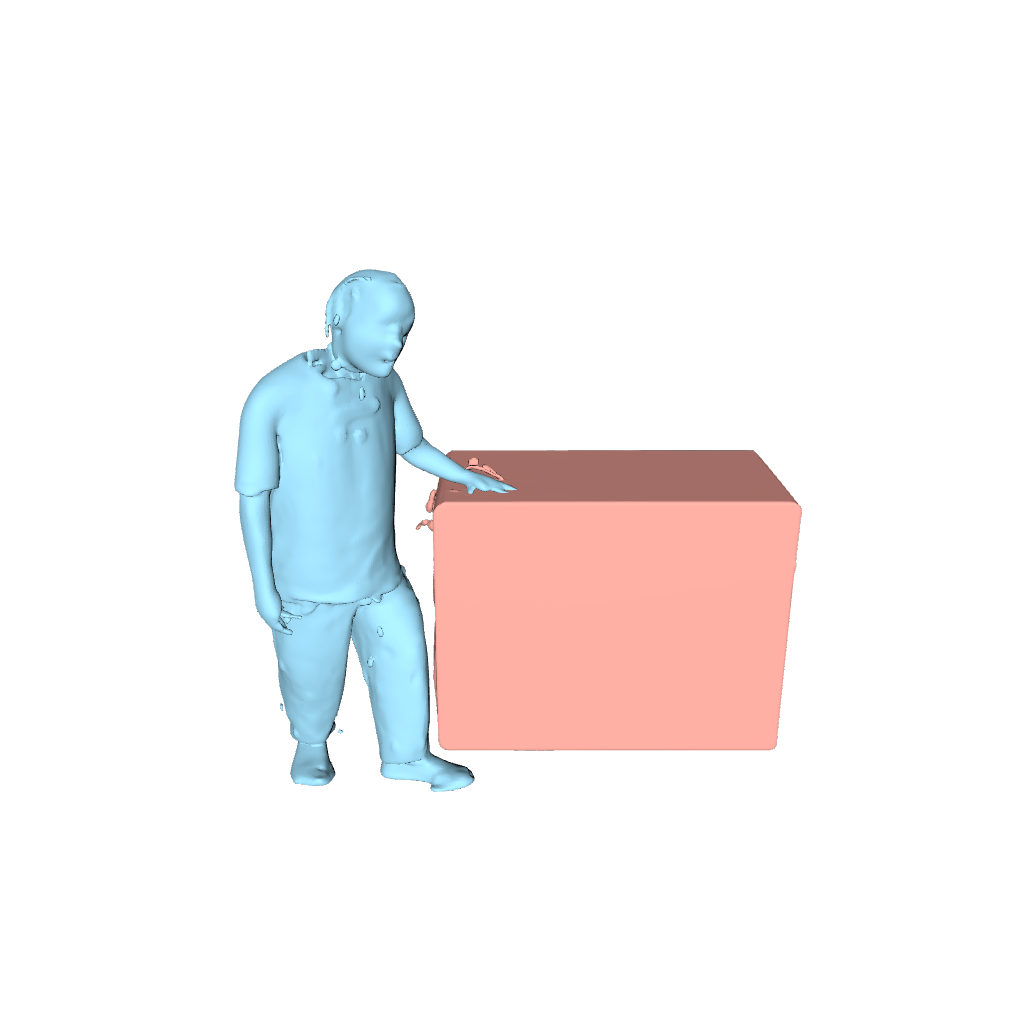} &
        \includegraphics[width=0.20\columnwidth, trim=100 0 100 0, clip]{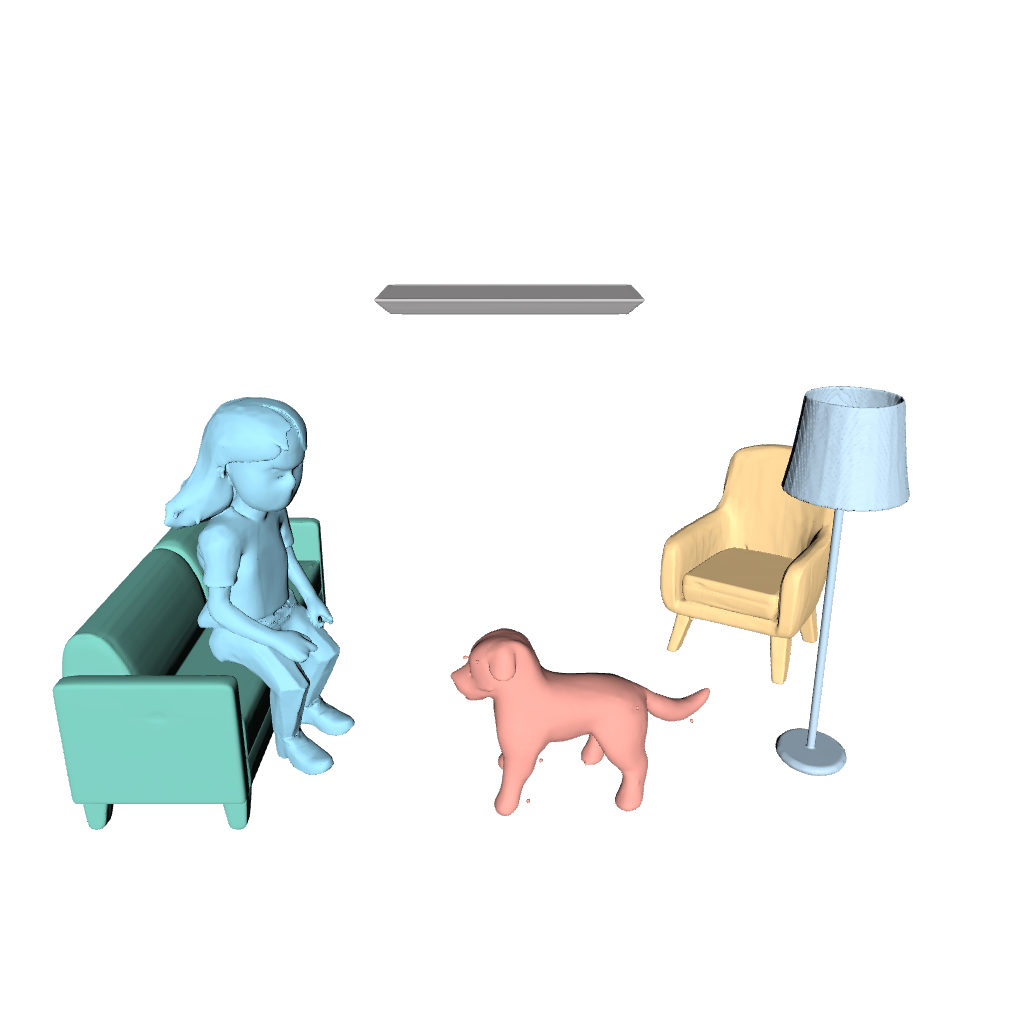} &
        \includegraphics[width=0.20\columnwidth, trim=200 0 0 0, clip]{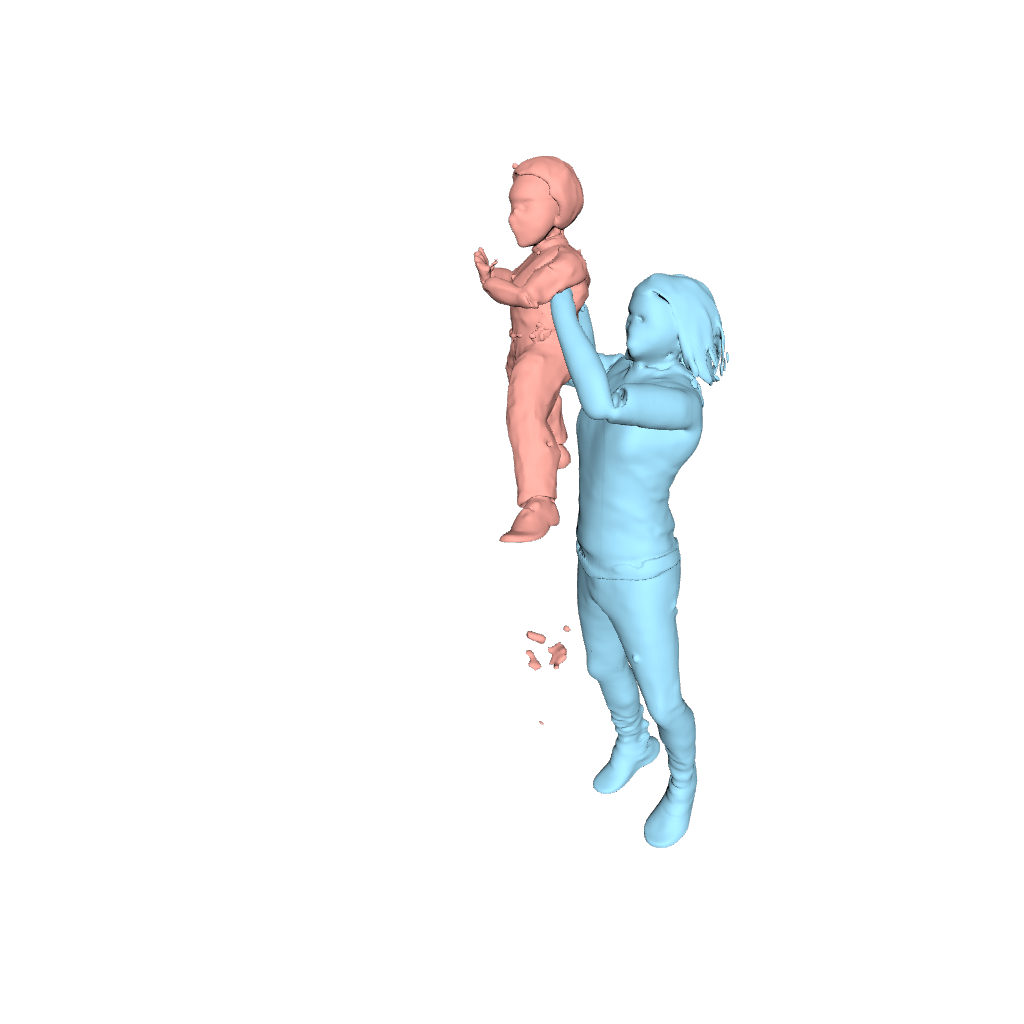} &
        \includegraphics[width=0.20\columnwidth, trim=100 0 100 0, clip]{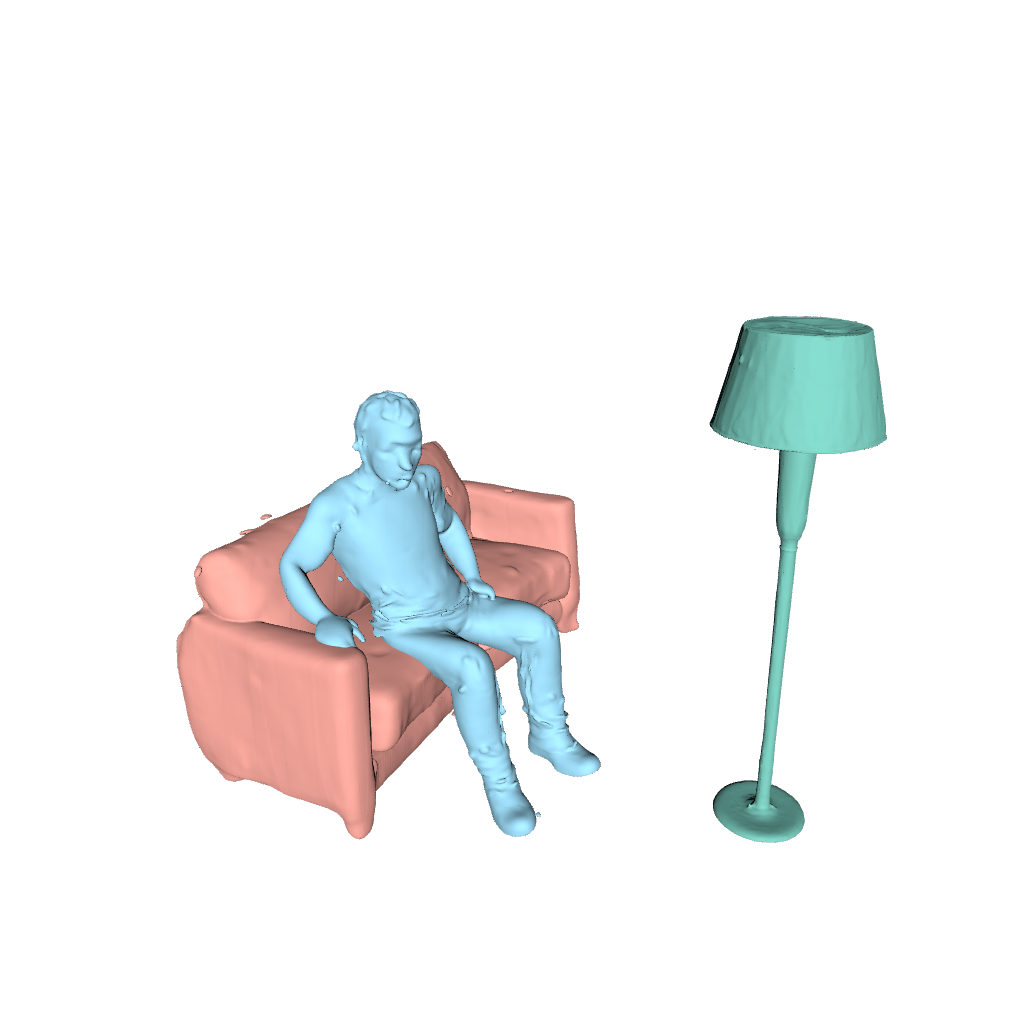} &
        \includegraphics[width=0.20\columnwidth, trim=100 0 100 200, clip]{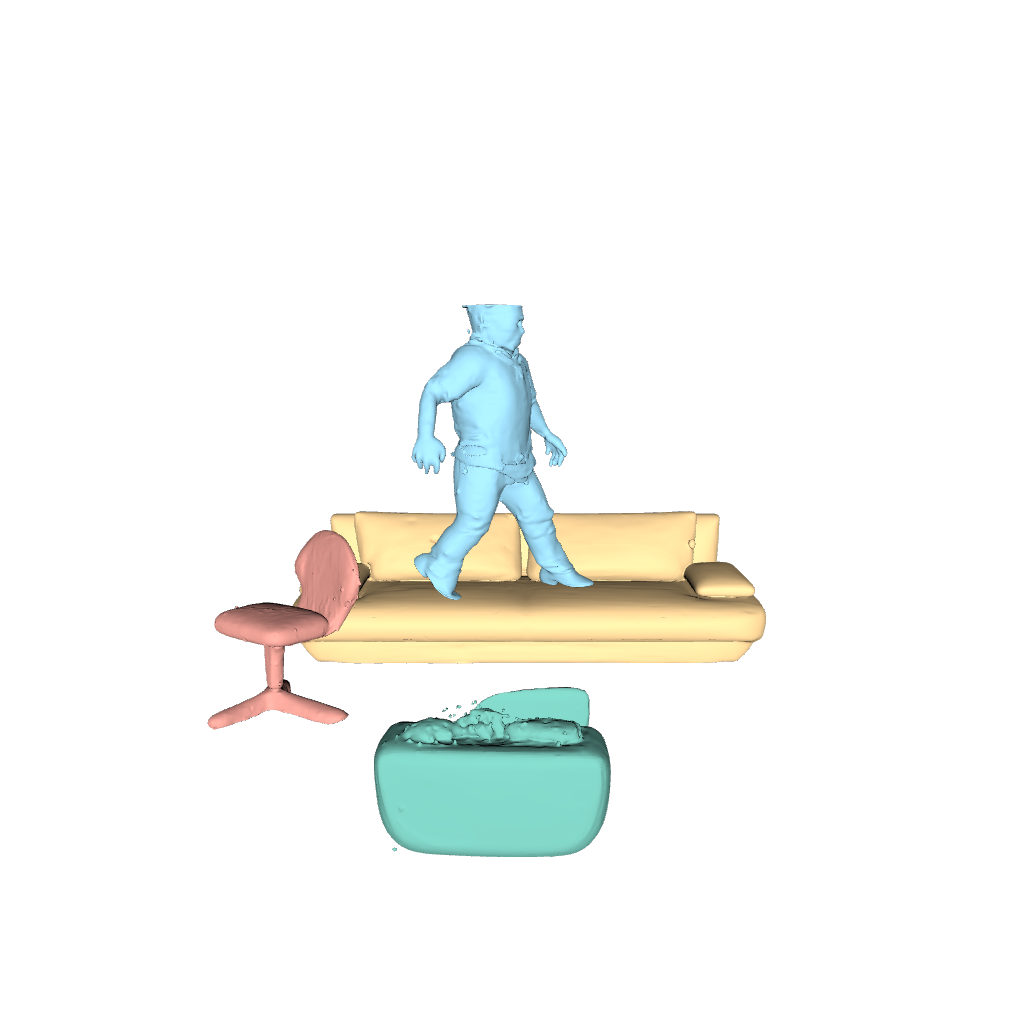}
    \end{tabular}
    \caption{Qualitative results on interaction scenarios. Top row shows input frames, and bottom shows 4D reconstructions.}
    \label{fig:interaction_results_appendix}
\end{figure}

\begin{figure}[h]
    \section{Comparison with Optimization-Based Baselines}
    We compare COM4D against the recent optimization-based method \emph{Shape of Motion (SOM)}~\cite{som2024}. SOM struggles to recover complete geometry under occlusions, often producing incomplete or noisy reconstructions (see \cref{fig:comparison_appendix}). 
    
    Quantitatively, SOM achieves a Chamfer Distance (CD) of \textbf{11.30\,cm}, whereas COM4D achieves \textbf{7.42\,cm} on~\cite{panoptic,Joo_2018_CVPR}. This gap highlights the advantage of our generative formulation, which leverages learned spatial and temporal priors to produce more complete and coherent reconstructions in challenging compositional settings.

    \vspace{0.5cm}

    \centering
    \setlength{\tabcolsep}{0pt}
    \renewcommand{\arraystretch}{0}
    \newcommand{\rowlabel}[1]{%
      \rotatebox{90}{\makebox[0.4cm][c]{\footnotesize #1}}%
    }
    \begin{tabular}{m{0.03\columnwidth} c c c c}
        \rowlabel{~~~~~~~~~~~~~~~~~~~~~~~Input} &
        \includegraphics[trim=750 100 550 450,clip,width=0.235\columnwidth,height=0.2\columnwidth]{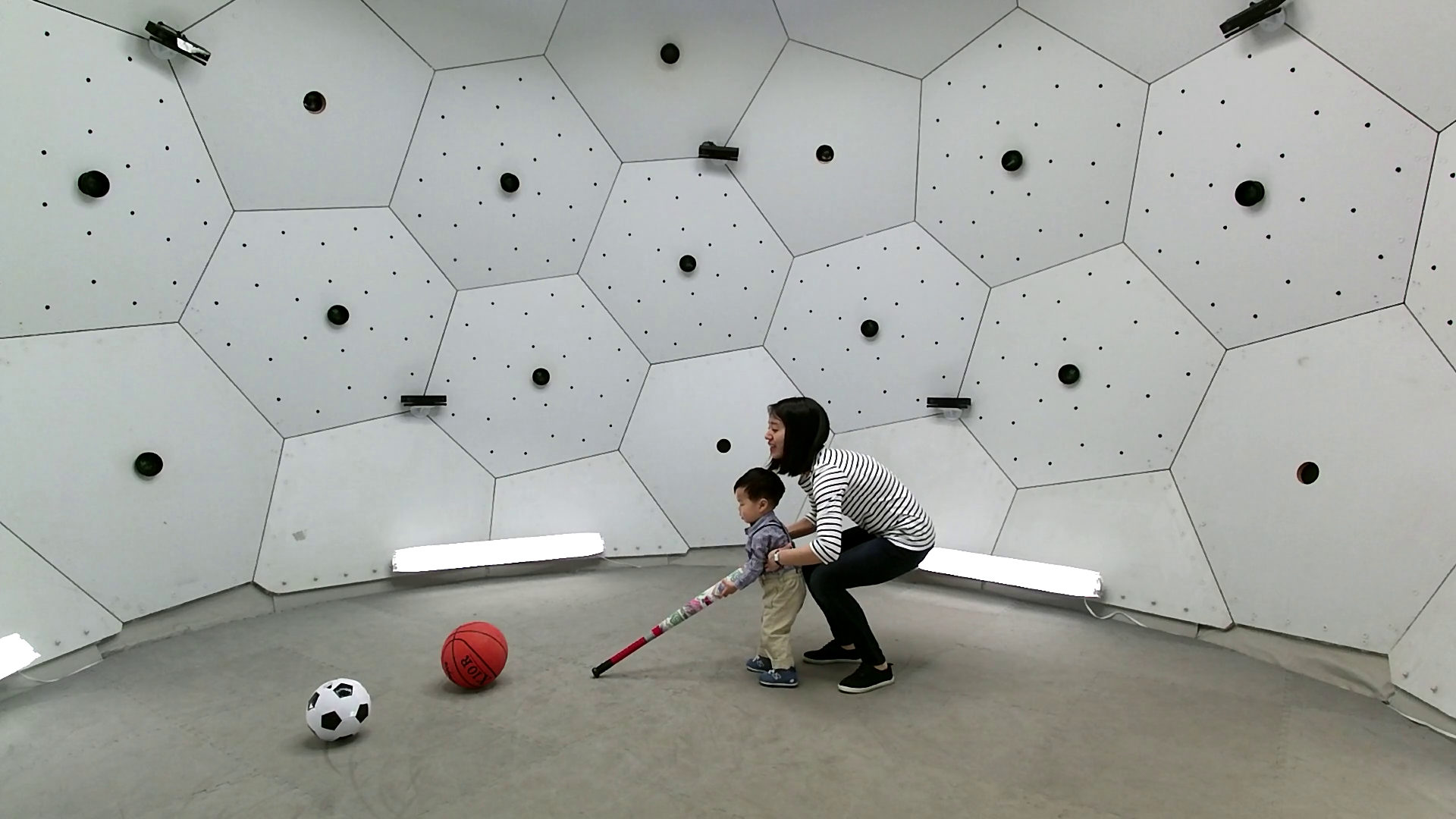} &
        \includegraphics[trim=500 15 600 350,clip,width=0.235\columnwidth,height=0.2\columnwidth]{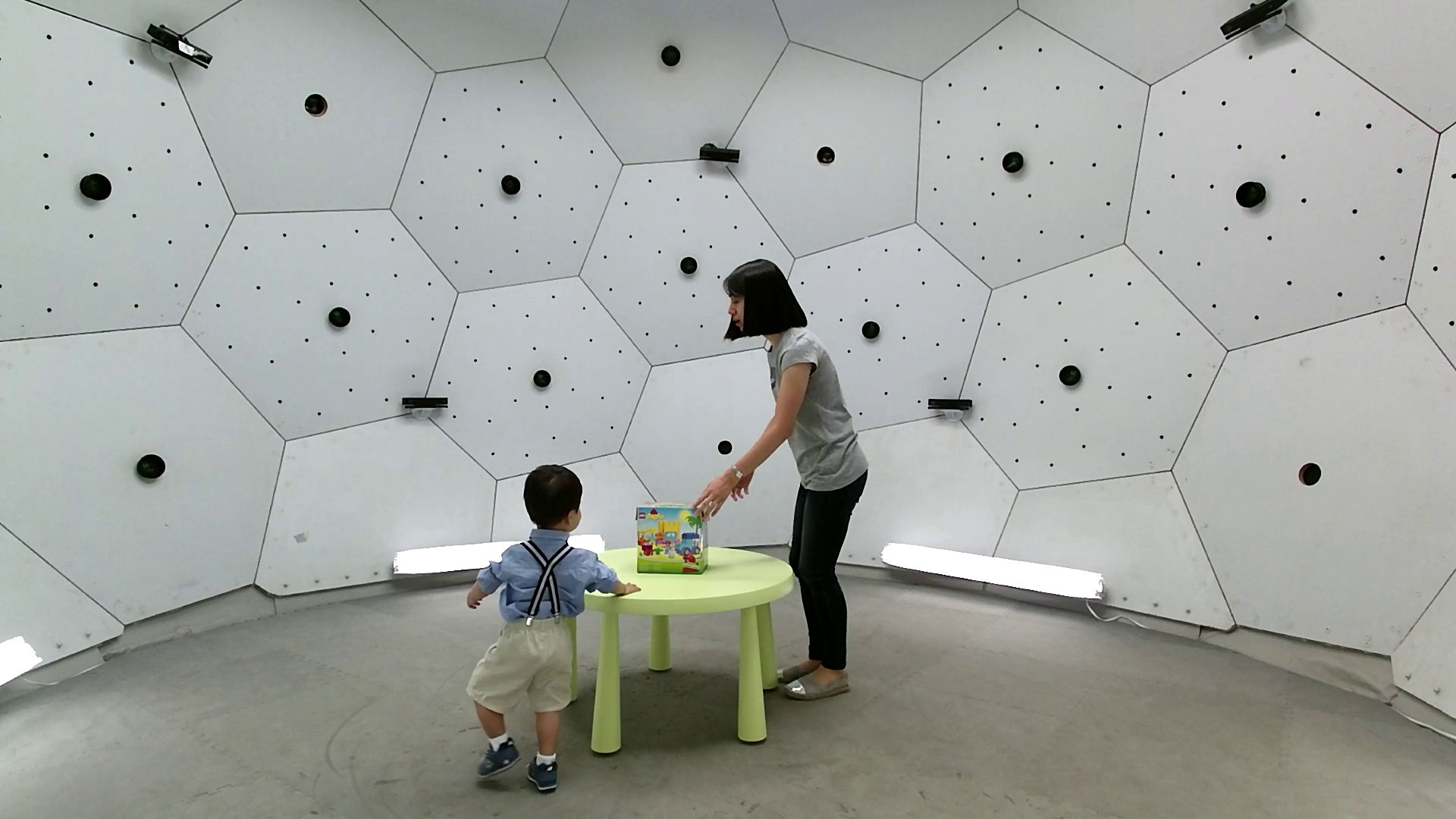} &
        \includegraphics[trim=200 0 700 210,clip,width=0.235\columnwidth,height=0.2\columnwidth]{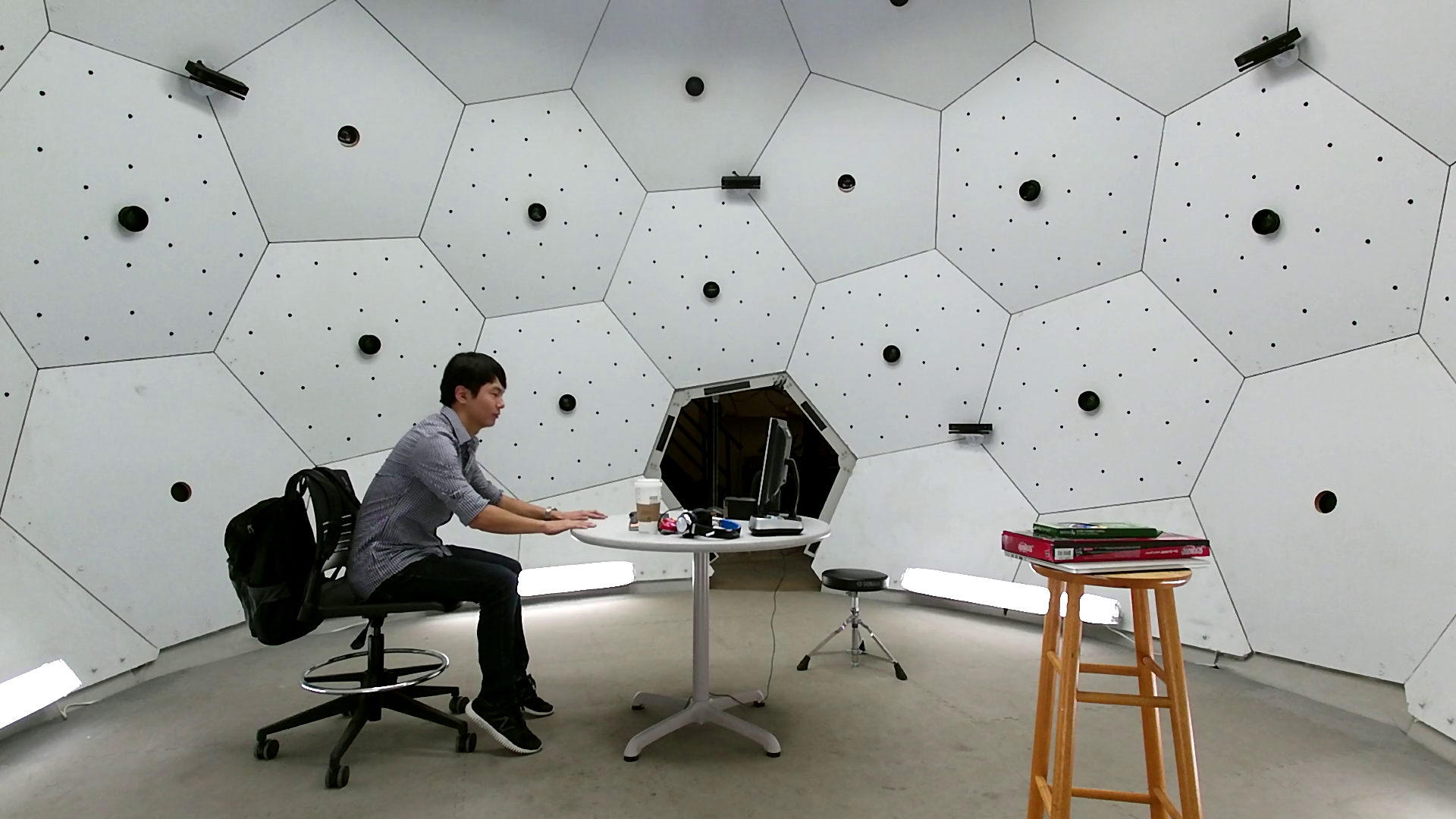} &
        \includegraphics[trim=100 120 100 180,clip,width=0.235\columnwidth,height=0.2\columnwidth]{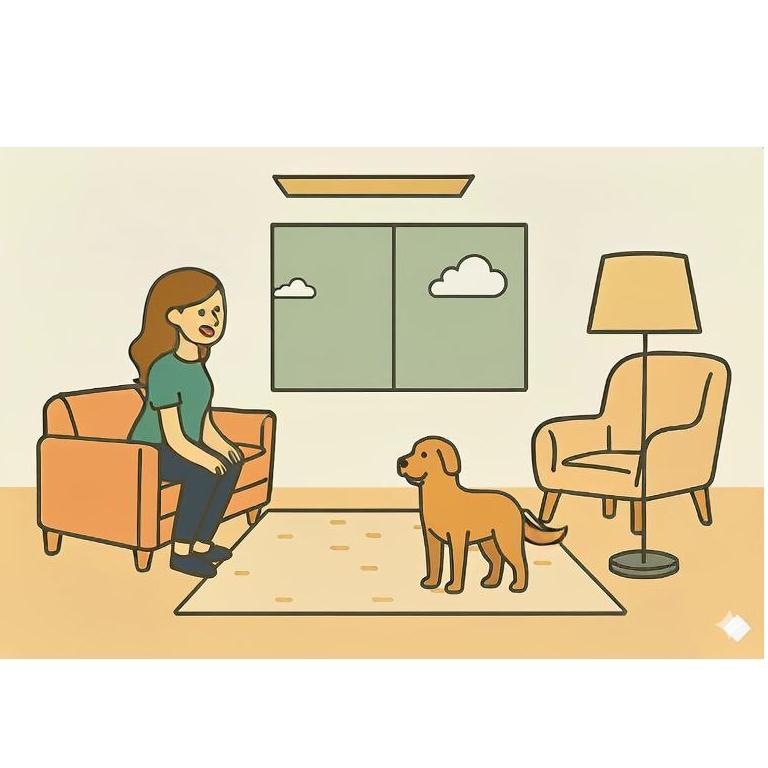} \\
        \rowlabel{~~~~~~~~~~~~~~~~~~SOM \cite{som2024}} &
        \includegraphics[trim=0 50 0 50,clip,width=0.235\columnwidth,height=0.15\columnwidth]{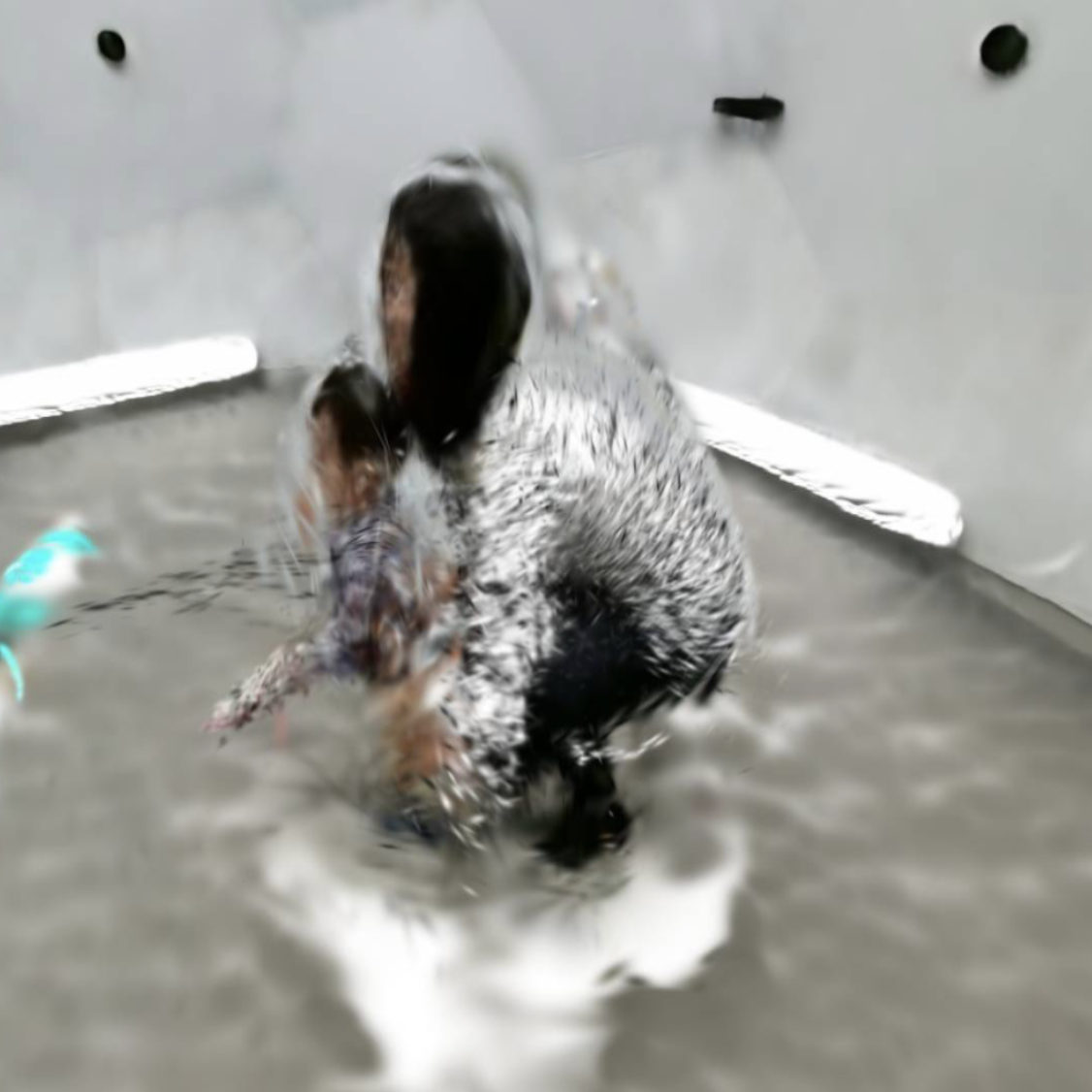} &
        \includegraphics[trim=0 50 0 50,clip,width=0.235\columnwidth,height=0.15\columnwidth]{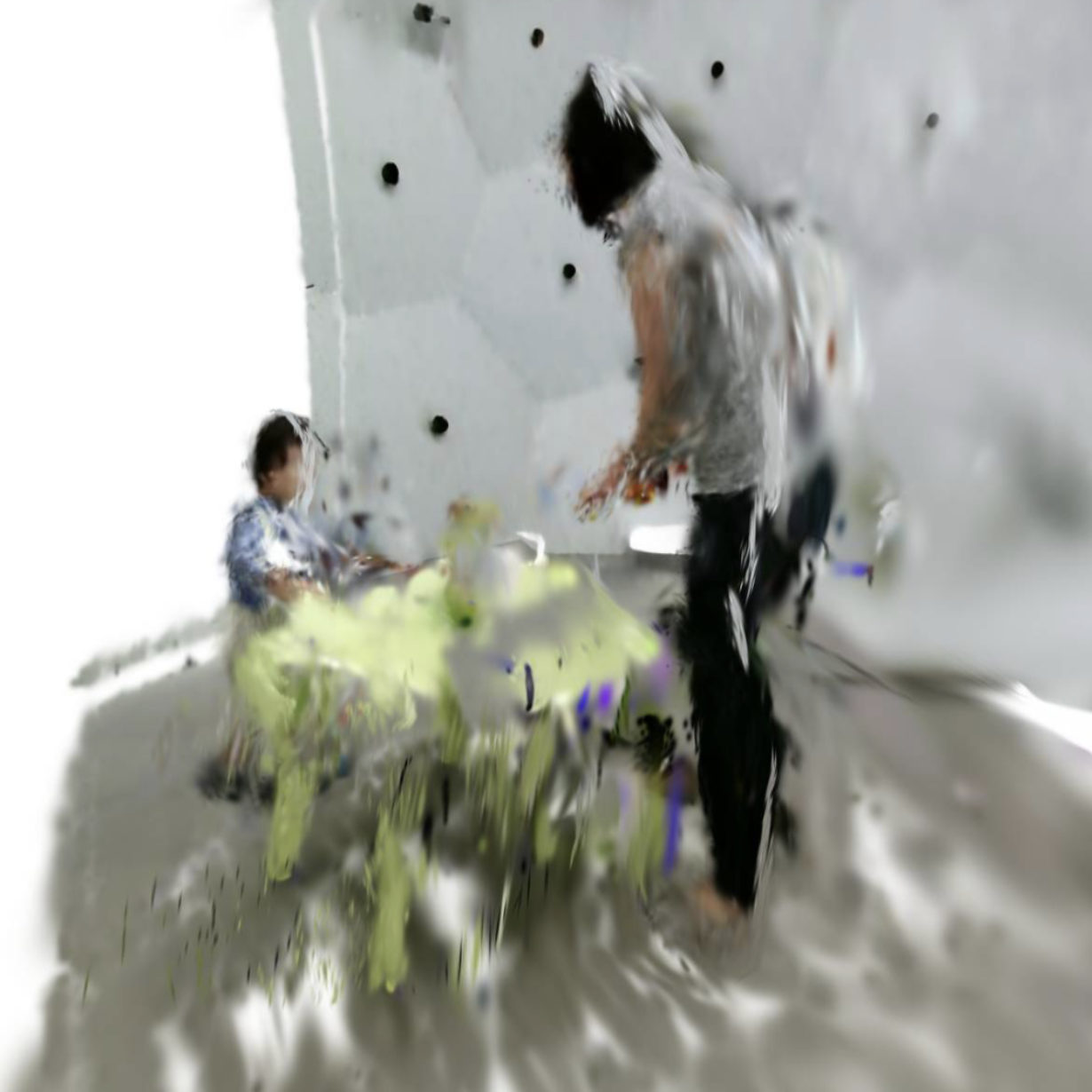} &
        \includegraphics[trim=0 50 0 50,clip,width=0.235\columnwidth,height=0.15\columnwidth]{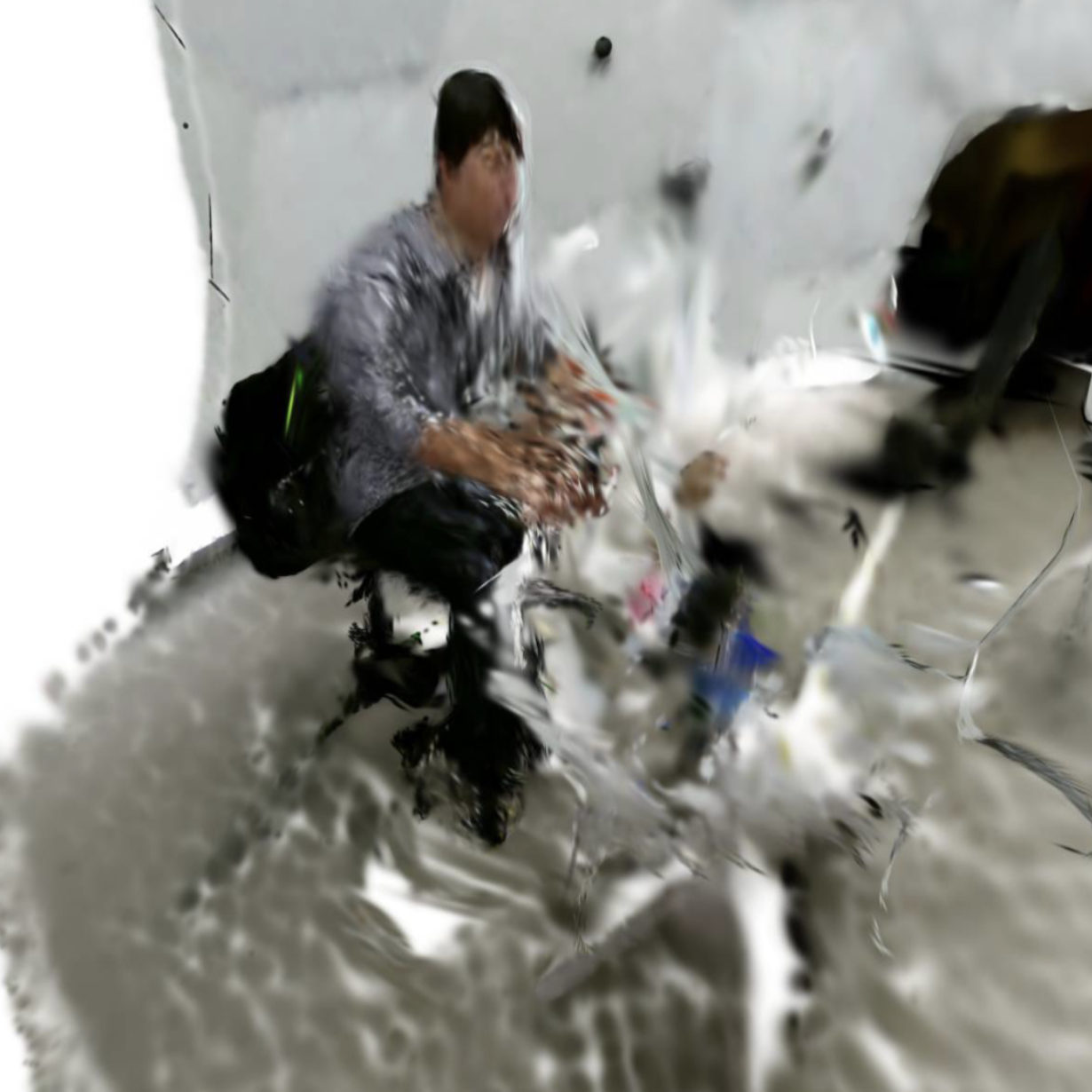} &
        \includegraphics[trim=0 50 0 50,clip,width=0.235\columnwidth,height=0.15\columnwidth]{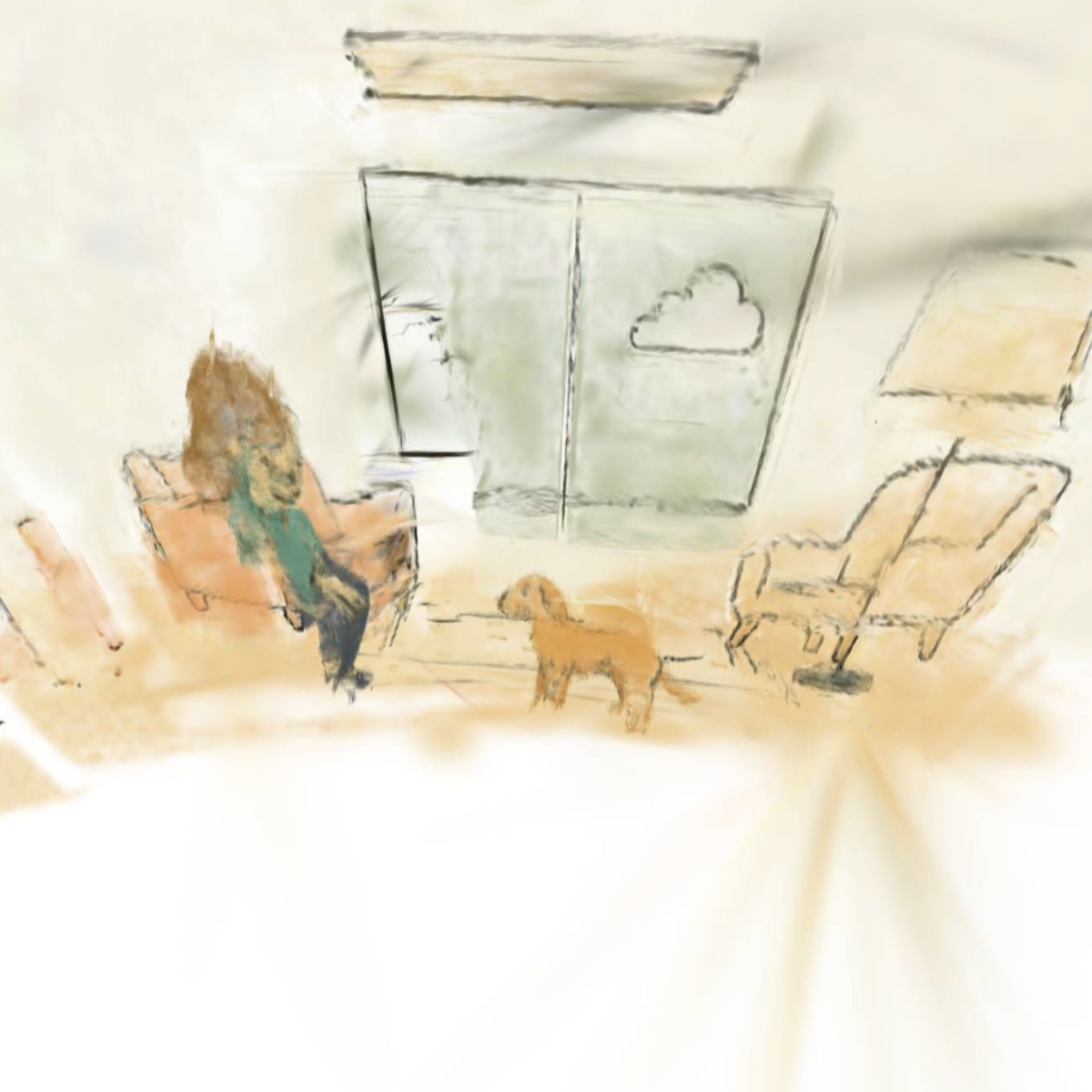} \\
        \rowlabel{~~~~~~~~~~~~~~~~~~~Ours} &
        \includegraphics[trim=200 100 200 350,clip,width=0.235\columnwidth,height=0.2\columnwidth]{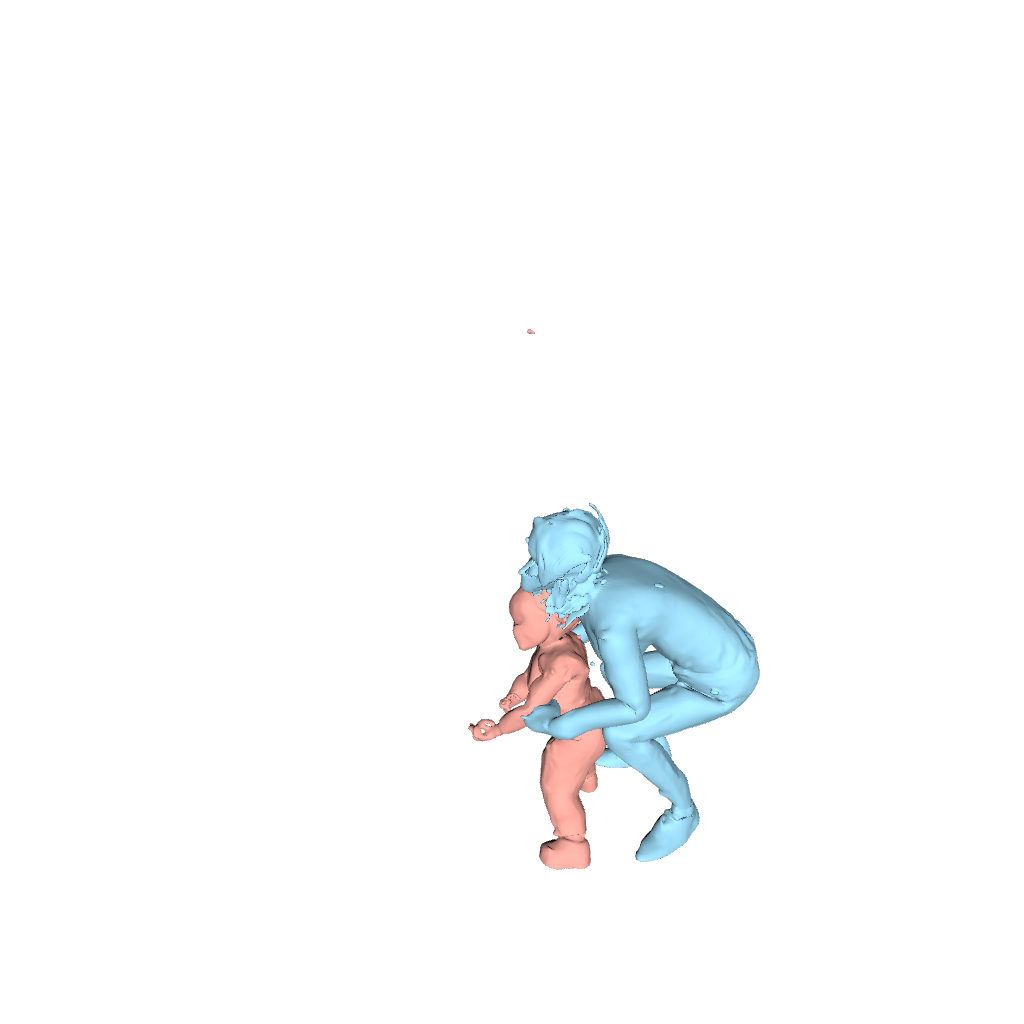} &
        \includegraphics[trim=100 100 100 150,clip,width=0.235\columnwidth,height=0.2\columnwidth]{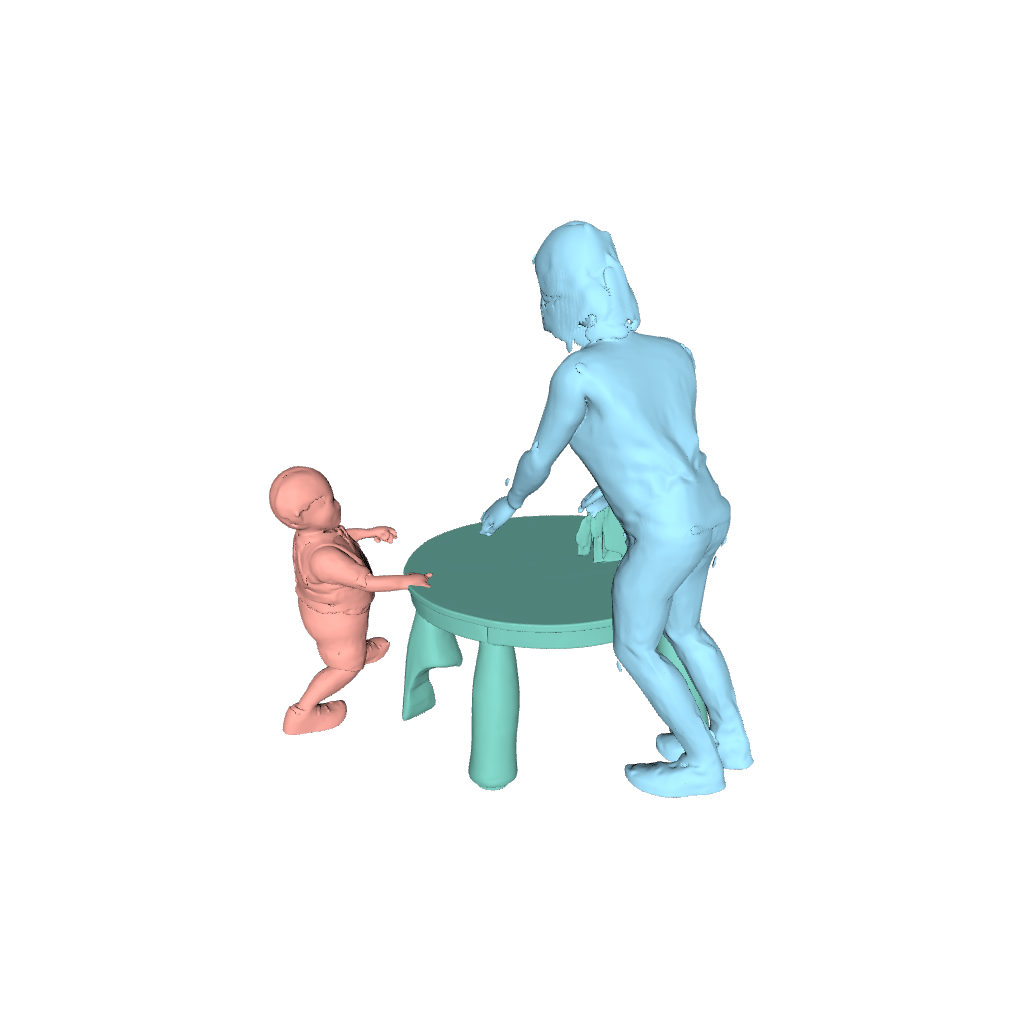} &
        \includegraphics[trim=100 0 0 150,clip,width=0.235\columnwidth,height=0.2\columnwidth]{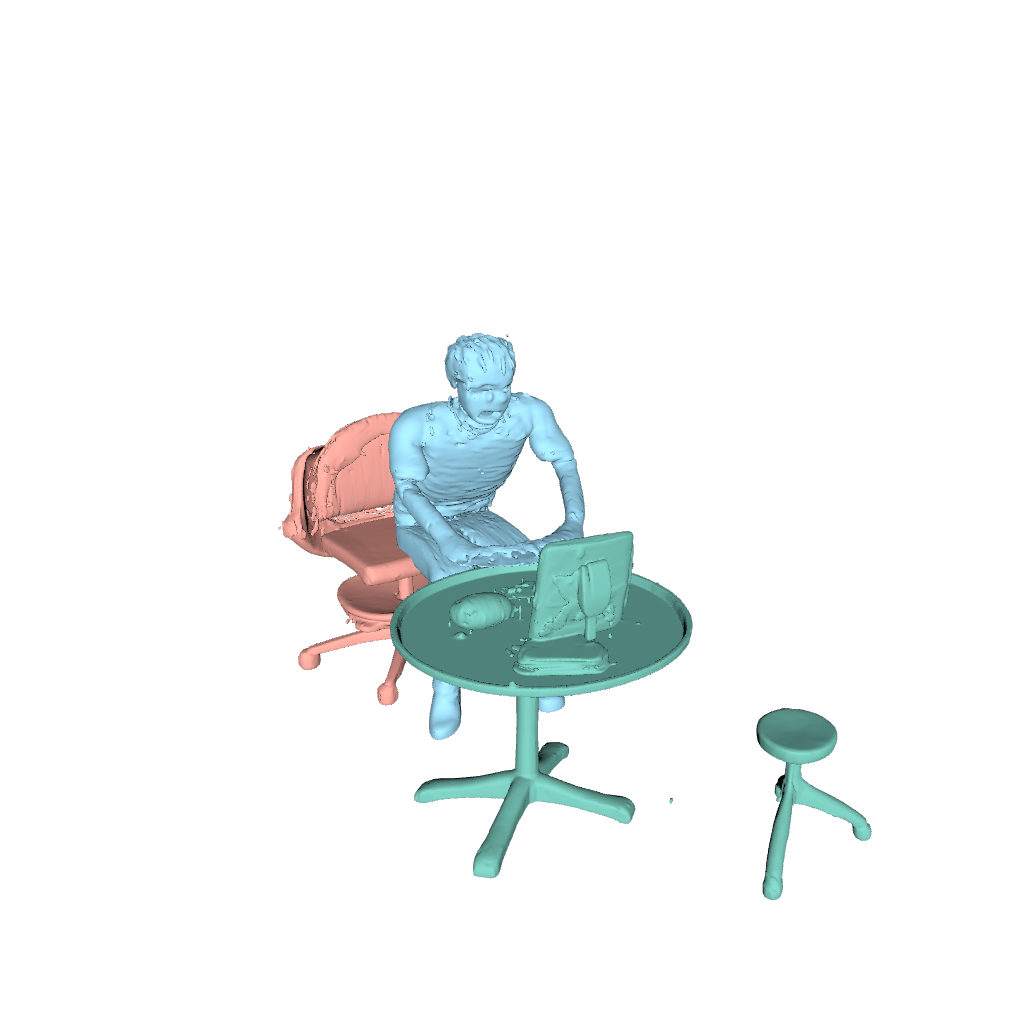} &
        \includegraphics[trim=0 0 0 50,clip,width=0.235\columnwidth,height=0.2\columnwidth]{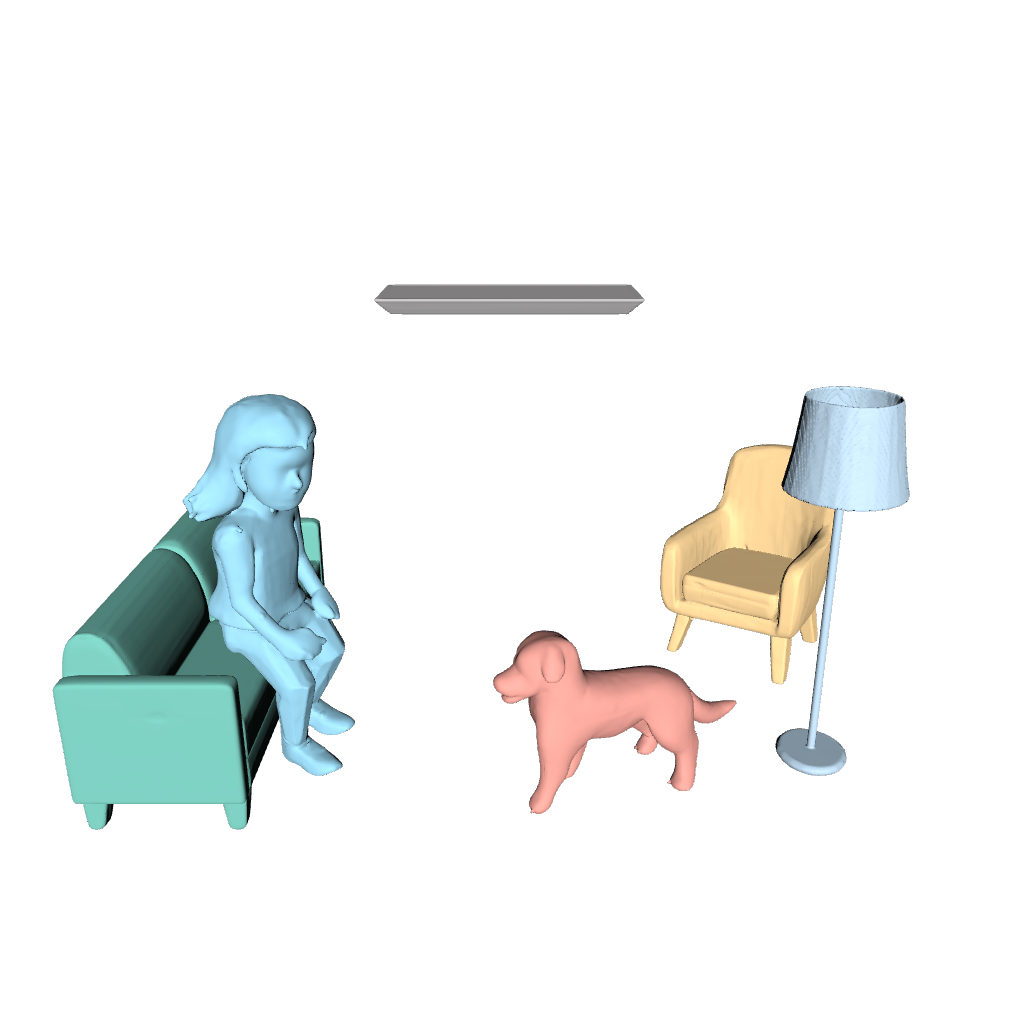}
    \end{tabular}
    \vspace{-10pt}
    \caption{Qualitative comparison across methods.}
    \label{fig:comparison_appendix}
\end{figure}

\begin{table}[h]

\section{Evaluation Using a Synthetic Dataset}

We additionally evaluate our method on a synthetic 4D dataset since a suitable real-captured 4D scene benchmark is unavailable. We first generate candidate synthetic scenes by composing one or more dynamic humanoid motion sequences with several static objects. Each dynamic object is selected from a motion sequence, sub-sampled with a fixed frame stride, and placed into a shared normalized world coordinate system. Static assets are then scaled, rotated around the vertical axis, and translated so that they fit around the dynamic trajectories without excessive overlap. For every time step, we export a combined GLB that contains the static geometry and the current pose of each dynamic object, and we also save metadata describing the source assets, transforms, per-frame bounds, and global centering applied to the scene.

For visualization, each frame is rendered from a fixed orthographic camera with consistent framing across the entire sequence. The renderer keeps the scene centered and uses the same camera and lighting conventions for all frames within a sample, which makes frame-to-frame motion directly comparable.

For quantitative evaluation, we compare the generated scene-level mesh against the static part of the reference scene and compute dynamic scores per moving object and per frame. For static geometry, we compute one CD/F-score pair per scene and report the arithmetic mean over scenes. For dynamic content, we first average all valid dynamic per-frame scores within each sample, and then average these per-sample dynamic means across the whole set. This prevents samples with more moving objects or more valid frames from dominating the final measure.

\vspace{10pt}

{
\centering \small
\begin{tabular}{lcc}
    \hline
    Component & CD $\downarrow$ & F-score $\uparrow$ \\
    \hline
    Static scene geometry & 0.293 & 0.418 \\
    Dynamic objects & 0.284 & 0.434 \\
    \hline
\end{tabular} \caption{Quantitative results on the synthetic 4D evaluation set. Lower CD is better; higher F-score is better.} \label{tab:synthetic_eval}
\par
}

\vspace{10pt}

\noindent
These synthetic-scene scores are worse than the corresponding single-object 4D and static compositional 3D scene results. The main reason is that, although the reconstructed objects themselves and their coarse arrangements are often visually consistent with the input, the relative positioning of objects is still somewhat inaccurate. Small errors in placement and pairwise spatial relationships accumulate at the full-scene level and are directly penalized by both CD and F-score, leading to lower quantitative performance than in simpler single-object or static-only settings.

\end{table}

\begin{figure}[h]
    \section{Additional Qualitative Results on Synthetic Compositional 4D Dataset}
    
    In \cref{fig:synthetic_com4d_appendix_qual}, \cref{fig:synthetic_com4d_appendix_qual_2}, \cref{fig:synthetic_com4d_appendix_qual_3} we show additional qualitative results on our synthetic compositional 4D dataset.
    \vspace{10pt}
      
  \centering
  \setlength{\tabcolsep}{0pt}
  \renewcommand{\arraystretch}{0}

  \begin{tabular}{@{}c@{\hspace{2pt}}cccc@{}}

        % ==================== SAMPLE 1: ortiz_Flair_20260324_005515 ====================
        \raisebox{-0.5\height}{\rotatebox{90}{\tiny Input}} &
        \raisebox{-0.5\height}{\includegraphics[width=0.23\columnwidth, trim=0 0 0 0, clip]{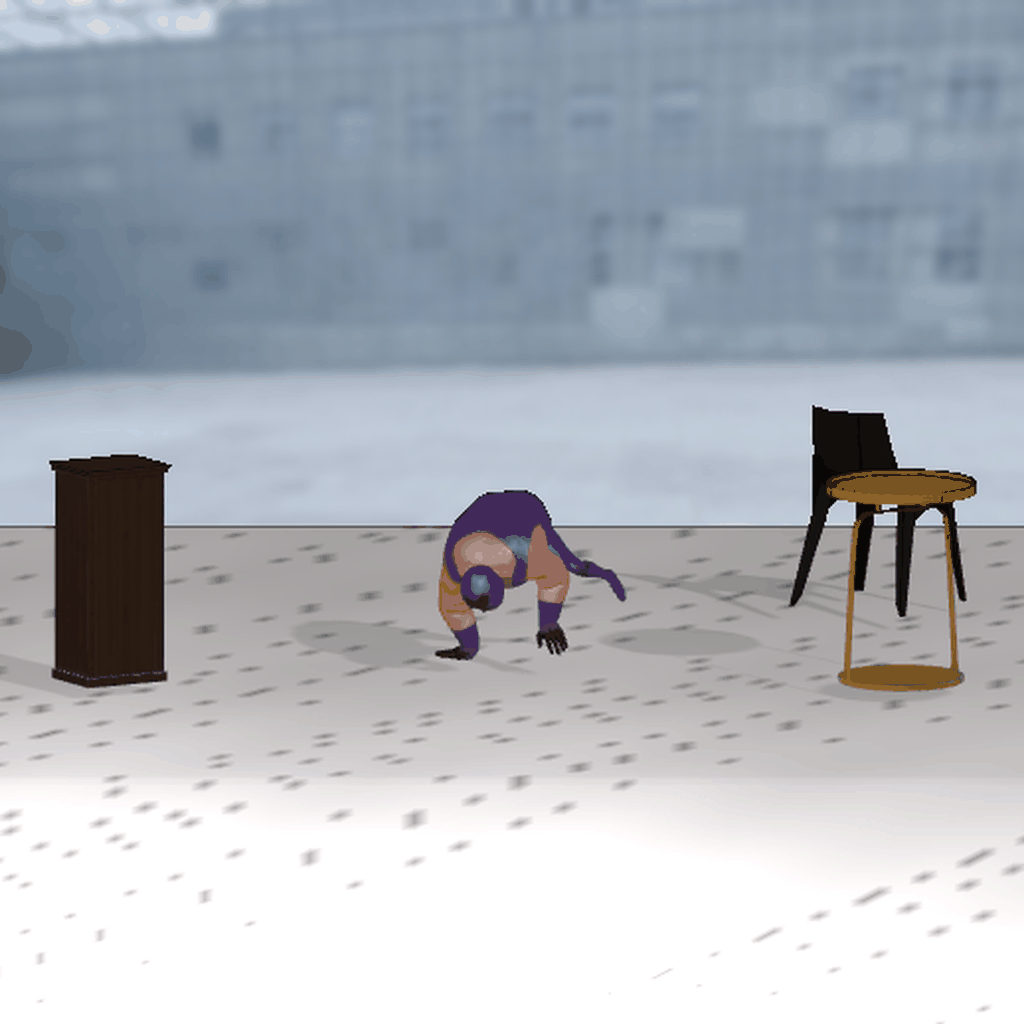}} &
        \raisebox{-0.5\height}{\includegraphics[width=0.23\columnwidth, trim=0 0 0 0, clip]{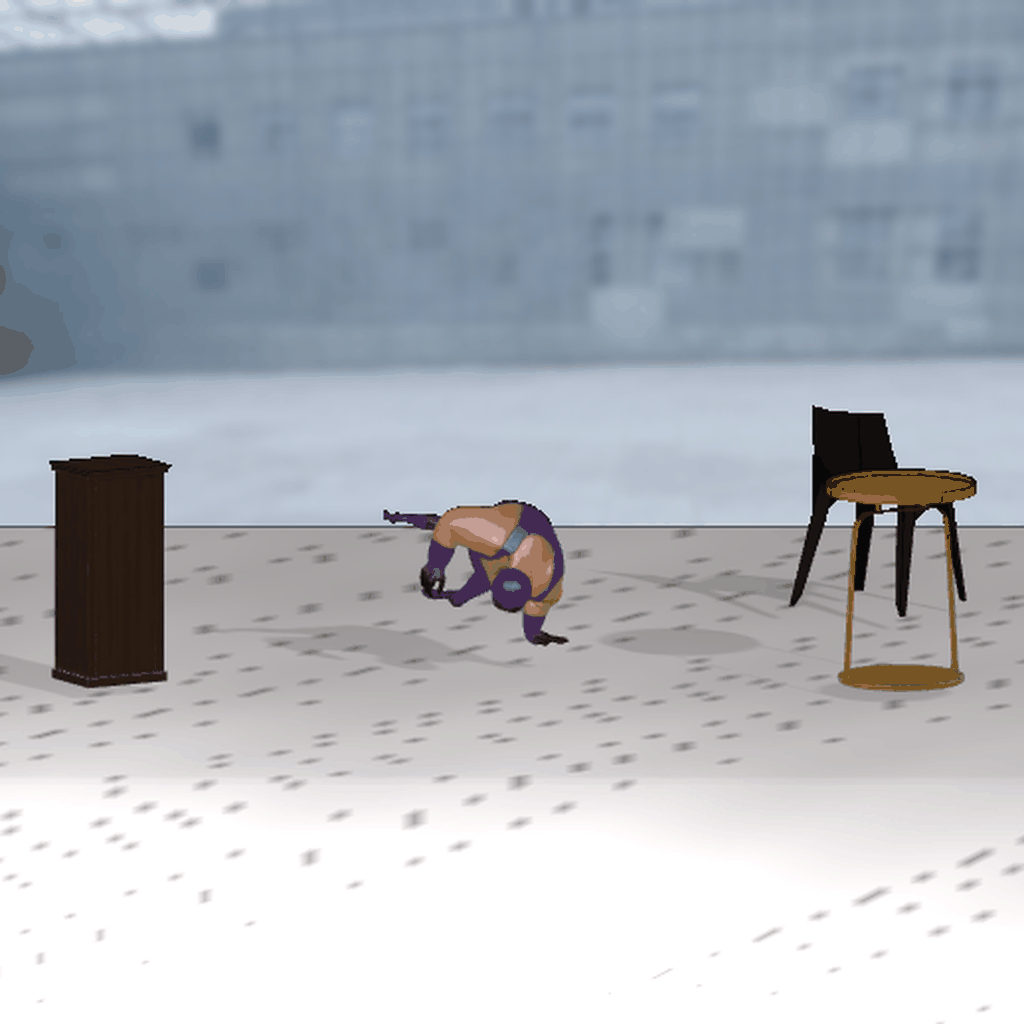}} &
        \raisebox{-0.5\height}{\includegraphics[width=0.23\columnwidth, trim=0 0 0 0, clip]{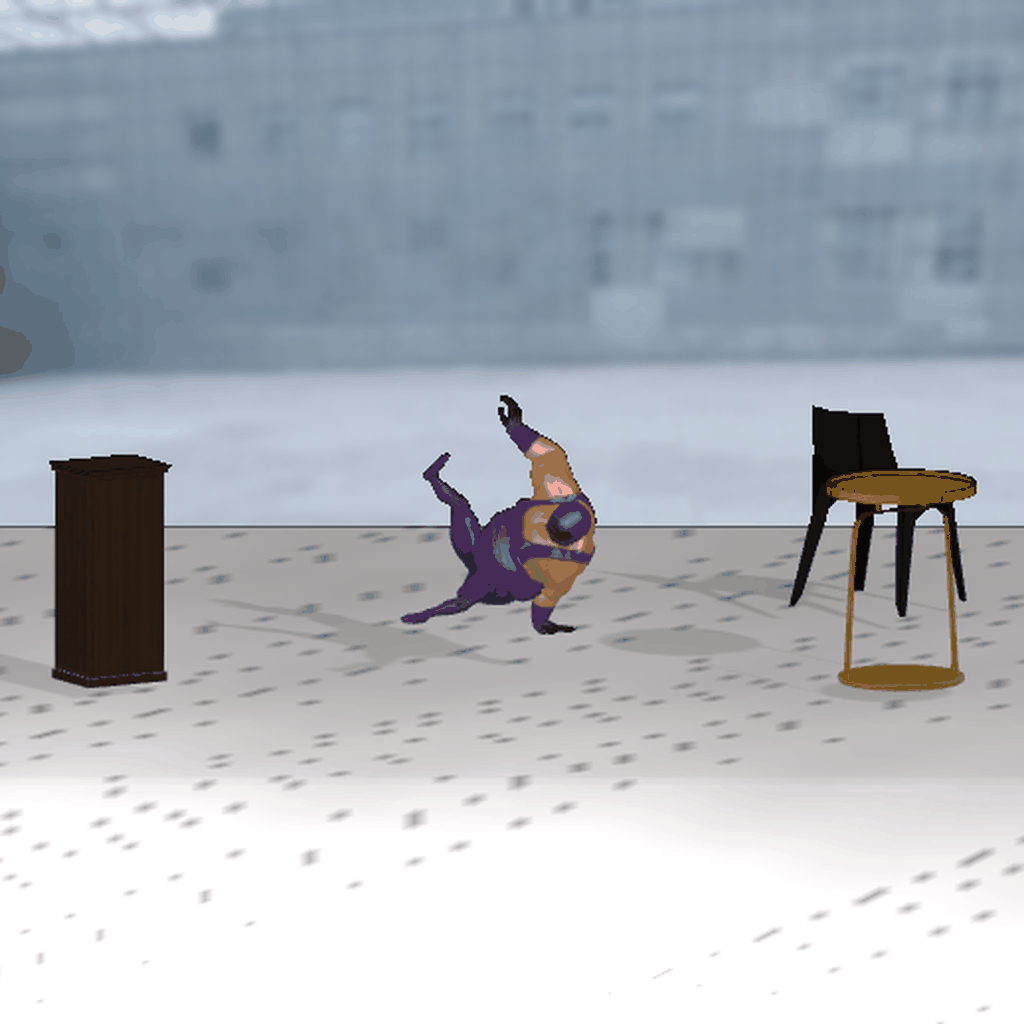}} &
        \raisebox{-0.5\height}{\includegraphics[width=0.23\columnwidth, trim=0 0 0 0, clip]{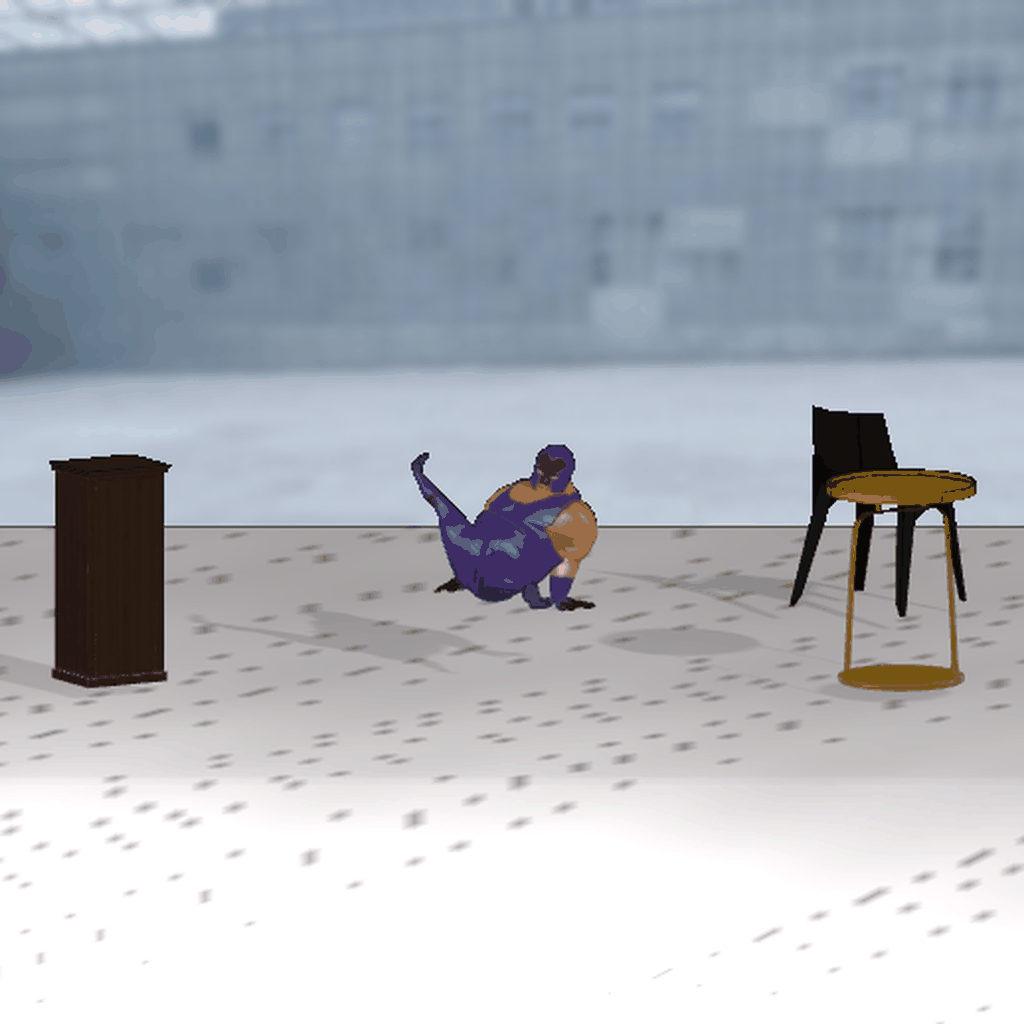}} \\[-0.5pt]
        \raisebox{-0.5\height}{\rotatebox{90}{\tiny COM4D}} &
        \raisebox{-0.5\height}{\includegraphics[width=0.23\columnwidth, trim=0 0 0 0, clip]{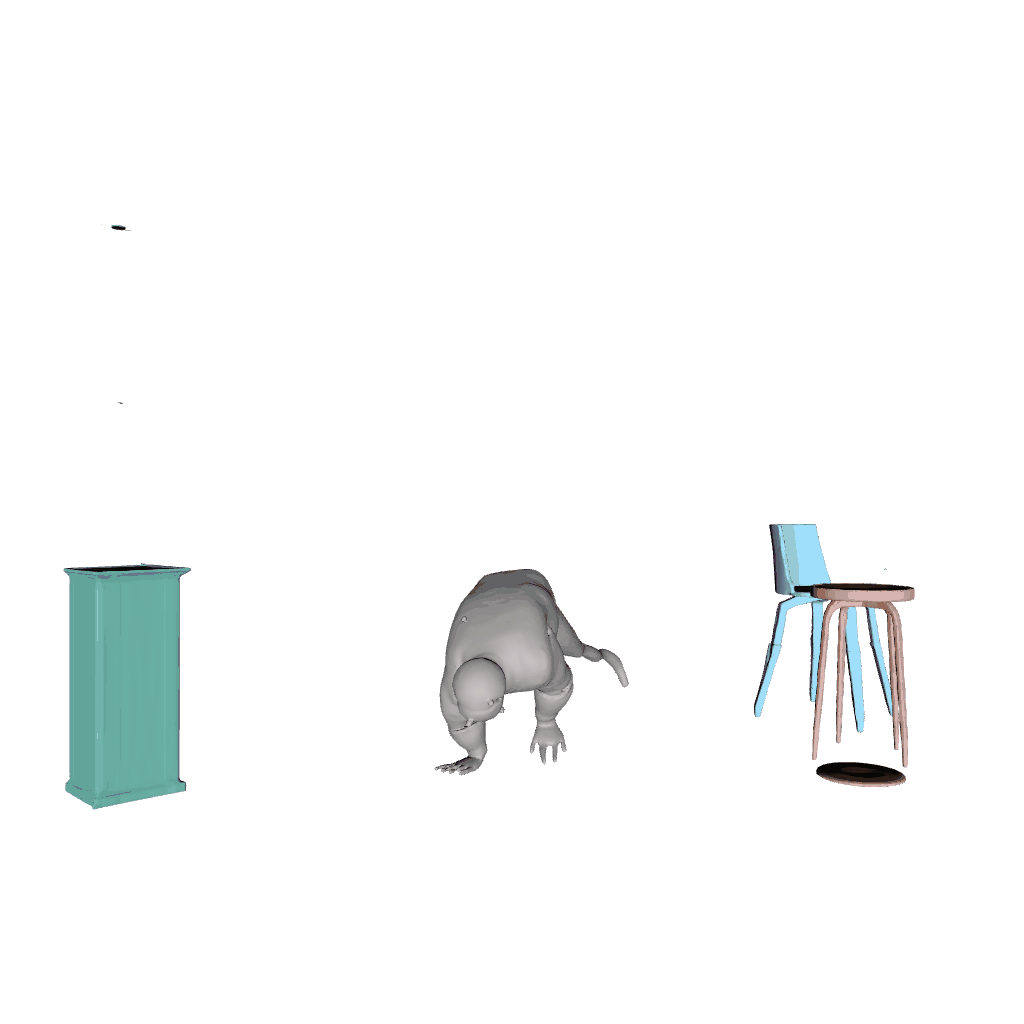}} &
        \raisebox{-0.5\height}{\includegraphics[width=0.23\columnwidth, trim=0 0 0 0, clip]{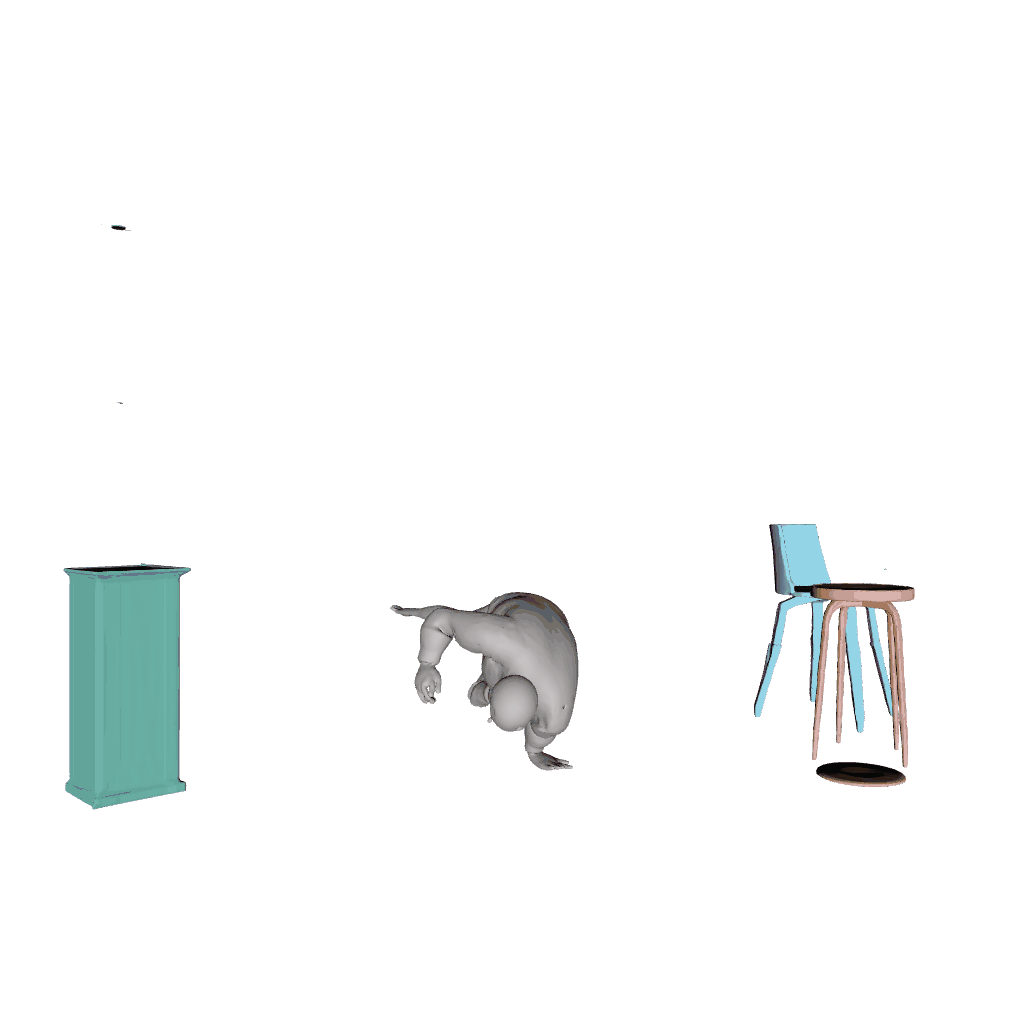}} &
        \raisebox{-0.5\height}{\includegraphics[width=0.23\columnwidth, trim=0 0 0 0, clip]{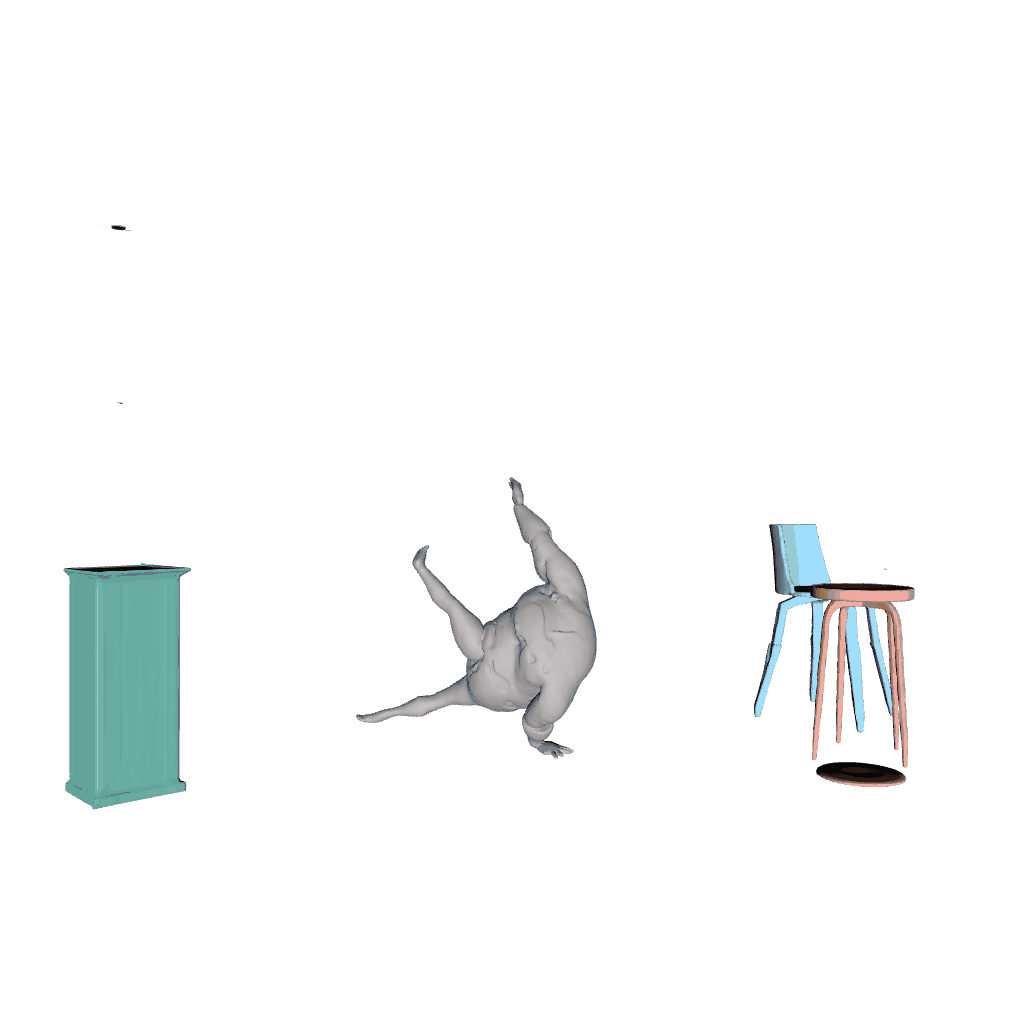}} &
        \raisebox{-0.5\height}{\includegraphics[width=0.23\columnwidth, trim=0 0 0 0, clip]{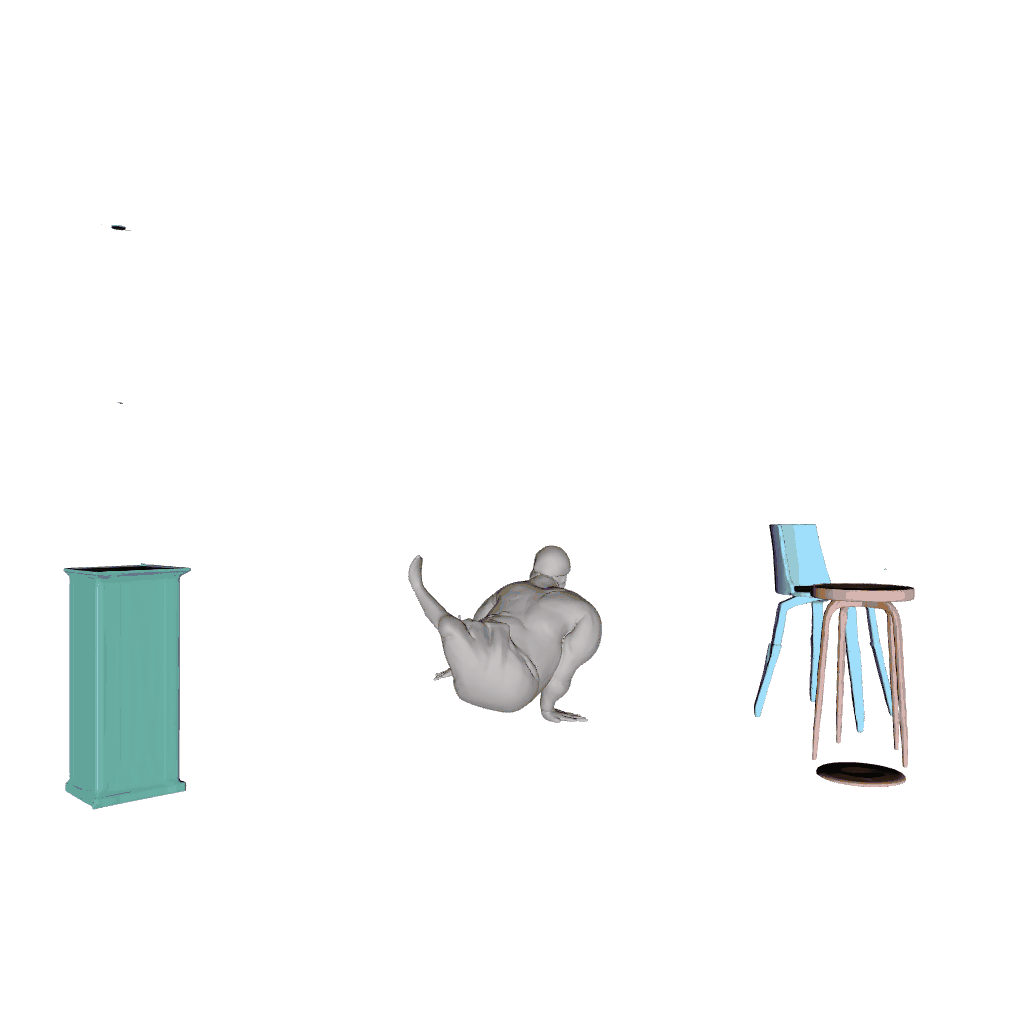}} \\[4pt]
        
        % ==================== SAMPLE 2: mousey_HipHopDancing_20260324_012223 ====================
        \raisebox{-0.5\height}{\rotatebox{90}{\tiny Input}} &
        \raisebox{-0.5\height}{\includegraphics[width=0.23\columnwidth, trim=0 0 0 0, clip]{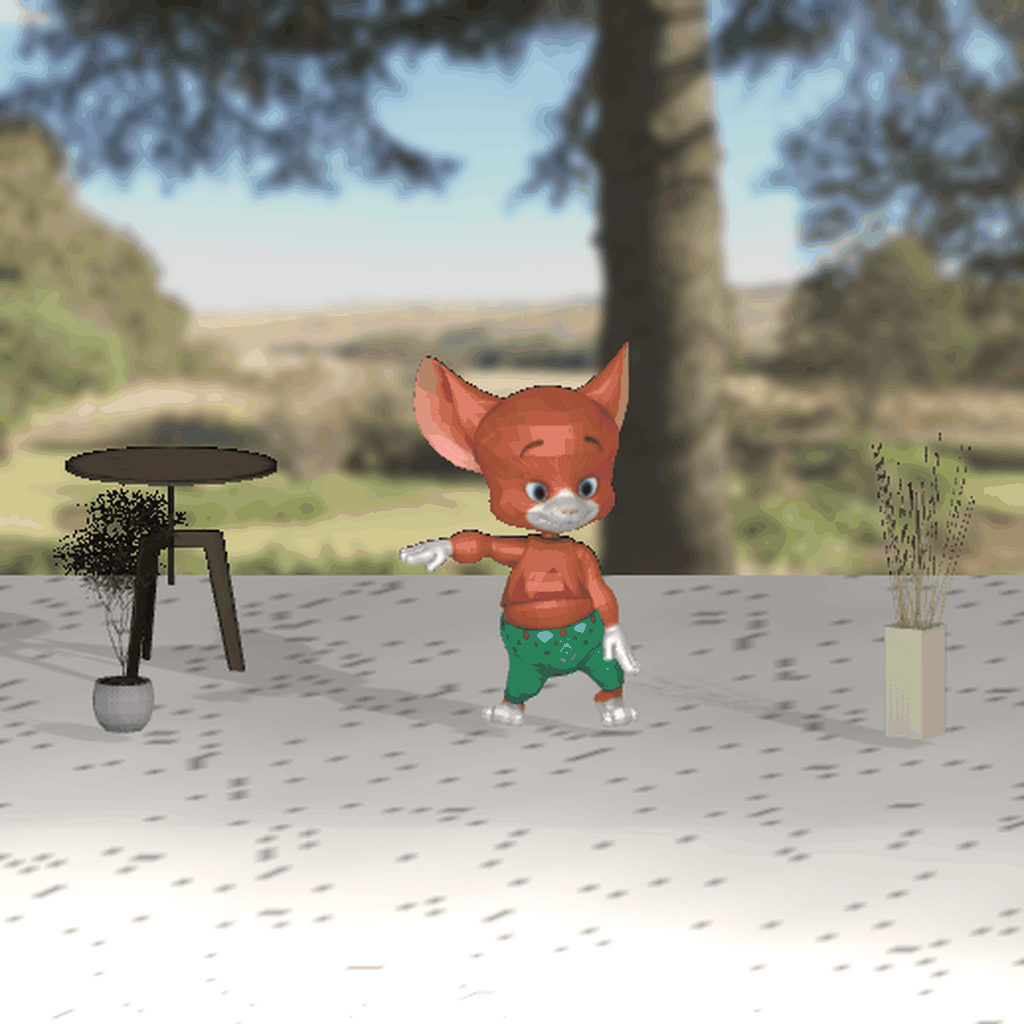}} &
        \raisebox{-0.5\height}{\includegraphics[width=0.23\columnwidth, trim=0 0 0 0, clip]{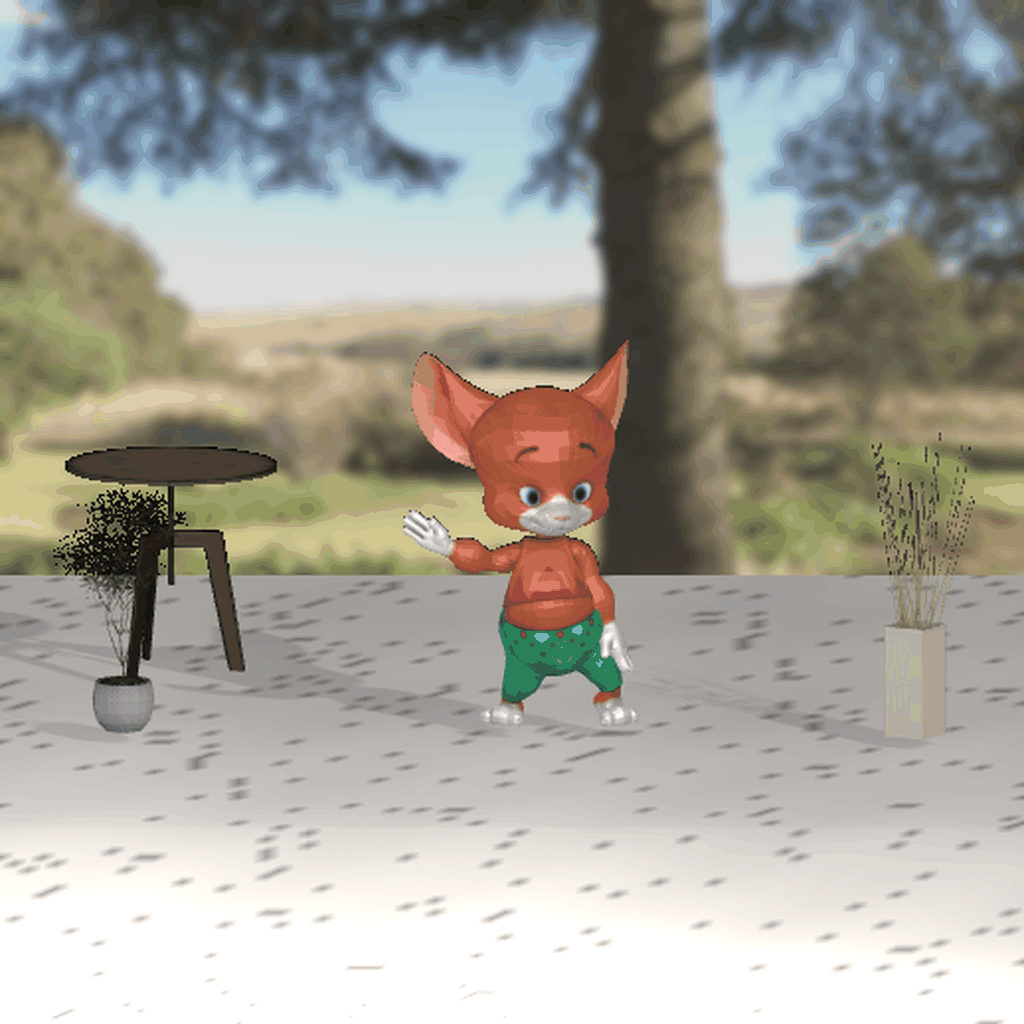}} &
        \raisebox{-0.5\height}{\includegraphics[width=0.23\columnwidth, trim=0 0 0 0, clip]{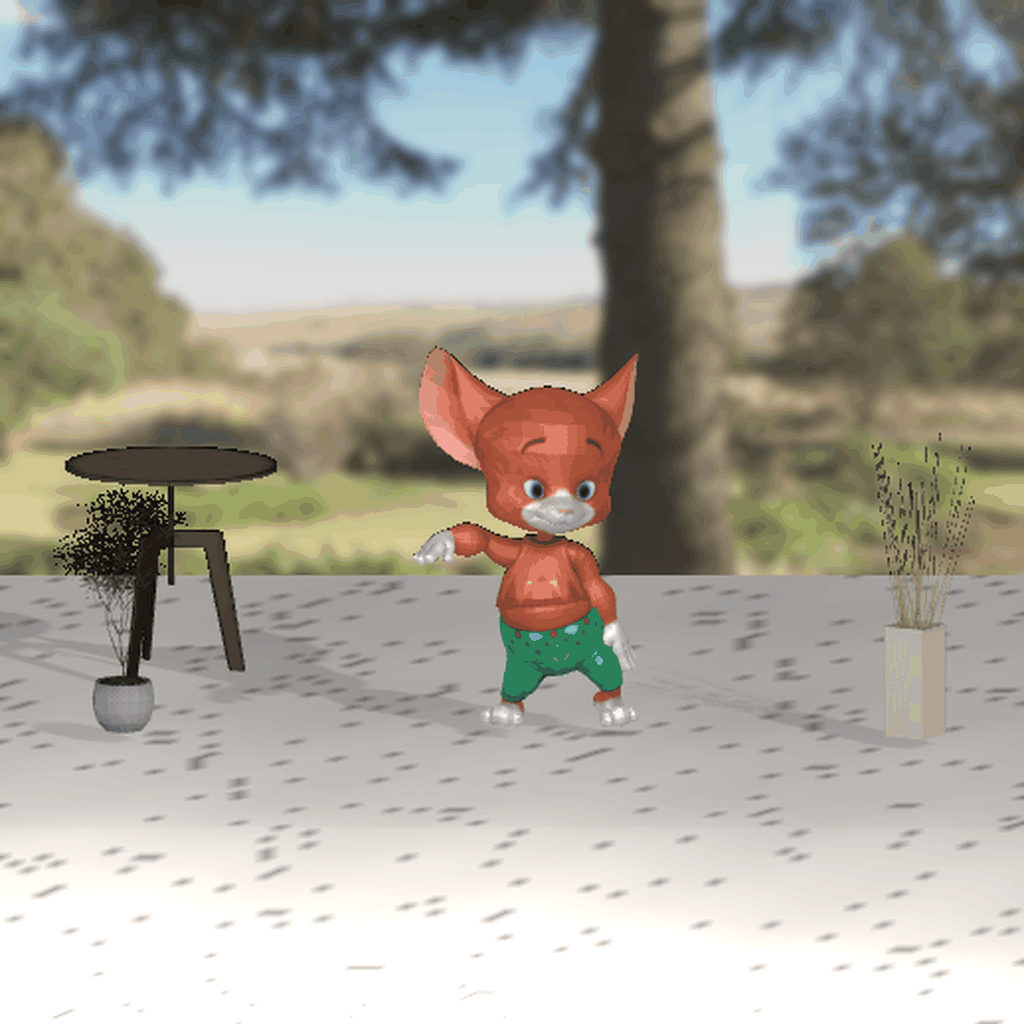}} &
        \raisebox{-0.5\height}{\includegraphics[width=0.23\columnwidth, trim=0 0 0 0, clip]{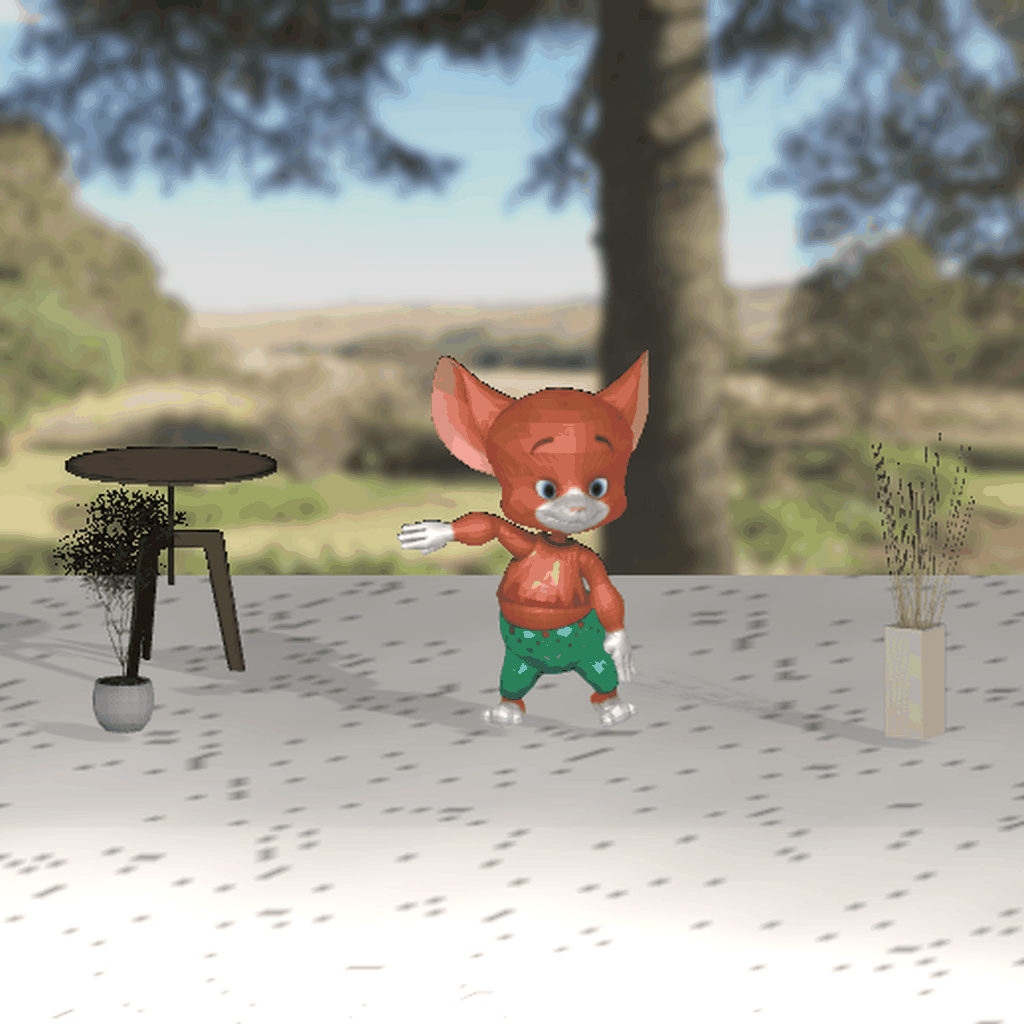}} \\[-0.5pt]
        \raisebox{-0.5\height}{\rotatebox{90}{\tiny COM4D}} &
        \raisebox{-0.5\height}{\includegraphics[width=0.23\columnwidth, trim=0 0 0 0, clip]{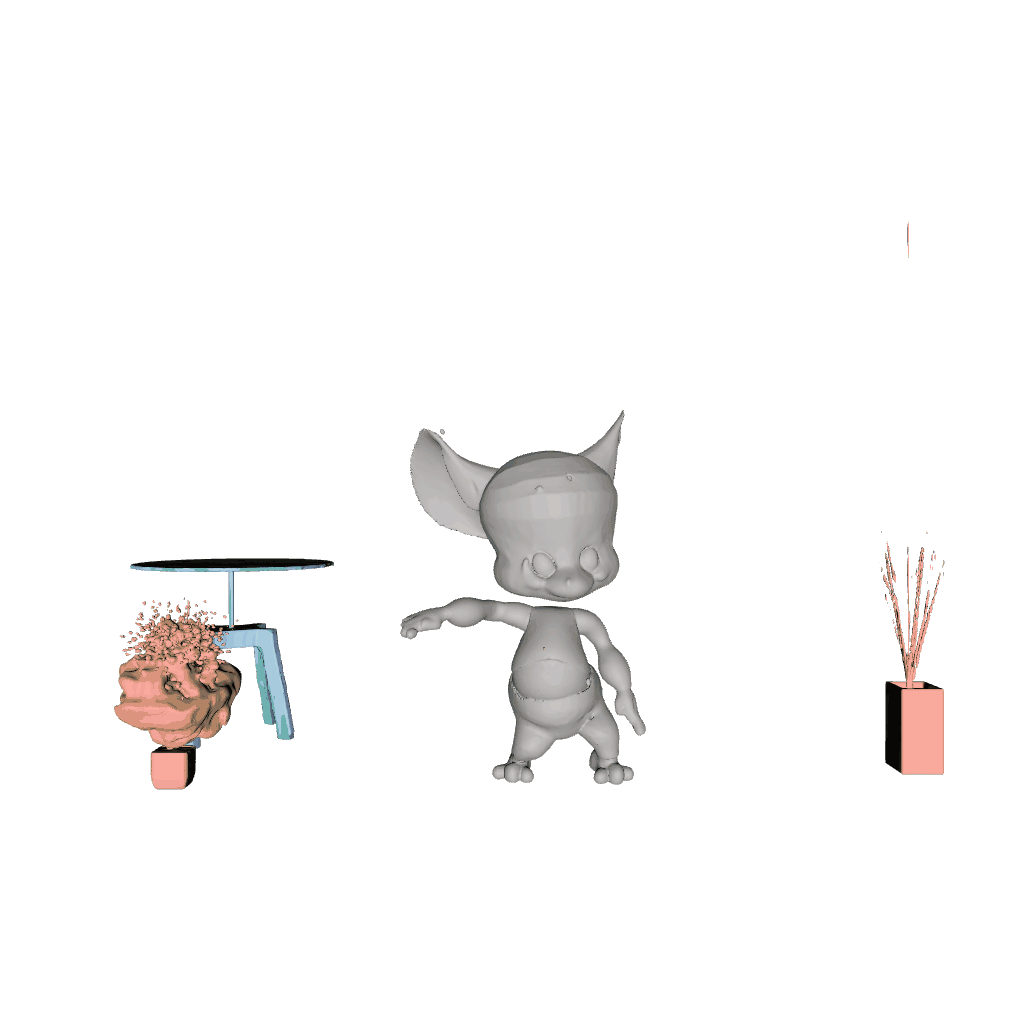}} &
        \raisebox{-0.5\height}{\includegraphics[width=0.23\columnwidth, trim=0 0 0 0, clip]{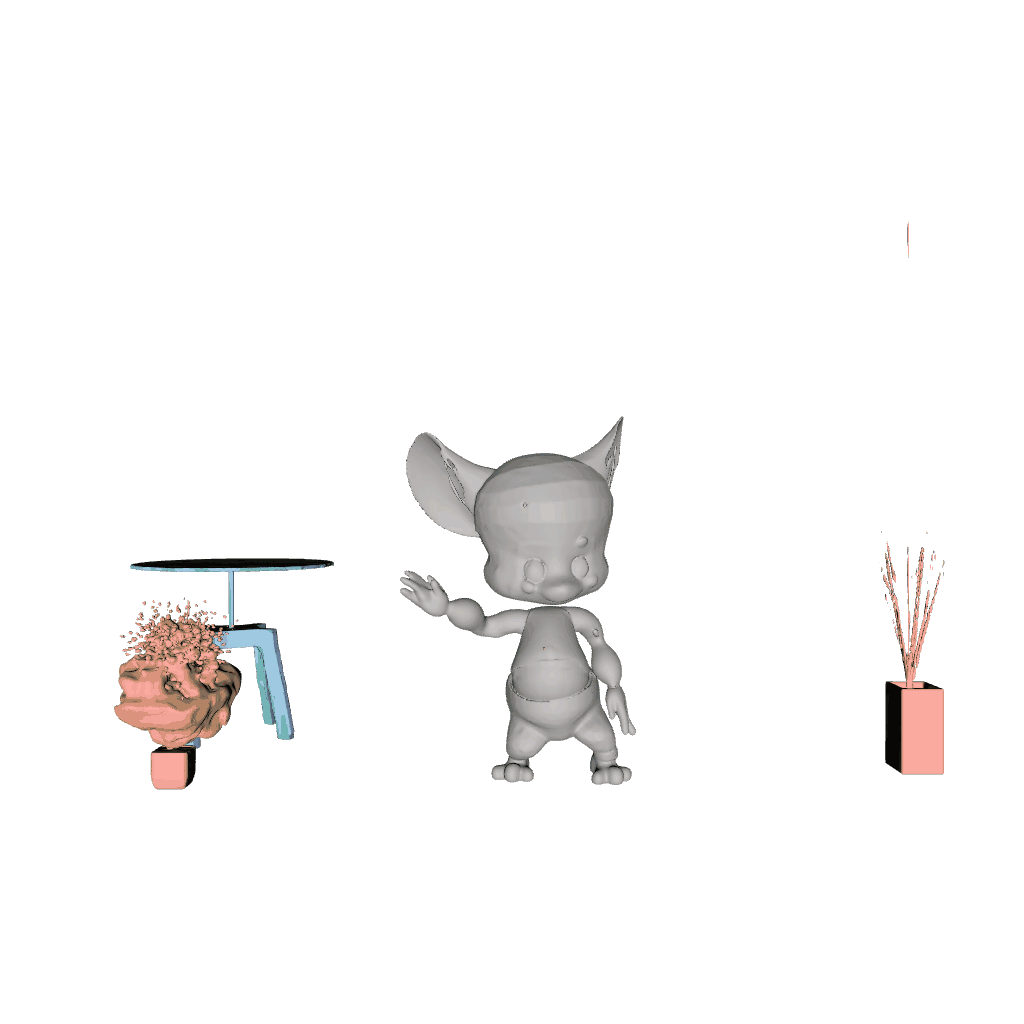}} &
        \raisebox{-0.5\height}{\includegraphics[width=0.23\columnwidth, trim=0 0 0 0, clip]{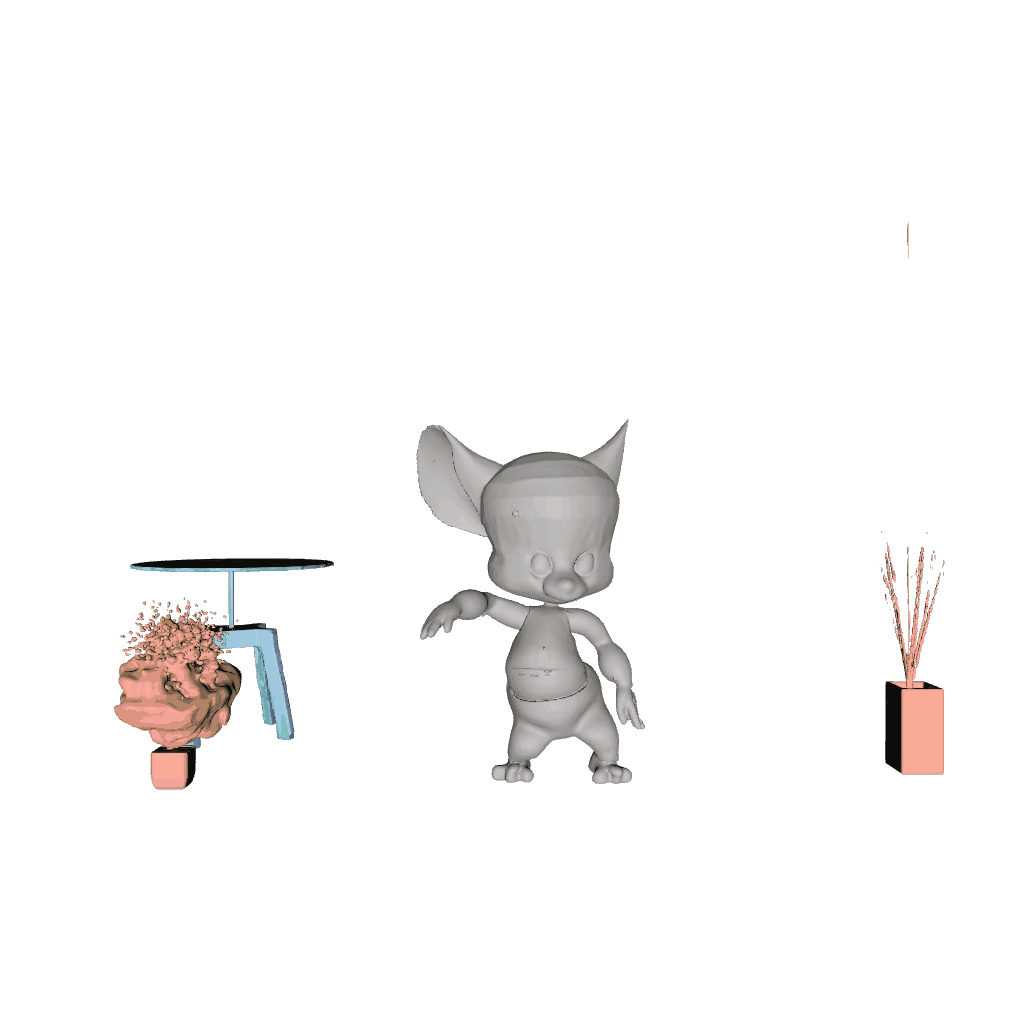}} &
        \raisebox{-0.5\height}{\includegraphics[width=0.23\columnwidth, trim=0 0 0 0, clip]{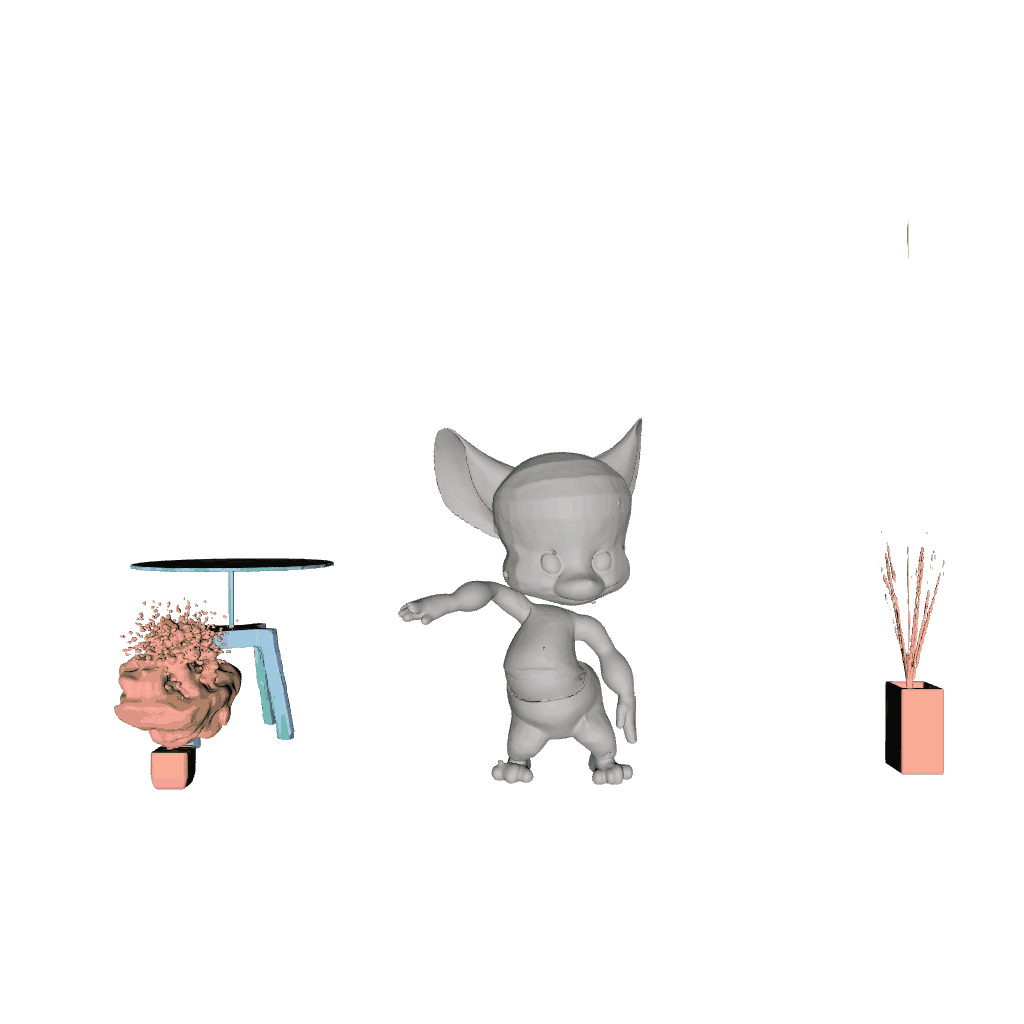}} \\[4pt]
        
        % ==================== SAMPLE 3: ninja_BigSideHit_20260324_003706 ====================
        \raisebox{-0.5\height}{\rotatebox{90}{\tiny Input}} &
        \raisebox{-0.5\height}{\includegraphics[width=0.23\columnwidth, trim=0 0 0 0, clip]{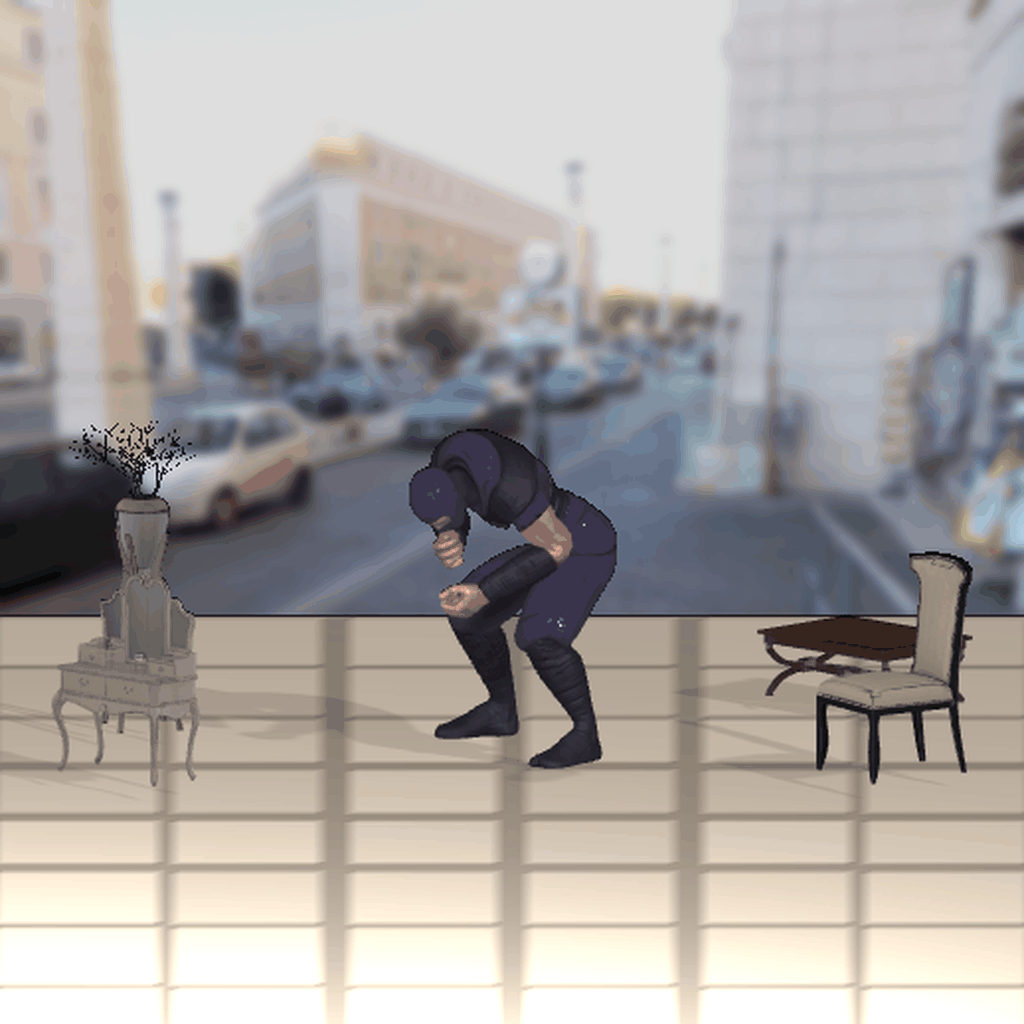}} &
        \raisebox{-0.5\height}{\includegraphics[width=0.23\columnwidth, trim=0 0 0 0, clip]{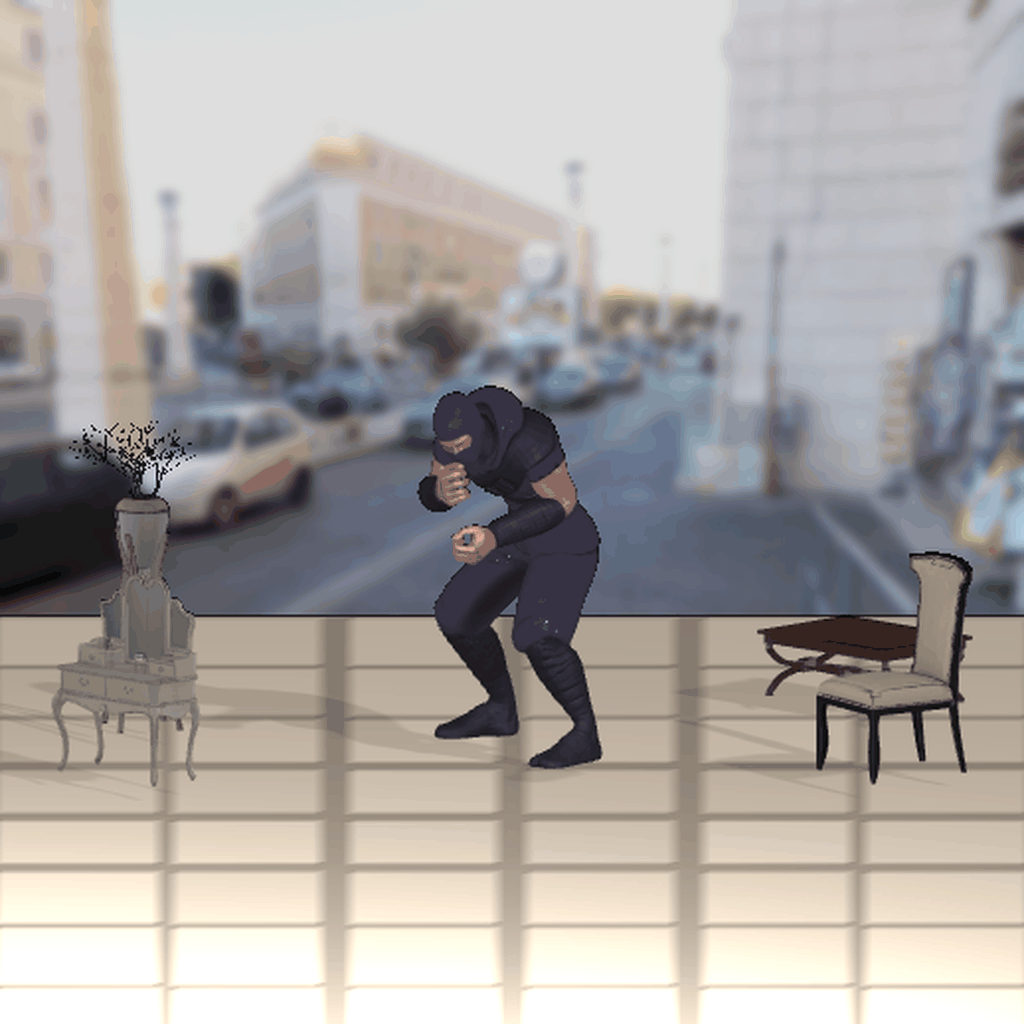}} &
        \raisebox{-0.5\height}{\includegraphics[width=0.23\columnwidth, trim=0 0 0 0, clip]{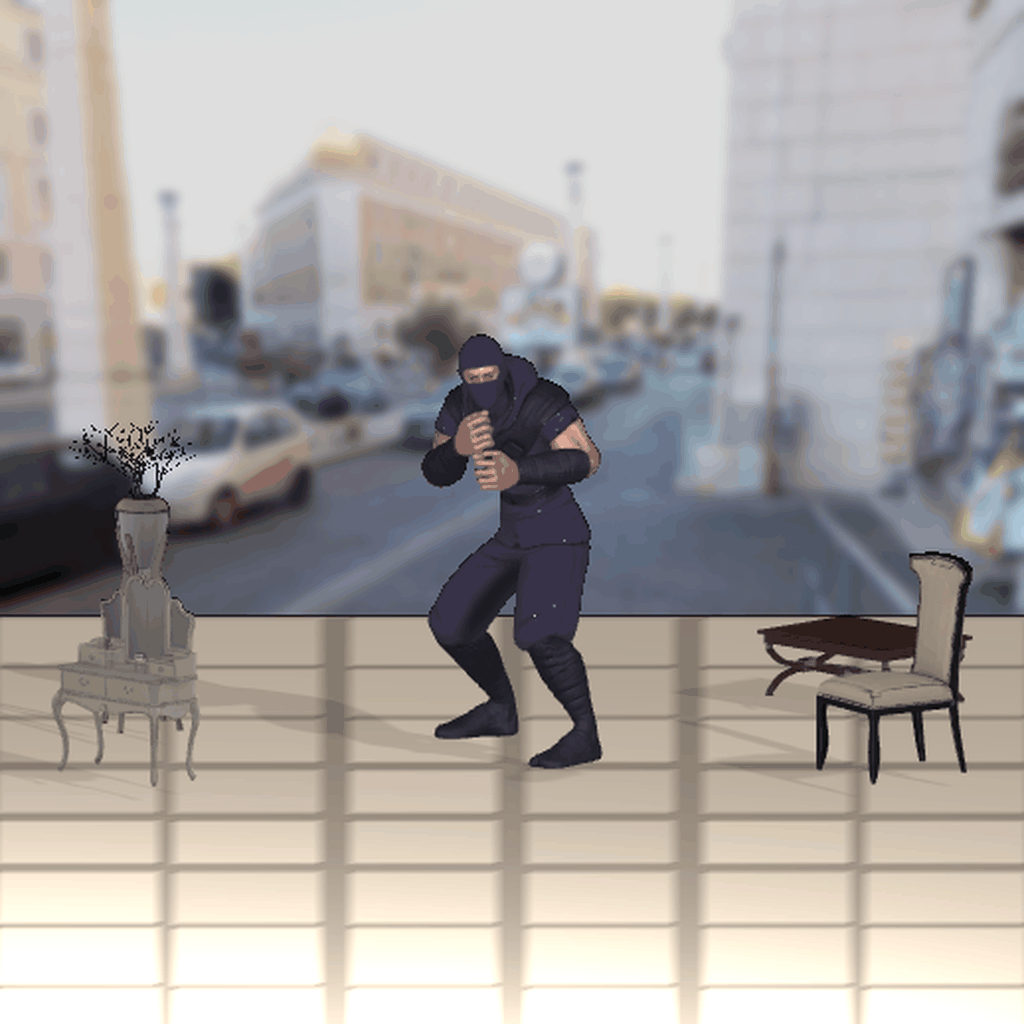}} &
        \raisebox{-0.5\height}{\includegraphics[width=0.23\columnwidth, trim=0 0 0 0, clip]{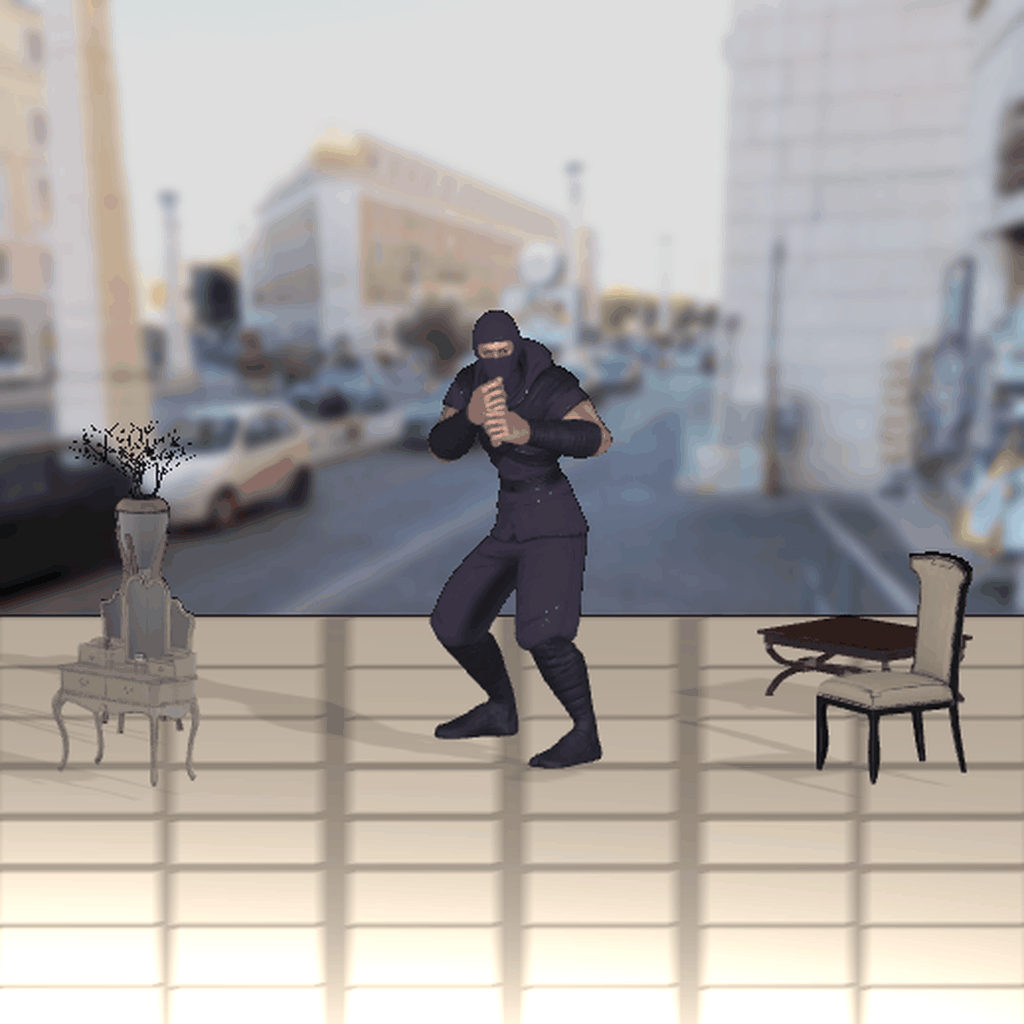}} \\[-0.5pt]
        \raisebox{-0.5\height}{\rotatebox{90}{\tiny COM4D}} &
        \raisebox{-0.5\height}{\includegraphics[width=0.23\columnwidth, trim=0 0 0 0, clip]{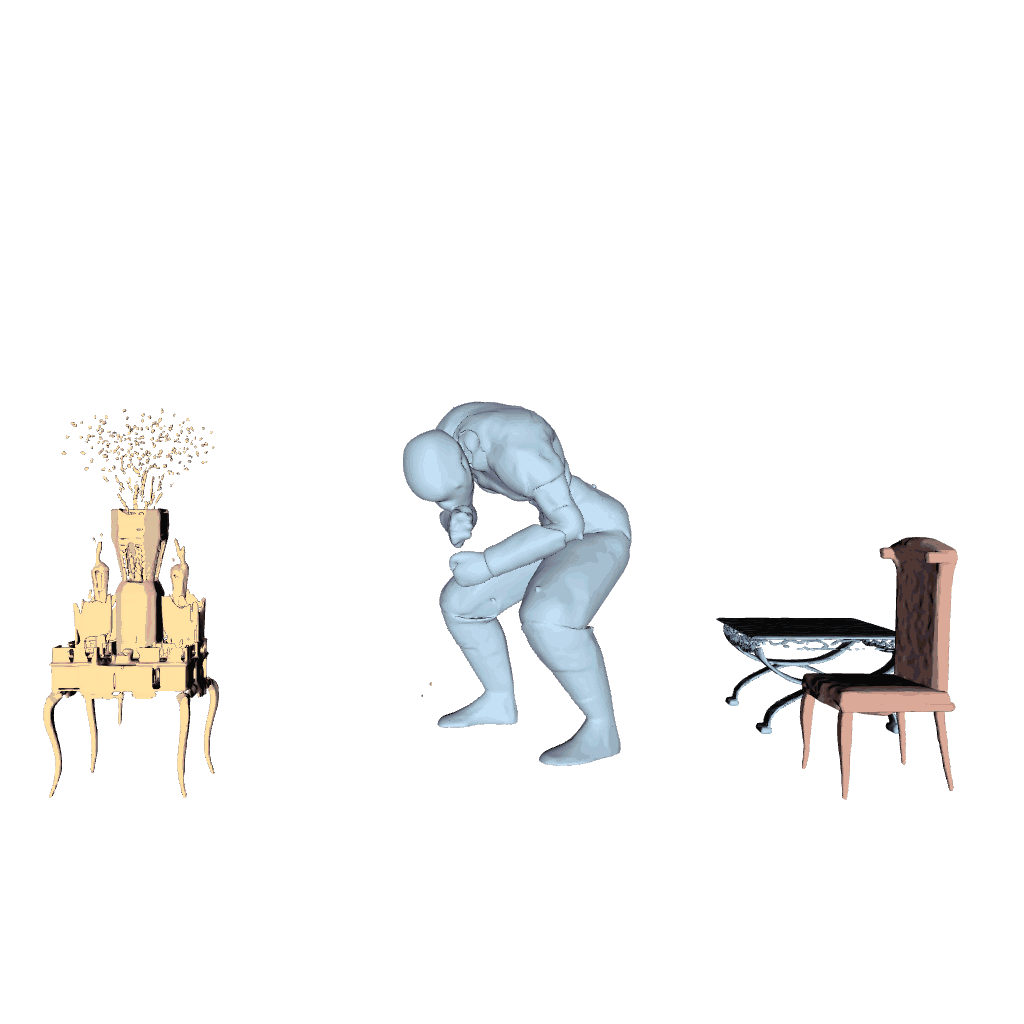}} &
        \raisebox{-0.5\height}{\includegraphics[width=0.23\columnwidth, trim=0 0 0 0, clip]{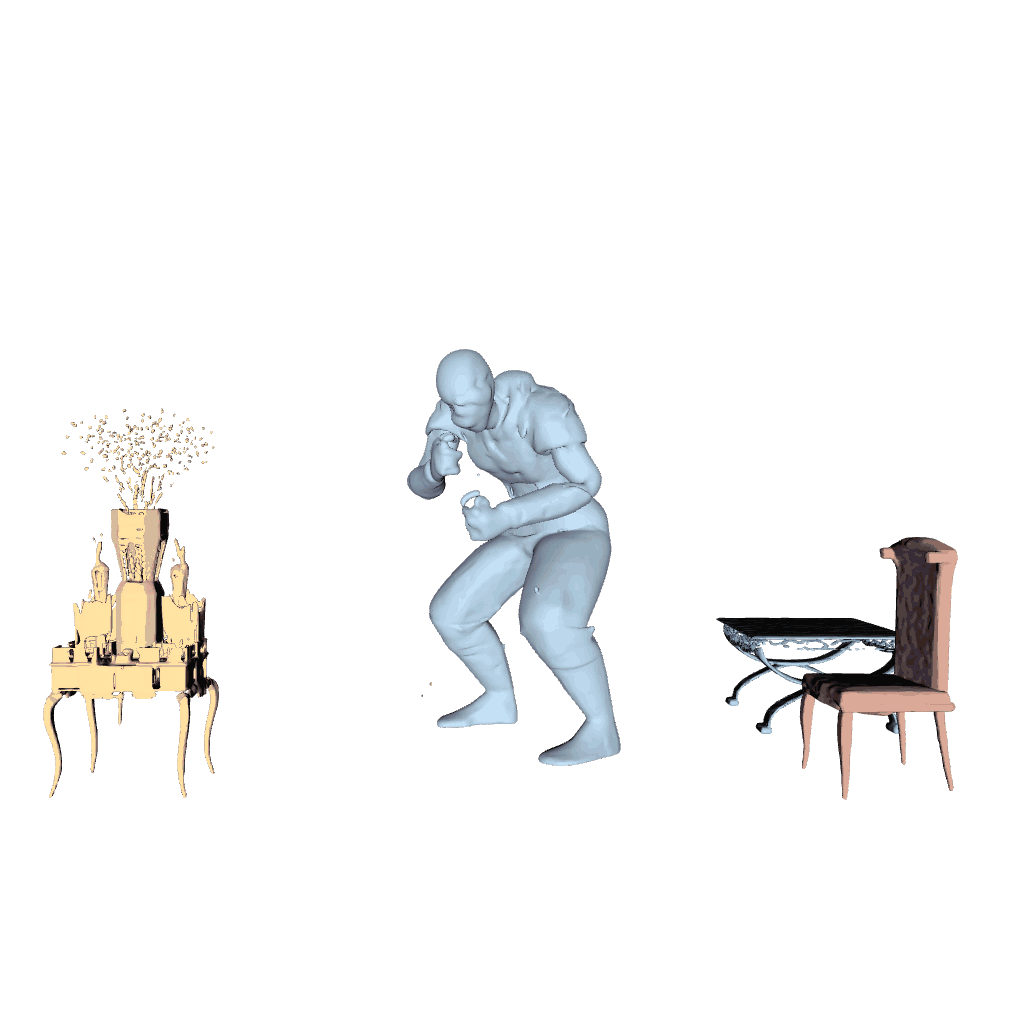}} &
        \raisebox{-0.5\height}{\includegraphics[width=0.23\columnwidth, trim=0 0 0 0, clip]{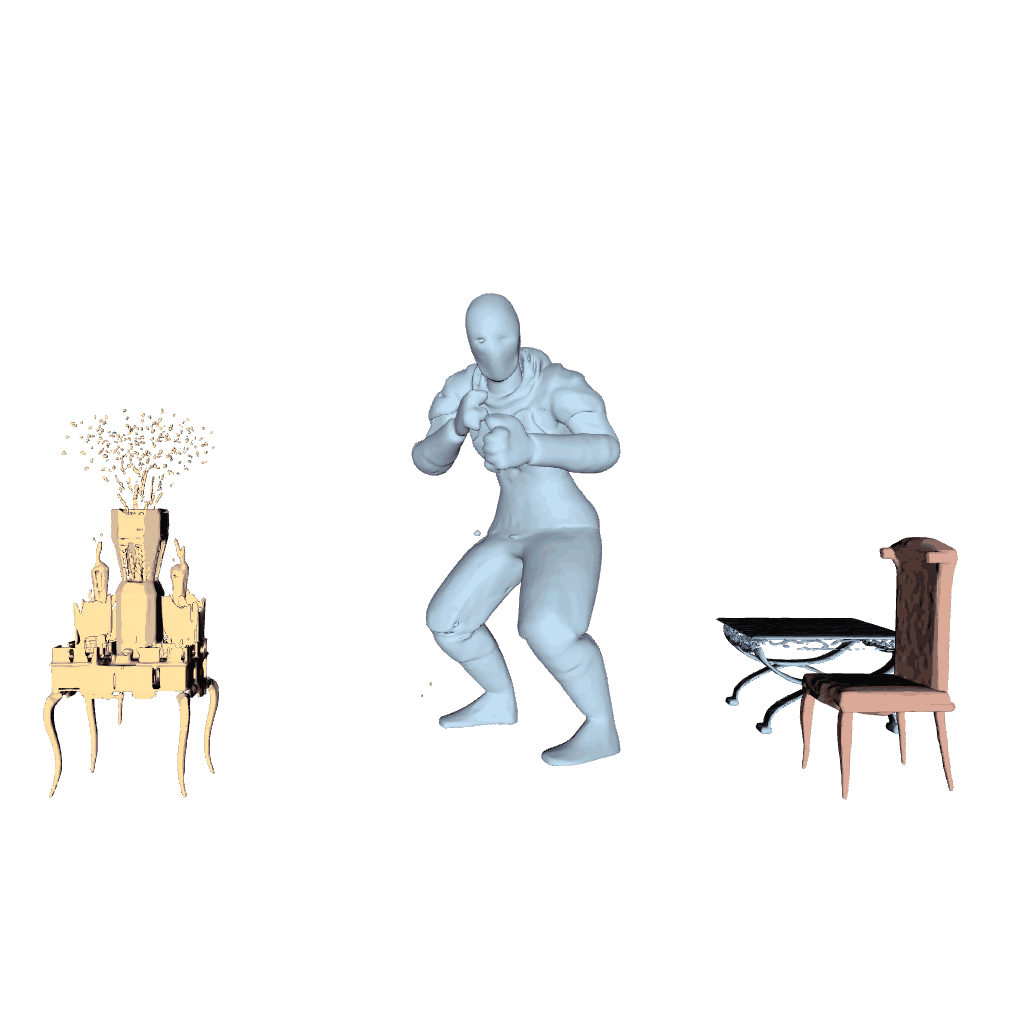}} &
        \raisebox{-0.5\height}{\includegraphics[width=0.23\columnwidth, trim=0 0 0 0, clip]{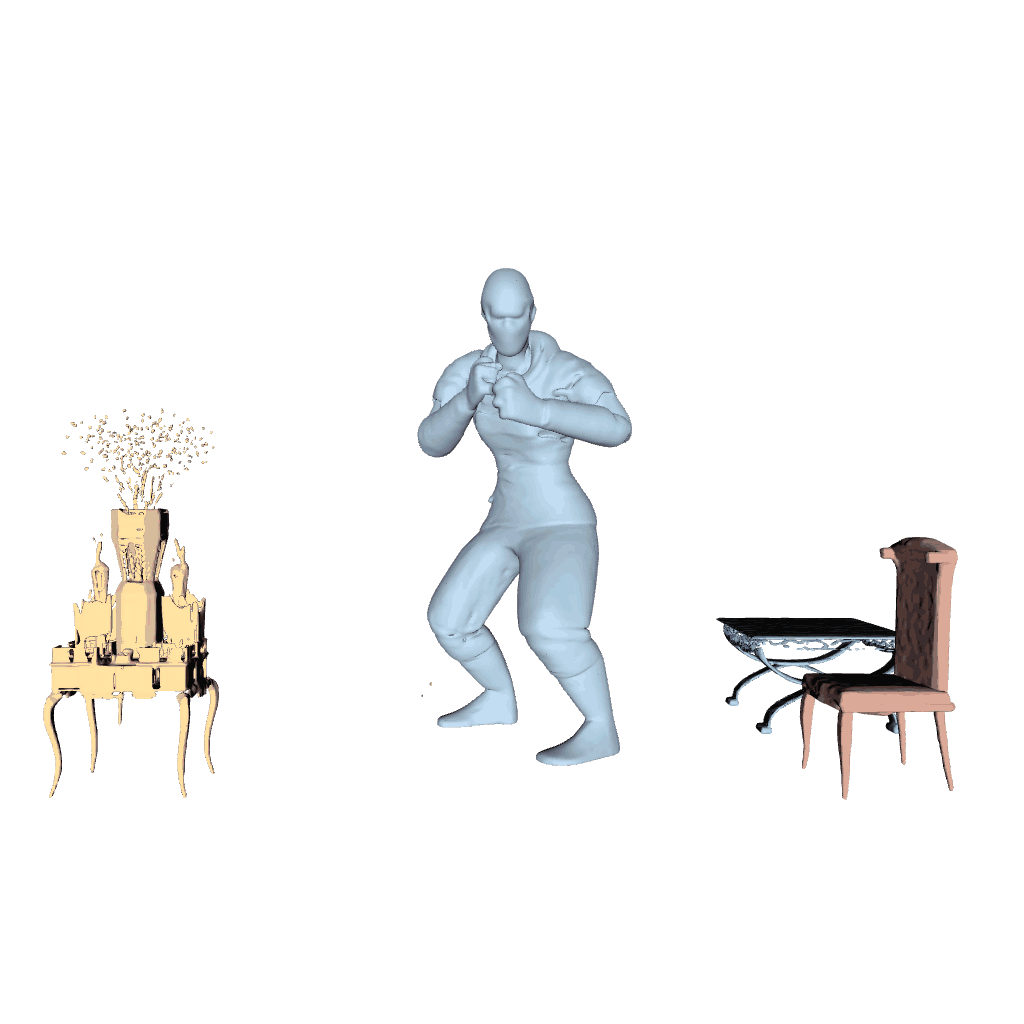}} \\[4pt]

        % ==================== SAMPLE 5: pumpkinhulk_GangnamStyle_20260324_014016 ====================
        \raisebox{-0.5\height}{\rotatebox{90}{\tiny Input}} &
        \raisebox{-0.5\height}{\includegraphics[width=0.23\columnwidth, trim=0 0 0 0, clip]{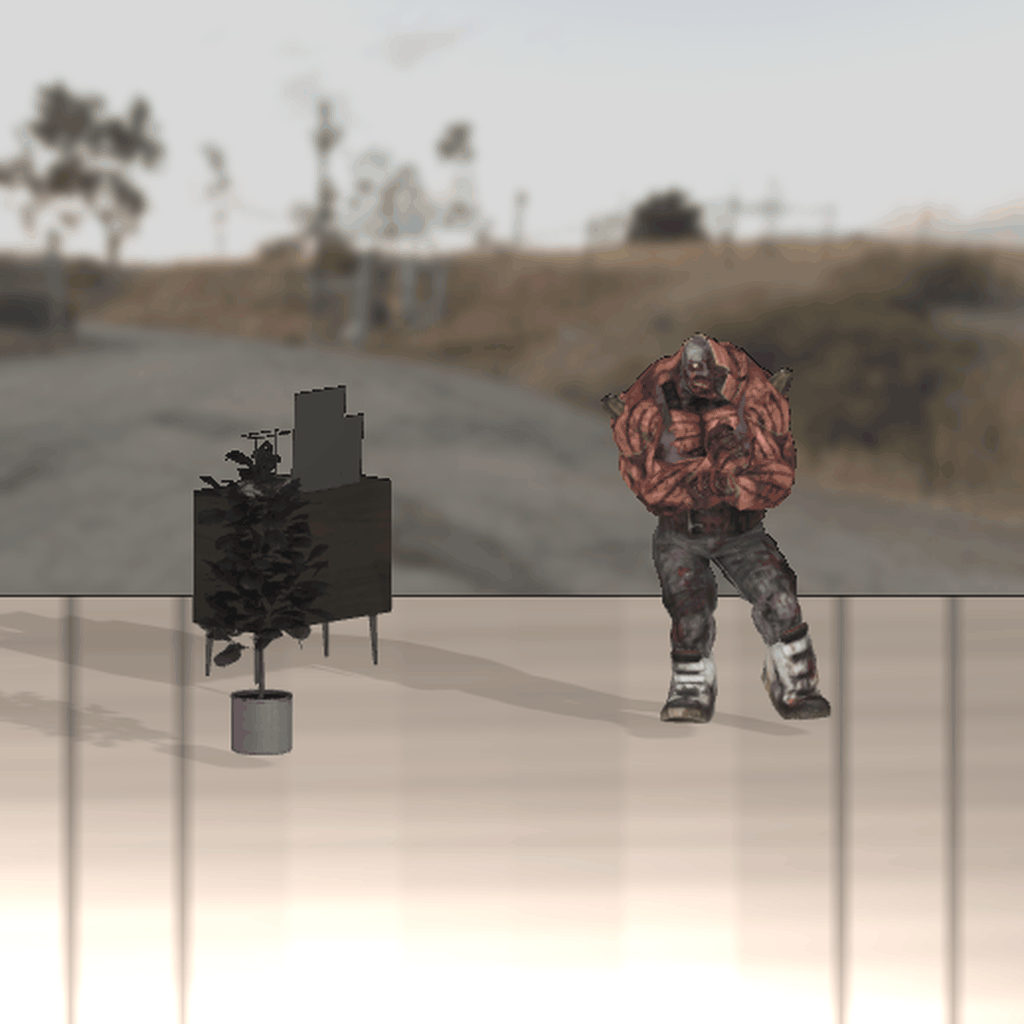}} &
        \raisebox{-0.5\height}{\includegraphics[width=0.23\columnwidth, trim=0 0 0 0, clip]{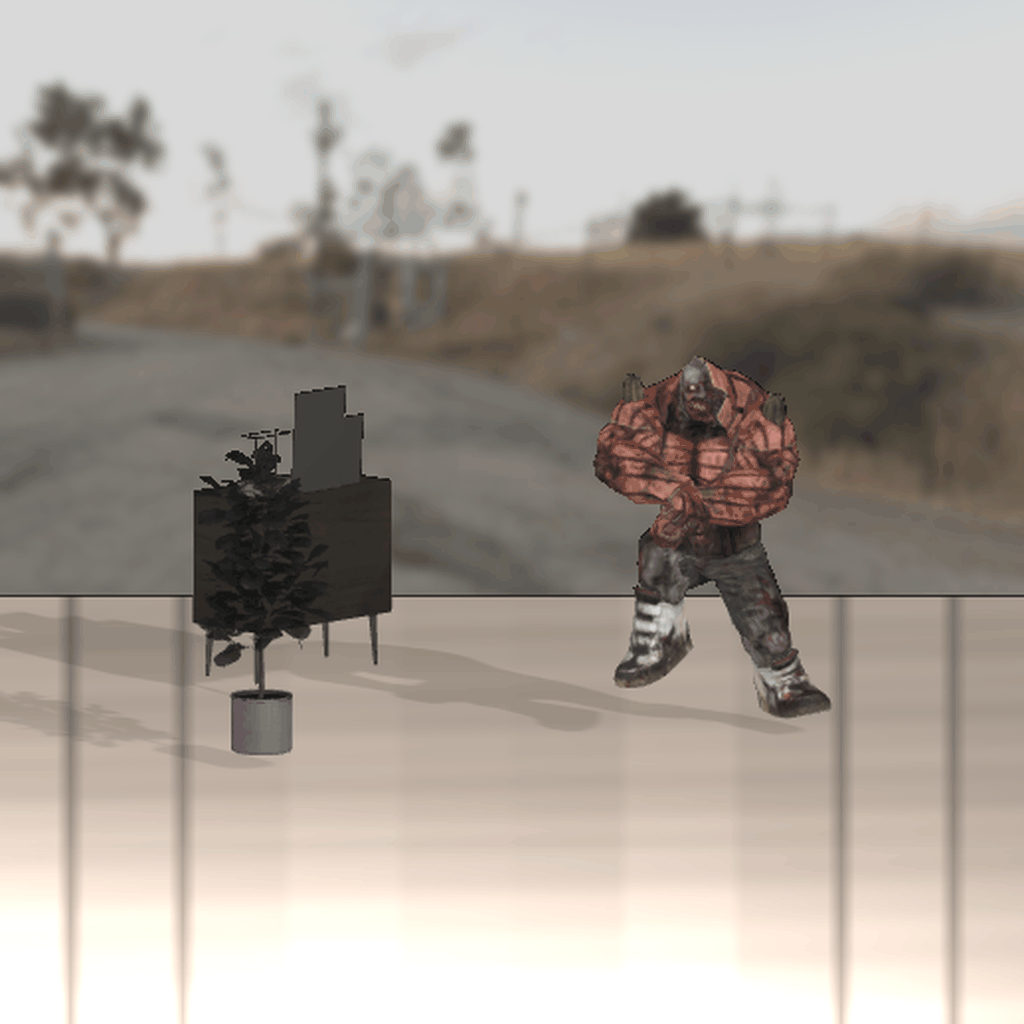}} &
        \raisebox{-0.5\height}{\includegraphics[width=0.23\columnwidth, trim=0 0 0 0, clip]{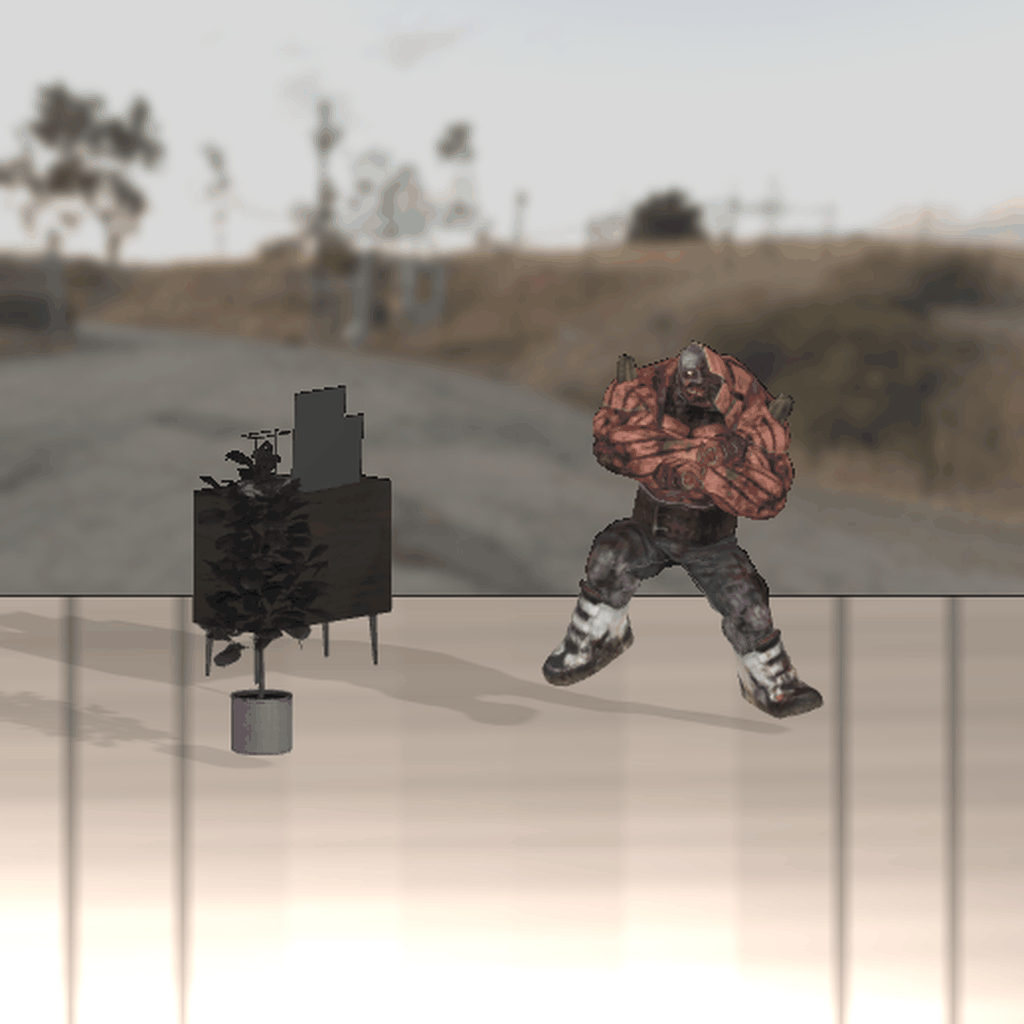}} &
        \raisebox{-0.5\height}{\includegraphics[width=0.23\columnwidth, trim=0 0 0 0, clip]{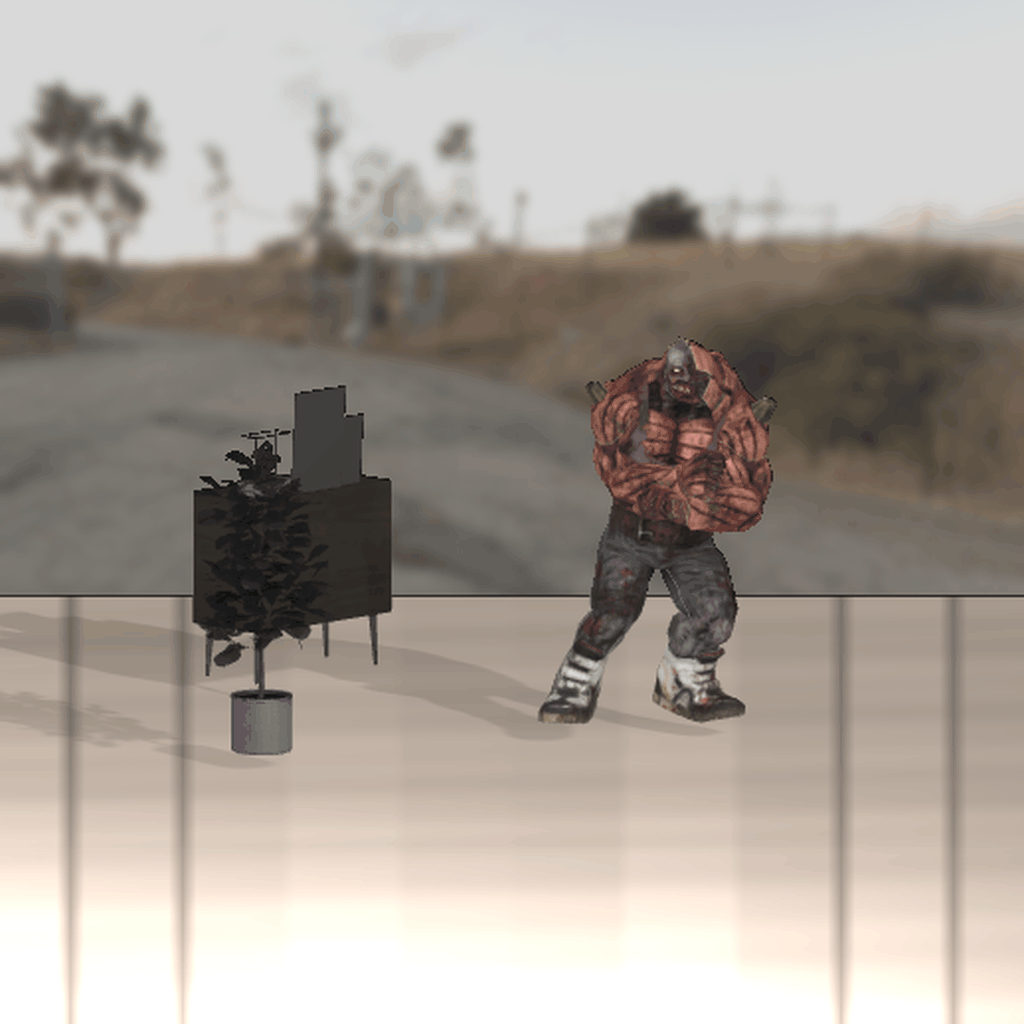}} \\[-0.5pt]
        \raisebox{-0.5\height}{\rotatebox{90}{\tiny COM4D}} &
        \raisebox{-0.5\height}{\includegraphics[width=0.23\columnwidth, trim=0 0 0 0, clip]{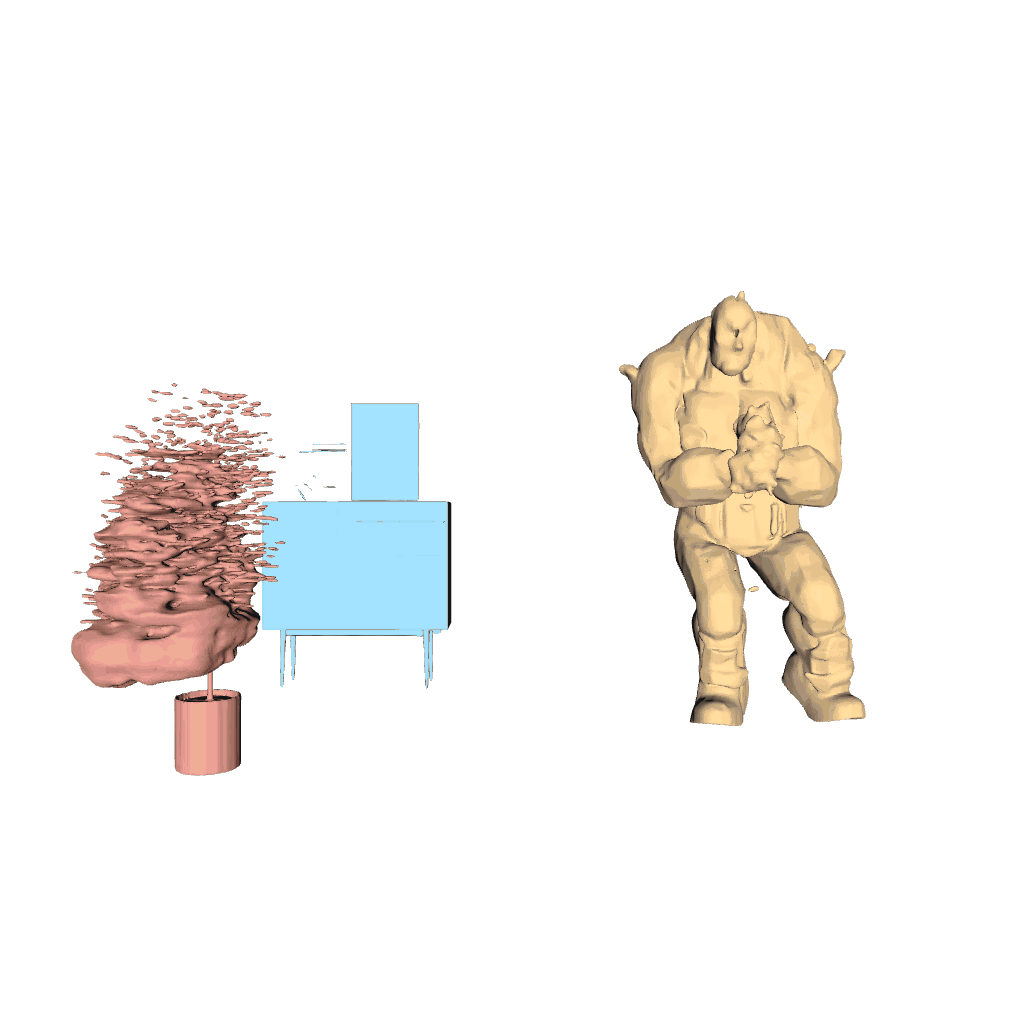}} &
        \raisebox{-0.5\height}{\includegraphics[width=0.23\columnwidth, trim=0 0 0 0, clip]{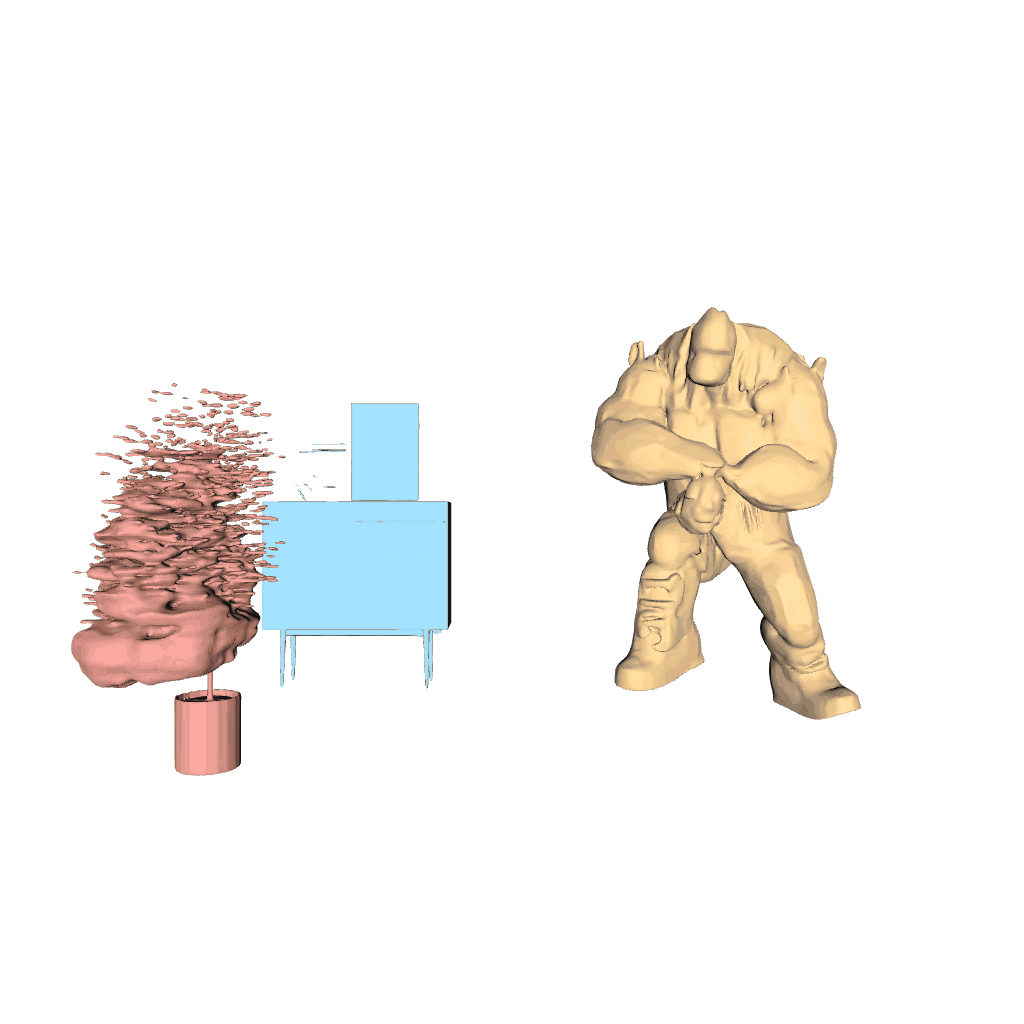}} &
        \raisebox{-0.5\height}{\includegraphics[width=0.23\columnwidth, trim=0 0 0 0, clip]{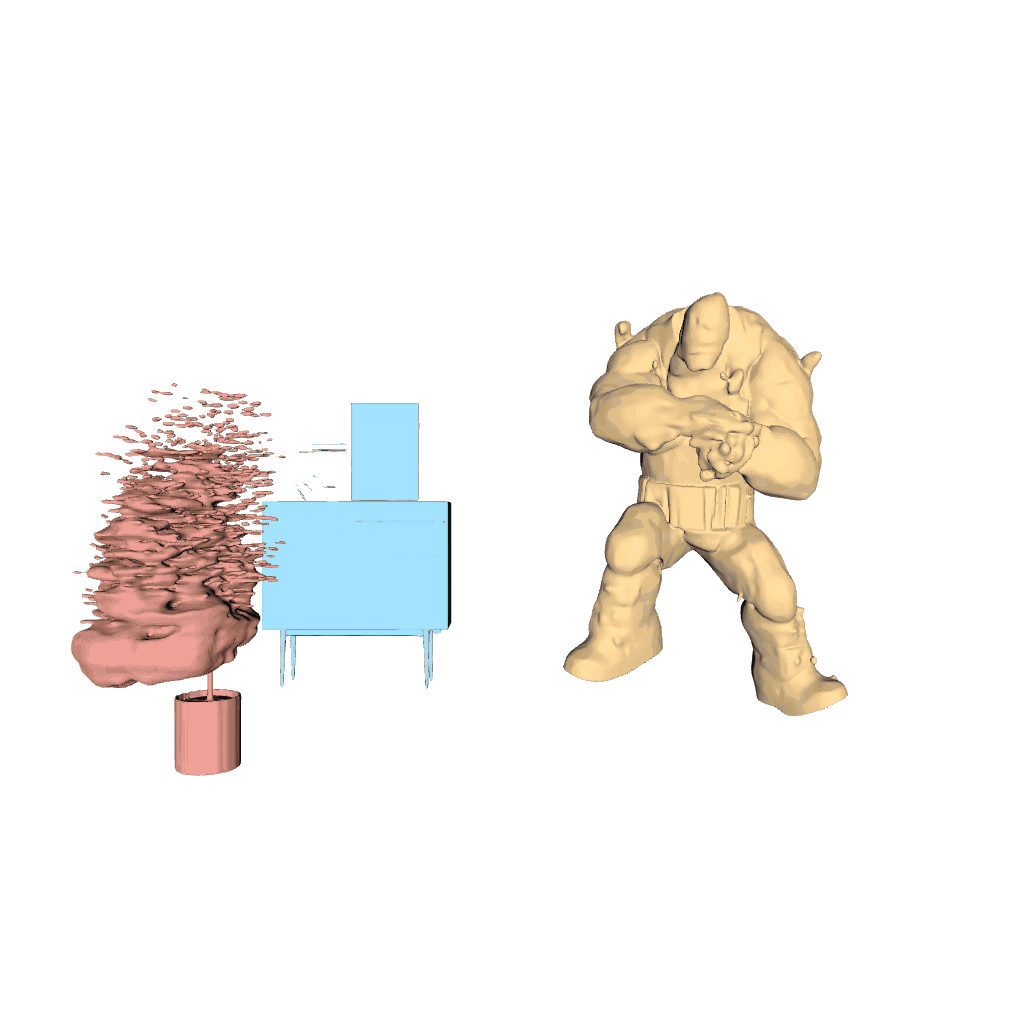}} &
        \raisebox{-0.5\height}{\includegraphics[width=0.23\columnwidth, trim=0 0 0 0, clip]{figures/appendix/synthetic/pumpkinhulk_GangnamStyle_20260324_014016/out/7.jpg}} \\[4pt]

  \end{tabular}

  \vspace{-3pt}
  \caption{
  Additional qualitative results on the synthetic compositional 4D dataset.
  }
  \label{fig:synthetic_com4d_appendix_qual}
\end{figure}

\begin{figure}[h]
  
  \centering
  \setlength{\tabcolsep}{0pt}
  \renewcommand{\arraystretch}{0}

  \begin{tabular}{@{}c@{\hspace{2pt}}cccc@{}}

% ==================== SAMPLE 1: pumpkinhulk_NorthernSoulSpinCombo_20260324_014245 ====================
\raisebox{-0.5\height}{\rotatebox{90}{\tiny Input}} &
\raisebox{-0.5\height}{\includegraphics[width=0.23\columnwidth, trim=0 0 0 0, clip]{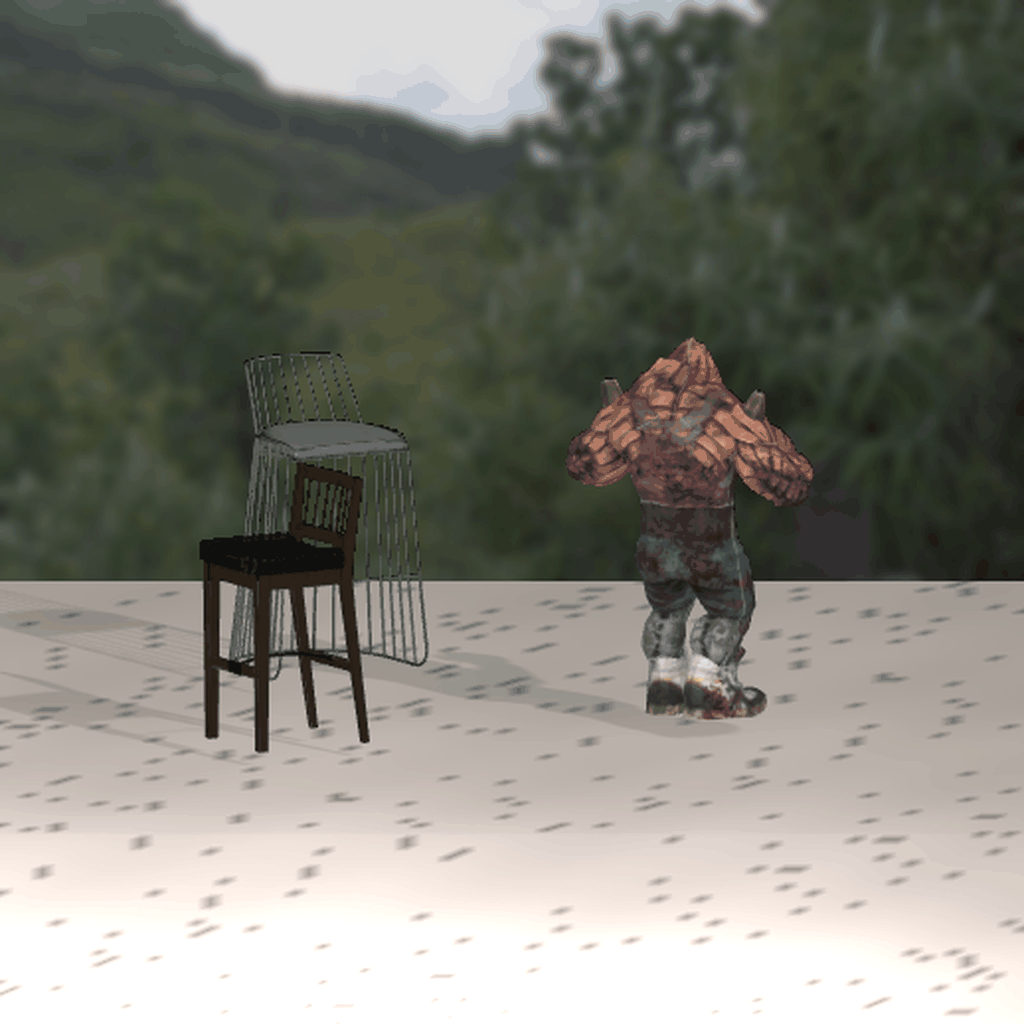}} &
\raisebox{-0.5\height}{\includegraphics[width=0.23\columnwidth, trim=0 0 0 0, clip]{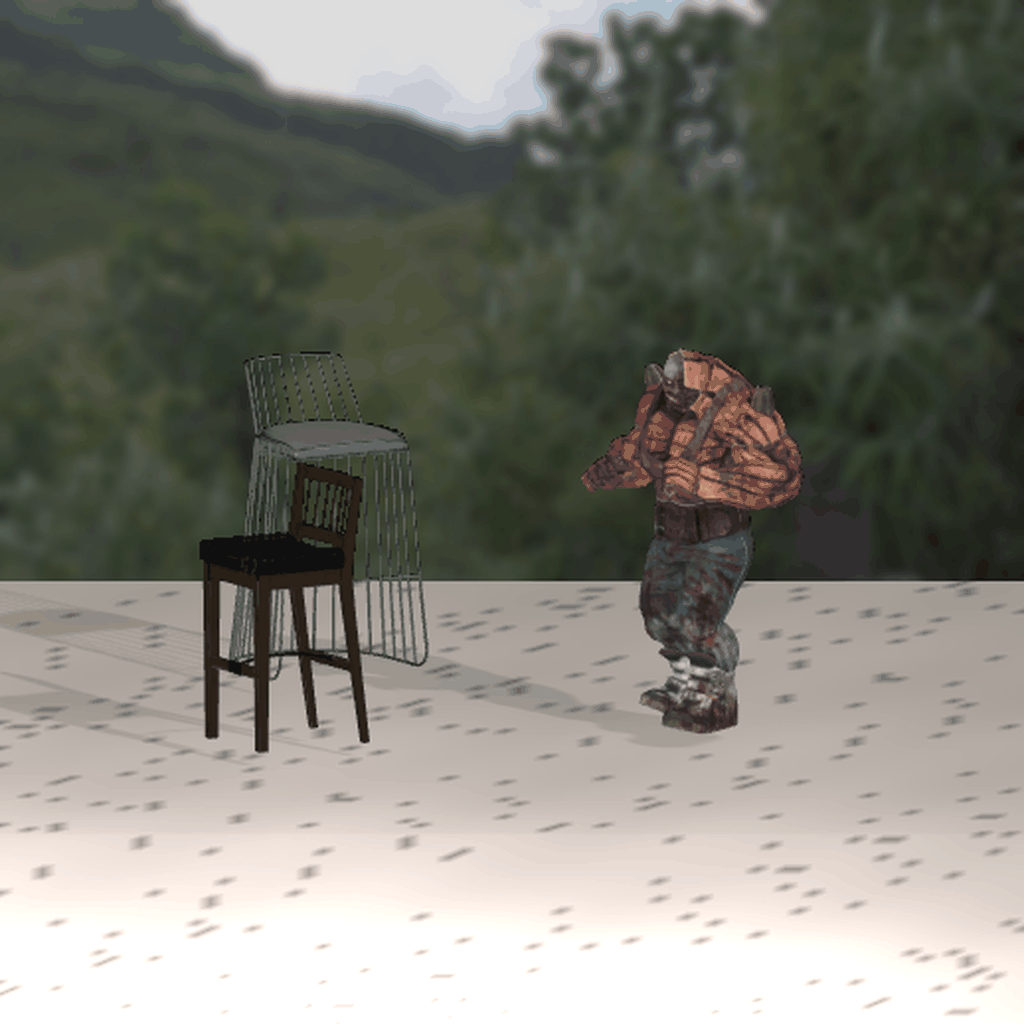}} &
\raisebox{-0.5\height}{\includegraphics[width=0.23\columnwidth, trim=0 0 0 0, clip]{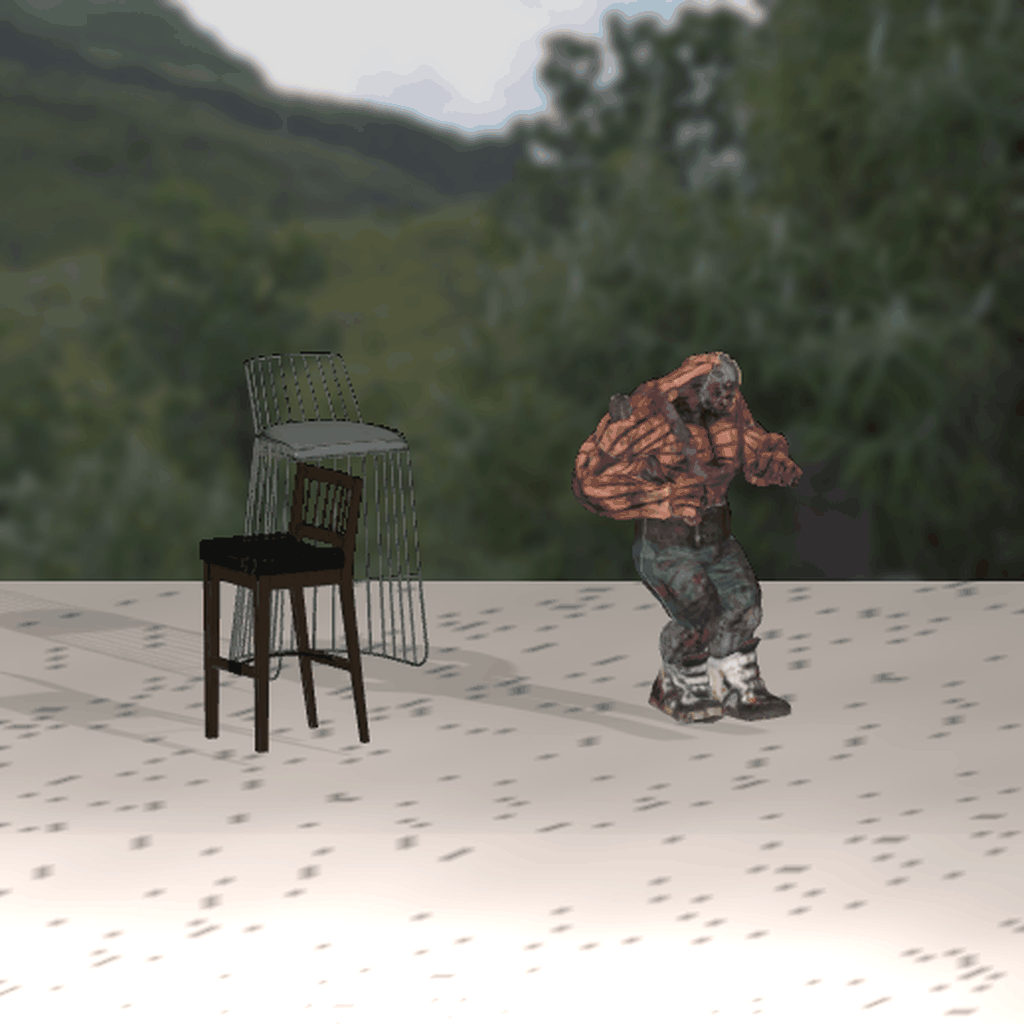}} &
\raisebox{-0.5\height}{\includegraphics[width=0.23\columnwidth, trim=0 0 0 0, clip]{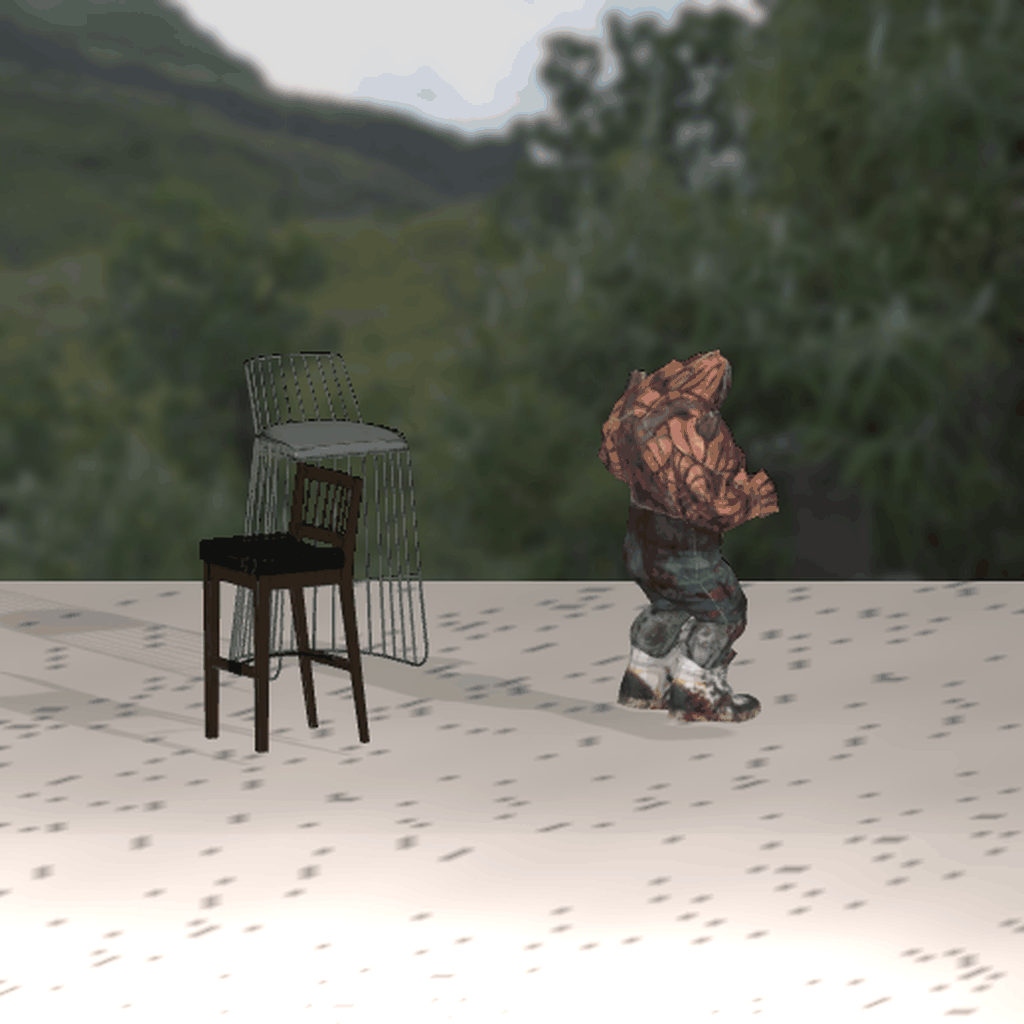}} \\[-0.5pt]
\raisebox{-0.5\height}{\rotatebox{90}{\tiny COM4D}} &
\raisebox{-0.5\height}{\includegraphics[width=0.23\columnwidth, trim=0 0 0 0, clip]{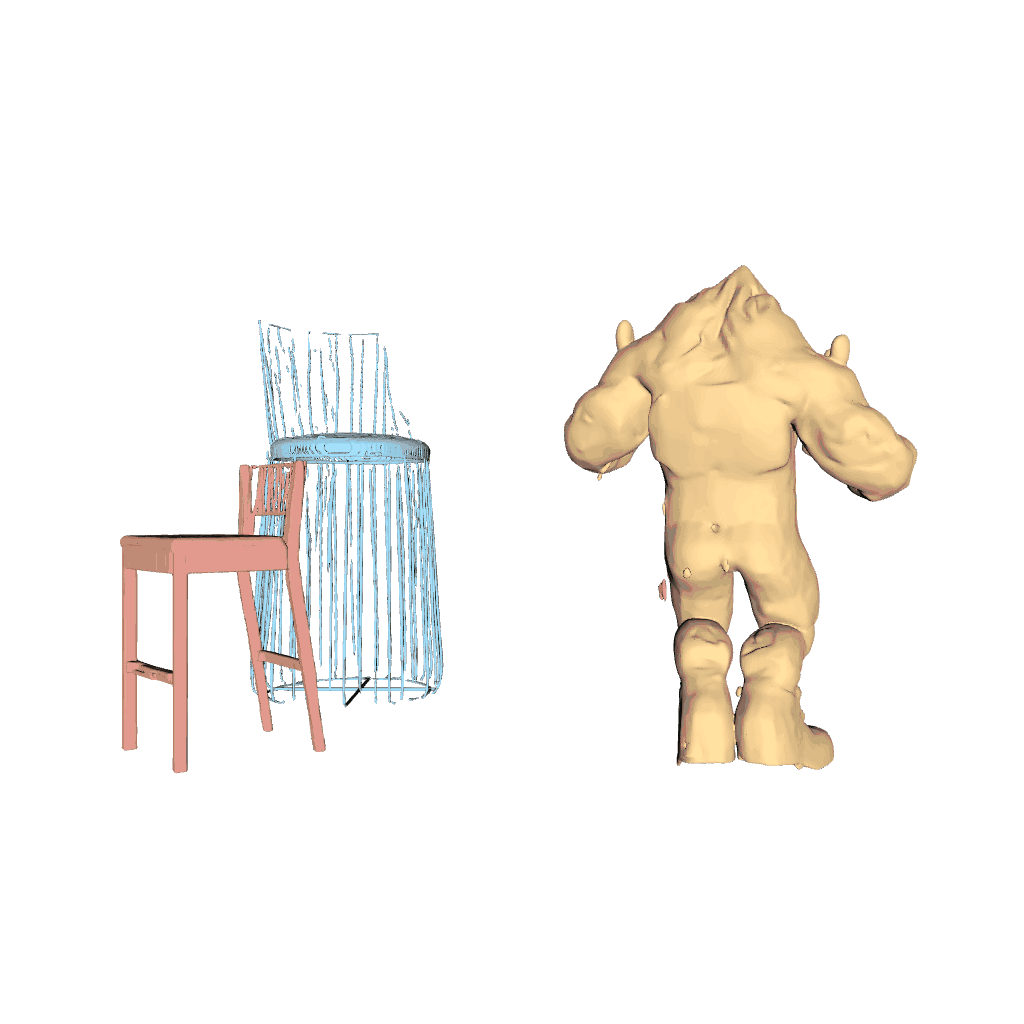}} &
\raisebox{-0.5\height}{\includegraphics[width=0.23\columnwidth, trim=0 0 0 0, clip]{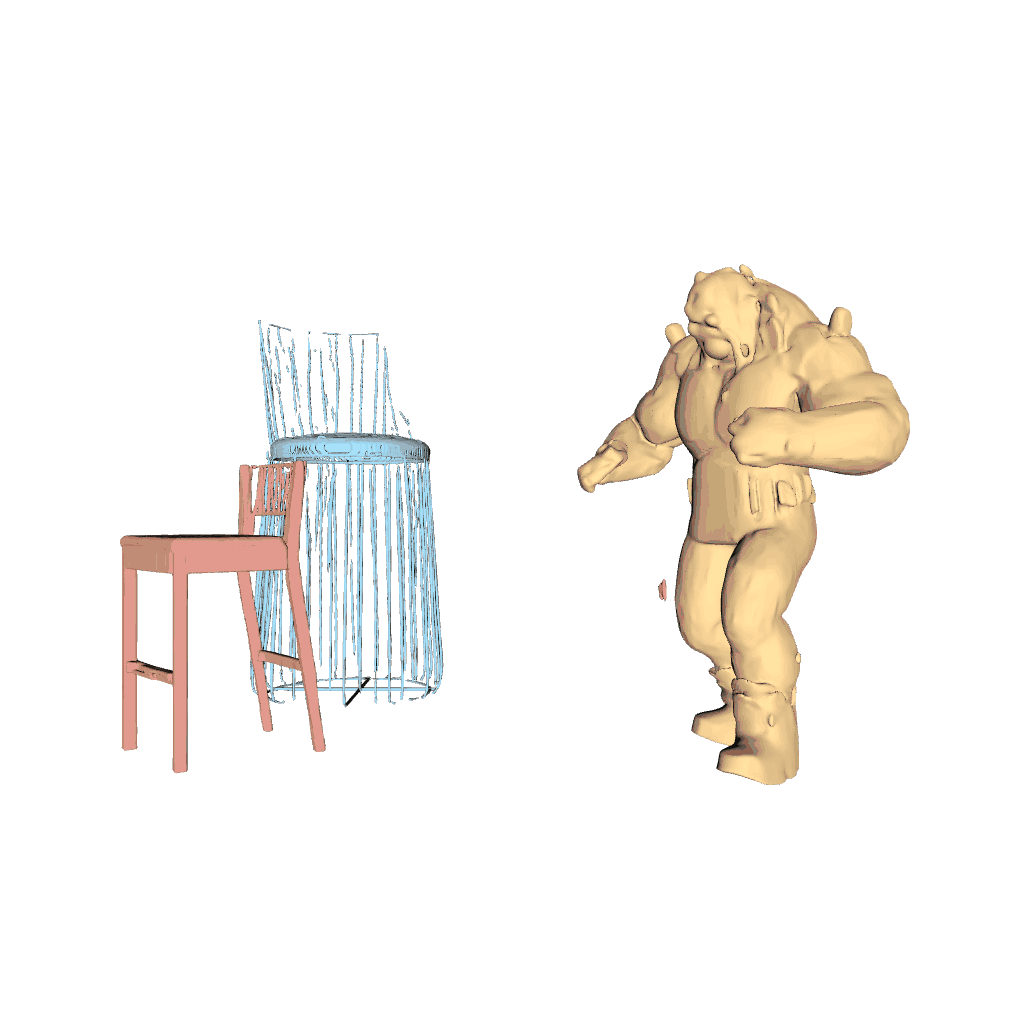}} &
\raisebox{-0.5\height}{\includegraphics[width=0.23\columnwidth, trim=0 0 0 0, clip]{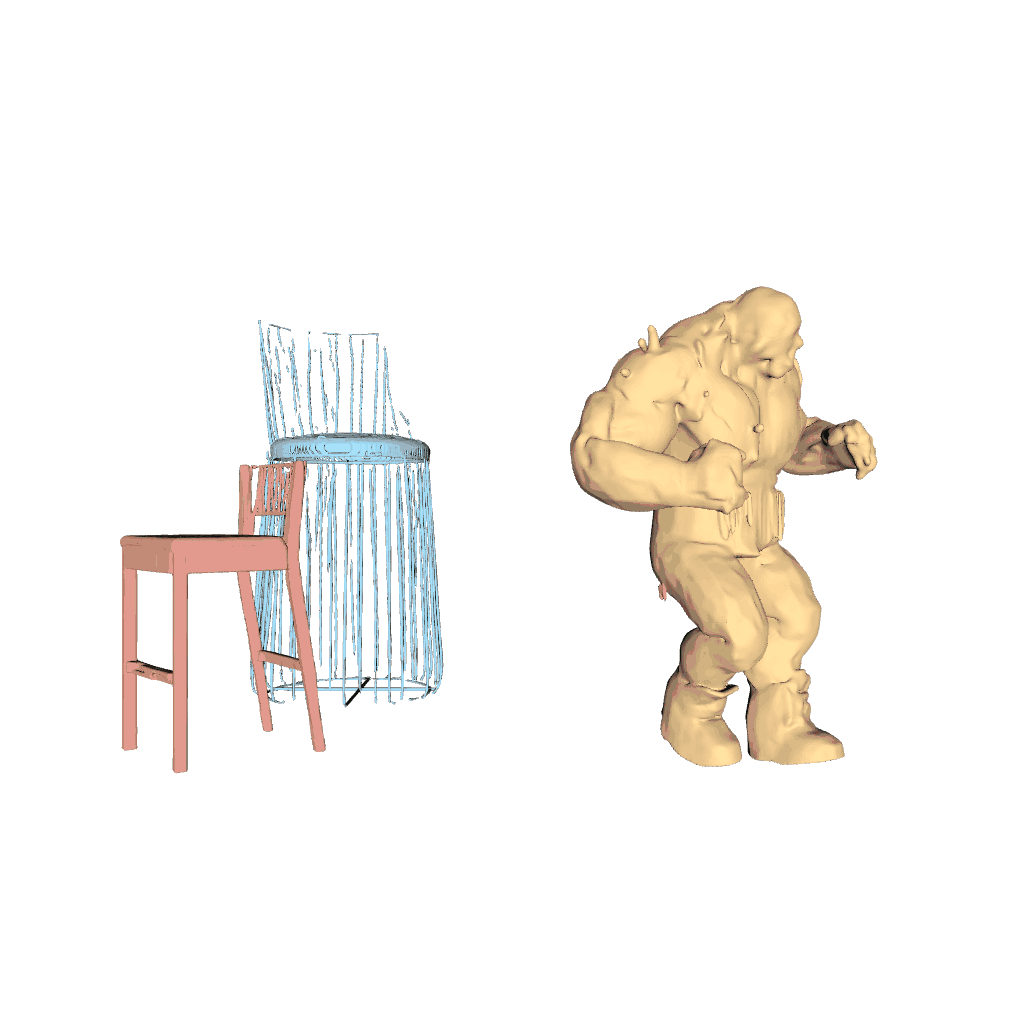}} &
\raisebox{-0.5\height}{\includegraphics[width=0.23\columnwidth, trim=0 0 0 0, clip]{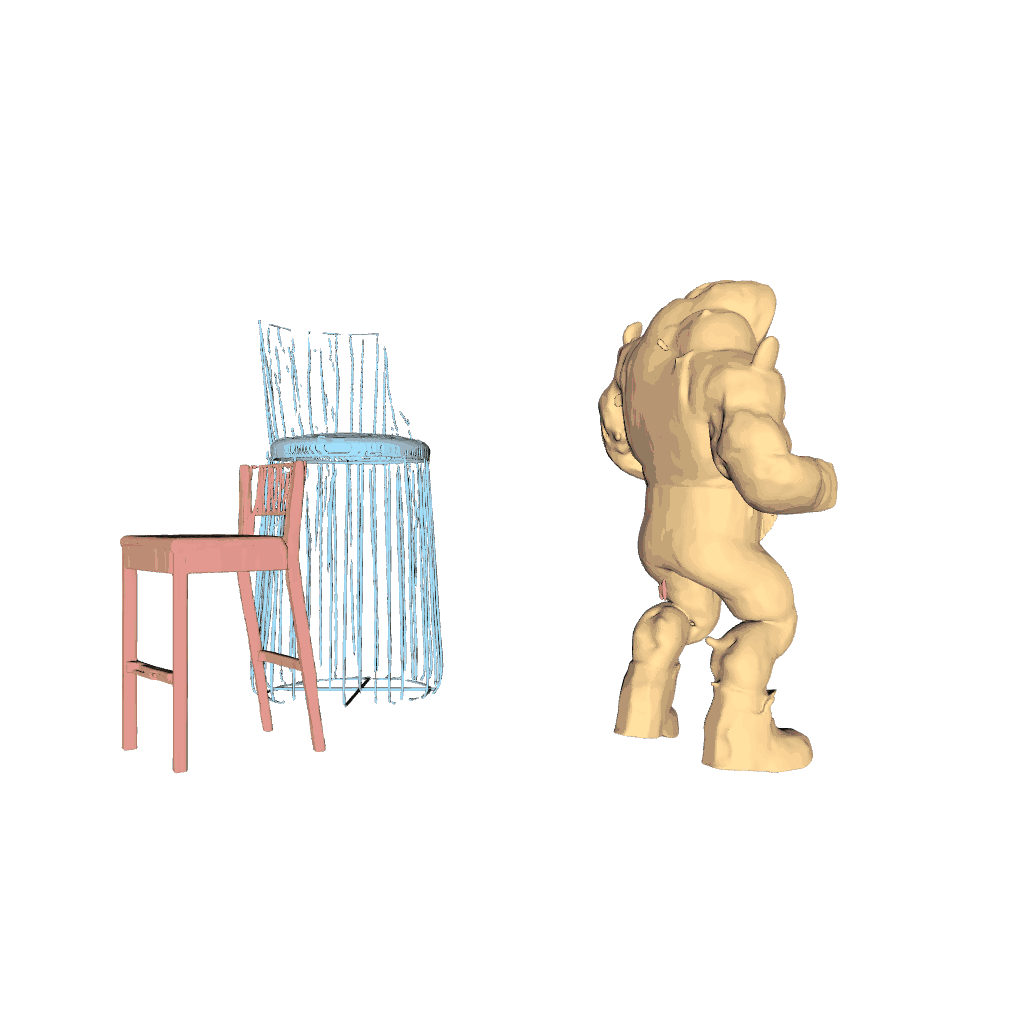}} \\[4pt]

% ==================== SAMPLE 2: mousey_Shuffling_20260324_012450 ====================
\raisebox{-0.5\height}{\rotatebox{90}{\tiny Input}} &
\raisebox{-0.5\height}{\includegraphics[width=0.23\columnwidth, trim=0 0 0 0, clip]{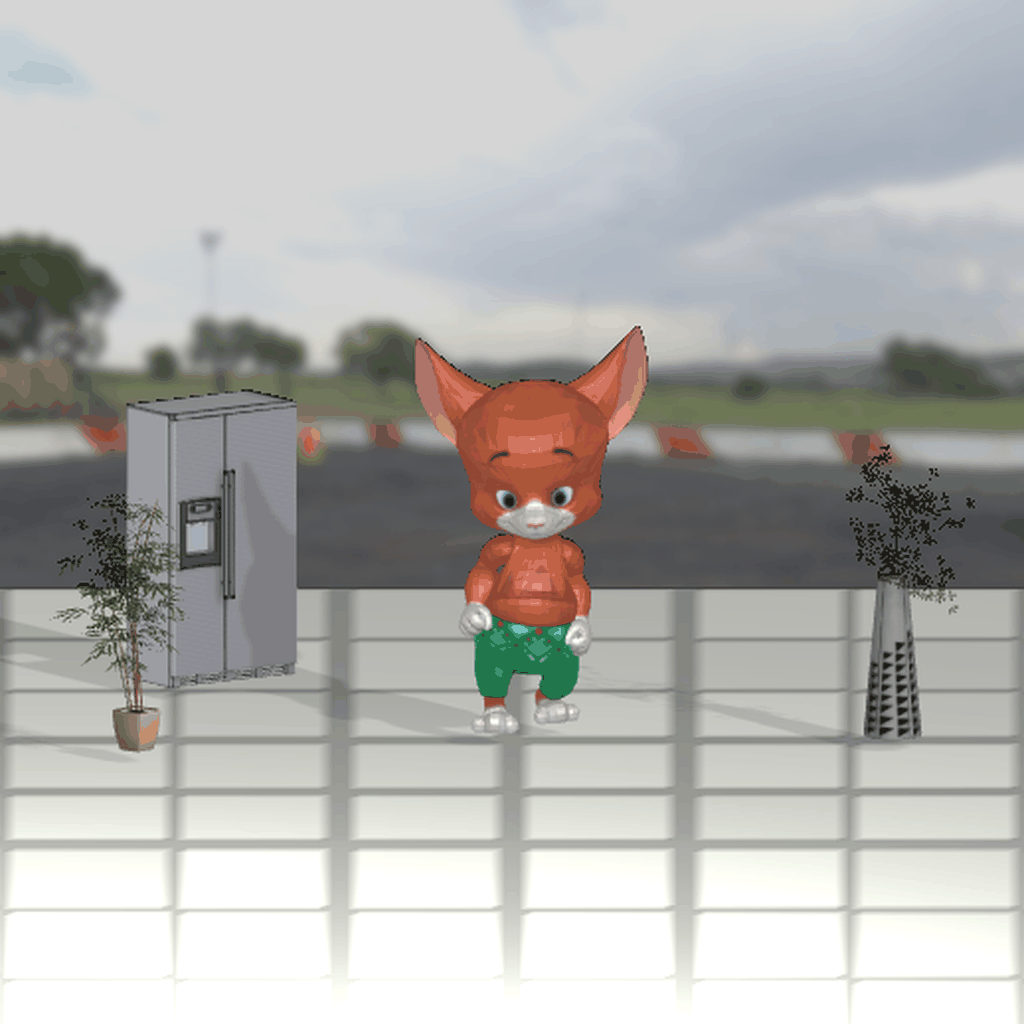}} &
\raisebox{-0.5\height}{\includegraphics[width=0.23\columnwidth, trim=0 0 0 0, clip]{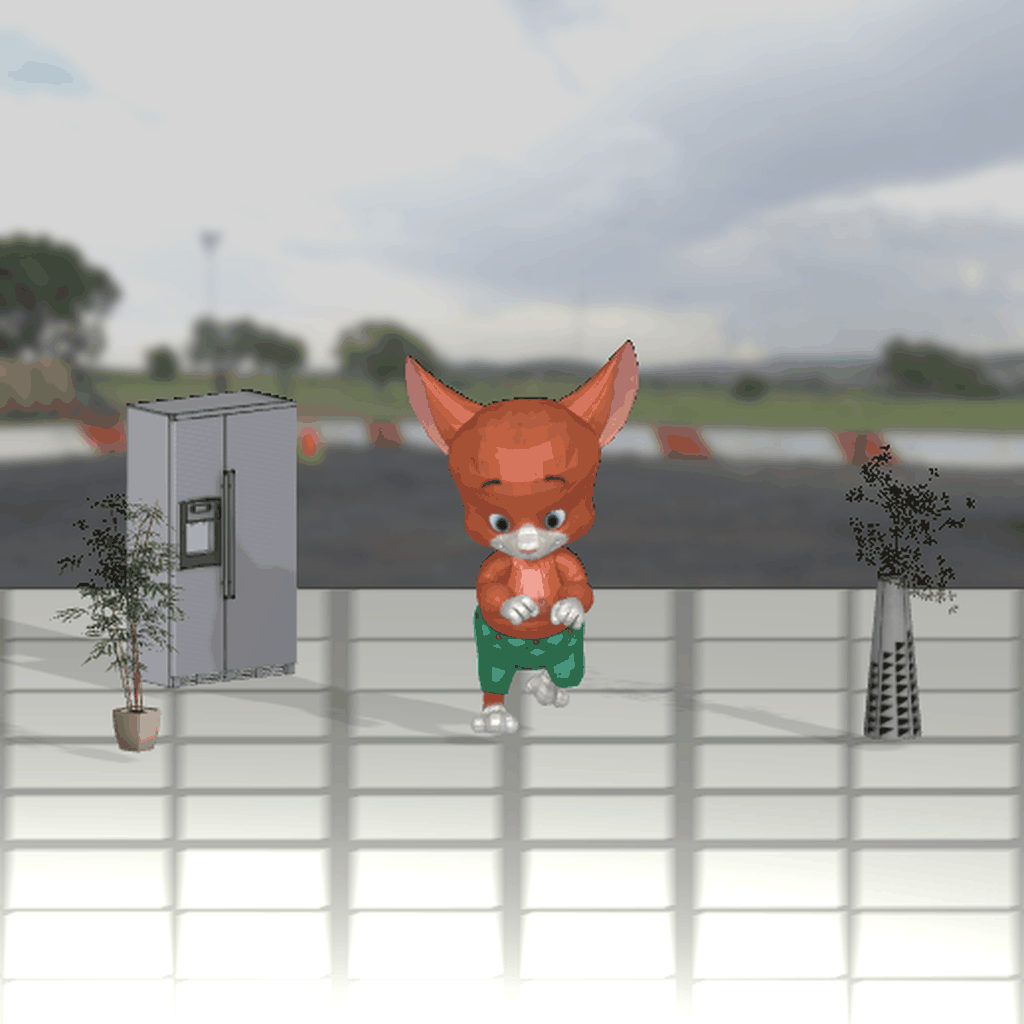}} &
\raisebox{-0.5\height}{\includegraphics[width=0.23\columnwidth, trim=0 0 0 0, clip]{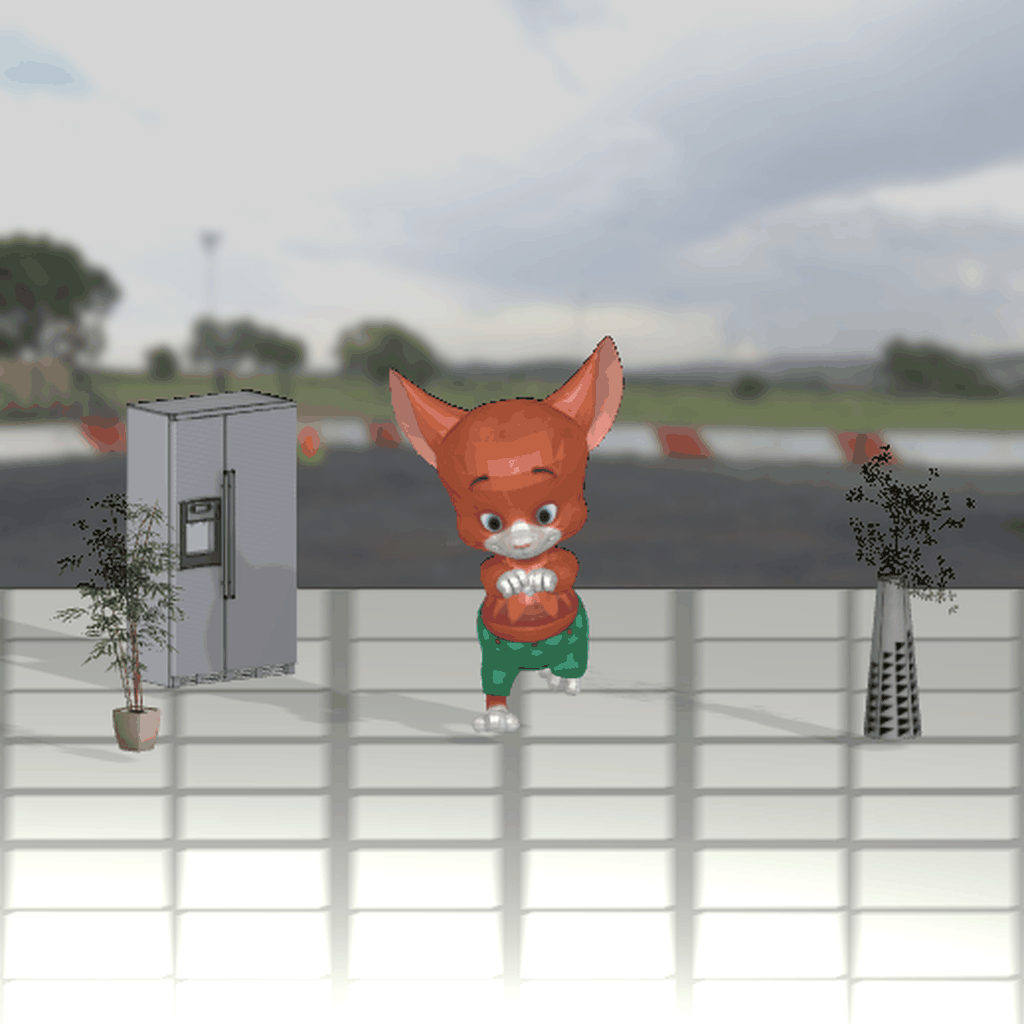}} &
\raisebox{-0.5\height}{\includegraphics[width=0.23\columnwidth, trim=0 0 0 0, clip]{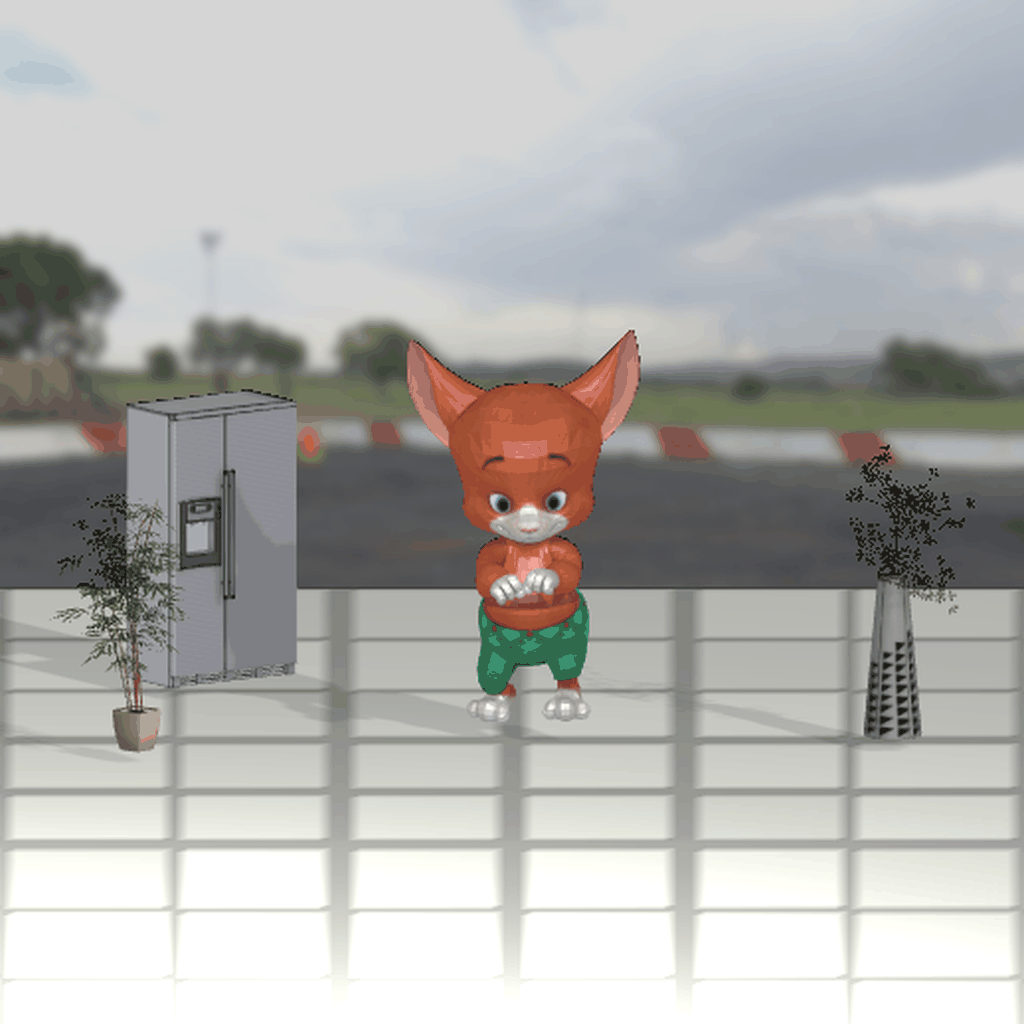}} \\[-0.5pt]
\raisebox{-0.5\height}{\rotatebox{90}{\tiny COM4D}} &
\raisebox{-0.5\height}{\includegraphics[width=0.23\columnwidth, trim=0 0 0 0, clip]{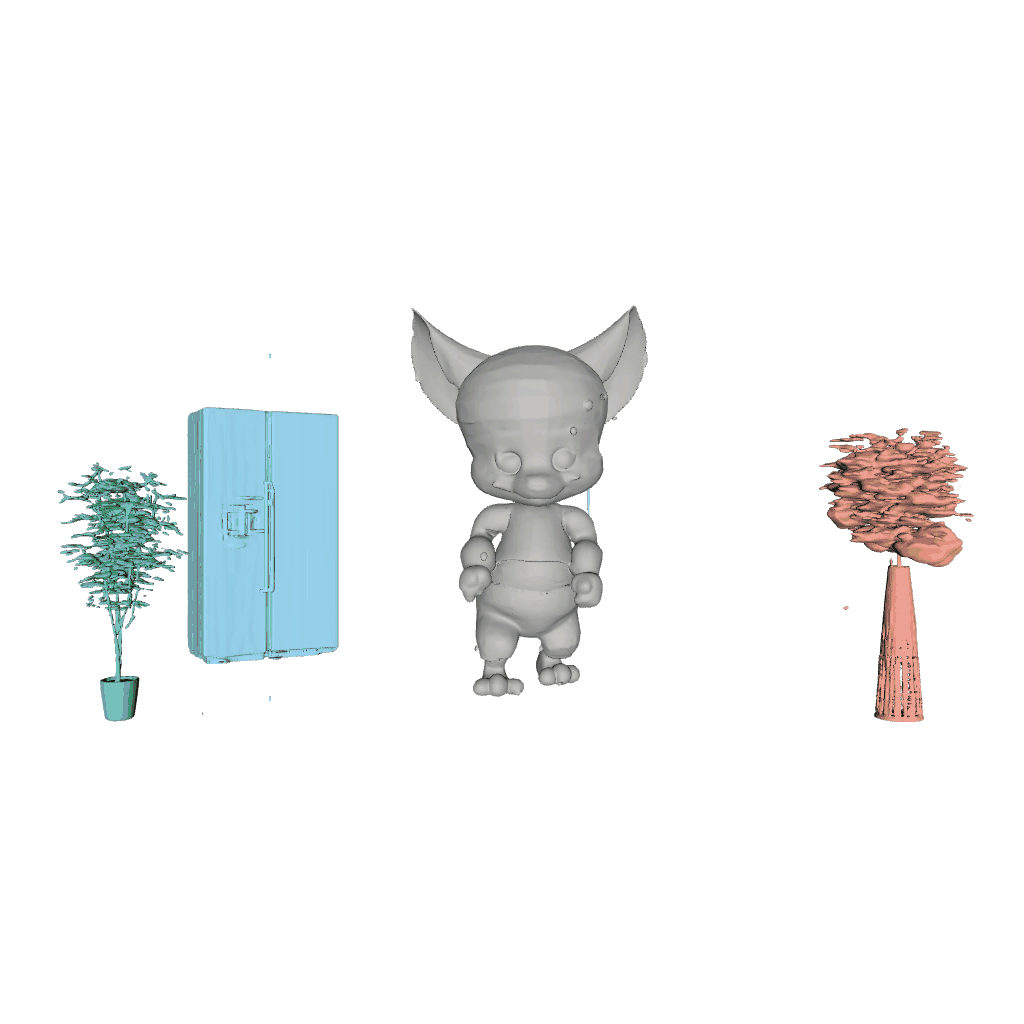}} &
\raisebox{-0.5\height}{\includegraphics[width=0.23\columnwidth, trim=0 0 0 0, clip]{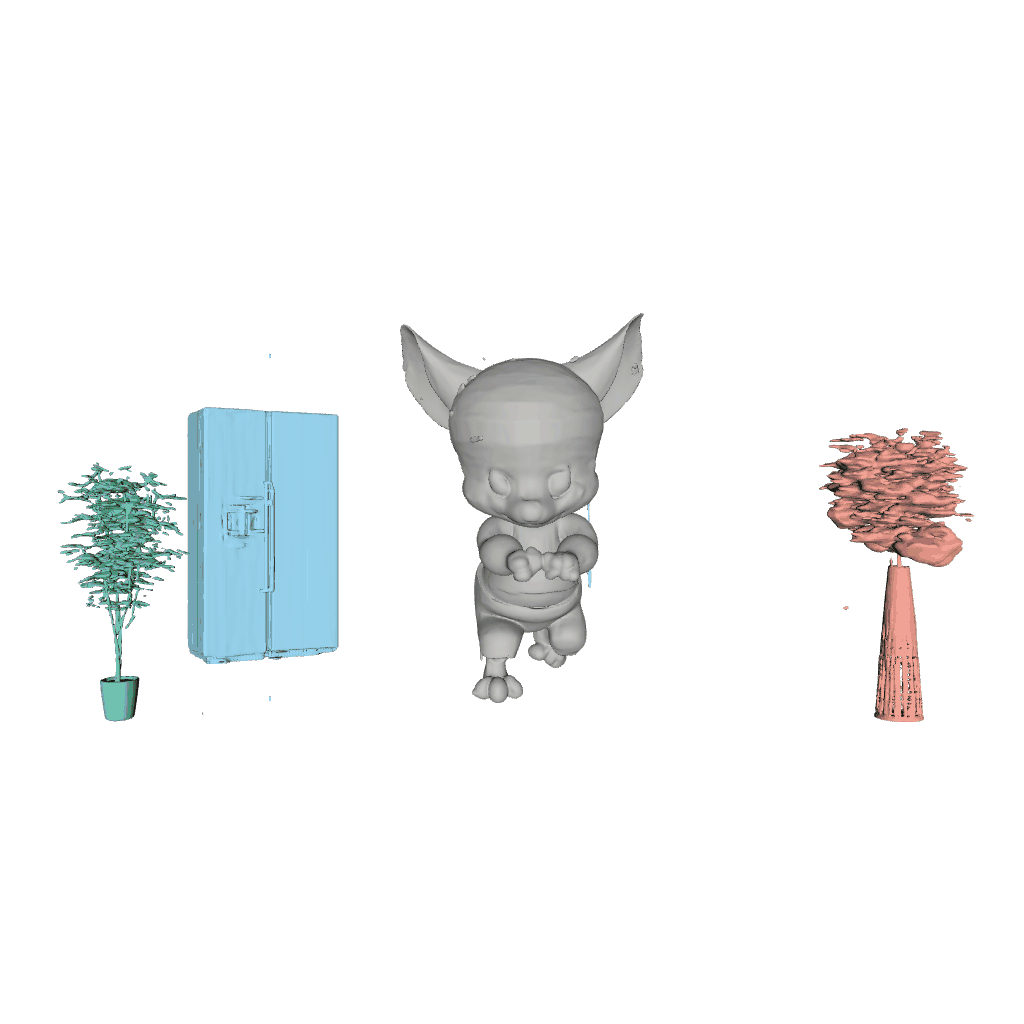}} &
\raisebox{-0.5\height}{\includegraphics[width=0.23\columnwidth, trim=0 0 0 0, clip]{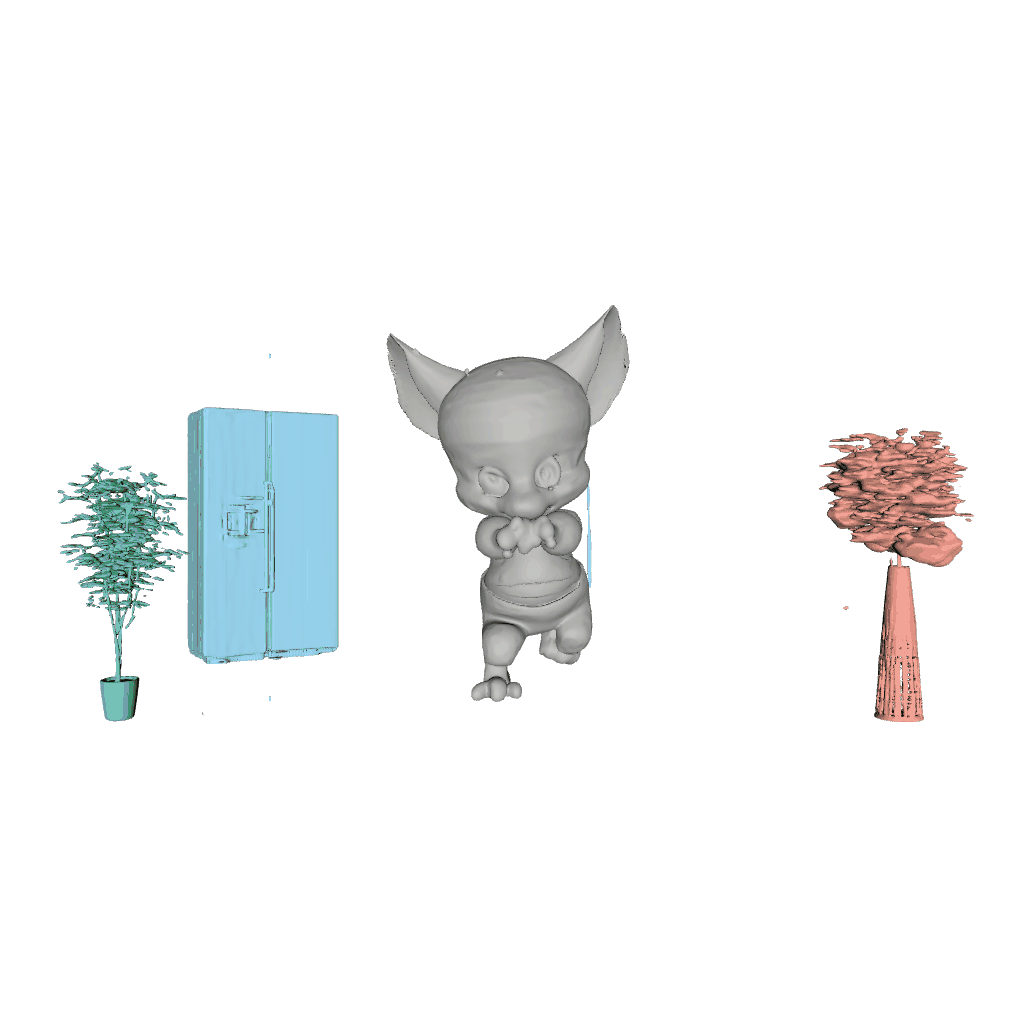}} &
\raisebox{-0.5\height}{\includegraphics[width=0.23\columnwidth, trim=0 0 0 0, clip]{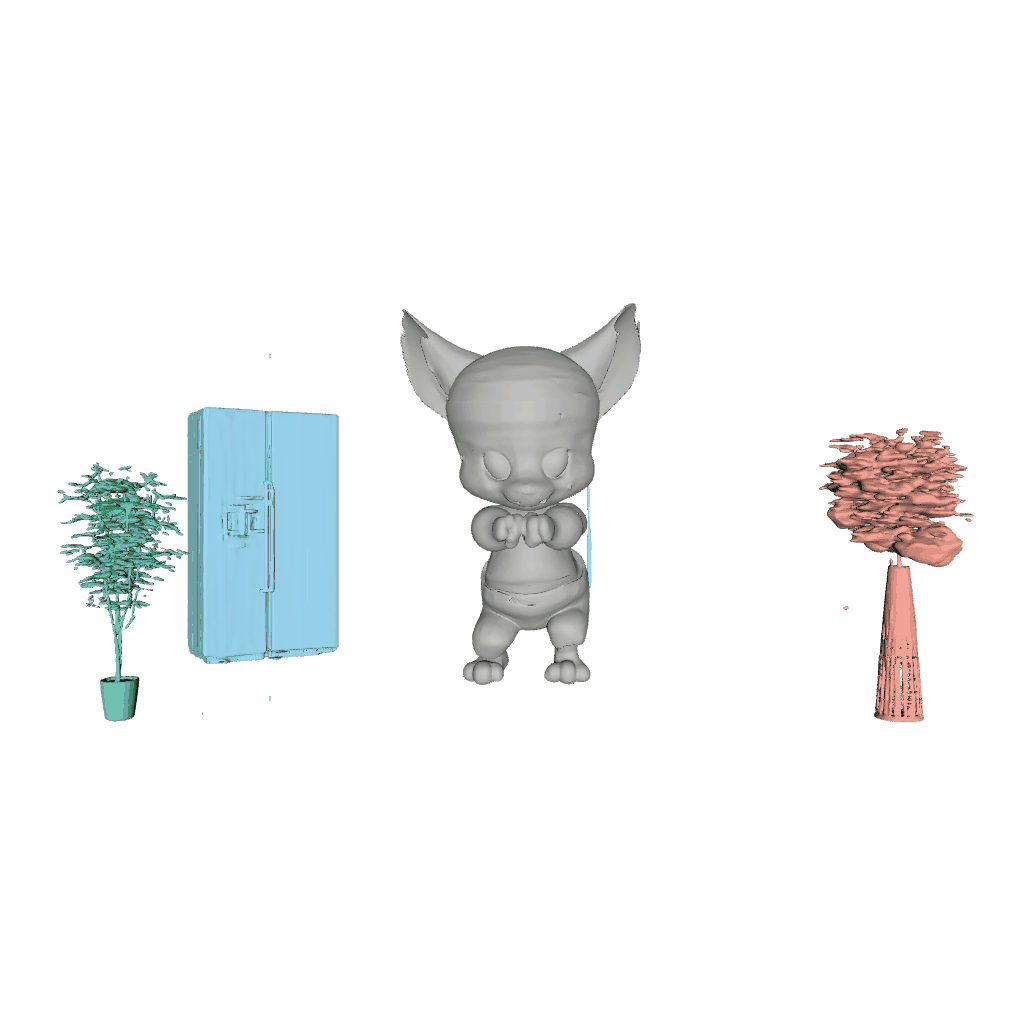}} \\[4pt]

% ==================== SAMPLE 3: jolleen_Climbing__markerman_InjuredWalkLeftTurn_3253_20260324_000651 ====================
\raisebox{-0.5\height}{\rotatebox{90}{\tiny Input}} &
\raisebox{-0.5\height}{\includegraphics[width=0.23\columnwidth, trim=0 0 0 0, clip]{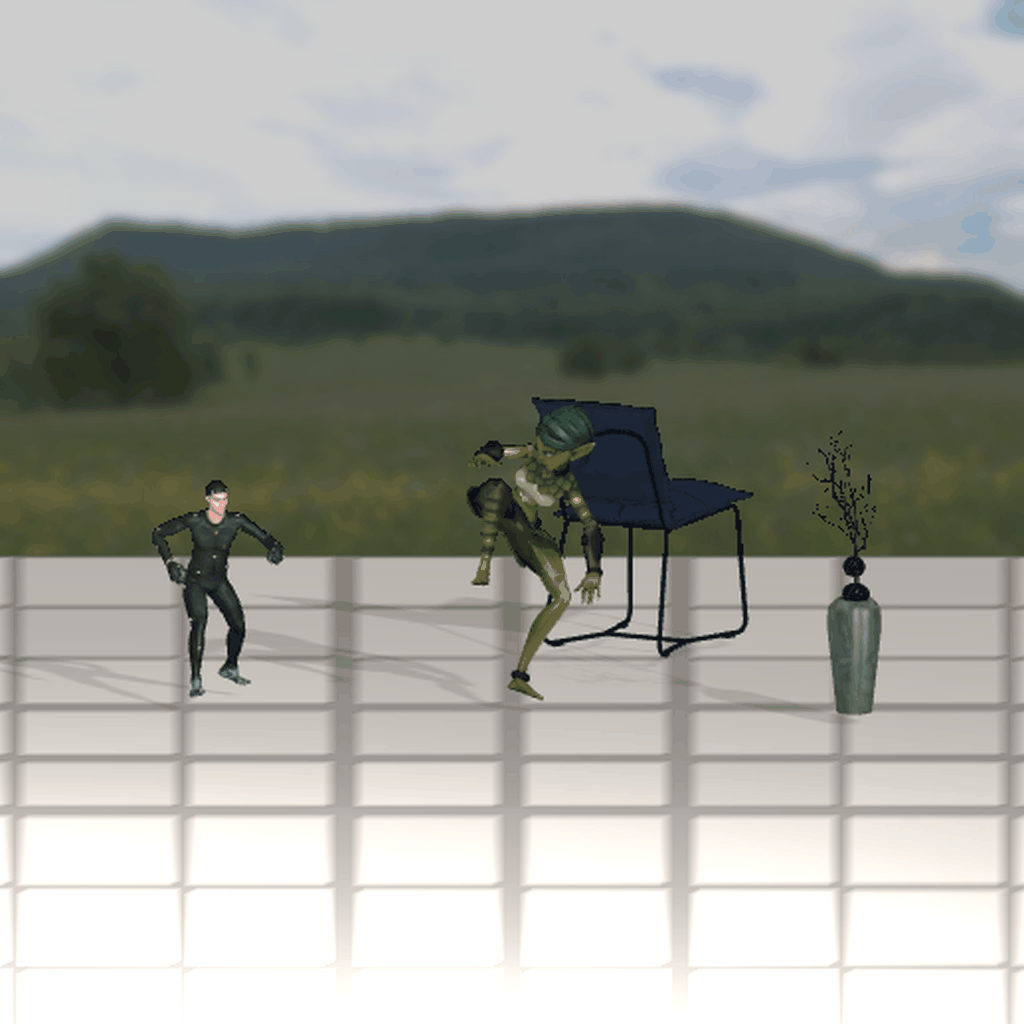}} &
\raisebox{-0.5\height}{\includegraphics[width=0.23\columnwidth, trim=0 0 0 0, clip]{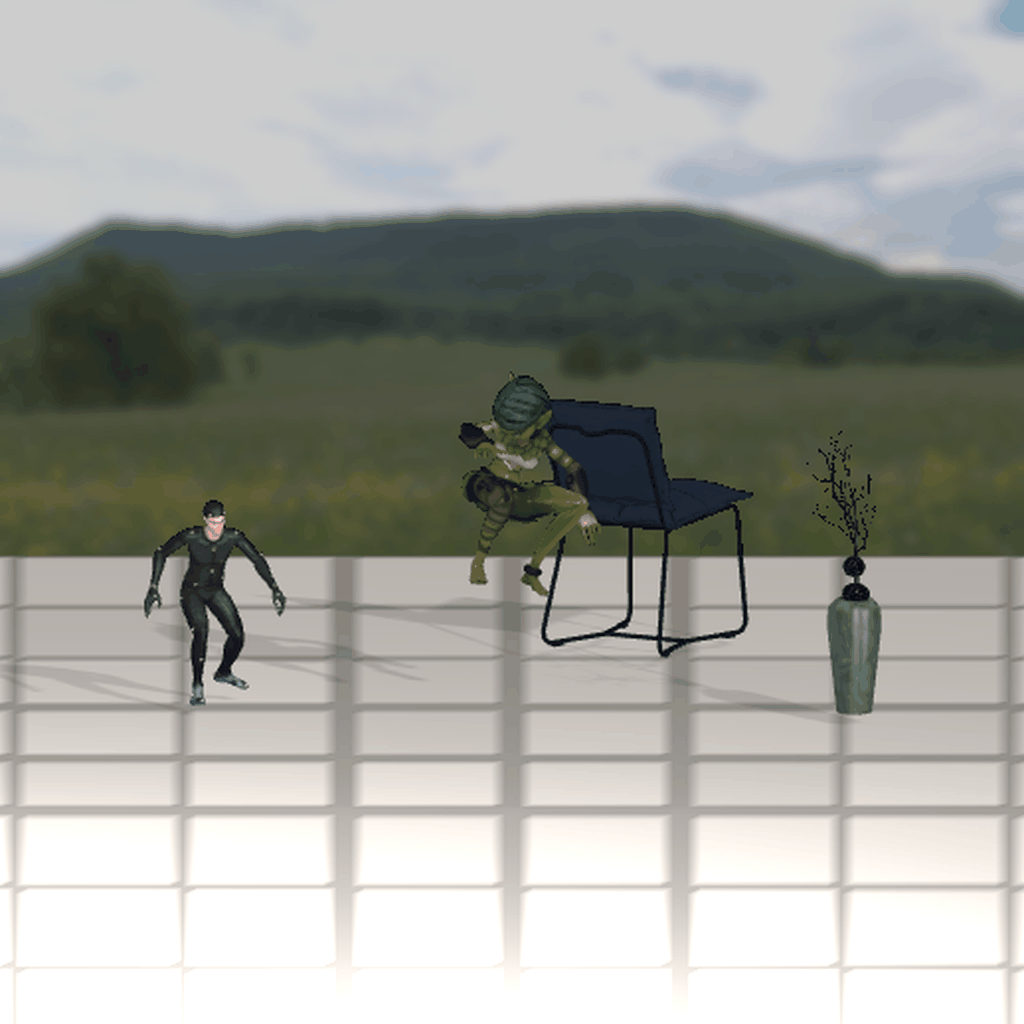}} &
\raisebox{-0.5\height}{\includegraphics[width=0.23\columnwidth, trim=0 0 0 0, clip]{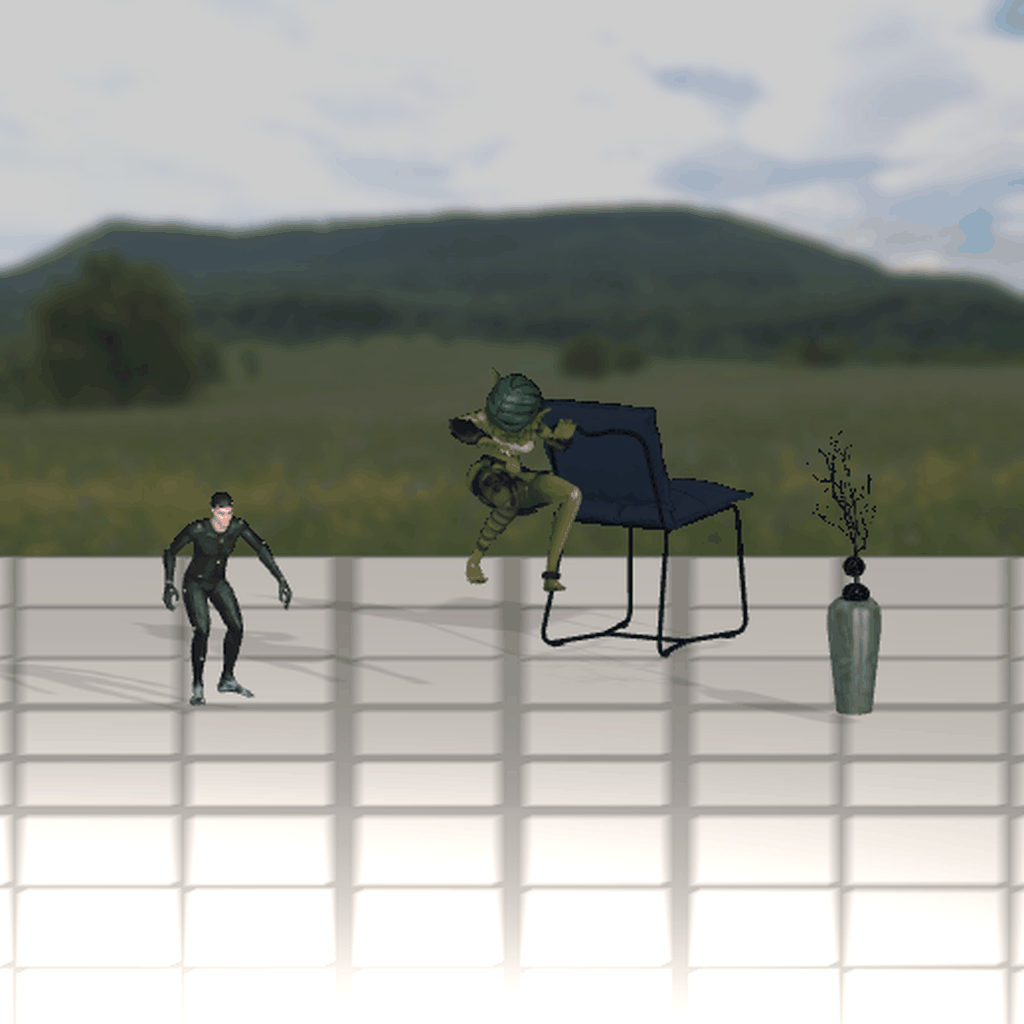}} &
\raisebox{-0.5\height}{\includegraphics[width=0.23\columnwidth, trim=0 0 0 0, clip]{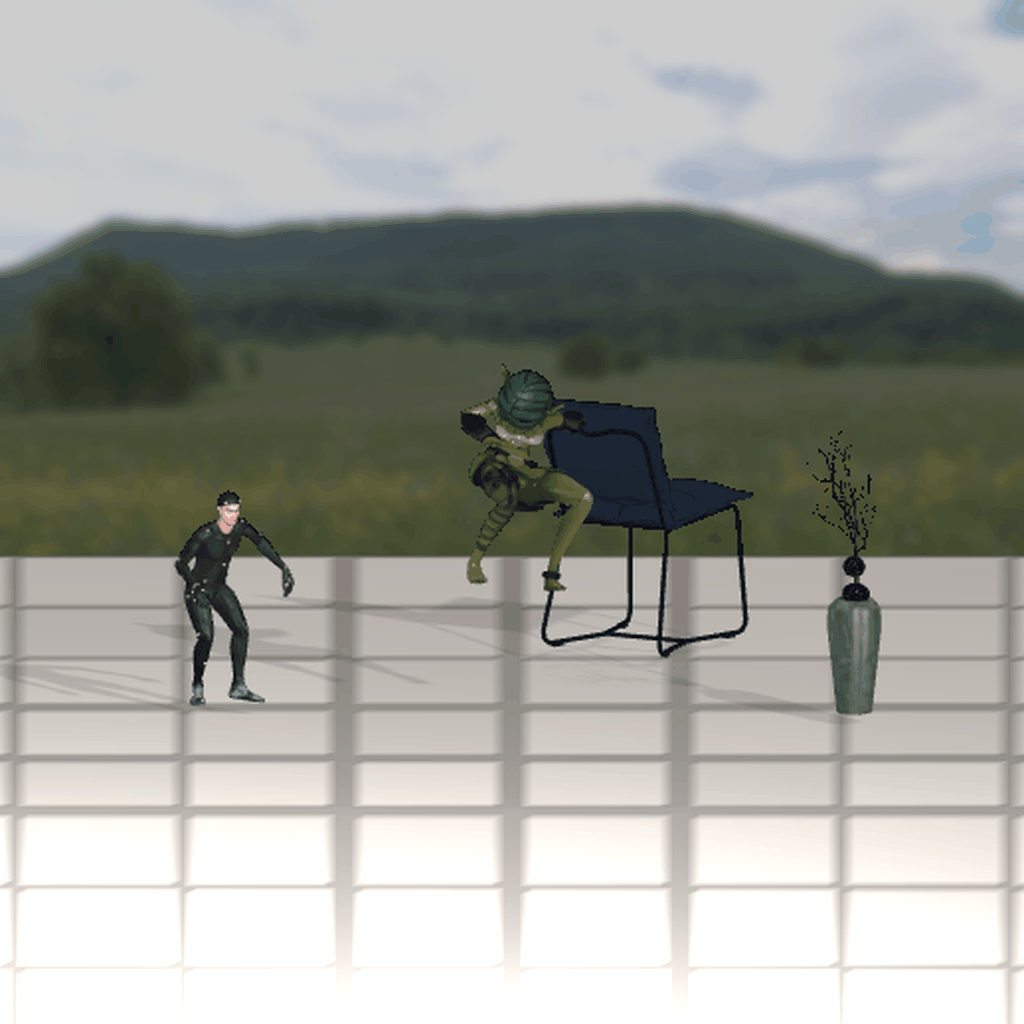}} \\[-0.5pt]
\raisebox{-0.5\height}{\rotatebox{90}{\tiny COM4D}} &
\raisebox{-0.5\height}{\includegraphics[width=0.23\columnwidth, trim=0 0 0 0, clip]{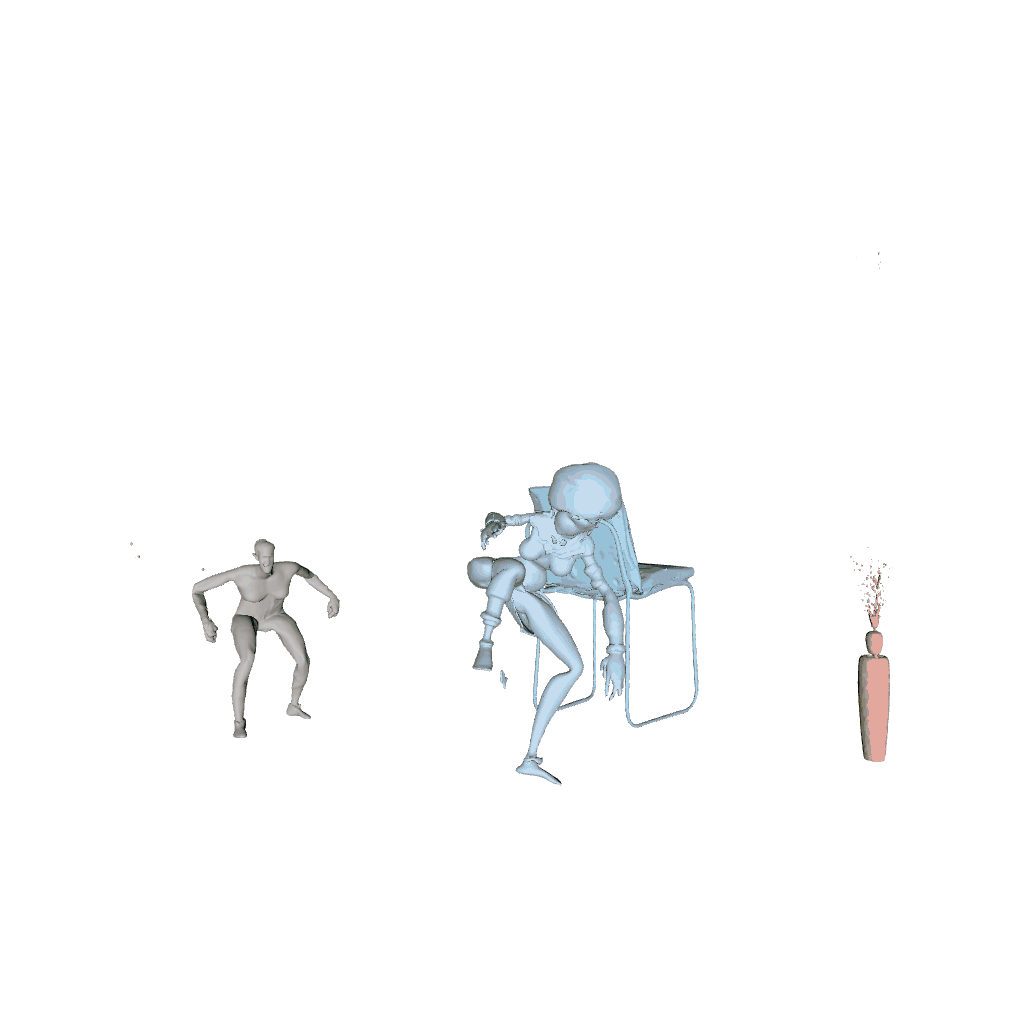}} &
\raisebox{-0.5\height}{\includegraphics[width=0.23\columnwidth, trim=0 0 0 0, clip]{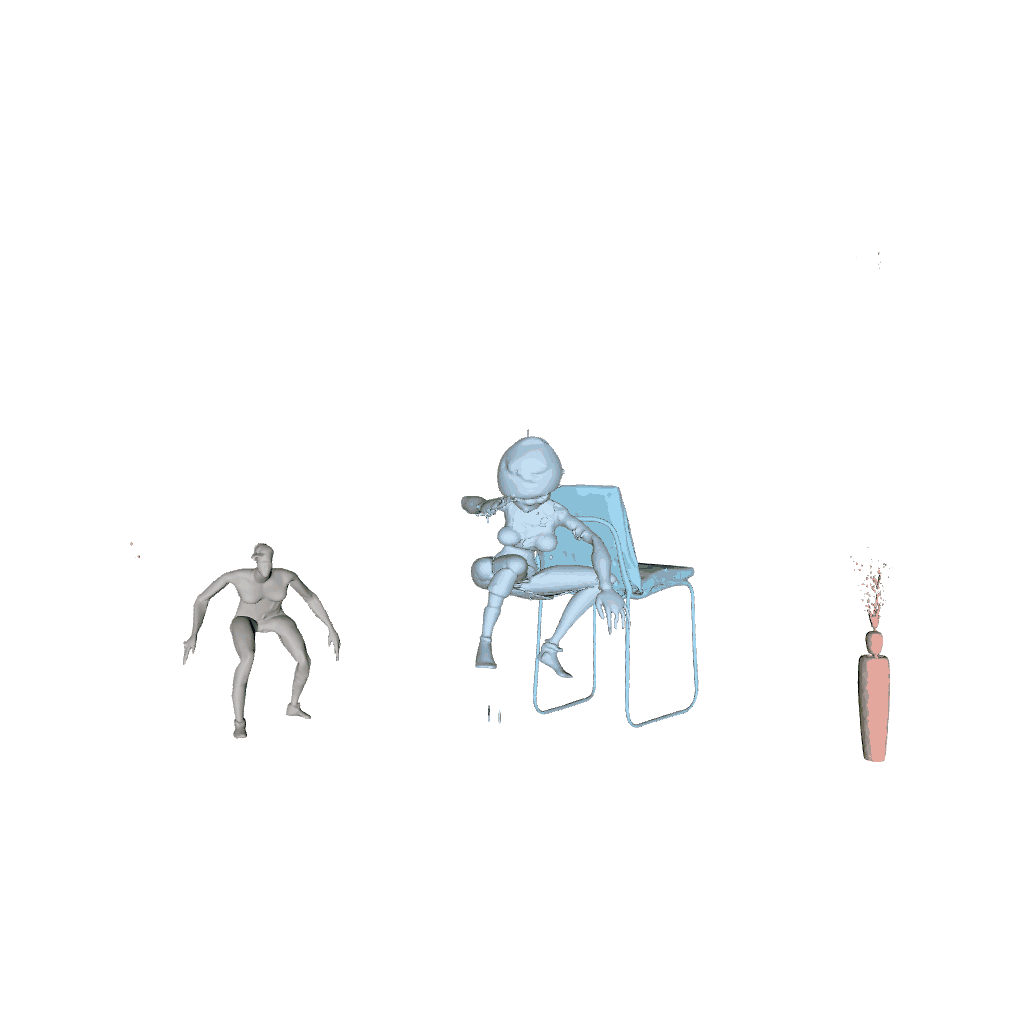}} &
\raisebox{-0.5\height}{\includegraphics[width=0.23\columnwidth, trim=0 0 0 0, clip]{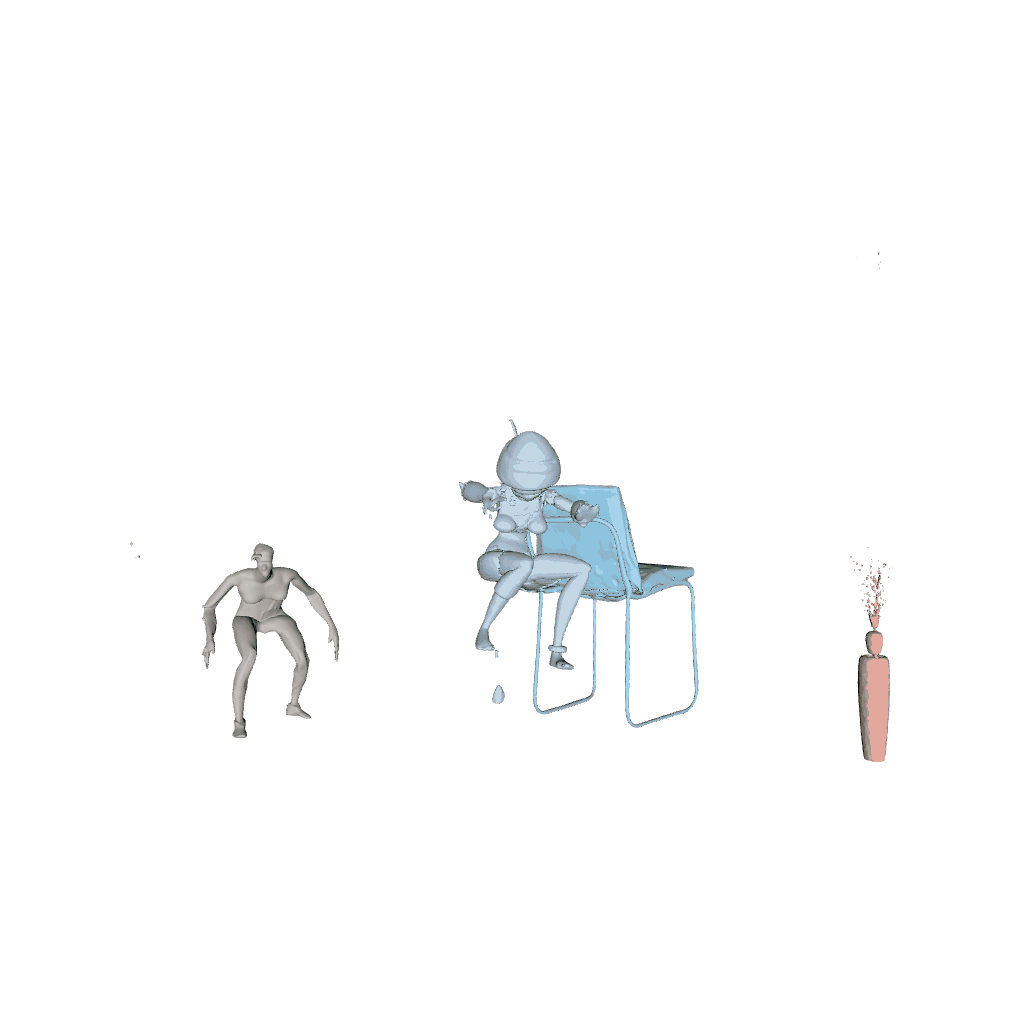}} &
\raisebox{-0.5\height}{\includegraphics[width=0.23\columnwidth, trim=0 0 0 0, clip]{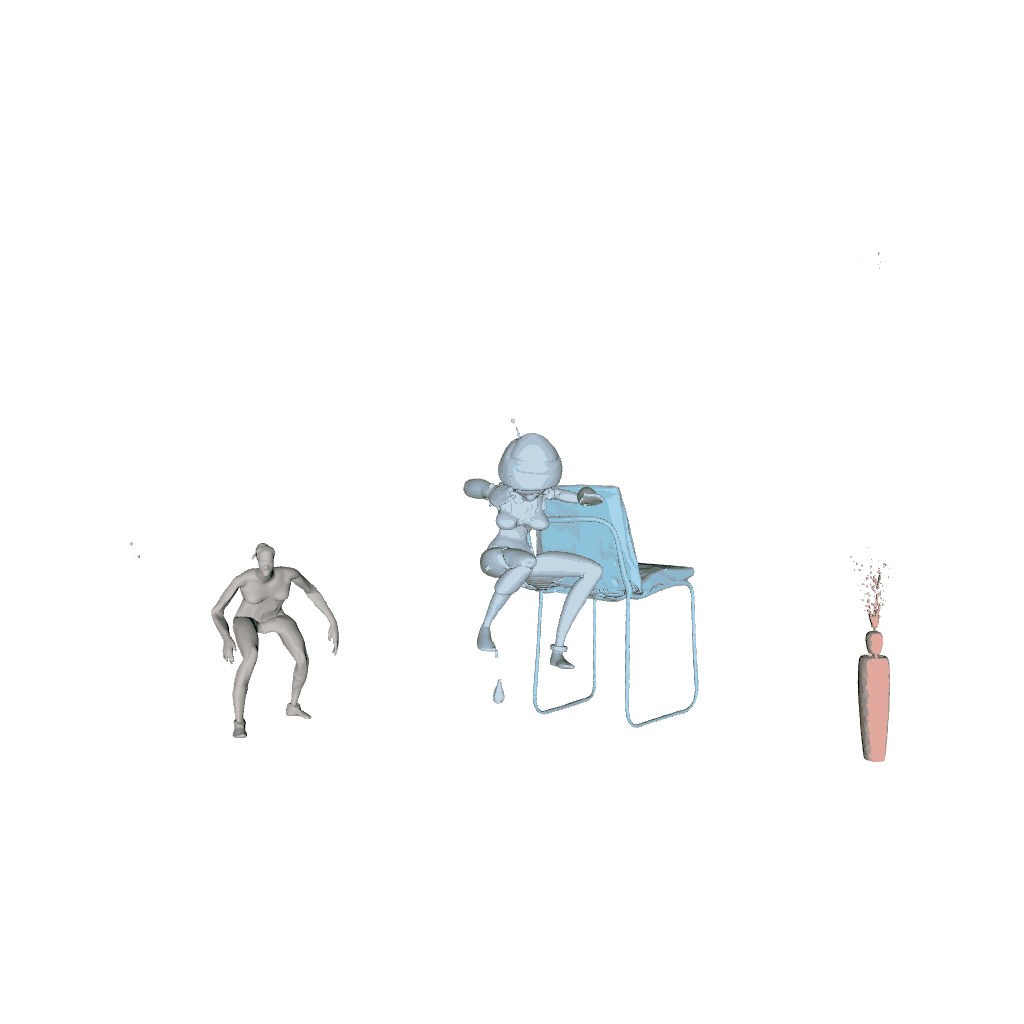}} \\[4pt]

% ==================== SAMPLE 4: arissa_GreatSwordSlash__markerman_WheelbarrowWalkTurn_20260323_205106 ====================
\raisebox{-0.5\height}{\rotatebox{90}{\tiny Input}} &
\raisebox{-0.5\height}{\includegraphics[width=0.23\columnwidth, trim=0 0 0 0, clip]{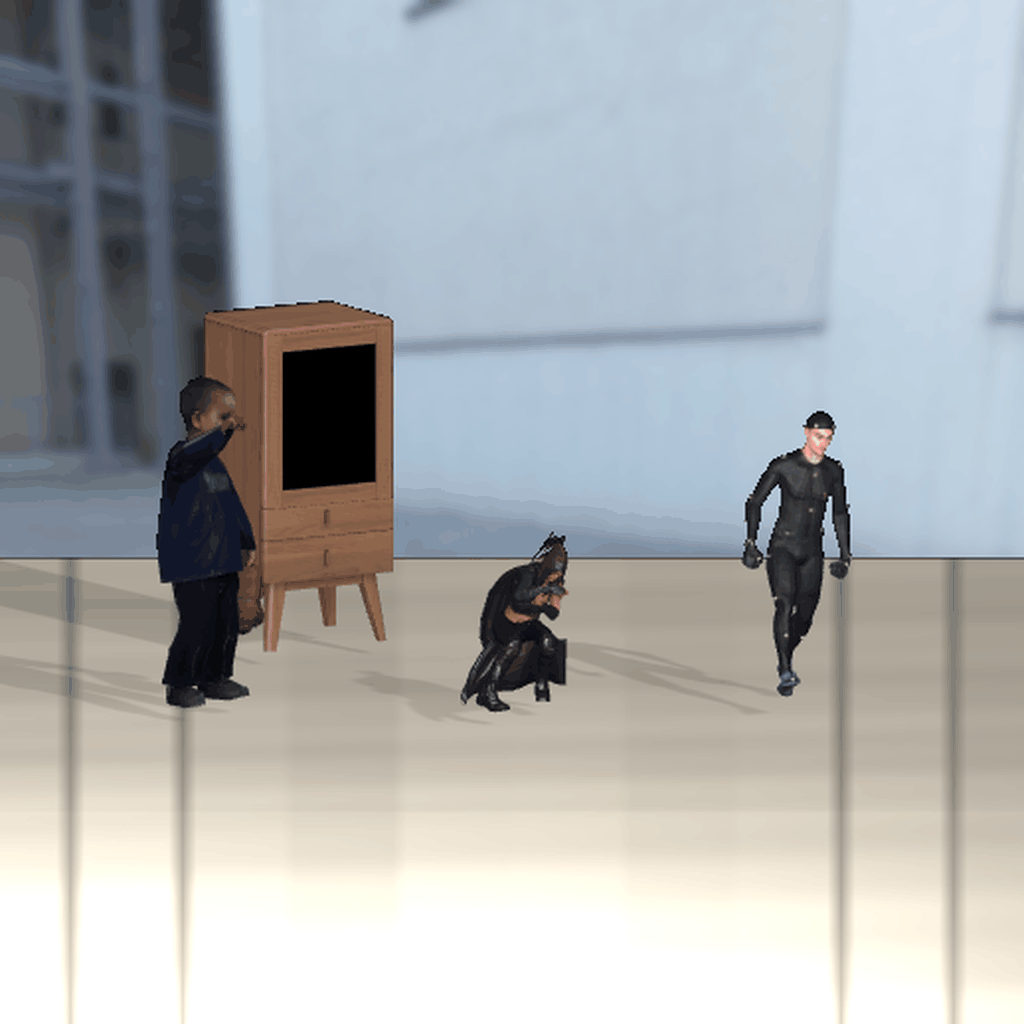}} &
\raisebox{-0.5\height}{\includegraphics[width=0.23\columnwidth, trim=0 0 0 0, clip]{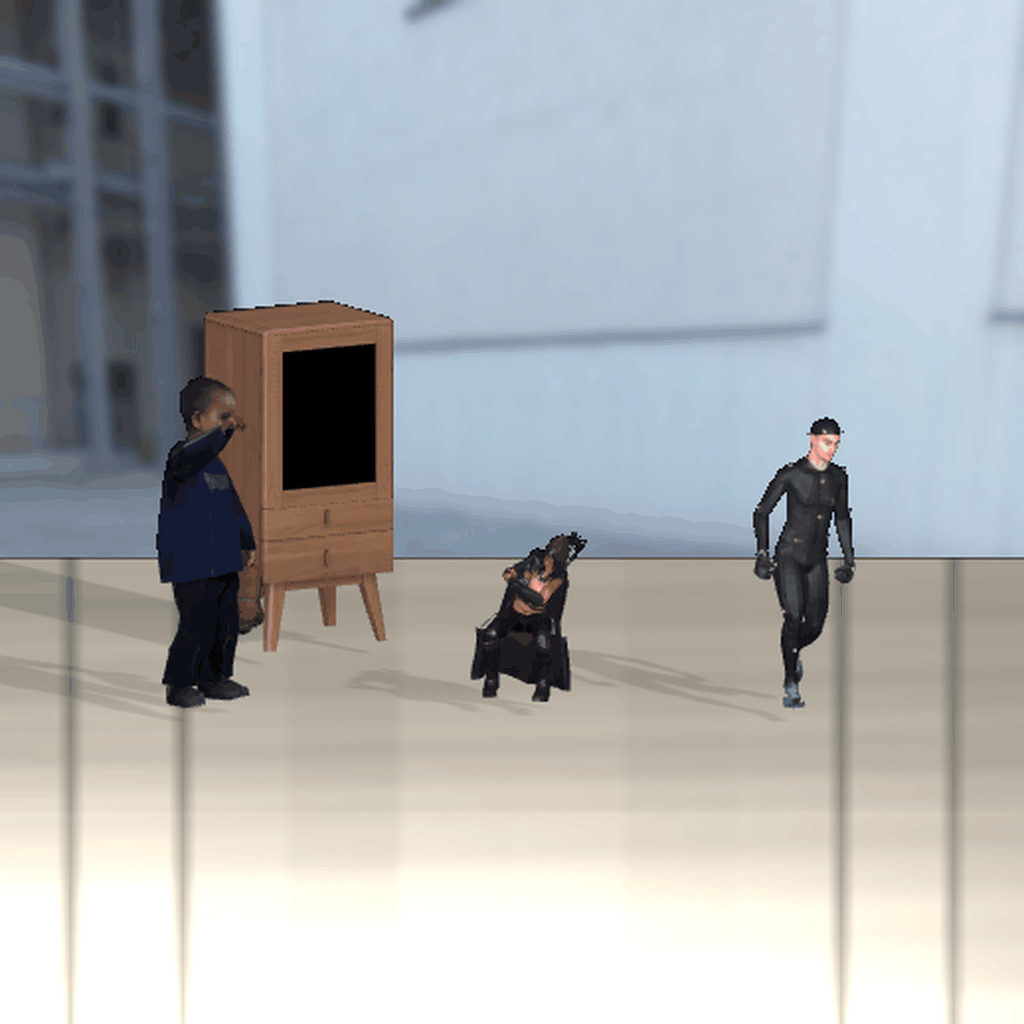}} &
\raisebox{-0.5\height}{\includegraphics[width=0.23\columnwidth, trim=0 0 0 0, clip]{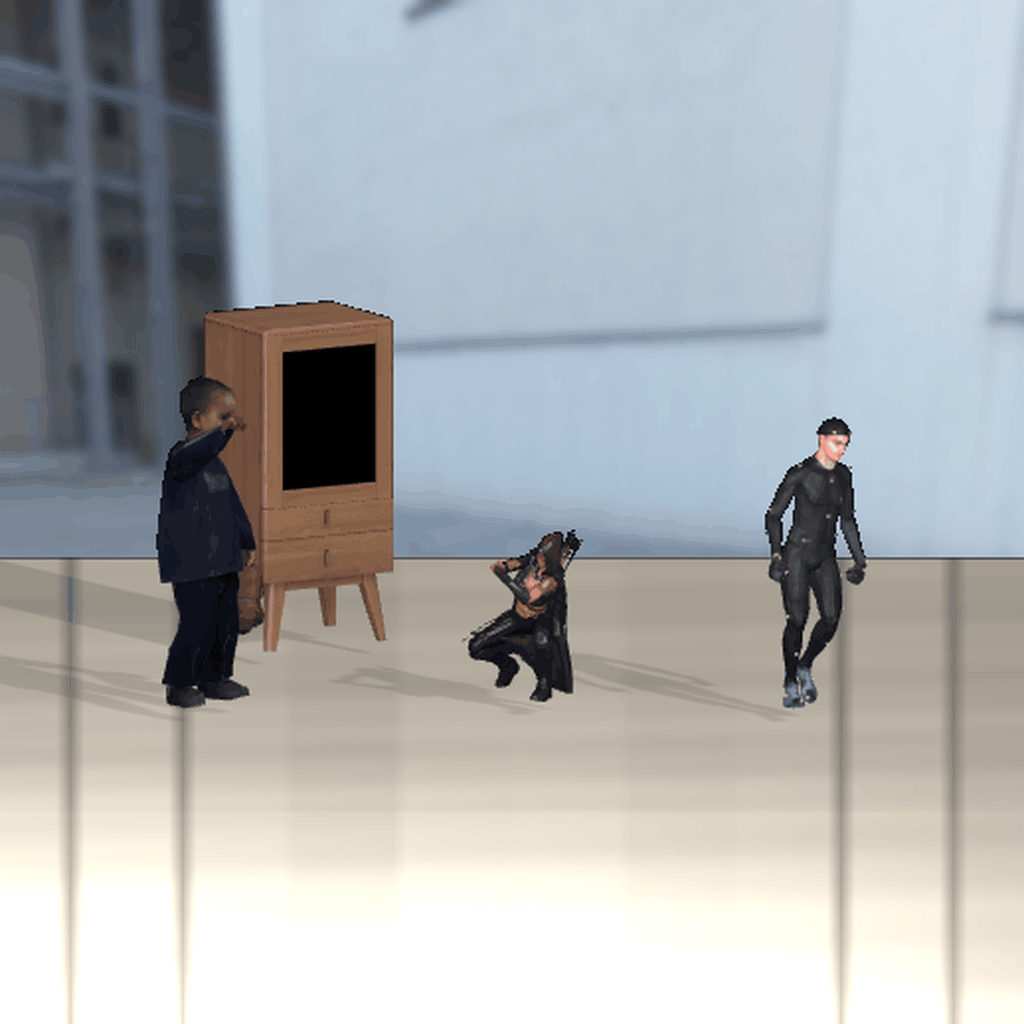}} &
\raisebox{-0.5\height}{\includegraphics[width=0.23\columnwidth, trim=0 0 0 0, clip]{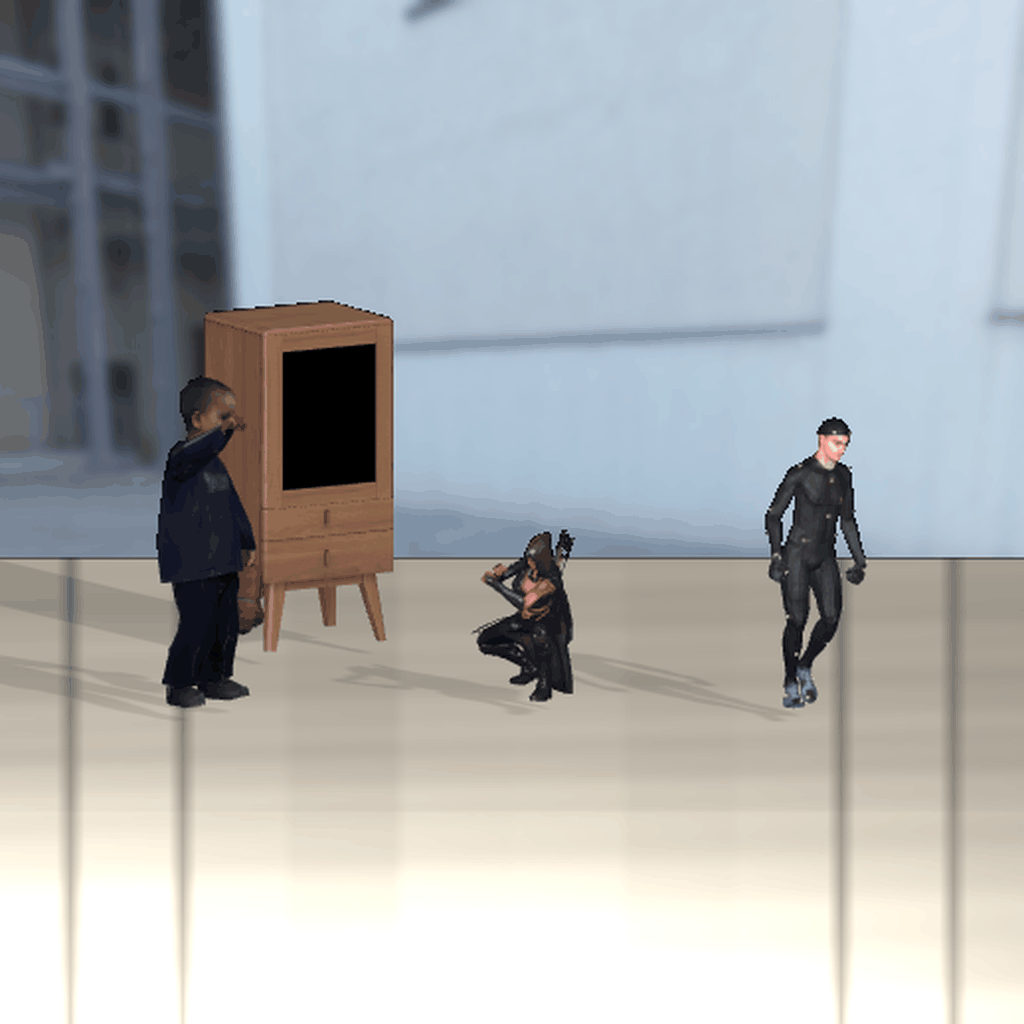}} \\[-0.5pt]
\raisebox{-0.5\height}{\rotatebox{90}{\tiny COM4D}} &
\raisebox{-0.5\height}{\includegraphics[width=0.23\columnwidth, trim=0 0 0 0, clip]{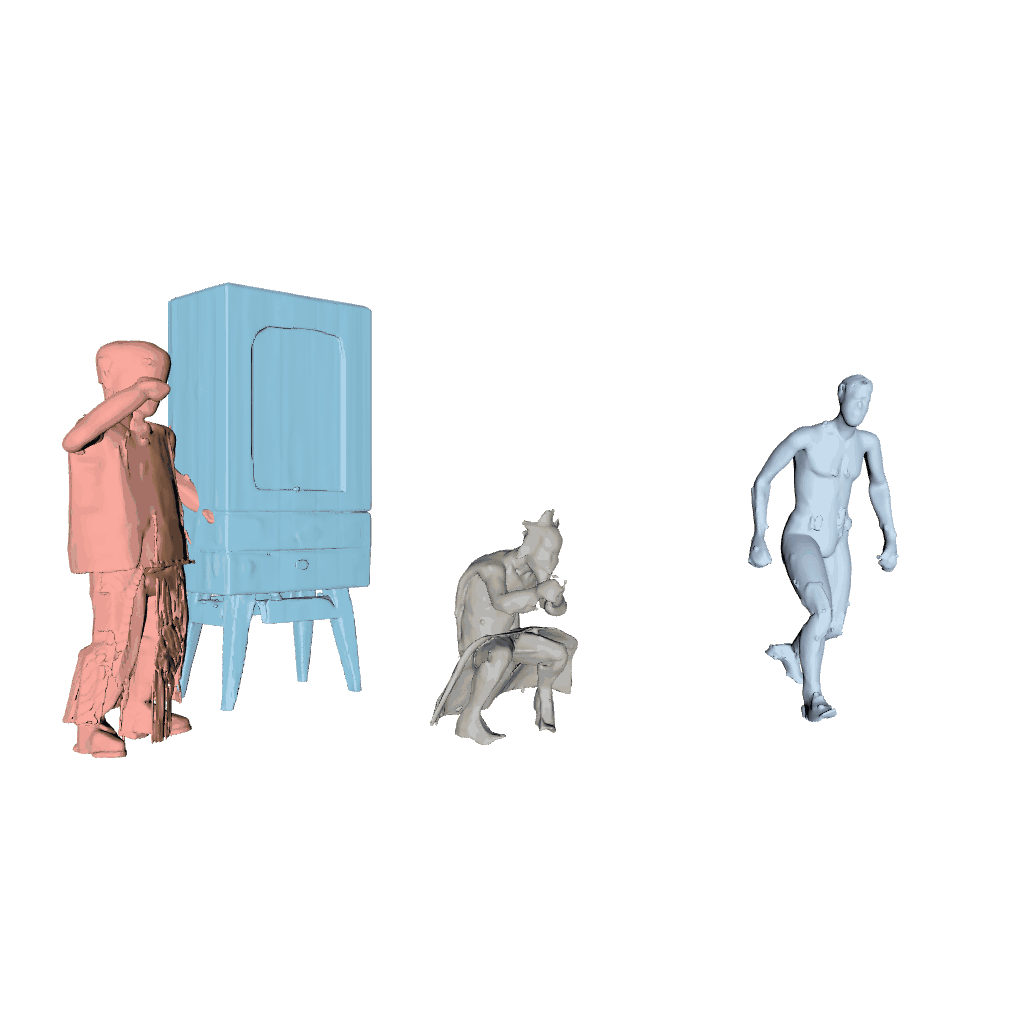}} &
\raisebox{-0.5\height}{\includegraphics[width=0.23\columnwidth, trim=0 0 0 0, clip]{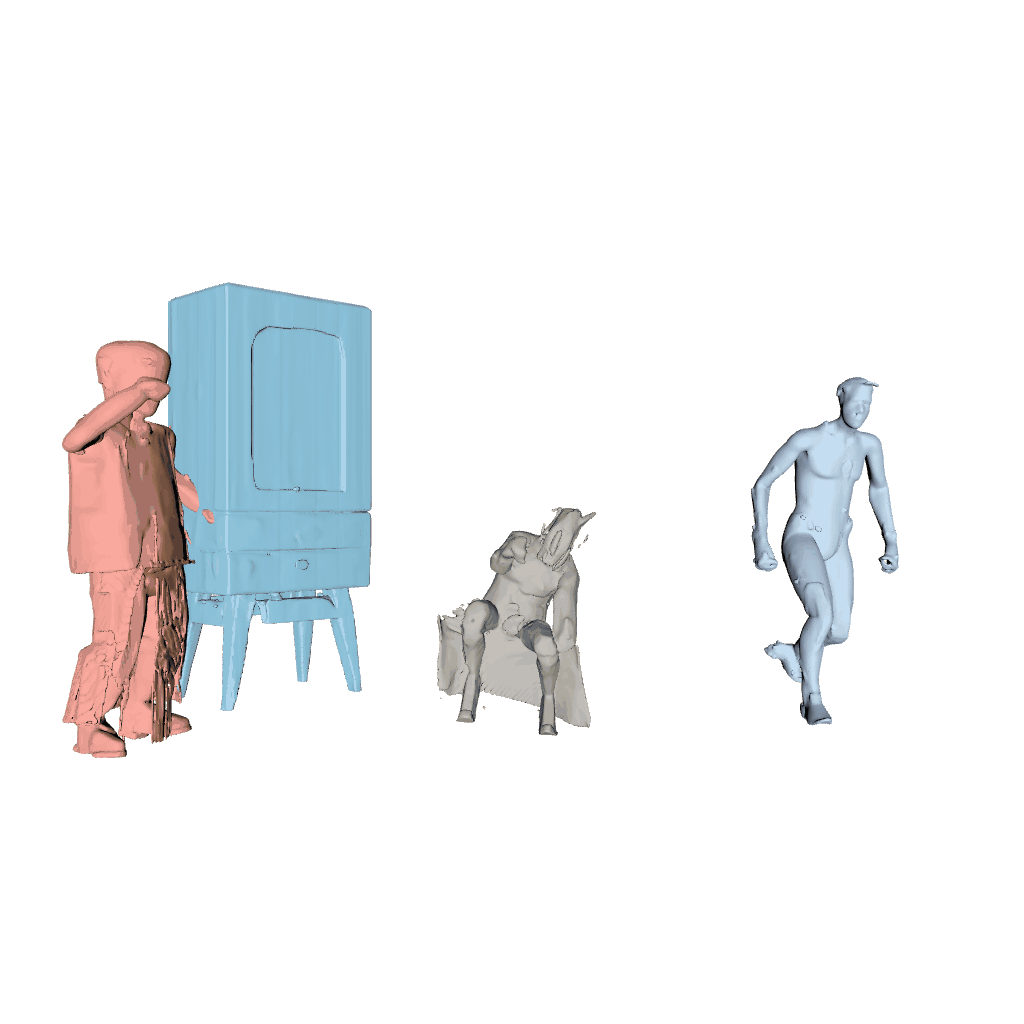}} &
\raisebox{-0.5\height}{\includegraphics[width=0.23\columnwidth, trim=0 0 0 0, clip]{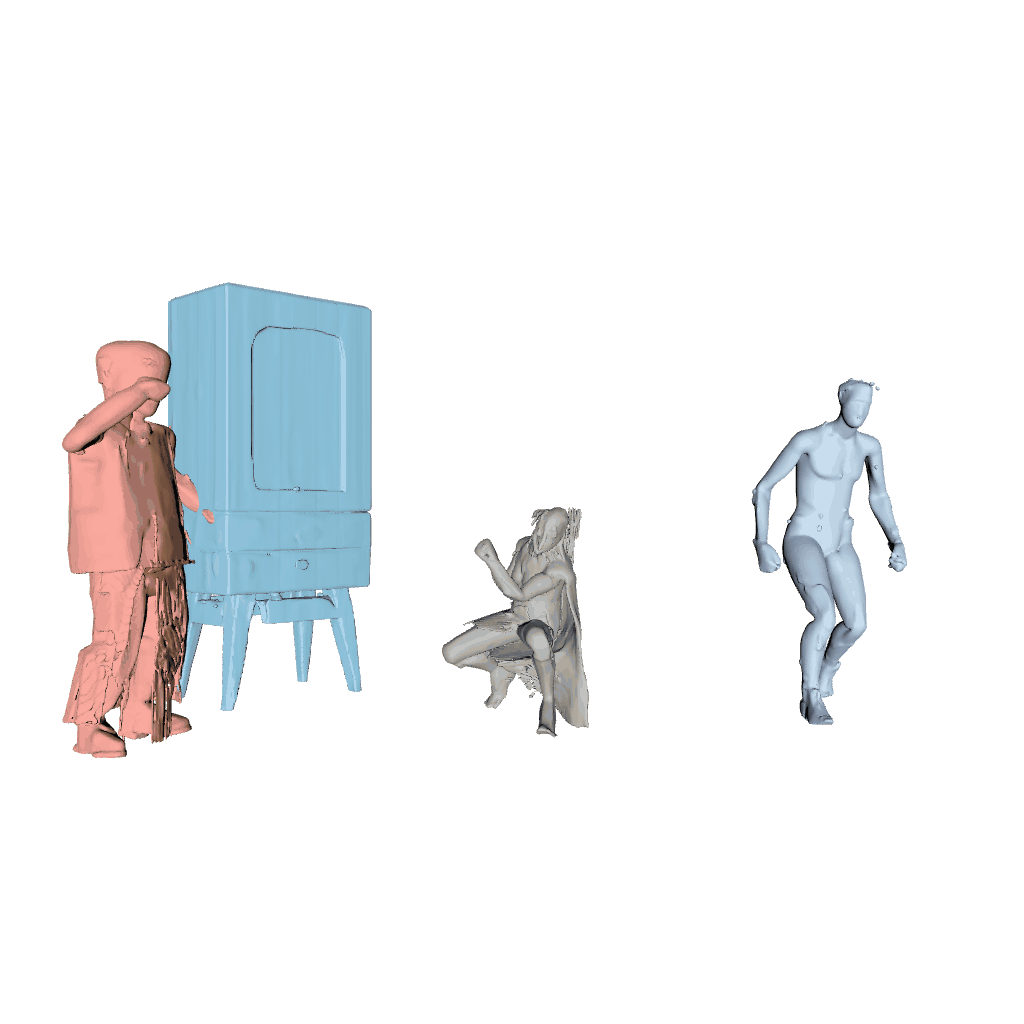}} &
\raisebox{-0.5\height}{\includegraphics[width=0.23\columnwidth, trim=0 0 0 0, clip]{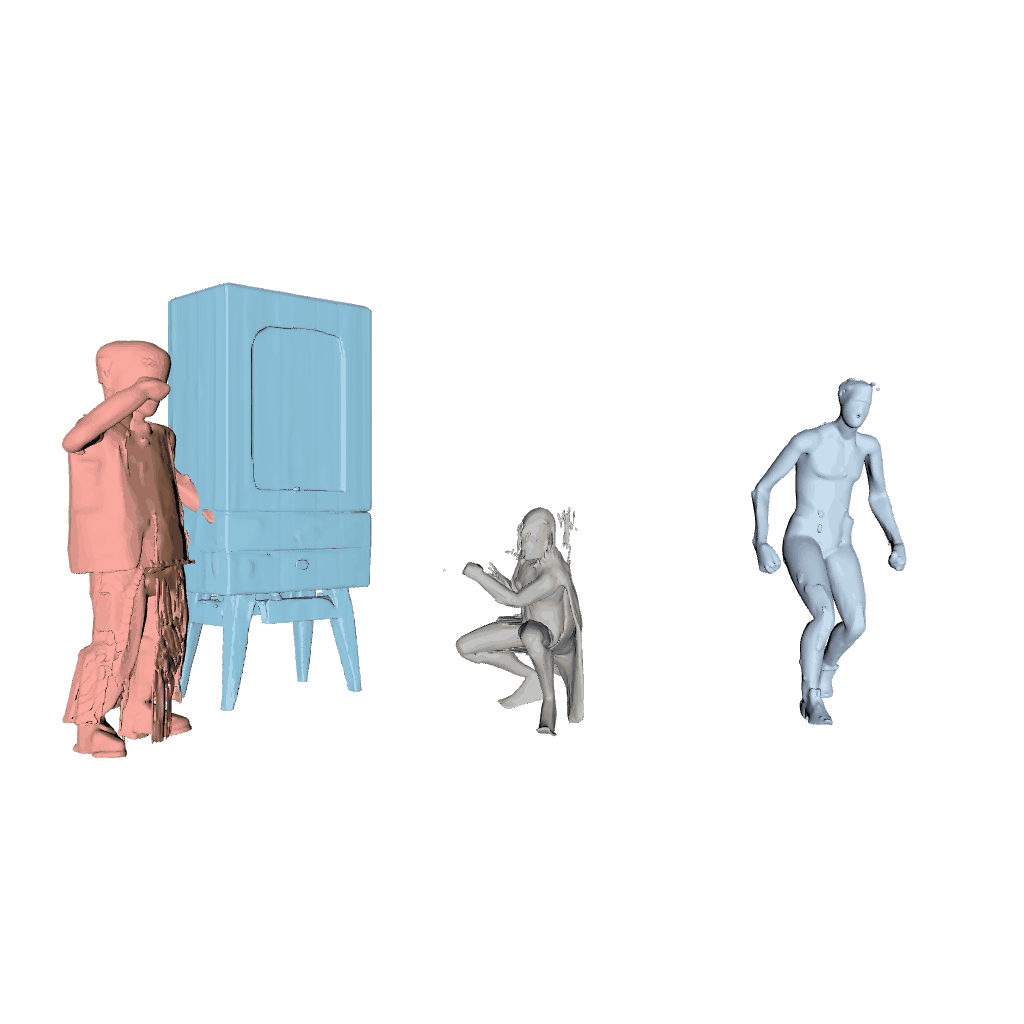}} \\

  \end{tabular}

  \vspace{-3pt}
  \caption{
  Additional qualitative results on the synthetic compositional 4D dataset.
  }
  \label{fig:synthetic_com4d_appendix_qual_2}
\end{figure}

\begin{figure}[h]
  
  \centering
  \setlength{\tabcolsep}{0pt}
  \renewcommand{\arraystretch}{0}

  \begin{tabular}{@{}c@{\hspace{2pt}}cccc@{}}

% ==================== SAMPLE 1: Maria_SteppingBackward__clarie_WheelbarrowWalk_6020_20260324_000626 ====================
\raisebox{-0.5\height}{\rotatebox{90}{\tiny Input}} &
\raisebox{-0.5\height}{\includegraphics[width=0.23\columnwidth, trim=0 0 0 0, clip]{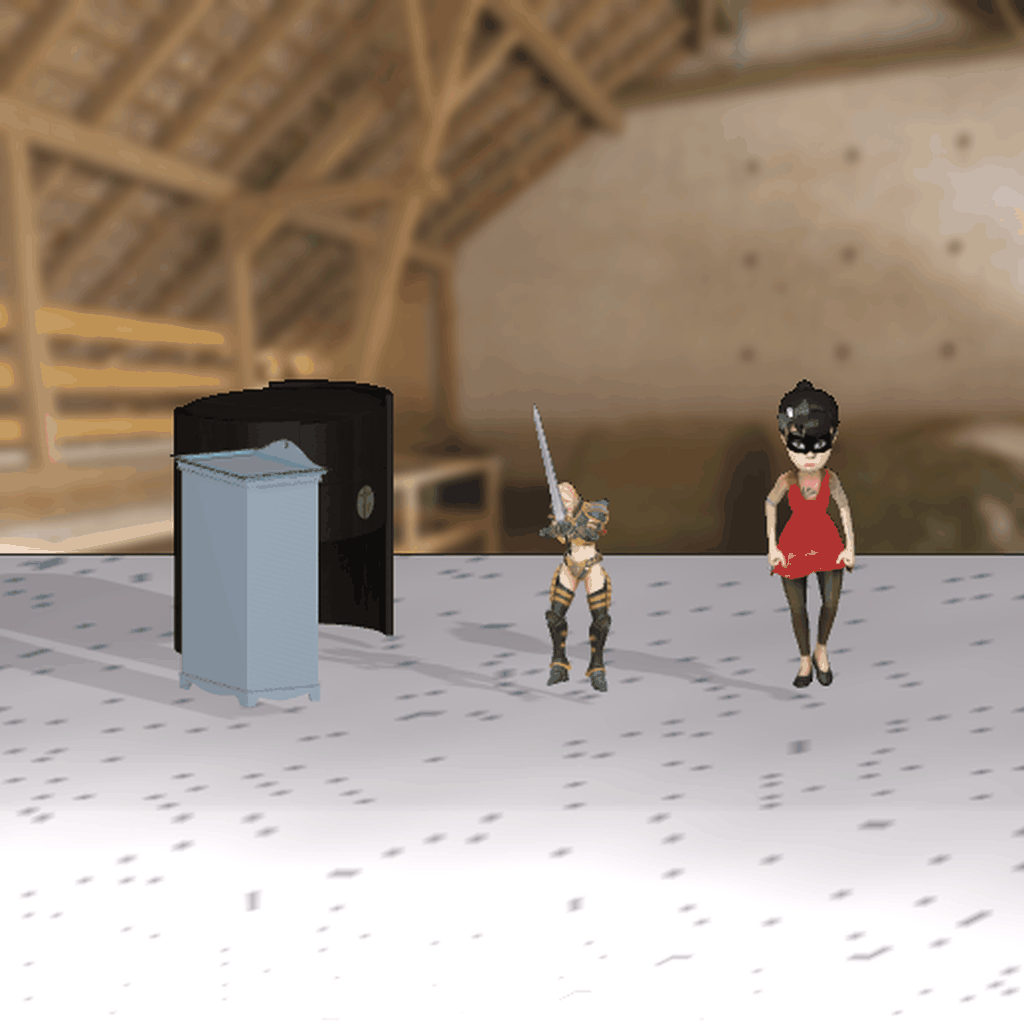}} &
\raisebox{-0.5\height}{\includegraphics[width=0.23\columnwidth, trim=0 0 0 0, clip]{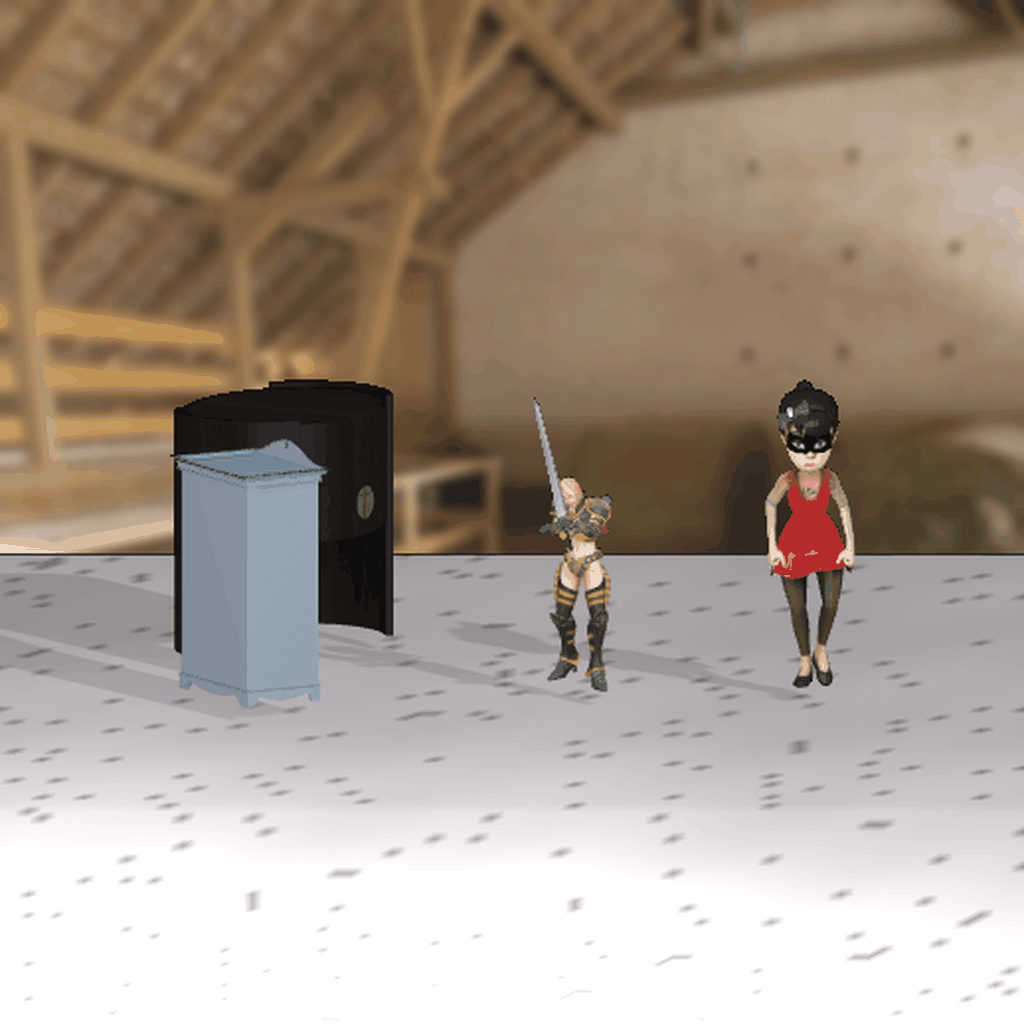}} &
\raisebox{-0.5\height}{\includegraphics[width=0.23\columnwidth, trim=0 0 0 0, clip]{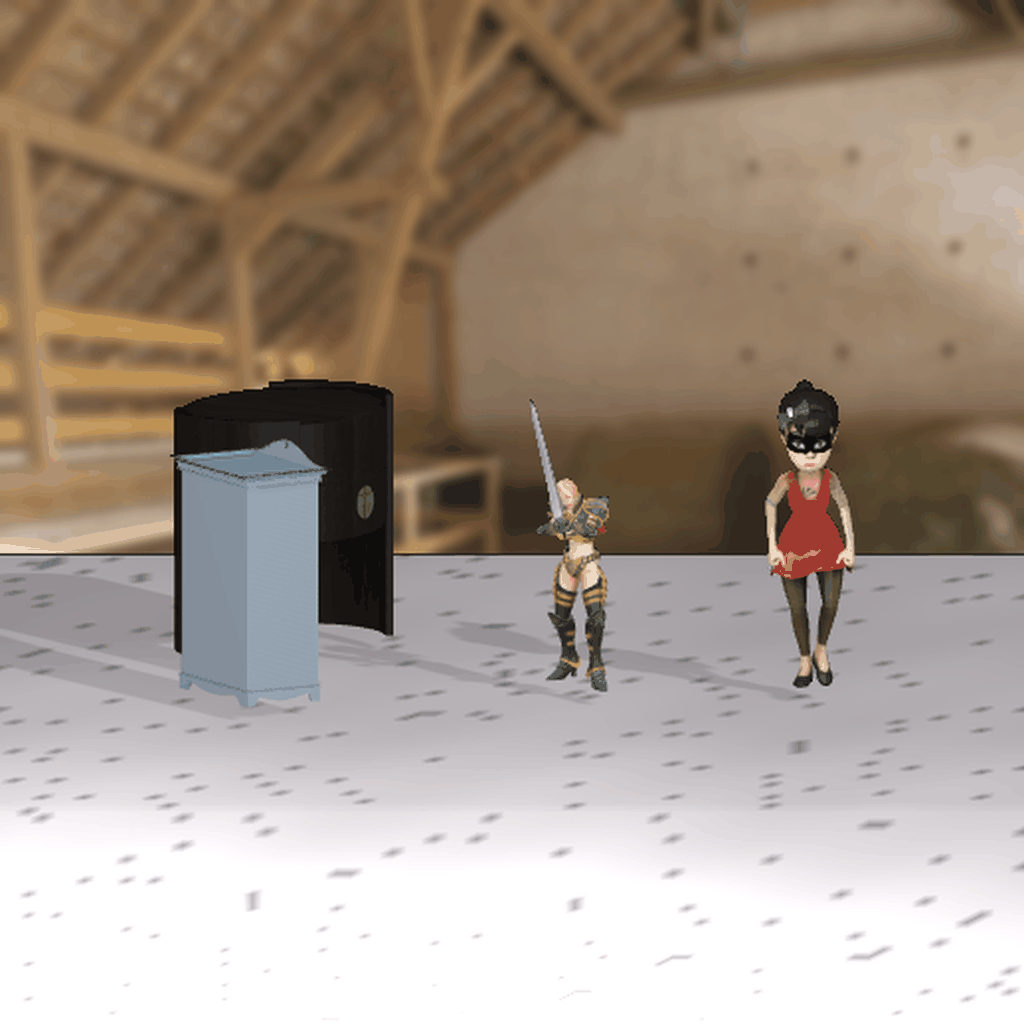}} &
\raisebox{-0.5\height}{\includegraphics[width=0.23\columnwidth, trim=0 0 0 0, clip]{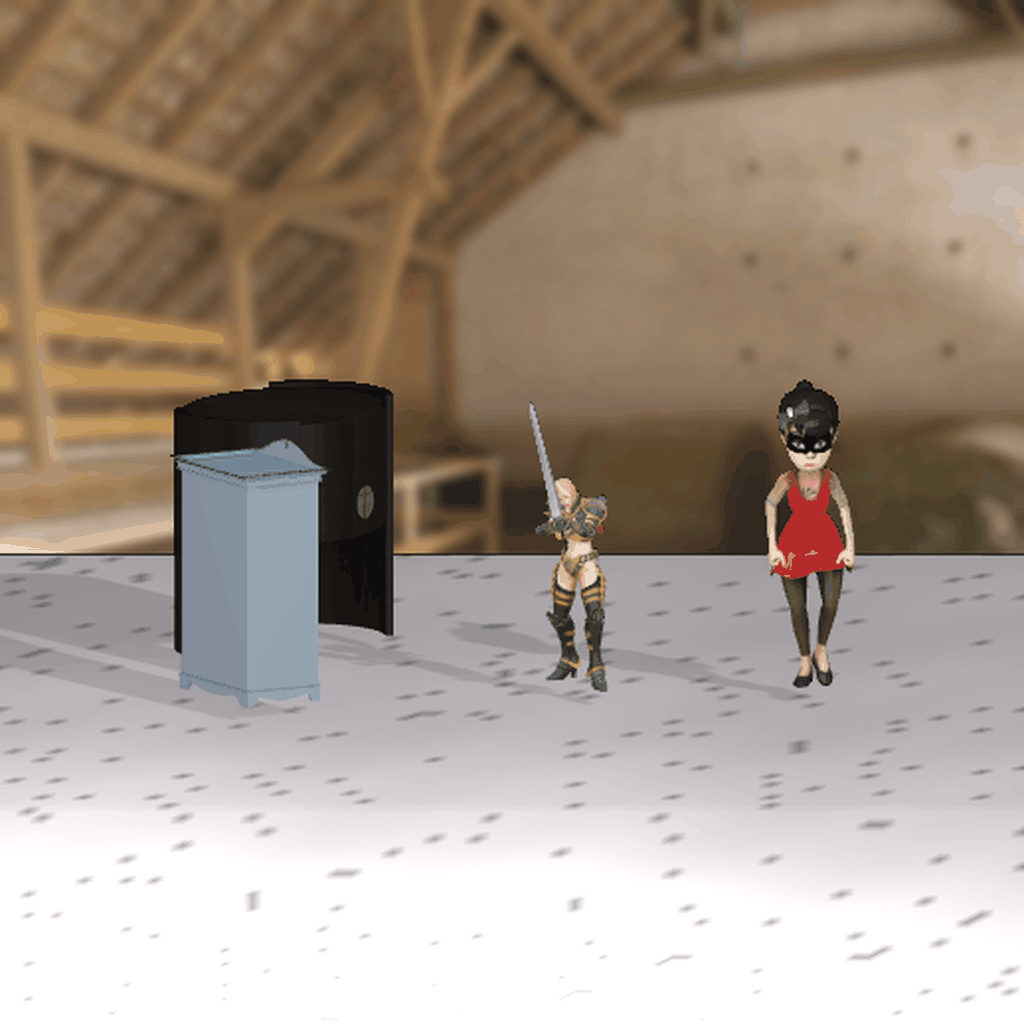}} \\[-0.5pt]
\raisebox{-0.5\height}{\rotatebox{90}{\tiny COM4D}} &
\raisebox{-0.5\height}{\includegraphics[width=0.23\columnwidth, trim=0 0 0 0, clip]{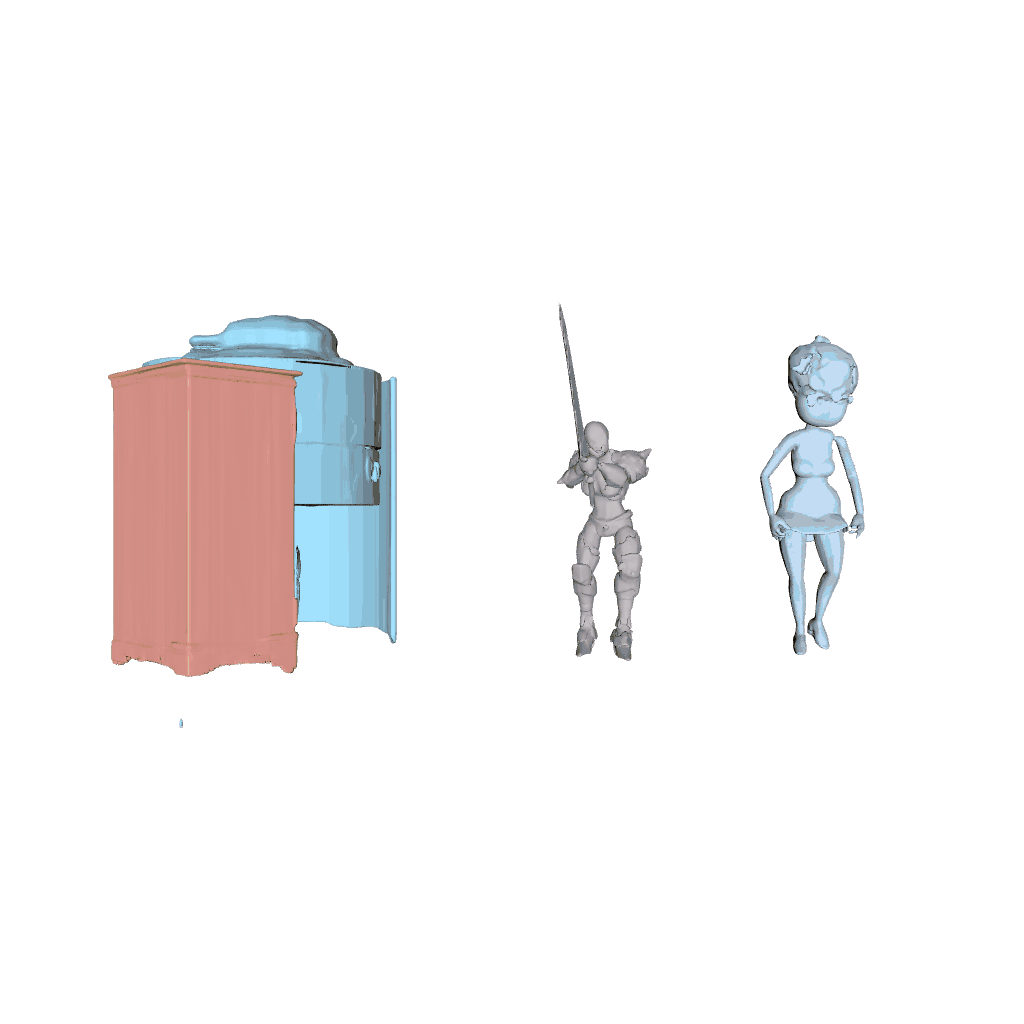}} &
\raisebox{-0.5\height}{\includegraphics[width=0.23\columnwidth, trim=0 0 0 0, clip]{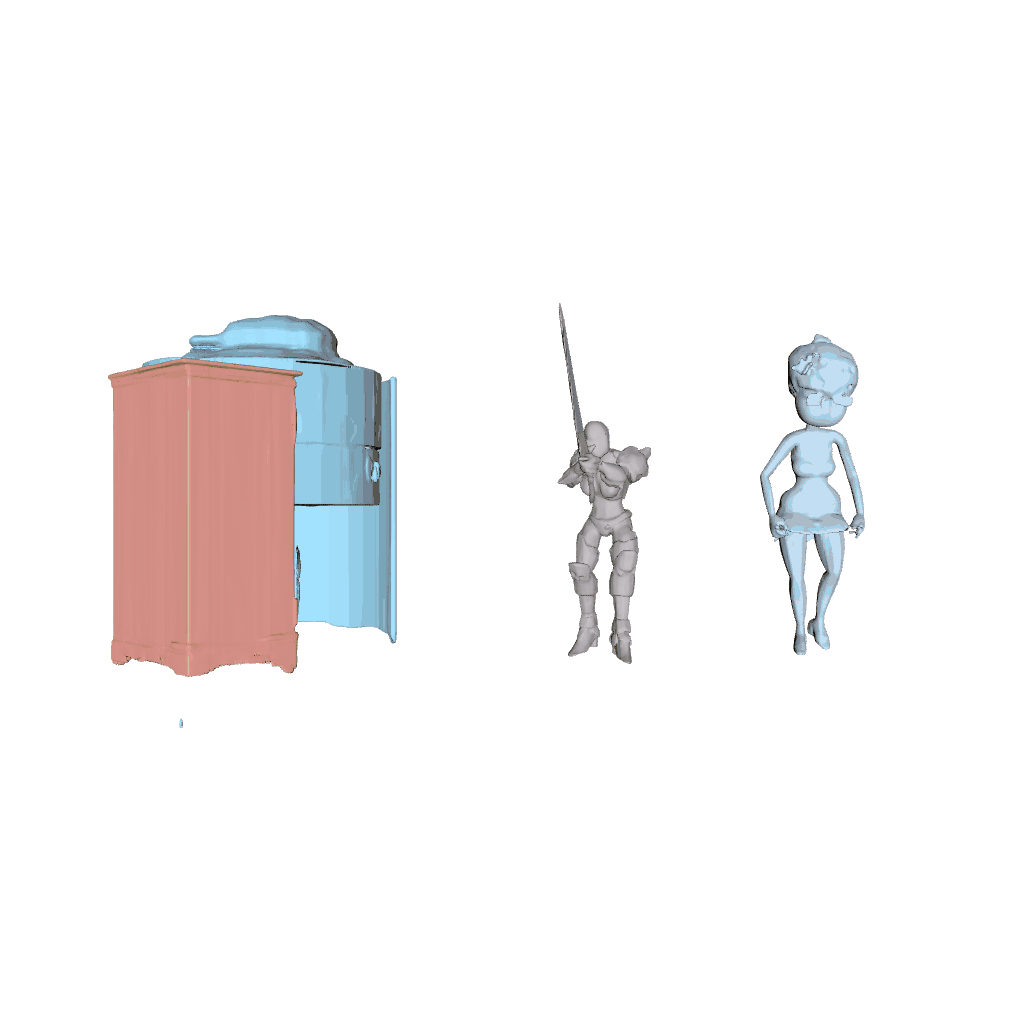}} &
\raisebox{-0.5\height}{\includegraphics[width=0.23\columnwidth, trim=0 0 0 0, clip]{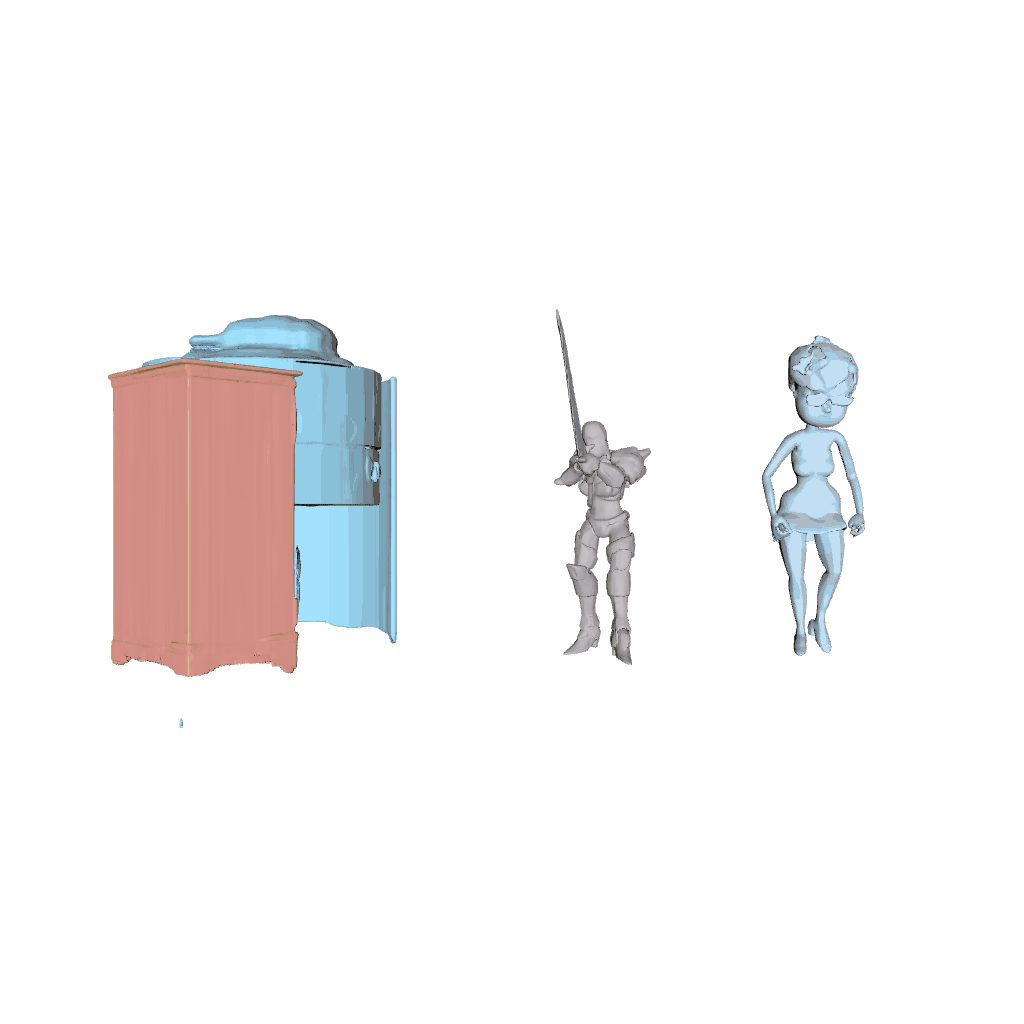}} &
\raisebox{-0.5\height}{\includegraphics[width=0.23\columnwidth, trim=0 0 0 0, clip]{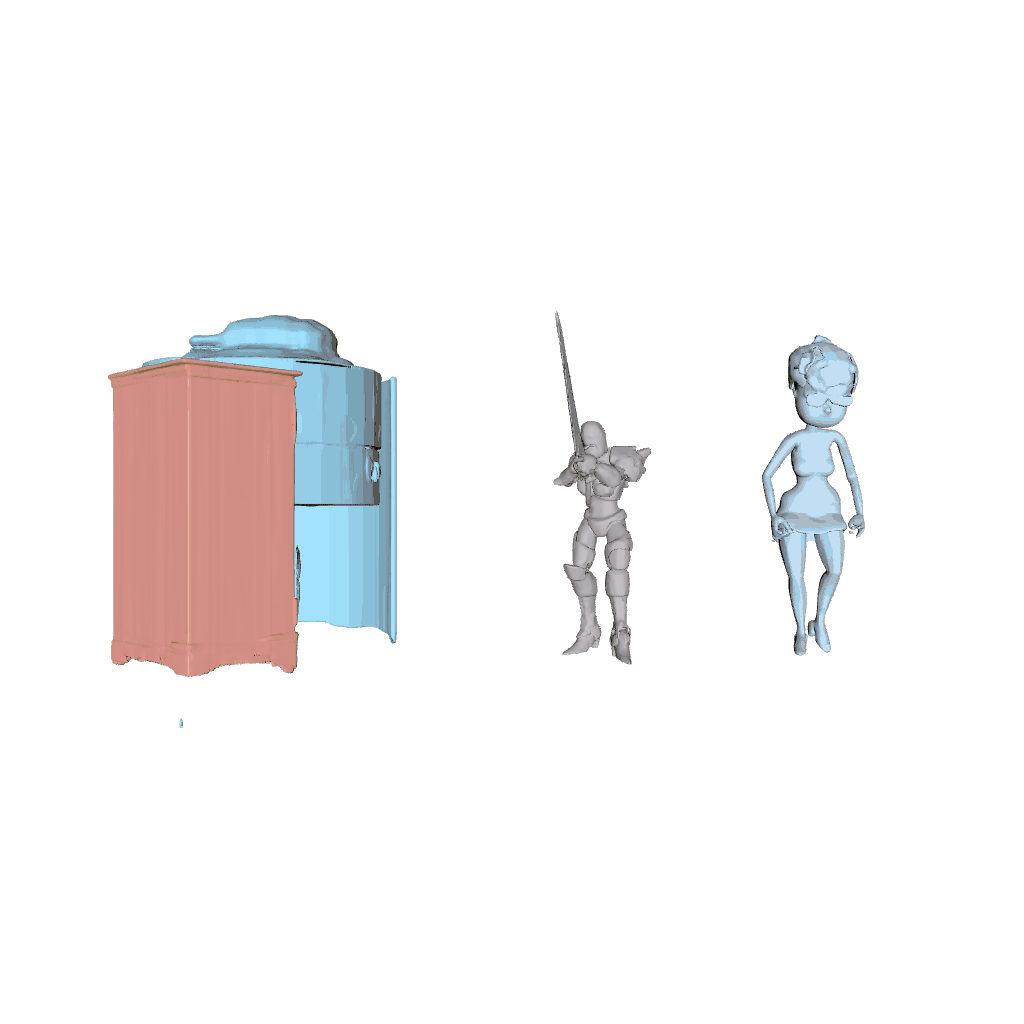}} \\[4pt]

% ==================== SAMPLE 2: Maria_SteppingBackward__vanguard_breakdance_4587_20260324_001209 ====================
\raisebox{-0.5\height}{\rotatebox{90}{\tiny Input}} &
\raisebox{-0.5\height}{\includegraphics[width=0.23\columnwidth, trim=0 0 0 0, clip]{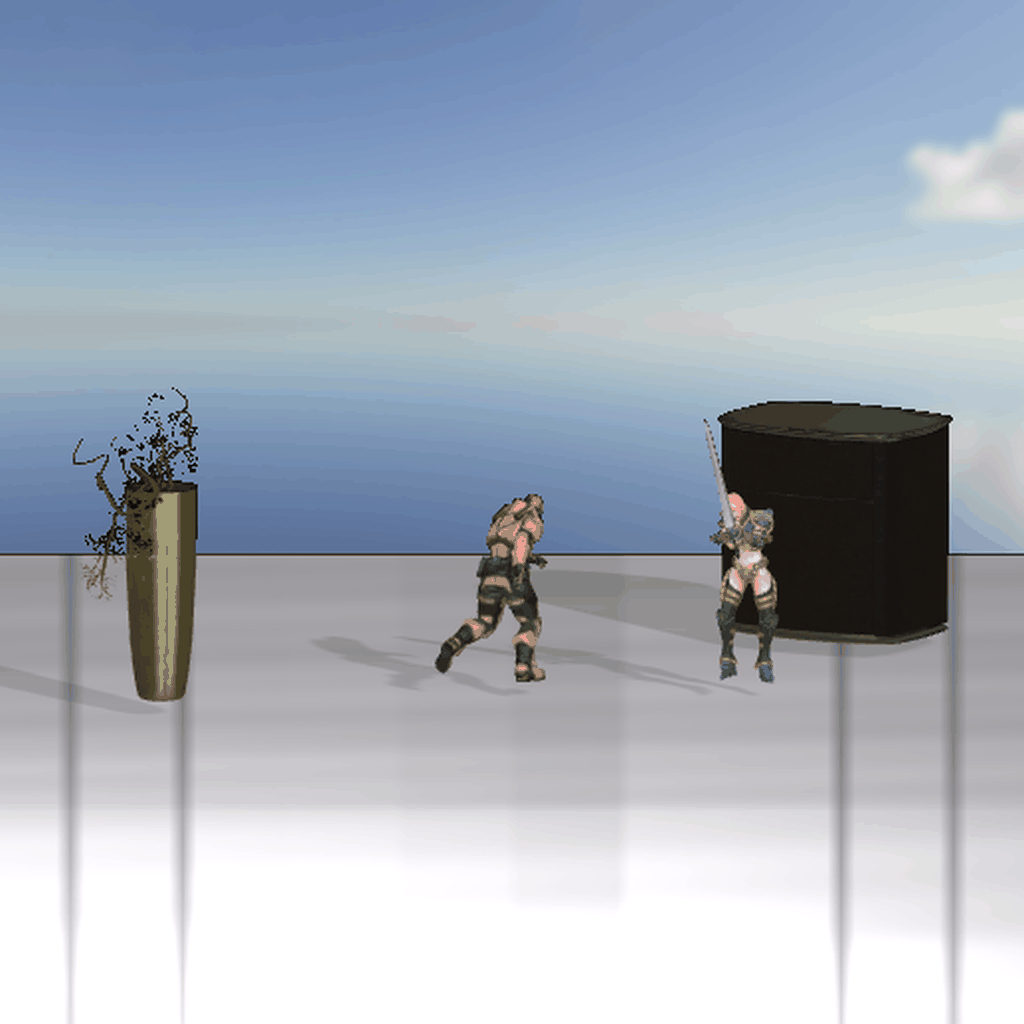}} &
\raisebox{-0.5\height}{\includegraphics[width=0.23\columnwidth, trim=0 0 0 0, clip]{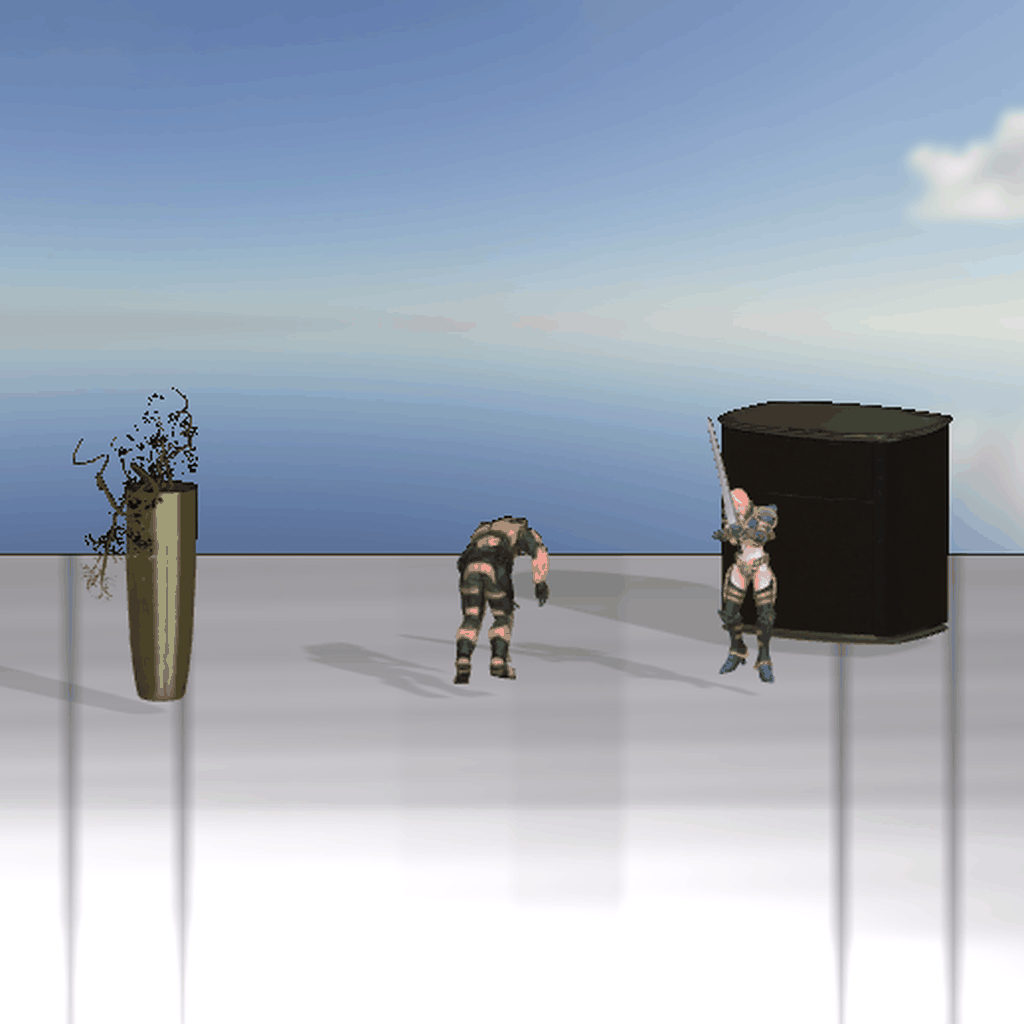}} &
\raisebox{-0.5\height}{\includegraphics[width=0.23\columnwidth, trim=0 0 0 0, clip]{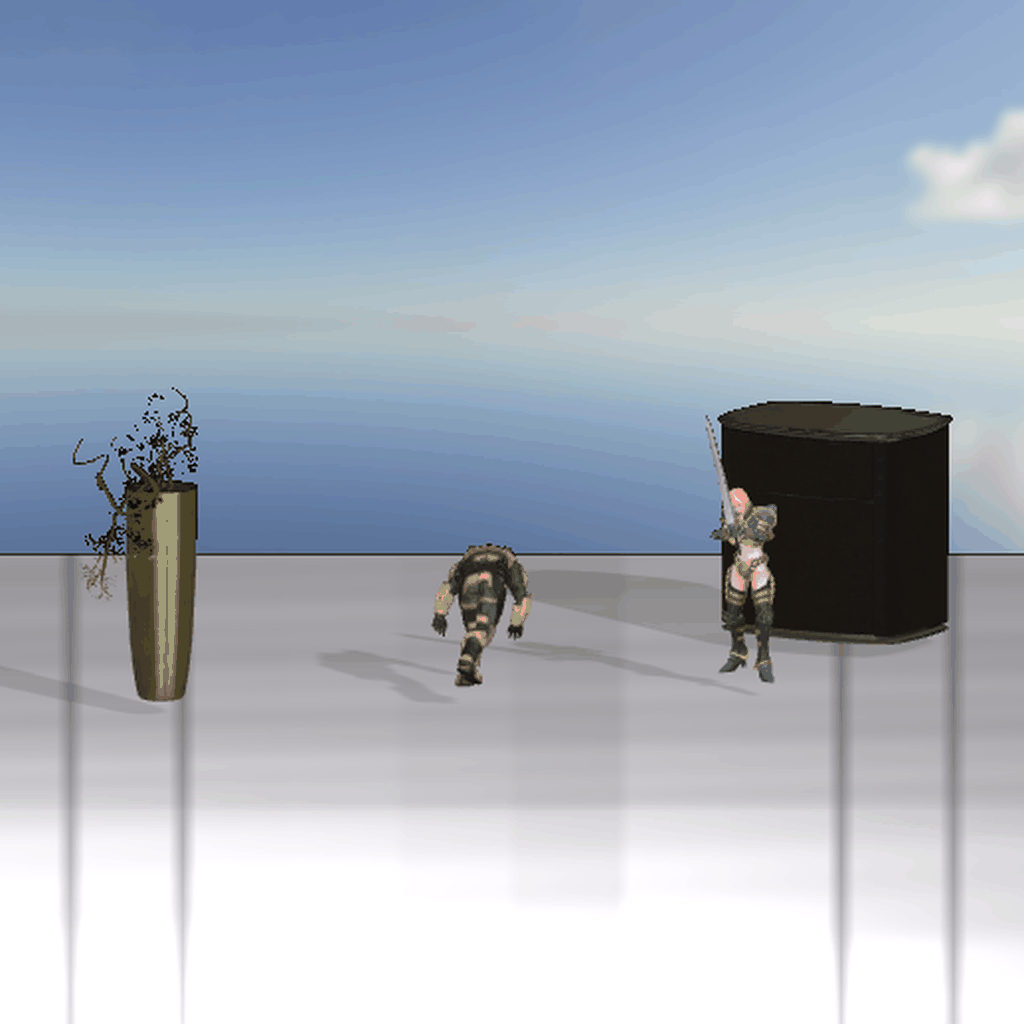}} &
\raisebox{-0.5\height}{\includegraphics[width=0.23\columnwidth, trim=0 0 0 0, clip]{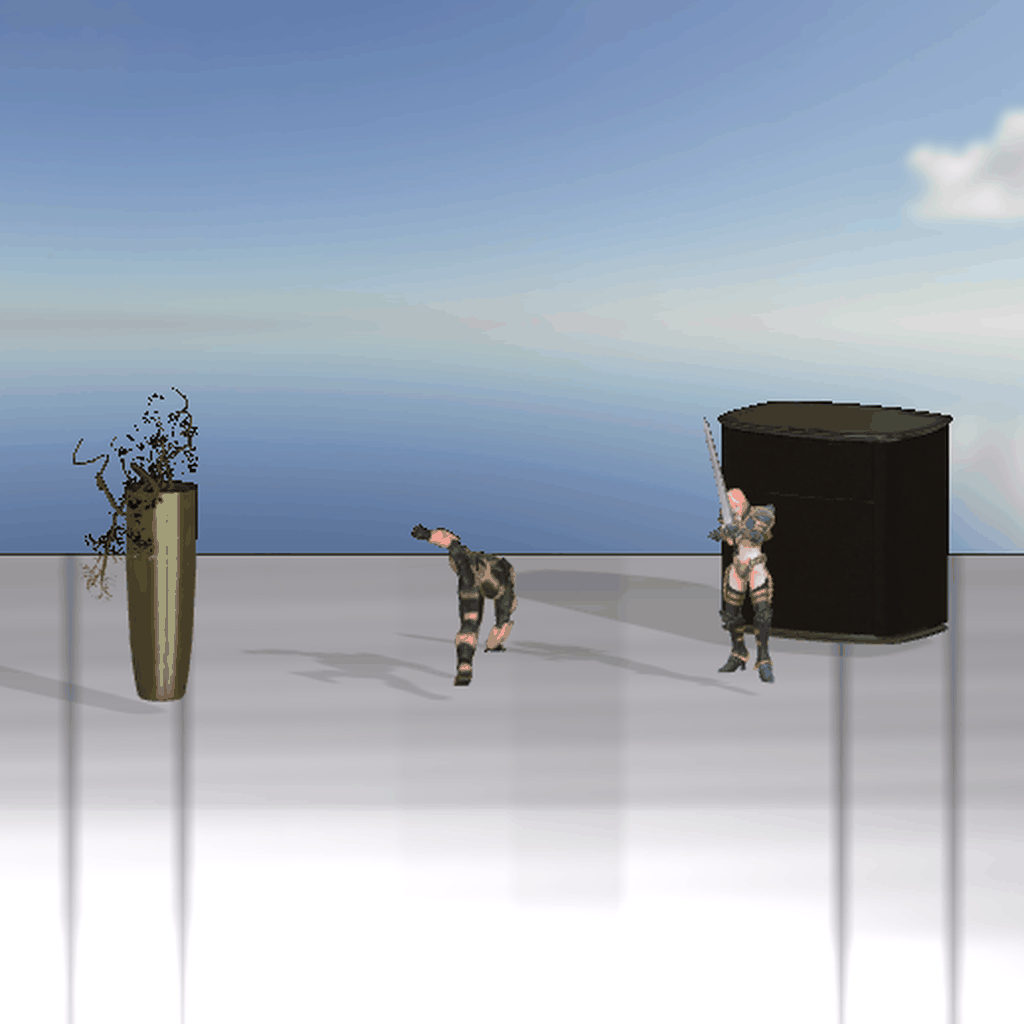}} \\[-0.5pt]
\raisebox{-0.5\height}{\rotatebox{90}{\tiny COM4D}} &
\raisebox{-0.5\height}{\includegraphics[width=0.23\columnwidth, trim=0 0 0 0, clip]{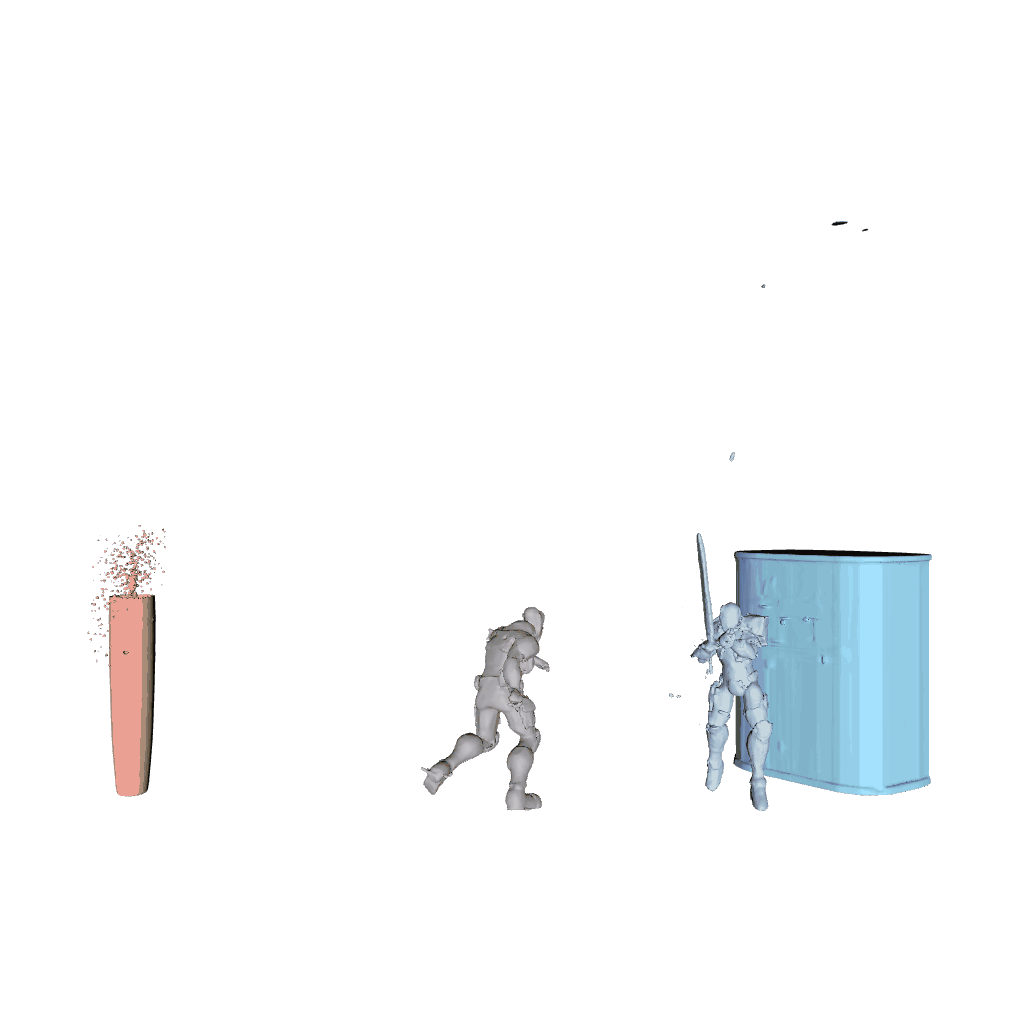}} &
\raisebox{-0.5\height}{\includegraphics[width=0.23\columnwidth, trim=0 0 0 0, clip]{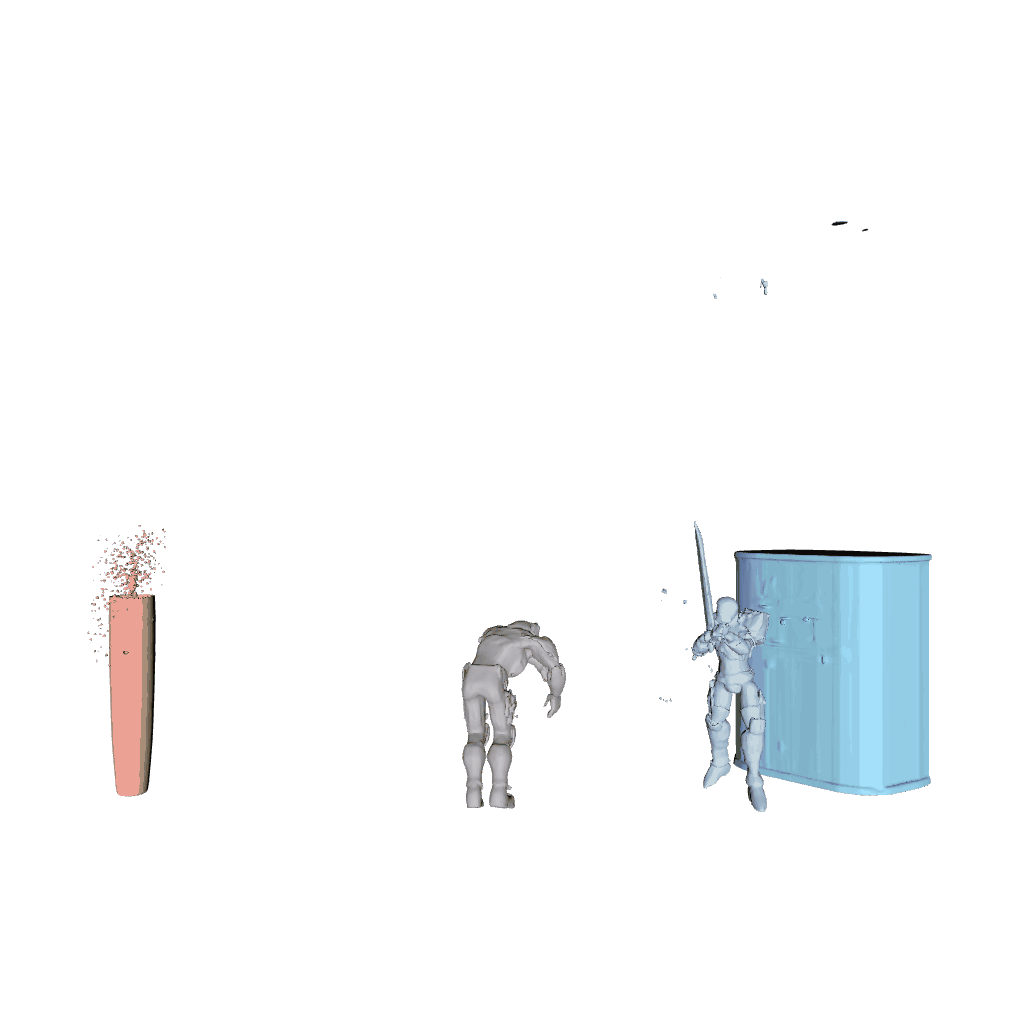}} &
\raisebox{-0.5\height}{\includegraphics[width=0.23\columnwidth, trim=0 0 0 0, clip]{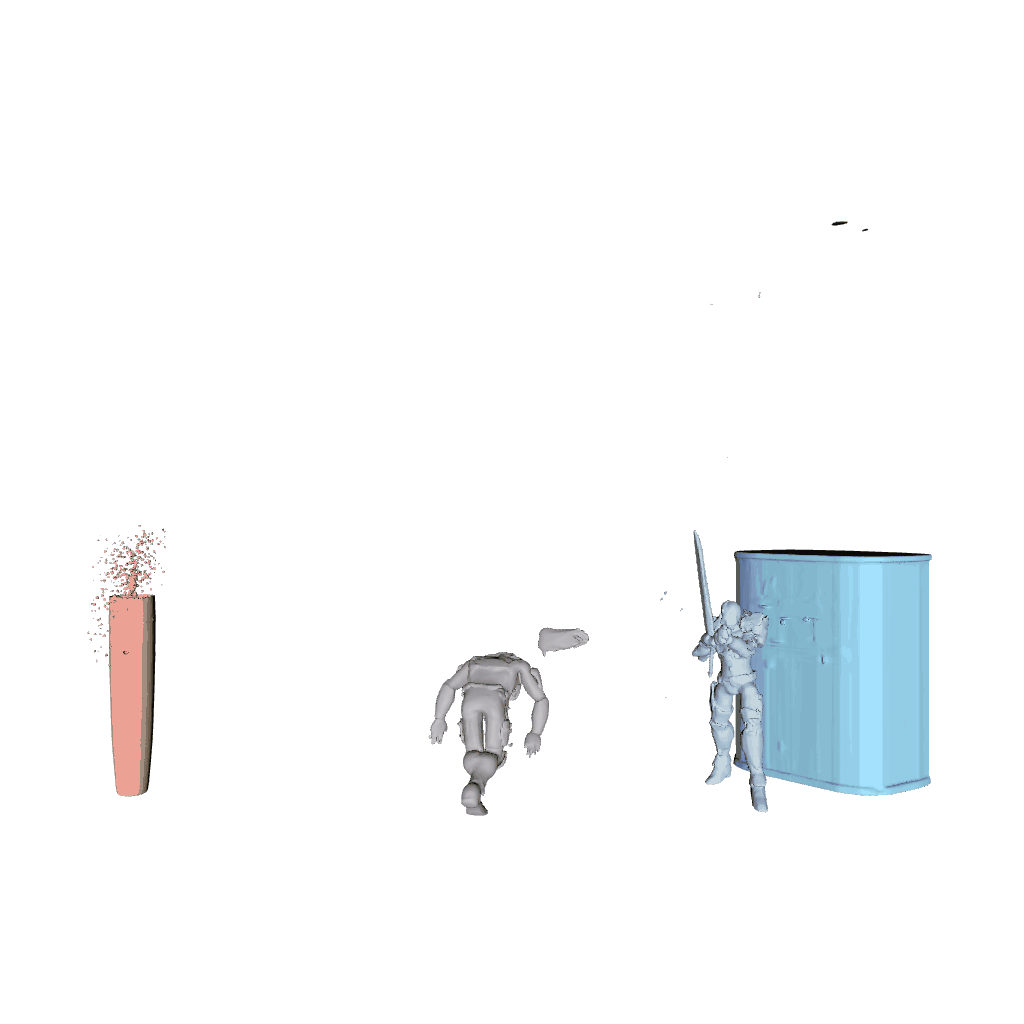}} &
\raisebox{-0.5\height}{\includegraphics[width=0.23\columnwidth, trim=0 0 0 0, clip]{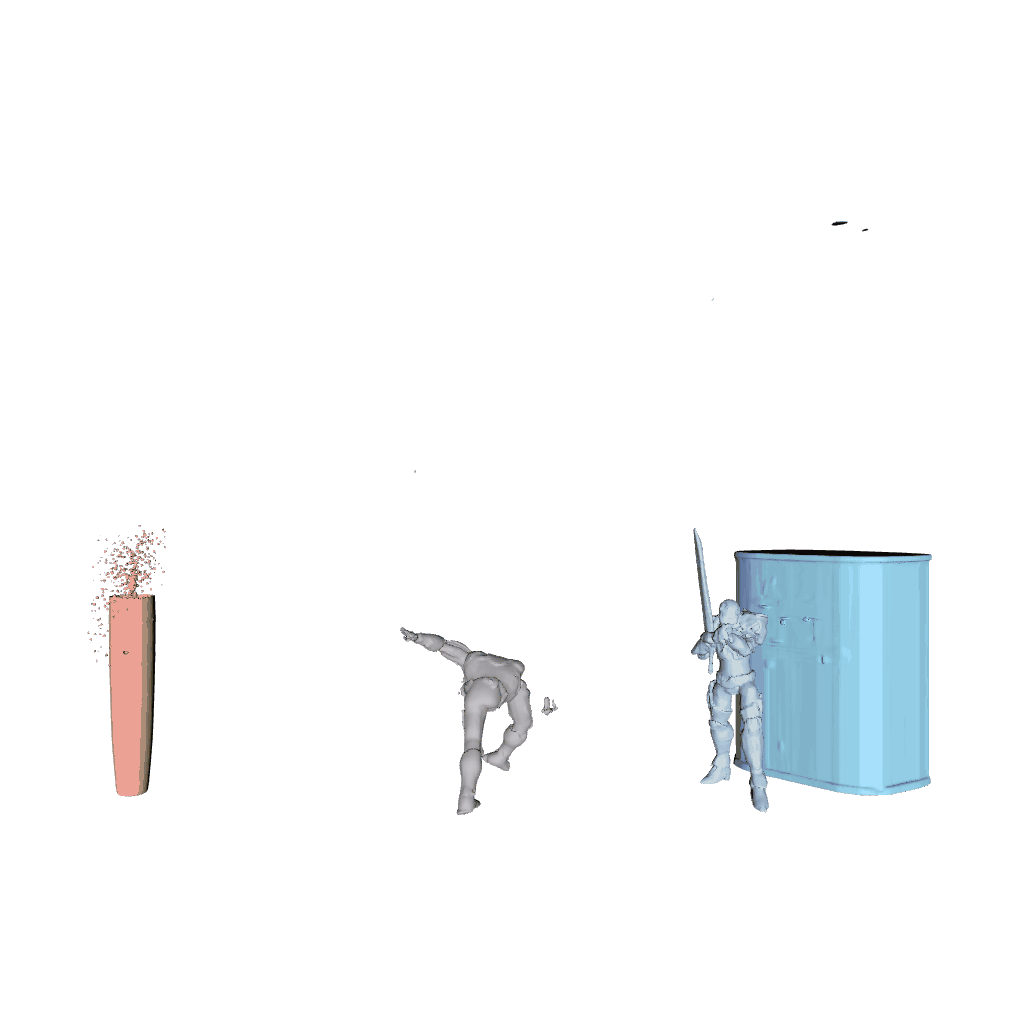}} \\[4pt]

% ==================== SAMPLE 3: markerman_JabCross__ninja_KettlebellSwing_5658_20260324_004510 ====================
\raisebox{-0.5\height}{\rotatebox{90}{\tiny Input}} &
\raisebox{-0.5\height}{\includegraphics[width=0.23\columnwidth, trim=0 0 0 0, clip]{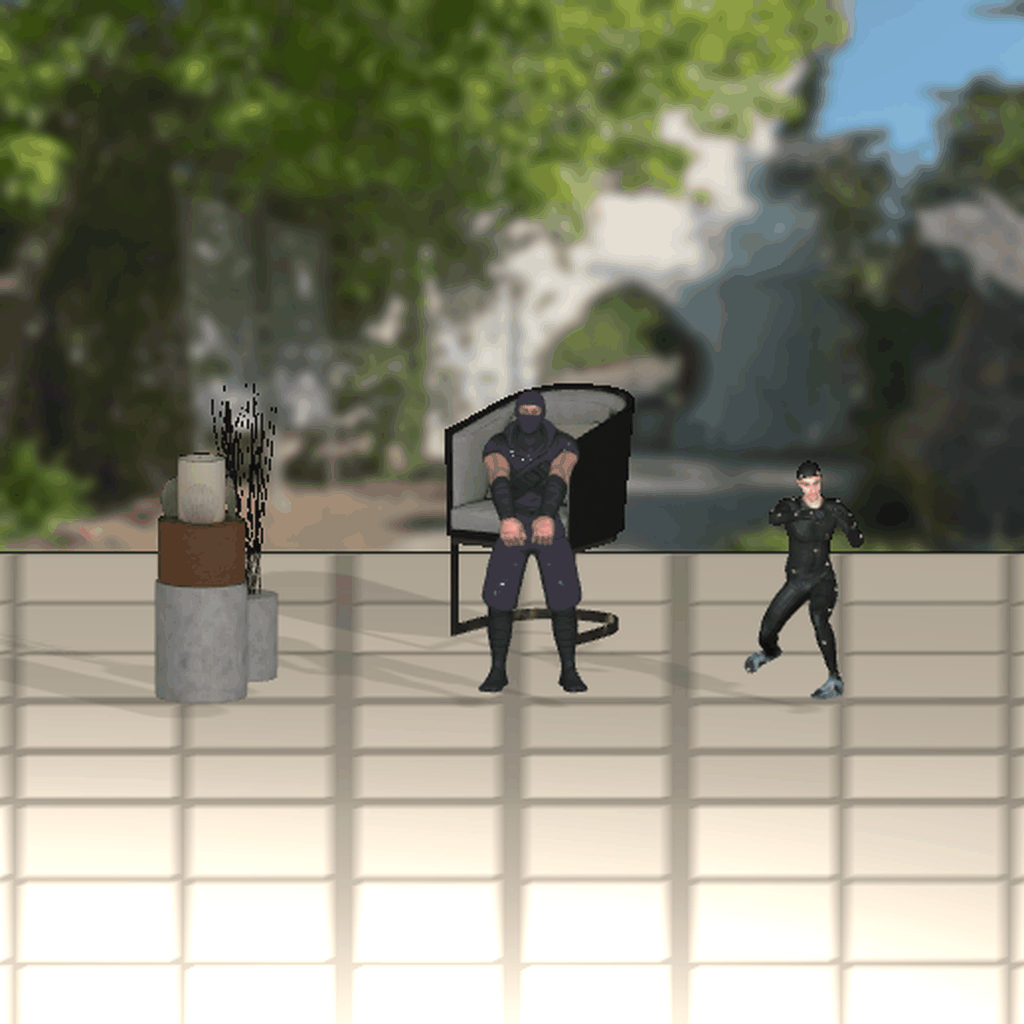}} &
\raisebox{-0.5\height}{\includegraphics[width=0.23\columnwidth, trim=0 0 0 0, clip]{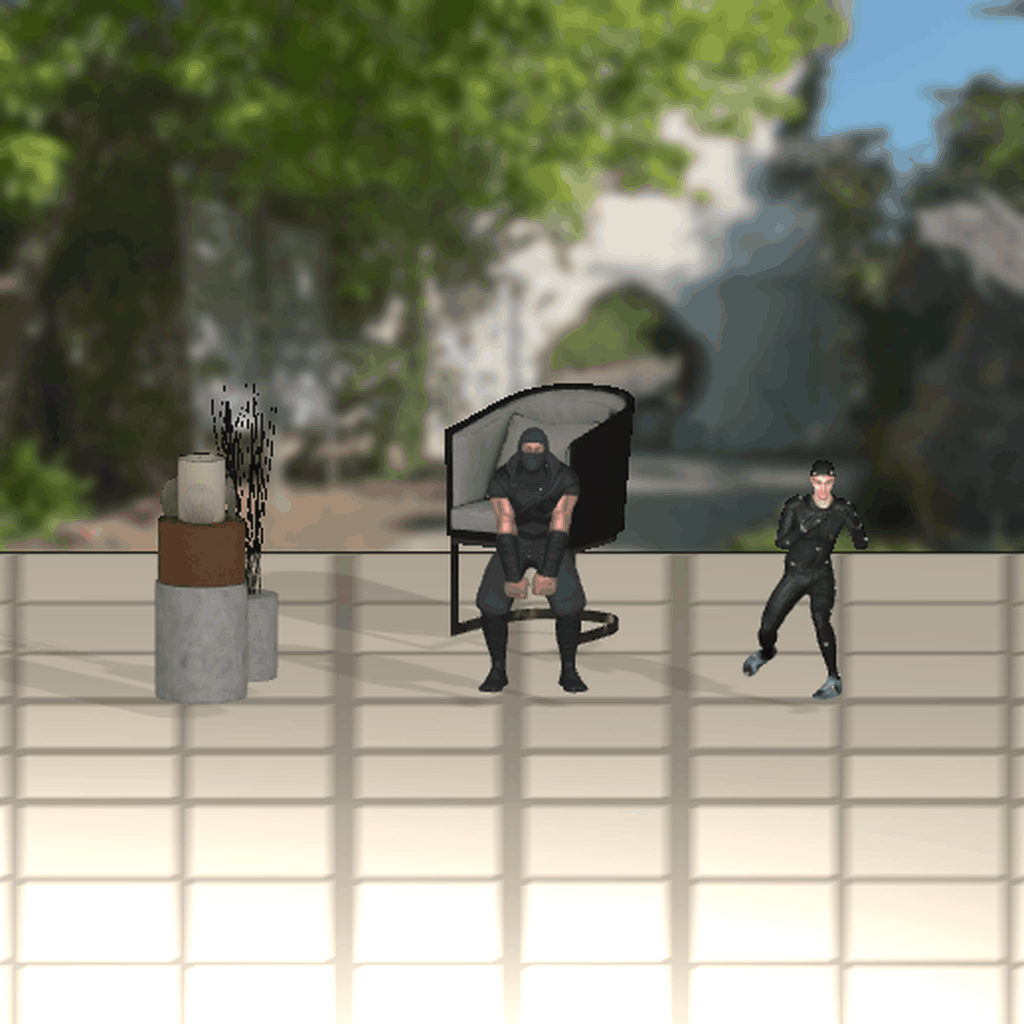}} &
\raisebox{-0.5\height}{\includegraphics[width=0.23\columnwidth, trim=0 0 0 0, clip]{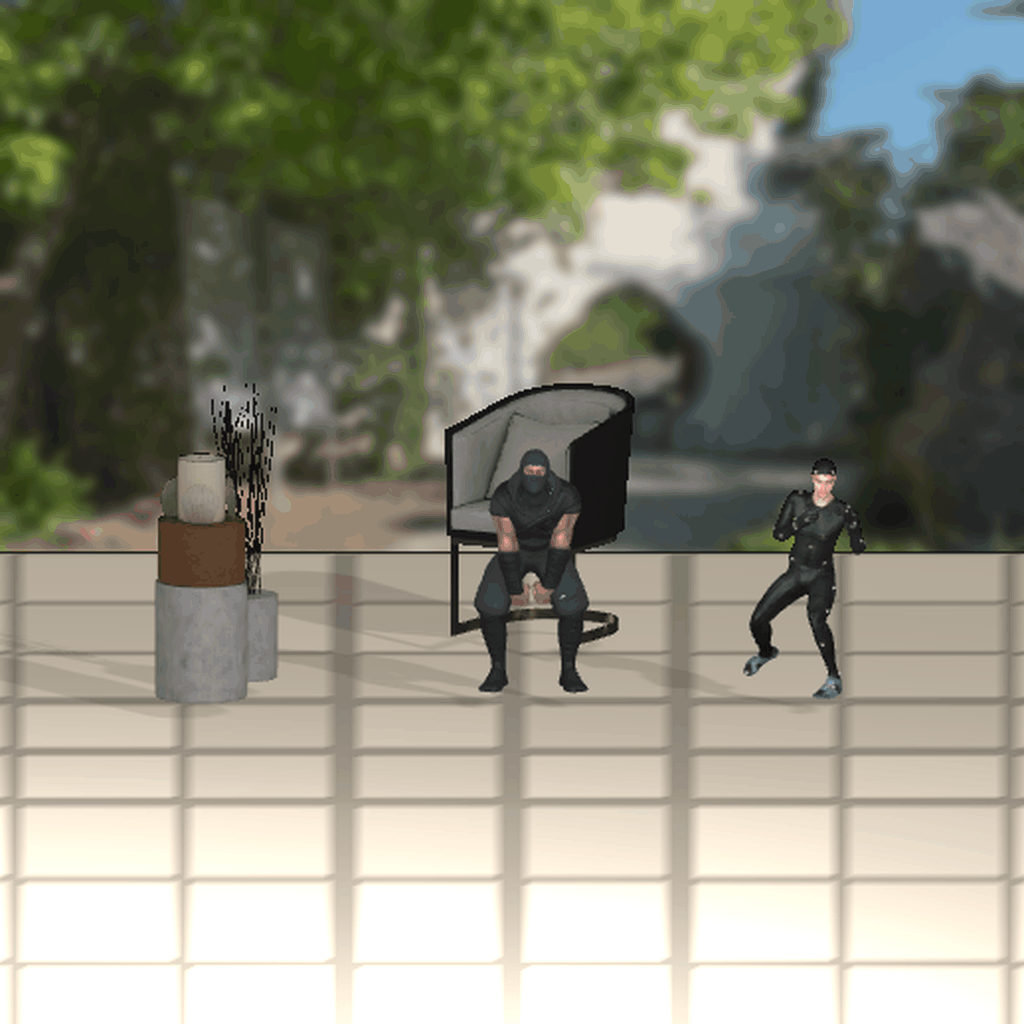}} &
\raisebox{-0.5\height}{\includegraphics[width=0.23\columnwidth, trim=0 0 0 0, clip]{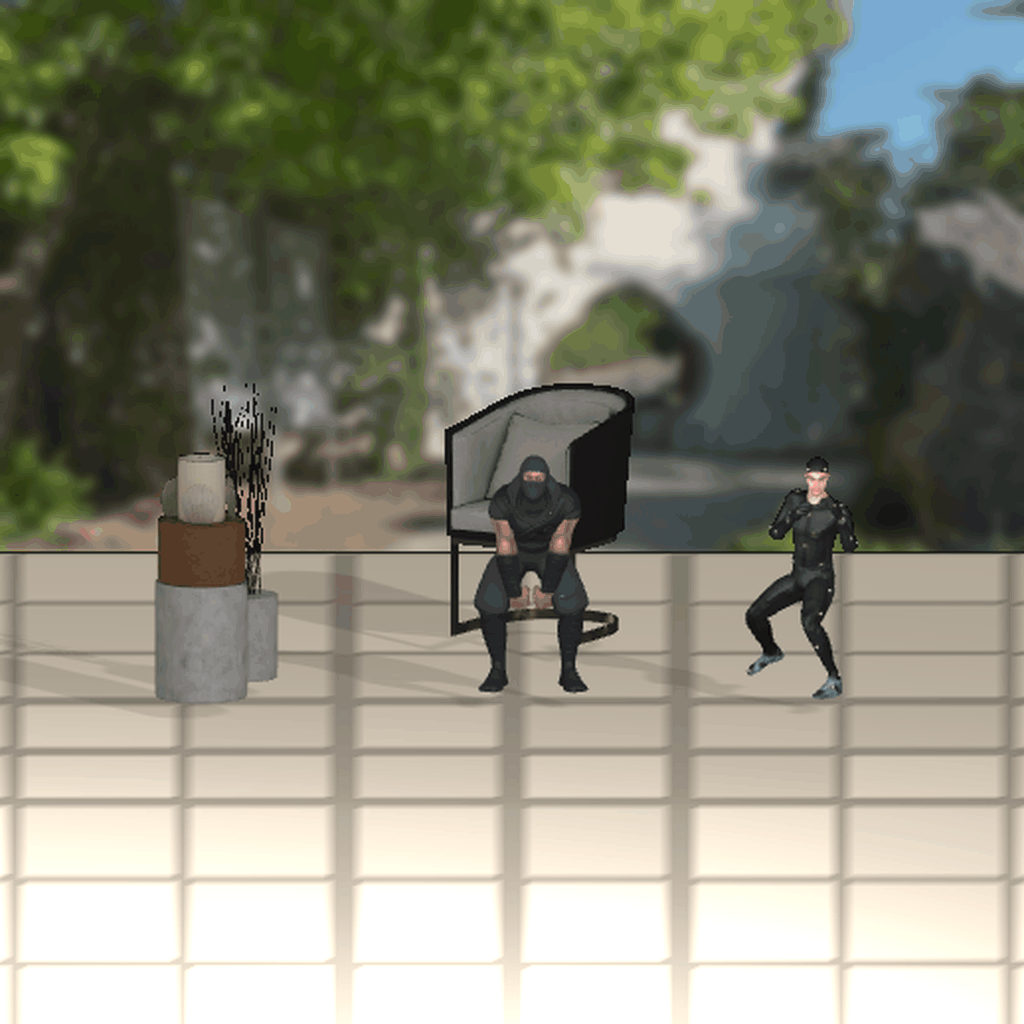}} \\[-0.5pt]
\raisebox{-0.5\height}{\rotatebox{90}{\tiny COM4D}} &
\raisebox{-0.5\height}{\includegraphics[width=0.23\columnwidth, trim=0 0 0 0, clip]{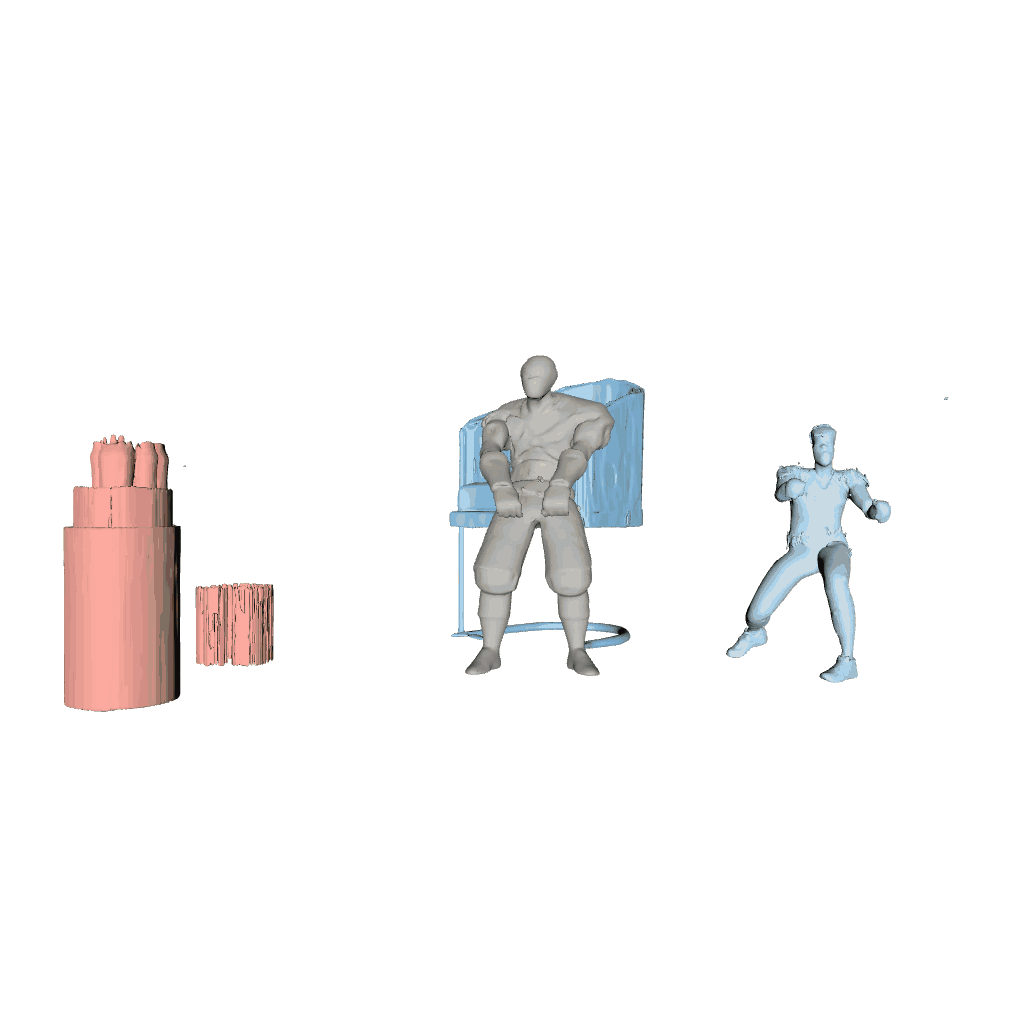}} &
\raisebox{-0.5\height}{\includegraphics[width=0.23\columnwidth, trim=0 0 0 0, clip]{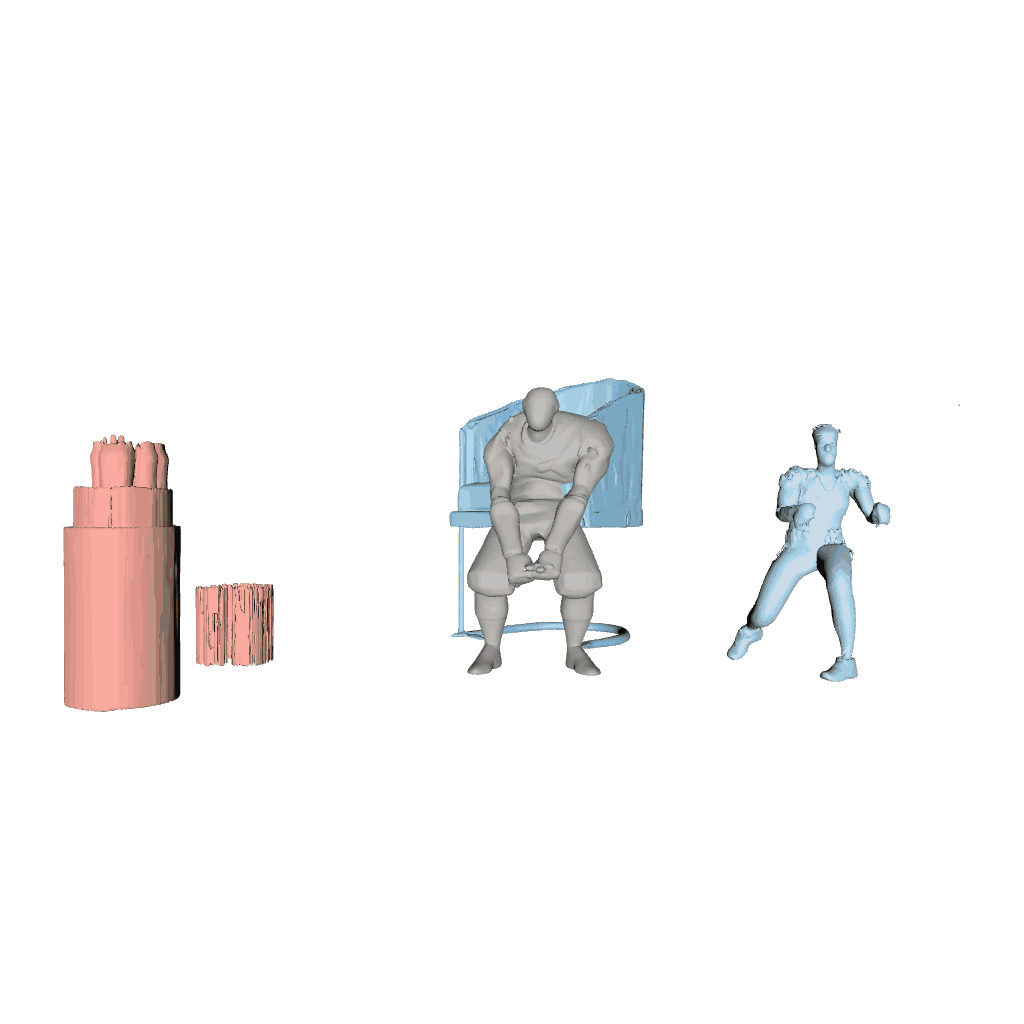}} &
\raisebox{-0.5\height}{\includegraphics[width=0.23\columnwidth, trim=0 0 0 0, clip]{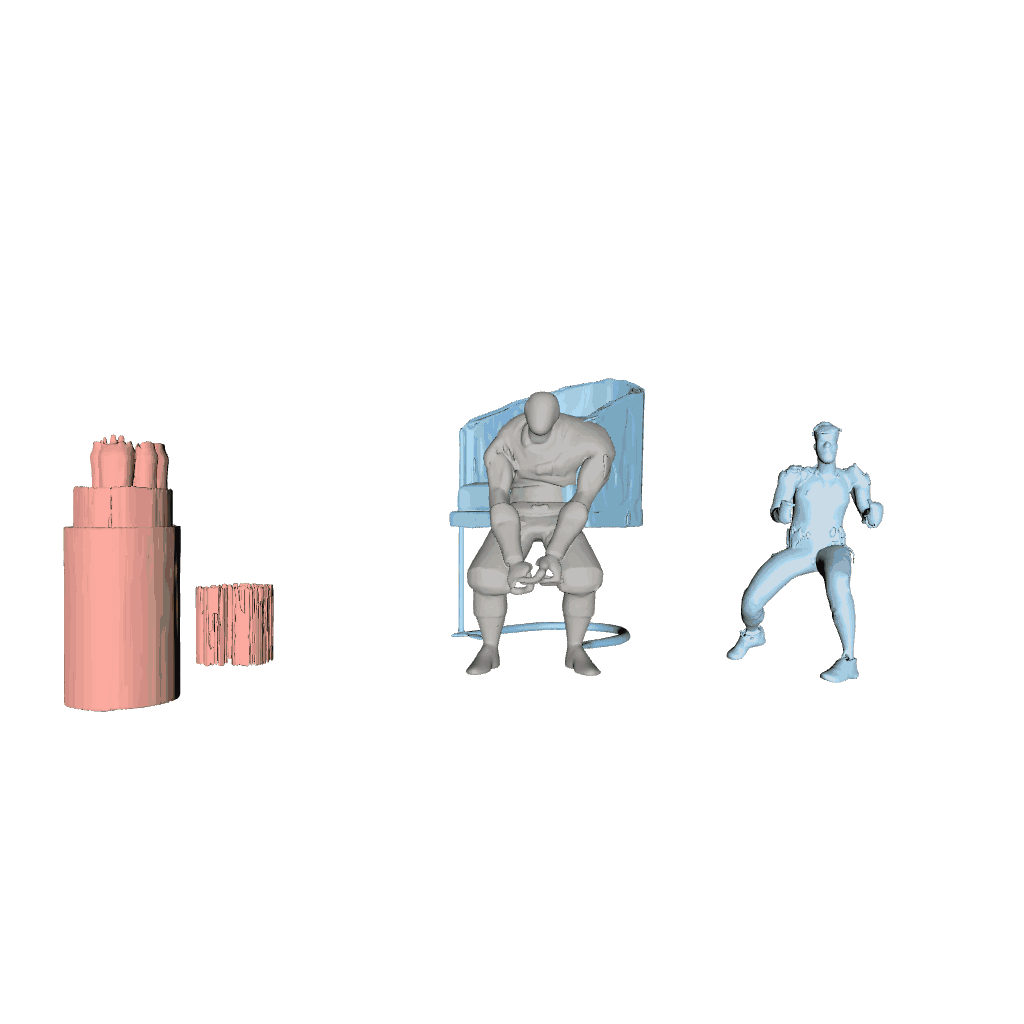}} &
\raisebox{-0.5\height}{\includegraphics[width=0.23\columnwidth, trim=0 0 0 0, clip]{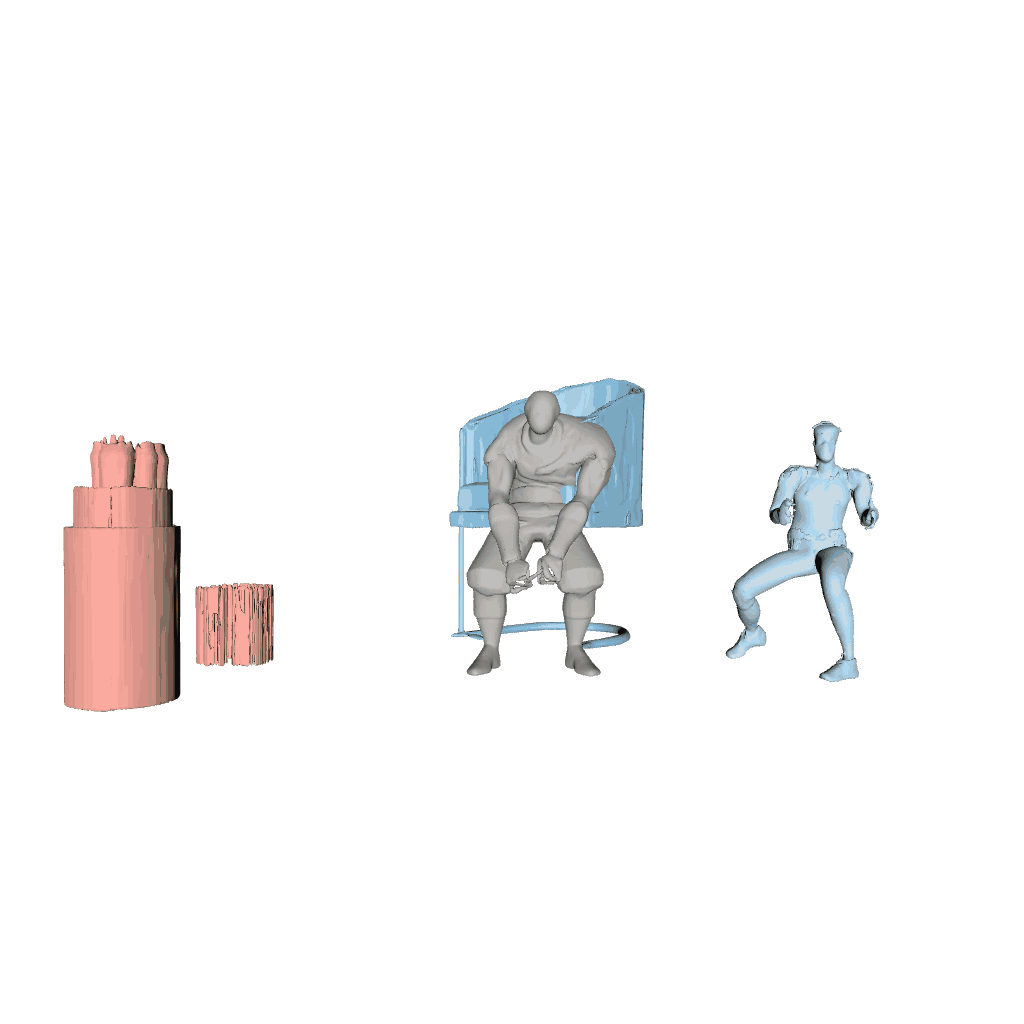}} \\[4pt]

% ==================== SAMPLE 4: prisoner_BboyUprockStart_20260324_011702 ====================
\raisebox{-0.5\height}{\rotatebox{90}{\tiny Input}} &
\raisebox{-0.5\height}{\includegraphics[width=0.23\columnwidth, trim=0 0 0 0, clip]{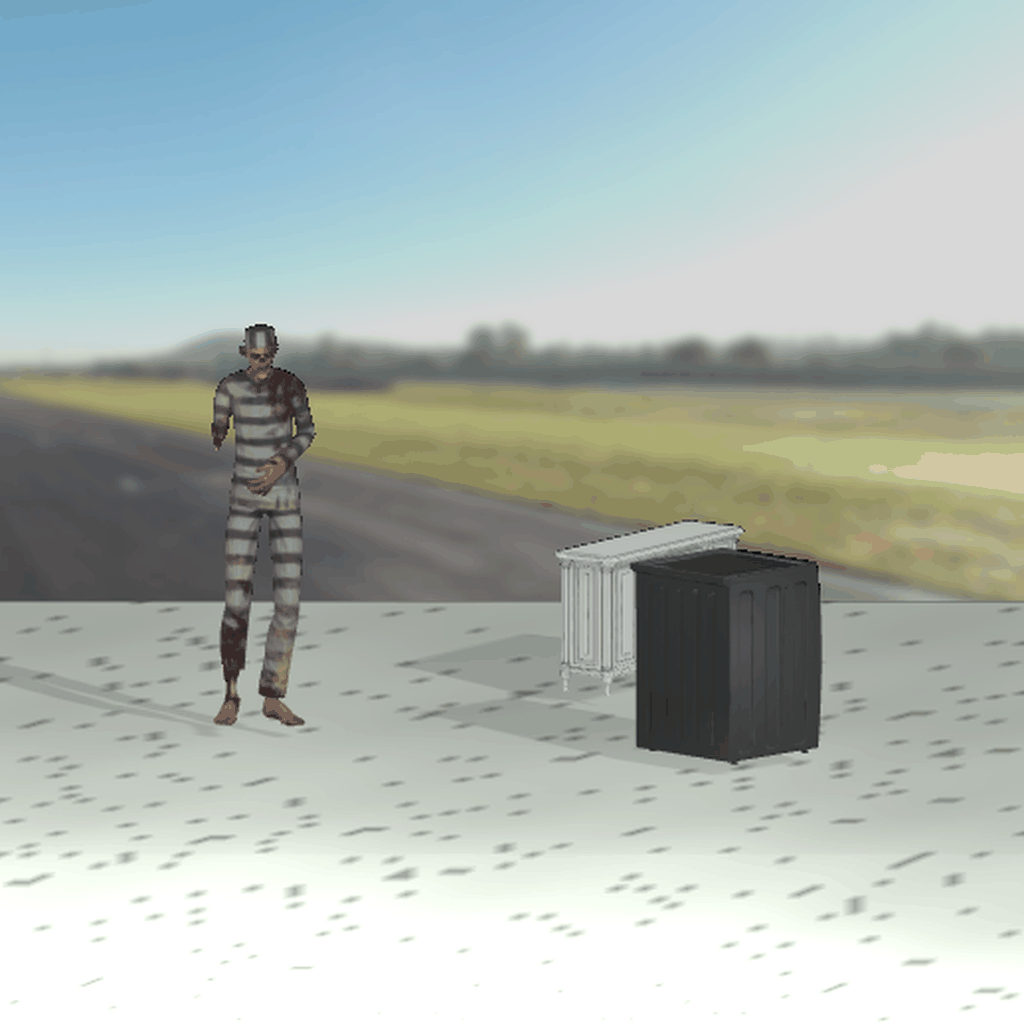}} &
\raisebox{-0.5\height}{\includegraphics[width=0.23\columnwidth, trim=0 0 0 0, clip]{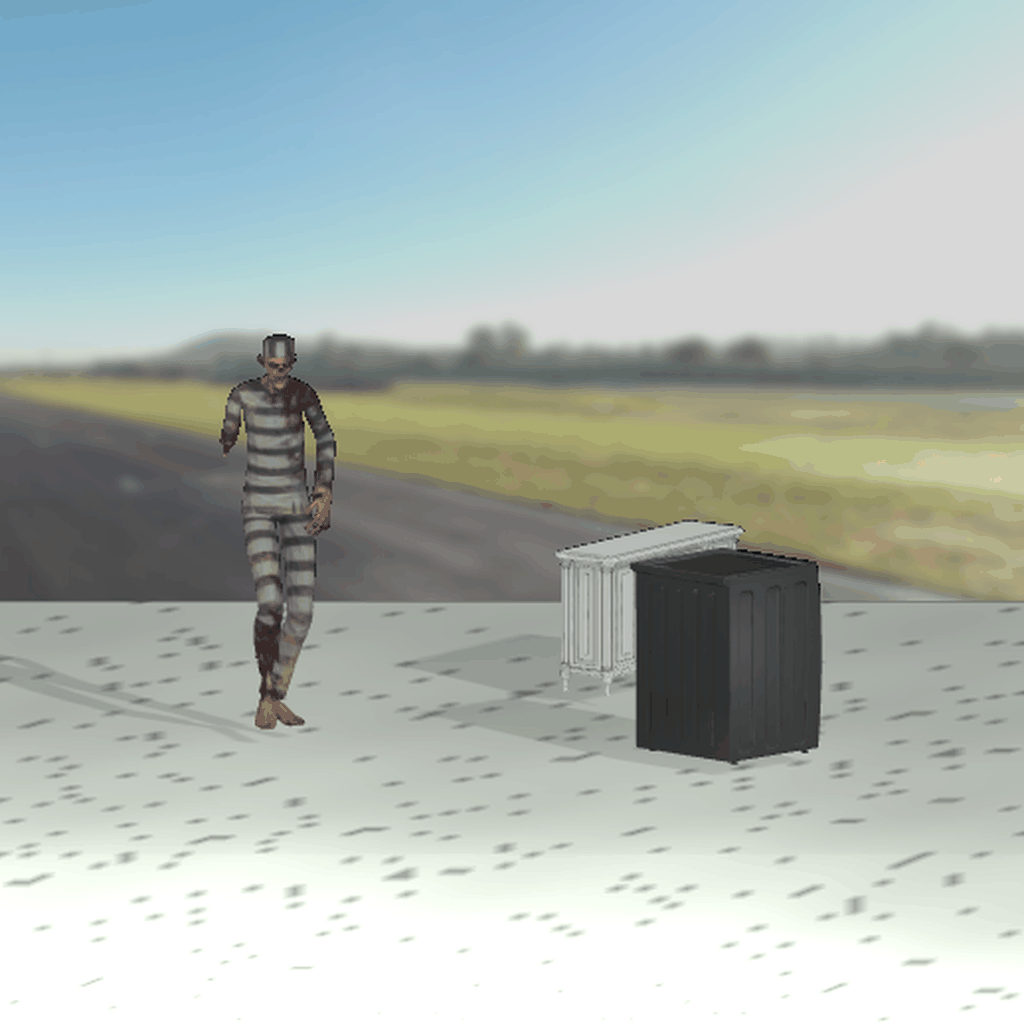}} &
\raisebox{-0.5\height}{\includegraphics[width=0.23\columnwidth, trim=0 0 0 0, clip]{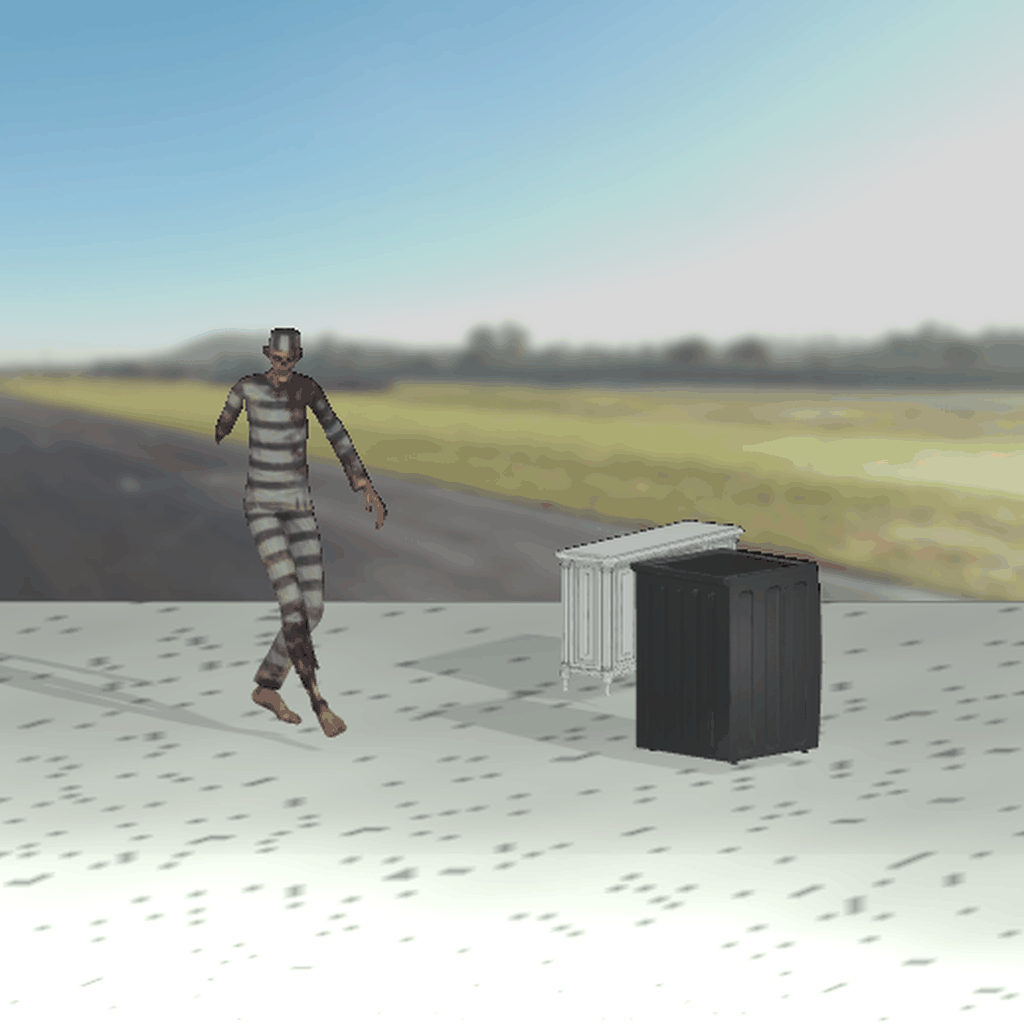}} &
\raisebox{-0.5\height}{\includegraphics[width=0.23\columnwidth, trim=0 0 0 0, clip]{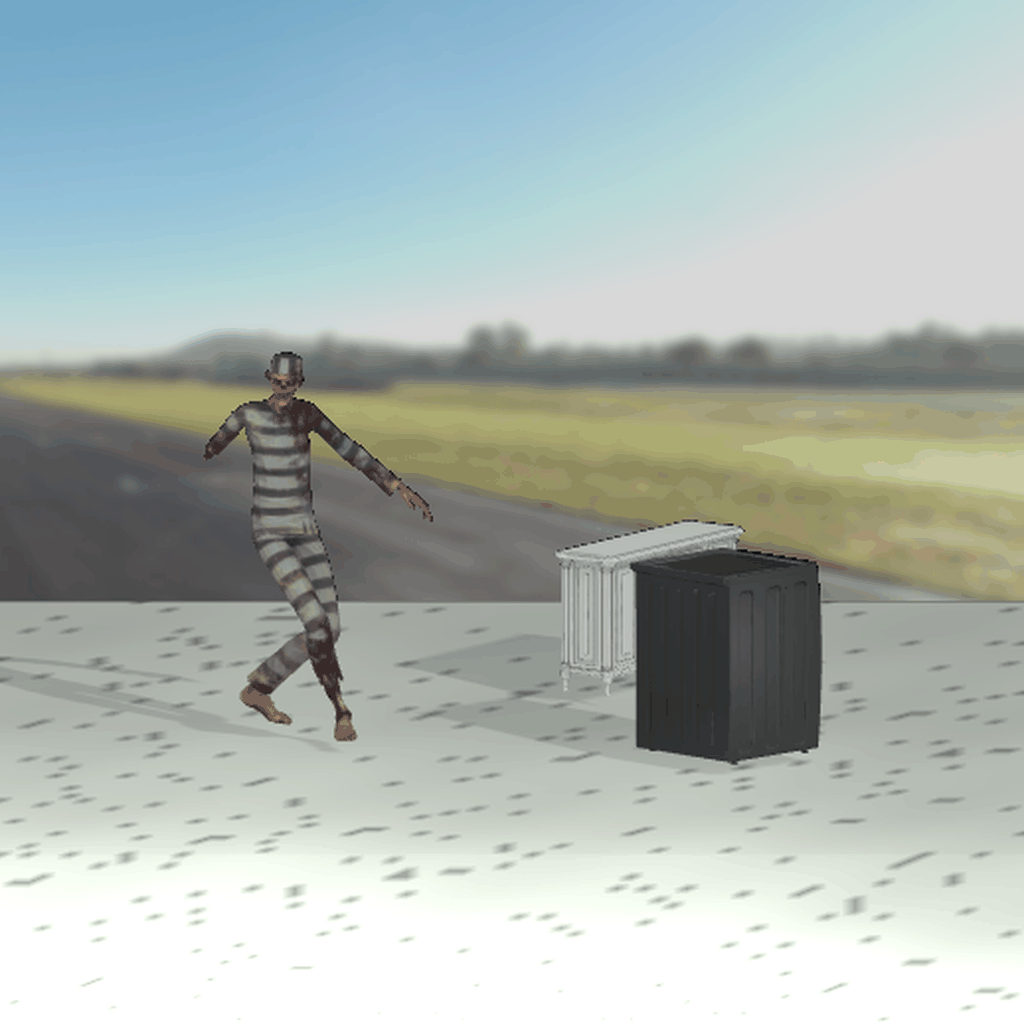}} \\[-0.5pt]
\raisebox{-0.5\height}{\rotatebox{90}{\tiny COM4D}} &
\raisebox{-0.5\height}{\includegraphics[width=0.23\columnwidth, trim=0 0 0 0, clip]{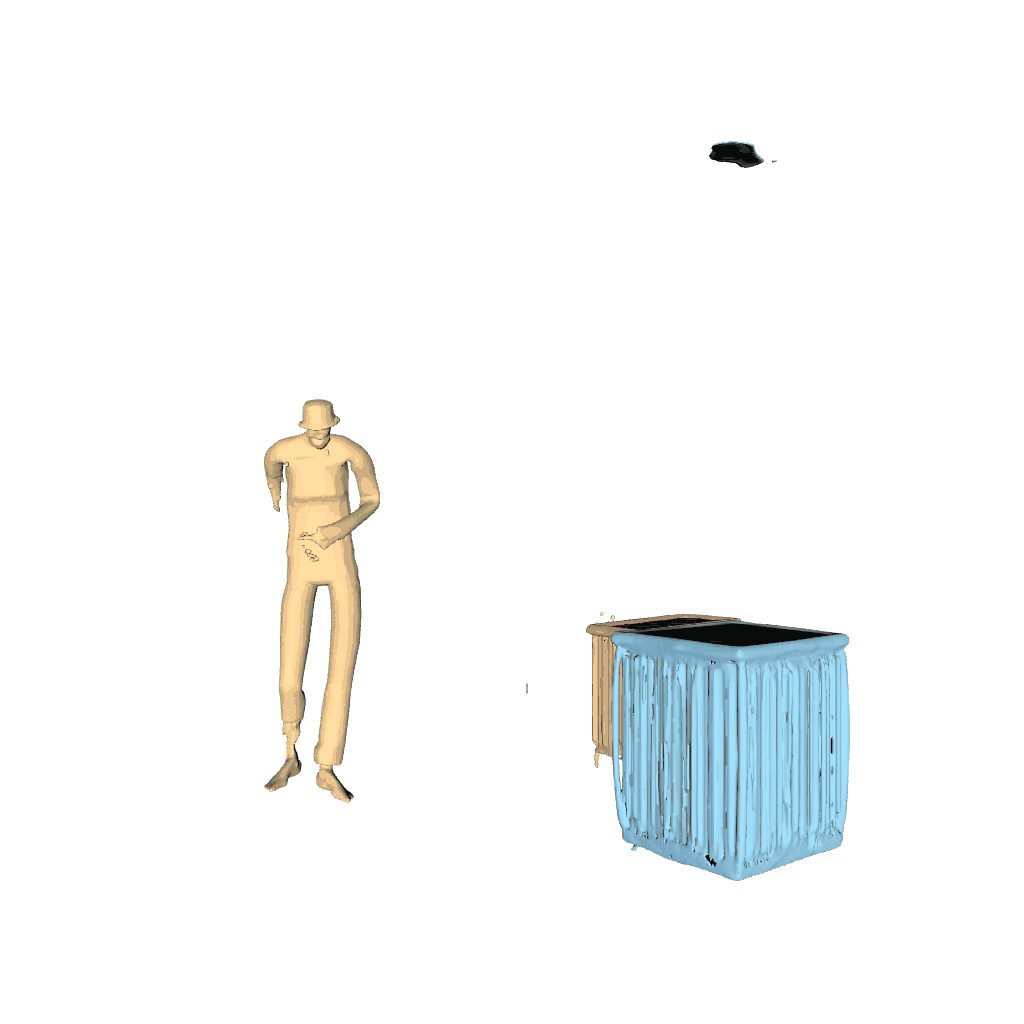}} &
\raisebox{-0.5\height}{\includegraphics[width=0.23\columnwidth, trim=0 0 0 0, clip]{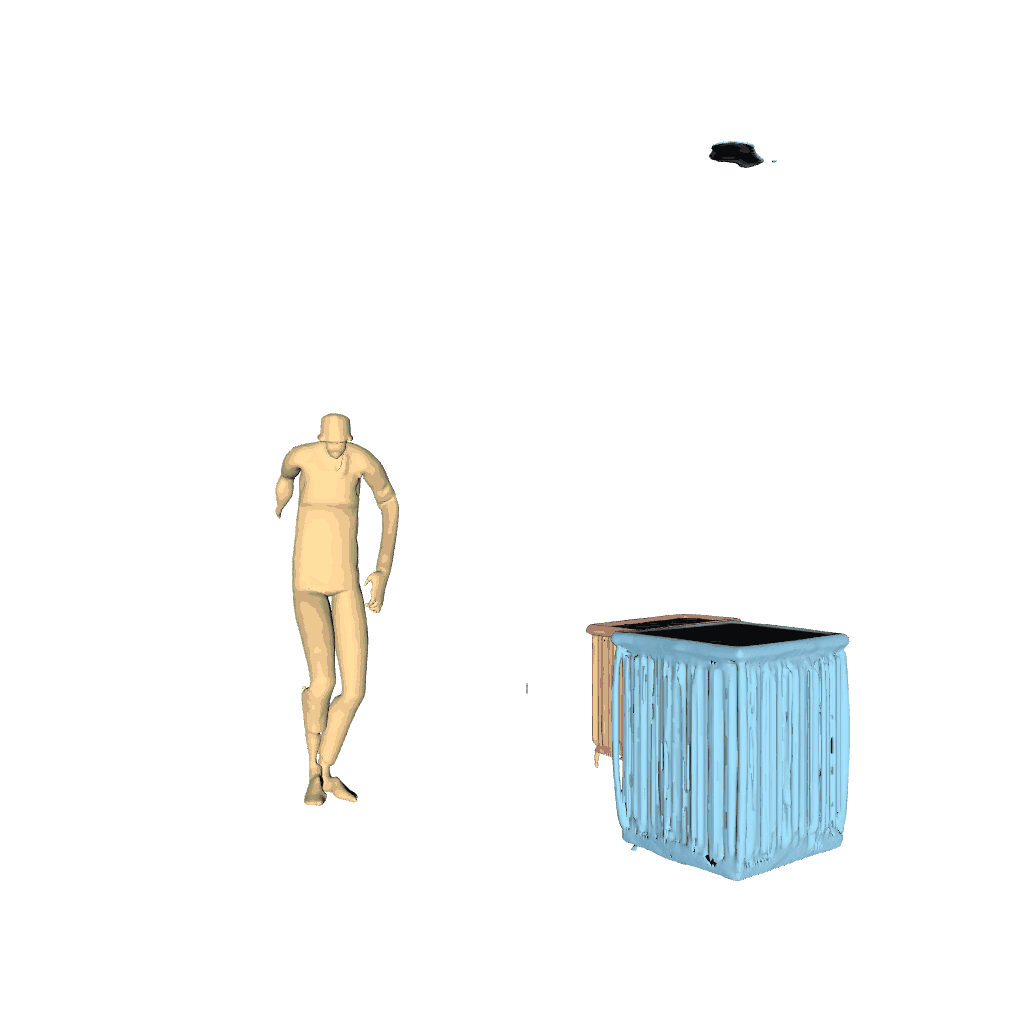}} &
\raisebox{-0.5\height}{\includegraphics[width=0.23\columnwidth, trim=0 0 0 0, clip]{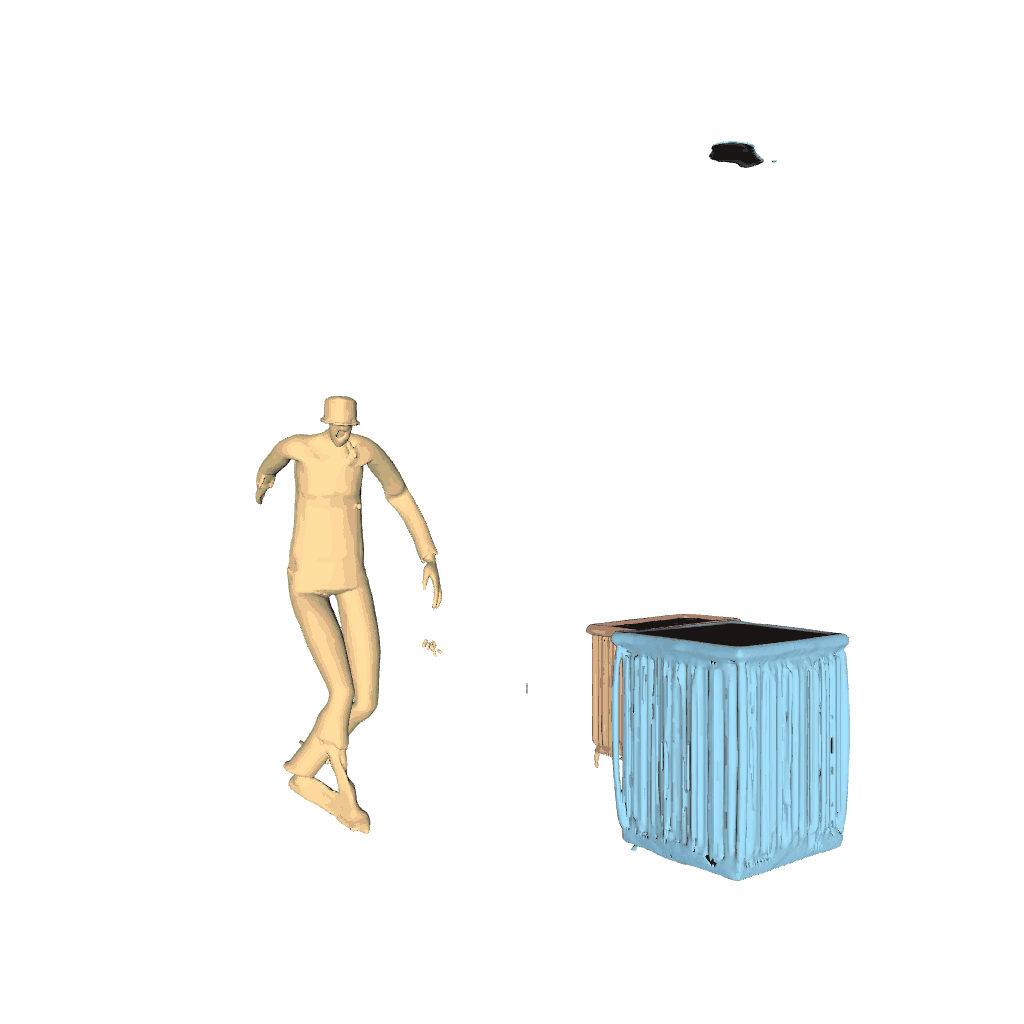}} &
\raisebox{-0.5\height}{\includegraphics[width=0.23\columnwidth, trim=0 0 0 0, clip]{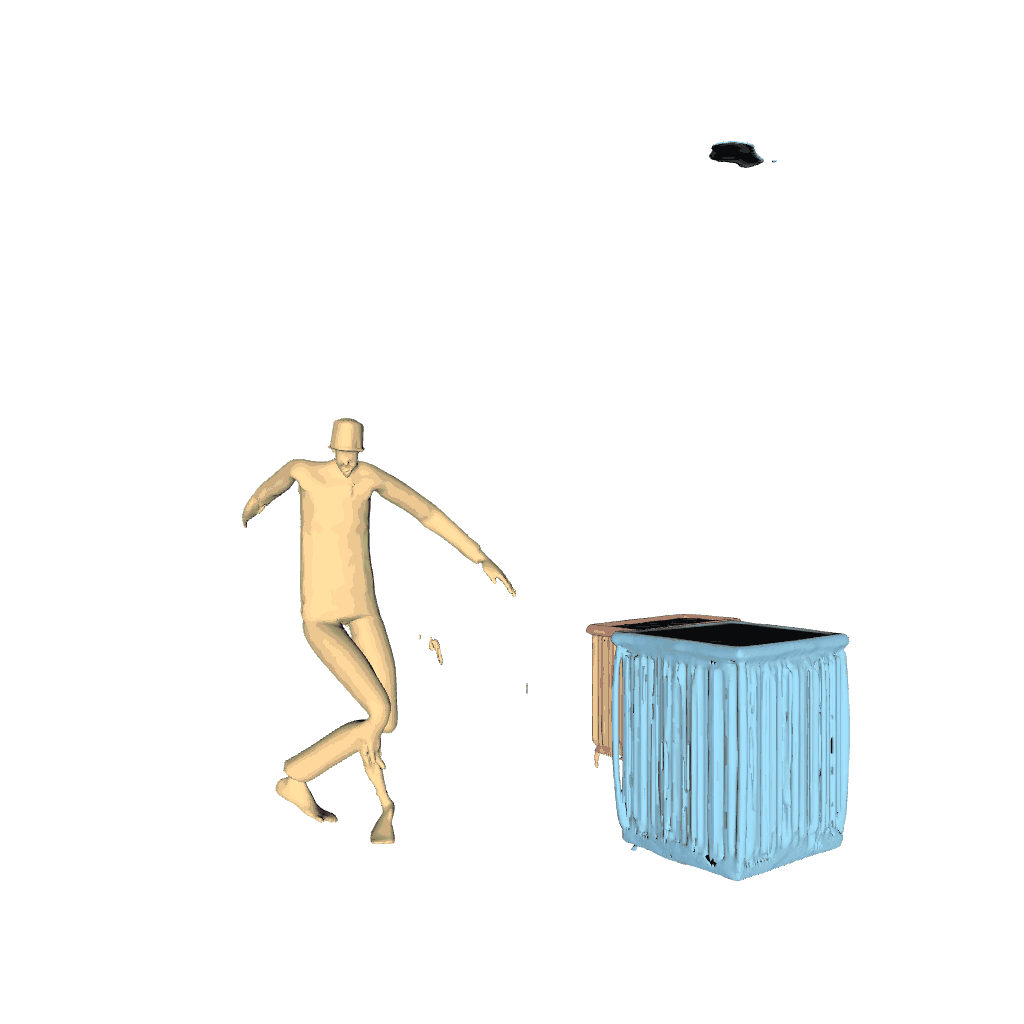}} \\

  \end{tabular}

  \vspace{-3pt}
  \caption{
  Additional qualitative results on the synthetic compositional 4D dataset.
  }
  \label{fig:synthetic_com4d_appendix_qual_3}
\end{figure}

\begin{figure}[h]
    \section{Additional Qualitative Results on 4D Object Reconstruction}
    In ~\cref{fig:qual_comp_4d_appendix_3_subjects}, ~\cref{fig:qual_comp_4d_appendix_5_6_full} and ~\cref{fig:qual_comp_4d_appendix_other} we show more qualitative results of our model and baselines on single object 4D object reconstruction. \\
  
  \centering
  % --- MODIFICATIONS ---
  \setlength{\tabcolsep}{0pt} % Zero horizontal spacing between images
  \renewcommand{\arraystretch}{0} % Remove default extra padding in rows

  % ==================== TABLE START ====================
  \begin{tabular}{@{}c@{\hspace{2pt}}cccccc@{}}
    % --- HEADERS ---
    & % Empty cell for the vertical text column
    \small{Input / GT} & 
    \textbf{\small{Ours}} &
    \small{TripoSG} &
    \small{V2M4} &
    \small{GVFD} &
    \small{L4GM} \\[3pt] 

% ==================== SUBJECT 1: 1eca489f... ====================
% -------- Frame 0 --------
\raisebox{-0.5\height}{\rotatebox{90}{\tiny Input $\angle 0^\circ$}} &
\raisebox{-0.5\height}{\includegraphics[width=0.16\columnwidth, trim=40 40 40 40, clip]{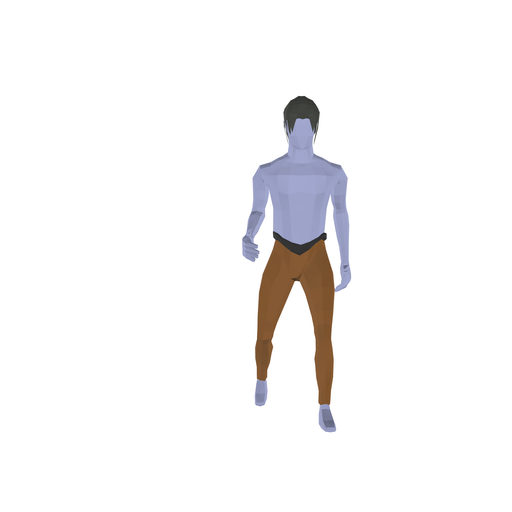}} &
\raisebox{-0.5\height}{\includegraphics[width=0.16\columnwidth, trim=20 20 20 20, clip]{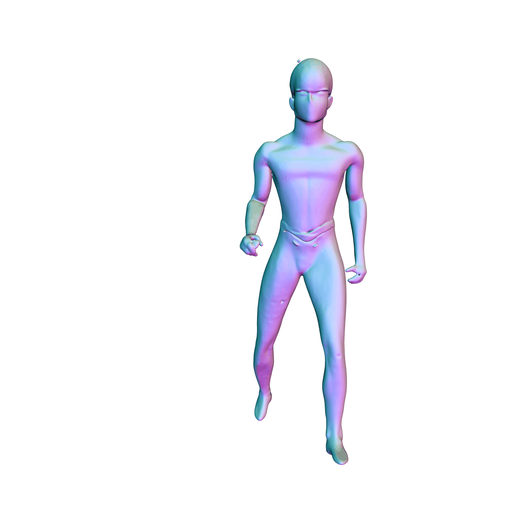}} &
\raisebox{-0.5\height}{\includegraphics[width=0.16\columnwidth, trim=40 40 40 40, clip]{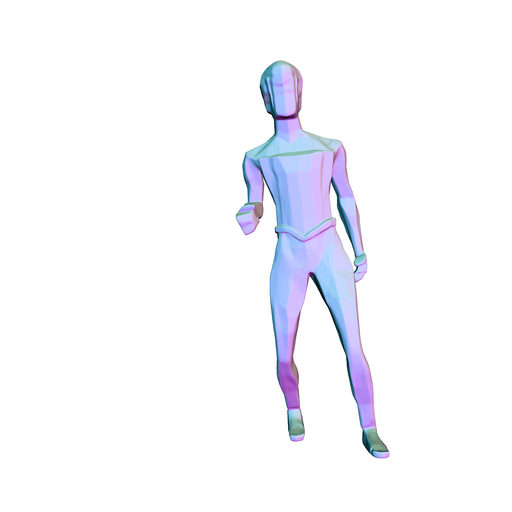}} &
\raisebox{-0.5\height}{\includegraphics[width=0.16\columnwidth, trim=40 40 40 40, clip]{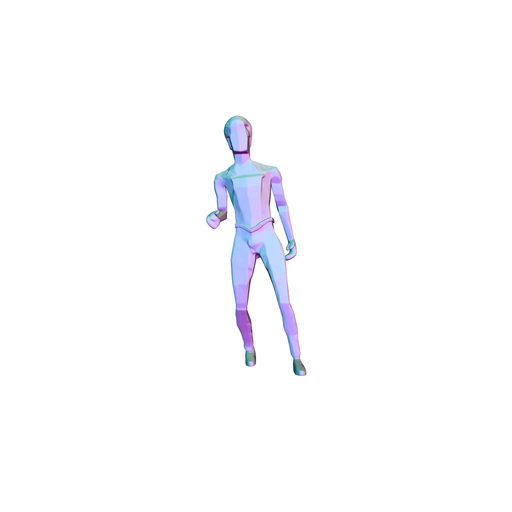}} &
\raisebox{-0.5\height}{\includegraphics[width=0.16\columnwidth, trim=40 40 40 40, clip]{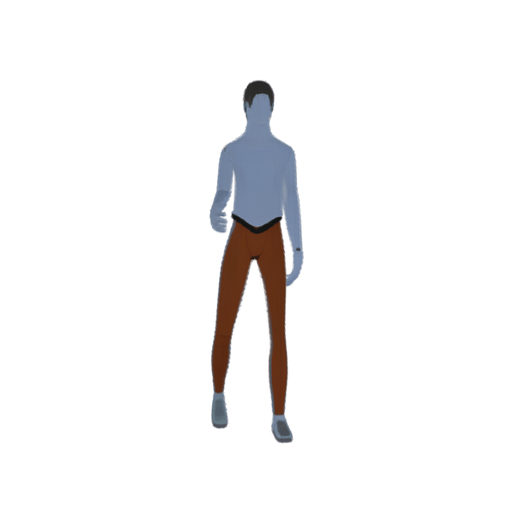}} &
\raisebox{-0.5\height}{\includegraphics[width=0.16\columnwidth, trim=40 40 40 40, clip]{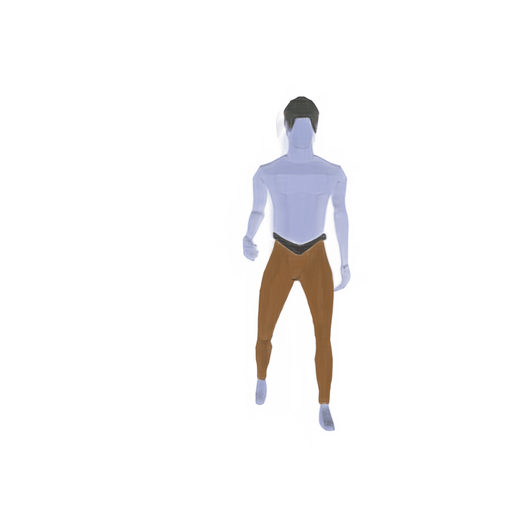}} \\[-0.5pt]
% --- New GT View Frame 0 ---
\raisebox{-0.5\height}{\rotatebox{90}{\tiny render $\angle 90^\circ$}} &
\raisebox{-0.5\height}{\includegraphics[width=0.16\columnwidth, trim=40 40 40 40, clip]{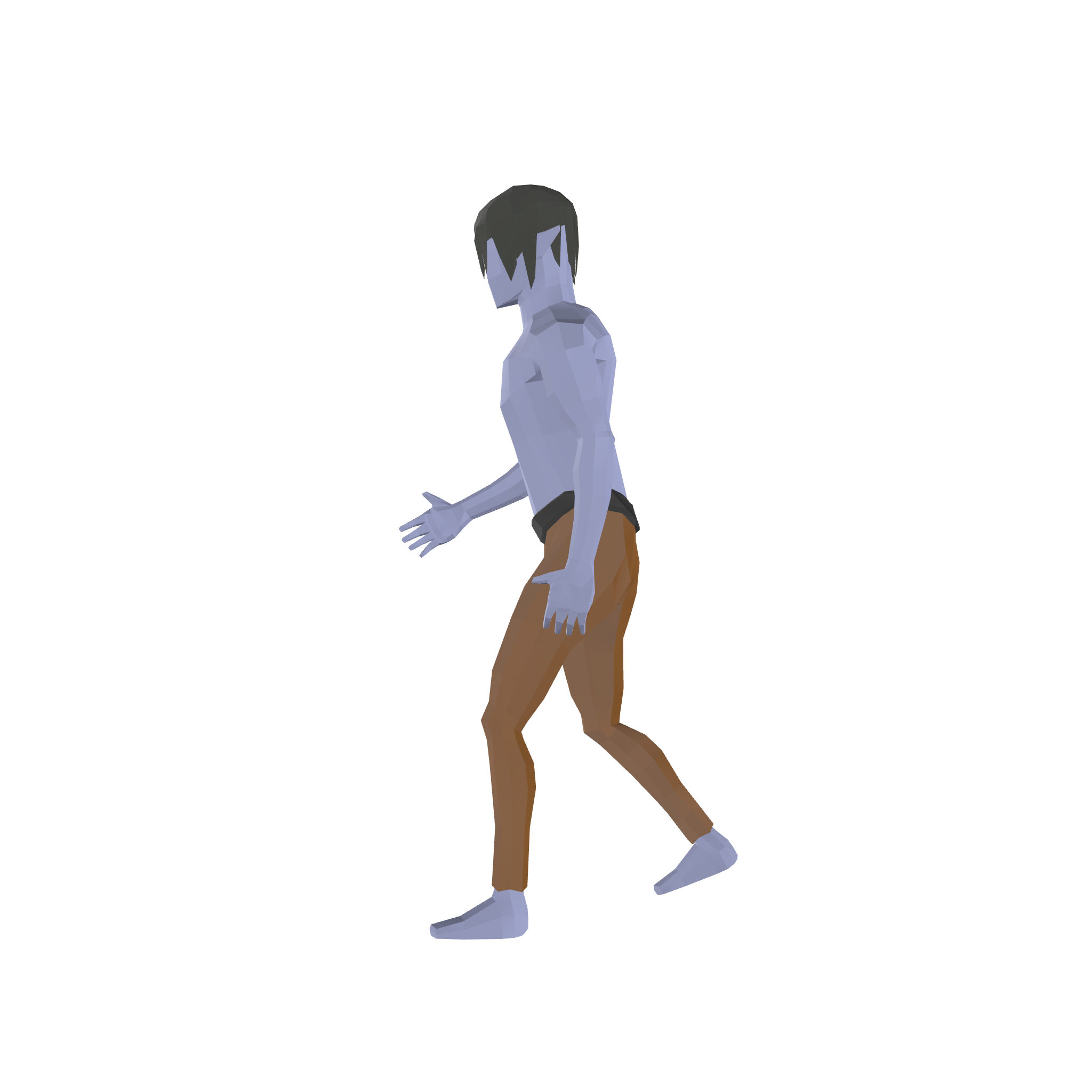}} & 
\raisebox{-0.5\height}{\includegraphics[width=0.16\columnwidth, trim=20 20 20 20, clip]{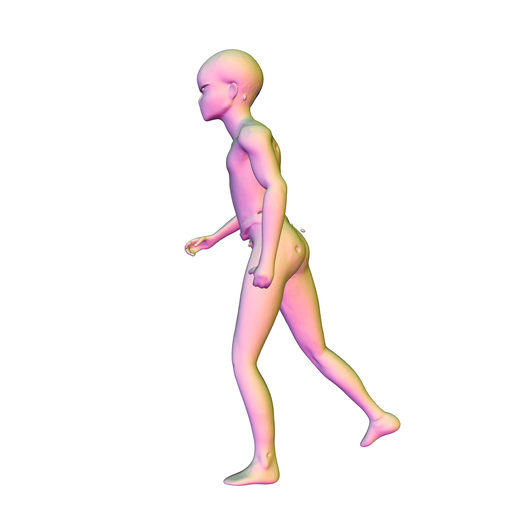}} &
\raisebox{-0.5\height}{\includegraphics[width=0.16\columnwidth, trim=40 40 40 40, clip]{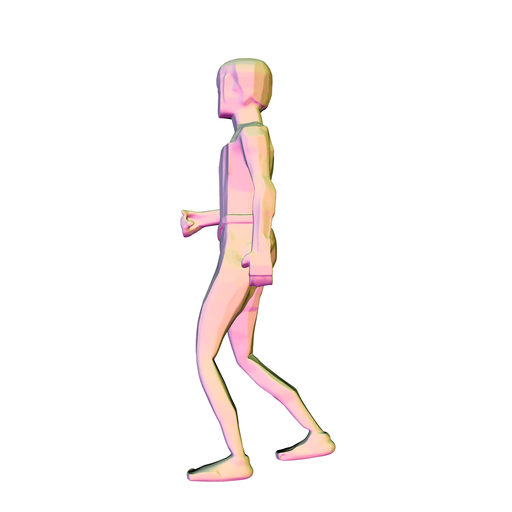}} &
\raisebox{-0.5\height}{\includegraphics[width=0.16\columnwidth, trim=40 40 40 40, clip]{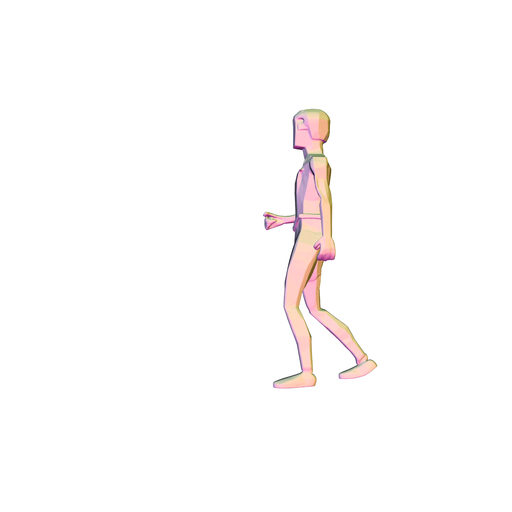}} &
\raisebox{-0.5\height}{\includegraphics[width=0.16\columnwidth, trim=40 40 40 40, clip]{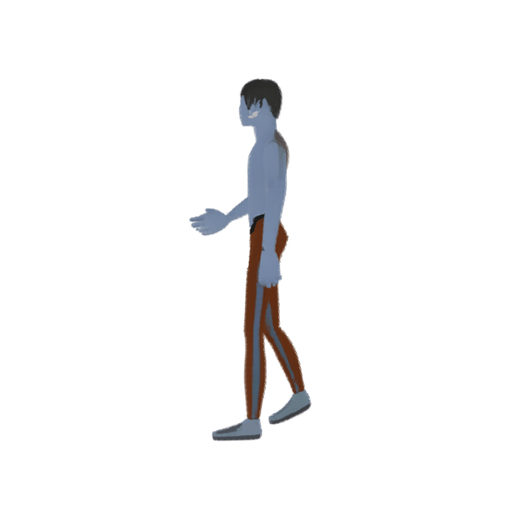}} &
\raisebox{-0.5\height}{\includegraphics[width=0.16\columnwidth, trim=40 40 40 40, clip]{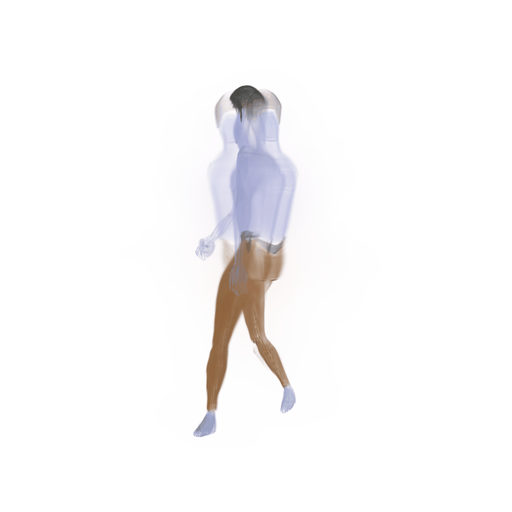}} \\[3pt]

% -------- Frame 8 --------
\raisebox{-0.5\height}{\rotatebox{90}{\tiny Input $\angle 0^\circ$}} &
\raisebox{-0.5\height}{\includegraphics[width=0.16\columnwidth, trim=40 40 40 40, clip]{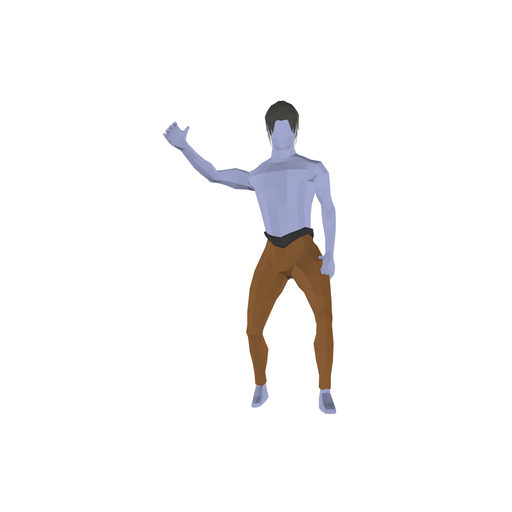}} &
\raisebox{-0.5\height}{\includegraphics[width=0.16\columnwidth, trim=20 20 20 20, clip]{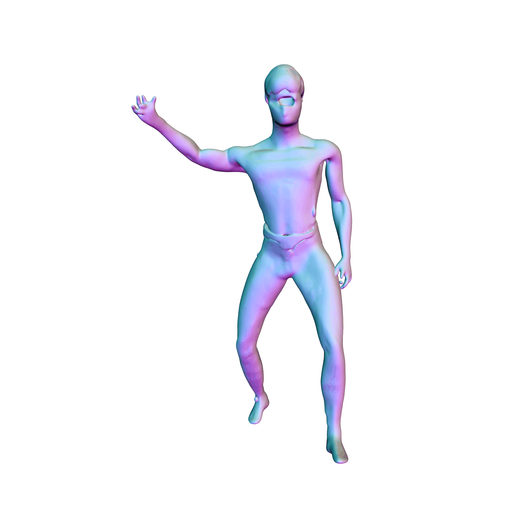}} &
\raisebox{-0.5\height}{\includegraphics[width=0.16\columnwidth, trim=40 40 40 40, clip]{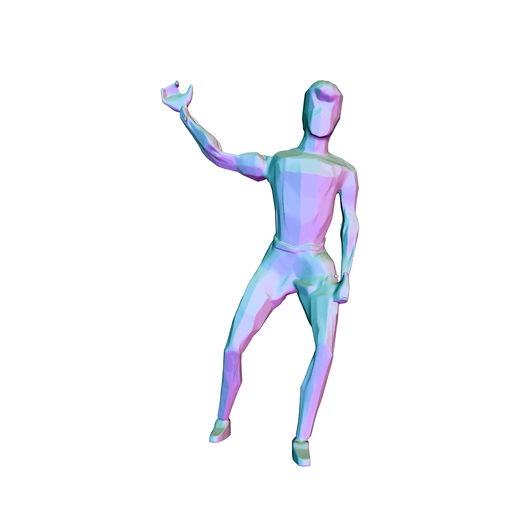}} &
\raisebox{-0.5\height}{\includegraphics[width=0.16\columnwidth, trim=40 40 40 40, clip]{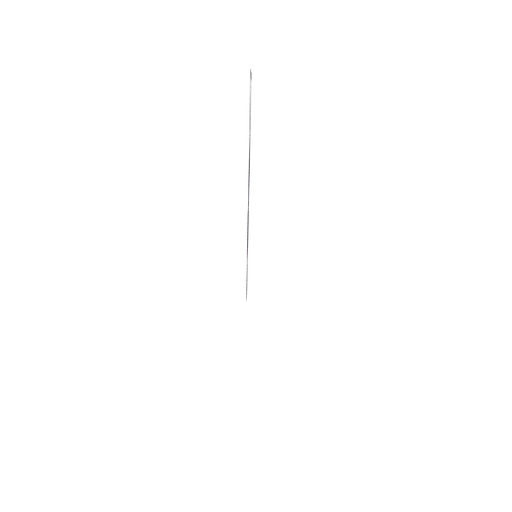}} &
\raisebox{-0.5\height}{\includegraphics[width=0.16\columnwidth, trim=40 40 40 40, clip]{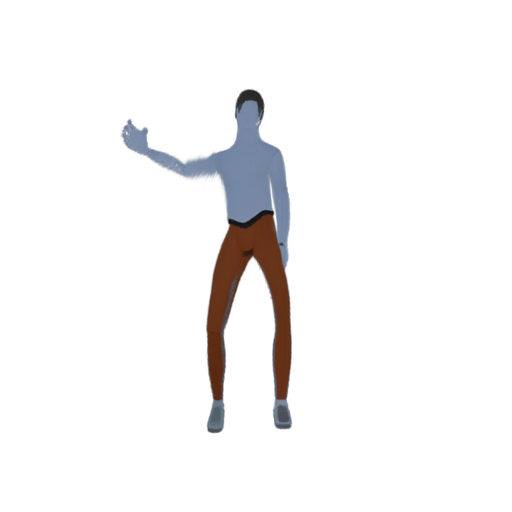}} &
\raisebox{-0.5\height}{\includegraphics[width=0.16\columnwidth, trim=40 40 40 40, clip]{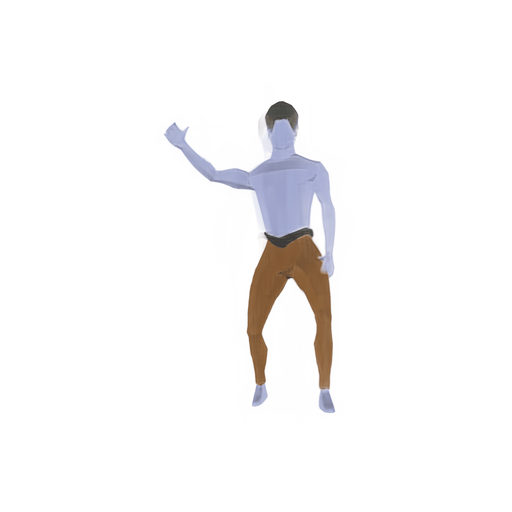}} \\[-0.5pt]
% --- New GT View Frame 8 ---
\raisebox{-0.5\height}{\rotatebox{90}{\tiny render $\angle 90^\circ$}} &
\raisebox{-0.5\height}{\includegraphics[width=0.16\columnwidth, trim=40 40 40 40, clip]{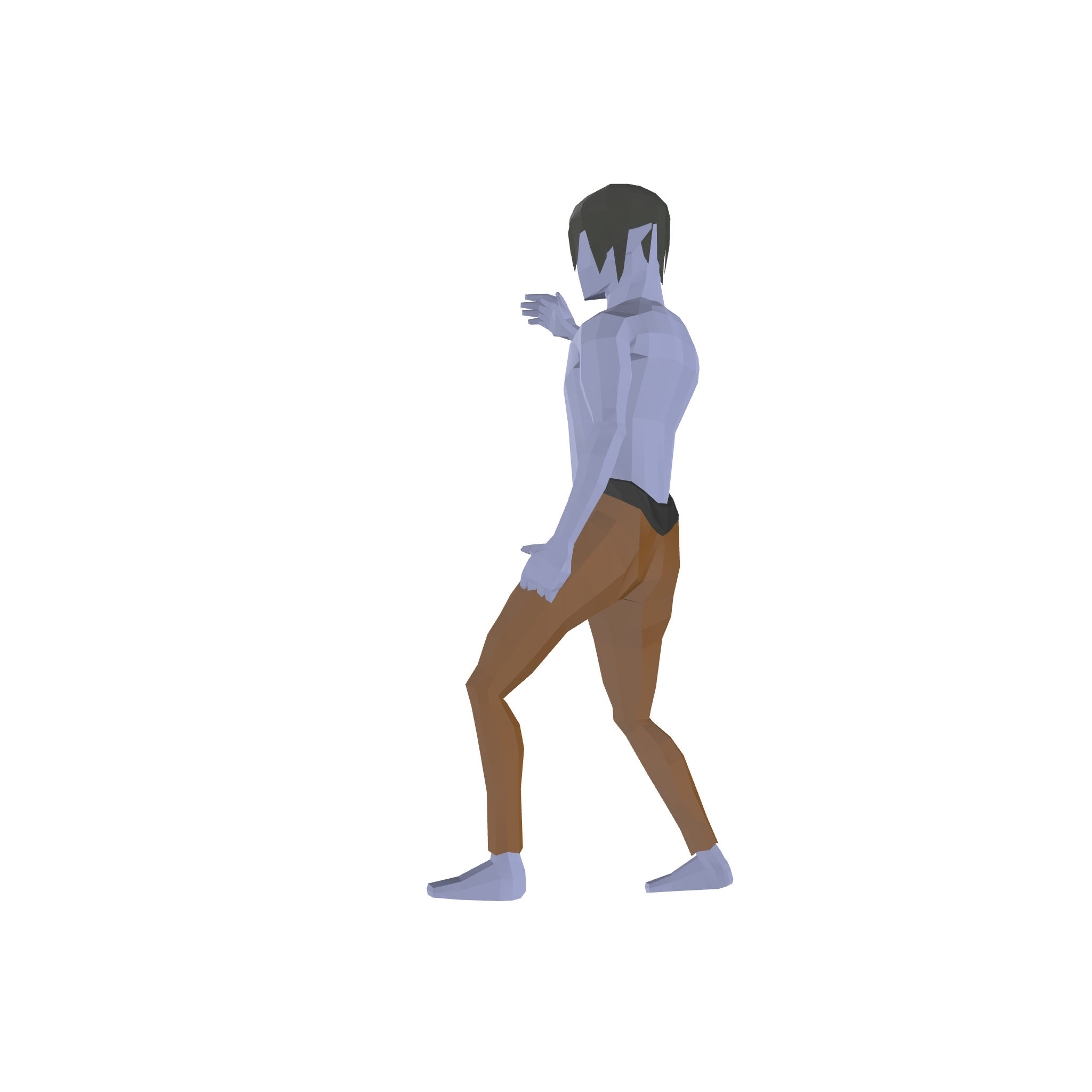}} & 
\raisebox{-0.5\height}{\includegraphics[width=0.16\columnwidth, trim=20 20 20 20, clip]{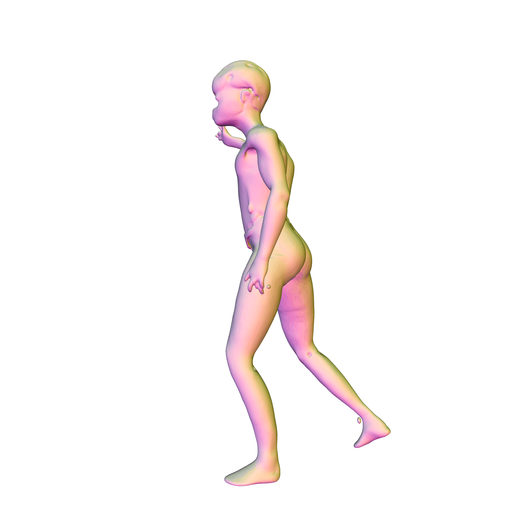}} &
\raisebox{-0.5\height}{\includegraphics[width=0.16\columnwidth, trim=40 40 40 40, clip]{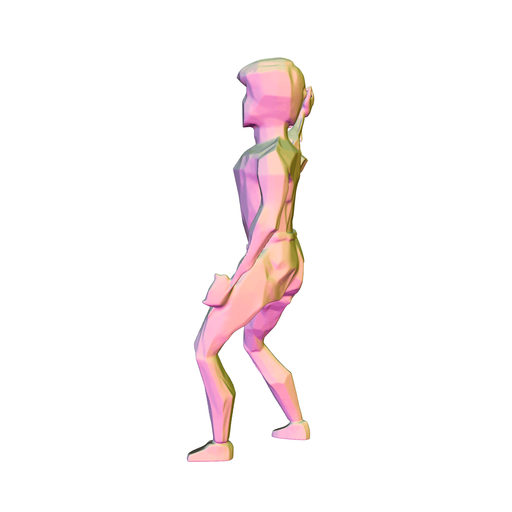}} &
\raisebox{-0.5\height}{\includegraphics[width=0.16\columnwidth, trim=40 40 40 40, clip]{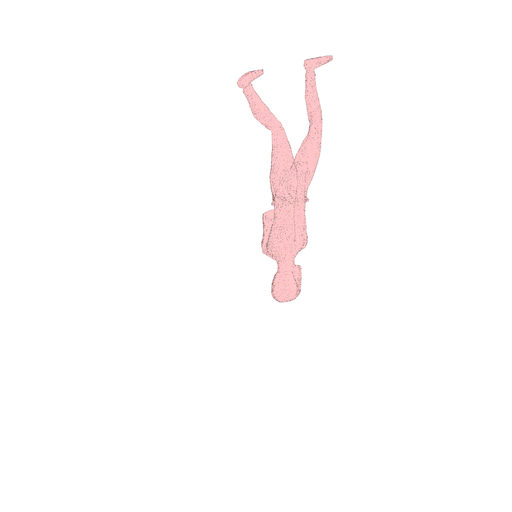}} &
\raisebox{-0.5\height}{\includegraphics[width=0.16\columnwidth, trim=40 40 40 40, clip]{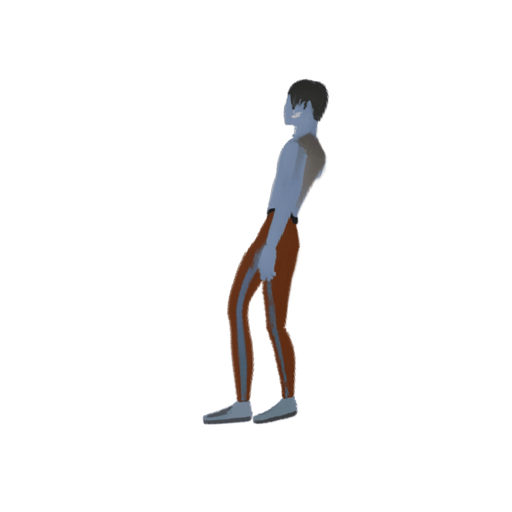}} &
\raisebox{-0.5\height}{\includegraphics[width=0.16\columnwidth, trim=40 40 40 40, clip]{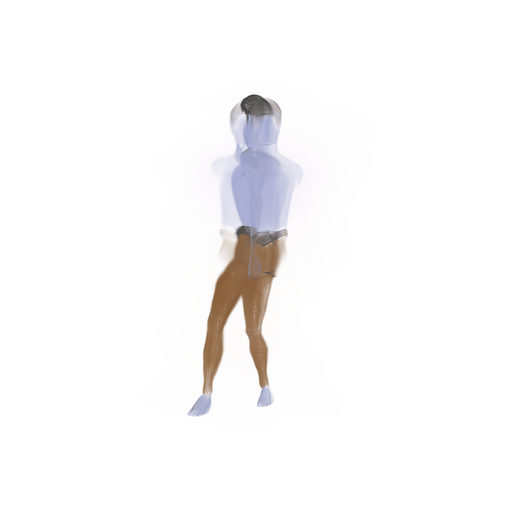}} \\[5pt]

% ==================== SUBJECT 2: 364ba80... ====================
% -------- Frame 0 --------
\raisebox{-0.5\height}{\rotatebox{90}{\tiny Input $\angle 0^\circ$}} &
\raisebox{-0.5\height}{\includegraphics[width=0.16\columnwidth, trim=40 40 40 40, clip]{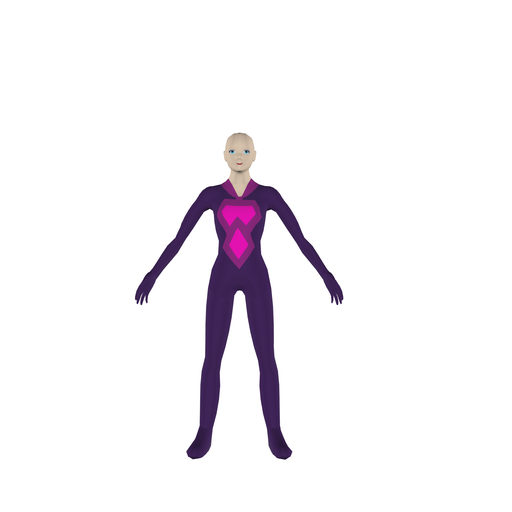}} &
\raisebox{-0.5\height}{\includegraphics[width=0.16\columnwidth, trim=20 20 20 20, clip]{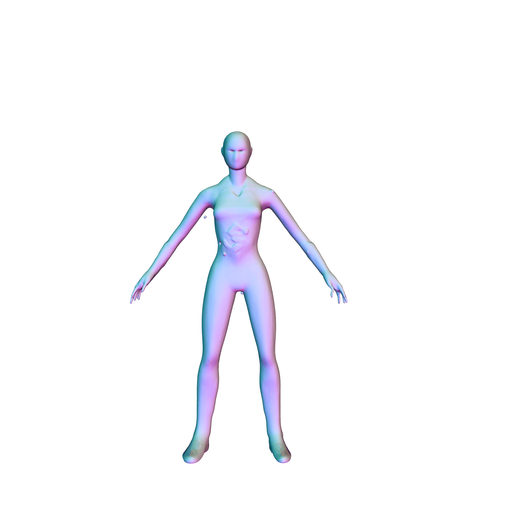}} &
\raisebox{-0.5\height}{\includegraphics[width=0.16\columnwidth, trim=40 40 40 40, clip]{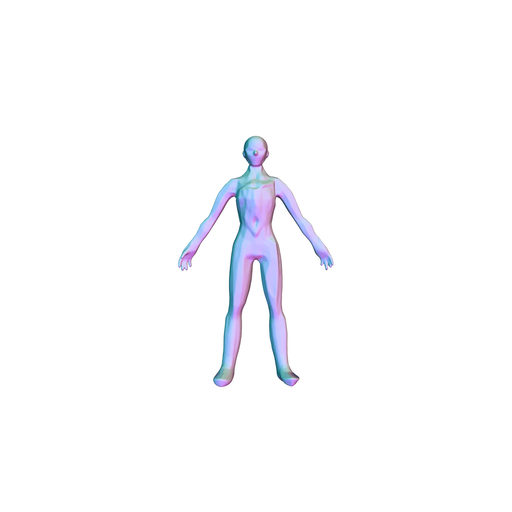}} &
\raisebox{-0.5\height}{\includegraphics[width=0.16\columnwidth, trim=40 40 40 40, clip]{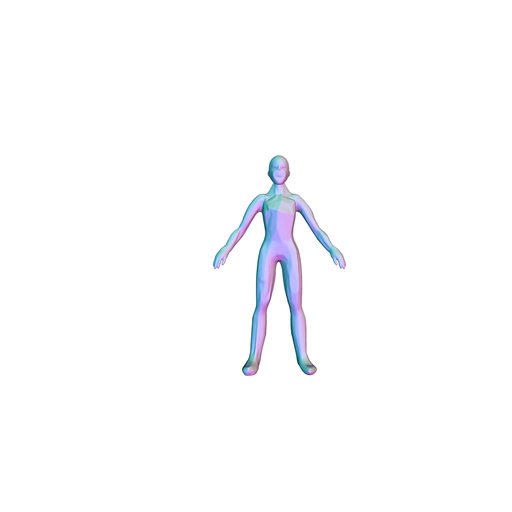}} &
\raisebox{-0.5\height}{\includegraphics[width=0.16\columnwidth, trim=40 40 40 40, clip]{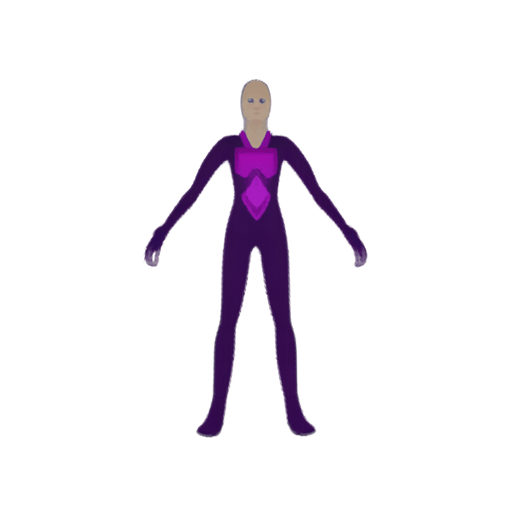}} &
\raisebox{-0.5\height}{\includegraphics[width=0.16\columnwidth, trim=40 40 40 40, clip]{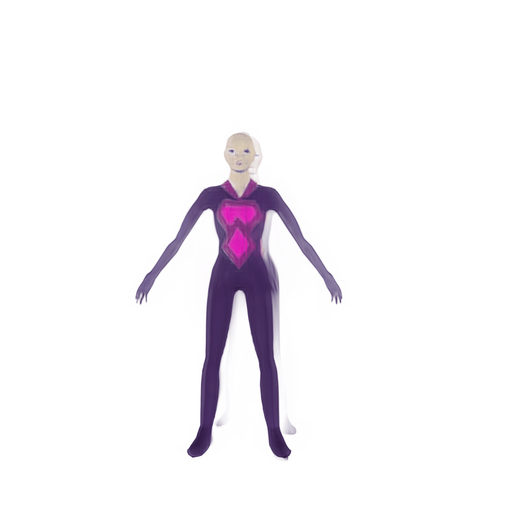}} \\[-0.5pt]
% --- New GT View Frame 0 ---
\raisebox{-0.5\height}{\rotatebox{90}{\tiny render $\angle 90^\circ$}} &
\raisebox{-0.5\height}{\includegraphics[width=0.16\columnwidth, trim=40 40 40 40, clip]{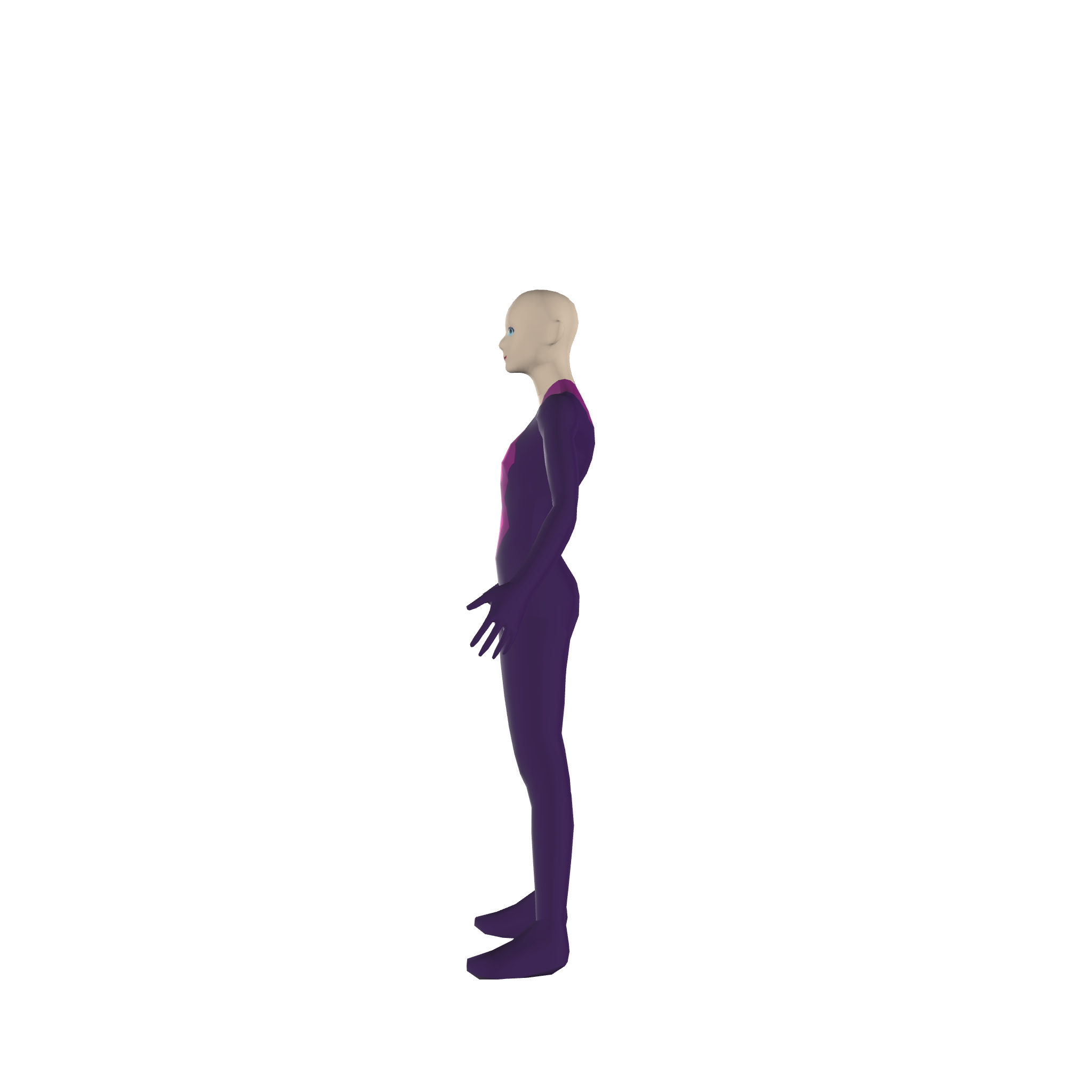}} & 
\raisebox{-0.5\height}{\includegraphics[width=0.16\columnwidth, trim=20 20 20 20, clip]{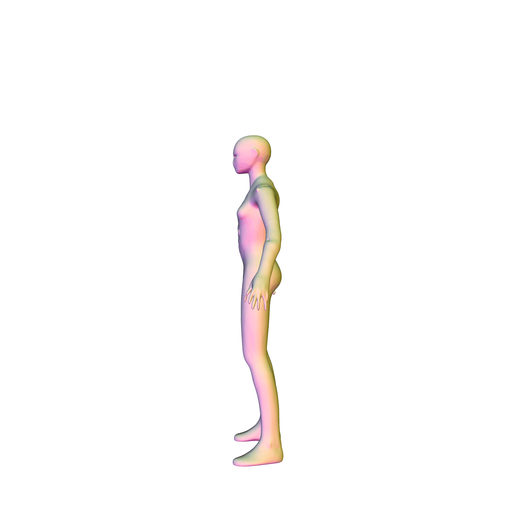}} &
\raisebox{-0.5\height}{\includegraphics[width=0.16\columnwidth, trim=40 40 40 40, clip]{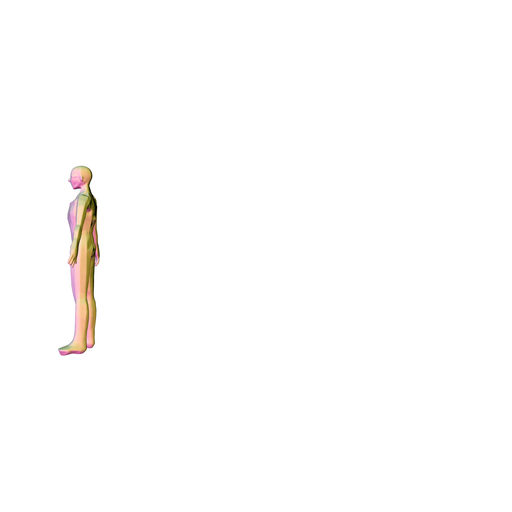}} &
\raisebox{-0.5\height}{\includegraphics[width=0.16\columnwidth, trim=40 40 40 40, clip]{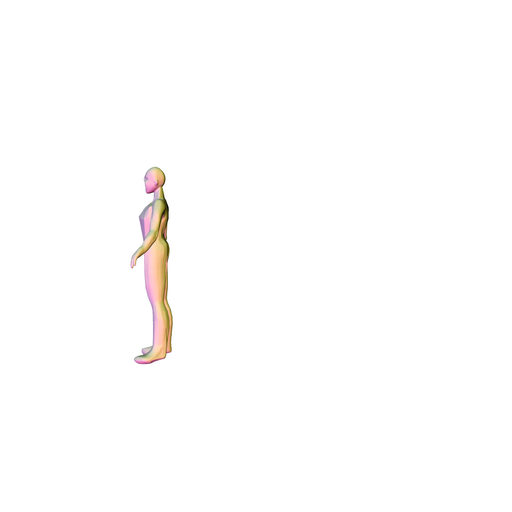}} &
\raisebox{-0.5\height}{\includegraphics[width=0.16\columnwidth, trim=40 40 40 40, clip]{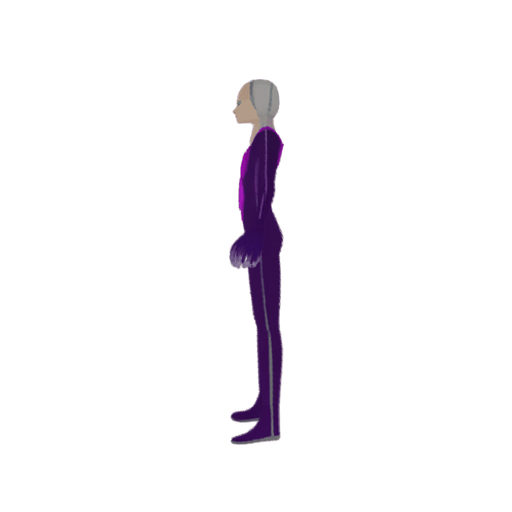}} &
\raisebox{-0.5\height}{\includegraphics[width=0.16\columnwidth, trim=40 40 40 40, clip]{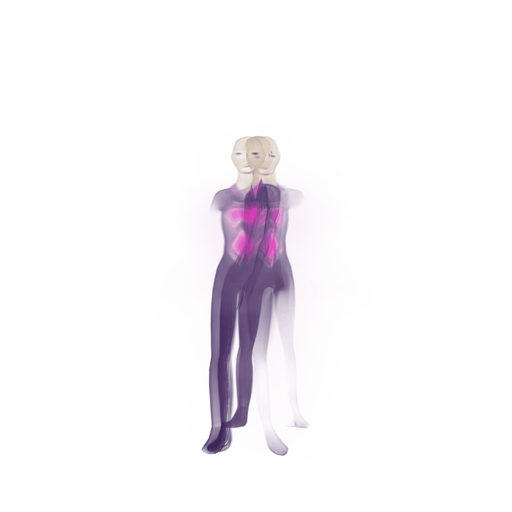}} \\[3pt]

% -------- Frame 8 --------
\raisebox{-0.5\height}{\rotatebox{90}{\tiny Input $\angle 0^\circ$}} &
\raisebox{-0.5\height}{\includegraphics[width=0.16\columnwidth, trim=40 40 40 40, clip]{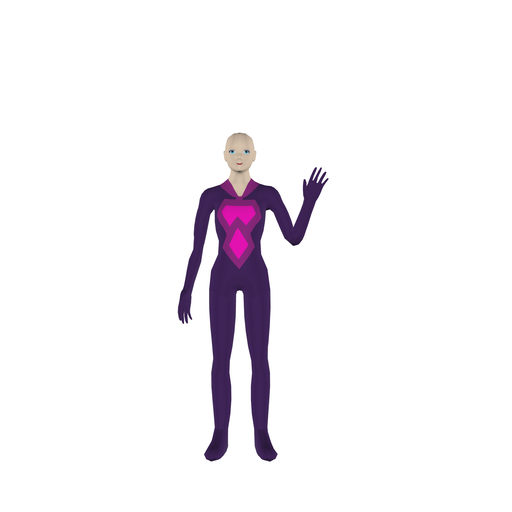}} &
\raisebox{-0.5\height}{\includegraphics[width=0.16\columnwidth, trim=20 20 20 20, clip]{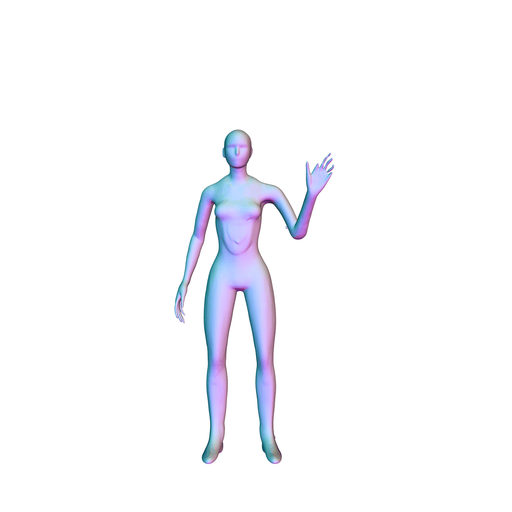}} &
\raisebox{-0.5\height}{\includegraphics[width=0.16\columnwidth, trim=40 40 40 40, clip]{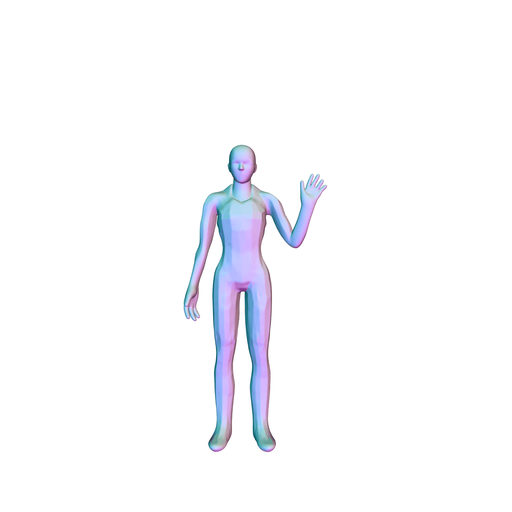}} &
\raisebox{-0.5\height}{\includegraphics[width=0.16\columnwidth, trim=40 40 40 40, clip]{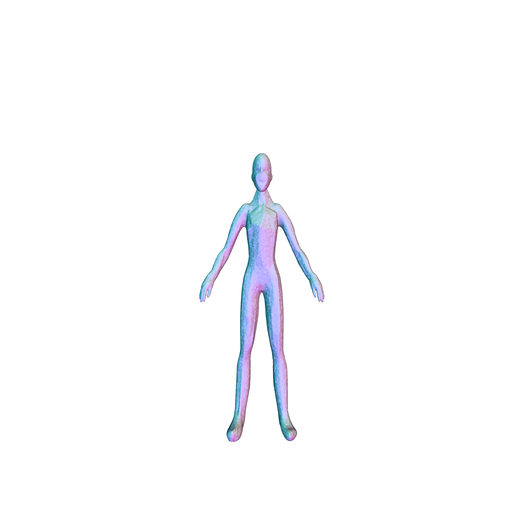}} &
\raisebox{-0.5\height}{\includegraphics[width=0.16\columnwidth, trim=40 40 40 40, clip]{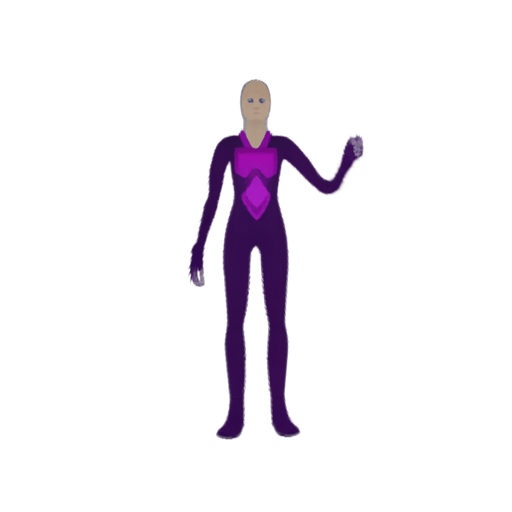}} &
\raisebox{-0.5\height}{\includegraphics[width=0.16\columnwidth, trim=40 40 40 40, clip]{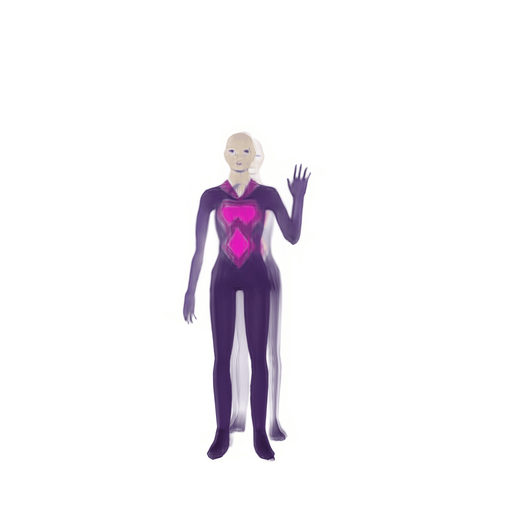}} \\[-0.5pt]
% --- New GT View Frame 8 ---
\raisebox{-0.5\height}{\rotatebox{90}{\tiny render $\angle 90^\circ$}} &
\raisebox{-0.5\height}{\includegraphics[width=0.16\columnwidth, trim=40 40 40 40, clip]{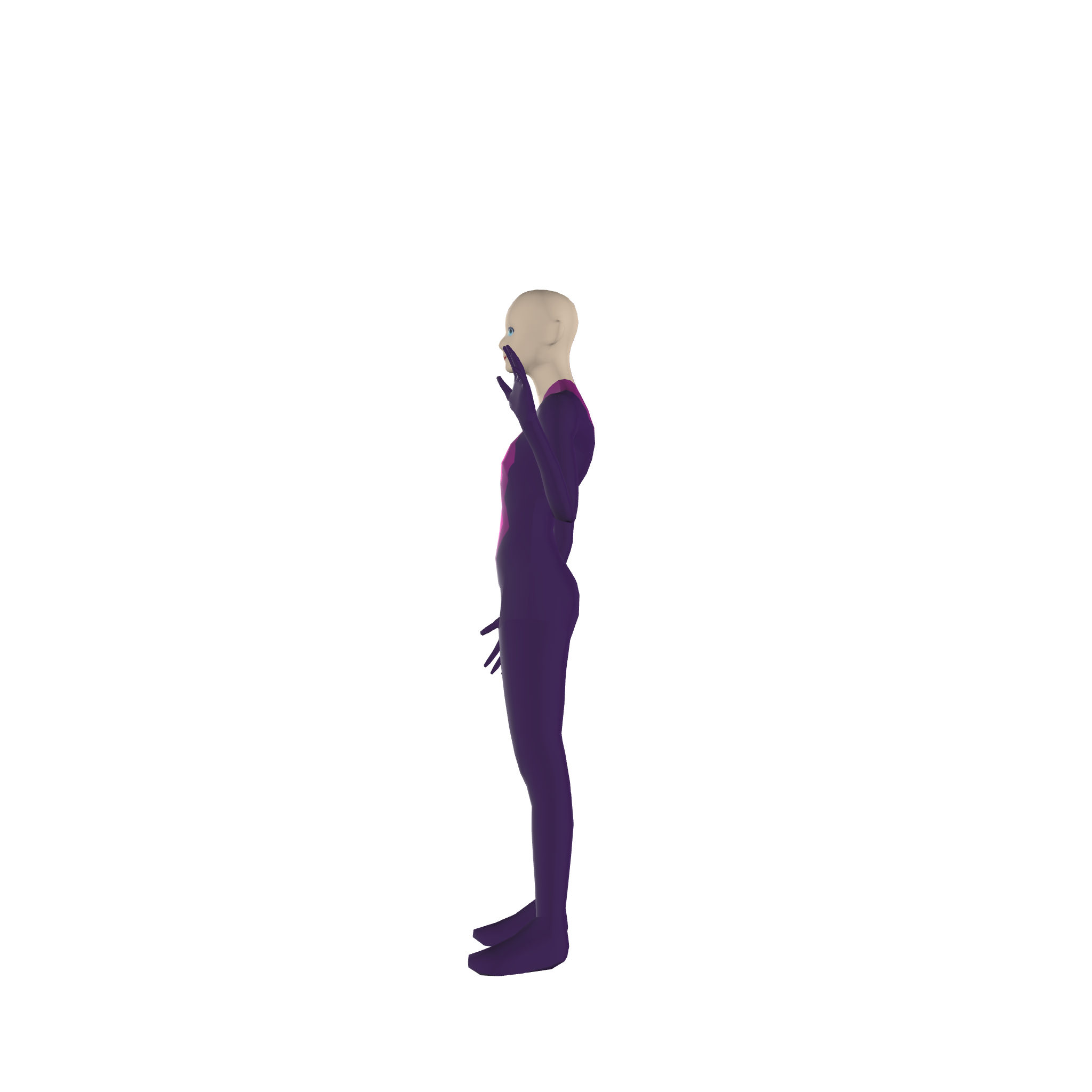}} & 
\raisebox{-0.5\height}{\includegraphics[width=0.16\columnwidth, trim=20 20 20 20, clip]{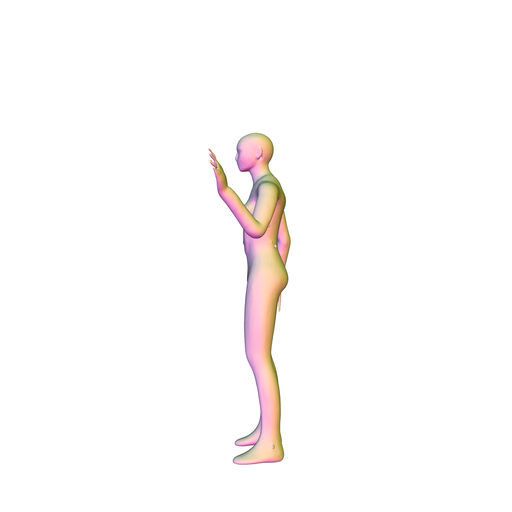}} &
\raisebox{-0.5\height}{\includegraphics[width=0.16\columnwidth, trim=40 40 40 40, clip]{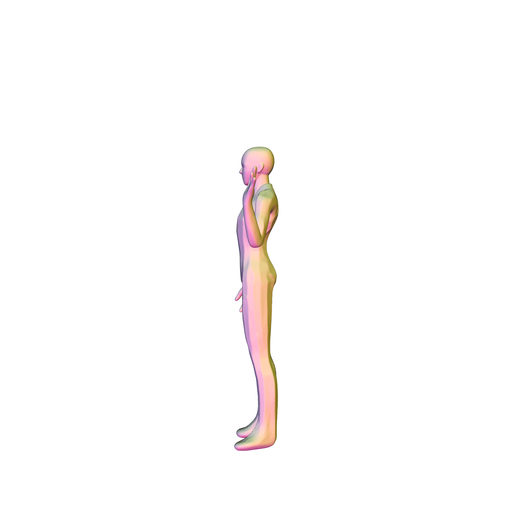}} &
\raisebox{-0.5\height}{\includegraphics[width=0.16\columnwidth, trim=40 40 40 40, clip]{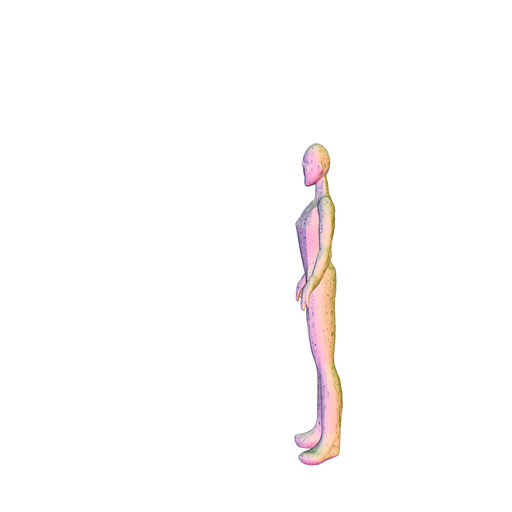}} &
\raisebox{-0.5\height}{\includegraphics[width=0.16\columnwidth, trim=40 40 40 40, clip]{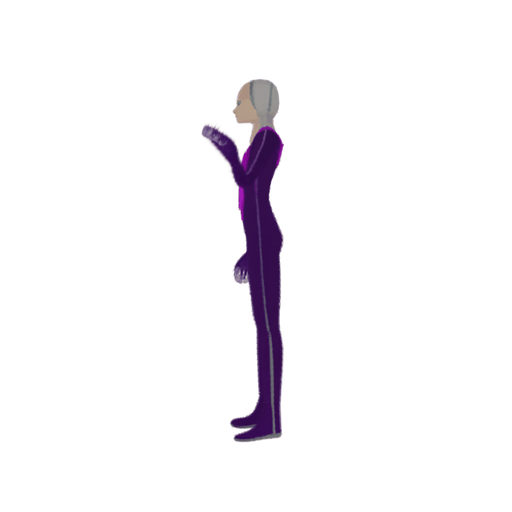}} &
\raisebox{-0.5\height}{\includegraphics[width=0.16\columnwidth, trim=40 40 40 40, clip]{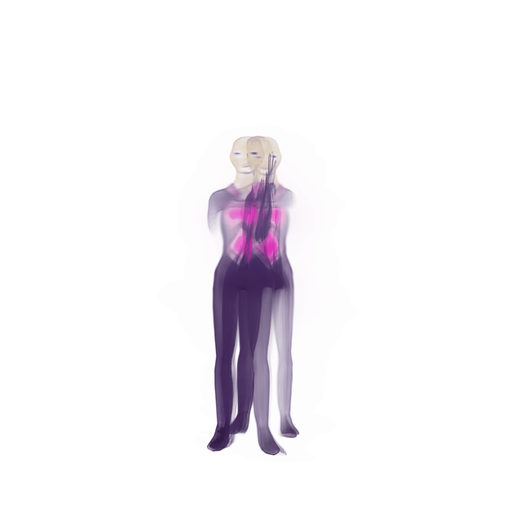}} \\[5pt]

% ==================== SUBJECT 3: 3d32db3... ====================
% -------- Frame 0 --------
\raisebox{-0.5\height}{\rotatebox{90}{\tiny Input $\angle 0^\circ$}} &
\raisebox{-0.5\height}{\includegraphics[width=0.16\columnwidth, trim=40 40 40 40, clip]{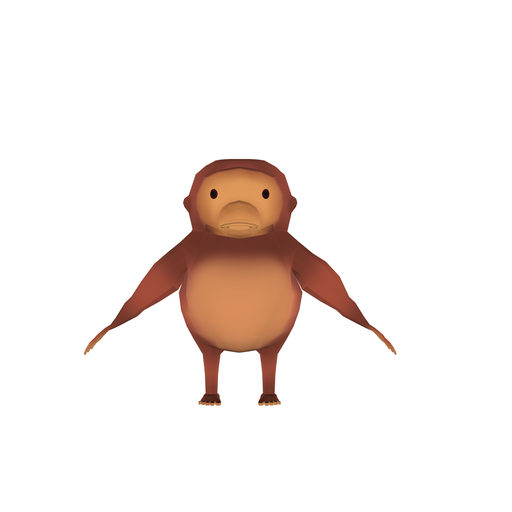}} &
\raisebox{-0.5\height}{\includegraphics[width=0.16\columnwidth, trim=20 20 20 20, clip]{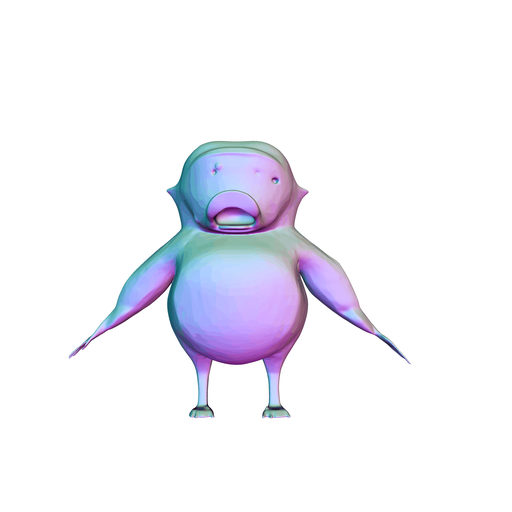}} &
\raisebox{-0.5\height}{\includegraphics[width=0.16\columnwidth, trim=40 40 40 40, clip]{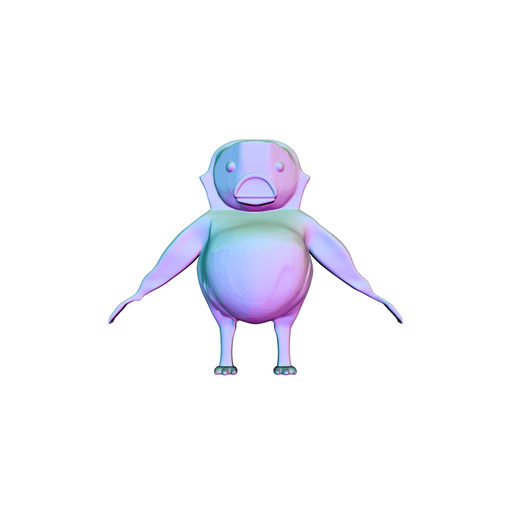}} &
\raisebox{-0.5\height}{\includegraphics[width=0.16\columnwidth, trim=40 40 40 40, clip]{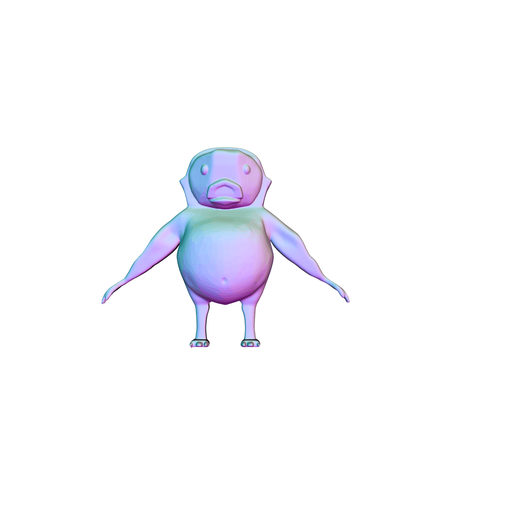}} &
\raisebox{-0.5\height}{\includegraphics[width=0.16\columnwidth, trim=40 40 40 40, clip]{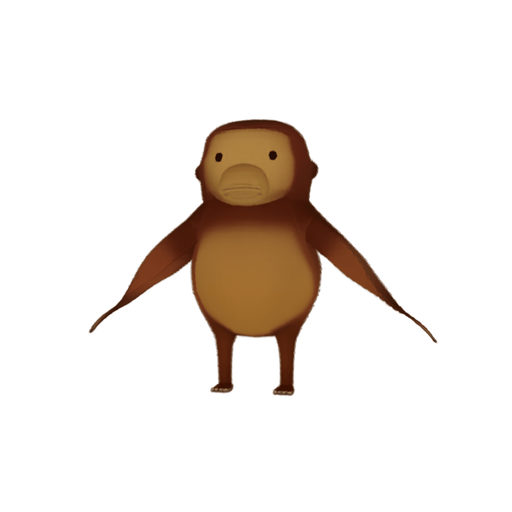}} &
\raisebox{-0.5\height}{\includegraphics[width=0.16\columnwidth, trim=40 40 40 40, clip]{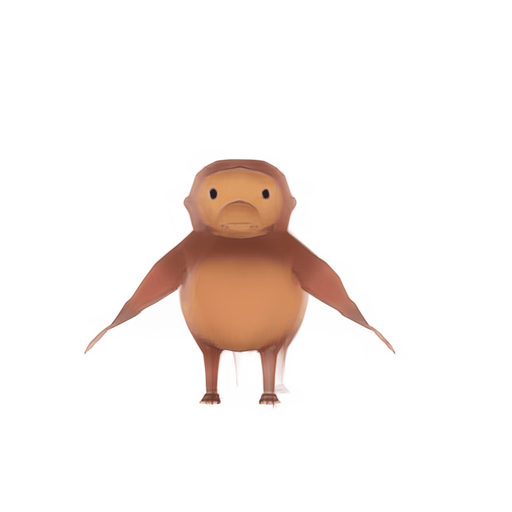}} \\[-0.5pt]
% --- New GT View Frame 0 ---
\raisebox{-0.5\height}{\rotatebox{90}{\tiny render $\angle 90^\circ$}} &
\raisebox{-0.5\height}{\includegraphics[width=0.16\columnwidth, trim=40 40 40 40, clip]{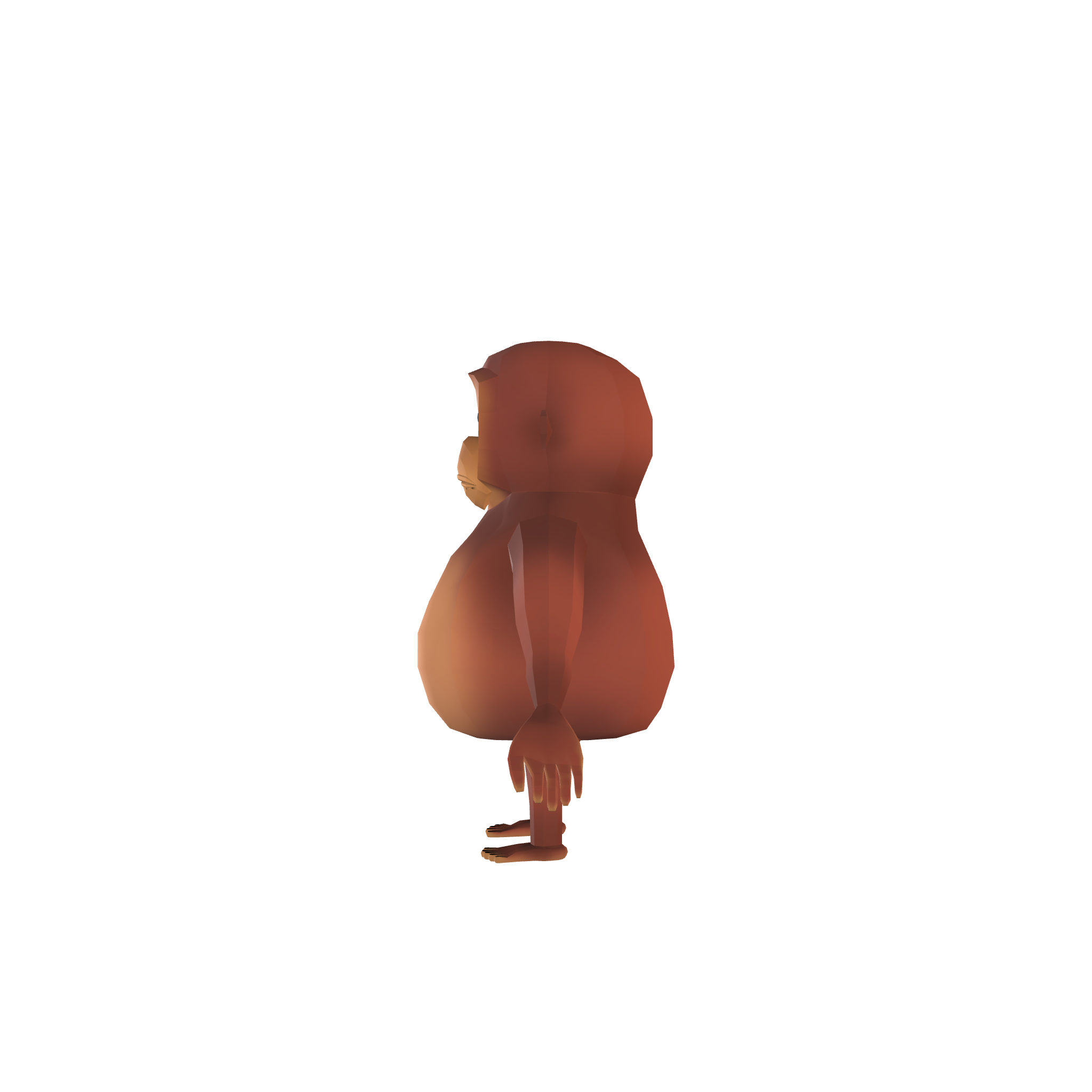}} & 
\raisebox{-0.5\height}{\includegraphics[width=0.16\columnwidth, trim=20 20 20 20, clip]{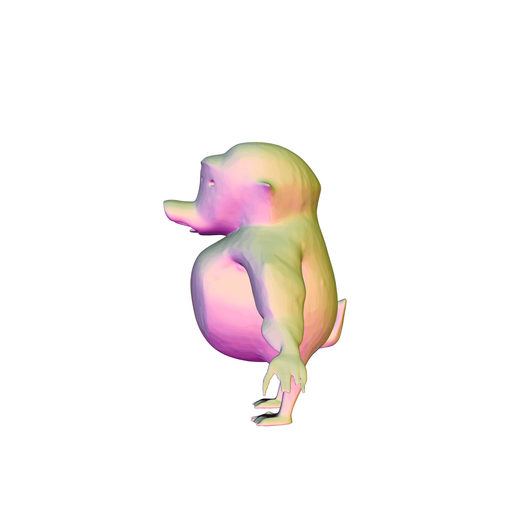}} &
\raisebox{-0.5\height}{\includegraphics[width=0.16\columnwidth, trim=40 40 40 40, clip]{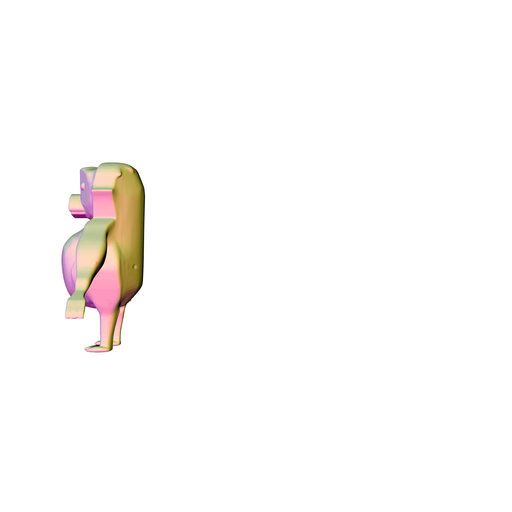}} &
\raisebox{-0.5\height}{\includegraphics[width=0.16\columnwidth, trim=40 40 40 40, clip]{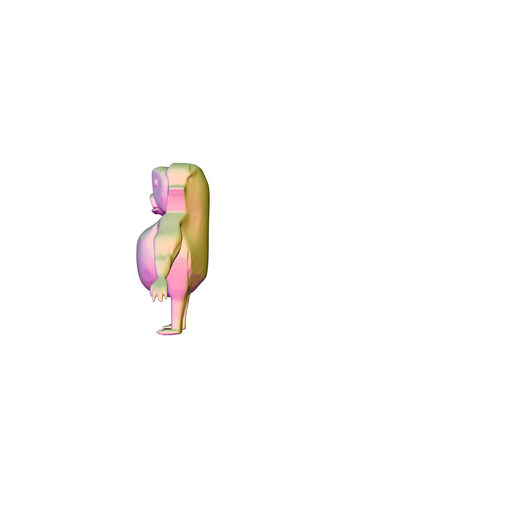}} &
\raisebox{-0.5\height}{\includegraphics[width=0.16\columnwidth, trim=40 40 40 40, clip]{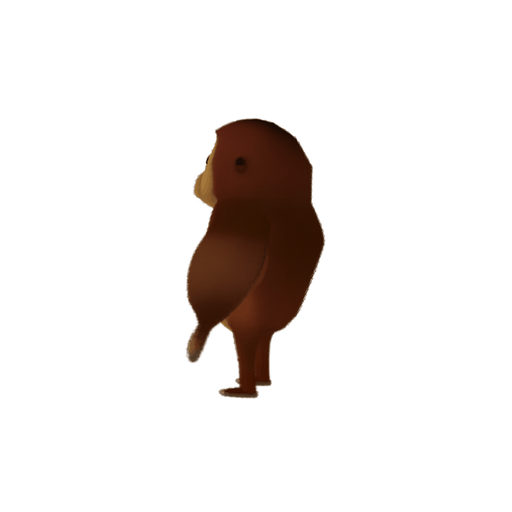}} &
\raisebox{-0.5\height}{\includegraphics[width=0.16\columnwidth, trim=40 40 40 40, clip]{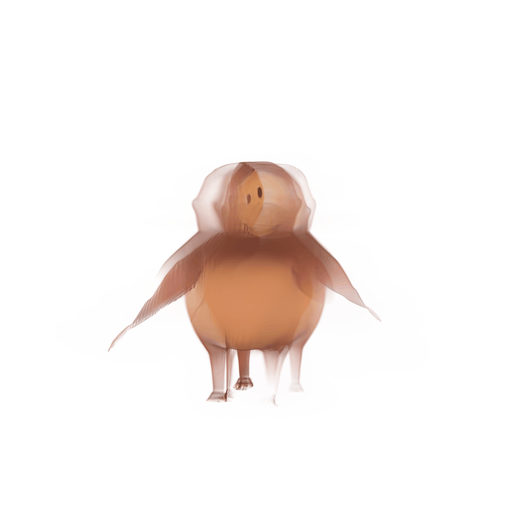}} \\[3pt]

% -------- Frame 8 --------
\raisebox{-0.5\height}{\rotatebox{90}{\tiny Input $\angle 0^\circ$}} &
\raisebox{-0.5\height}{\includegraphics[width=0.16\columnwidth, trim=40 40 40 40, clip]{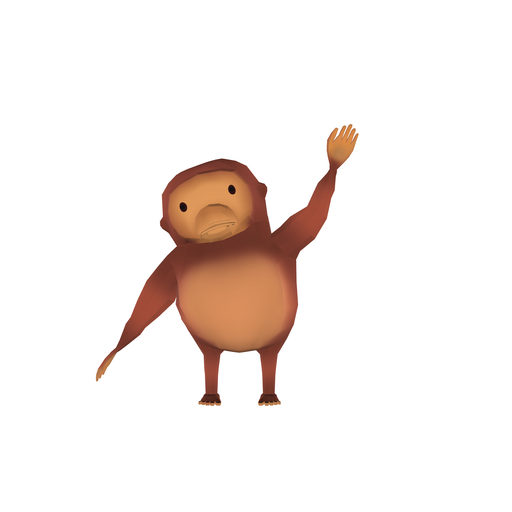}} &
\raisebox{-0.5\height}{\includegraphics[width=0.16\columnwidth, trim=20 20 20 20, clip]{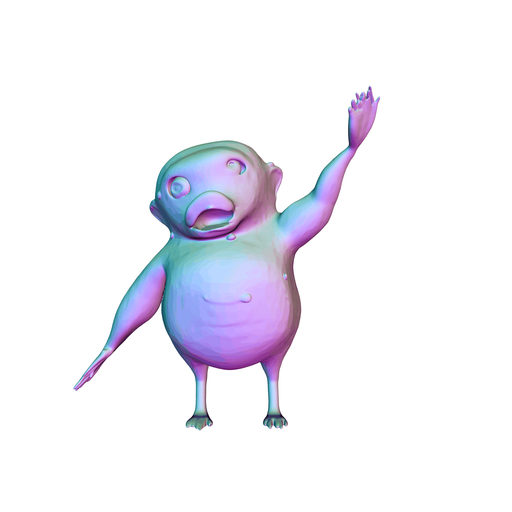}} &
\raisebox{-0.5\height}{\includegraphics[width=0.16\columnwidth, trim=40 40 40 40, clip]{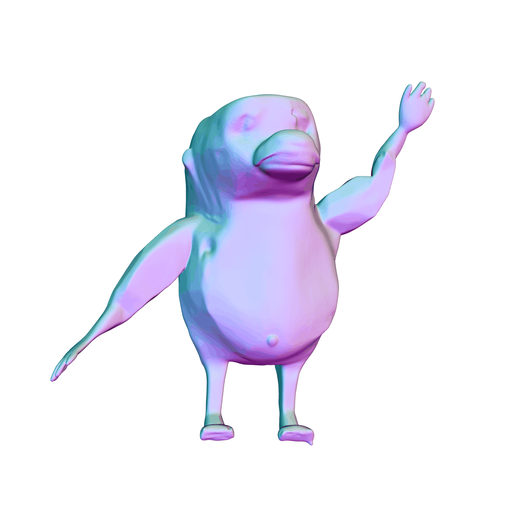}} &
\raisebox{-0.5\height}{\includegraphics[width=0.16\columnwidth, trim=40 40 40 40, clip]{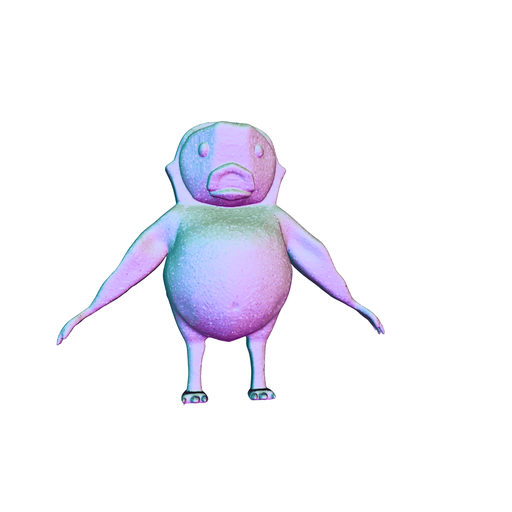}} &
\raisebox{-0.5\height}{\includegraphics[width=0.16\columnwidth, trim=40 40 40 40, clip]{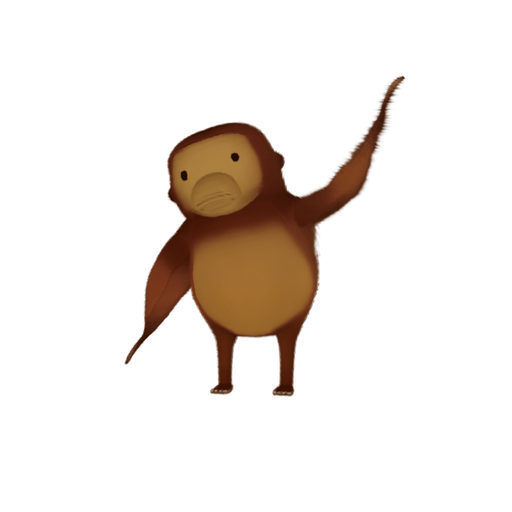}} &
\raisebox{-0.5\height}{\includegraphics[width=0.16\columnwidth, trim=40 40 40 40, clip]{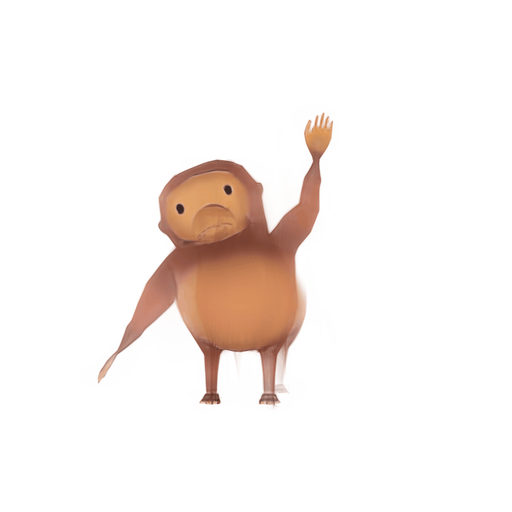}} \\[-0.5pt]
% --- New GT View Frame 8 ---
\raisebox{-0.5\height}{\rotatebox{90}{\tiny render $\angle 90^\circ$}} &
\raisebox{-0.5\height}{\includegraphics[width=0.16\columnwidth, trim=80 80 80 80, clip]{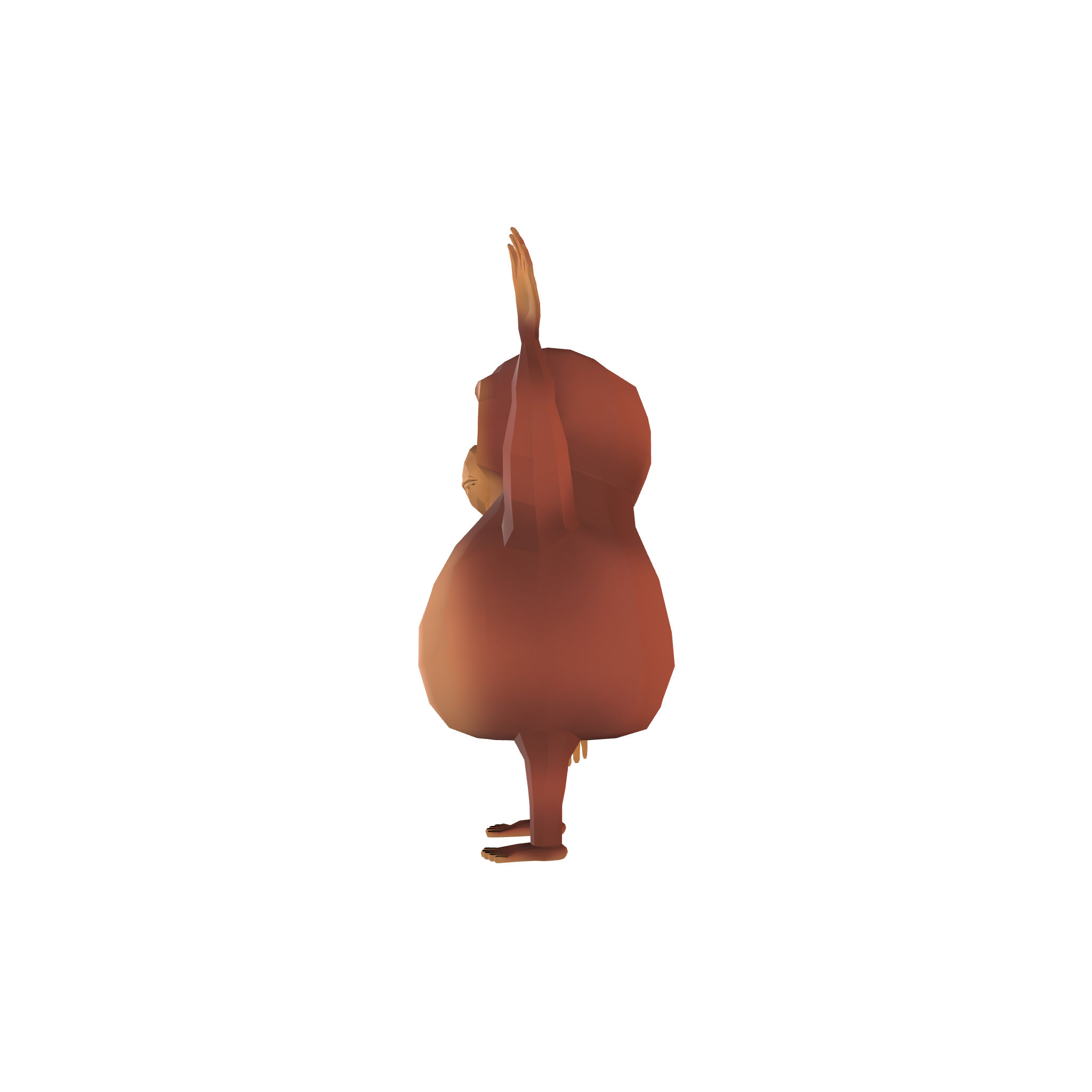}} & 
\raisebox{-0.5\height}{\includegraphics[width=0.16\columnwidth, trim=20 20 20 20, clip]{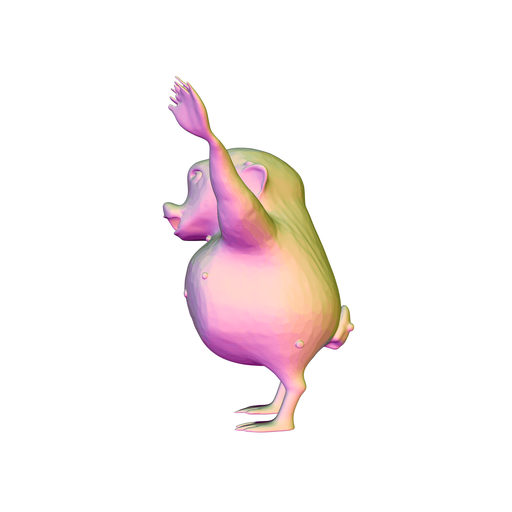}} &
\raisebox{-0.5\height}{\includegraphics[width=0.16\columnwidth, trim=40 40 40 40, clip]{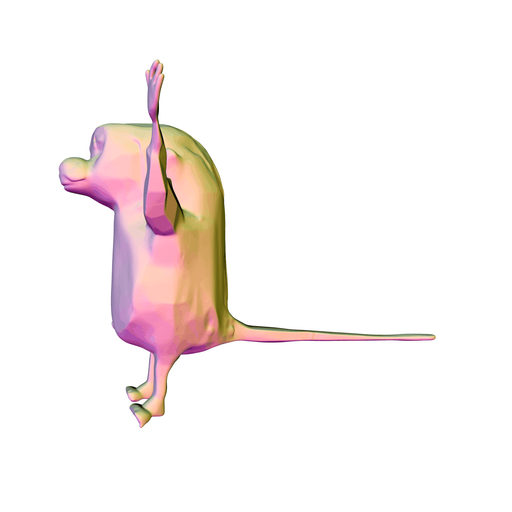}} &
\raisebox{-0.5\height}{\includegraphics[width=0.16\columnwidth, trim=40 40 40 40, clip]{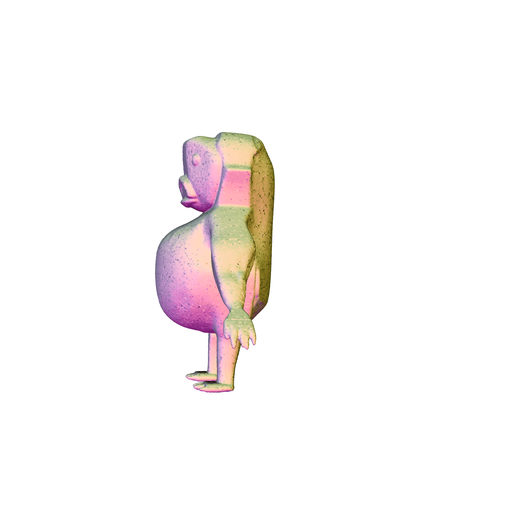}} &
\raisebox{-0.5\height}{\includegraphics[width=0.16\columnwidth, trim=40 40 40 40, clip]{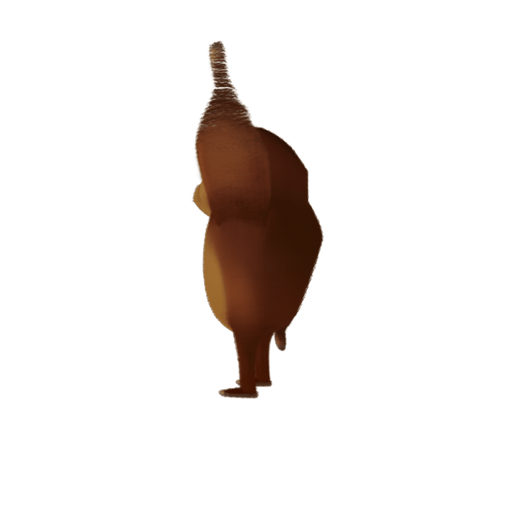}} &
\raisebox{-0.5\height}{\includegraphics[width=0.16\columnwidth, trim=40 40 40 40, clip]{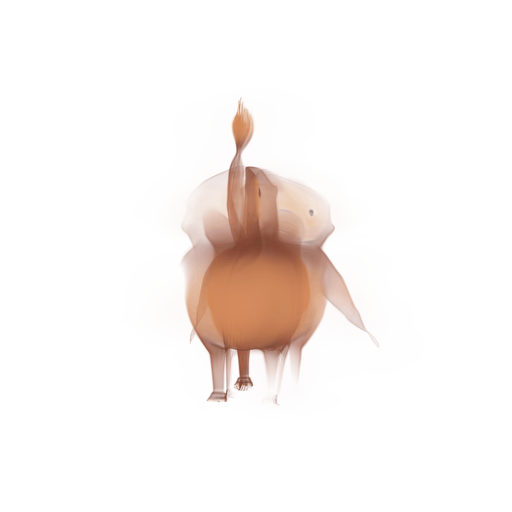}} \\

  \end{tabular}
  % ==================== TABLE END ====================
  
  \vspace{-3pt}
  \caption{
  Qualitative 4D generation comparisons from Objaverse~\cite{deitke2022objaverseuniverseannotated3d} showing three subjects at two time steps. For each model, we show the reconstructed input view (top) and a rendered novel view (bottom). We display the ground truth novel view in the bottom left.
  }
  \label{fig:qual_comp_4d_appendix_3_subjects}
\end{figure}

\begin{figure}[h]
  \centering
  % --- MODIFICATIONS ---
  \setlength{\tabcolsep}{0pt} % Zero horizontal spacing between images
  \renewcommand{\arraystretch}{0} % Remove default extra padding in rows

  % ==================== TABLE START ====================
  % Added a column 'c' at the start for vertical text
  \begin{tabular}{@{}c@{\hspace{2pt}}cccccc@{}}
    % --- HEADERS ---
    & % Empty cell for the vertical text column
    \small{Input / GT} &
    \textbf{\small{Ours}} &
    \small{TripoSG} &
    \small{V2M4} &
    \small{GVFD} &
    \small{L4GM} \\[3pt] % Keep a small gap between titles and images

% ==================== SUBJECT 5: 7af0f72... ====================
% -------- Frame 0 --------
\raisebox{-0.5\height}{\rotatebox{90}{\tiny Input $\angle 0^\circ$}} &
\raisebox{-0.5\height}{\includegraphics[width=0.16\columnwidth, trim=40 40 40 40, clip]{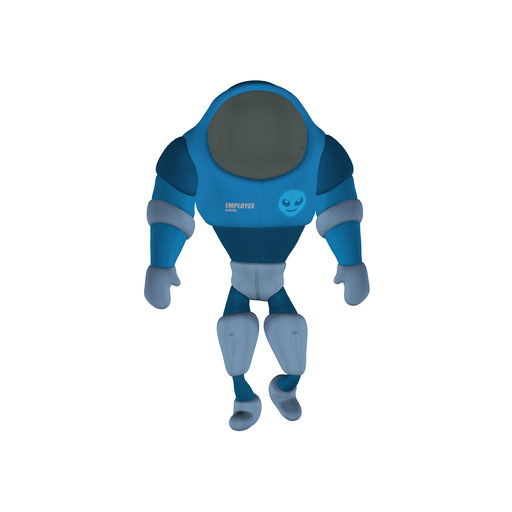}} &
\raisebox{-0.5\height}{\includegraphics[width=0.16\columnwidth, trim=20 20 20 20, clip]{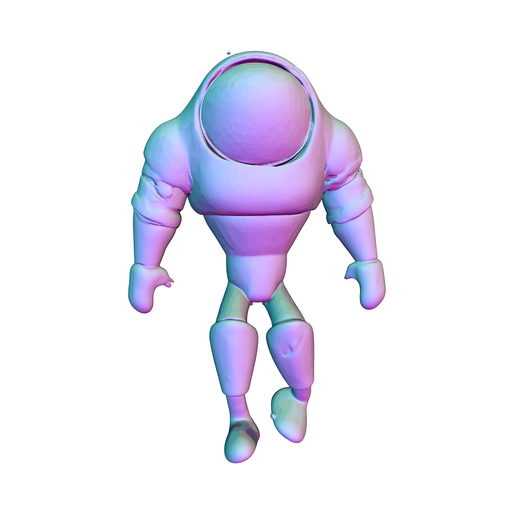}} &
\raisebox{-0.5\height}{\includegraphics[width=0.16\columnwidth, trim=40 40 40 40, clip]{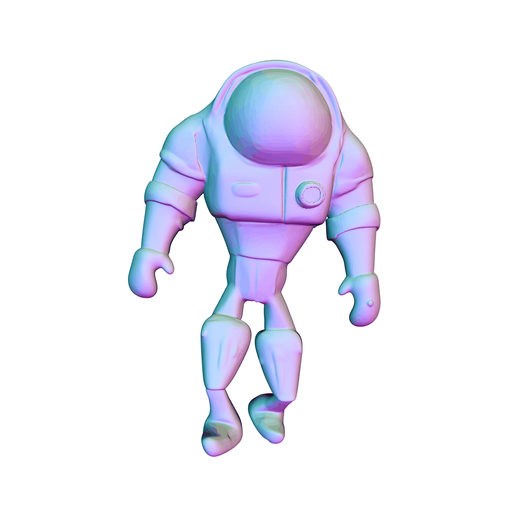}} &
\raisebox{-0.5\height}{\includegraphics[width=0.16\columnwidth, trim=40 40 40 40, clip]{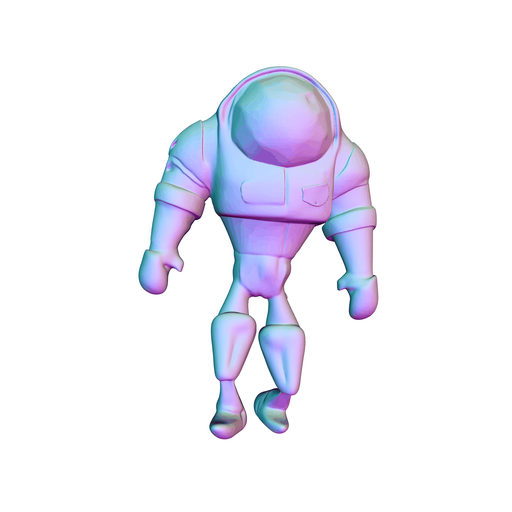}} &
\raisebox{-0.5\height}{\includegraphics[width=0.16\columnwidth, trim=40 40 40 40, clip]{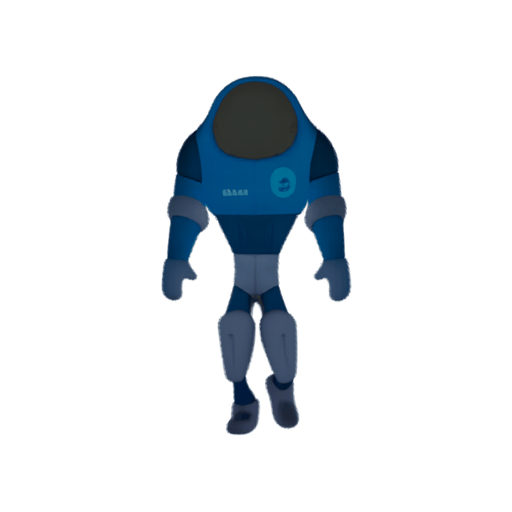}} &
\raisebox{-0.5\height}{\includegraphics[width=0.16\columnwidth, trim=40 40 40 40, clip]{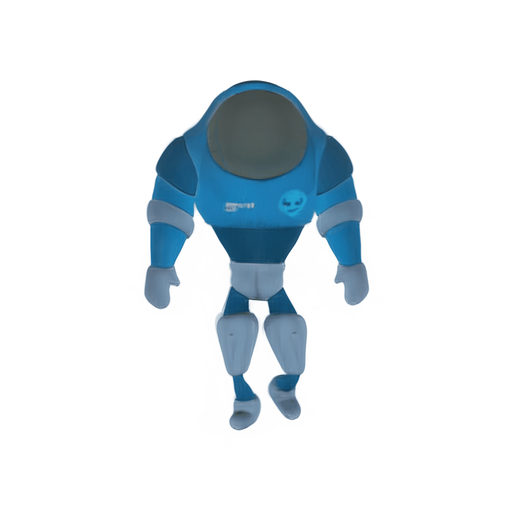}} \\[-0.5pt]
% --- GT Frame 0 ---
\raisebox{-0.5\height}{\rotatebox{90}{\tiny render $\angle 90^\circ$}} &
\raisebox{-0.5\height}{\includegraphics[width=0.16\columnwidth, trim=40 40 40 40, clip]{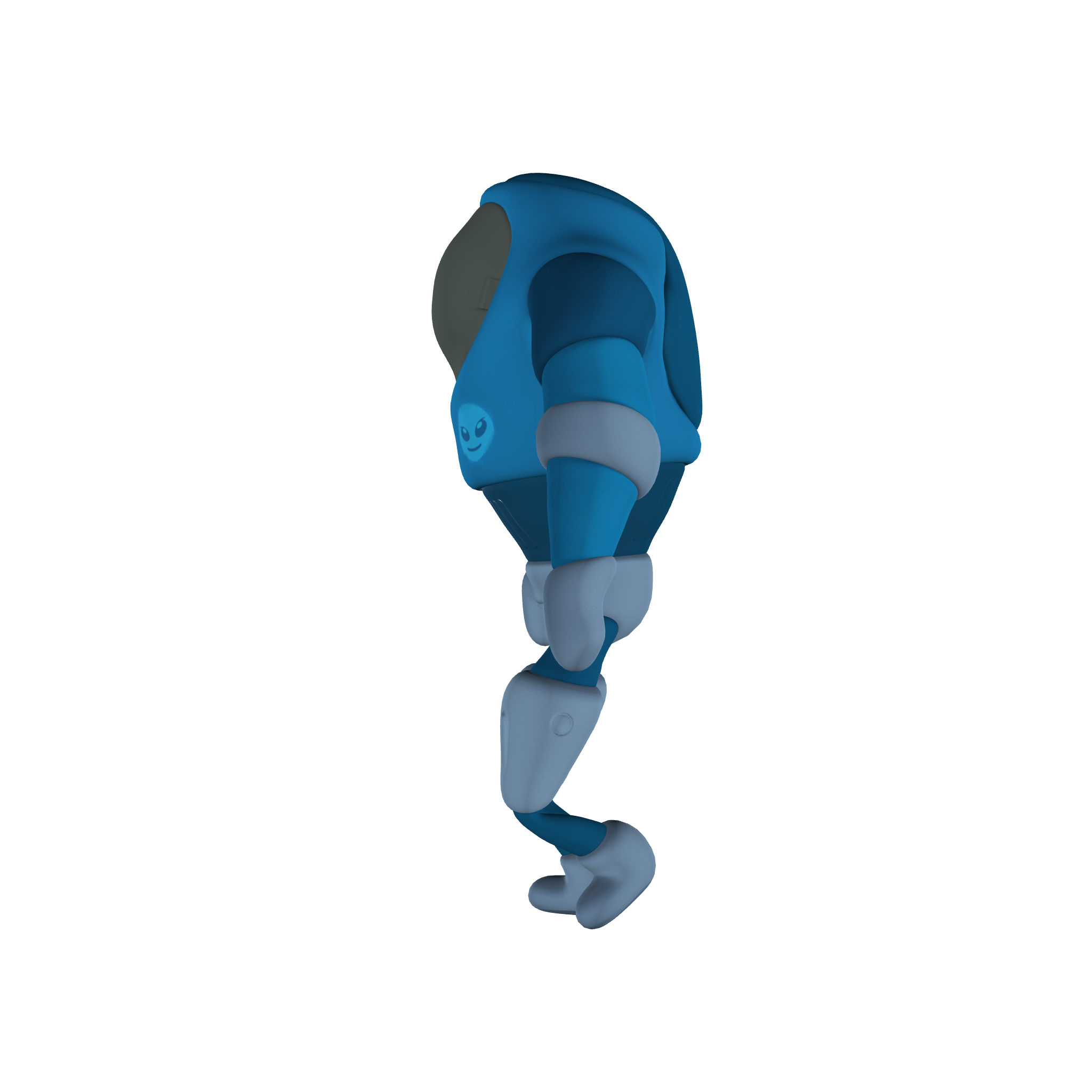}} & 
\raisebox{-0.5\height}{\includegraphics[width=0.16\columnwidth, trim=20 20 20 20, clip]{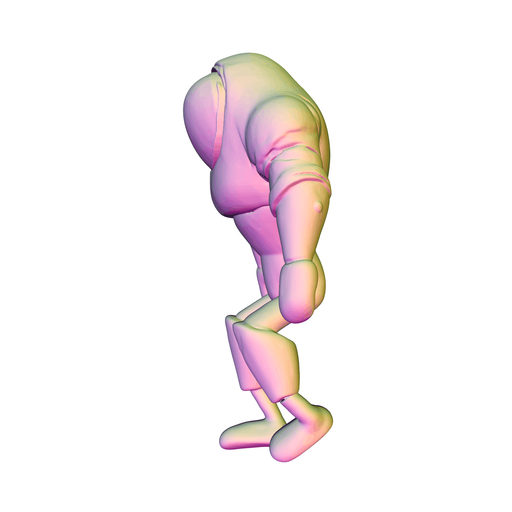}} &
\raisebox{-0.5\height}{\includegraphics[width=0.16\columnwidth, trim=40 40 40 40, clip]{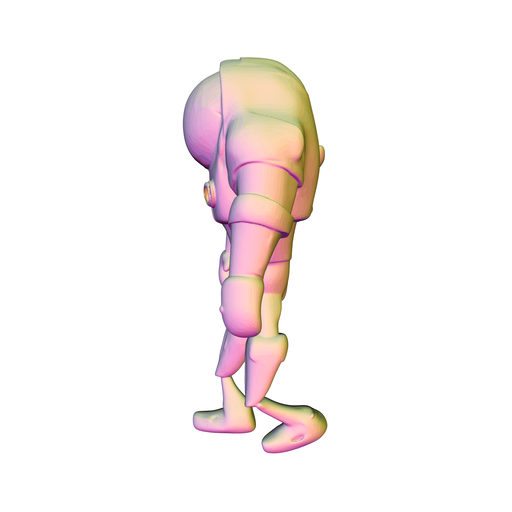}} &
\raisebox{-0.5\height}{\includegraphics[width=0.16\columnwidth, trim=40 40 40 40, clip]{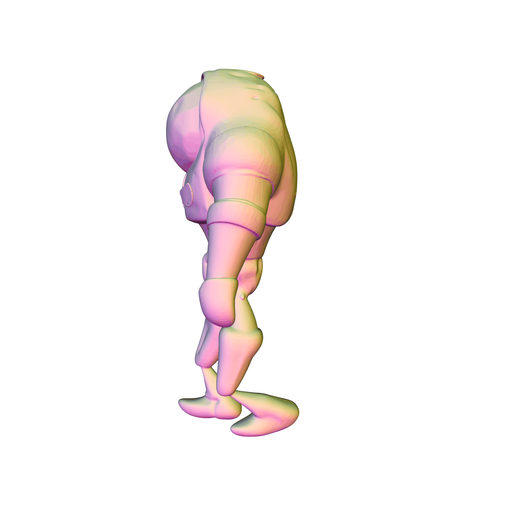}} &
\raisebox{-0.5\height}{\includegraphics[width=0.16\columnwidth, trim=40 40 40 40, clip]{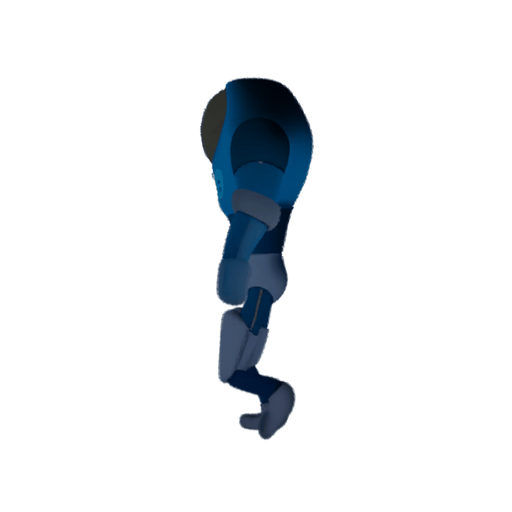}} &
\raisebox{-0.5\height}{\includegraphics[width=0.16\columnwidth, trim=40 40 40 40, clip]{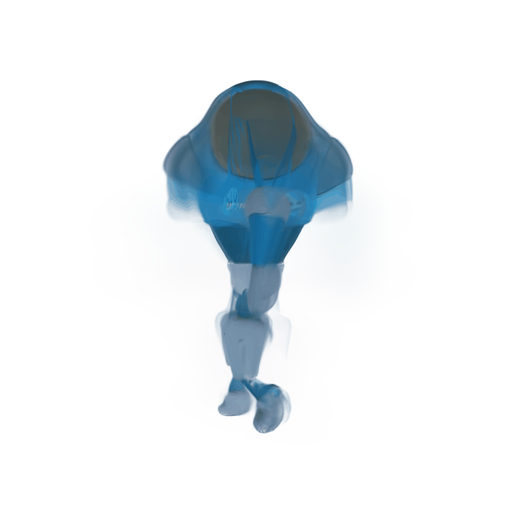}} \\[3pt]

% -------- Frame 8 --------
\raisebox{-0.5\height}{\rotatebox{90}{\tiny Input $\angle 0^\circ$}} &
\raisebox{-0.5\height}{\includegraphics[width=0.16\columnwidth, trim=40 40 40 40, clip]{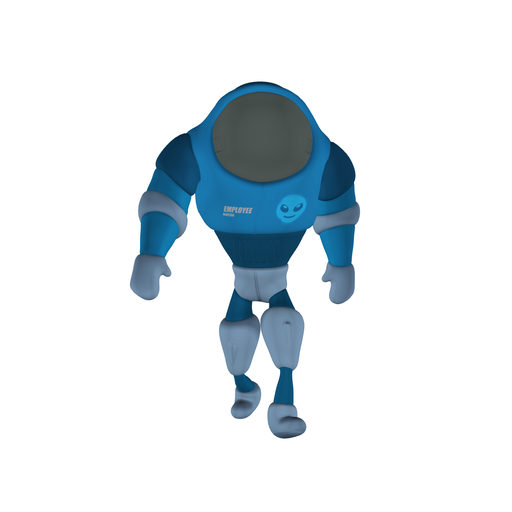}} &
\raisebox{-0.5\height}{\includegraphics[width=0.16\columnwidth, trim=20 20 20 20, clip]{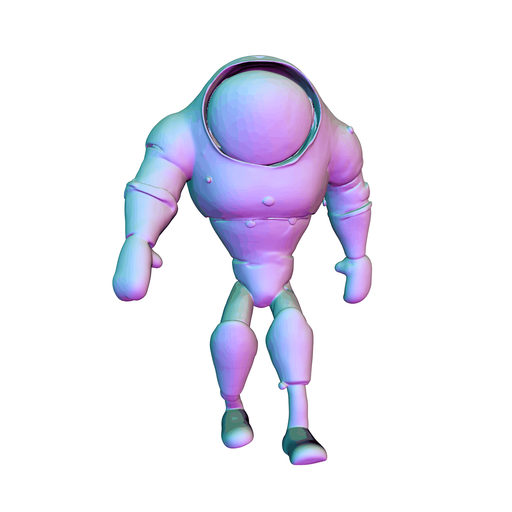}} &
\raisebox{-0.5\height}{\includegraphics[width=0.16\columnwidth, trim=40 40 40 40, clip]{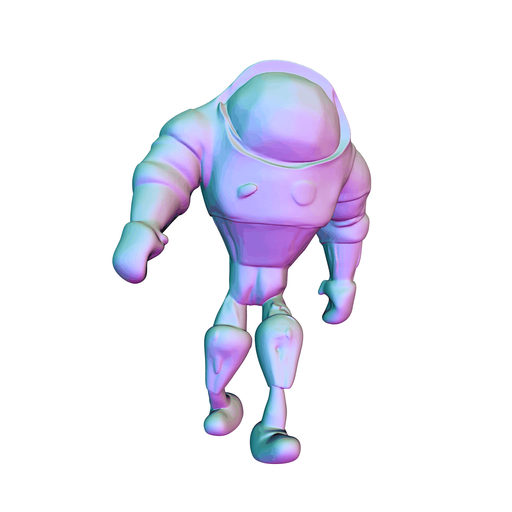}} &
\raisebox{-0.5\height}{\includegraphics[width=0.16\columnwidth, trim=40 40 40 40, clip]{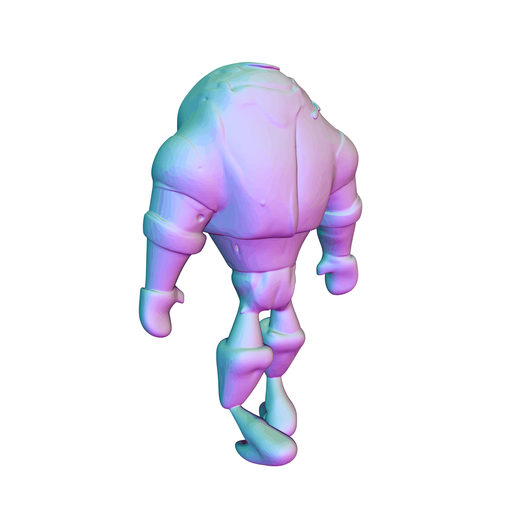}} &
\raisebox{-0.5\height}{\includegraphics[width=0.16\columnwidth, trim=40 40 40 40, clip]{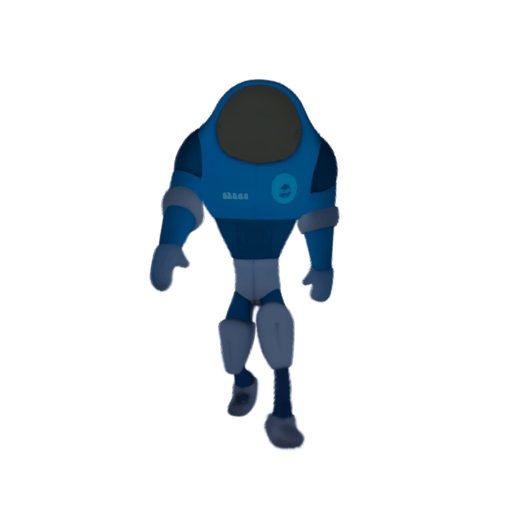}} &
\raisebox{-0.5\height}{\includegraphics[width=0.16\columnwidth, trim=40 40 40 40, clip]{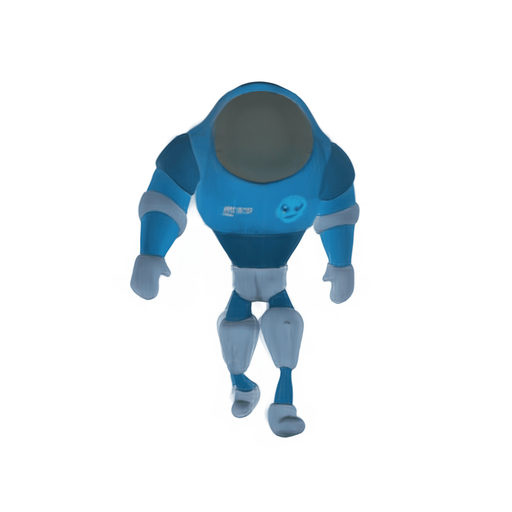}} \\[-0.5pt]
% --- GT Frame 8 ---
\raisebox{-0.5\height}{\rotatebox{90}{\tiny render $\angle 90^\circ$}} &
\raisebox{-0.5\height}{\includegraphics[width=0.16\columnwidth, trim=40 40 40 40, clip]{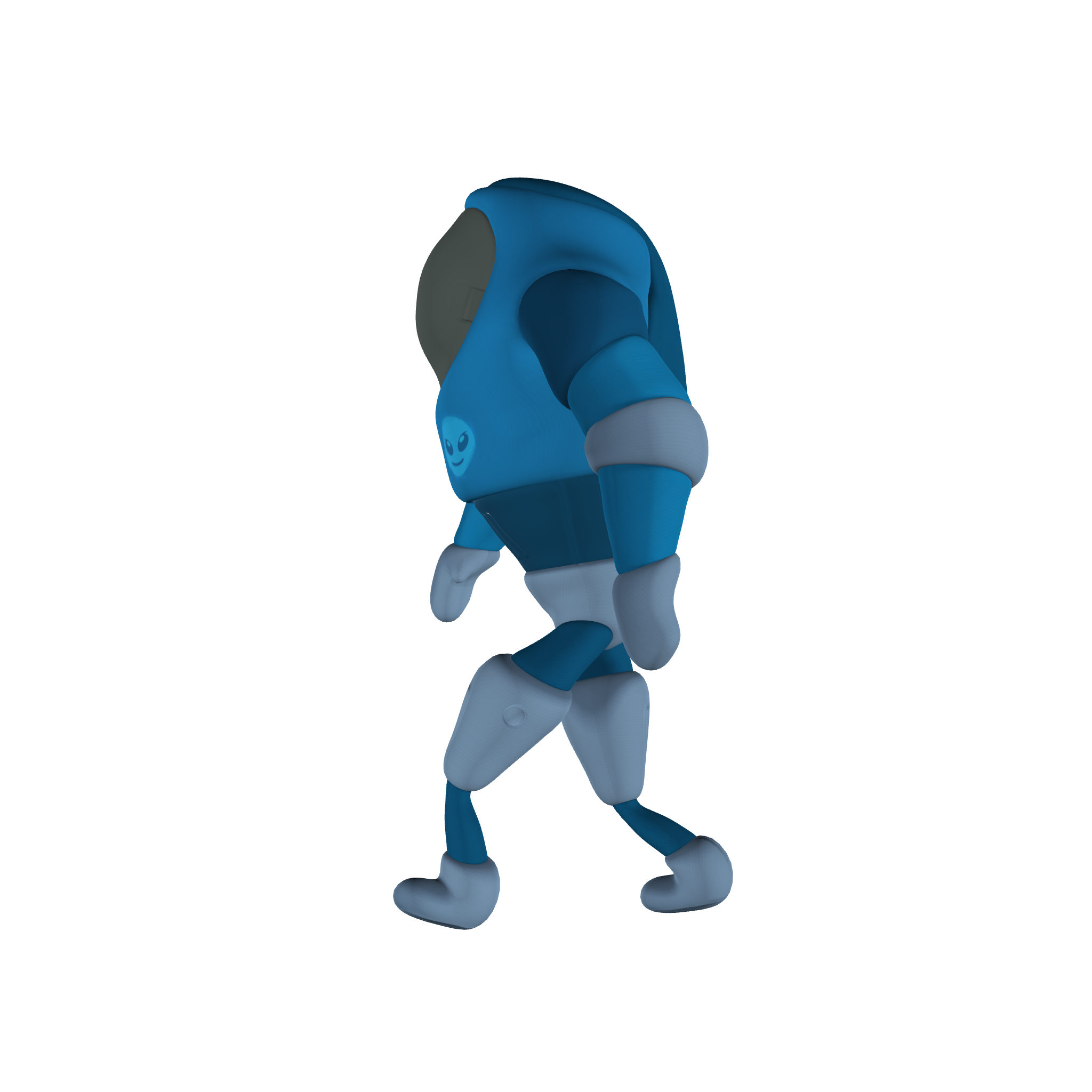}} & 
\raisebox{-0.5\height}{\includegraphics[width=0.16\columnwidth, trim=20 20 20 20, clip]{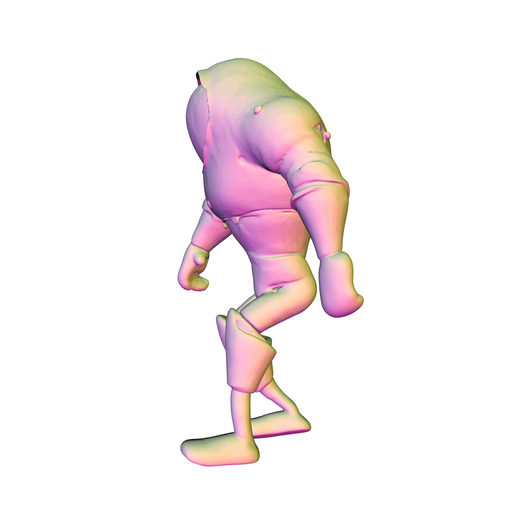}} &
\raisebox{-0.5\height}{\includegraphics[width=0.16\columnwidth, trim=40 40 40 40, clip]{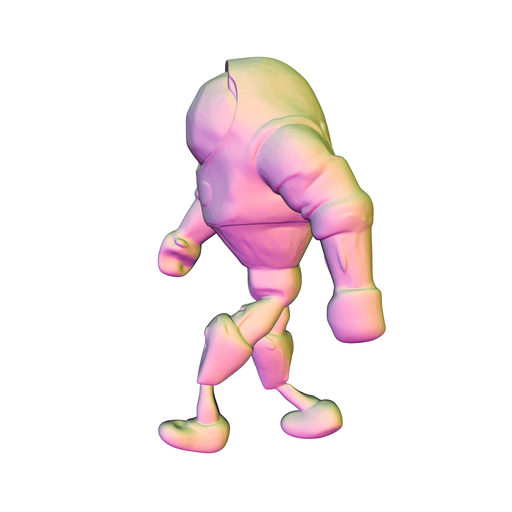}} &
\raisebox{-0.5\height}{\includegraphics[width=0.16\columnwidth, trim=40 40 40 40, clip]{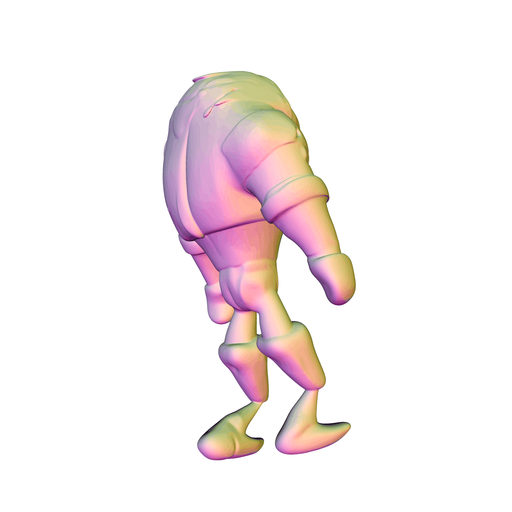}} &
\raisebox{-0.5\height}{\includegraphics[width=0.16\columnwidth, trim=40 40 40 40, clip]{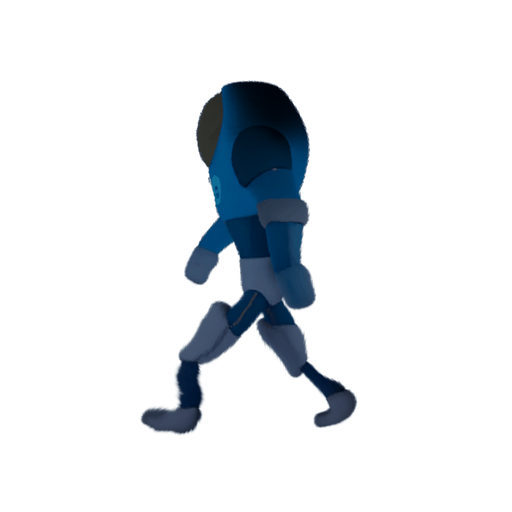}} &
\raisebox{-0.5\height}{\includegraphics[width=0.16\columnwidth, trim=40 40 40 40, clip]{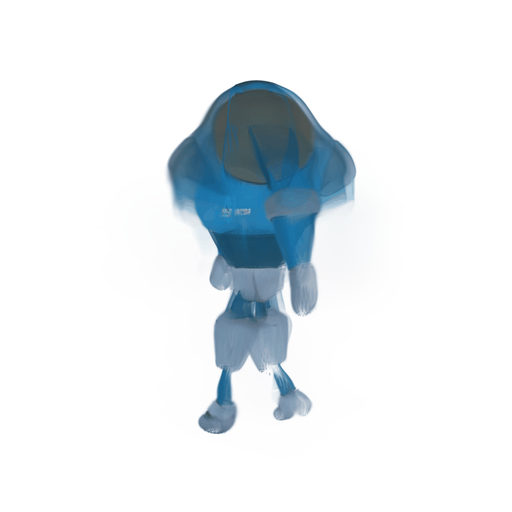}} \\[3pt]

% -------- Frame 15 --------
\raisebox{-0.5\height}{\rotatebox{90}{\tiny Input $\angle 0^\circ$}} &
\raisebox{-0.5\height}{\includegraphics[width=0.16\columnwidth, trim=40 40 40 40, clip]{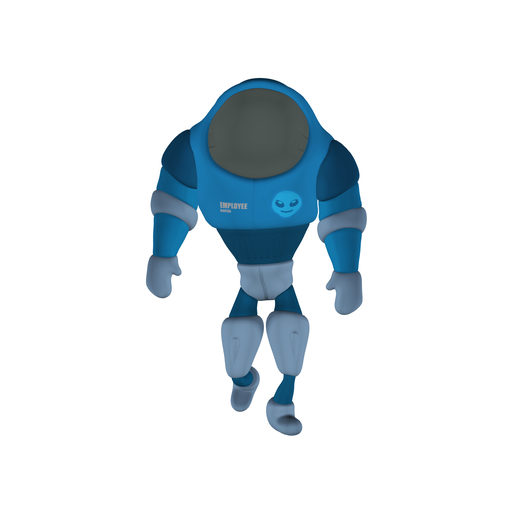}} &
\raisebox{-0.5\height}{\includegraphics[width=0.16\columnwidth, trim=20 20 20 20, clip]{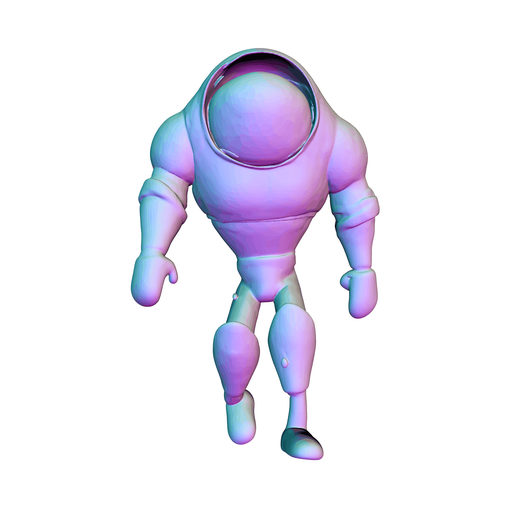}} &
\raisebox{-0.5\height}{\includegraphics[width=0.16\columnwidth, trim=40 40 40 40, clip]{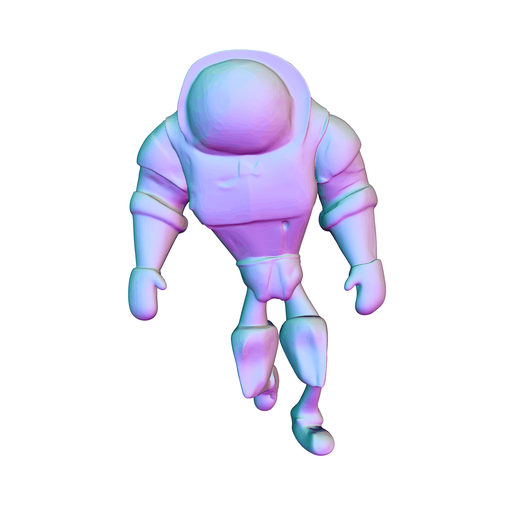}} &
\raisebox{-0.5\height}{\includegraphics[width=0.16\columnwidth, trim=40 40 40 40, clip]{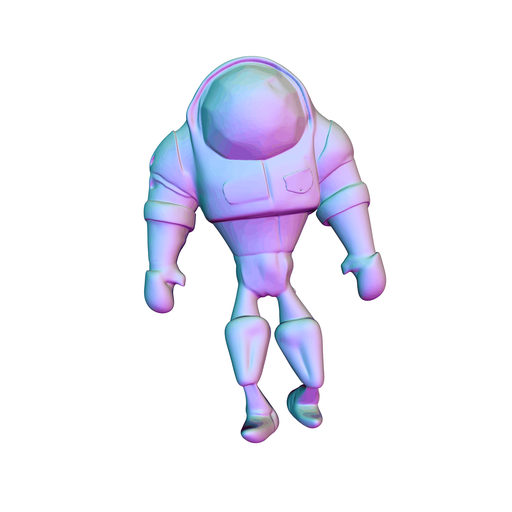}} &
\raisebox{-0.5\height}{\includegraphics[width=0.16\columnwidth, trim=40 40 40 40, clip]{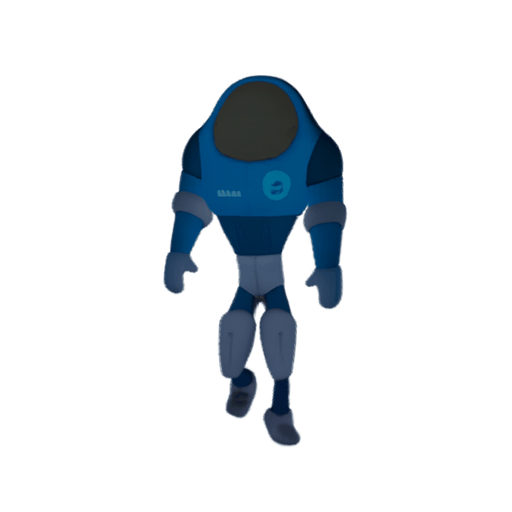}} &
\raisebox{-0.5\height}{\includegraphics[width=0.16\columnwidth, trim=40 40 40 40, clip]{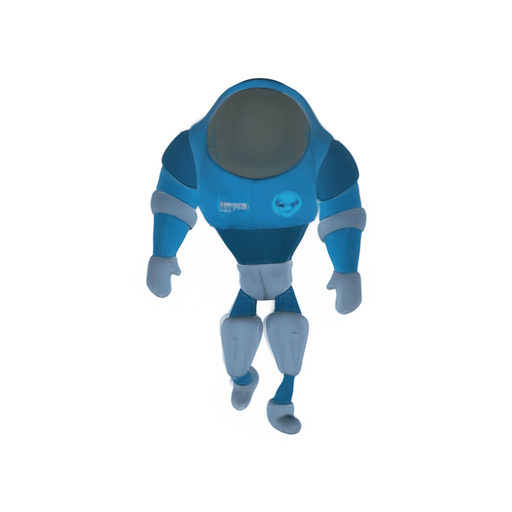}} \\[-0.5pt]
% --- GT Frame 15 ---
\raisebox{-0.5\height}{\rotatebox{90}{\tiny render $\angle 90^\circ$}} &
\raisebox{-0.5\height}{\includegraphics[width=0.16\columnwidth, trim=40 40 40 40, clip]{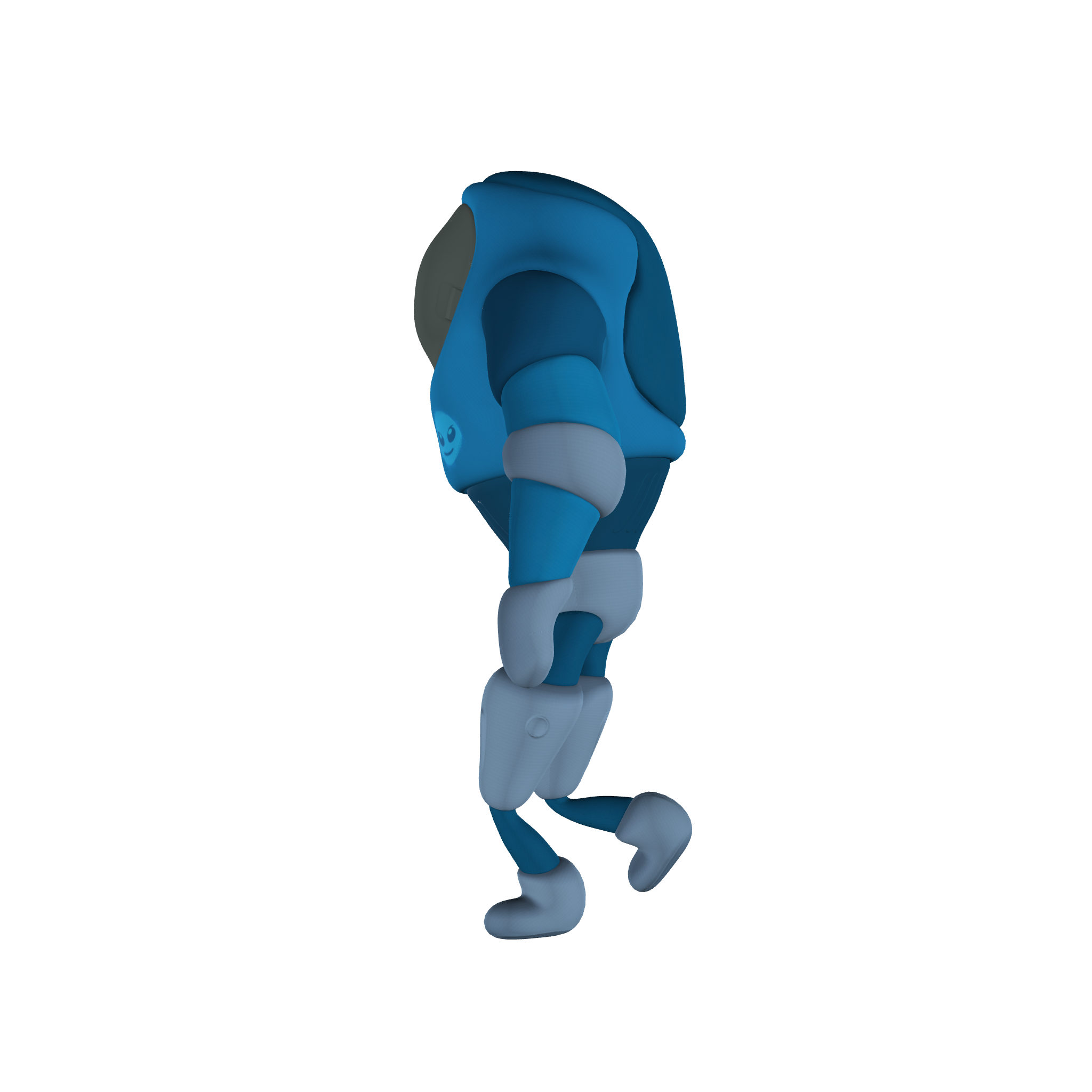}} & 
\raisebox{-0.5\height}{\includegraphics[width=0.16\columnwidth, trim=20 20 20 20, clip]{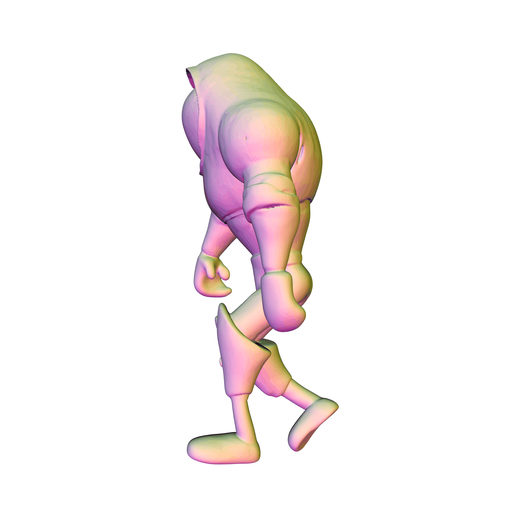}} &
\raisebox{-0.5\height}{\includegraphics[width=0.16\columnwidth, trim=40 40 40 40, clip]{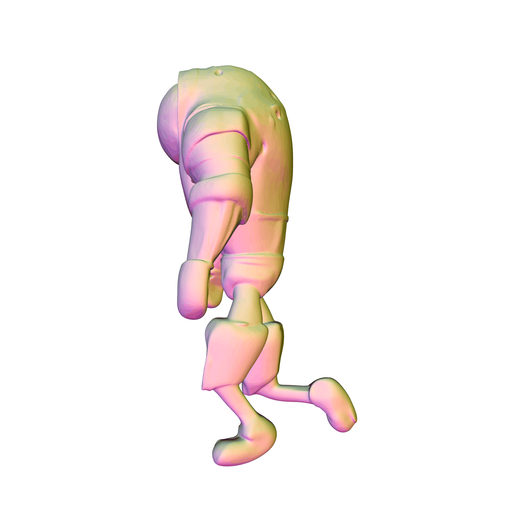}} &
\raisebox{-0.5\height}{\includegraphics[width=0.16\columnwidth, trim=40 40 40 40, clip]{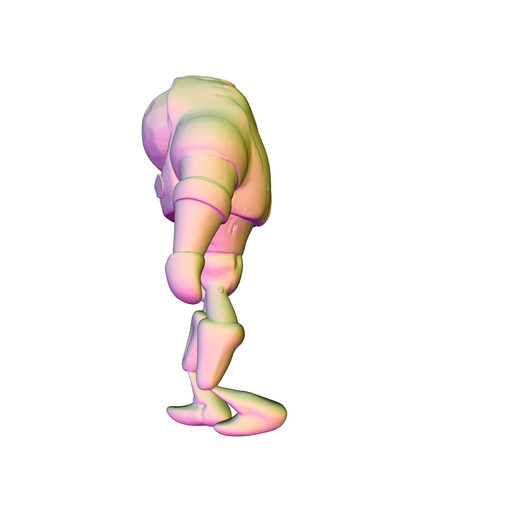}} &
\raisebox{-0.5\height}{\includegraphics[width=0.16\columnwidth, trim=40 40 40 40, clip]{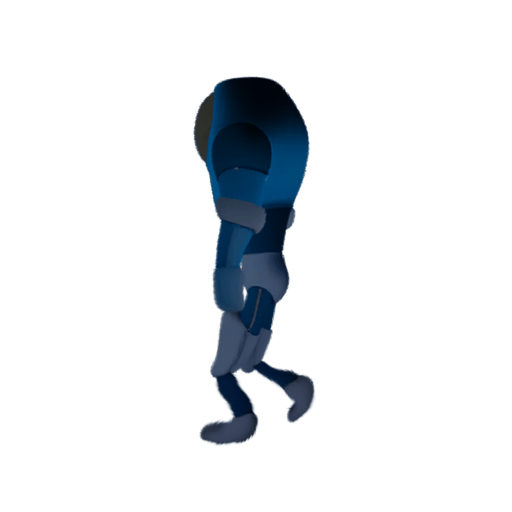}} &
\raisebox{-0.5\height}{\includegraphics[width=0.16\columnwidth, trim=40 40 40 40, clip]{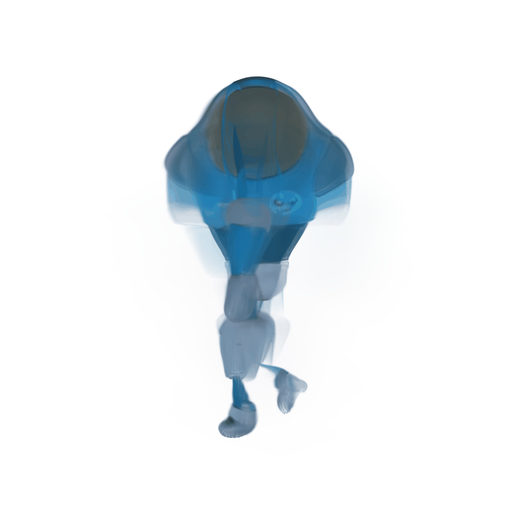}} \\[5pt]

% ==================== SUBJECT 6: 8723c78... ====================
% -------- Frame 0 --------
\raisebox{-0.5\height}{\rotatebox{90}{\tiny Input $\angle 0^\circ$}} &
\raisebox{-0.5\height}{\includegraphics[width=0.16\columnwidth, trim=40 40 40 40, clip]{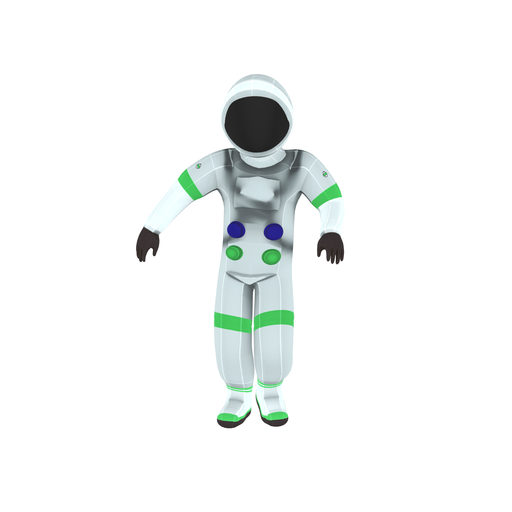}} &
\raisebox{-0.5\height}{\includegraphics[width=0.16\columnwidth, trim=20 20 20 20, clip]{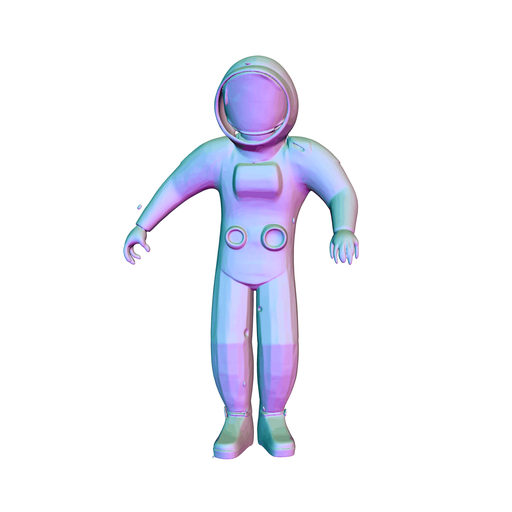}} &
\raisebox{-0.5\height}{\includegraphics[width=0.16\columnwidth, trim=40 40 40 40, clip]{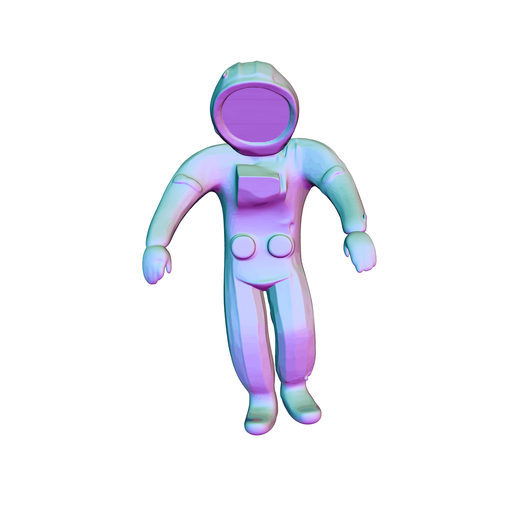}} &
\raisebox{-0.5\height}{\includegraphics[width=0.16\columnwidth, trim=40 40 40 40, clip]{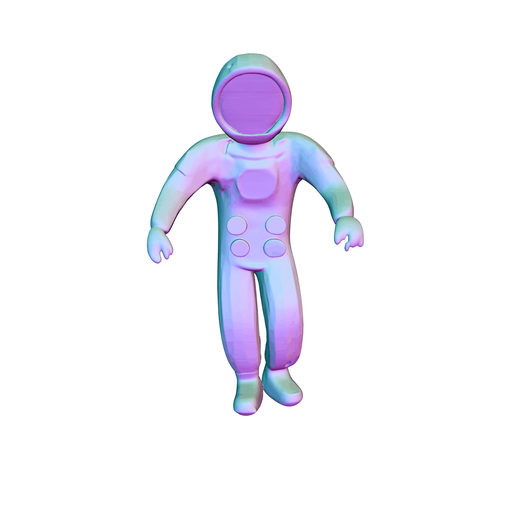}} &
\raisebox{-0.5\height}{\includegraphics[width=0.16\columnwidth, trim=40 40 40 40, clip]{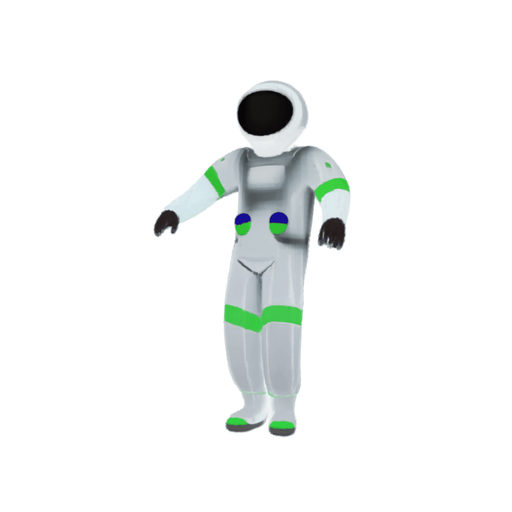}} &
\raisebox{-0.5\height}{\includegraphics[width=0.16\columnwidth, trim=40 40 40 40, clip]{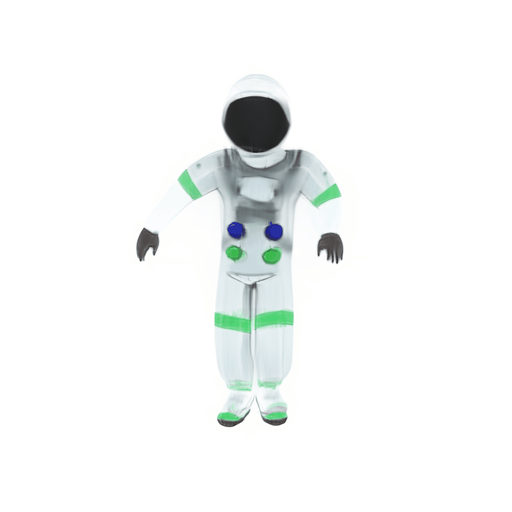}} \\[-0.5pt]
% --- GT Frame 0 ---
\raisebox{-0.5\height}{\rotatebox{90}{\tiny render $\angle 90^\circ$}} &
\raisebox{-0.5\height}{\includegraphics[width=0.16\columnwidth, trim=40 40 40 40, clip]{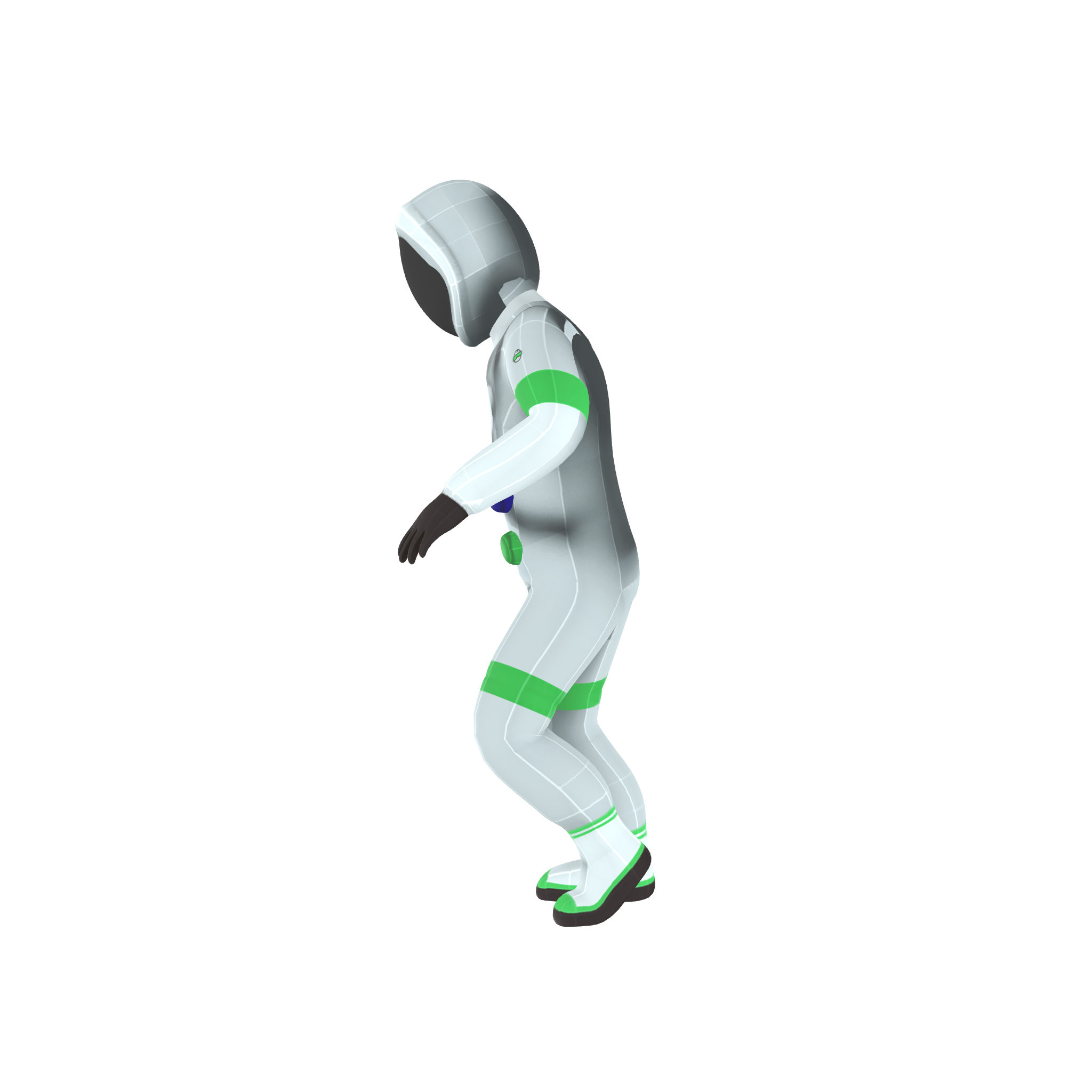}} & 
\raisebox{-0.5\height}{\includegraphics[width=0.16\columnwidth, trim=20 20 20 20, clip]{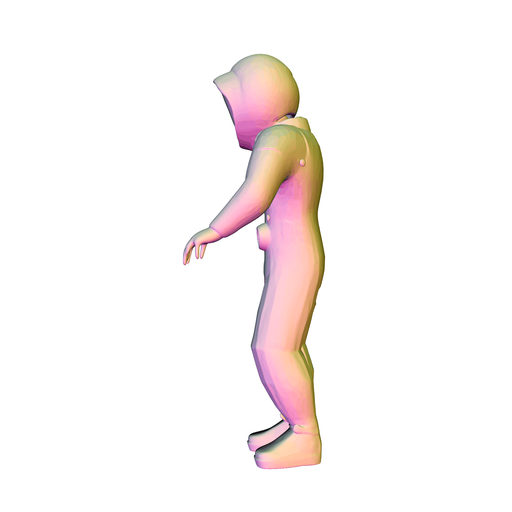}} &
\raisebox{-0.5\height}{\includegraphics[width=0.16\columnwidth, trim=40 40 40 40, clip]{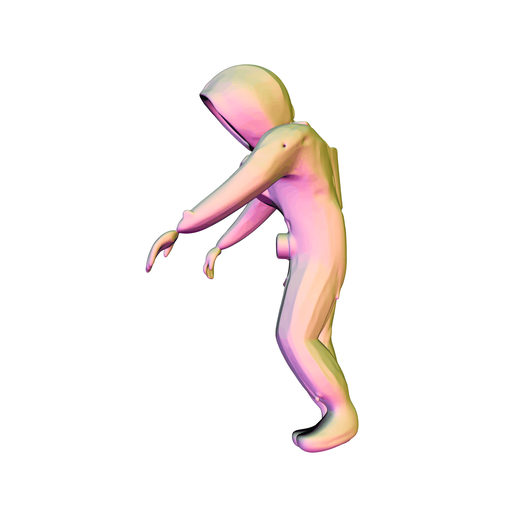}} &
\raisebox{-0.5\height}{\includegraphics[width=0.16\columnwidth, trim=40 40 40 40, clip]{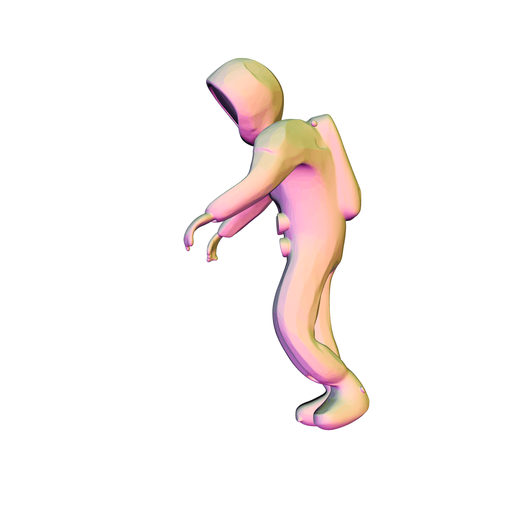}} &
\raisebox{-0.5\height}{\includegraphics[width=0.16\columnwidth, trim=40 40 40 40, clip]{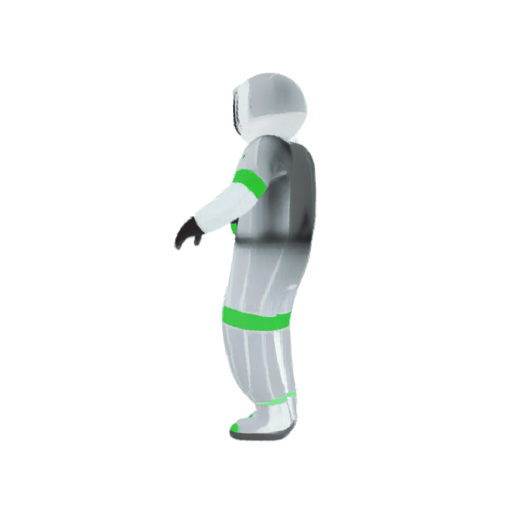}} &
\raisebox{-0.5\height}{\includegraphics[width=0.16\columnwidth, trim=40 40 40 40, clip]{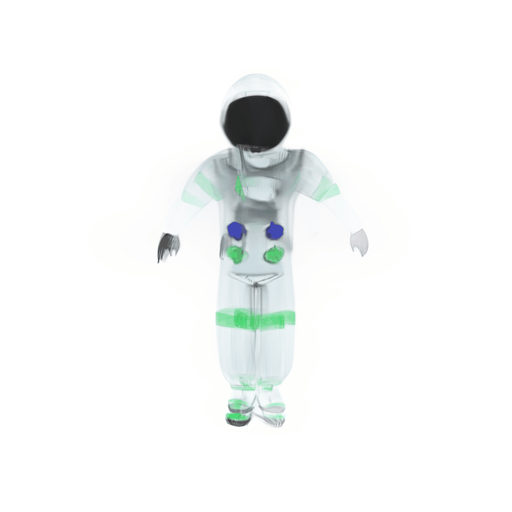}} \\[3pt]

% -------- Frame 8 --------
\raisebox{-0.5\height}{\rotatebox{90}{\tiny Input $\angle 0^\circ$}} &
\raisebox{-0.5\height}{\includegraphics[width=0.16\columnwidth, trim=40 40 40 40, clip]{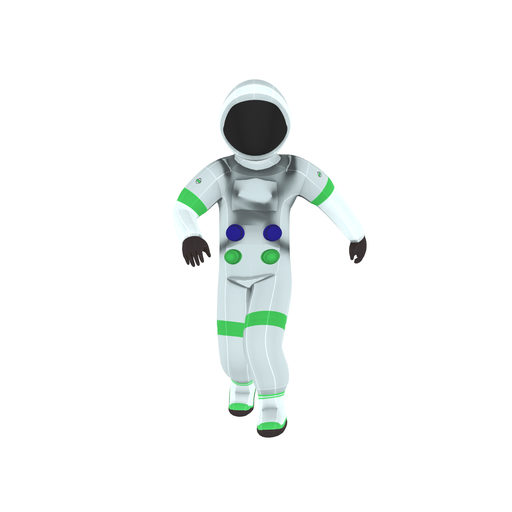}} &
\raisebox{-0.5\height}{\includegraphics[width=0.16\columnwidth, trim=20 20 20 20, clip]{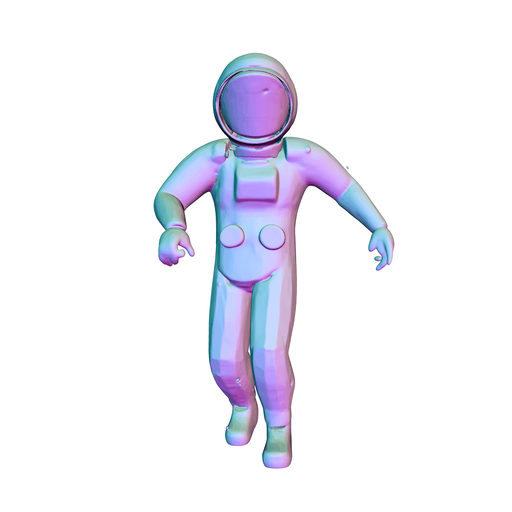}} &
\raisebox{-0.5\height}{\includegraphics[width=0.16\columnwidth, trim=40 40 40 40, clip]{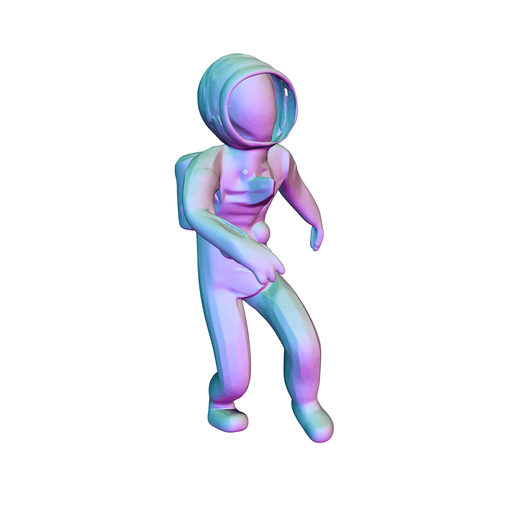}} &
\raisebox{-0.5\height}{\includegraphics[width=0.16\columnwidth, trim=40 40 40 40, clip]{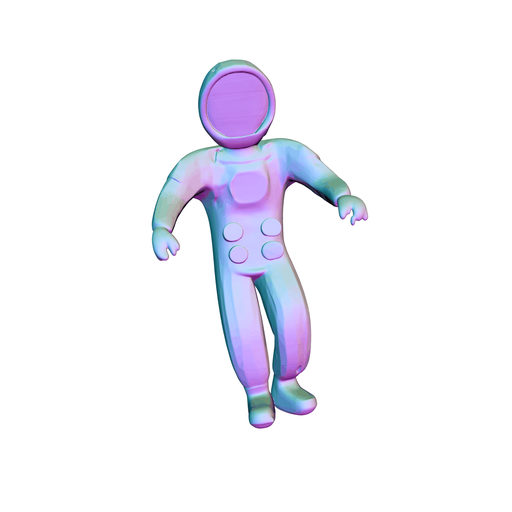}} &
\raisebox{-0.5\height}{\includegraphics[width=0.16\columnwidth, trim=40 40 40 40, clip]{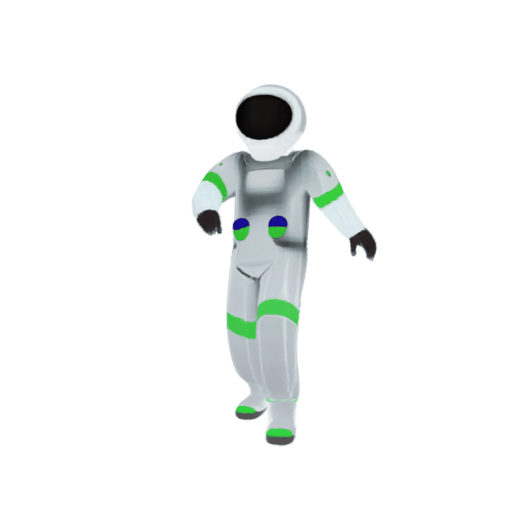}} &
\raisebox{-0.5\height}{\includegraphics[width=0.16\columnwidth, trim=40 40 40 40, clip]{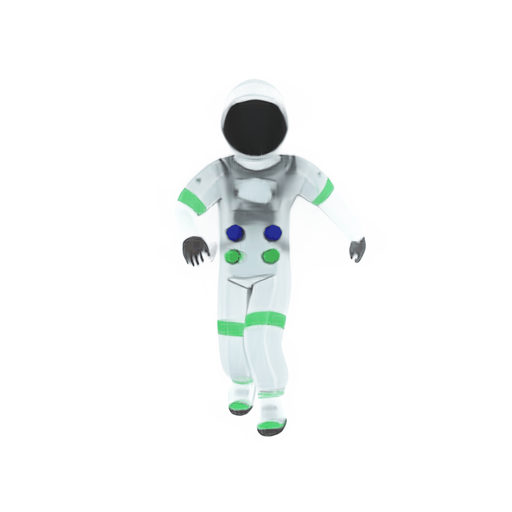}} \\[-0.5pt]
% --- GT Frame 8 ---
\raisebox{-0.5\height}{\rotatebox{90}{\tiny render $\angle 90^\circ$}} &
\raisebox{-0.5\height}{\includegraphics[width=0.16\columnwidth, trim=40 40 40 40, clip]{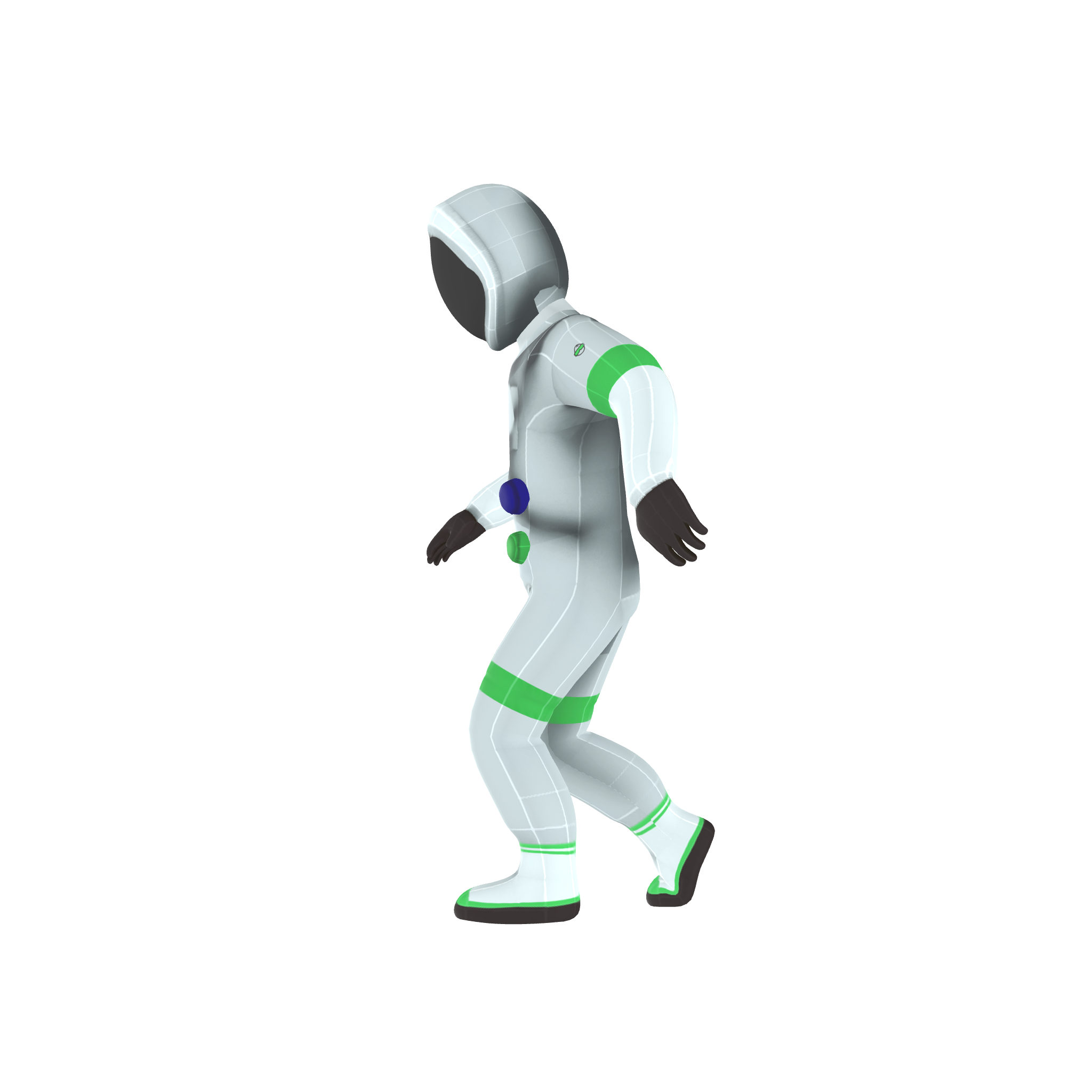}} & 
\raisebox{-0.5\height}{\includegraphics[width=0.16\columnwidth, trim=20 20 20 20, clip]{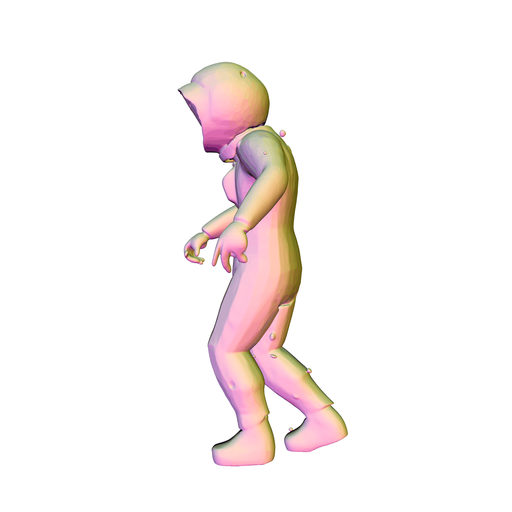}} &
\raisebox{-0.5\height}{\includegraphics[width=0.16\columnwidth, trim=40 40 40 40, clip]{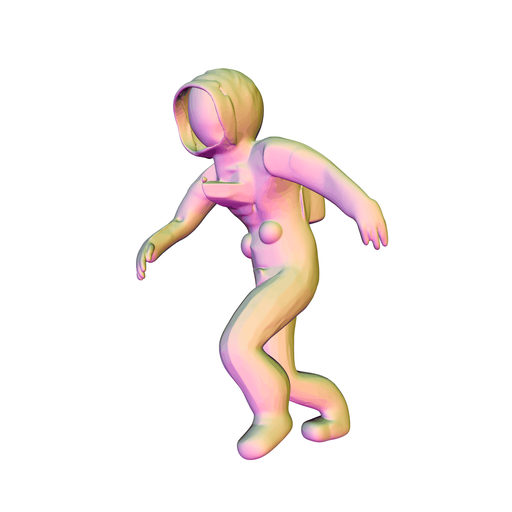}} &
\raisebox{-0.5\height}{\includegraphics[width=0.16\columnwidth, trim=40 40 40 40, clip]{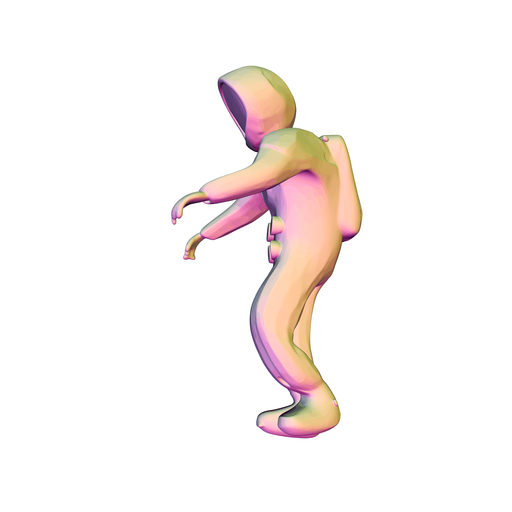}} &
\raisebox{-0.5\height}{\includegraphics[width=0.16\columnwidth, trim=40 40 40 40, clip]{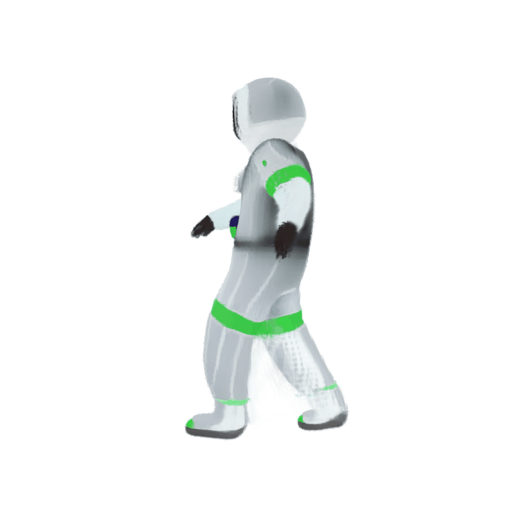}} &
\raisebox{-0.5\height}{\includegraphics[width=0.16\columnwidth, trim=40 40 40 40, clip]{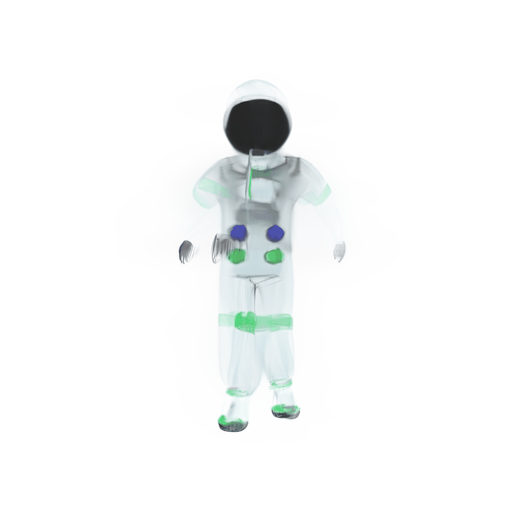}} \\[3pt]

% -------- Frame 15 --------
\raisebox{-0.5\height}{\rotatebox{90}{\tiny Input $\angle 0^\circ$}} &
\raisebox{-0.5\height}{\includegraphics[width=0.16\columnwidth, trim=40 40 40 40, clip]{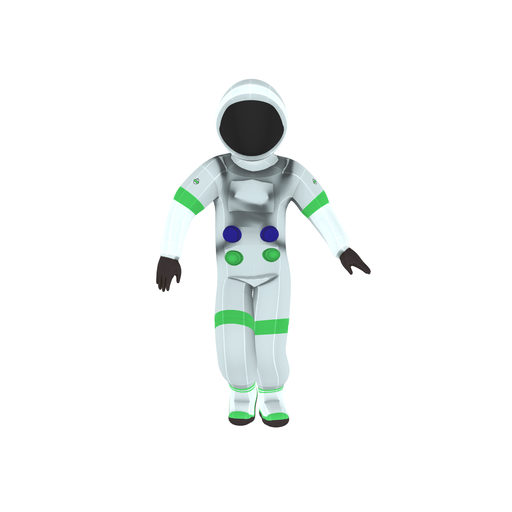}} &
\raisebox{-0.5\height}{\includegraphics[width=0.16\columnwidth, trim=20 20 20 20, clip]{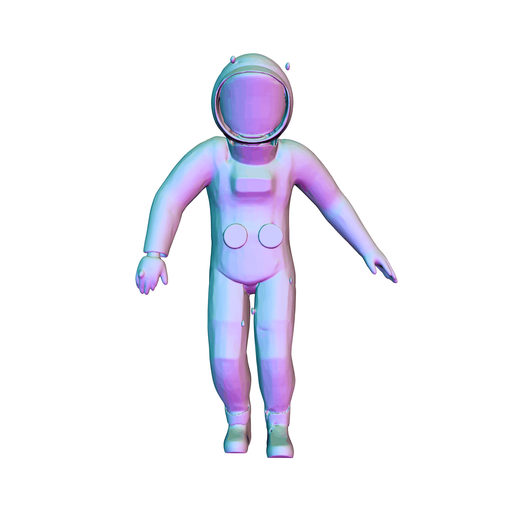}} &
\raisebox{-0.5\height}{\includegraphics[width=0.16\columnwidth, trim=40 40 40 40, clip]{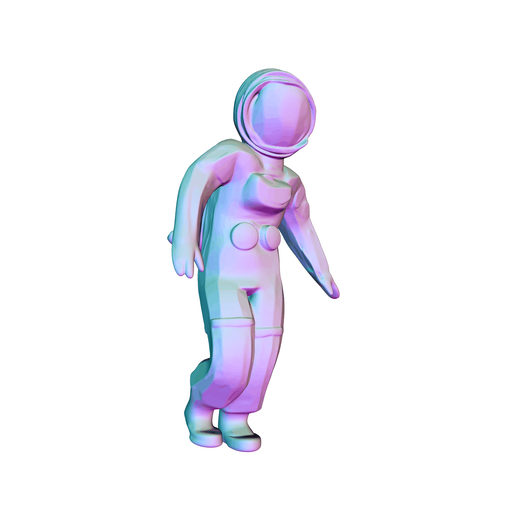}} &
\raisebox{-0.5\height}{\includegraphics[width=0.16\columnwidth, trim=40 40 40 40, clip]{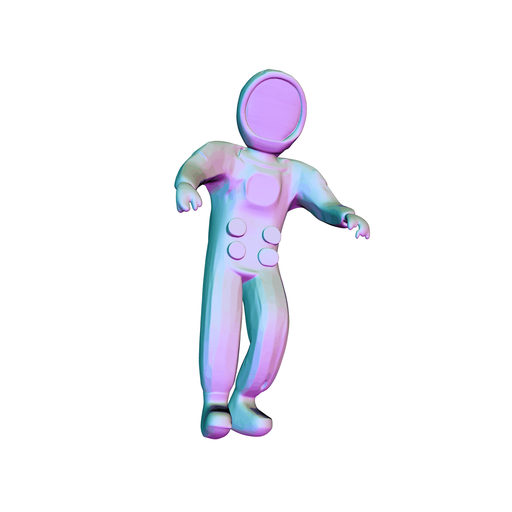}} &
\raisebox{-0.5\height}{\includegraphics[width=0.16\columnwidth, trim=40 40 40 40, clip]{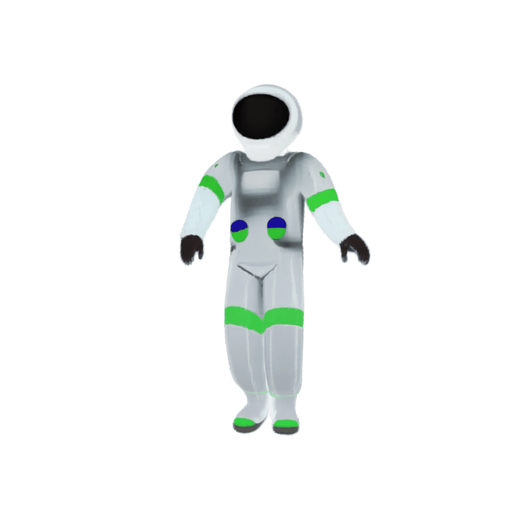}} &
\raisebox{-0.5\height}{\includegraphics[width=0.16\columnwidth, trim=40 40 40 40, clip]{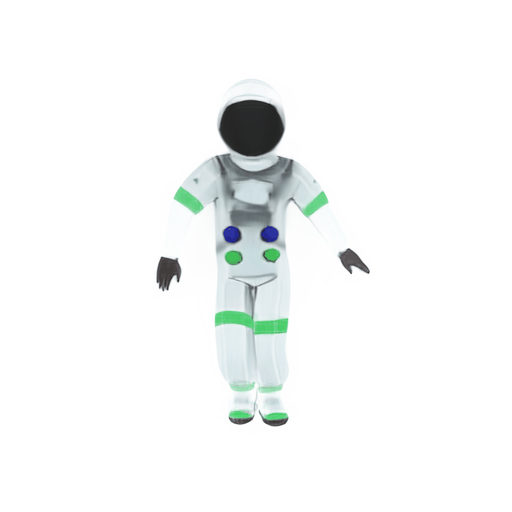}} \\[-0.5pt]
% --- GT Frame 15 ---
\raisebox{-0.5\height}{\rotatebox{90}{\tiny render $\angle 90^\circ$}} &
\raisebox{-0.5\height}{\includegraphics[width=0.16\columnwidth, trim=40 40 40 40, clip]{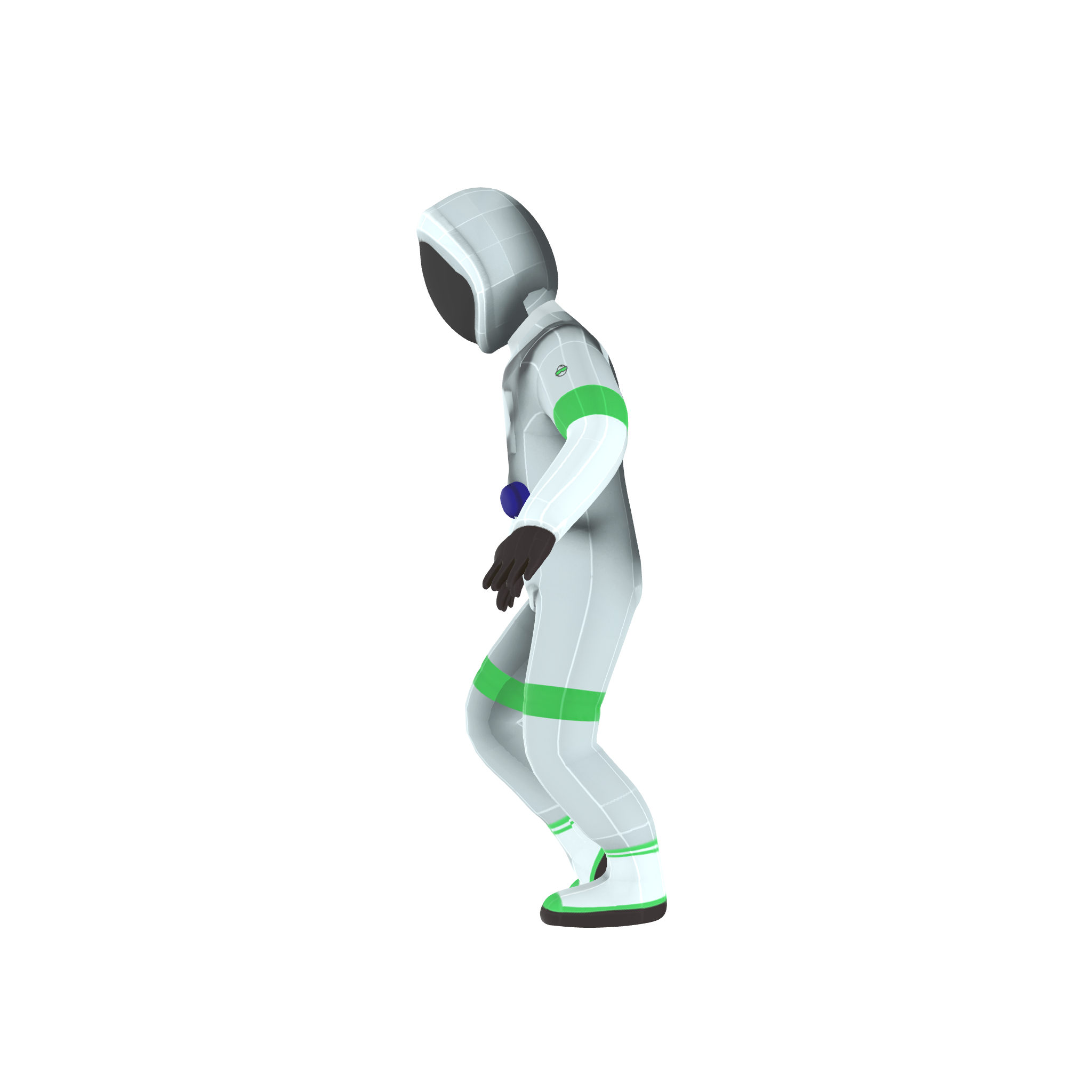}} & 
\raisebox{-0.5\height}{\includegraphics[width=0.16\columnwidth, trim=20 20 20 20, clip]{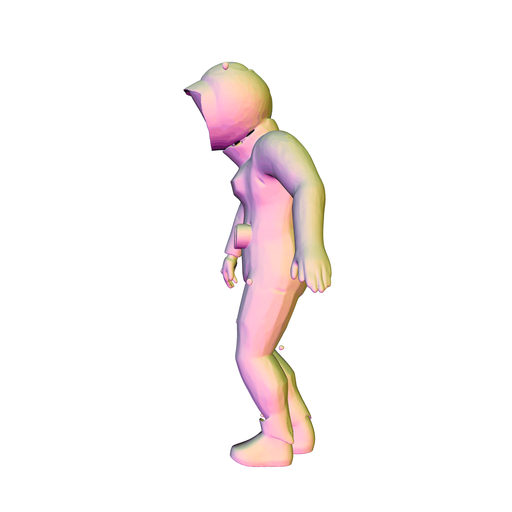}} &
\raisebox{-0.5\height}{\includegraphics[width=0.16\columnwidth, trim=40 40 40 40, clip]{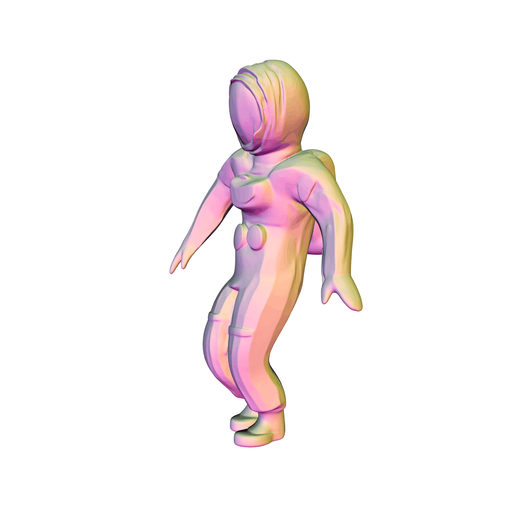}} &
\raisebox{-0.5\height}{\includegraphics[width=0.16\columnwidth, trim=40 40 40 40, clip]{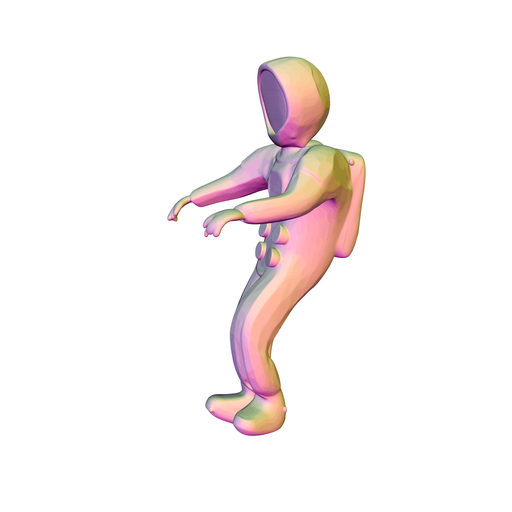}} &
\raisebox{-0.5\height}{\includegraphics[width=0.16\columnwidth, trim=40 40 40 40, clip]{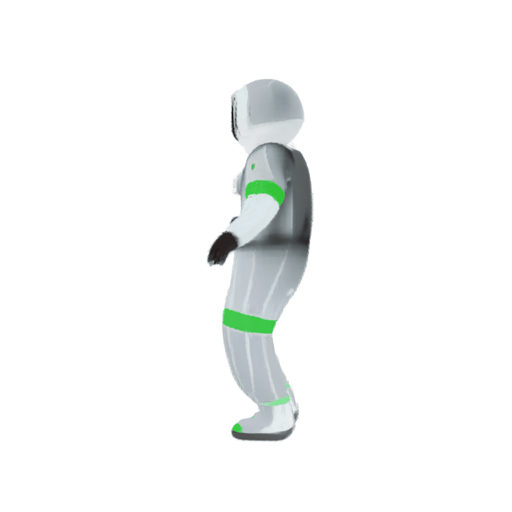}} &
\raisebox{-0.5\height}{\includegraphics[width=0.16\columnwidth, trim=40 40 40 40, clip]{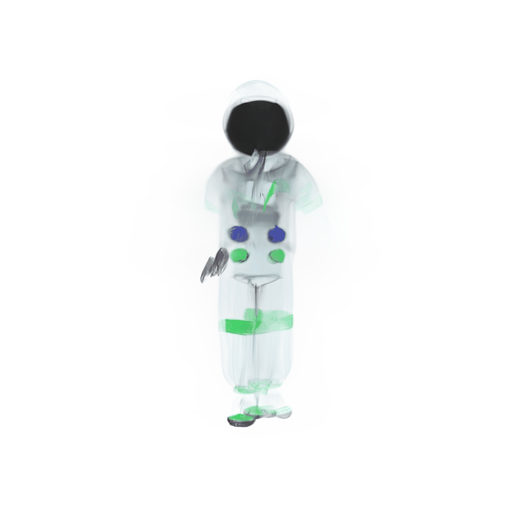}} \\

  \end{tabular}
  % ==================== TABLE END ====================
  
  \vspace{-3pt}
  \caption{
  Further qualitative 4D generation comparisons from Objaverse~\cite{deitke2022objaverseuniverseannotated3d} showing two subjects of Objaverse~\cite{deitke2022objaverseuniverseannotated3d}, at three time steps. For each model, we show the reconstructed input view (top) and a rendered novel view (bottom). We display the ground truth novel view in the bottom left. The novel view in particular highlights the discrepancies of the methods' outputs from the ground truth. Both TripoSG and V2M4 show moderate and consistent misalignment. Similar misalignment, particularly in rotation and skeletal pose is apparent in GVFD, while L4GM often fails to provide good novel views. Typically, our method shows far less shape or pose misalignment, as reflected in the quantitative metrics.
  }
  \label{fig:qual_comp_4d_appendix_5_6_full}
\end{figure}

\begin{figure}[h]
  \centering
  % --- MODIFICATIONS ---
  \setlength{\tabcolsep}{0pt} % Zero horizontal spacing between images
  \renewcommand{\arraystretch}{0} % Remove default extra padding in rows

  % ==================== TABLE START ====================
  % Added a column 'c' at the start for vertical text
  \begin{tabular}{@{}c@{\hspace{2pt}}ccccc@{}}
    % --- HEADERS ---
    & % Empty cell for the vertical text column
    \small{Input / GT} &
    \textbf{\small{Ours}} &
    \small{TripoSG} &
    \small{GVFD} &
    \small{L4GM} \\[3pt] % Keep a small gap between titles and images

% ==================== SUBJECT: deerA4K_sleep ====================
% -------- Frame 0 --------
\raisebox{-0.5\height}{\rotatebox{90}{\tiny Input $\angle 0^\circ$}} &
\raisebox{-0.5\height}{\includegraphics[width=0.19\columnwidth, trim=40 40 40 40, clip]{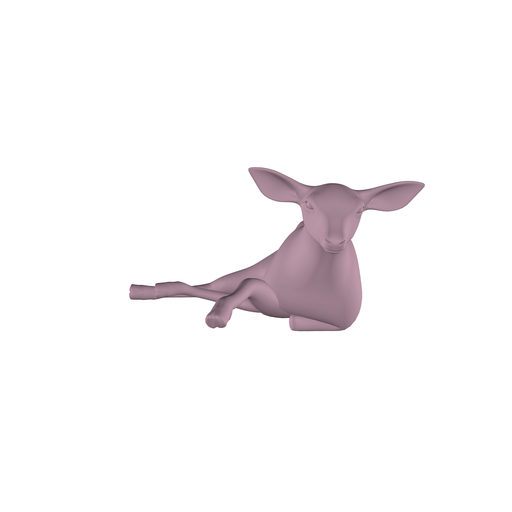}} &
\raisebox{-0.5\height}{\includegraphics[width=0.19\columnwidth, trim=20 20 20 20, clip]{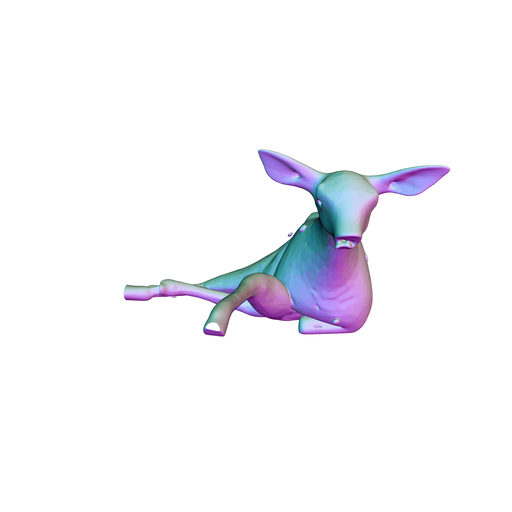}} &
\raisebox{-0.5\height}{\includegraphics[width=0.19\columnwidth, trim=40 40 40 40, clip]{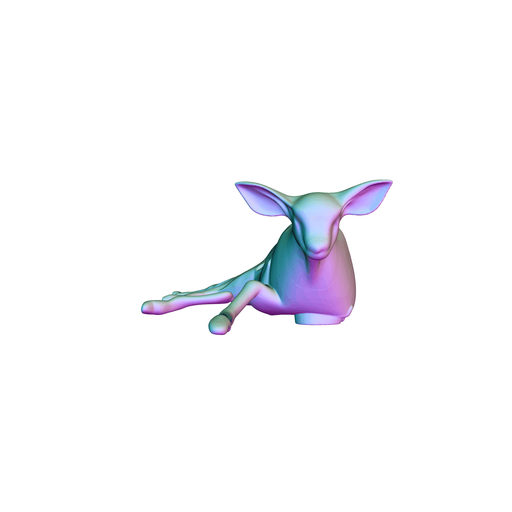}} &
\raisebox{-0.5\height}{\includegraphics[width=0.19\columnwidth, trim=40 40 40 40, clip]{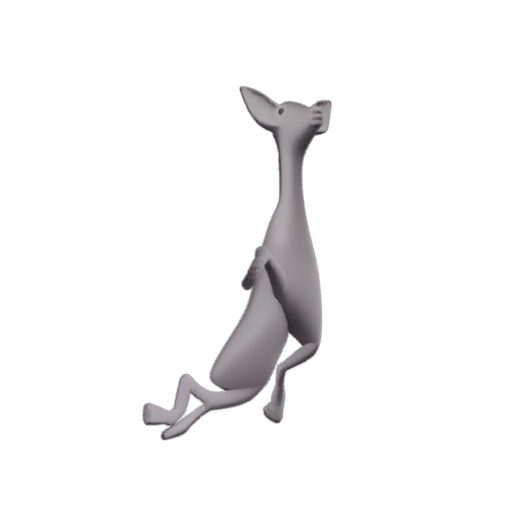}} &
\raisebox{-0.5\height}{\includegraphics[width=0.19\columnwidth, trim=40 40 40 40, clip]{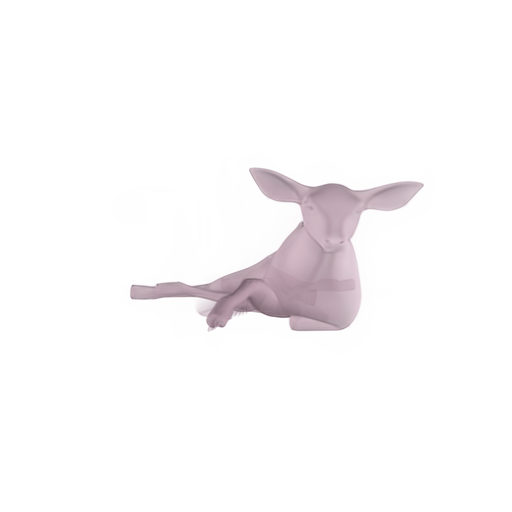}} \\[-0.5pt]
% --- GT Frame 0 ---
\raisebox{-0.5\height}{\rotatebox{90}{\tiny render $\angle 90^\circ$}} &
\raisebox{-0.5\height}{\includegraphics[width=0.19\columnwidth, trim=40 40 40 40, clip]{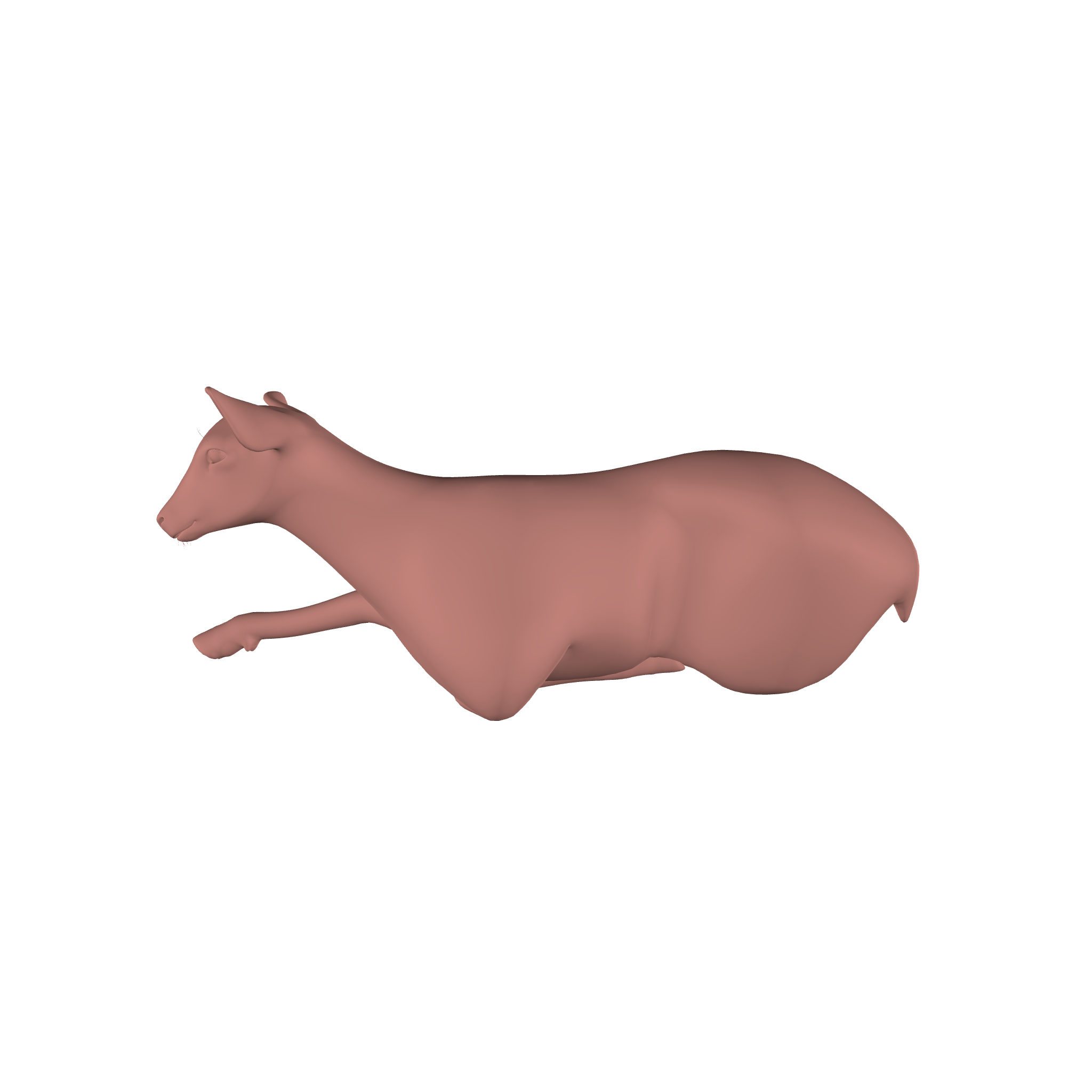}} & 
\raisebox{-0.5\height}{\includegraphics[width=0.19\columnwidth, trim=20 20 20 20, clip]{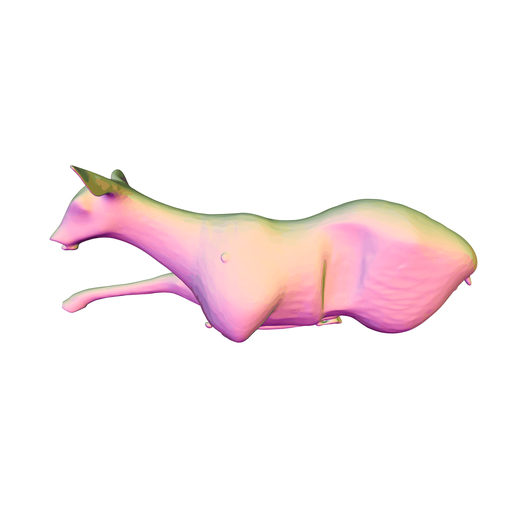}} &
\raisebox{-0.5\height}{\includegraphics[width=0.19\columnwidth, trim=40 40 40 40, clip]{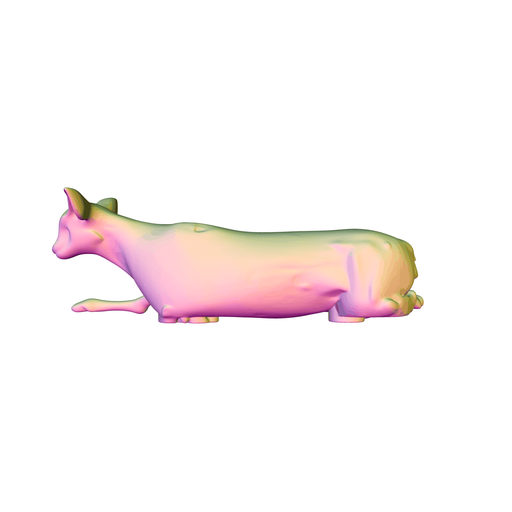}} &
\raisebox{-0.5\height}{\includegraphics[width=0.19\columnwidth, trim=40 40 40 40, clip]{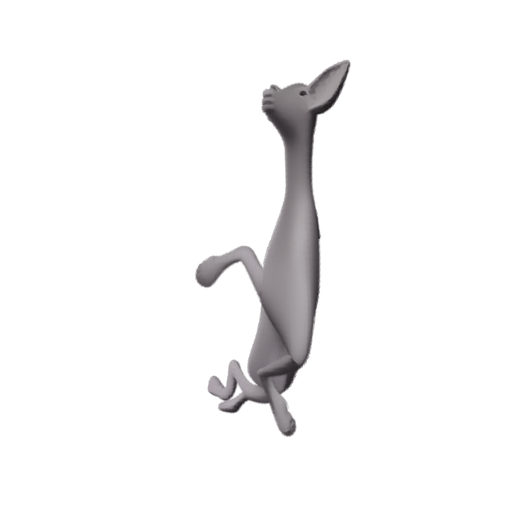}} &
\raisebox{-0.5\height}{\includegraphics[width=0.19\columnwidth, trim=40 40 40 40, clip]{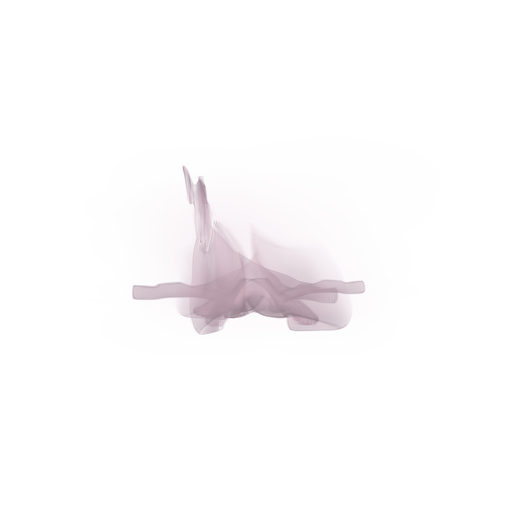}} \\[3pt]

% -------- Frame 8 --------
\raisebox{-0.5\height}{\rotatebox{90}{\tiny Input $\angle 0^\circ$}} &
\raisebox{-0.5\height}{\includegraphics[width=0.19\columnwidth, trim=40 40 40 40, clip]{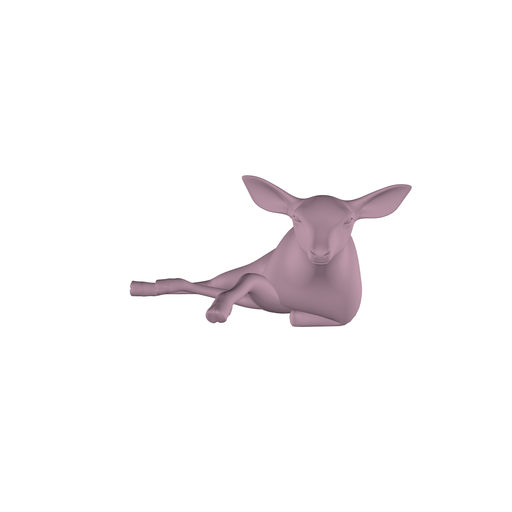}} &
\raisebox{-0.5\height}{\includegraphics[width=0.19\columnwidth, trim=20 20 20 20, clip]{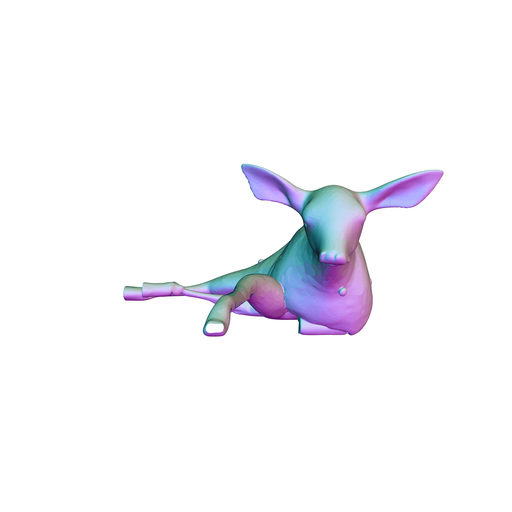}} &
\raisebox{-0.5\height}{\includegraphics[width=0.19\columnwidth, trim=40 40 40 40, clip]{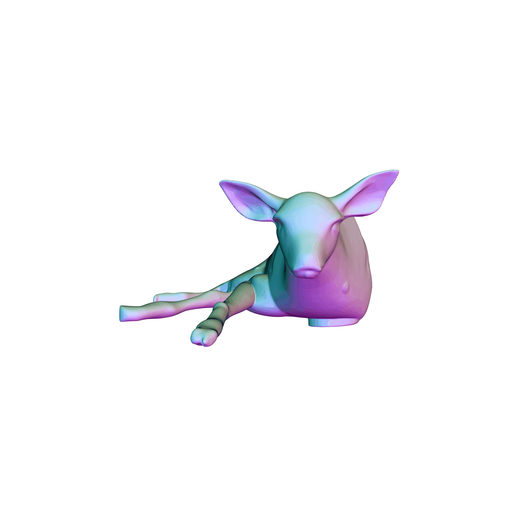}} &
\raisebox{-0.5\height}{\includegraphics[width=0.19\columnwidth, trim=40 40 40 40, clip]{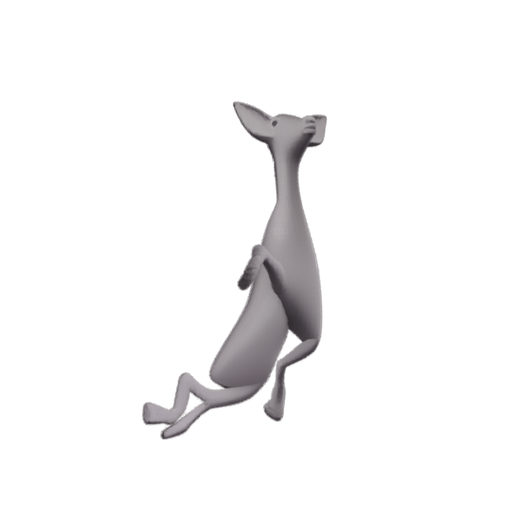}} &
\raisebox{-0.5\height}{\includegraphics[width=0.19\columnwidth, trim=40 40 40 40, clip]{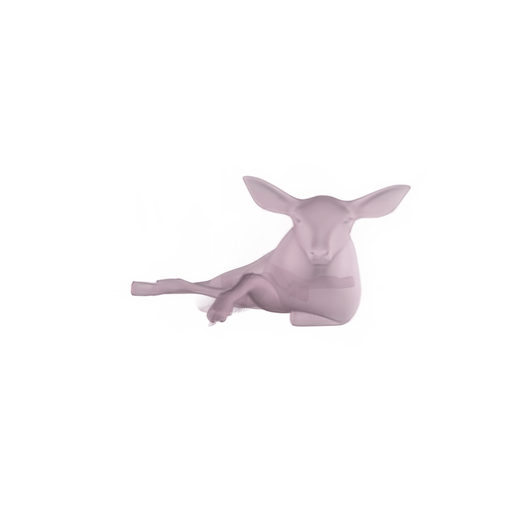}} \\[-0.5pt]
% --- GT Frame 8 ---
\raisebox{-0.5\height}{\rotatebox{90}{\tiny render $\angle 90^\circ$}} &
\raisebox{-0.5\height}{\includegraphics[width=0.19\columnwidth, trim=40 40 40 40, clip]{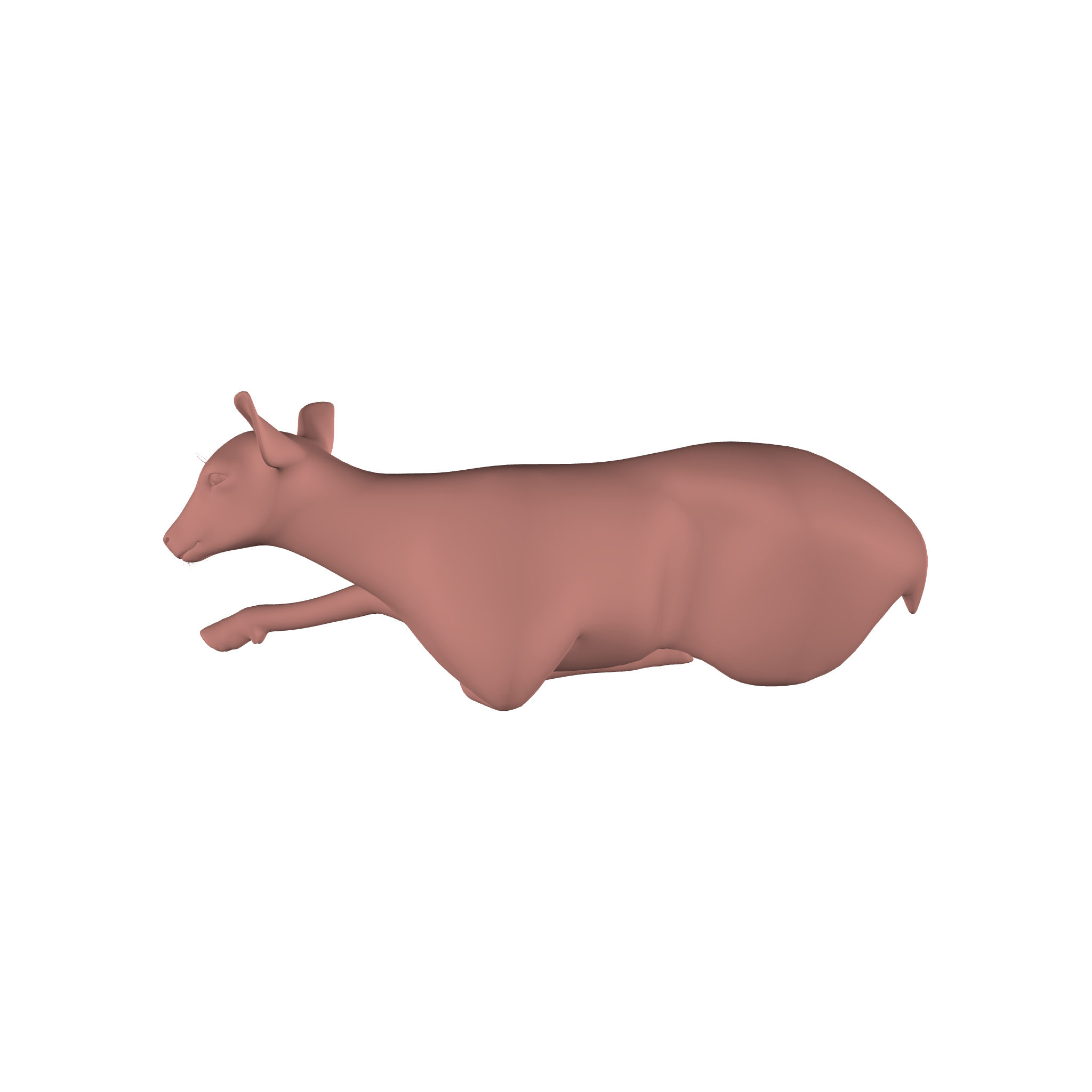}} & 
\raisebox{-0.5\height}{\includegraphics[width=0.19\columnwidth, trim=20 20 20 20, clip]{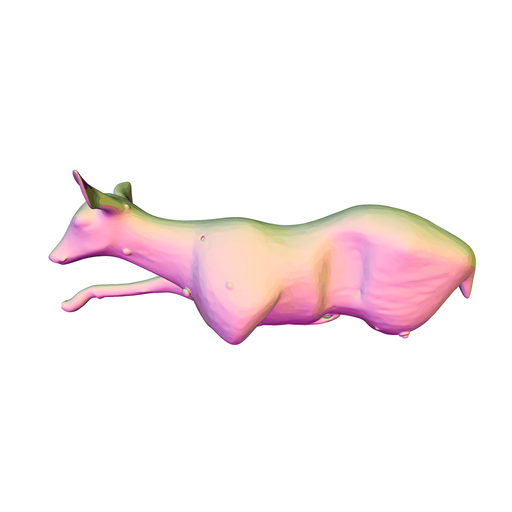}} &
\raisebox{-0.5\height}{\includegraphics[width=0.19\columnwidth, trim=40 40 40 40, clip]{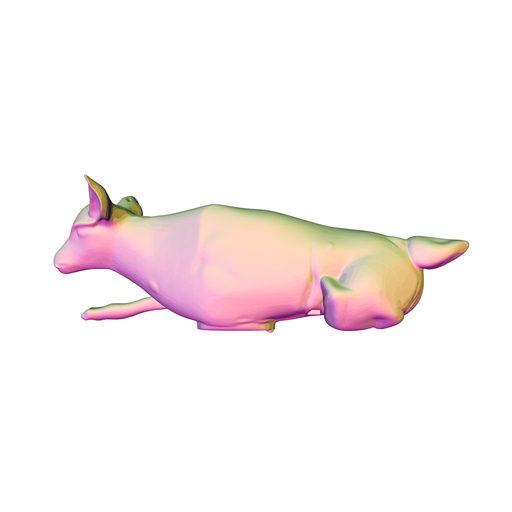}} &
\raisebox{-0.5\height}{\includegraphics[width=0.19\columnwidth, trim=40 40 40 40, clip]{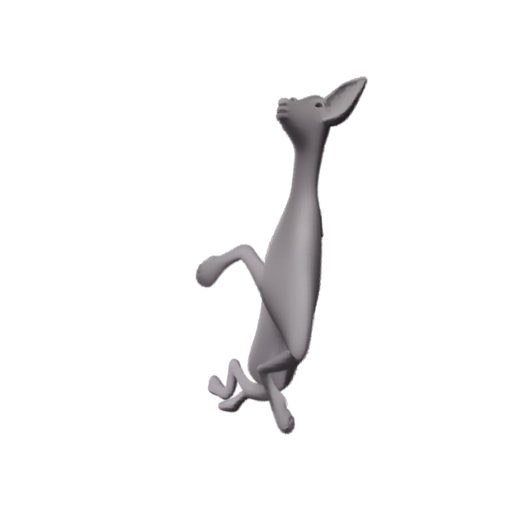}} &
\raisebox{-0.5\height}{\includegraphics[width=0.19\columnwidth, trim=40 40 40 40, clip]{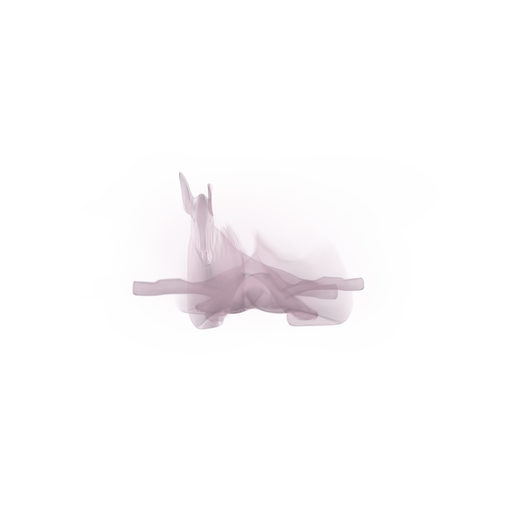}} \\[3pt]

% -------- Frame 15 --------
\raisebox{-0.5\height}{\rotatebox{90}{\tiny Input $\angle 0^\circ$}} &
\raisebox{-0.5\height}{\includegraphics[width=0.19\columnwidth, trim=40 40 40 40, clip]{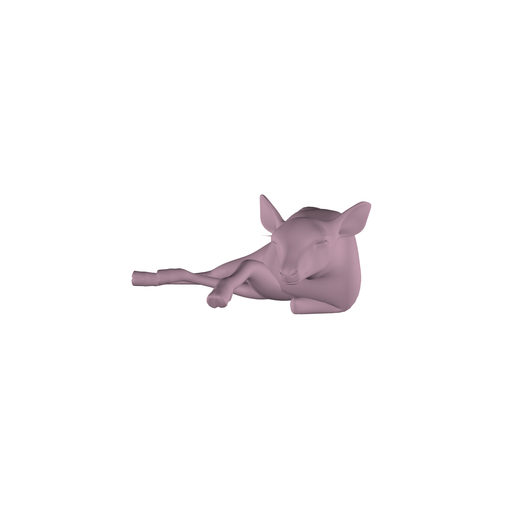}} &
\raisebox{-0.5\height}{\includegraphics[width=0.19\columnwidth, trim=20 20 20 20, clip]{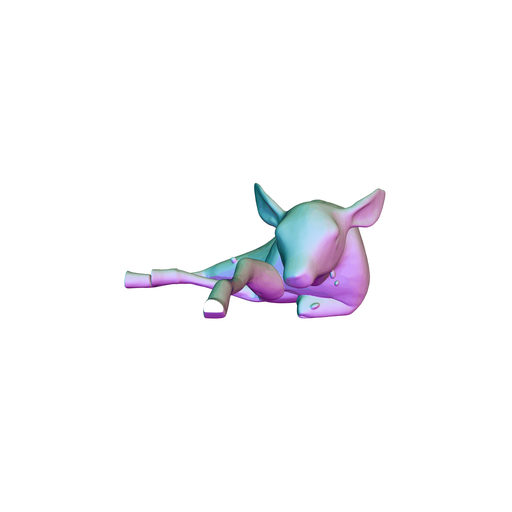}} &
\raisebox{-0.5\height}{\includegraphics[width=0.19\columnwidth, trim=40 40 40 40, clip]{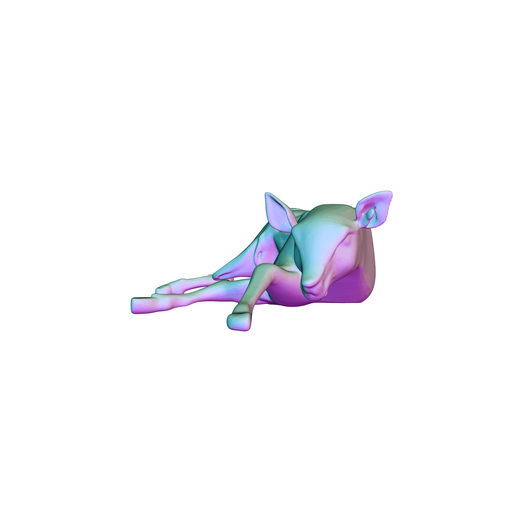}} &
\raisebox{-0.5\height}{\includegraphics[width=0.19\columnwidth, trim=40 40 40 40, clip]{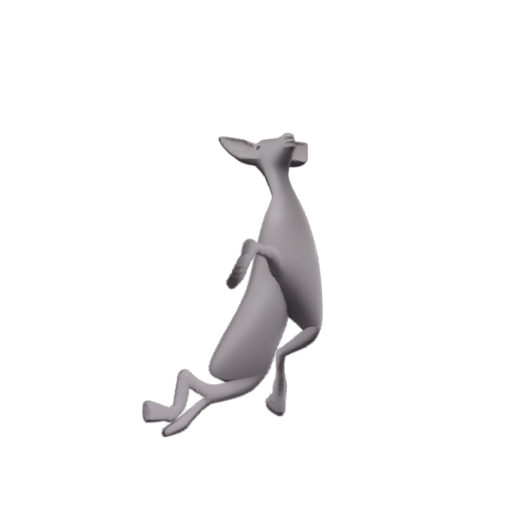}} &
\raisebox{-0.5\height}{\includegraphics[width=0.19\columnwidth, trim=40 40 40 40, clip]{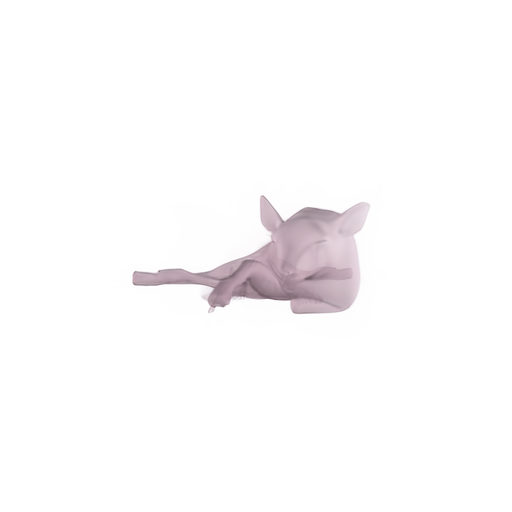}} \\[-0.5pt]
% --- GT Frame 15 ---
\raisebox{-0.5\height}{\rotatebox{90}{\tiny render $\angle 90^\circ$}} &
\raisebox{-0.5\height}{\includegraphics[width=0.19\columnwidth, trim=40 40 40 40, clip]{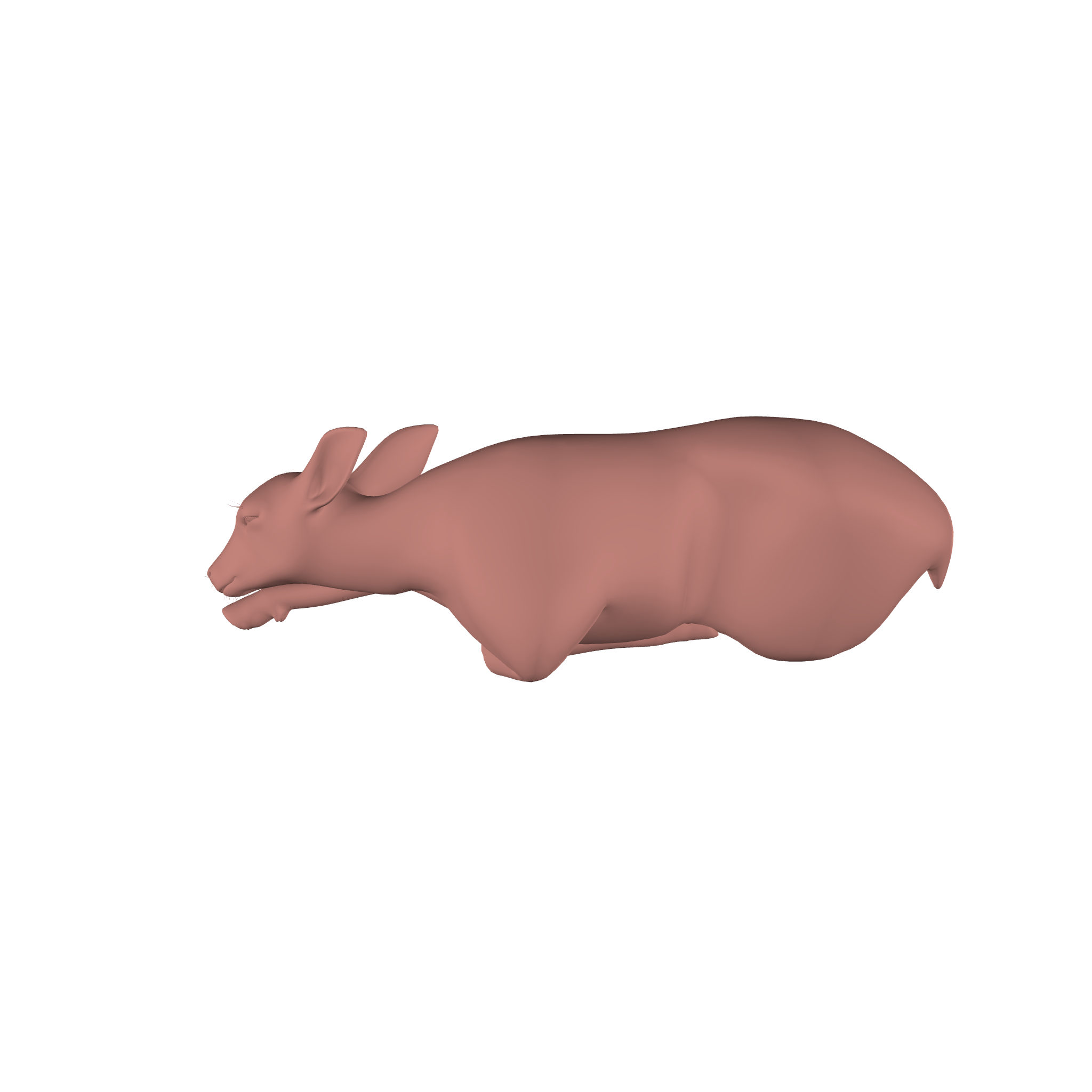}} & 
\raisebox{-0.5\height}{\includegraphics[width=0.19\columnwidth, trim=20 20 20 20, clip]{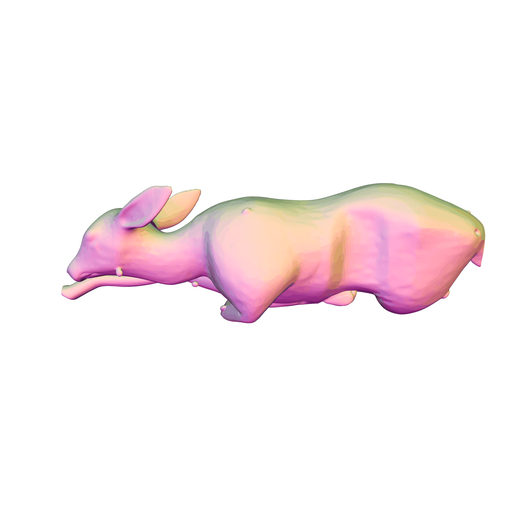}} &
\raisebox{-0.5\height}{\includegraphics[width=0.19\columnwidth, trim=40 40 40 40, clip]{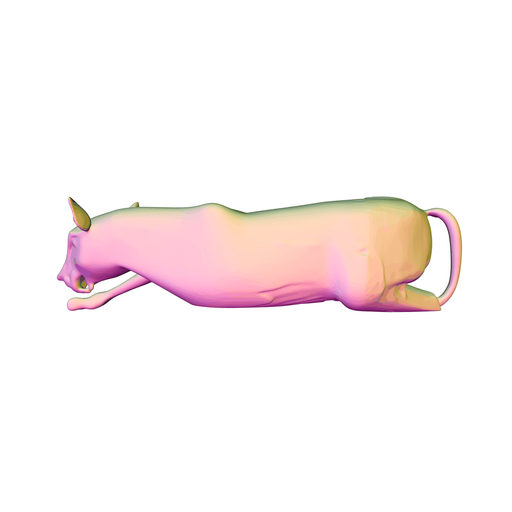}} &
\raisebox{-0.5\height}{\includegraphics[width=0.19\columnwidth, trim=40 40 40 40, clip]{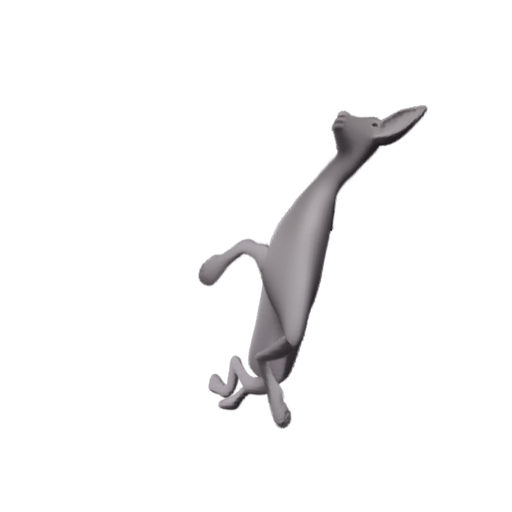}} &
\raisebox{-0.5\height}{\includegraphics[width=0.19\columnwidth, trim=40 40 40 40, clip]{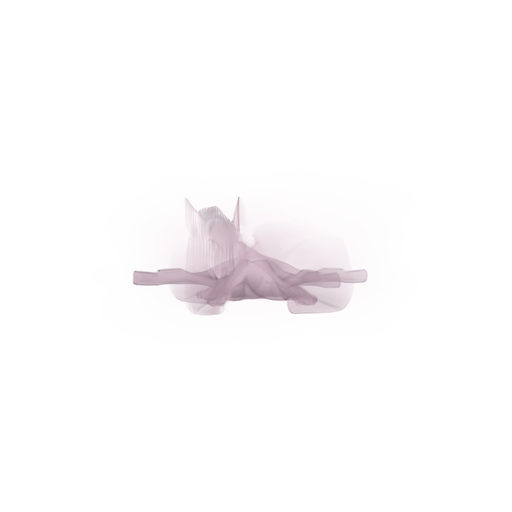}} \\[5pt]

% ==================== SUBJECT: ironman_BreakdanceFreezeVar4 ====================
% -------- Frame 0 --------
\raisebox{-0.5\height}{\rotatebox{90}{\tiny Input $\angle 0^\circ$}} &
\raisebox{-0.5\height}{\includegraphics[width=0.19\columnwidth, trim=40 40 40 40, clip]{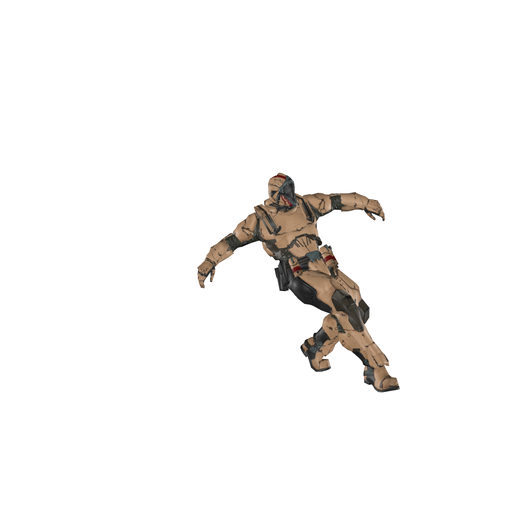}} &
\raisebox{-0.5\height}{\includegraphics[width=0.19\columnwidth, trim=20 20 20 20, clip]{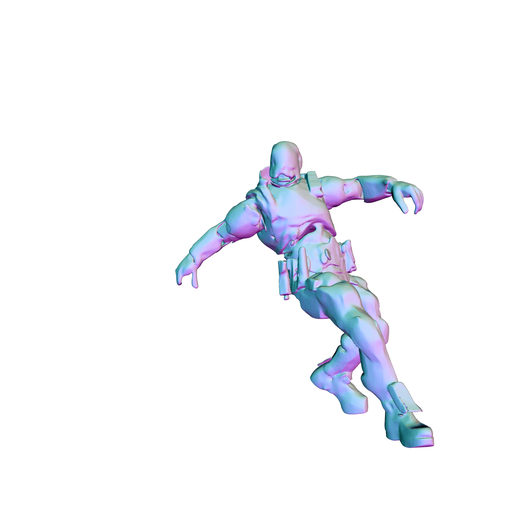}} &
\raisebox{-0.5\height}{\includegraphics[width=0.19\columnwidth, trim=40 40 40 40, clip]{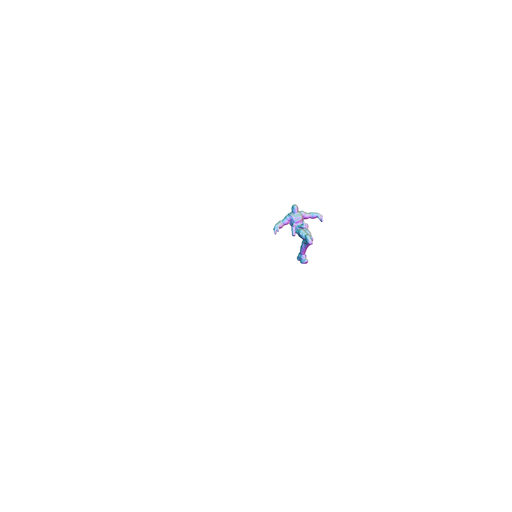}} &
\raisebox{-0.5\height}{\includegraphics[width=0.19\columnwidth, trim=40 40 40 40, clip]{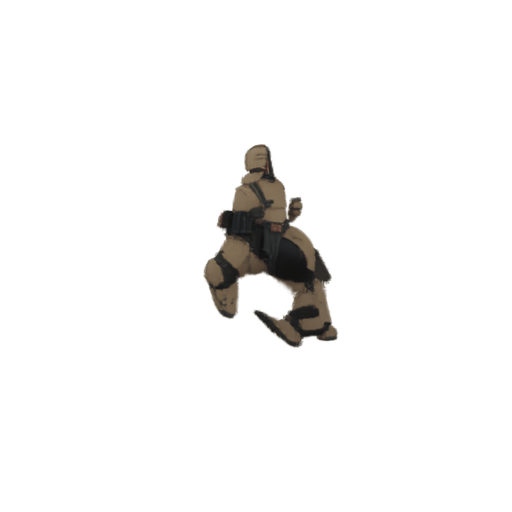}} &
\raisebox{-0.5\height}{\includegraphics[width=0.19\columnwidth, trim=40 40 40 40, clip]{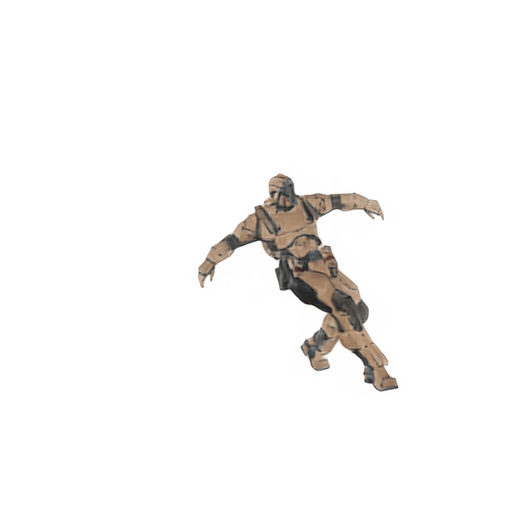}} \\[-0.5pt]
% --- GT Frame 0 ---
\raisebox{-0.5\height}{\rotatebox{90}{\tiny render $\angle 90^\circ$}} &
\raisebox{-0.5\height}{\includegraphics[width=0.19\columnwidth, trim=40 40 40 40, clip]{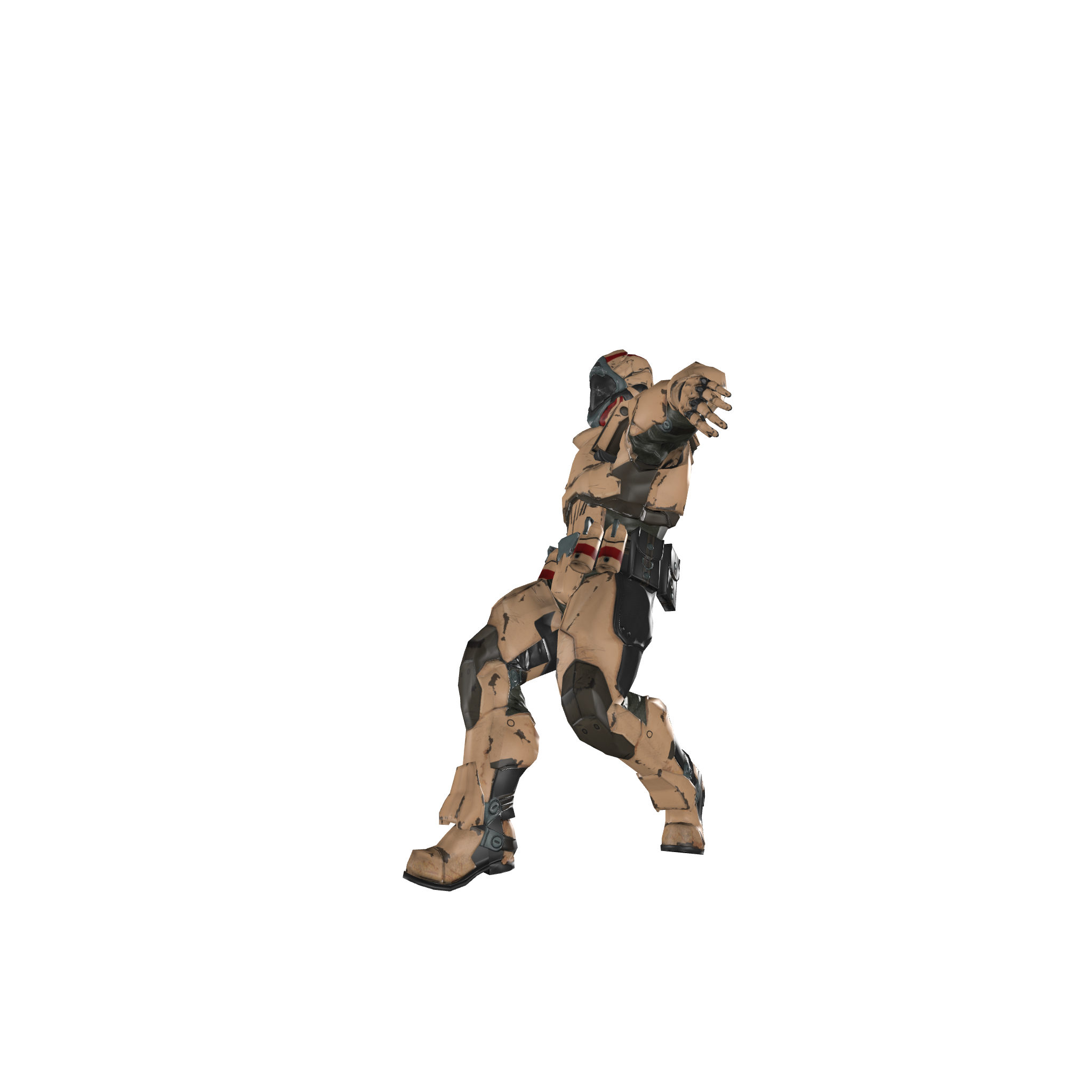}} & 
\raisebox{-0.5\height}{\includegraphics[width=0.19\columnwidth, trim=20 20 20 20, clip]{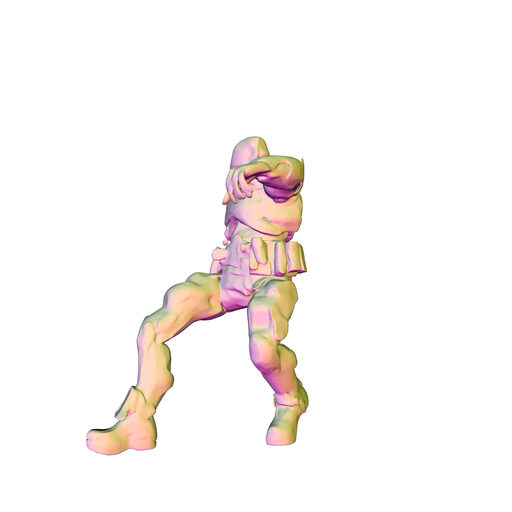}} &
\raisebox{-0.5\height}{\includegraphics[width=0.19\columnwidth, trim=40 40 40 40, clip]{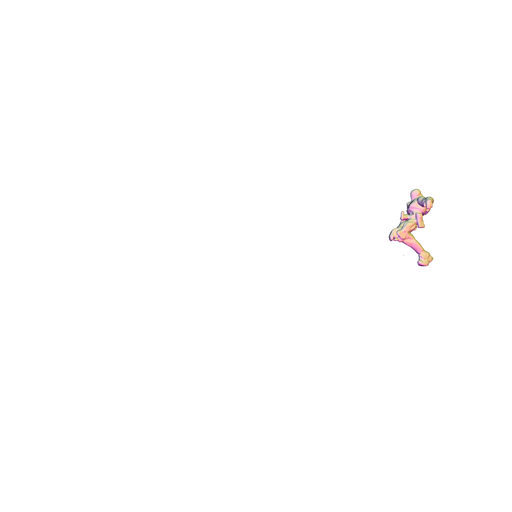}} &
\raisebox{-0.5\height}{\includegraphics[width=0.19\columnwidth, trim=40 40 40 40, clip]{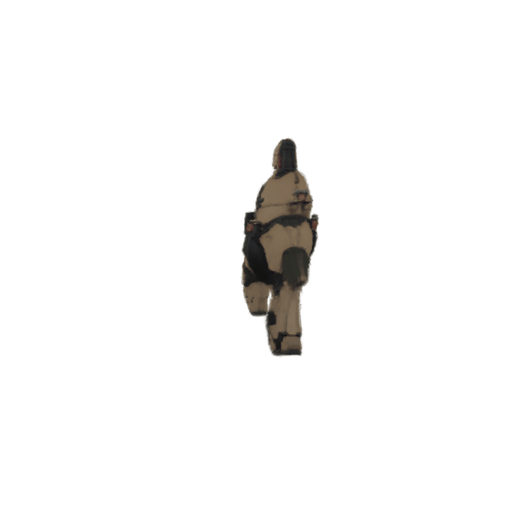}} &
\raisebox{-0.5\height}{\includegraphics[width=0.19\columnwidth, trim=40 40 40 40, clip]{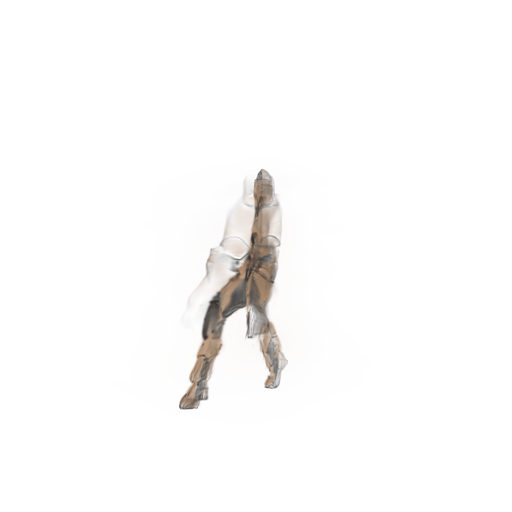}} \\[3pt]

% -------- Frame 8 --------
\raisebox{-0.5\height}{\rotatebox{90}{\tiny Input $\angle 0^\circ$}} &
\raisebox{-0.5\height}{\includegraphics[width=0.19\columnwidth, trim=40 40 40 40, clip]{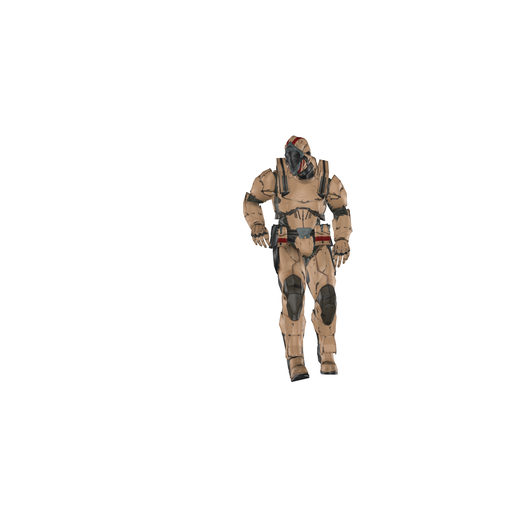}} &
\raisebox{-0.5\height}{\includegraphics[width=0.19\columnwidth, trim=20 20 20 20, clip]{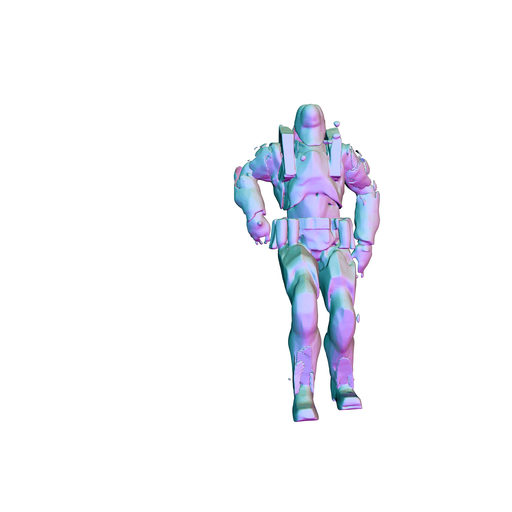}} &
\raisebox{-0.5\height}{\includegraphics[width=0.19\columnwidth, trim=40 40 40 40, clip]{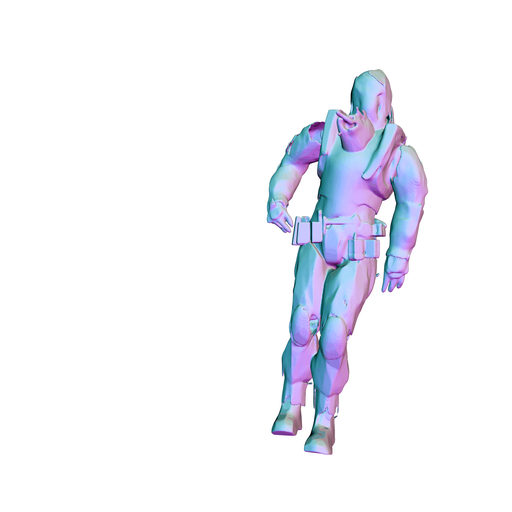}} &
\raisebox{-0.5\height}{\includegraphics[width=0.19\columnwidth, trim=40 40 40 40, clip]{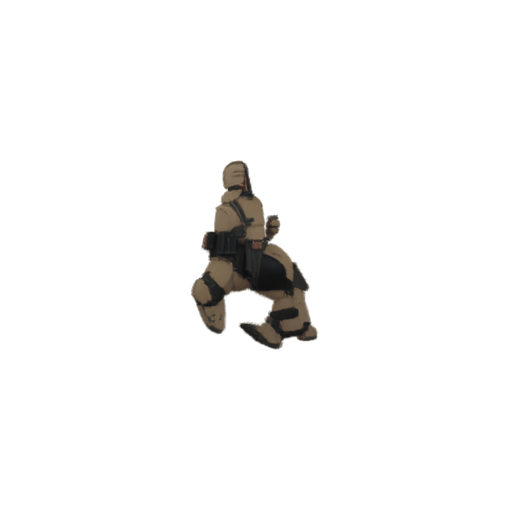}} &
\raisebox{-0.5\height}{\includegraphics[width=0.19\columnwidth, trim=40 40 40 40, clip]{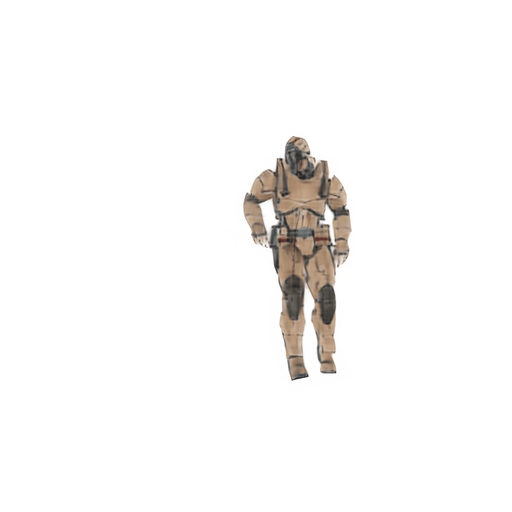}} \\[-0.5pt]
% --- GT Frame 8 ---
\raisebox{-0.5\height}{\rotatebox{90}{\tiny render $\angle 90^\circ$}} &
\raisebox{-0.5\height}{\includegraphics[width=0.19\columnwidth, trim=40 40 40 40, clip]{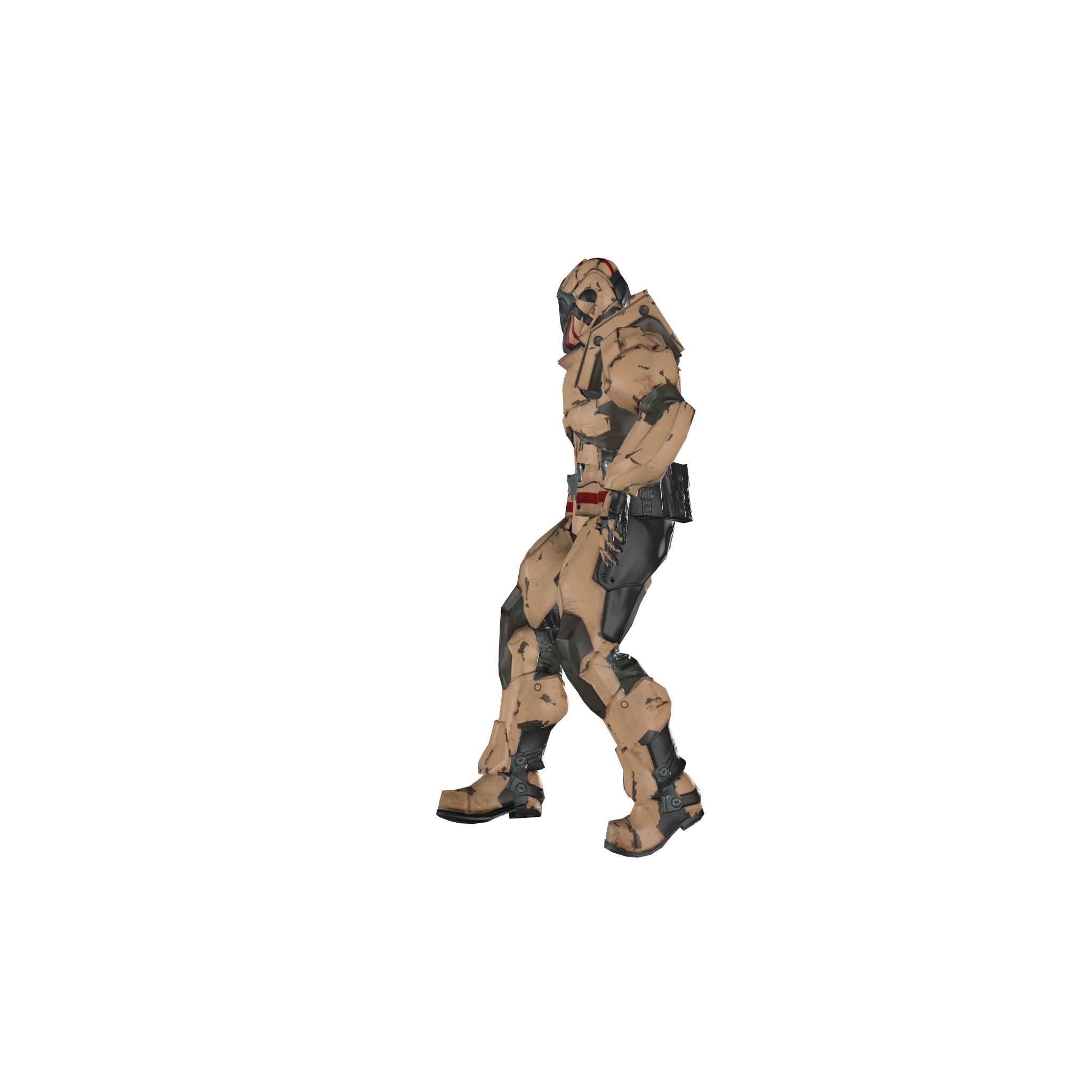}} & 
\raisebox{-0.5\height}{\includegraphics[width=0.19\columnwidth, trim=20 20 20 20, clip]{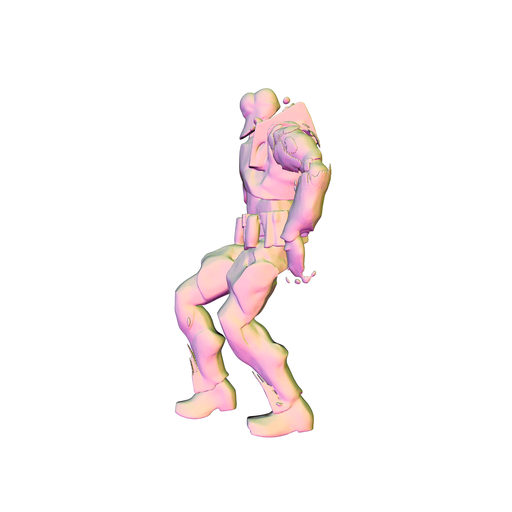}} &
\raisebox{-0.5\height}{\includegraphics[width=0.19\columnwidth, trim=40 40 40 40, clip]{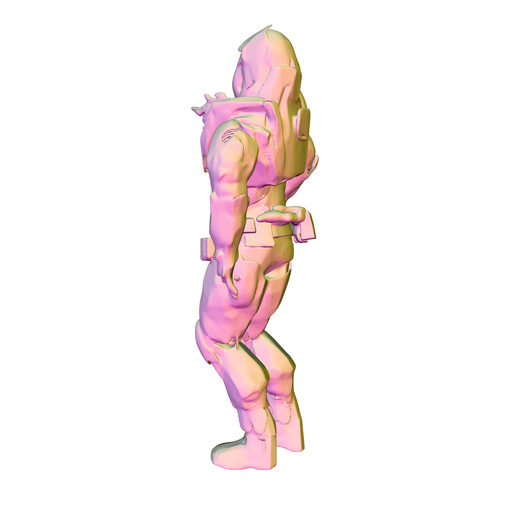}} &
\raisebox{-0.5\height}{\includegraphics[width=0.19\columnwidth, trim=40 40 40 40, clip]{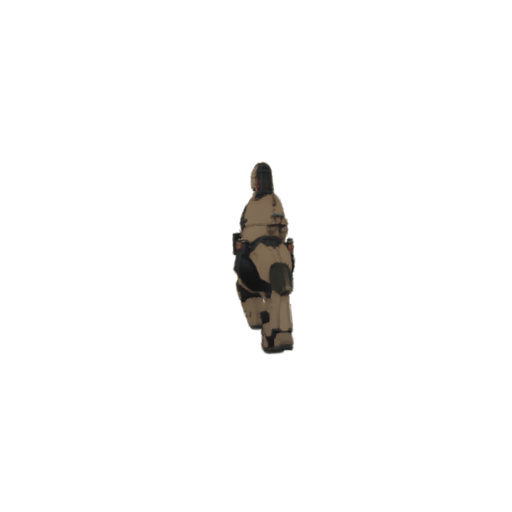}} &
\raisebox{-0.5\height}{\includegraphics[width=0.19\columnwidth, trim=40 40 40 40, clip]{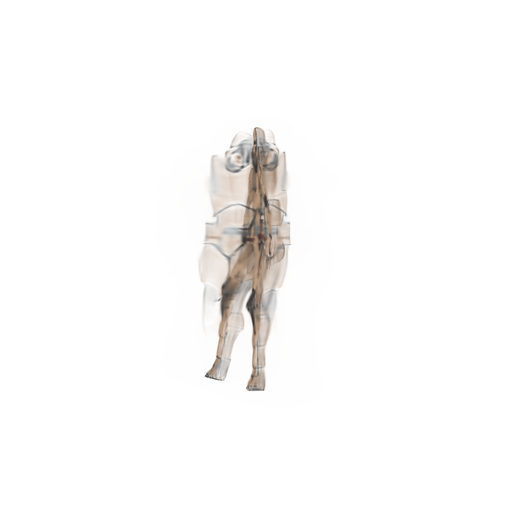}} \\[3pt]

% -------- Frame 15 --------
\raisebox{-0.5\height}{\rotatebox{90}{\tiny Input $\angle 0^\circ$}} &
\raisebox{-0.5\height}{\includegraphics[width=0.19\columnwidth, trim=40 40 40 40, clip]{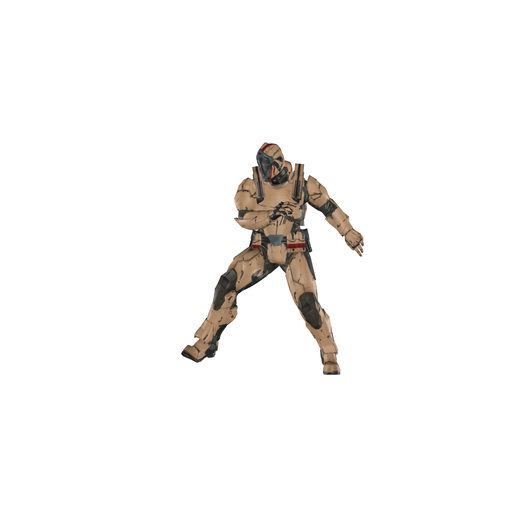}} &
\raisebox{-0.5\height}{\includegraphics[width=0.19\columnwidth, trim=20 20 20 20, clip]{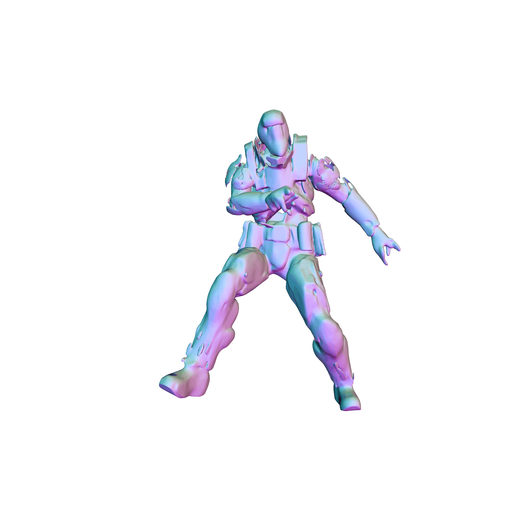}} &
\raisebox{-0.5\height}{\includegraphics[width=0.19\columnwidth, trim=40 40 40 40, clip]{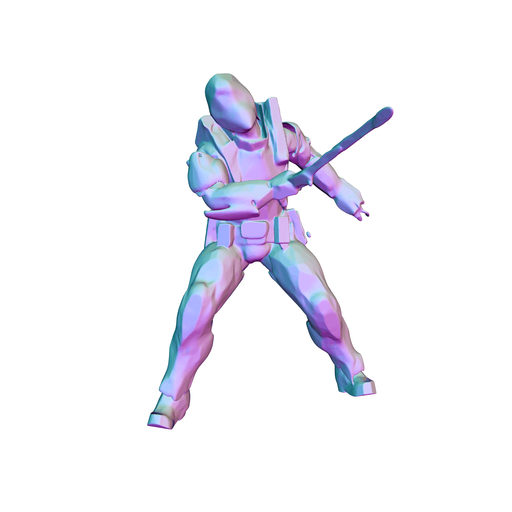}} &
\raisebox{-0.5\height}{\includegraphics[width=0.19\columnwidth, trim=40 40 40 40, clip]{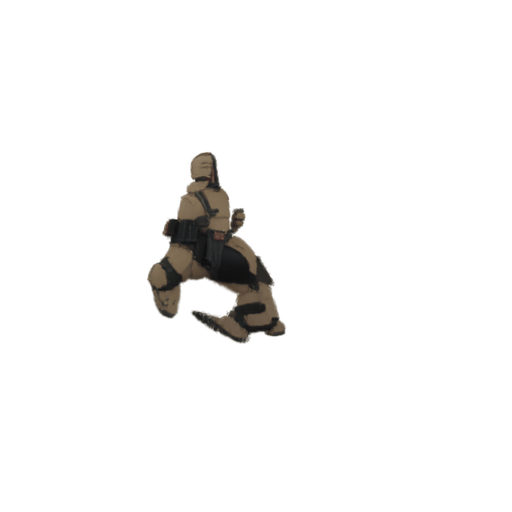}} &
\raisebox{-0.5\height}{\includegraphics[width=0.19\columnwidth, trim=40 40 40 40, clip]{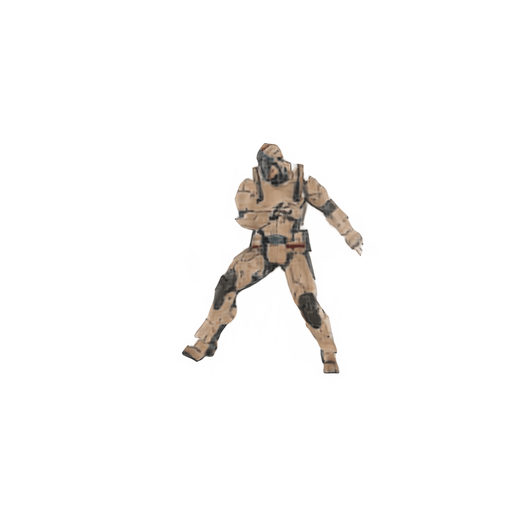}} \\[-0.5pt]
% --- GT Frame 15 ---
\raisebox{-0.5\height}{\rotatebox{90}{\tiny render $\angle 90^\circ$}} &
\raisebox{-0.5\height}{\includegraphics[width=0.19\columnwidth, trim=40 40 40 40, clip]{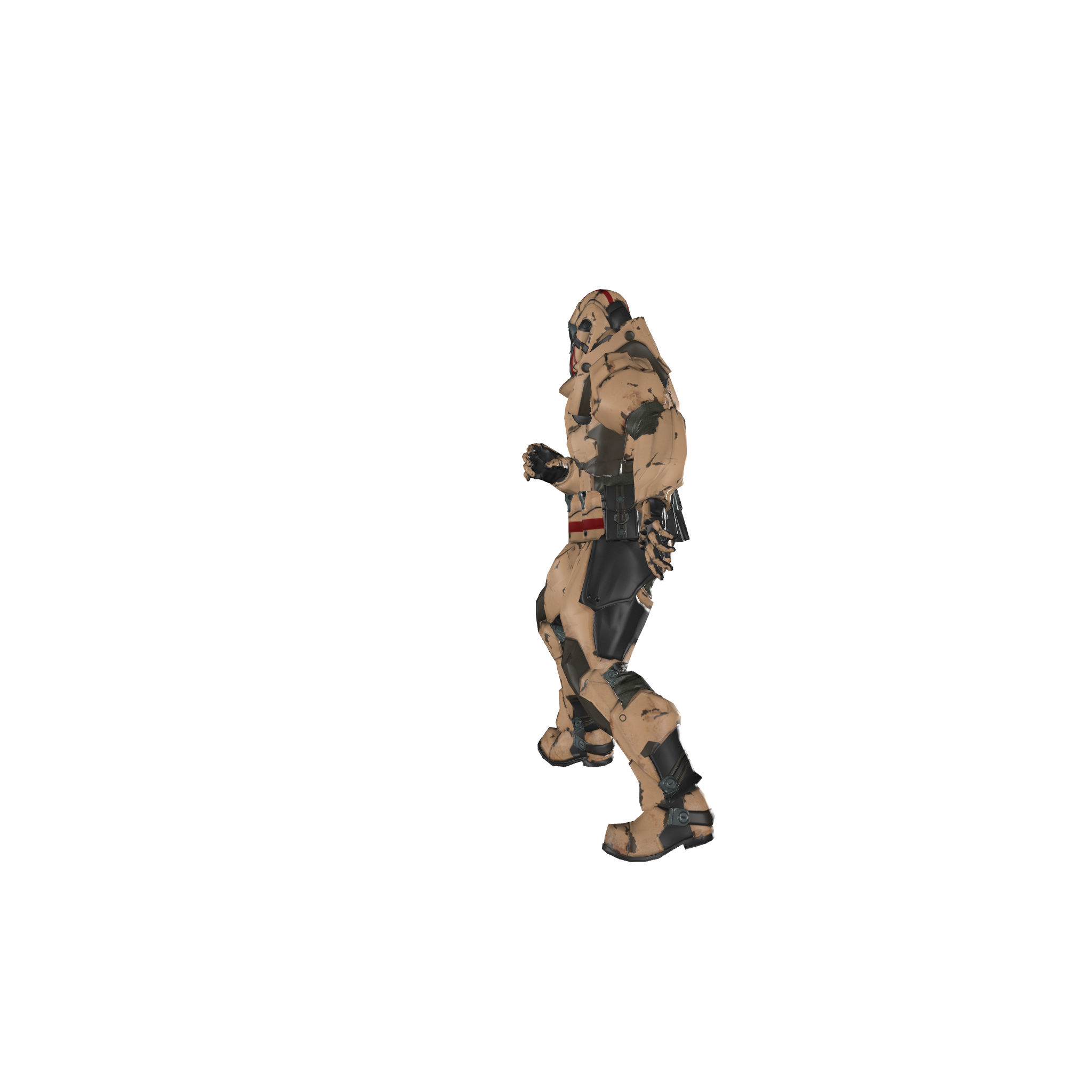}} & 
\raisebox{-0.5\height}{\includegraphics[width=0.19\columnwidth, trim=20 20 20 20, clip]{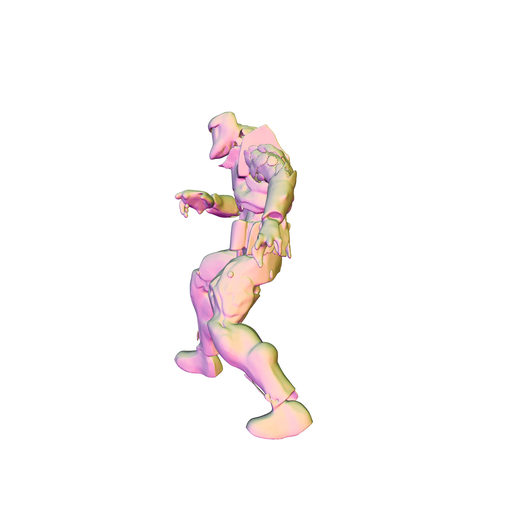}} &
\raisebox{-0.5\height}{\includegraphics[width=0.19\columnwidth, trim=40 40 40 40, clip]{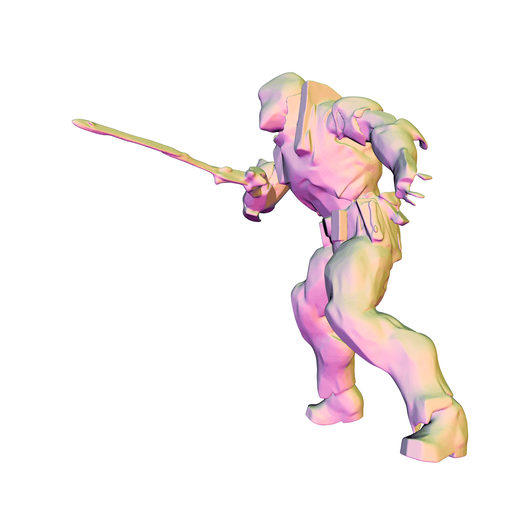}} &
\raisebox{-0.5\height}{\includegraphics[width=0.19\columnwidth, trim=40 40 40 40, clip]{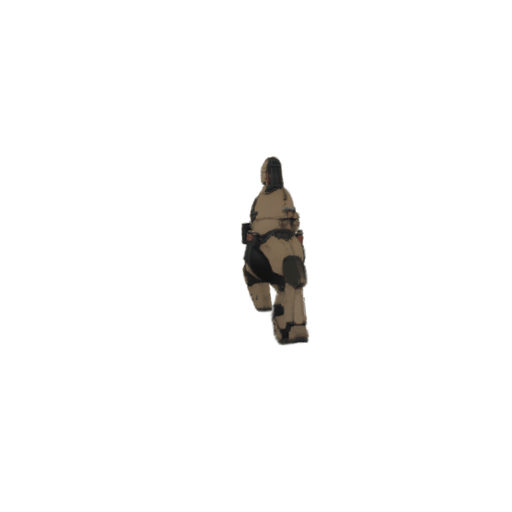}} &
\raisebox{-0.5\height}{\includegraphics[width=0.19\columnwidth, trim=40 40 40 40, clip]{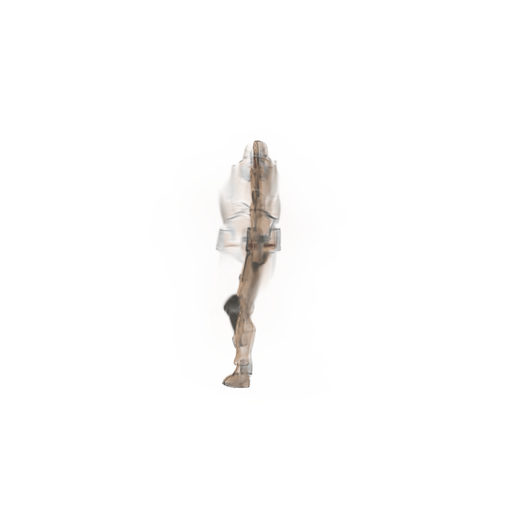}} \\

  \end{tabular}
  % ==================== TABLE END ====================
  
  \vspace{-3pt}
  \caption{
  Qualitative 4D generation comparisons from DeformingThings~\cite{li20214dcomplete} on two subjects at three time steps. For each model, we show the reconstructed input view (top) and a rendered novel view (bottom). We display the ground truth novel view in the bottom left. Compared to \cref{fig:qual_comp_4d_appendix_5_6_full}, the sequences contain stronger motion thus showing even larger difference in performances of our method. In particular, V2M4 fails completely due to the large motion.
  }
  \label{fig:qual_comp_4d_appendix_other}
\end{figure}

\begin{figure}[h]
    \section{Additional Qualitative Results on 3D Scene Generation}
  \centering
  \setlength{\tabcolsep}{0.2pt} % horizontal space
  \renewcommand{\arraystretch}{0} % vertical compression
  \begin{tabular}{@{}cccc@{}}
    Input &
    \textbf{Ours} &
    PartCrafter~\cite{lin2025partcrafterstructured3dmesh} &
    MIDI~\cite{huang2025midimultiinstancediffusionsingle} \\
    
    % This is the original row you provided
    \raisebox{-0.5\height}{\includegraphics[width=0.23\linewidth]{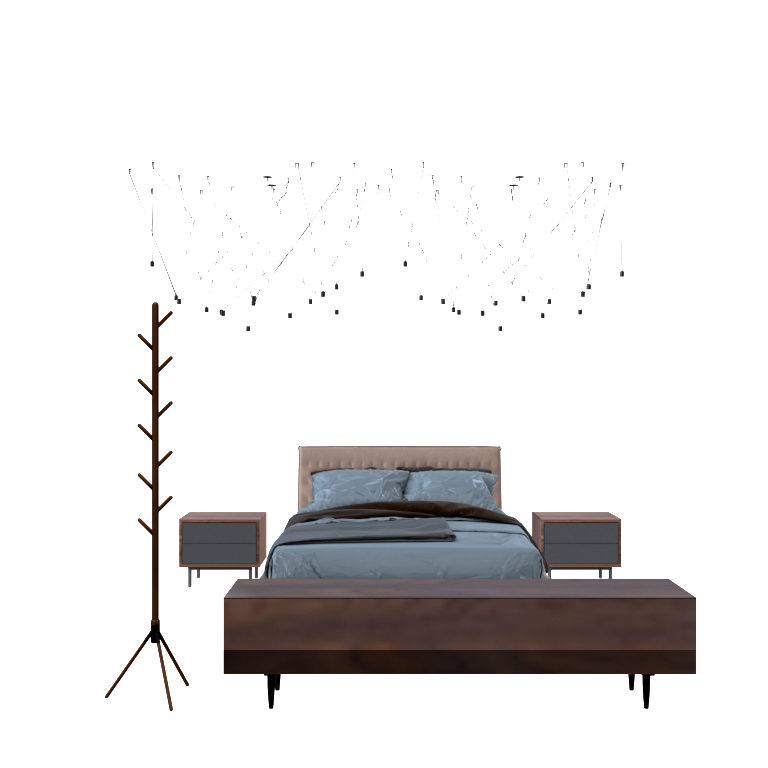}} &
    \raisebox{-0.5\height}{\includegraphics[width=0.23\linewidth]{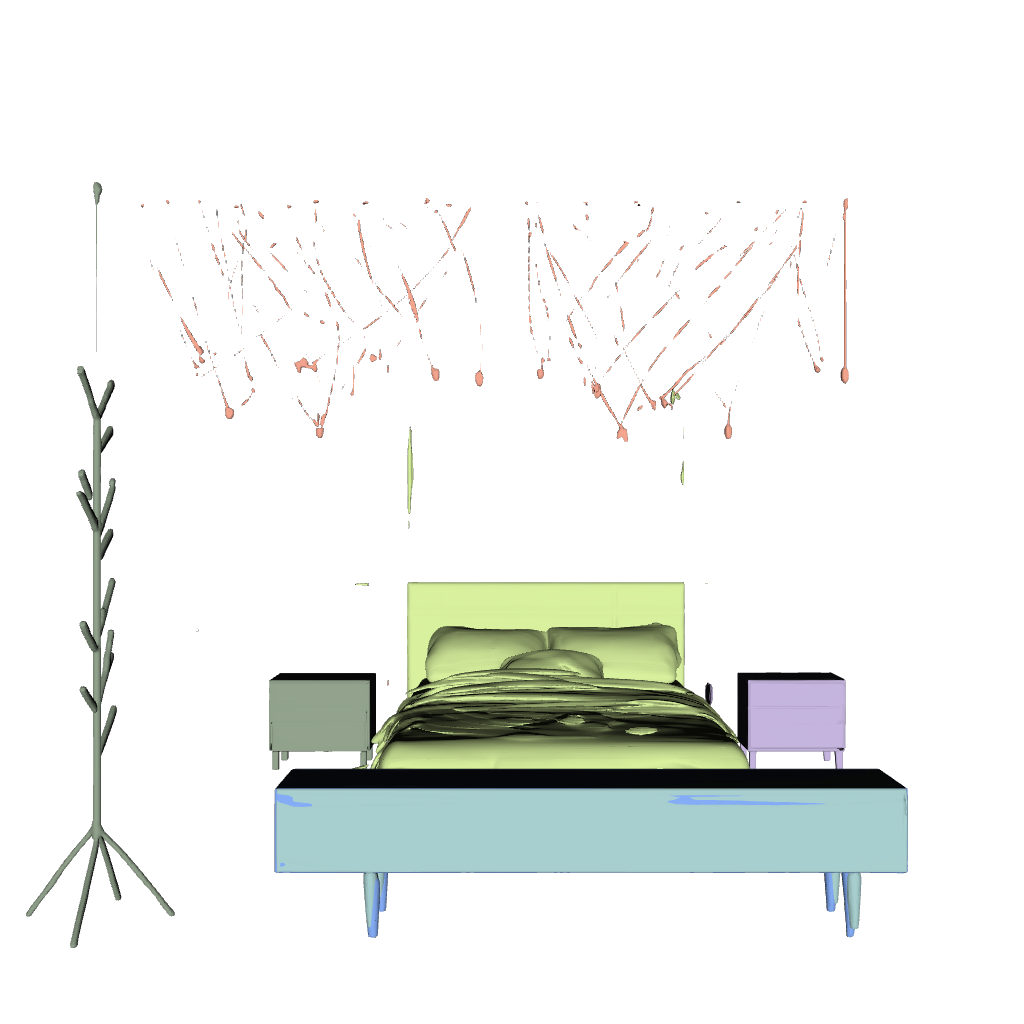}} &
    \raisebox{-0.5\height}{\includegraphics[width=0.23\linewidth]{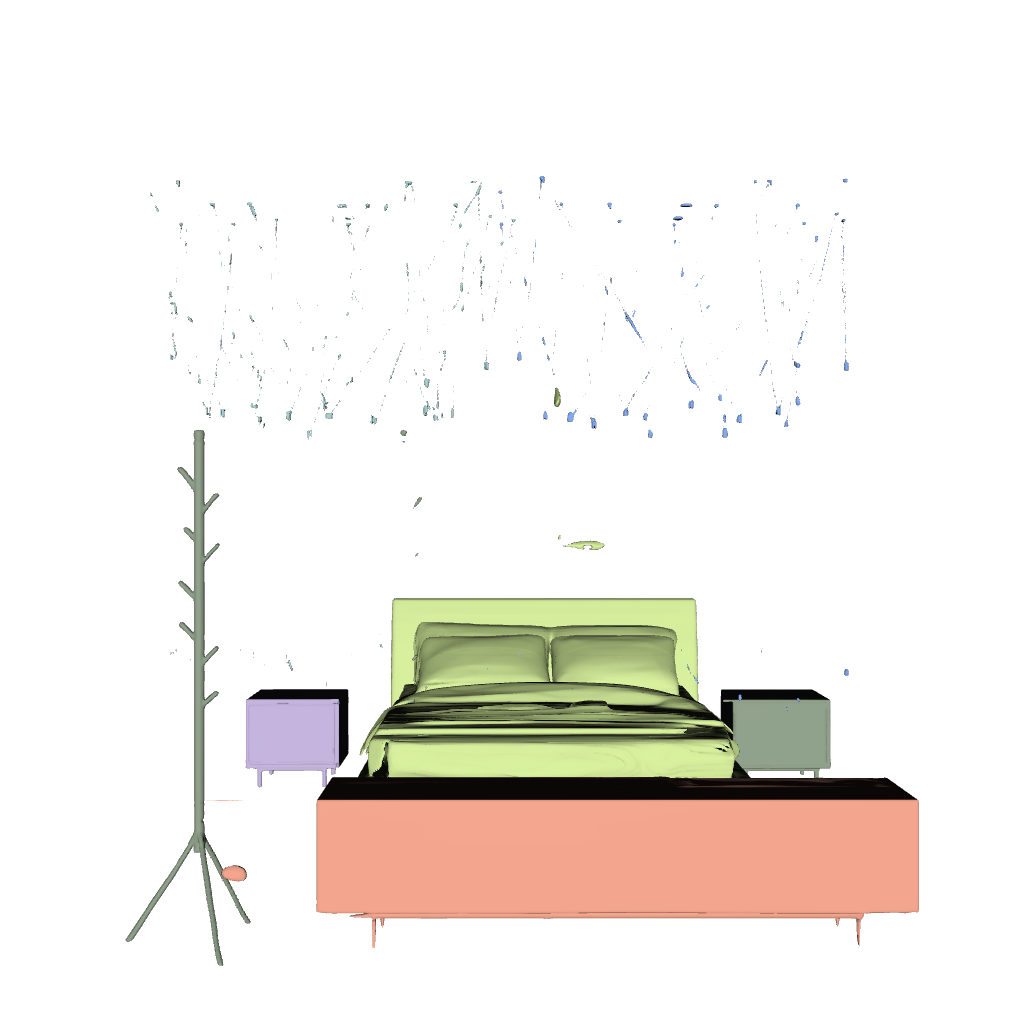}} &
    \raisebox{-0.5\height}{\includegraphics[width=0.23\linewidth]{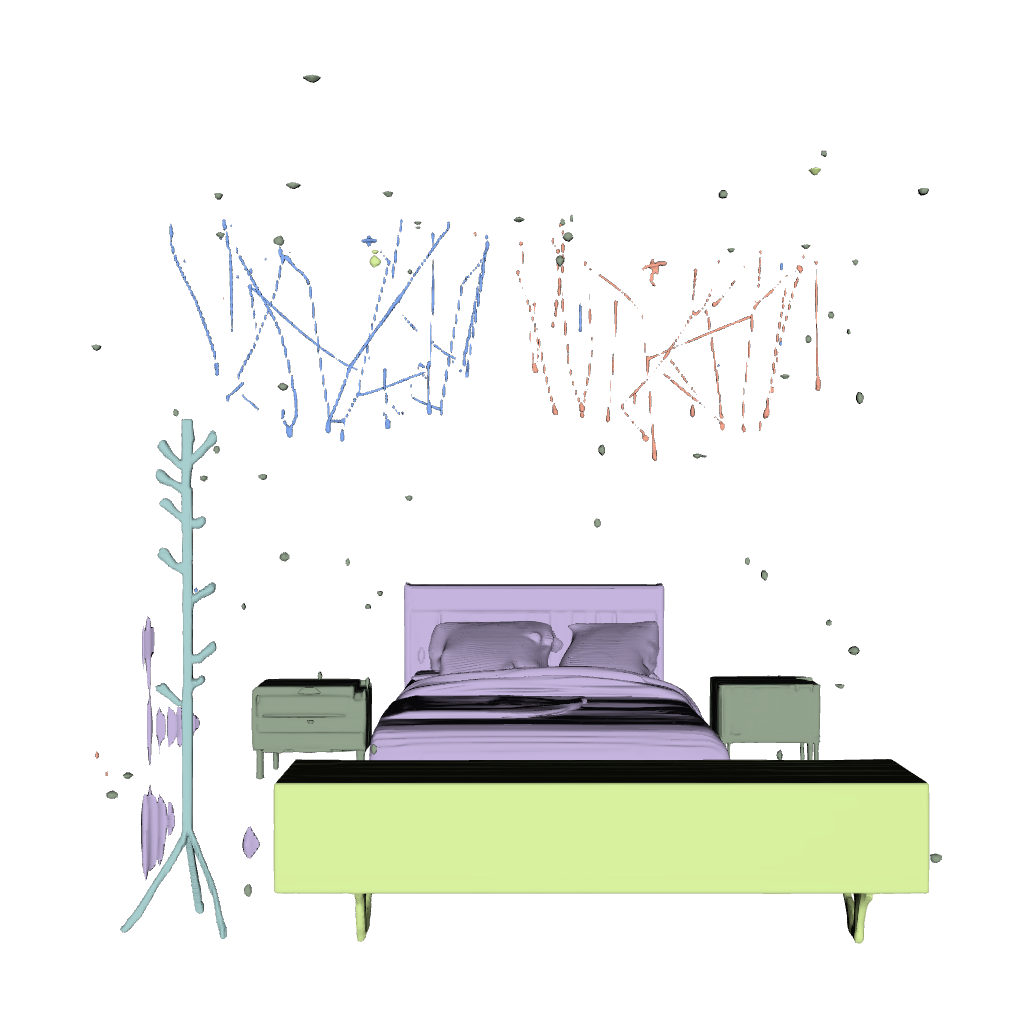}} \\
    
    % New rows start here
    % Row 1
    \raisebox{-0.5\height}{\includegraphics[width=0.23\linewidth]{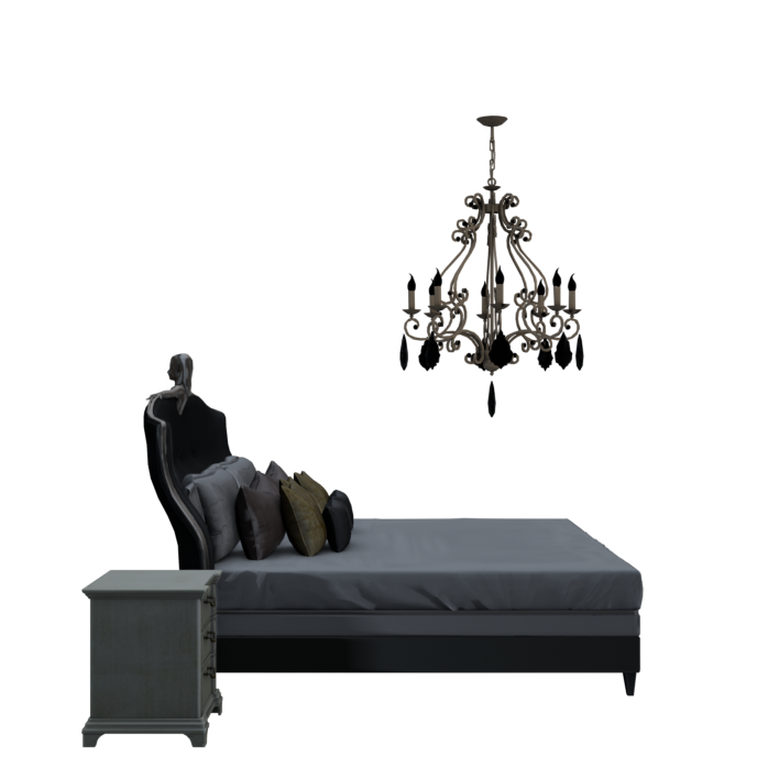}} &
    \raisebox{-0.5\height}{\includegraphics[width=0.23\linewidth]{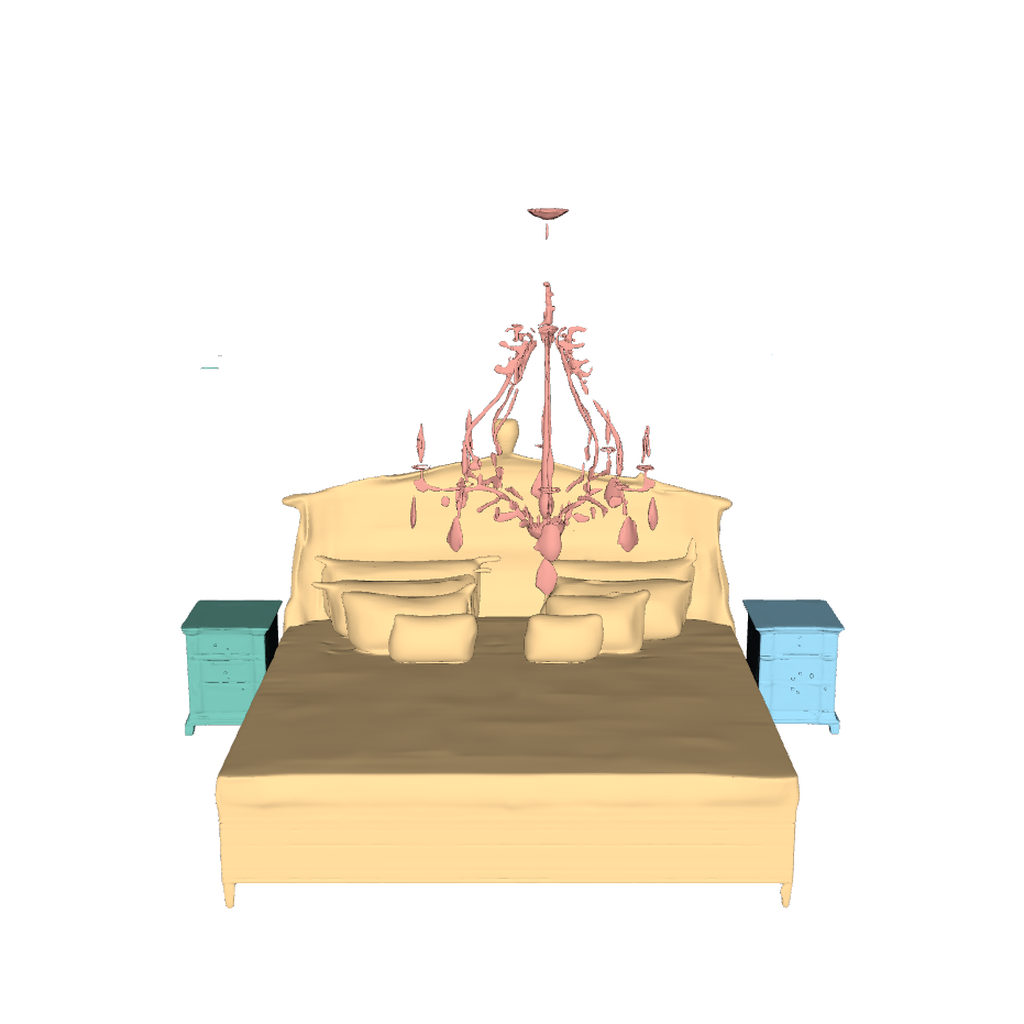}} &
    \raisebox{-0.5\height}{\includegraphics[width=0.23\linewidth]{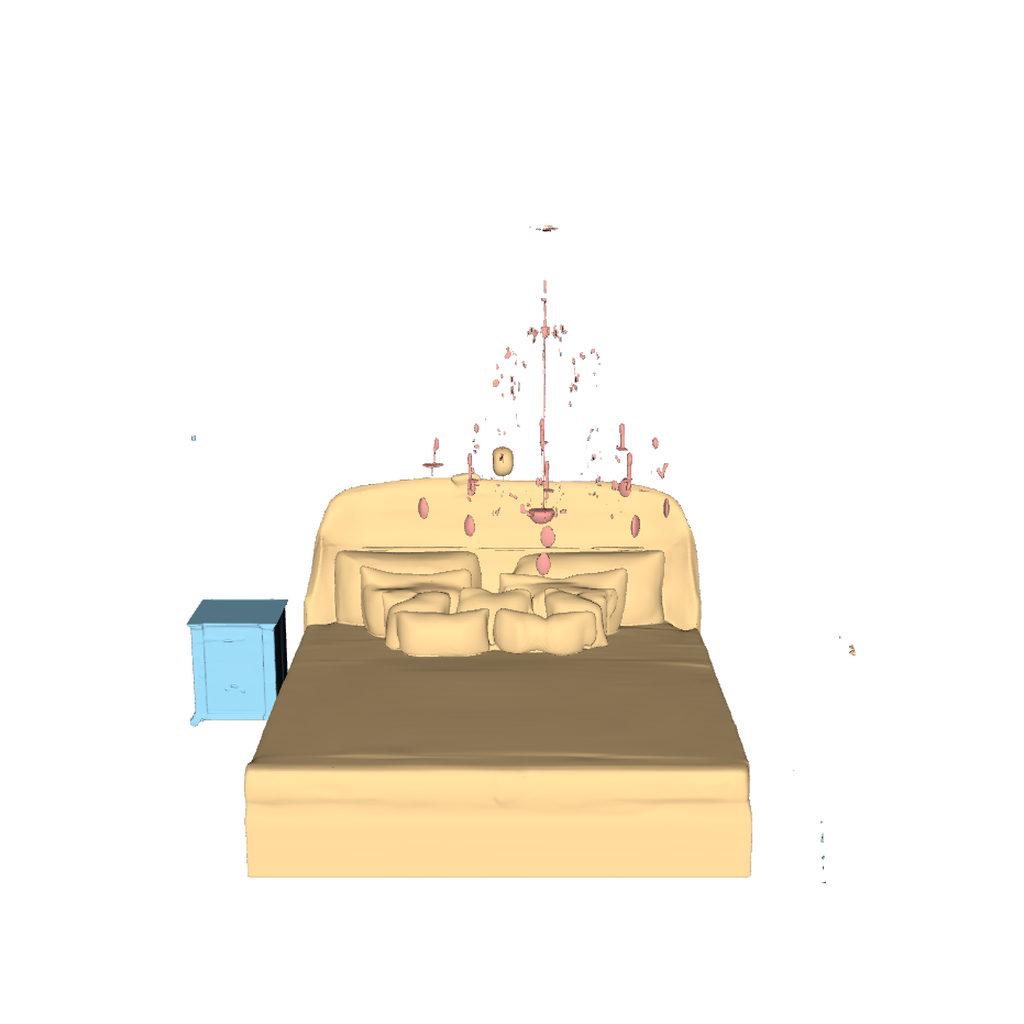}} &
    \raisebox{-0.5\height}{\includegraphics[width=0.23\linewidth]{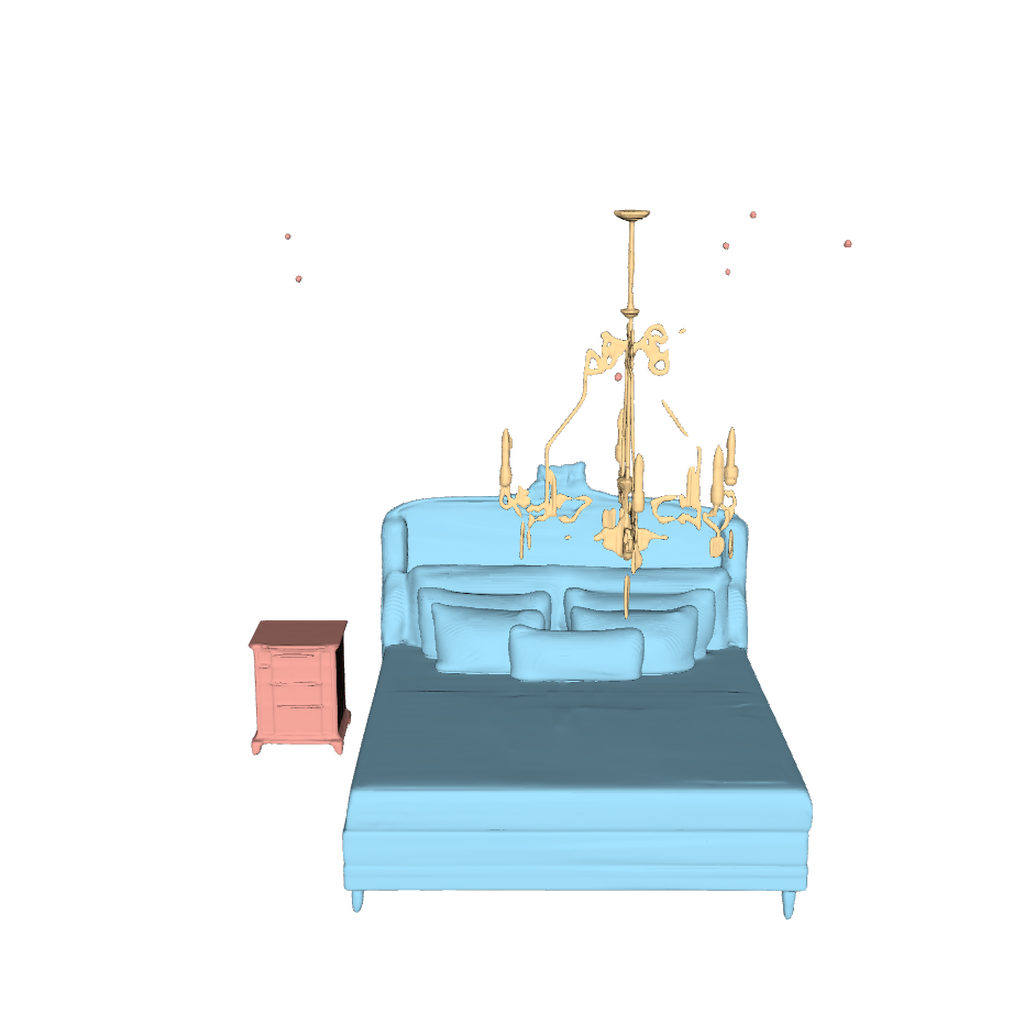}} \\
    
    % Row 2
    \raisebox{-0.5\height}{\includegraphics[width=0.23\linewidth]{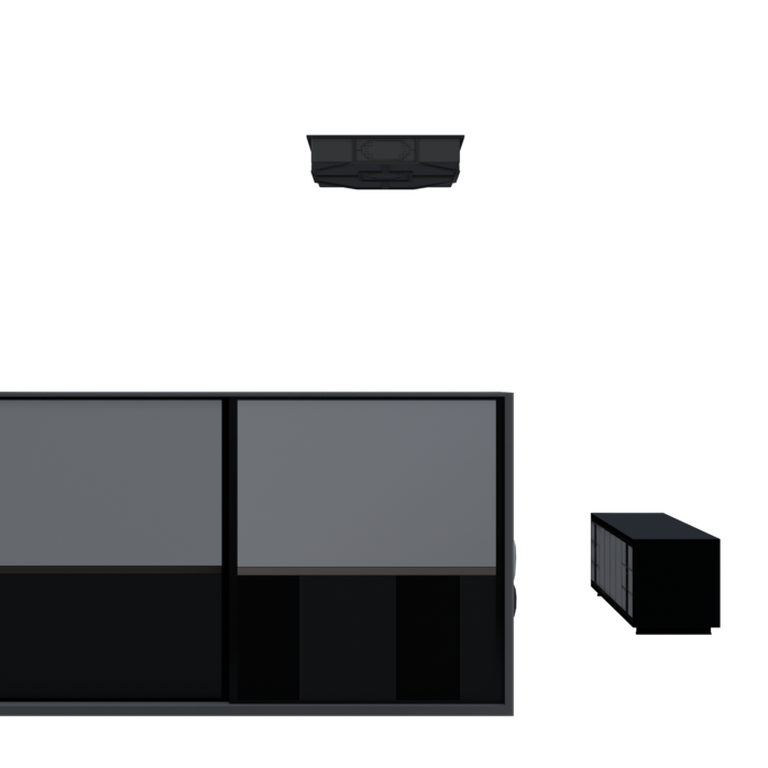}} &
    \raisebox{-0.5\height}{\includegraphics[width=0.23\linewidth]{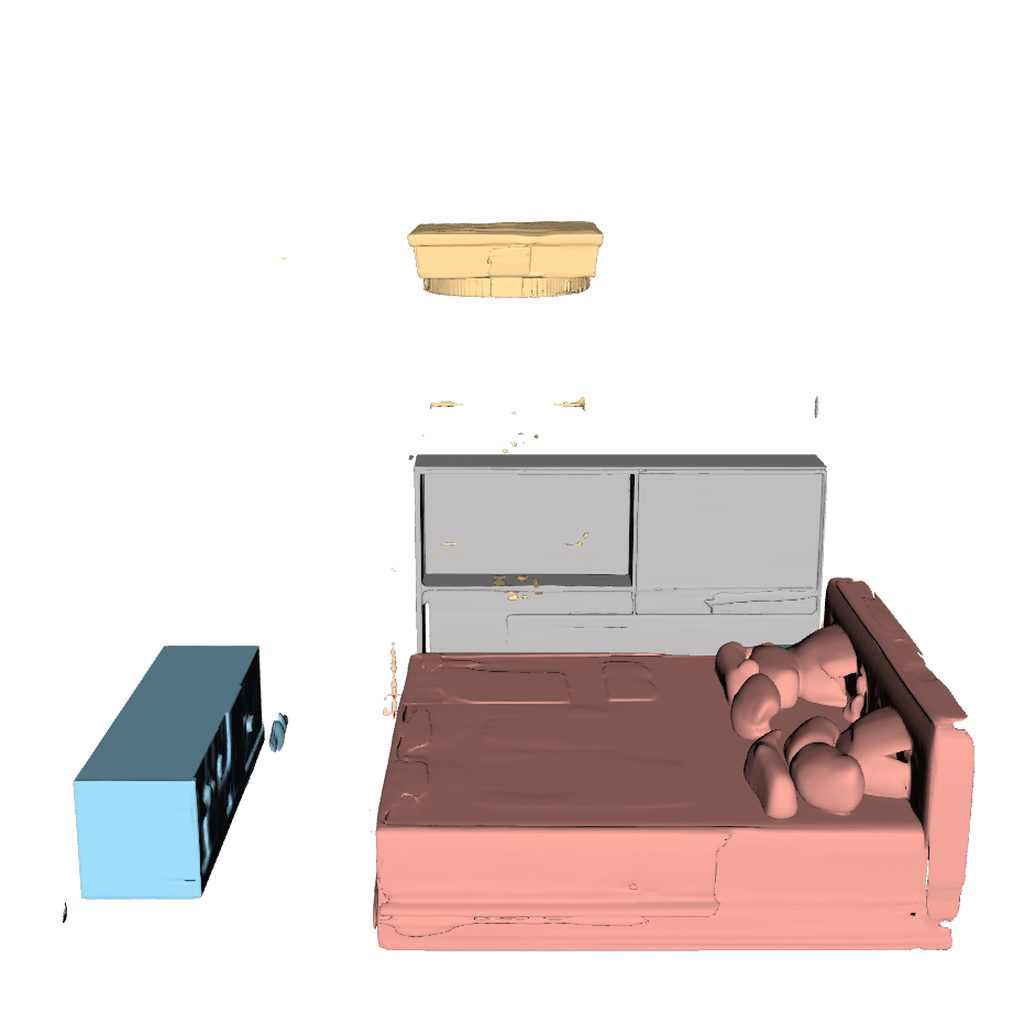}} &
    \raisebox{-0.5\height}{\includegraphics[width=0.23\linewidth]{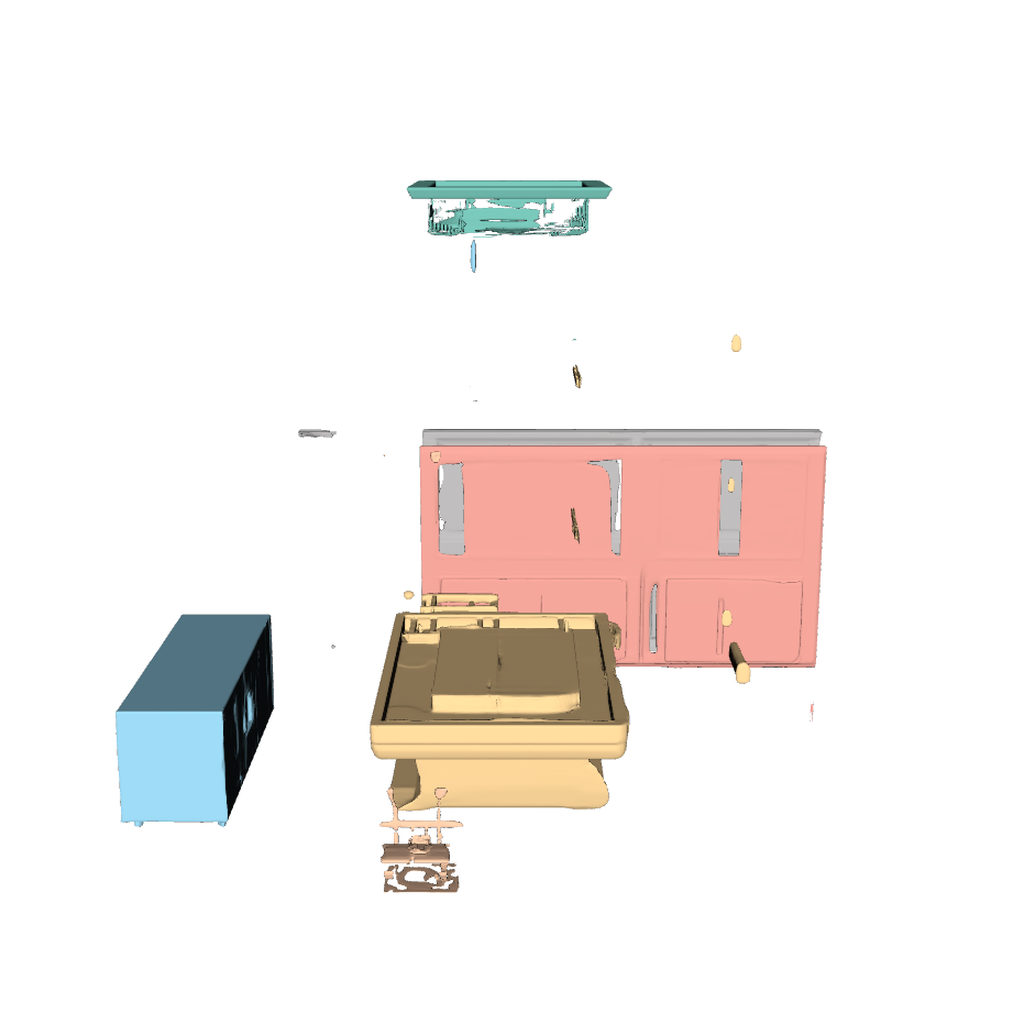}} &
    \raisebox{-0.5\height}{\includegraphics[width=0.23\linewidth]{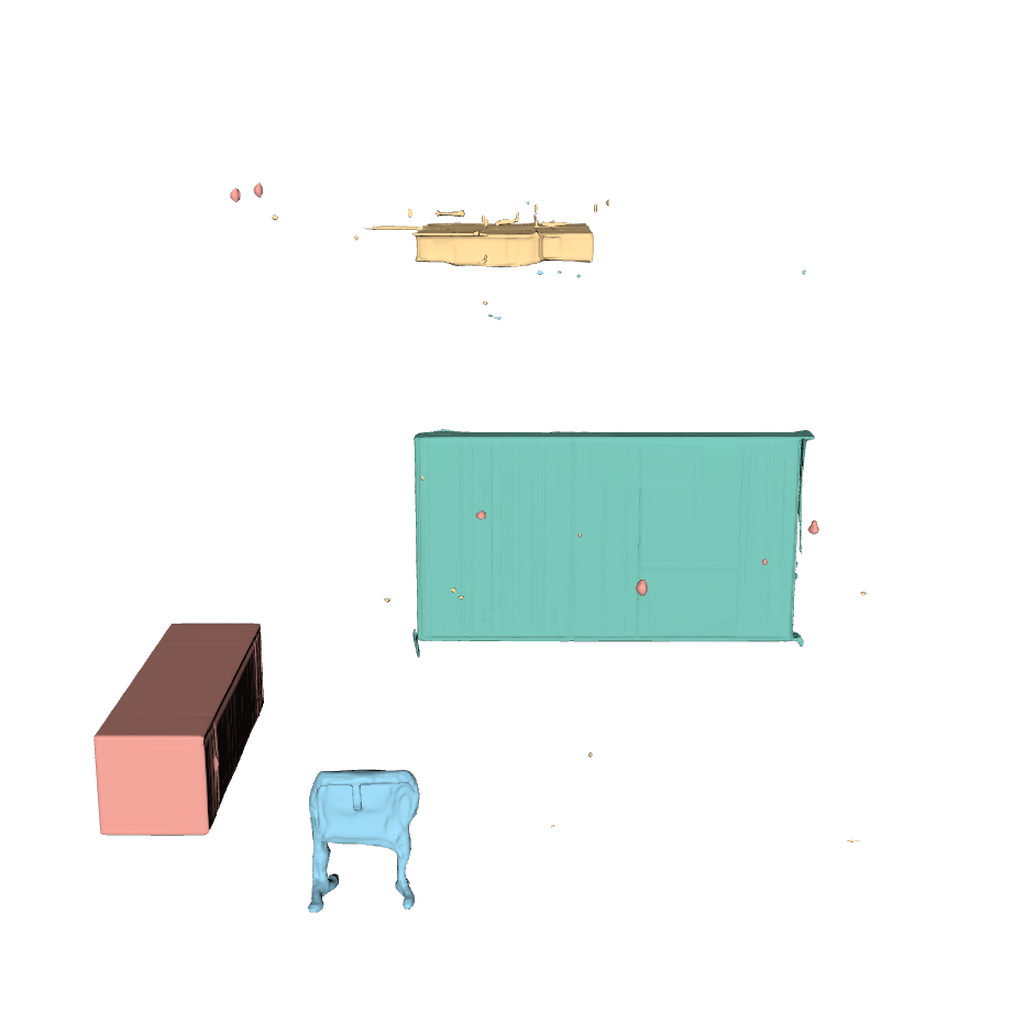}} \\
    
    % Row 3
    \raisebox{-0.5\height}{\includegraphics[width=0.23\linewidth]{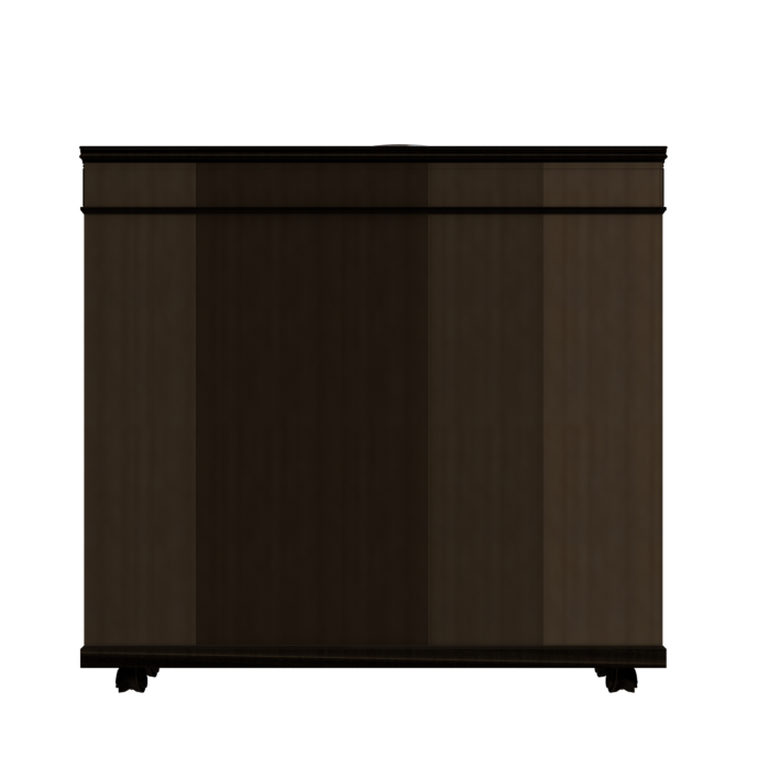}} &
    \raisebox{-0.5\height}{\includegraphics[width=0.23\linewidth]{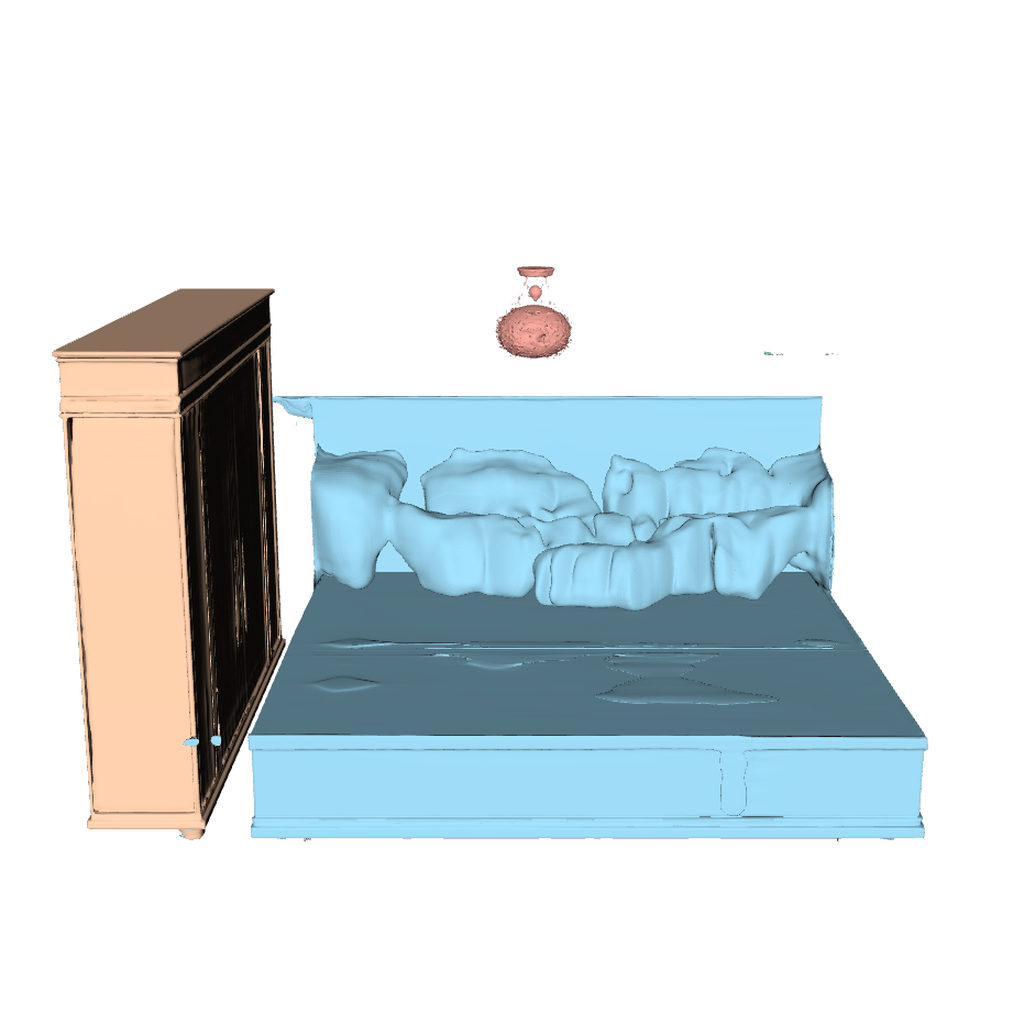}} &
    \raisebox{-0.5\height}{\includegraphics[width=0.23\linewidth]{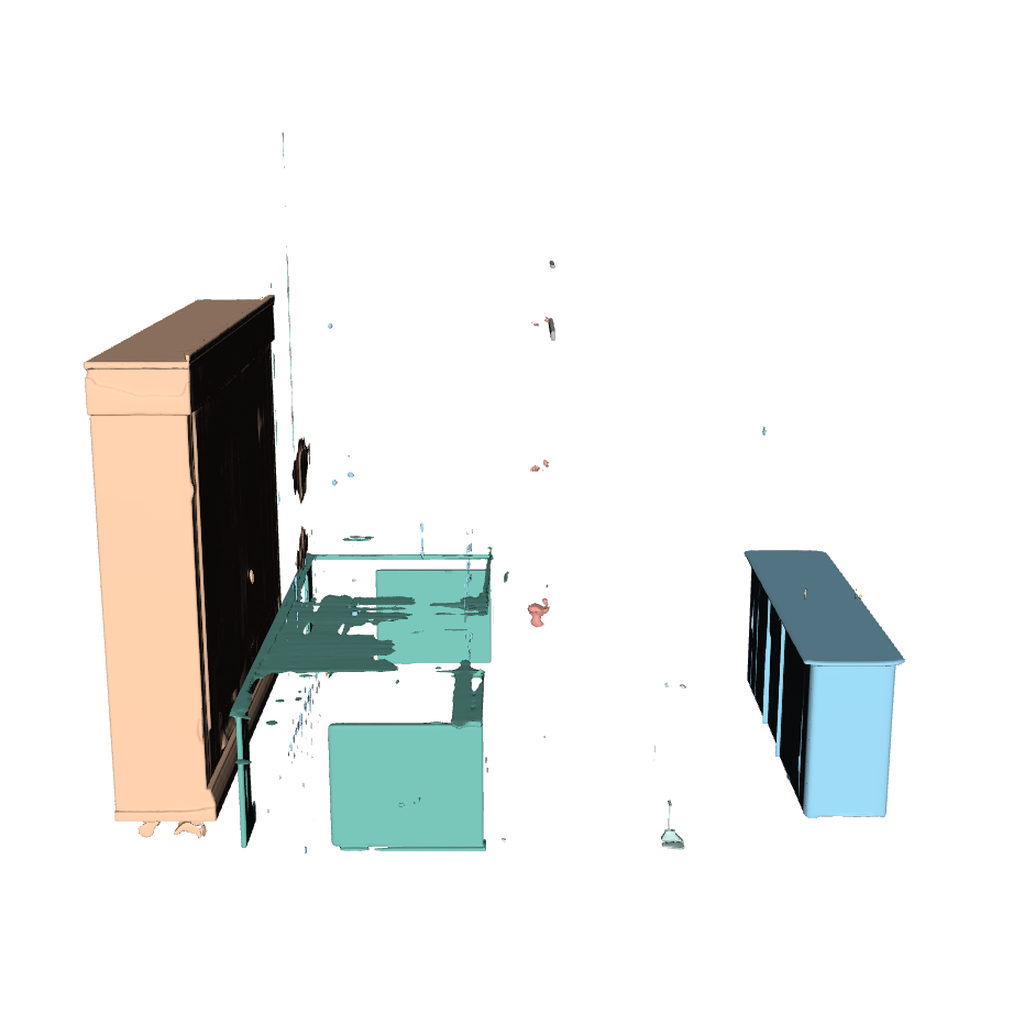}} &
    \raisebox{-0.5\height}{\includegraphics[width=0.23\linewidth]{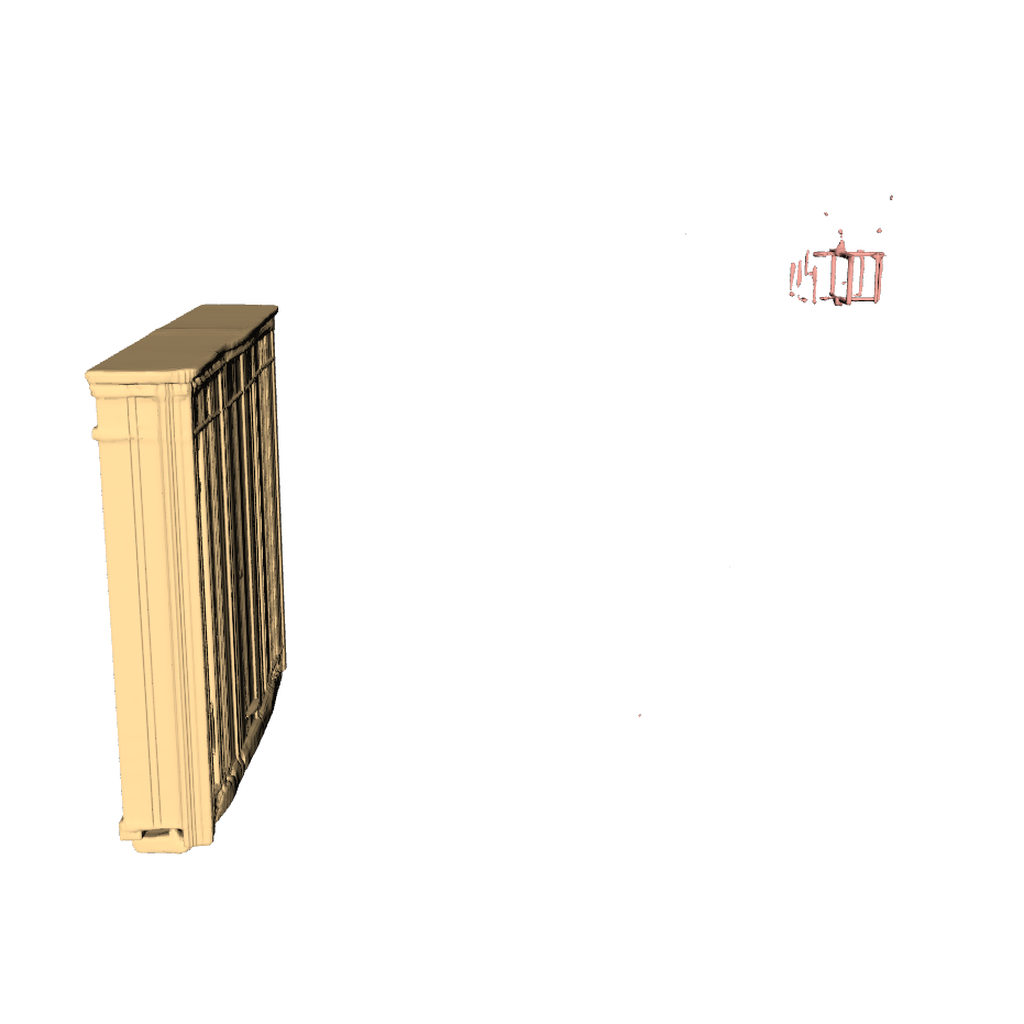}} \\
    
    % Row 4
    \raisebox{-0.5\height}{\includegraphics[width=0.23\linewidth]{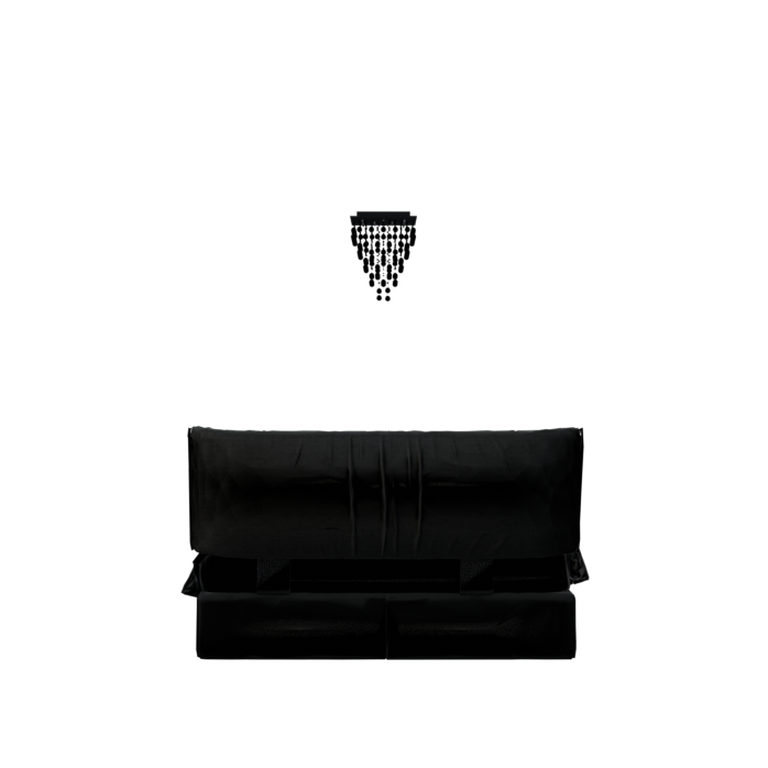}} &
    \raisebox{-0.5\height}{\includegraphics[width=0.23\linewidth]{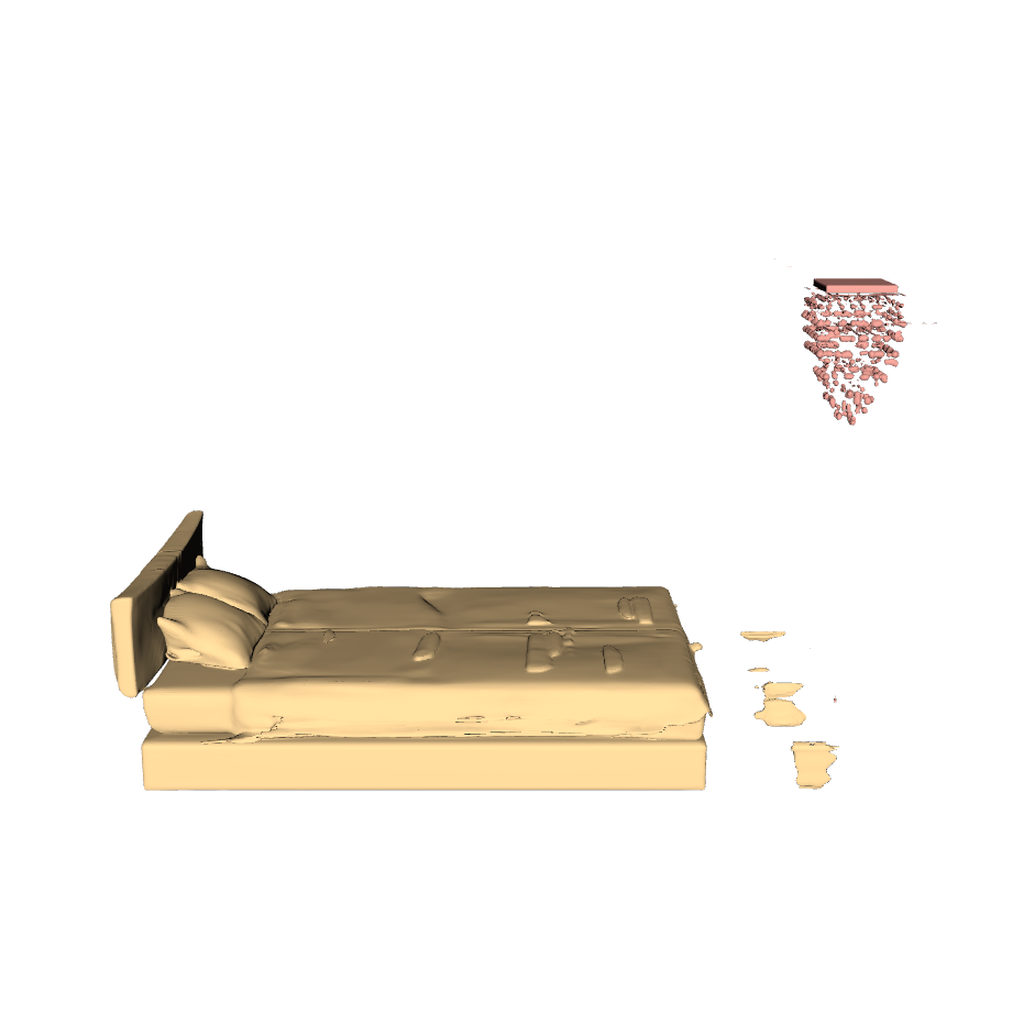}} &
    \raisebox{-0.5\height}{\includegraphics[width=0.23\linewidth]{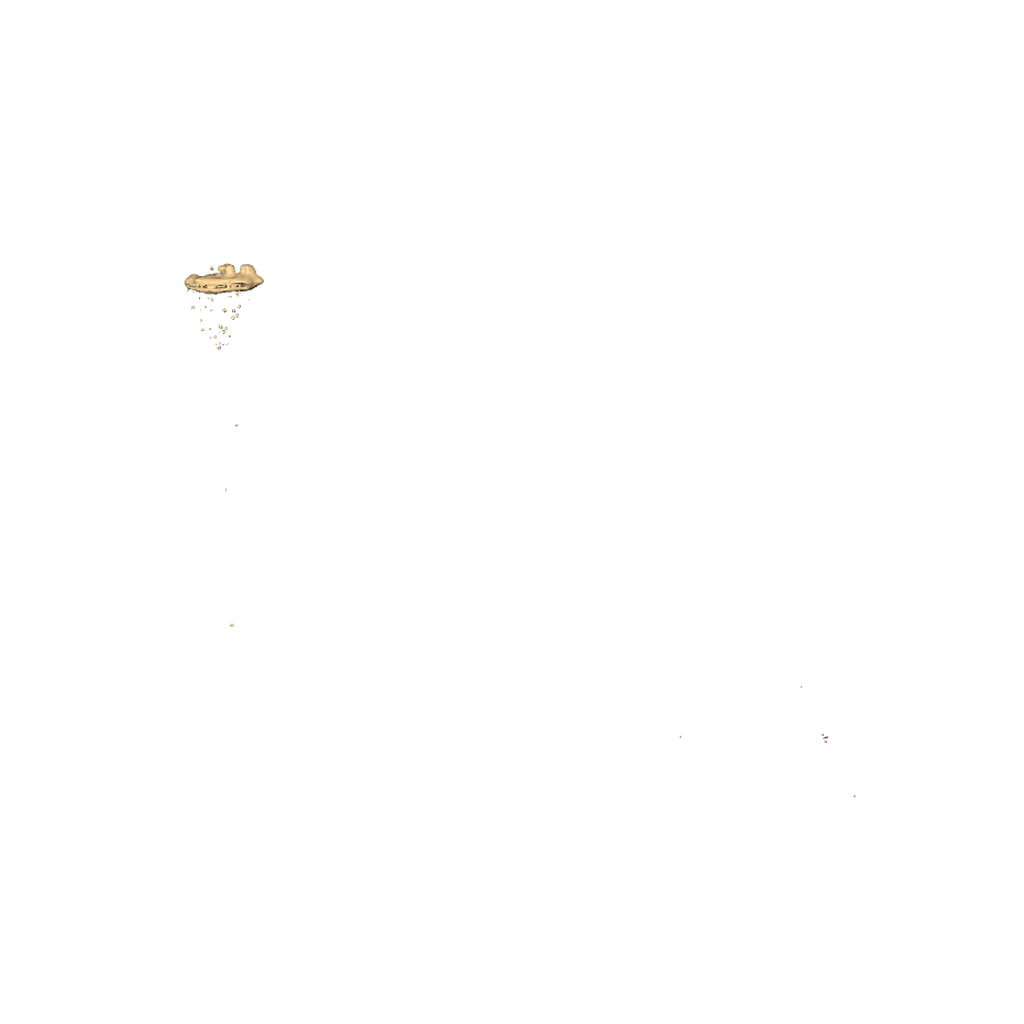}} &
    \raisebox{-0.5\height}{\includegraphics[width=0.23\linewidth]{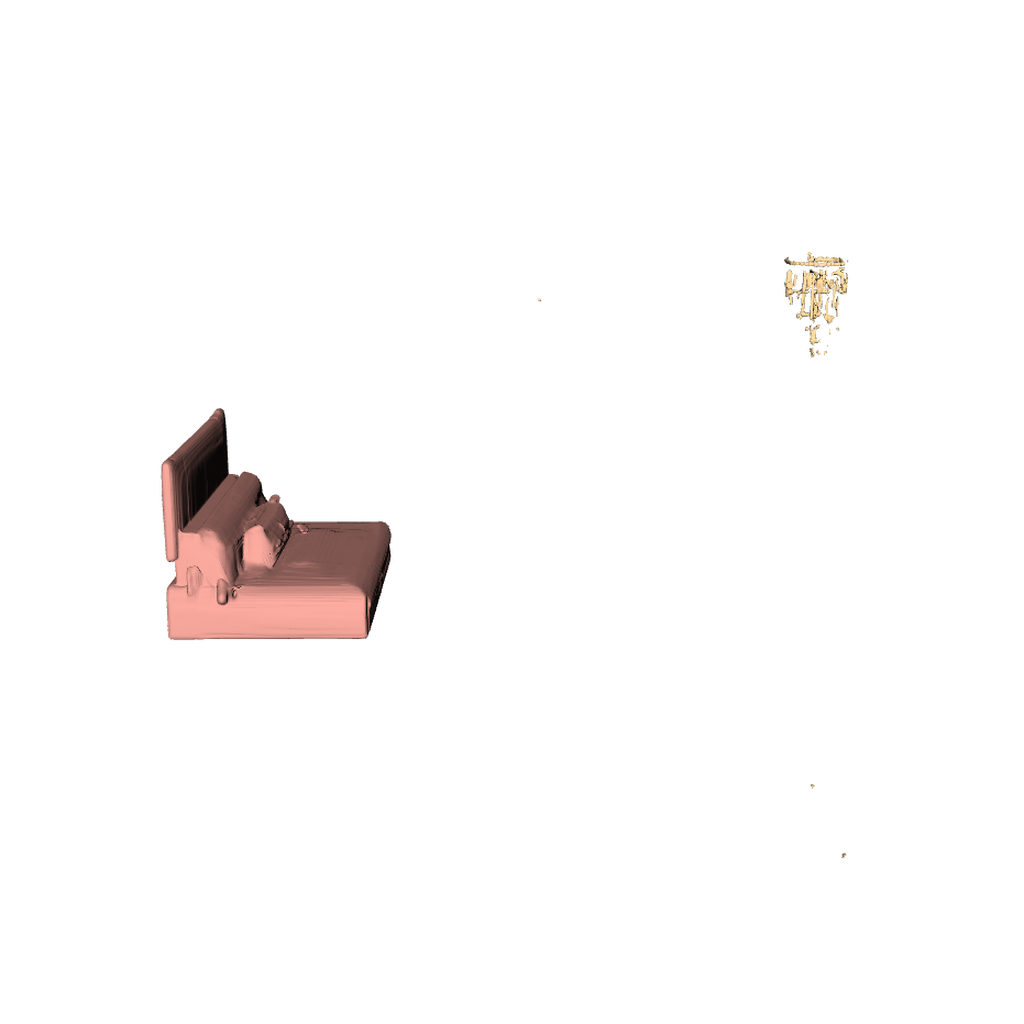}} \\
    
    % Row 5
    \raisebox{-0.5\height}{\includegraphics[width=0.23\linewidth]{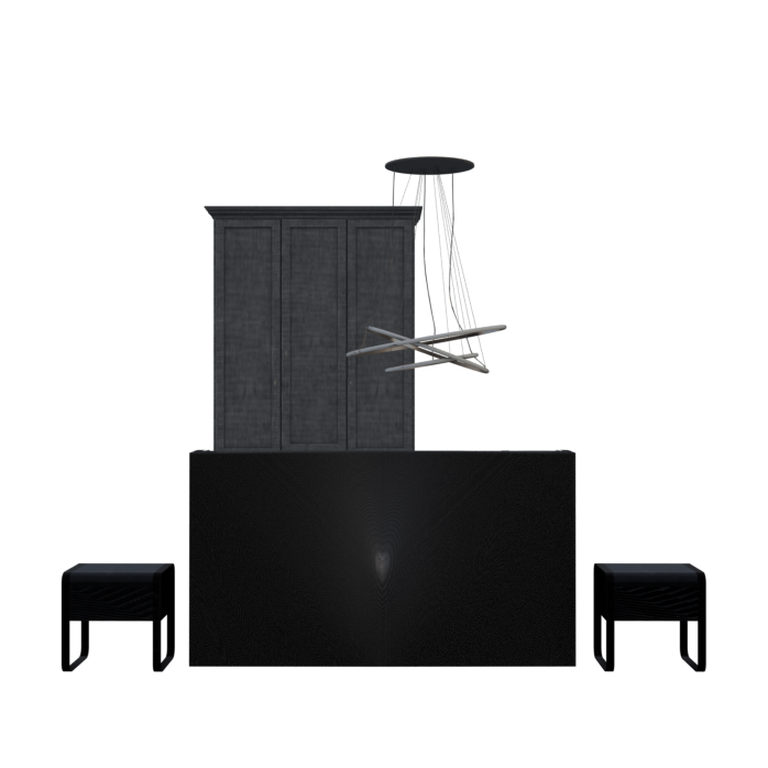}} &
    \raisebox{-0.5\height}{\includegraphics[width=0.23\linewidth]{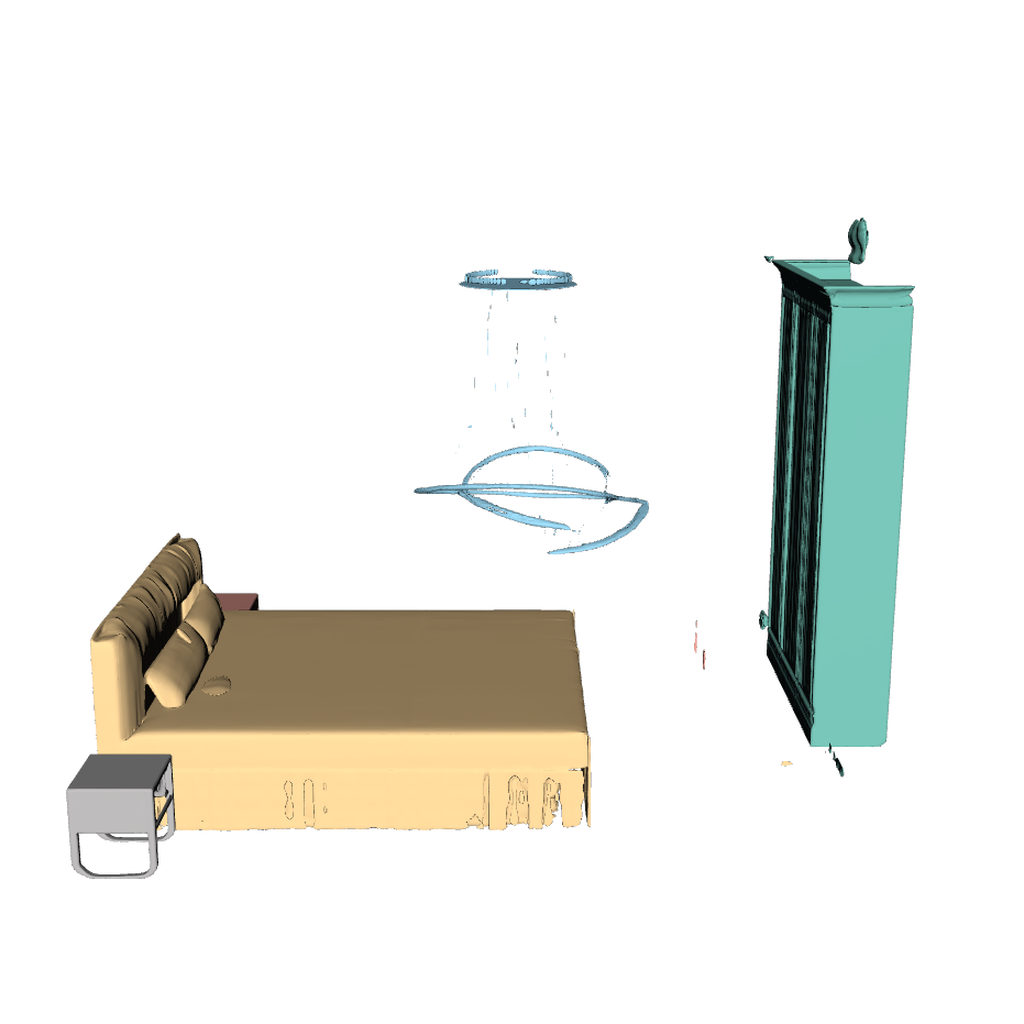}} &
    \raisebox{-0.5\height}{\includegraphics[width=0.23\linewidth]{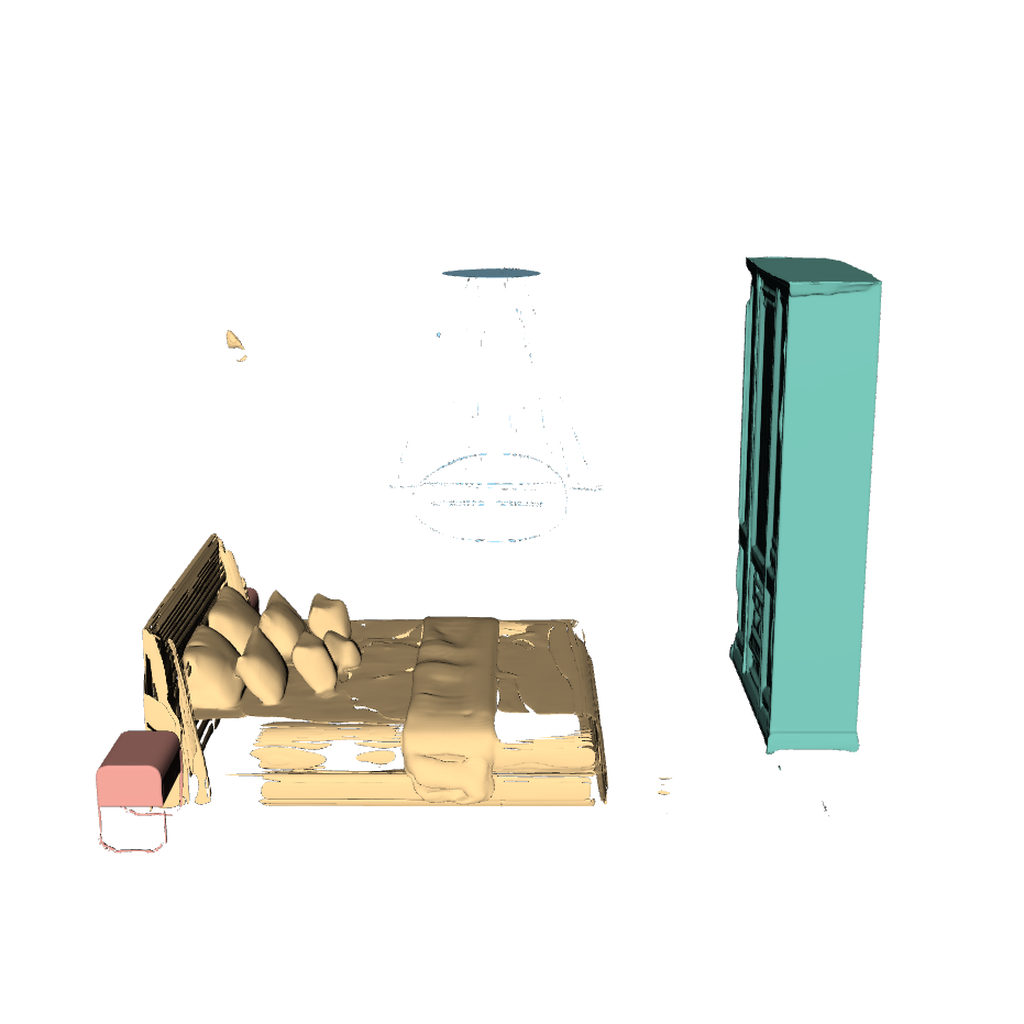}} &
    \raisebox{-0.5\height}{\includegraphics[width=0.23\linewidth]{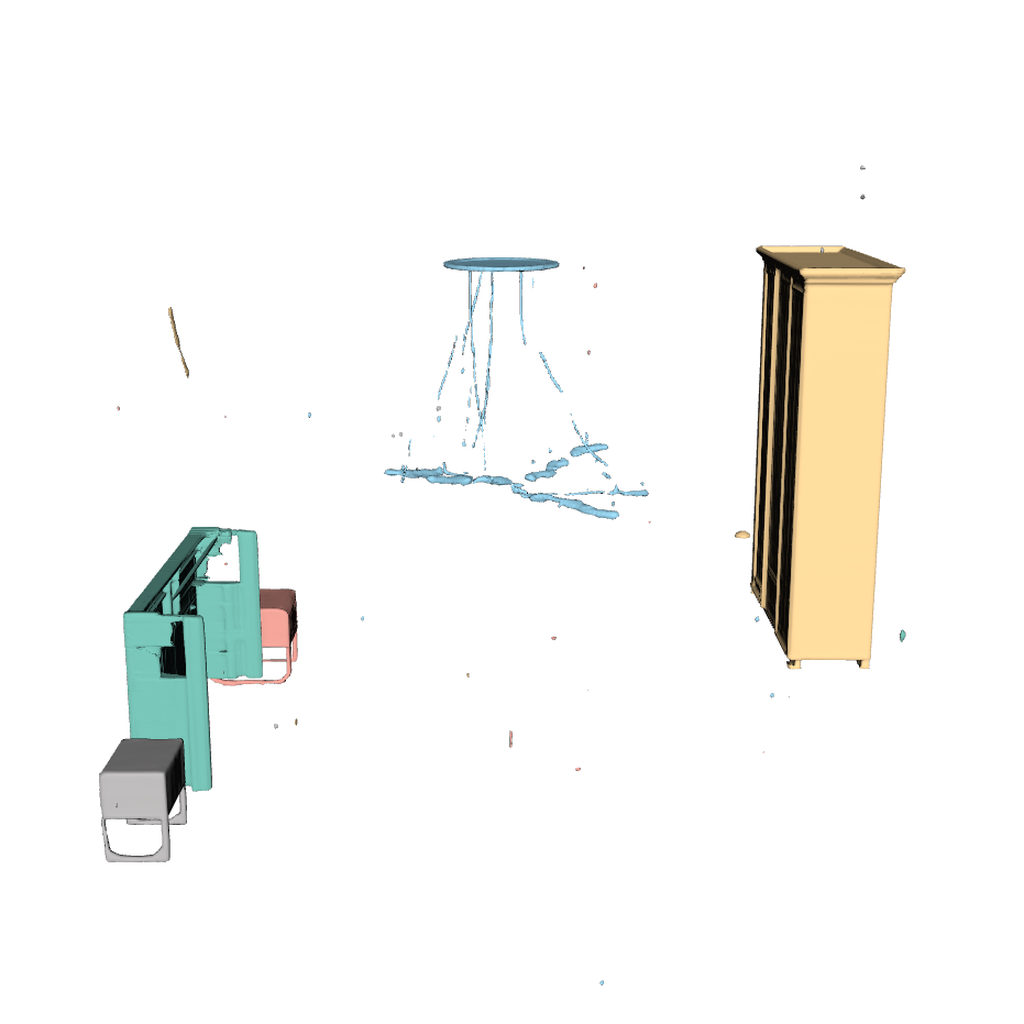}} \\
    
    % Row 6
    \raisebox{-0.5\height}{\includegraphics[width=0.23\linewidth]{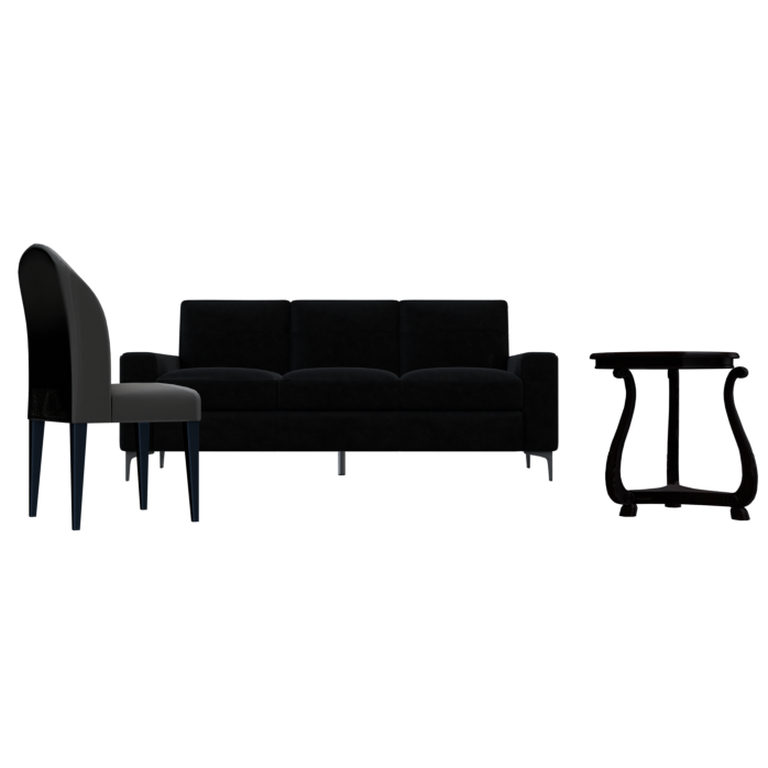}} &
    \raisebox{-0.5\height}{\includegraphics[width=0.23\linewidth]{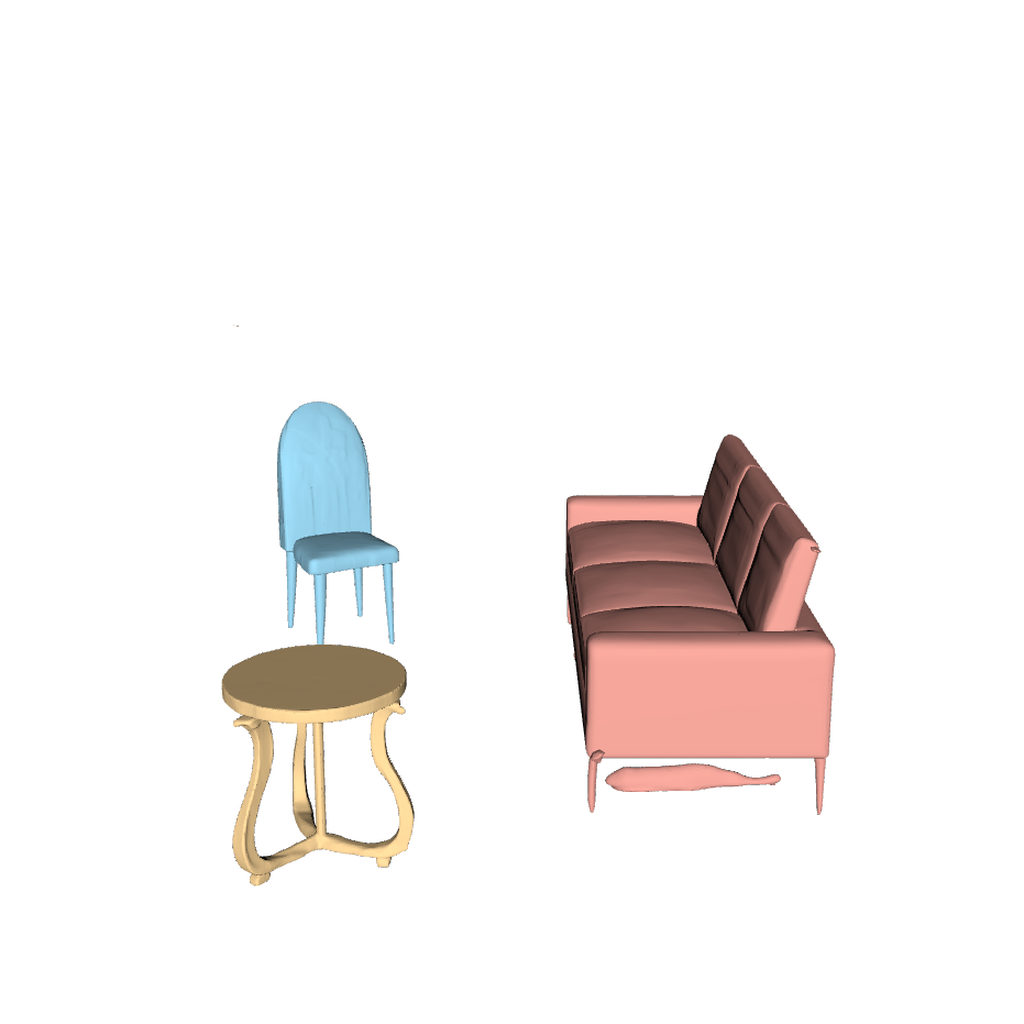}} &
    \raisebox{-0.5\height}{\includegraphics[width=0.23\linewidth]{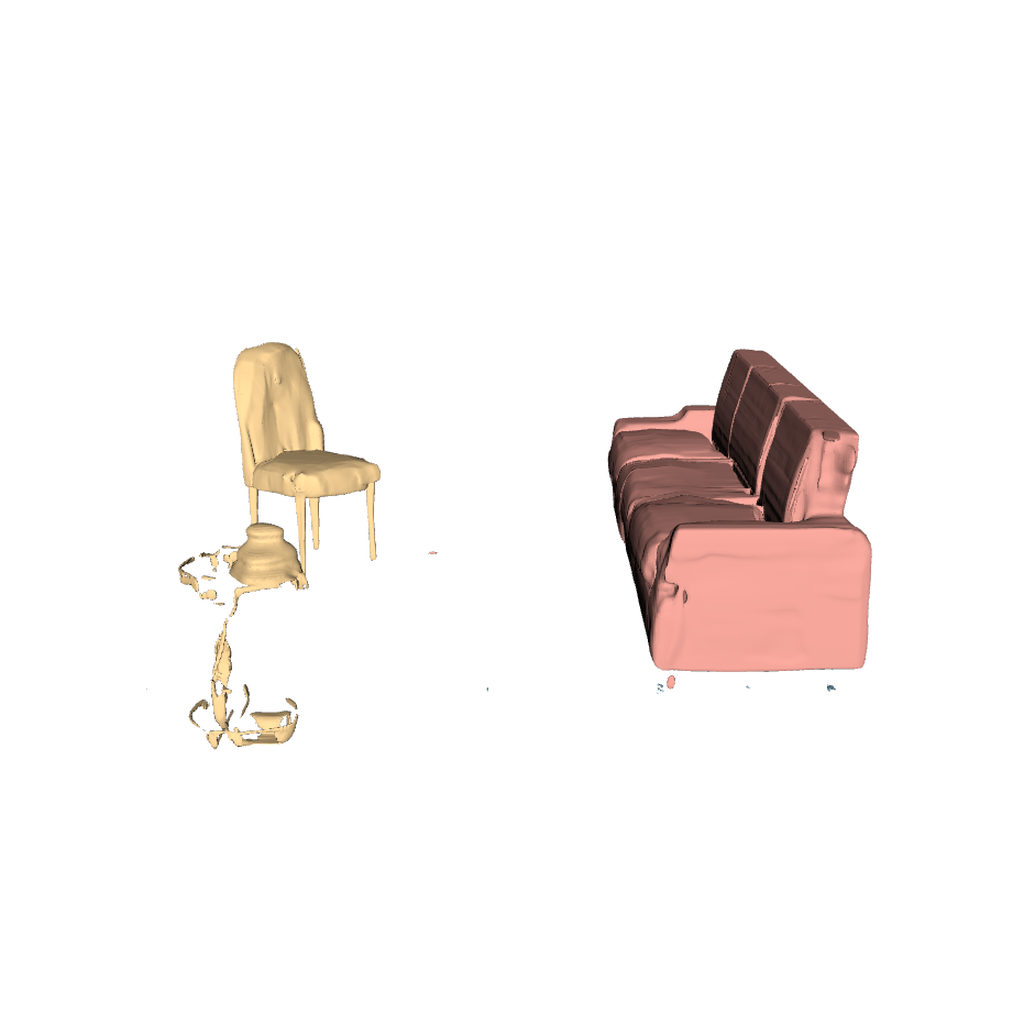}} &
    \raisebox{-0.5\height}{\includegraphics[width=0.23\linewidth]{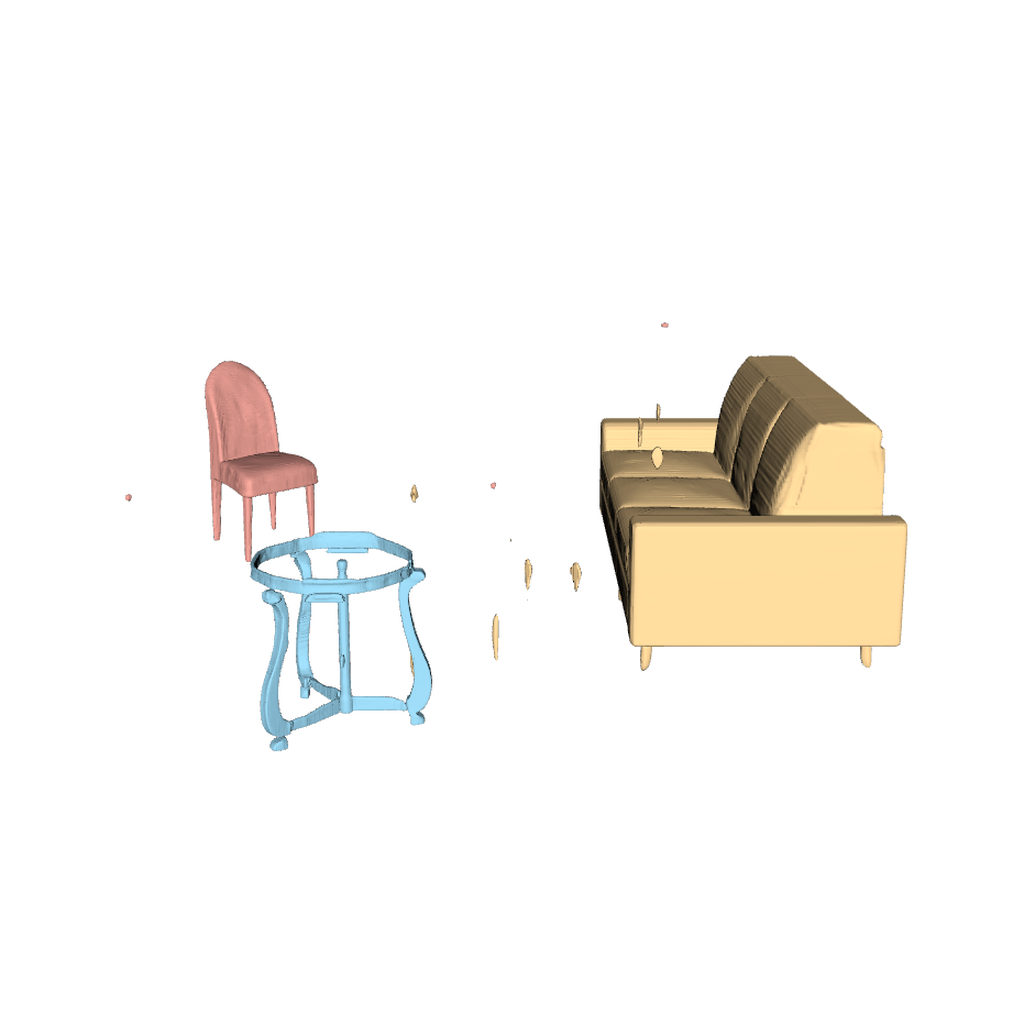}} \\
    
    % Row 7
    \raisebox{-0.5\height}{\includegraphics[width=0.23\linewidth]{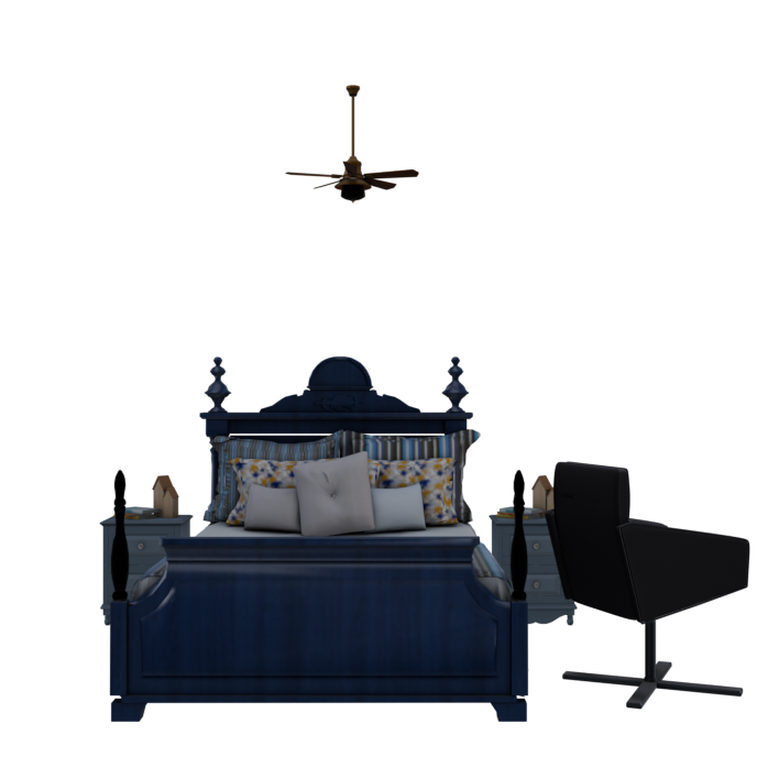}} &
    \raisebox{-0.5\height}{\includegraphics[width=0.23\linewidth]{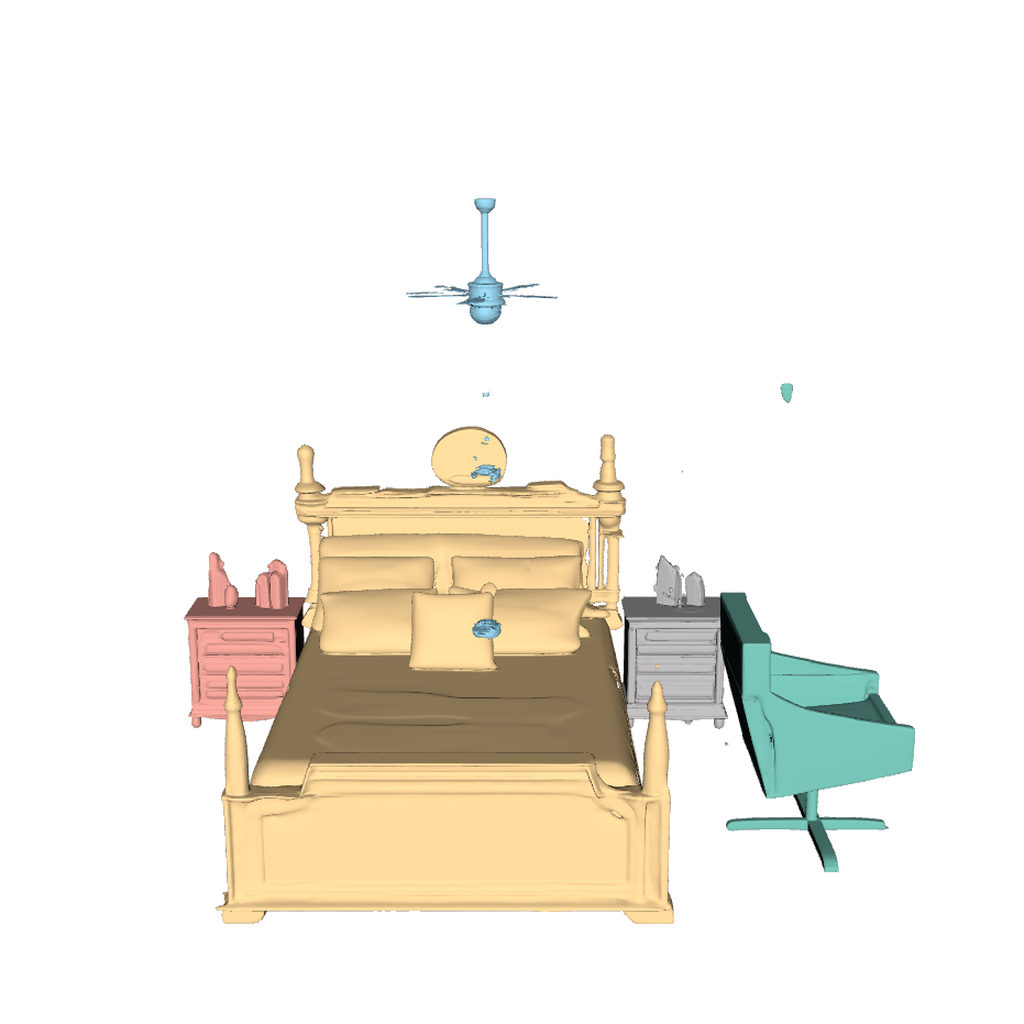}} &
    \raisebox{-0.5\height}{\includegraphics[width=0.23\linewidth]{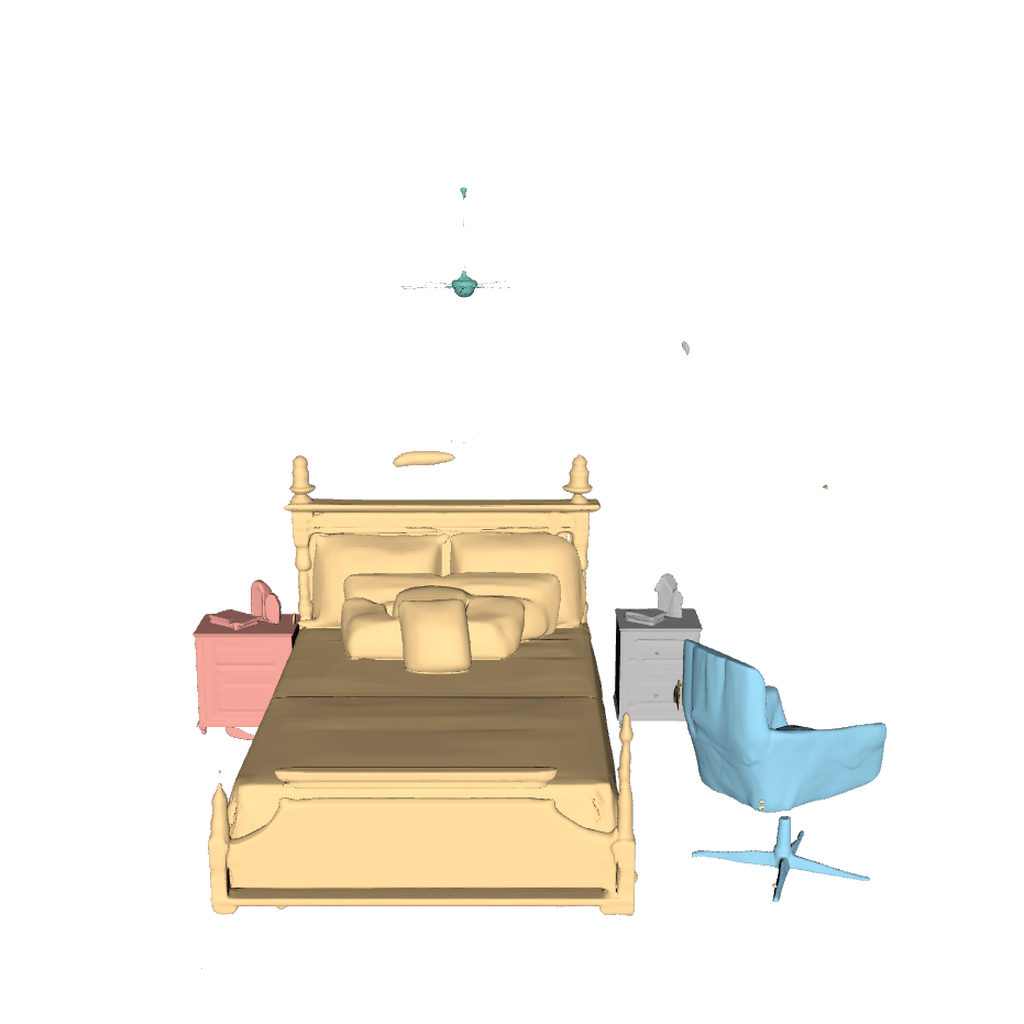}} &
    \raisebox{-0.5\height}{\includegraphics[width=0.23\linewidth]{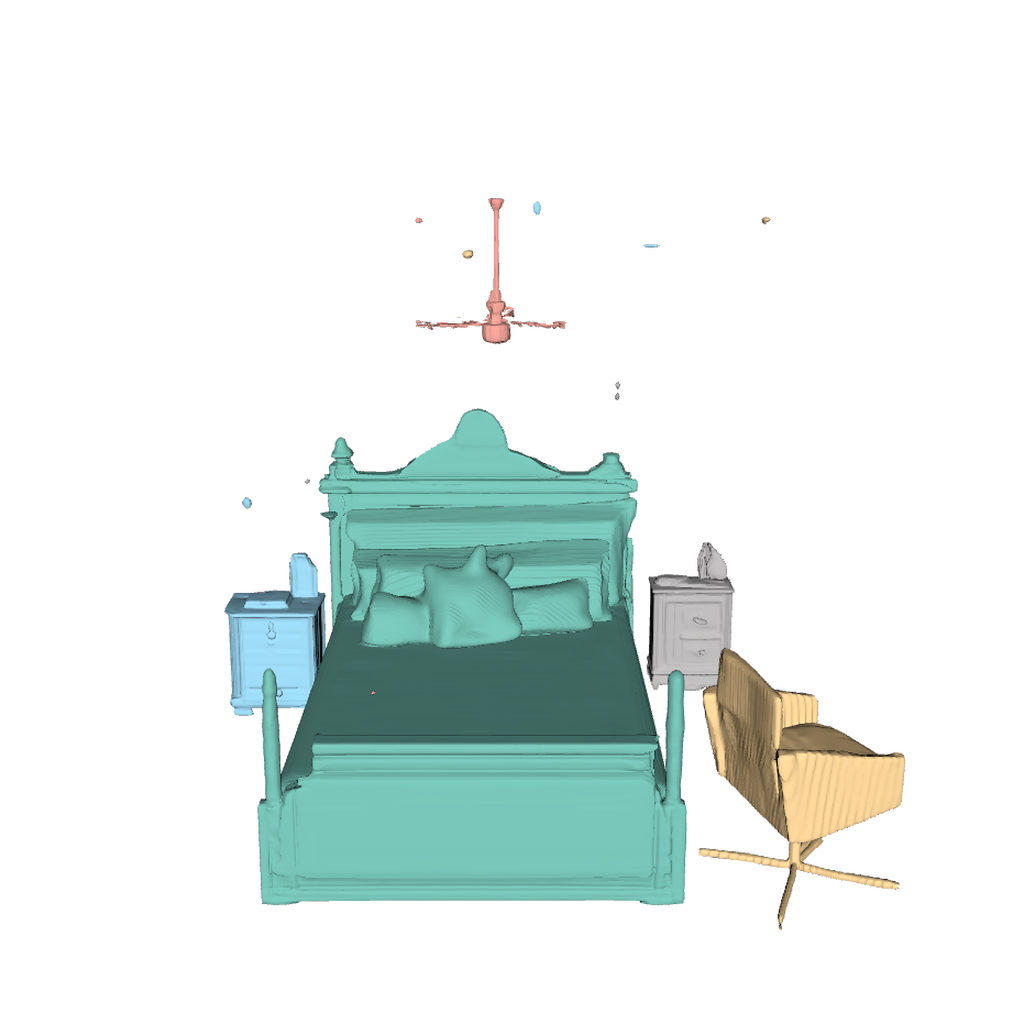}} \\
    
    % Row 8
    \raisebox{-0.5\height}{\includegraphics[width=0.23\linewidth]{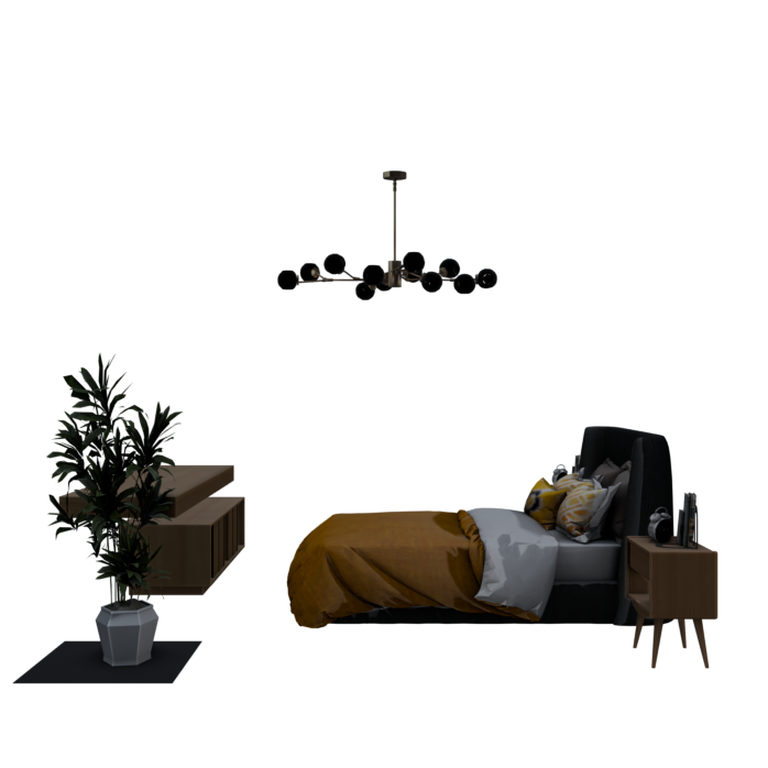}} &
    \raisebox{-0.5\height}{\includegraphics[width=0.23\linewidth]{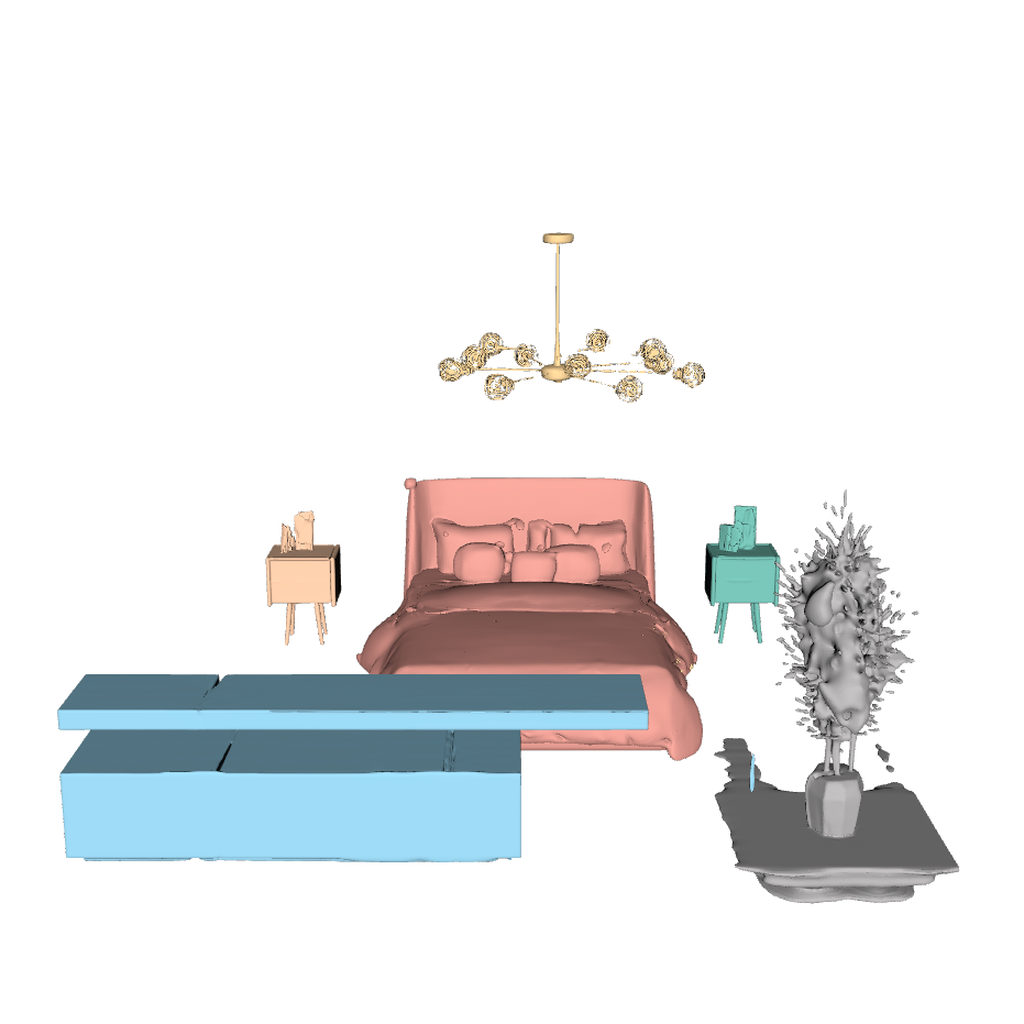}} &
    \raisebox{-0.5\height}{\includegraphics[width=0.23\linewidth]{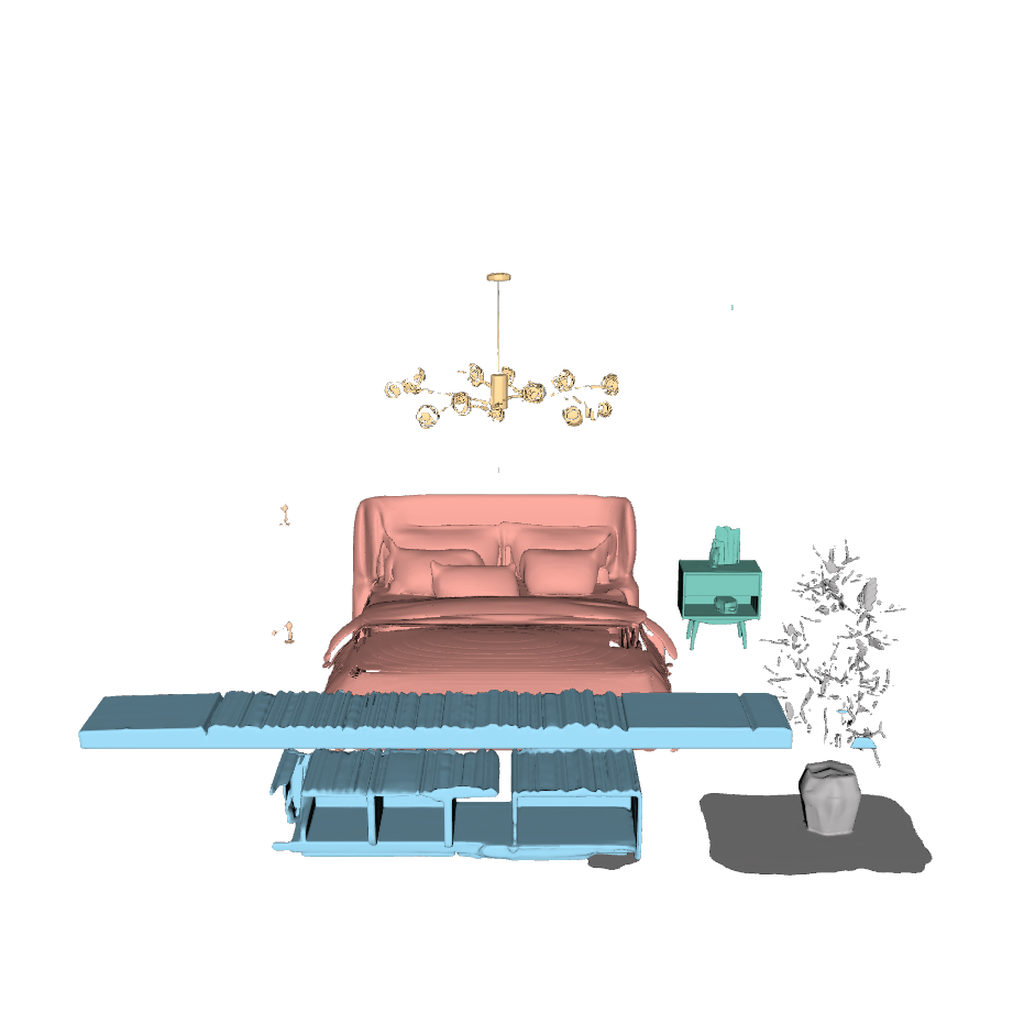}} &
    \raisebox{-0.5\height}{\includegraphics[width=0.23\linewidth]{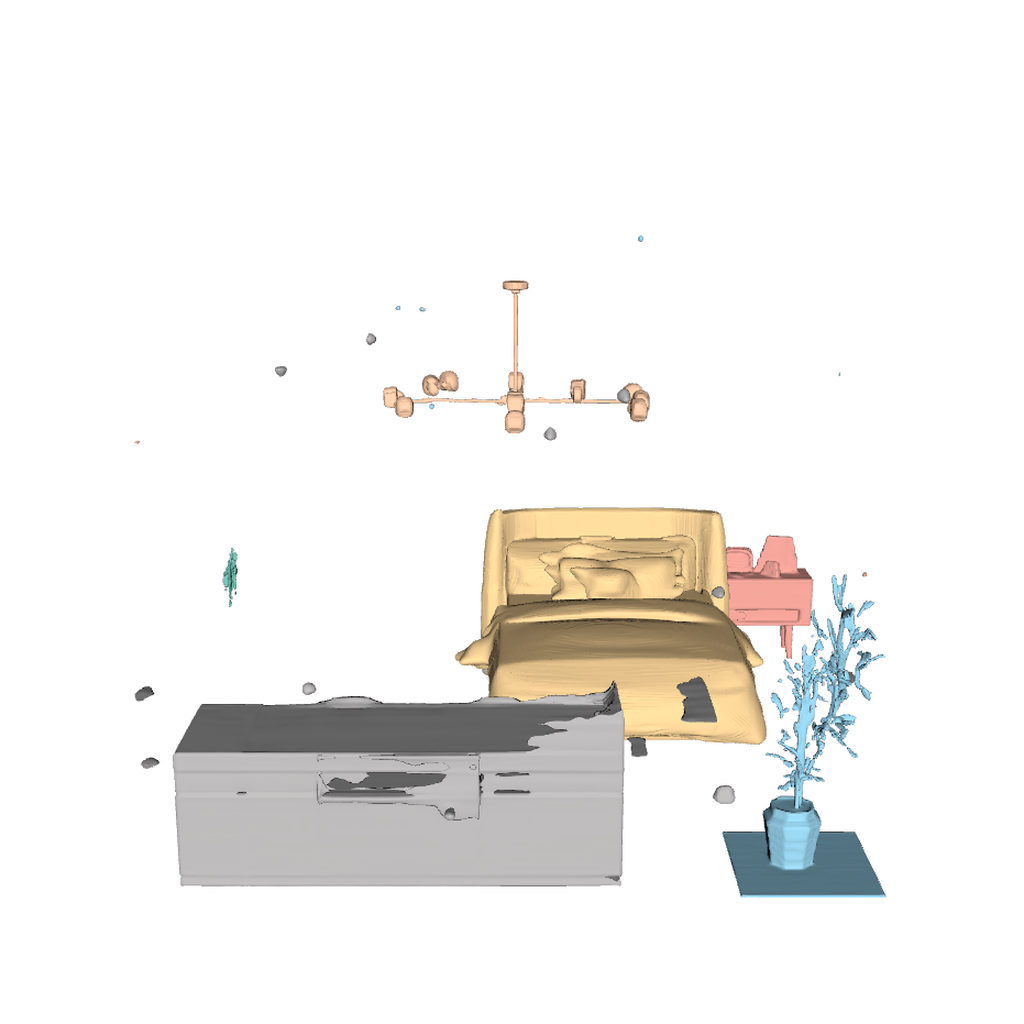}} \\
    
    % Row 9
    \raisebox{-0.5\height}{\includegraphics[width=0.23\linewidth]{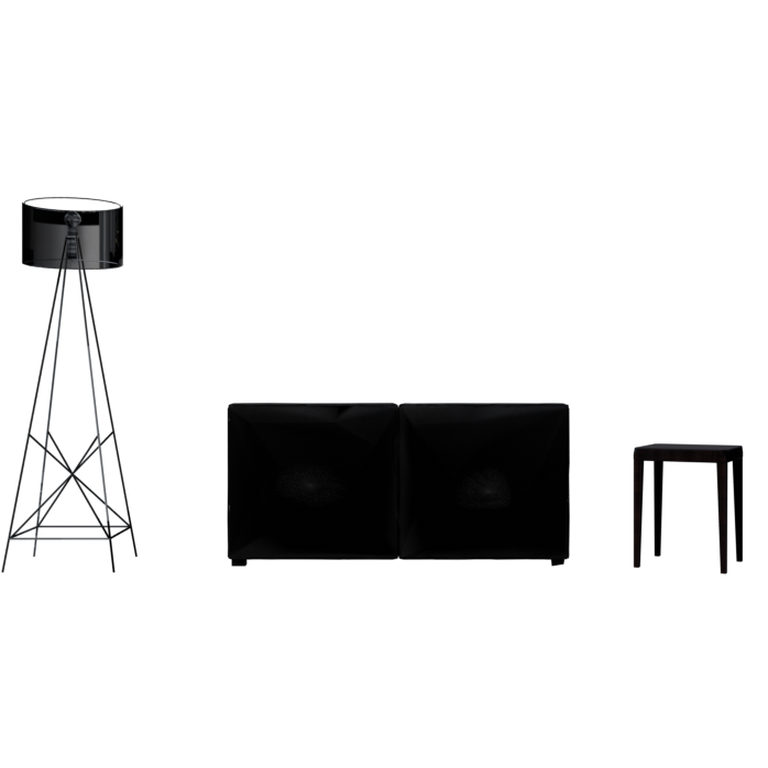}} &
    \raisebox{-0.5\height}{\includegraphics[width=0.23\linewidth]{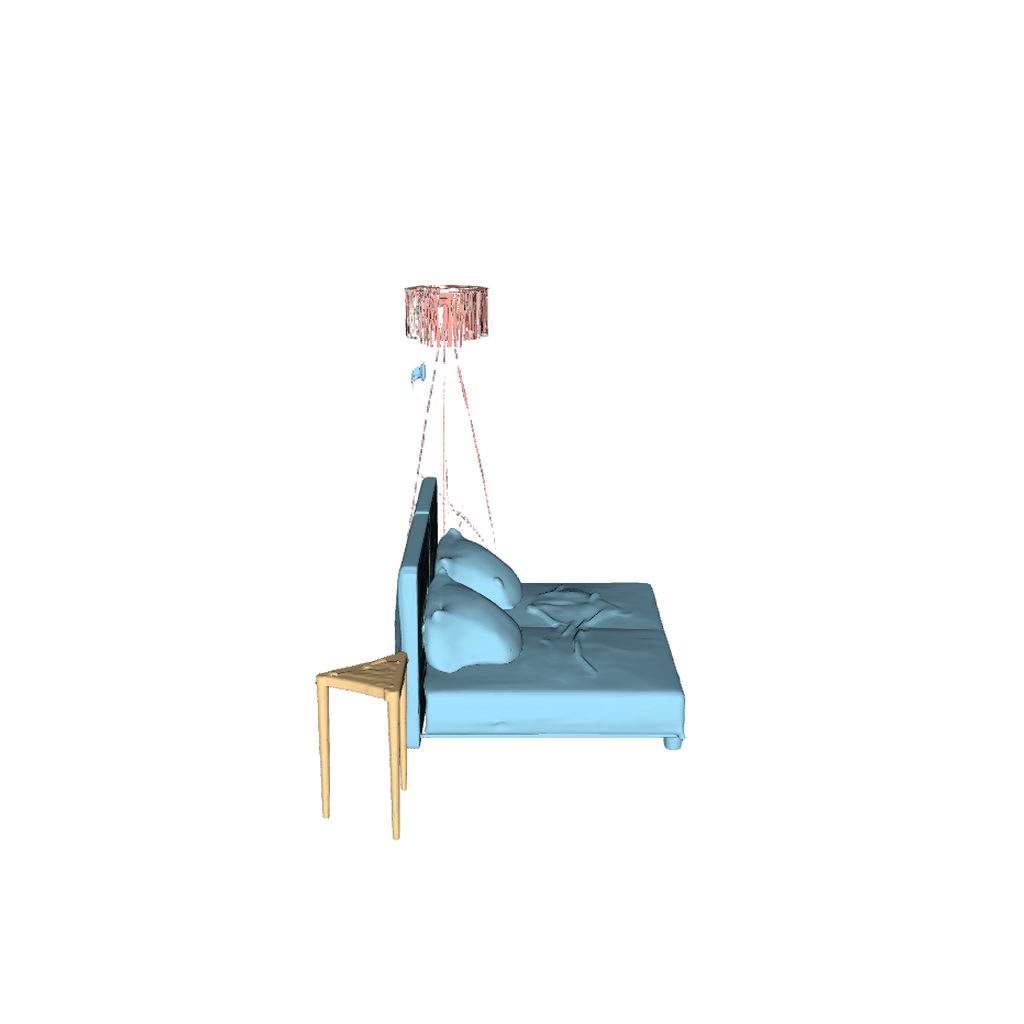}} &
    \raisebox{-0.5\height}{\includegraphics[width=0.23\linewidth]{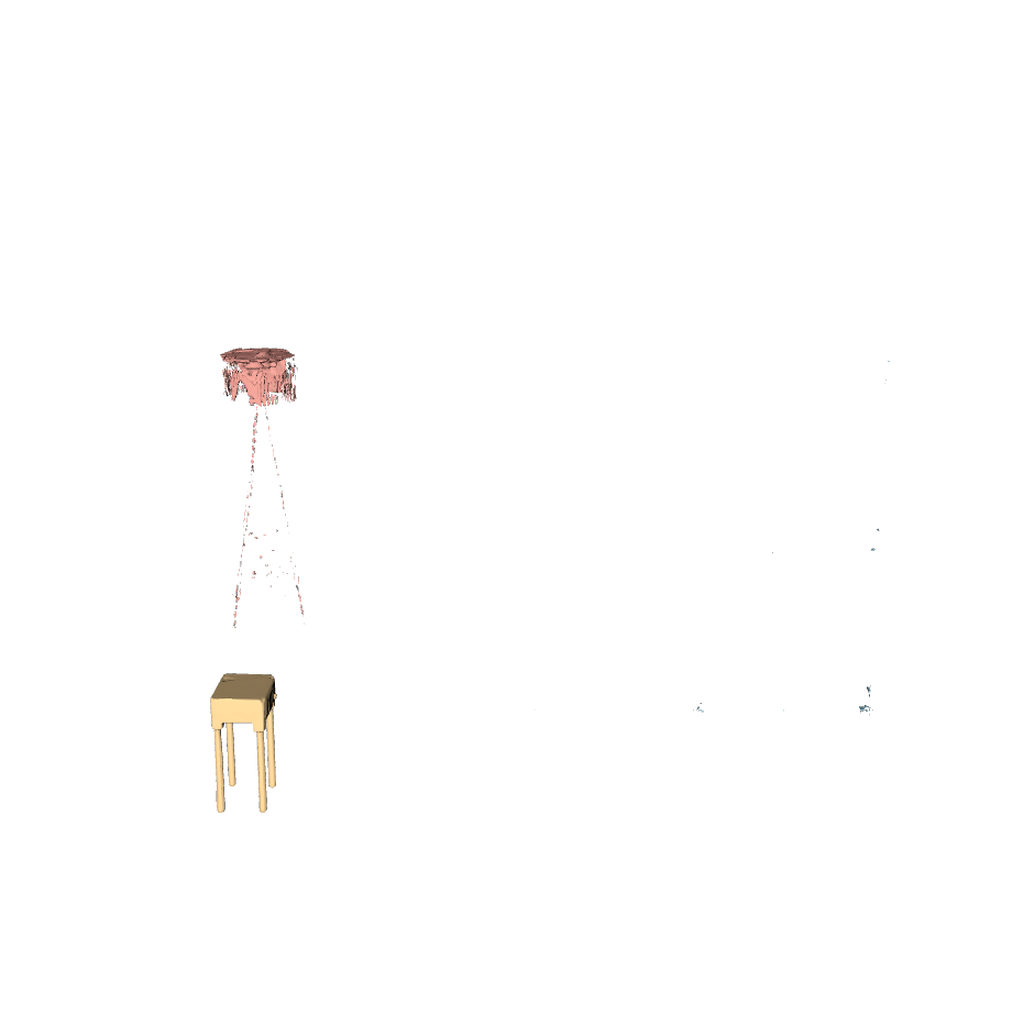}} &
    \raisebox{-0.5\height}{\includegraphics[width=0.23\linewidth]{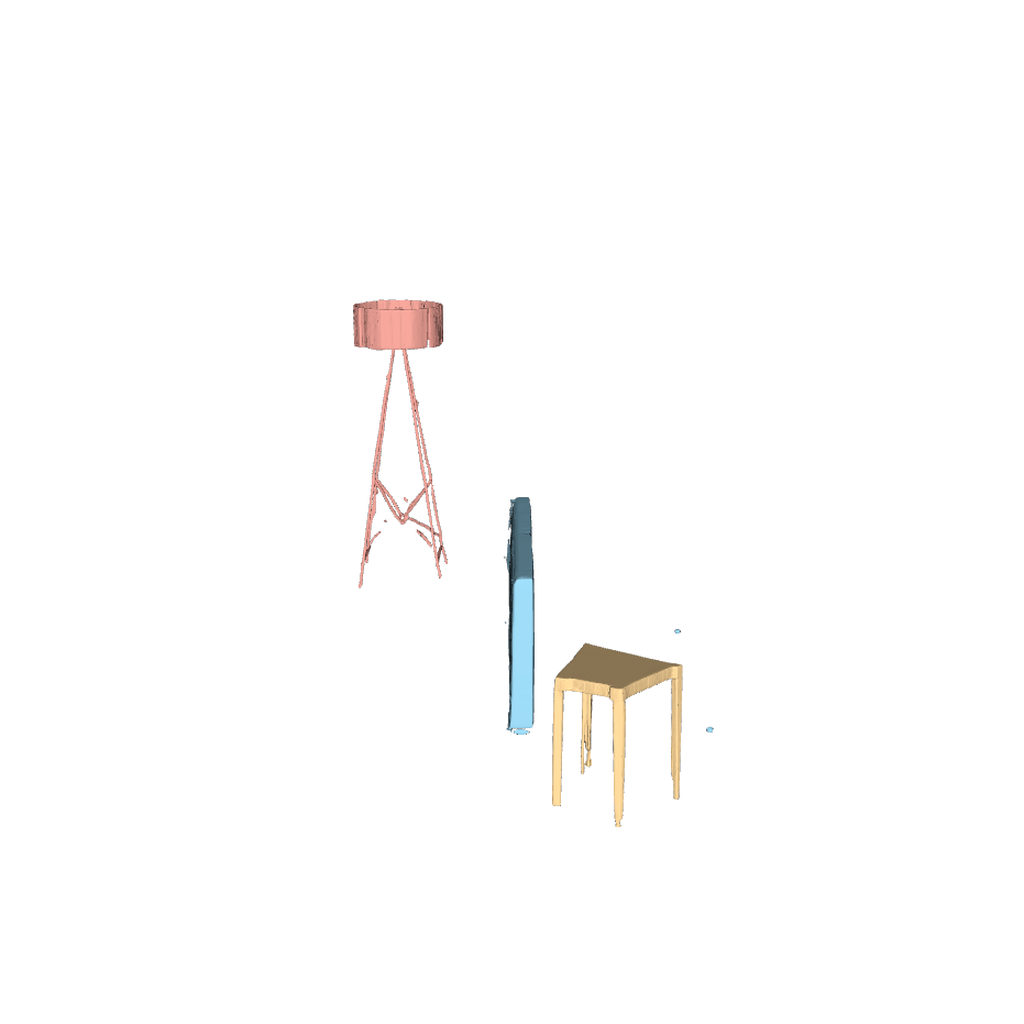}} \\
    
  \end{tabular}
  \caption{
  Qualitative comparison across ours, PartCrafter~\cite{lin2025partcrafterstructured3dmesh} and MIDI~\cite{huang2025midimultiinstancediffusionsingle}. We show qualitatively how our method shows better performance, as shown by the quantitative metrics. Largely, this is due to the consistent reconstruction of all parts and their accurate composition. Both PartCrafter~\cite{lin2025partcrafterstructured3dmesh} and MIDI~\cite{huang2025midimultiinstancediffusionsingle} often miss large objects, \eg, the bed. 
  }
  \label{fig:qual_comp_3d_appendix}
\end{figure}
\end{document}